\documentclass{article}

\usepackage[final]{neurips_2024}

\usepackage[utf8]{inputenc} 
\usepackage[T1]{fontenc}    
\usepackage{hyperref}       
\usepackage{url}            
\usepackage{booktabs}       
\usepackage{amsfonts}       
\usepackage{nicefrac}       
\usepackage{microtype}      
\usepackage{xcolor}         

\usepackage{amsmath}
\usepackage{bm}
\usepackage{bbm}
\usepackage{graphicx}
\usepackage{caption}
\usepackage{subcaption}
\usepackage{placeins}
\usepackage{amsthm}
\newtheorem{condition}{Condition}[section]
\newtheorem{theorem}{Theorem}[section]
\newtheorem{lemma}{Lemma}[section]
\newtheorem{definition}{Definition}[section]
\newtheorem{claim}{Claim}

\usepackage{etoc}
\etocdepthtag.toc{mtchapter}
\etocsettagdepth{mtchapter}{subsection}
\etocsettagdepth{mtappendix}{none}

\title{Unveil Benign Overfitting for Transformer in Vision: Training Dynamics, Convergence, and Generalization}

\author{
  Jiarui Jiang\thanks{Equal contribution}\hspace{0.5em}\textsuperscript{1}, Wei Huang\footnotemark[1]\hspace{0.5em}\textsuperscript{2}, Miao Zhang\thanks{Corresponding author}\hspace{0.5em}\textsuperscript{1}, Taiji Suzuki\textsuperscript{3 2}, Liqiang Nie\textsuperscript{1} \\
  \textsuperscript{1}Harbin Institute of Technology, Shenzhen
  \\
  \textsuperscript{2}RIKEN AIP
  \\
  \textsuperscript{3}University of Tokyo
  \\
  \texttt{jiaruij@outlook.com},\hspace{0.5em}\texttt{wei.huang.vr@riken.jp},\hspace{0.5em}\texttt{zhangmiao@hit.edu.cn}
  \\
  \texttt{taiji@mist.i.u-tokyo.ac.jp},\hspace{0.5em}\texttt{nieliqiang@hit.edu.cn}
}

\begin{document}

\allowdisplaybreaks[4]

\maketitle
\setcounter{footnote}{0}

\begin{abstract}
  Transformers have demonstrated great power in the recent development of large foundational models. In particular, the Vision Transformer (ViT) has brought revolutionary changes to the field of vision, achieving significant accomplishments on the experimental side. However, their theoretical capabilities, particularly in terms of generalization when trained to overfit training data, are still not fully understood. To address this gap, this work delves deeply into the \textit{benign overfitting} perspective of transformers in vision. To this end, we study the optimization of a Transformer composed of a self-attention layer with softmax followed by a fully connected layer under gradient descent on a certain data distribution model. By developing techniques that address the challenges posed by softmax and the interdependent nature of multiple weights in transformer optimization, we successfully characterized the training dynamics and achieved generalization in post-training. Our results establish a sharp condition that can distinguish between the small test error phase and the large test error regime, based on the signal-to-noise ratio in the data model. The theoretical results are further verified by experimental simulation. To the best of our knowledge, this is the first work to characterize benign overfitting for Transformers.

\end{abstract}

\section{Introduction}

Transformers \citep{vaswani2017attention} have revolutionized numerous fields in artificial intelligence, ranging from natural language processing \citep{devlin2018bert,openai2023gpt4}, computer vision \cite{dosovitskiy2020image,he2022masked}, graph \cite{yun2019graph,hu2020heterogeneous} to reinforcement learning \cite{chen2021decision,janner2021offline}. In particular, Vision Transformer (ViT) \cite{dosovitskiy2020image} has been developed to advancing the computer vision tasks \citep{liu2021swin,khan2022transformers,guo2022attention,tolstikhin2021mlp,lin2014microsoft} compared to Convolutional neural networks. Since then, with the high performance, ViT emerged as a hot area of research and application. Numerous technologies and methods began to surface, including enhancing ViT’s data and resource efficiency \citep{touvron2021training,hassani2021escaping,chen2021chasing,touvron2022resmlp}, reducing heavy computation cost \cite{chen2022auto,zhu2024vision,wang2023image}.

Recent success brought by ViT has been inspiring an increasing number of works to understand the ViT through empirical and theoretical studies. The empirical investigation works study the robustness \citep{bhojanapalli2021understanding,naseer2021intriguing,paul2022vision,bai2021transformers}; and the role in self-supervision \citep{caron2021emerging,chen2021empirical,oquab2023dinov2} and others. On the other hand side, the theoretical works have been conducted to understand the ViT from the perspective of expressivity \citep{vuckovic2020mathematical,edelman2022inductive}, optimization \citep{zhang2020adaptive,tian2023scan,tian2023joma}, generalization \citep{li2023theoretical}, and in-context learning \citep{huang2023context,garg2022can,bai2024transformers,ahn2024transformers}.

Despite the insightful understanding provided by the above investigations, it remains unclear how ViTs generalize to unseen data when they are trained to overfit the training data. In particular, the conditions under which ViTs can exhibit \textit{benign overfitting} \citep{bartlett2020benign}, where the test error remains small despite overfitting to the training data, are not well understood. Moreover, prior work on the optimization and generalization of ViTs often focuses on simplified settings, such as linear transformers or unrealistic loss functions, due to technical limitations. In this work, we aim to fill this gap through a feature learning theory \citep{allen2020towards,cao2022benign}. Based on a data generative model with $M$ tokens composed of signal tokens and noise tokens and a two-layer ViT with softmax attention, we characterize the training dynamics from a random initialization and the generalization ability of the ViT after convergence. We establish a sharp separation in condition to distinct the benign overfitting and harmful overfitting regime for ViTs.  Our contributions are summarized as follows:
\begin{itemize}
    \item We successfully characterize the optimization of the ViT through three different phases in dynamics that exhibit rich and unique behaviors related to self-attention. Building on the convergence results, we further characterize the generalization bounds on unseen datasets.

    \item Technically, we develop novel methods to handle non-linear attention and the inter-dependent nature of multiple weights in transformer optimization when training from scratch.

    \item We establish a sharp separation condition in the signal-to-noise ratio in data to distinguish the benign overfitting and harmful overfitting regimes of ViTs. This separation is then verified by experimental simulation.

\end{itemize}

\section{Related Work}

\paragraph{Benign Overfiting in Deep Learning}

The most prominent behavior of deep learning is its breaking of bias-variance trade-off in statistical learning theory, i.e., good generalization on unseen data even with overfitting on training data.
This line of research starts from \cite{bartlett2020benign} who studied benign overfitting in linear regression, learning data generated by a linear model with additive Gaussian noises. It is shown that at large input dimensions (which leads to over-parameterization), the excessive risk of the interpolation can be asymptotically optimal. \cite{hastie2022surprises,wu2020optimal} studied linear regression when both the dimension and the number of samples scale together with a fixed ratio. \cite{timor2023implicit} proposed a unified data model for linear predictors and studied conditions under which benign overfitting occurs in different problems. Besides the studies on linear models \citep{wang2021benign,zou2021benign,zhou2023implicit}, several recent works tried to study benign overfitting in neural networks \citep{frei2022benign,kornowski2024tempered,chatterji2022interplay,xu2023benign}. In particular, \citet{li2021towards}~investigated the benign overfitting in two-layer neural networks with the first layer parameters fixed at random initialization. Furthermore, \textit{benign overfitting} was characterized in convolution neural networks \citep{cao2022benign,kou2023benign}, OOD \citep{chen2024understanding}, federated learning \citep{huang2023understanding}, graph neural network \citep{huang2023graph} by feature learning theory \citep{allen2020towards,cao2022benign}. Unlike existing research on benign overfitting, this work focuses on ViTs.

\paragraph{Optimization of Transformers} Towards understanding the optimization of Transformers, \cite{zhang2020adaptive} provided the analysis for adaptive gradient methods.
\cite{jelassi2022vision} proposed a spatially structured dataset and a simplified ViT model and showed that ViT implicitly learns the
spatial structure of the dataset while generalizing. Besides, \cite{li2023transformers} provided fine-grained mechanistic understanding of how transformers learn ``semantic structure'' for BERT-like framework \citep{devlin2018bert}. Furthermore, \cite{tian2023scan,tian2023joma}  characterizes the SGD training
dynamics of a 1-layer Transformer and multi-layer Transformer respectively. They focused on the unique phenomena and the role of attention in the training dynamics. \cite{huang2023context} studied the learning dynamics of a one-layer
transformer with softmax attention trained via gradient descent in order to in-context learn linear function
classes. In addition, \cite{li2024training} provided a theoretical analysis of the training dynamics of Transformers with nonlinear self-attention within In-Context Learning. In particular, \cite{li2023theoretical} is most relevant to us, as they also study the training dynamics of ViTs with a similar data model. However, they considered hinge loss and a specific initialization to simplify analysis and did not characterize the harmful overfitting regime. In summary, while the above work studies the optimization dynamics of attention-based models, they do not characterize benign overfitting as we investigate.

\section{Problem Setup}
\label{problem_setup}
In this section, we outline the data generation model, the Vision Transformer model, and the gradient descent training algorithm.

\textbf{Notations.} For two sequences $ \{ x_n \} $ and $ \{ y_n \} $, we denote $ x_n = O(y_n) $ if there exist some absolute constant $ C > 0 $ and $ N > 0 $ such that $ |x_n| \le C |y_n| $ for all $n \ge N$, 
denote $ x_n = \Omega(y_n) $ if $ y_n = O(x_n) $, and denote $ x_n = \Theta(y_n) $ if $ x_n = O(y_n) $ and $ x_n = \Omega(y_n) $. We also denote $ x_n = o(y_n) $ if lim $ |x_n / y_n| = 0 $. 
Finally, we use $\widetilde{O}(\cdot)$, $\widetilde{\Omega}(\cdot)$ and $\widetilde{\Theta}(\cdot)$ to hide logarithmic factors in these notations respectively.

\begin{definition}[\textbf{Data Generation Model}]
    \label{data_def}
    Let \(\bm{\mu}_+, \bm{\mu}_- \in \mathbb{R}^d\) be fixed vectors representing the signals contained in data points, where \(\Vert \bm{\mu}_+ \Vert_2 = \Vert \bm{\mu}_- \Vert_2 = \Vert \bm{\mu} \Vert_2 \) and \( \langle \bm{\mu}_+, \bm{\mu}_- \rangle = 0 \) . Then each data point \((\bm{X}, y)\) with \(\bm{X} = (\bm{x}_1, \bm{x}_2, \dots, \bm{x}_M)^\top \in \mathbb{R}^{M \times d}\) and \(y \in \{-1, 1\}\) is generated from the following distribution \(D\):
    \begin{enumerate}
        \item The label \(y\) is generated as a Rademacher random variable.
        \item If \(y = 1\) then \(\bm{x}_1\) is given as \(\bm{\mu}_+\), if \(y = -1\) then \(\bm{x}_1\) is given as \(\bm{\mu}_-\), which represents signals.
        \item A noise vector \(\bm{\xi}_2\) is generated from the Gaussian distribution \(\mathcal{N}(0, \tilde{\sigma}_p^2 \cdot (\bm{I} - \bm{\mu}_+\bm{\mu}_+^{\top} \cdot \Vert \bm{\mu} \Vert_2^{-2} - \bm{\mu}_-\bm{\mu}_-^{\top} \cdot \Vert \bm{\mu} \Vert_2^{-2}))\).
        \item Noise vectors \(\bm{\xi}_3, \dots, \bm{\xi}_M\) is generated i.i.d from the Gaussian distribution \(\mathcal{N}(0, \sigma_p^2 \cdot (\bm{I} - \bm{\mu}_+\bm{\mu}_+^{\top} \cdot \Vert \bm{\mu} \Vert_2^{-2} - \bm{\mu}_-\bm{\mu}_-^{\top} \cdot \Vert \bm{\mu} \Vert_2^{-2}))\).
        \item \(\bm{x}_2, \dots, \bm{x}_M\) are given by \(\bm{\xi}_2, \dots, \bm{\xi}_M\), which represent noises.
    \end{enumerate}
\end{definition}
\paragraph{}
Our data model is envisioned by considering \(M\) tokens of the data points: \(\bm{x}_1, \bm{x}_2, \dots, \bm{x}_M\), which can be seen as patches derived from image data. 
\(\bm{x}_1\) embodies the signal that is inherently linked to the data's class label, while \(\bm{x}_2, \bm{x}_3, \dots, \bm{x}_M\) represents the noise, which is not associated with the label. 
For simplicity, we assume that the noise patches is independently drawn from Gaussian distribution \(\mathcal{N}(0, \tilde{\sigma}_p^2 \cdot (\bm{I} - \bm{\mu}_+\bm{\mu}_+^{\top} \cdot \Vert \bm{\mu} \Vert_2^{-2} - \bm{\mu}_-\bm{\mu}_-^{\top} \cdot \Vert \bm{\mu} \Vert_2^{-2}))\) and \(\mathcal{N}(0, \sigma_p^2 \cdot (\bm{I} - \bm{\mu}_+\bm{\mu}_+^{\top} \cdot \Vert \bm{\mu} \Vert_2^{-2} - \bm{\mu}_-\bm{\mu}_-^{\top} \cdot \Vert \bm{\mu} \Vert_2^{-2}))\), ensuring the noise vectors remains orthogonal to the signal vectors \(\bm{\mu}_+\) and \(\bm{\mu}_-\). 
Also, if \(\tilde{\sigma}_p\) is sufficiently larger than \(\sigma_p\), the input tokens will become sparse, which is consistent with the empirical observation that the \textit{attention} and activation values in Transformer-based models are usually sparse \citep{child2019generating, robinson2023sparse}.
We denote \(\mathrm{SNR} = \Vert \bm{\mu} \Vert_2 / (\sigma_p \sqrt{d})\) to represent the signal-to-noise ratio.

\textbf{Two-layer Transformer.} We consider a two layer Transformer network with a self-attention layer and a fixed linear layer, which is defined as:
\begin{equation}
    \label{ViT}
    f(\bm{X}, \theta) = \frac{1}{M} \sum\limits_{l=1}^M \bm{\varphi} (\bm{x}^\top_l \bm{W}_{Q} \bm{W}_{K}^{\top} \bm{X}^\top ) \bm{X} \bm{W}_V \bm{w}_O.  
\end{equation}
Here, \(\bm{\varphi}(\cdot):\mathbb{R}^M \rightarrow \mathbb{R}^M\) denote the softmax function, \(\bm{W}_Q, \bm{W}_K \in \mathbb{R}^{d \times d_h}, \bm{W}_V \in \mathbb{R}^{d \times d_v}\) denote the query matrix, key matrix, and value matrix, respectively, and \(\bm{w}_O \in \mathbb{R}^{d_v}\) denote the weight for the linear layer. We use \(\theta\) to denote the collection of all the model weights.
This model is a simplified version of the ViT model from \cite{dosovitskiy2020image}, making our analysis focus on the self-attention mechanism which is the most critical component of the ViTs.

Given a training data set \(S = \{(\bm{X}_n, y_n)\}_{n=1}^{N}\) generated from the distribution \(D\) defined in Definition \ref{data_def}, where the subscript $n$ represents the $n$-th sample, we train the two-layer Transformer by minimizing the empirical cross-entropy loss function:
\[
    L_S(\theta) = \frac{1}{N} \sum\limits_{n=1}^{N} \ell(y_{n}f(\bm{X}_{n}, \theta)),
\]
where \(\ell(z)= \log (1+ \exp (-z))\) and \(f(\bm{X}, \theta)\) is the two-layer Transformer. We further defined the population loss (test loss) \(L_D(\theta) := \mathbb{E}_{(\bm{X}, y) \sim D} \ell(yf(\bm{X}, \theta))\).

\textbf{Training algorithm.} We consider Gaussian initialization for the network weights \(\bm{W}_Q, \bm{W}_K\) and \(\bm{W}_V\), where each entry of \(\bm{W}_Q\) and \(\bm{W}_K\) is sampled from a Gaussian distribution \(\mathcal{N}(0, \sigma_h^2)\), and \(\bm{W}_V\) is sampled from \(\mathcal{N}(0, \sigma_V^2)\) at initialization. 
We use gradient descent to optimize training loss \(L_S(\theta)\), and the update of \(\bm{W}_Q, \bm{W}_K\) and \(\bm{W}_V\) can be written as follows:
\[
    \bm{W}^{(t+1)} = \bm{W}^{(t)} - \eta \cdot \nabla_{\bm{W}} L_S(\theta(t)),
\]
where \(\bm{W}^{(t)}\) can be designated as \(\bm{W}_Q, \bm{W}_K\) or \(\bm{W}_V\).
\paragraph{}
Recalling the ViT defined in Eq. \eqref{ViT}, we can intuitively analyze its training dynamics and generalization: if more attention is paid to signals, i.e., \( \bm{x}^\top_l \bm{W}_{Q} \bm{W}_{K}^{\top} \bm{\mu}_\pm \ge \bm{x}^\top_l \bm{W}_{Q} \bm{W}_{K}^{\top} \bm{\xi}_i \) for \( i \in [M]/\{1\} \), then the vector \( \bm{\varphi} (\bm{x}^\top_l \bm{W}_{Q} \bm{W}_{K}^{\top} \bm{X}^\top ) \bm{X} \) has a higher similarity with signals \(\bm{\mu}_\pm\) rather than with noises \( \bm{\xi}_i \). In turn, \(\bm{W}_V\) together with \(\bm{w}_O\) have more chance to learn signals \( \bm{\mu}_\pm \) and utilize them to make prediction. At the same time, if vector \( \bm{W}_V \bm{w}_O  \) learns signals \(\bm{\mu}_\pm\) more than noises \( \bm{\xi}_i \), i.e., \( \bm{\mu}_\pm \bm{W}_V \bm{w}_O \ge \bm{\xi}_i \bm{W}_V \bm{w}_O \), then during the gradient descent training, the gradient send to \( \bm{x}^\top_l \bm{W}_{Q} \bm{W}_{K}^{\top} \bm{\mu}_\pm \) will be larger than that send to \( \bm{x}^\top_l \bm{W}_{Q} \bm{W}_{K}^{\top} \bm{\xi}_i \). As a result, more and more attention is paid on signals and the similarity between signals \(\bm{\mu}_\pm\) and \( \bm{W}_V \bm{w}_O  \) becomes increasingly high. Therefore, the Transformer can perform well on new data points. On the contrary, if much attention is paid to some noises in the training dataset and the model utilizes them for making prediction and optimizing the training loss, then the model can fit the training dataset well but might not perform well on the new test data.

\section{Main Results}
This section presents our main theoretical results which characterize the convergence and generalization of the ViT model under different sample size $N$ and signal-to-noise ratio \(\mathrm{SNR} = \Vert \bm{\mu} \Vert_2 / (\sigma_p \sqrt{d})\). The results are based on the following conditions.
\begin{condition}
    \label{cond}
    Given a sufficiently small failure probability \(\delta>0\) and a target training loss \(\epsilon>0\), suppose that:(1) Dimension \(d_h = \widetilde{\Omega}\footnote{We use \(\widetilde{\Omega}(\cdot)\) and \(\widetilde{O}(\cdot)\) to omit logarithmic terms in the notation. } \Big( \max \{\mathrm{SNR}^4, \mathrm{SNR}^{-4}\} N^2 \epsilon^{-2} \Big) \). (2) Dimension \(d = \widetilde{\Omega} \Big( \epsilon^{-2} N^2 d_h \Big) \). (3) Training sample size \(N = \Omega(\mathrm{polylog}(d))\).(4) The number of input tokens \(M = \Theta(1)\). (5) The $\ell_2$-norm of linear layer weights \(\Vert \bm{w}_O \Vert_2 = \Theta(1)\). (6) The learning rate \(\eta \le \widetilde{O} ( \min \{ \Vert \bm{\mu} \Vert_2^{-2}, ( \sigma_p^2 d )^{-1} \} \cdot d_h^{-\frac{1}{2}} )\). (7) The standard deviation of Gaussian initialization \(\sigma_V\) satisfies: \(\sigma_V \le \widetilde{O} \big( \Vert \bm{w}_O \Vert_2^{-1} \cdot \min \{ \Vert \bm{\mu} \Vert_2^{-1}, ( \sigma_p \sqrt{d} )^{-1} \} \cdot d_h^{-\frac{1}{4}} \big) \). (8) The variance of Gaussian initialization \(\sigma_h^2\) satisfies: \( \min \{ \Vert \bm{\mu} \Vert_2^{-2}, ( \sigma_p^2 d )^{-1} \} \cdot d_h^{-\frac{1}{2}} \cdot \big( \log ( 6N^2M^2 / \delta ) \big)^{-2} \le \sigma_h^2 \le \min \{ \Vert \bm{\mu} \Vert_2^{-2}, ( \sigma_p^2 d )^{-1} \} \cdot d_h^{-\frac{1}{2}} \cdot \big( \log ( 6N^2M^2 / \delta ) \big)^{-\frac{3}{2}} \). (9) Target training loss \( \epsilon \le O(1/\mathrm{polylog}(d)) \). (10) The relationship between \(\sigma_p\) and \(\tilde{\sigma}_p\) satisfies \(\tilde{\sigma}_p = C_p \sigma_p\) and \(C_p = 5 \sqrt{M}\).
\end{condition}
\paragraph{}
Conditions (1) and (2) ensure an over-parametrized learning setting, and similar conditions have been made in the theoretical analysis of CNN models \citep{cao2022benign, kou2023benign}. Condition (3) ensures that there are enough samples in each class with high probability. Conditions (4)-(5) are intended to simplify the calculation, and can be easily generalized to \(M = \Omega(1)\), \(\Vert \bm{w}_O \Vert_2 = o(1)\) or \(\Vert \bm{w}_O \Vert_2 = \omega(1)\) setting.
Conditions (6)-(8) ensure that the Transformer can be effectively trained well. Condition (9) ensures that the Transformer is sufficiently overfitting the training data. Condition (10) ensures the sparsity of the input token features.

\begin{theorem}[\textbf{Benign Overfiting}]
    \label{benign}
    Under Condition \ref{cond}, if \(N \cdot \mathrm{SNR}^2 = \Omega(1)\), then with probability at least \(1 - d^{-1}\), there exist \(T = \Theta ( \eta^{-1} \epsilon^{-1} \Vert \bm{\mu} \Vert_2^{-2} \Vert \bm{w}_O \Vert_2^{-2} )\) such that:
    \begin{enumerate}
        \item The training loss converges to \(\epsilon\): \(L_S(\theta(T)) \le \epsilon\).
        \item The test loss is nearly zero: \(L_D(\theta(T)) \le o(1)\).
    \end{enumerate}
\end{theorem}
\paragraph{}
Theorem \ref{benign} characterizes the case of \textit{benign overfitting}. It shows that as long as \(N \cdot \mathrm{SNR}^2 = \Omega(1)\), the Transformer can generalize well, even though it overfit the training data. To complement the above result and highlight the sharpness of this condition, we present the following theorem for the regime of \textit{harmful overfitting}.
\begin{theorem}[\textbf{Harmful Overfiting}]
    \label{harmful}
    Under Condition \ref{cond}, if \(N^{-1} \cdot \mathrm{SNR}^{-2} = \Omega(1)\), then with probability at least \(1 - d^{-1}\), there exist \(T = \Theta ( N \eta^{-1} \epsilon^{-1} \sigma_p^{-2} d^{-1} \Vert \bm{w}_O \Vert_2^{-2}  )\) such that:
    \begin{enumerate}
        \item The training loss converges to \(\epsilon\): \(L_S(\theta(T)) \le \epsilon\).
        \item The test loss is high: \(L_D(\theta(T)) = \Theta(1) \).
    \end{enumerate}
\end{theorem}
\paragraph{}
Theorem \ref{harmful} shows that if \(N^{-1} \cdot \mathrm{SNR}^{-2} = \Omega(1)\), the trained Transformer has a high test loss as a result of overfitting the noises in the training data. Theorem \ref{benign} and Theorem \ref{harmful} reveals a sharp phase transition between \textit{benign overfitting} and \textit{harmful overfitting}.
\paragraph{Comparison with other results.} 
Firstly, compared with CNNs with \(\mathrm{ReLU}^q\) network result \citep{cao2022benign}, when signal-to-noise ratio is small (\(\mathrm{SNR} \le 1\)), our results show that ViTs require less number of samples to gengeralize well, which reflects the advantage of Transformers.
Secondly, previous theoretical benign overfitting analysis of CNNs with ReLU activations rely heavily on signal strength \(\Vert \bm{\mu} \Vert_2\) and signal-to-noise ratio \(\mathrm{SNR}\) \citep{kou2023benign, meng2023benign}, i.e., \(\mathrm{SNR}^2 = \widetilde{O}(1/N)\) and require \(\Vert \bm{\mu} \Vert_2\) to be large enough, while our analysis does not need to impose restrictions on these conditions. 
Thirdly, previous works only show when Transformer can generalize well \citep{li2023theoretical, deora2023optimization}, while we demonstrate harmful overfitting as a complementary, which reflect that our benign overfitting condition is tighter and more precise.
\section{Proof Sketch}
This section discusses our main challenges in studying ViT's training dynamic, and presents our technical solutions to overcoming them. The complete proofs are given in the appendix.
\subsection{Vectorized Q \& K and scalarized V}
Our first main challenge is to deal with the three matrices \(\bm{W}_Q\), \(\bm{W}_K\) and \(\bm{W}_V\). In contrast to CNN models, whose convolutional kernels can be treated as vectors and thus allow for a more straightforward analytical approach \citep{cao2022benign, kou2023benign}, the QKV matrices within the Transformer model are inherently complex to analyze. Moreover, their mutual interactions further complicate the analysis. 
To circumvent this complexity, \cite{oymak2023role} and \cite{tian2023scan} merge the key-query weights(e.g. \(\bm{W}:= \bm{W}_Q \bm{W}_K^\top\)), \cite{huang2023context, zhang2024trained, jelassi2022vision} employ a specific initialization (e.g. \(\bm{W}_Q^{(0)} = \bm{W}_K^{(0)} = \bm{I}\)). Although their approach simplifies the analysis, it is not conducive to showing the dynamics of QKV interactions, i.e., how \(\bm{W}_Q\), \(\bm{W}_K\), and \(\bm{W}_V\) affect each other. 

In order to simplify the analysis of QKV dynamics without losing rigor, we propose key techniques called \textit{Vectorized Q \& K} and \textit{scalarized V}. 
The basic idea comes from the property that the product of token feature vectors with the QKV matrices results in vectors, e.g., \(\bm{\mu}_+^\top \bm{W}_Q, \bm{\xi}_2^\top \bm{W}_Q\), etc., which are more amenable to analysis. Further, each entry of matrix \(\bm{X} \bm{W}_{Q} \bm{W}_{K}^{\top} \bm{X}^\top\) can be regarded as the inner product of two vectors, e.g., \( \bm{\mu}_+^\top \bm{W}_{Q} \bm{W}_{K}^{\top} \bm{\mu}_+ = \langle \bm{\mu}_+^\top \bm{W}_{Q}, \bm{\mu}_+^\top \bm{W}_{K} \rangle \). Therefore, the dynamics of \textit{attention} can be studied by analyzing the dynamics of the \textit{vectorized Q \& K} defined as follows:
\begin{definition}[Vectorized Q \& K]
    \label{vectorize_qk}
    Let \(\bm{W}_Q^{(t)}\) and \(\bm{W}_K^{(t)}\) be the QK matrices of the ViT at the t-th iteration of gradient descent. Then we define the vectorized Q and vectorized K as follows
    \[ \bm{q}_+^{(t)} = \bm{\mu}_+^\top \bm{W}_Q^{(t)},  \qquad \bm{q}_-^{(t)} = \bm{\mu}_-^\top \bm{W}_Q^{(t)}, \qquad \bm{q}_{n, i}^{(t)} = \bm{\xi}_{n, i}^{\top} \bm{W}_Q^{(t)}, \]
    \[ \bm{k}_+^{(t)} = \bm{\mu}_+^\top \bm{W}_K^{(t)},  \qquad \bm{k}_-^{(t)} = \bm{\mu}_-^\top \bm{W}_K^{(t)}, \qquad \bm{k}_{n, i}^{(t)} = \bm{\xi}_{n, i}^{\top} \bm{W}_K^{(t)}, \]
    for \( i \in [M] \backslash \{1\}, n \in [N] \).
\end{definition}
\paragraph{}
With Definition \ref{vectorize_qk}, denoting \(S_+ := \{ n \in [N] : y_n = 1 \} \), \(S_- := \{ n \in [N] : y_n = -1 \} \), we further analyze the dynamics of the \textit{vectorized Q \& K}. By carefully computing the dynamic of \(\bm{q}_+^{(t)}\), we have
\begin{equation}
\begin{split}
    \label{q_p_dynamic}
    \bm{q}_+^{(t + 1)} - \bm{q}_+^{(t)}
    = \frac{\eta}{NM} \sum\limits_{n \in S_+} - \ell_n^{\prime(t)} \Vert \bm{\mu} \Vert_2^2 \bm{w}_O^\top \bm{W}_V^{(t)\top} \bm{X}_n^{\top} (diag(\bm{\varphi}_{n, 1}^{(t)}) - \bm{\varphi}_{n, 1}^{(t)\top} \bm{\varphi}_{n, 1}^{(t)}) \bm{X}_n \bm{W}_K^{(t)},
\end{split}
\end{equation}
where \(\bm{\varphi}_{n, i}^{(t)} :=  \bm{\varphi} (\bm{x}_{n, i}^\top \bm{W}_{Q}^{(t)} \bm{W}_{K}^{(t)\top} \bm{X}_n^\top ) \) is a shorthand notation,
and \(\bm{w}_O^\top \bm{W}_V^{(t)\top} \bm{X}_n^{\top}\) and \( \bm{X}_n \bm{W}_K^{(t)} \) can be viewed in the following forms
\[
    \bm{w}_O^\top \bm{W}_V^{(t)\top} \bm{X}_n^{\top} = \big( \bm{\mu}_+^\top \bm{W}_V^{(t)} \bm{w}_O, \bm{\xi}_{n, 2}^{\top} \bm{W}_V^{(t)} \bm{w}_O, \dots, \bm{\xi}_{n, M}^{\top} \bm{W}_V^{(t)} \bm{w}_O \big),
\]
\[
    \bm{X}_n \bm{W}_K^{(t)} = \big( \bm{k}_+^{(t)\top}, \bm{k}_{n, 2}^{(t)\top}, \dots, \bm{k}_{n, M}^{(t)\top} \big)^\top.
\]
Thus \(\bm{q}_+^{(t + 1)} - \bm{q}_+^{(t)}\) can be decomposed into a linear combination of \( \bm{k}_+^{(t)} \) and \( \bm{k}_{n, i}^{(t)} \). Therefore, Eq. \eqref{q_p_dynamic} can be further expanded as \( \bm{q}_+^{(t + 1)} - \bm{q}_+^{(t)} = \alpha_{+, +}^{(t)} \bm{k}_+^{(t)} + \sideset{}{_{n \in S_+}}{\sum} \sideset{}{_{i=2}^M}{\sum} \alpha_{n, +, i}^{(t)} \bm{k}_{n, i}^{(t)} \), where \( (\alpha_{+, +}^{(t)}, \alpha_{n, +, 2}^{(t)}, \dots, \alpha_{n, +, M}^{(t)}) =  \frac{\eta}{NM} \sum\limits_{n \in S_+} - \ell_n^{\prime(t)} \Vert \bm{\mu} \Vert_2^2 \bm{w}_O^\top \bm{W}_V^{(t)\top} \bm{X}_n^{\top} (diag(\bm{\varphi}_{n, 1}^{(t)}) - \bm{\varphi}_{n, 1}^{(t)\top} \bm{\varphi}_{n, 1}^{(t)}) \).
Intuitively, if \(\alpha_{+, +}^{(t)}\) is larger enough than \(\alpha_{n, +, i}^{(t)}\), then \(\langle \bm{q}_+^{(t)}, \bm{k}_+^{(t)} \rangle\) will grow faster than \(\langle \bm{q}_+^{(t)}, \bm{k}_{n, i}^{(t)} \rangle\), which means token \(\bm{\mu}_+\) will pay more \textit{attention} to \(\bm{\mu}_+\) rather than \(\bm{\xi}_{n, i}\). 
The dynamics of \( \bm{q}_-^{(t)} \), \( \bm{q}_{n, i}^{(t)} \), \( \bm{k}_+^{(t)} \), \( \bm{k}_-^{(t)} \) and \( \bm{k}_{n, i}^{(t)} \) are similar to \(\bm{q}_+^{(t)}\), and we can study the dynamics of \(QK\) matries by analysing the linear combination coefficients such as \(\alpha_{+, +}^{(t)}\), \(\alpha_{n, +, i}^{(t)}\).

As \(\bm{x}_{n, i}^{\top} \bm{W}_V^{(t)} \bm{w}_O\) are scalars and their dynamics contain a factor \(\Vert \bm{w}_O \Vert_2^2\), we define the \textit{scalarized V} as follows.
\begin{definition}[Scalarized V]
    \label{scalarized_V}
    Let \(\bm{W}_V^{(t)}\) be the V matrix of the ViT at the t-th iteration of gradient descent. Then there exist coefficients \( \gamma_{V, +}^{(t)} \), \( \gamma_{V, -}^{(t)} \), \( \rho_{V, n, i}^{(t)} \) such that
    \[
        \bm{\mu}_+^\top \bm{W}_V^{(t)} \bm{w}_O = \bm{\mu}_+^\top \bm{W}_V^{(0)} \bm{w}_O + \gamma_{V, +}^{(t)} \Vert \bm{w}_O \Vert_2^2,
    \]
    \[
        \bm{\mu}_-^\top \bm{W}_V^{(t)} \bm{w}_O = \bm{\mu}_-^\top \bm{W}_V^{(0)} \bm{w}_O + \gamma_{V, -}^{(t)} \Vert \bm{w}_O \Vert_2^2,
    \]
    \[
        \bm{\xi}_{n, i}^{\top} \bm{W}_V^{(t)} \bm{w}_O = \bm{\xi}_{n, i}^{\top} \bm{W}_V^{(0)} \bm{w}_O + \rho_{V, n, i}^{(t)} \Vert \bm{w}_O \Vert_2^2
    \]
    for \( i \in [M] \backslash \{1\}, n \in [N] \).
\end{definition}
\paragraph{}
With the \textit{Vectorized Q \& K} and \textit{scalarized V}, one can simplify the study of the Transformer learning process to a meticulous calculation of the coefficients such as \( \alpha \), \( \gamma \), \( \rho \) throughout the training period. So how to calculate the dynamics of these coefficients is a key point in our analysis. In the next subsection we describe how to handle \textit{softmax} function and further give bounds for these coefficients.

\subsection{Dealing with the softmax function}
\label{technique2}
Our second challenge is to deal with the \textit{softmax} function, which is the critical component and introduces non-linear transformation in our Transformer model.

As \(\bm{W}_Q\) and \(\bm{W}_K\) are within the \textit{softmax} function and \(\bm{W}_V\) is outside the \textit{softmax} function, we divide the dynamic of training process into two key aspects: (1) How \(\bm{W}_Q\) and \(\bm{W}_K\) affect \(\bm{W}_V\); (2) How \(\bm{W}_V\) affects \(\bm{W}_Q\) and \(\bm{W}_K\). Next, we present our approach to addressing these two critical issues.
\paragraph{1. How \(\bm{W}_Q\) and \(\bm{W}_K\) affect \(\bm{W}_V\):} Without loss of generality, we take a data point \((\bm{X}_n, y_n)\) with \(y_n = 1\) as an example and provide the dynamics for \(\gamma_{V, +}^{(t)}\) and \(\rho_{V, n, i}^{(t)}\) in \(\bm{W}_V\) as follows:
\begin{lemma}[Dynamics of \(\gamma\) and \(\rho\)]
    \label{gamma_rho_dy_main}
    \begin{align*}
        \gamma_{V, +}^{(t + 1)} - \gamma_{V, +}^{(t)}
        &= \frac{\eta \Vert \bm{\mu} \Vert_2^2}{NM} \sum\limits_{n \in S_+, s \in [M]} \frac{ - \ell_n^{\prime(t)} \exp ( \bm{x}_{n, s}^{\top} \bm{W}_{Q}^{(t)} \bm{W}_{K}^{(t)\top} \bm{\mu}_+ )}{  \exp ( \bm{x}_{n, s}^{\top} \bm{W}_{Q}^{(t)} \bm{W}_{K}^{(t)\top} \bm{\mu}_+ ) + \sum\limits_{k=2}^M  \exp ( \bm{x}_{n, s}^{\top} \bm{W}_{Q}^{(t)} \bm{W}_{K}^{(t)\top} \bm{x}_{n, k} )}
    \end{align*}
    \begin{align*}
        \vert \rho_{V, n, i}^{(t+1)} - \rho_{V, n, i}^{(t)} \vert
        \le \Big\vert \frac{2 \eta C_p^2 \sigma_p^2 d}{NM} \sum\limits_{s \in [M]} \frac{ - \ell_n^{\prime(t)} \exp ( \bm{x}_{n, s}^{\top} \bm{W}_{Q}^{(t)} \bm{W}_{K}^{(t)\top} \bm{\xi}_{n, i} )}{  \exp ( \bm{x}_{n, s}^{\top} \bm{W}_{Q}^{(t)} \bm{W}_{K}^{(t)\top} \bm{\mu}_+ ) + \sum\limits_{k=2}^M  \exp ( \bm{x}_{n, s}^{\top} \bm{W}_{Q}^{(t)} \bm{W}_{K}^{(t)\top} \bm{\xi}_{n, k} )} \Big\vert
    \end{align*}
    for \(i \in [M] \backslash \{1\}, n \in S_+\), where \(\ell_n^{\prime(t)} := \ell^{\prime}(y_n f(\bm{X}_n, \theta(t) ) )\) is a shorthand notation. 
\end{lemma}
In benign overfitting regime where \(N \cdot \mathrm{SNR}^2 = \Omega(1)\), the factor \( \sum\limits_{n \in S_+} \frac{\eta \Vert \bm{\mu} \Vert_2^2}{NM}\) is larger than \( \frac{2 \eta C_p^2 \sigma_p^2 d}{NM}\). Therefore, as long as \(\bm{x}_{n, s}^{\top} \bm{W}_{Q}^{(t)} \bm{W}_{K}^{(t)\top} \bm{\mu}_+\) are not less than \(\bm{x}_{n, s}^{\top} \bm{W}_{Q}^{(t)} \bm{W}_{K}^{(t)\top} \bm{\xi}_{n, i}\), \(\gamma_{V, +}^{(t)}\) will grow faster than \(\rho_{V, n, i}^{(t)}\). In other words, if \(\bm{q}_\pm^{(t)}\) and \(\bm{q}_{n, i}^{(t)}\) prefer to align with \(\bm{k}_\pm^{(t)}\) rather than \(\bm{k}_{n, i}^{(t)}\), then \(\bm{W}_V\) would prefer to learn signals rather than memorize noises.
\paragraph{2. How \(\bm{W}_V\) affects \(\bm{W}_Q\) and \(\bm{W}_K\):}
Recalling that \( (\alpha_{+, +}^{(t)}, \alpha_{n, +, 2}^{(t)}, \dots, \alpha_{n, +, M}^{(t)}) =  \frac{\eta}{NM} \sum\limits_{n \in S_+} - \ell_n^{\prime(t)} \Vert \bm{\mu} \Vert_2^2 \bm{w}_O^\top \bm{W}_V^{(t)\top} \bm{X}_n^{\top} (diag(\bm{\varphi}_{n, 1}^{(t)}) - \bm{\varphi}_{n, 1}^{(t)\top} \bm{\varphi}_{n, 1}^{(t)}) \), where the most complicated part is the matrix \((diag(\bm{\varphi}_{n, 1}^{(t)}) - \bm{\varphi}_{n, 1}^{(t)\top} \bm{\varphi}_{n, 1}^{(t)})\).
We observe that the matrix \(diag(\bm{\varphi}_{n, 1}^{(t)}) - \bm{\varphi}_{n, 1}^{(t)\top} \bm{\varphi}_{n, 1}^{(t)}\) has two important properties:
\begin{enumerate}
    \item \label{properties1} The diagonal elements are positive, while the elements on the off-diagonal are negative.
    \item \label{properties2} The sum of each row and column of this matrix is 0.
\end{enumerate}
Based on properties \ref{properties1} and \ref{properties2}, we can deduce the following conclusion: as long as \(\bm{\mu}_+^\top \bm{W}_V^{(t)} \bm{w}_O\) is larger enough than \(\bm{\xi}_{n, i}^{\top} \bm{W}_V^{(t)} \bm{w}_O\), we have \(\alpha_{+, +}^{(t)} \ge 0\) and \(\alpha_{n, +, i}^{(t)} \le 0\), implying that \(\bm{q}_+^{(t)}\) prefers to align \(\bm{k}_+^{(t)}\) rather than \(\bm{k}_{n, i}^{(t)}\). 

\paragraph{}
In summary, if more \textit{attention} is paid to signals, \(\bm{W}_V\) learns the signals faster than it memorizes the noises; conversely, if \(\bm{W}_V\) learns the signals significantly more than it memorizes the noises, more \textit{attention} will be paid to signals rather than noises. In other words, \(\bm{W}_Q\), \(\bm{W}_K\) and \(\bm{W}_V\) promote each other.

\subsection{Three-Stage Decoupling}
Our third main challenge is to deal with the complicated relation among the coefficients, e.g., \(\gamma\), \(\rho\), \(\alpha\). Inspired by the two-stage analysis utilized by \cite{cao2022benign} to decouple the coefficients in CNN models, we analyze the ViT training process in three stage. In the following, we explain the key steps for proving Theorem \ref{benign}. The proof for Theorem \ref{harmful} is similar and we detail it in the appendix.

\paragraph{Stage 1:} 
The analysis in \ref{technique2} shows that \(\bm{W}_Q\), \(\bm{W}_K\) and \(\bm{W}_V\) can promote each other. But this process of mutual reinforcement requires some conditions, e.g. \(\bm{\mu}_+^\top \bm{W}_V^{(t)} \bm{w}_O\) is sufficiently larger than \(\bm{\xi}_{n, i}^{\top} \bm{W}_V^{(t)} \bm{w}_O\) which does not necessarily hold for Gaussian initialization, complicating the proof. Stage 1 was introduced to solve this problem, and Lemma \ref{stage1_main} formally describes this stage:
\begin{lemma}[V's Beginning of Learning Signals]
    \label{stage1_main}
    Under the same conditions as Theorem \ref{benign}, there exist \(T_1 = \frac{10M(3M + 1) N}{\eta d_h^{\frac{1}{4}} (N \Vert \bm{\mu} \Vert_2^2 - 60M^2 C_p^2 \sigma_p^2 d) \Vert \bm{w}_O \Vert_2^2} \) such that 
    the first element of vector \( \bm{X}_n \bm{W}_V^{(t)} \bm{w}_O \) dominate its other elements, that is, \( \bm{\mu}_+^\top \bm{W}_V^{(t)} \bm{w}_O \ge 3 M \cdot \vert \bm{\xi}_{n, i}^{\top} \bm{W}_V^{(t)} \bm{w}_O \vert \) for all \( n \in S_+ \), \( i \in [M] \backslash \{1\} \) and
    \( \bm{\mu}_-^\top \bm{W}_V^{(t)} \bm{w}_O \le - 3 M \cdot \vert \bm{\xi}_{n, i}^{\top} \bm{W}_V^{(t)} \bm{w}_O \vert \) for all \( n \in S_- \), \( i \in [M] \backslash \{1\} \).
\end{lemma}
With Lemma \ref{stage1_main}, we can guarantee the monotonicity of \textit{attention} on signals for \(t \ge T_1\), which allows us to concentrate on analyzing the growth rate of \textit{attention} in the following stages.

\paragraph{Stage 2:} 
Note that the output of \textit{softmax} function has an upper bound 1, thus the output of ViT can be bounded by \(\max \{\vert \bm{\mu}_\pm^\top \bm{W}_V^{(t)} \bm{w}_O \vert, \vert \bm{\xi}_{n, i}^{\top} \bm{W}_V^{(t)} \bm{w}_O \vert\}\). By Lemma \ref{gamma_rho_dy_main}, it can be proved that there exists \(T_2 = \Theta \Big( \eta^{-1} \Vert \bm{\mu} \Vert_2^{-2} \Vert \bm{w}_O \Vert_2^{-2} \log (6N^2M^2 / \delta)^{-1} \Big)\) such that: \(\max \{\vert \bm{\mu}_\pm^\top \bm{W}_V^{(t)} \bm{w}_O \vert, \vert \bm{\xi}_{n, i}^{\top} \bm{W}_V^{(t)} \bm{w}_O \vert\} = o(1)\), and further get \(1/2 - o(1) \le - \ell_n^{\prime(t)} \le 1/2 + o(1)\). Therefore, one can simplify the dynamics of the coefficients (e.g., \(\gamma\), \(\rho\), \(\alpha\)) by plugging the tight bounds of \(- \ell_n^{\prime(t)}\). The following lemma provides some bounds for the dynamics of \(\bm{W}_Q\), \(\bm{W}_K\) and\(\bm{W}_V\) in Stage 2.

\begin{lemma}[Dynamics of QKV in Stage 2]
    \label{dynamics_QK}
    Under Condition \ref{cond}, if \(N \cdot \mathrm{SNR}^2 = \Omega(1)\), then with probability at least \(1 - d^{-1}\), there exist constant \(C\), \(T_1 = \Theta \Big( \eta^{-1} d_h^{- \frac{1}{4}} \Vert \bm{\mu} \Vert_2^{-2} \Vert \bm{w}_O \Vert_2^{-2} \Big)\) and \(T_2 = \widetilde{O} \Big( \eta^{-1} \Vert \bm{\mu} \Vert_2^{-2} \Vert \bm{w}_O \Vert_2^{-2} \Big)\) such that:
    \[
        \bm{\mu}_+^\top \bm{W}_V^{(t)} \bm{w}_O \ge \eta C \Vert \bm{\mu} \Vert_2^2 \Vert \bm{w}_O \Vert_2^2 (t - T_1), \quad \bm{\mu}_+^\top \bm{W}_V^{(t)} \bm{w}_O \ge 3 M \cdot \vert \bm{\xi}_{n, i}^{\top} \bm{W}_V^{(T)} \bm{w}_O \vert,
    \]
    \[
        \bm{\mu}_-^\top \bm{W}_V^{(t)} \bm{w}_O \le - \eta C \Vert \bm{\mu} \Vert_2^2 \Vert \bm{w}_O \Vert_2^2 (t - T_1), \quad \bm{\mu}_-^\top \bm{W}_V^{(t)} \bm{w}_O \le - 3 M \cdot \vert \bm{\xi}_{n, i}^{\top} \bm{W}_V^{(T)} \bm{w}_O \vert,
    \]
    \begin{align*}
    \resizebox{.95\linewidth}{!}{$
        \langle \bm{q}_\pm^{(t)}, \bm{k}_\pm^{(t)} \rangle - \langle \bm{q}_\pm^{(t)}, \bm{k}_{n, j}^{(t)} \rangle \ge \log \Big( \exp \big( \langle \bm{q}_\pm^{(T_1)}, \bm{k}_\pm^{(T_1)} \rangle - \langle \bm{q}_\pm^{(T_1)}, \bm{k}_{n, j}^{(T_1)} \rangle \big) + \frac{\eta^2 C \Vert \bm{\mu} \Vert_2^4 \Vert \bm{w}_O \Vert_2^2 d_h^{\frac{1}{2}}}{N \big( \log (6N^2M^2 / \delta) \big)^{2} } \cdot (t - T_1)(t - T_1 - 1) \Big)
        $}
    \end{align*}
    \begin{align*}
    \resizebox{.95\linewidth}{!}{$
        \langle \bm{q}_{n, i}^{(t)}, \bm{k}_\pm^{(t)} \rangle - \langle \bm{q}_{n, i}^{(t)}, \bm{k}_{n, j}^{(t)} \rangle \ge \log \Big( \exp \big( \langle \bm{q}_{n, i}^{(T_1)}, \bm{k}_\pm^{(T_1)} \rangle - \langle \bm{q}_{n, i}^{(T_1)}, \bm{k}_{n, j}^{(T_1)} \rangle \big) + \frac{\eta^2 C \sigma_p^2 d \Vert \bm{\mu} \Vert_2^2 \Vert \bm{w}_O \Vert_2^2 d_h^{\frac{1}{2}} }{N \big( \log (6N^2M^2 / \delta) \big)^{2}} \cdot (t - T_1)(t - T_1 - 1) \Big)
        $}        
    \end{align*}
    for \(i, j \in [M] \backslash \{1\}, n \in [N], t \in [T_1, T_2] \).
\end{lemma}
\paragraph{}

Lemma \ref{dynamics_QK} presents the training dynamics of \(\bm{W}_Q\), \(\bm{W}_K\) and \(\bm{W}_V\) under benign overfitting regime in two aspects: \textit{direction} and \textit{speed}, that is,
\begin{itemize}
    \item \textbf{direction:} \(\bm{W}_V \bm{w}_O\) prefer to learn the signals rather than memorize the noises; more and more \textit{attention} is paid to signals, while the \textit{attention} on noises is decreasing.
    \item \textbf{speed:} The inner products of \(\bm{\mu}_\pm\) and \(\bm{W}_V \bm{w}_O\) grow at a linear rate; the inner products of label-related \textit{vectorized Q \& K} grow at a logarithmic rate.
\end{itemize}
\paragraph{Stage 3:} As the training process going on, the loss begins to converge, and the loss derivatives \(- \ell_n^{\prime(t)}\) no longer remain near \(1/2\). 
Therefore, the increasing rate of the inner products of \textit{vectorize Q \& K} and \(\bm{\mu}_\pm^\top \bm{W}_V \bm{w}_O\) begins to diminish. Based on the analysis in Stage 2, the \textit{attention} is sufficiently sparse at the beginning of Stage 3. In other words, the \textit{attention} on the signals is nearly 1, while the \textit{attention} on the noises is nearly 0. According to this sparsity, \(\bm{\mu}_\pm^\top \bm{W}_V^{(t)} \bm{w}_O\) will grow much faster than \(\bm{\xi}_{n, i}^{\top} \bm{W}_V^{(t)} \bm{w}_O \) by Lemma \ref{gamma_rho_dy_main}, and Lemma \ref{stage3_main} provides the lower bound of \(\bm{\mu}_\pm^\top \bm{W}_V^{(t)} \bm{w}_O\).
\begin{lemma}
    \label{stage3_main}
    Under the same conditions as Theorem \ref{benign}, there exists \(T_3 = \Theta \Big( \eta^{-1} \epsilon^{-1} \Vert \bm{\mu} \Vert_2^{-2} \Vert \bm{w}_O \Vert_2^{-2} \log (6N^2M^2 / \delta)^{-1} \Big)\) such that:
    \[
        \bm{\mu}_+^\top \bm{W}_V^{(t)} \bm{w}_O \ge \log \Big( \exp ( \bm{\mu}_+^\top \bm{W}_V^{(T_2)} \bm{w}_O ) + \eta C \Vert \bm{\mu} \Vert_2^2 \Vert \bm{w}_O \Vert_2^2 (t - T_2) \Big), 
    \]
    \[
        \bm{\mu}_-^\top \bm{W}_V^{(t)} \bm{w}_O \le - \log \Big( \exp ( - \bm{\mu}_-^\top \bm{W}_V^{(T_2)} \bm{w}_O ) + \eta C \Vert \bm{\mu} \Vert_2^2 \Vert \bm{w}_O \Vert_2^2 (t - T_2) \Big)
    \]
    for \(t \in [T_2, T_3]\), where \(C\) is a constant, \(T_2\) is the last iteration of stage 2.
\end{lemma}
\paragraph{Convergence:}
Lemma \ref{stage3_main} provides logarithmic lower bounds for \(\bm{\mu}_\pm^\top \bm{W}_V^{(t)} \bm{w}_O\). Plugging \(t = T_3\) into these inequality, we have \(\vert \bm{\mu}_\pm^\top \bm{W}_V^{(t)} \bm{w}_O \vert \ge \log \big( \Theta(1/\epsilon) \big) \) , then as long as \( \vert \bm{\xi}_{n, i}^{\top} \bm{W}_V^{(t)} \bm{w}_O \vert \) is sufficiently small, it can be proved that \(y_n f(\bm{X}_n, \theta(t) ) \ge \log \big( 1/\epsilon \big) \) for all \( n \in [N] \), thus \(\ell(y_n f(\bm{X}_n, \theta(t) ) ) = \log ( 1 + \exp( - \log ( 1/\epsilon ) ) ) \le \epsilon\) and \(L_S(\theta(t)) \le \epsilon\). The convergence of ViT is accordingly obtained.

\paragraph{Generalization:} Consider a new data point \((\bm{X}, y)\) generated from the distribution defined in Definition \ref{data_def}. Without loss of generality, we suppose that the signal token is \(\bm{\mu}_+\) and the label is 1, i.e. \(\bm{X} = ( \bm{\mu}_+, \bm{\xi}_2, \dots, \bm{\xi}_M) \), \(y = 1\). It is clear that \(\bm{\xi}_i^\top \bm{W}_{Q} \bm{W}_{K}^{\top} \bm{\mu}_\pm\) has a mean zero, which implies that the \textit{attention} on the signals in the test data may not necessarily be as high as that in the training data. However, the sparsity of \textit{attention} during training process creates conditions for \(\bm{W}_V \bm{w}_O\) to utilize the signals to make predictions, facilitating the generalization of the ViT. We provide the lower bound for the output of ViT on the unseen data with high probability as follows:
\begin{lemma}
    \label{output_bound_main}
    Under the same conditions as Theorem \ref{benign}, there exists \(T_3 = \Theta \Big( \eta^{-1} \epsilon^{-1} \Vert \bm{\mu} \Vert_2^{-2} \Vert \bm{w}_O \Vert_2^{-2} \log (6N^2M^2 / \delta)^{-1} \Big)\), with probability at least \(1 - \delta / N^2M\),
    \[
        y f(\bm{X}, \theta (T_3)) \ge \frac{\log ( C / \epsilon )}{C^\prime} - 1,
    \]
\end{lemma}
where \(C\) and \(C^\prime\) are constants. Lemma \ref{output_bound_main} shows that \(y f(\bm{X}, \theta (T_3))\) is large with high probability, and through careful calculation, we can provide a small bound for test loss. More details are in Appendix \ref{population_loss_section}.

\section{Experimental Verification}
\label{experimental_verification}
We present simulation results on synthetic data and MNIST dataset to verify our theoretical results.

\textbf{Synthetic data experiments setting:} We follow Definition \ref{data_def} to generate the training set and test set. Specifically, we set token size \(M = 16\) and feature dimension \(d = 1024\). Without loss of generality, we set \(\bm{\mu}_+ = \Vert \bm{\mu} \Vert_2 \cdot [1, 0, \dots, 0]^\top \) and \(\bm{\mu}_- = \Vert \bm{\mu} \Vert_2 \cdot [0, 1, 0, \dots, 0]^\top \). We generate noise vector \(\bm{\xi}_2\) from the Gaussian distribution \(\mathcal{N}(0, \tilde{\sigma}_p^2 \bm{I})\), where \(\tilde{\sigma}_p\) is fixed to 4. Similarly, we generate the other noise vectors \(\bm{\xi}_i\) for \(i \in [M] / \{1, 2\}\) from the Gaussian distribution \(\mathcal{N}(0, \sigma_p^2 \bm{I})\) where \(\sigma_p\) is fixed to 0.2.

We consider a two-layer Transformer defined in Section \ref{problem_setup}. The dimensions of matrix \(\bm{W}_Q\), \(\bm{W}_K\) and \(\bm{W}_V\) are set as \(d_h = d_v = 512\). The ViT parameters are initialized using PyTorch's default initialization method, and then they are divided by 16 to ensure that the weights are initialized small enough. We train the ViT with full-batch gradient descent and learning rate \(\eta = 0.1\), target training loss \(\epsilon = 0.01\). We consider different traning sample size \(N\) ranging from 2 to 20, and different signal-to noise-ratio \(\mathrm{SNR}\) ranging from 0.16 to 15.6. We evaluate the test loss with 100 test data points after training loss converges to \(\epsilon\). All experiments are performed on an NVIDIA A100 GPU.
\begin{figure}[htbp]
  \centering
  \begin{subfigure}[b]{0.45\textwidth}
    \includegraphics[width=\textwidth]{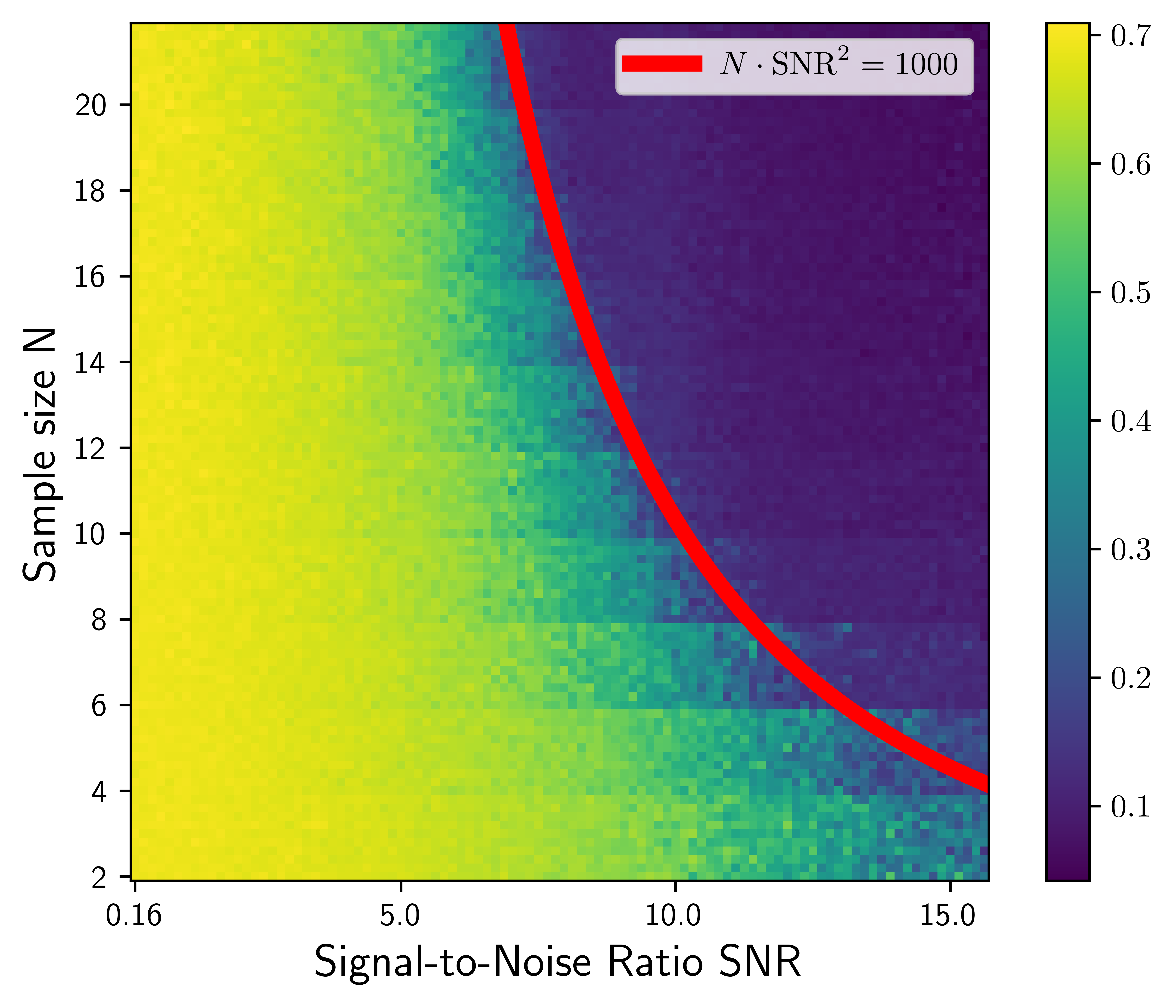}
    \caption{Test Loss Heatmap}
    \label{fig:sub1}
  \end{subfigure}
  \hfill
  \begin{subfigure}[b]{0.45\textwidth}
    \includegraphics[width=\textwidth]{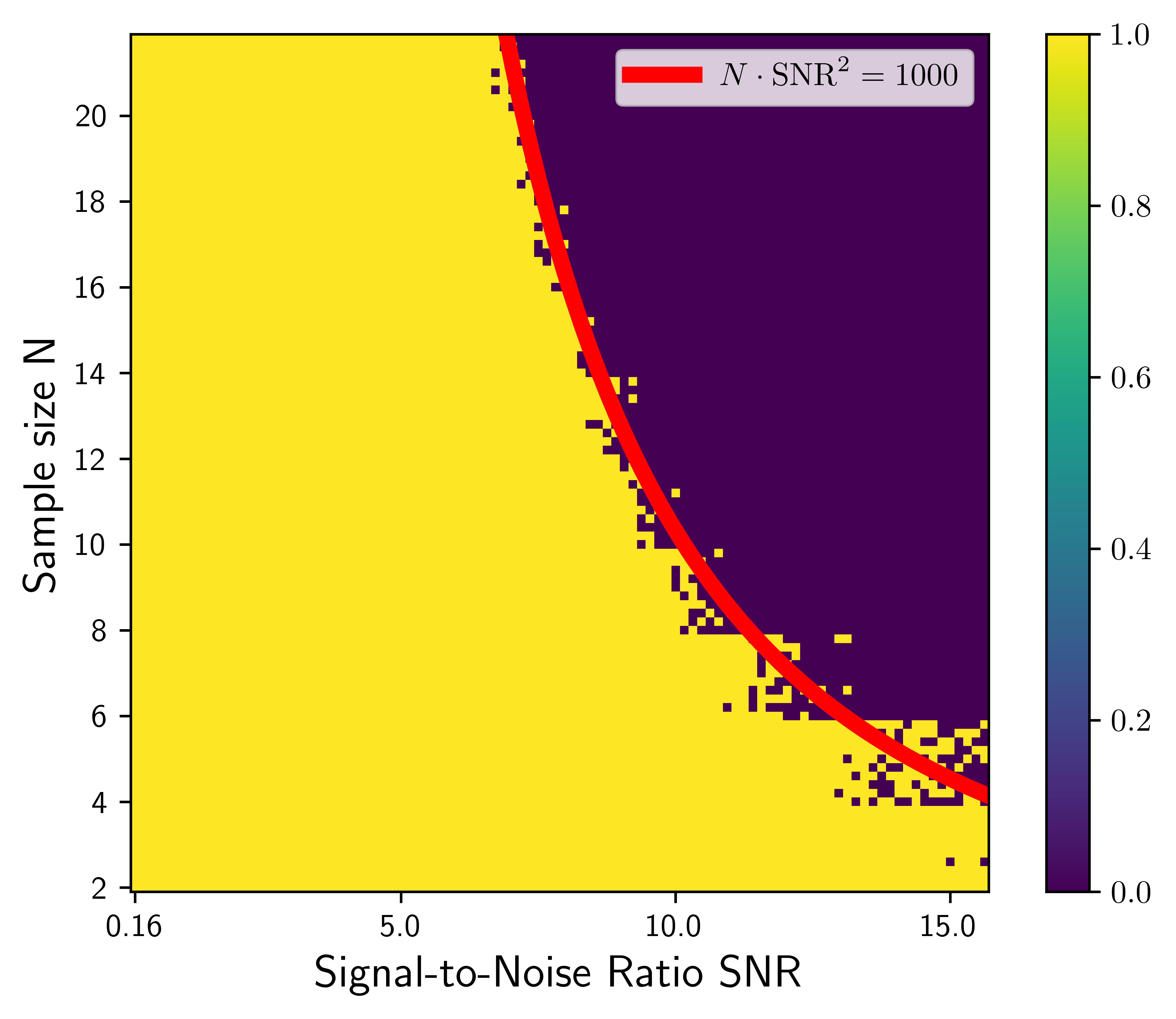}
    \caption{Cutoff Test Loss Heatmap}
    \label{fig:sub2}
  \end{subfigure}
  \caption{(a) is a heatmap of test loss on synthetic data across various signal-to-noise ratios (\(\mathrm{SNR}\)) and sample sizes (N). High test losses are indicated with yellow, while low test losses are indicated with purple. (b) is a heatmap that applies a cutoff value 0.2. It categorizes values below 0.2 as 0 (purple), and above 0.2 as 1 (yellow). The expression for the red curves in (a) and (b) is \( N \cdot \mathrm{SNR}^2 = 1000 \).}
  \label{fig:heatmap}
\end{figure}

\vspace{-1em}
\textbf{Synthetic data experiments results:} Figure \ref{fig:sub1} shows that as \(N\) and \(\mathrm{SNR}\) increase, the test loss tends to decrease. As Figure \ref{fig:sub2} shows, the theoretically derived red curve (\(N \cdot \mathrm{SNR}^2 = 1000\)) almost follows the experimental dividing line between the yellow area (test loss > 0.2) and the purple area (test loss < 0.2). These experimental results further validate our theoretical results, that is, the sharp condition separation between benign and harmful overfitting, where \(N \cdot \mathrm{SNR}^2 = \Omega(1)\) is a precise condition for benign overfitting and \(N^{-1} \cdot \mathrm{SNR}^{-2} = \Omega(1)\) is a precise condition for harmful overfitting. More experimental results on the dynamics of QKV can be found in Appendix \ref{training_dynamics}.

\begin{figure}[htbp]
    \centering
    \begin{subfigure}[b]{0.31\textwidth}
      \includegraphics[width=\textwidth]{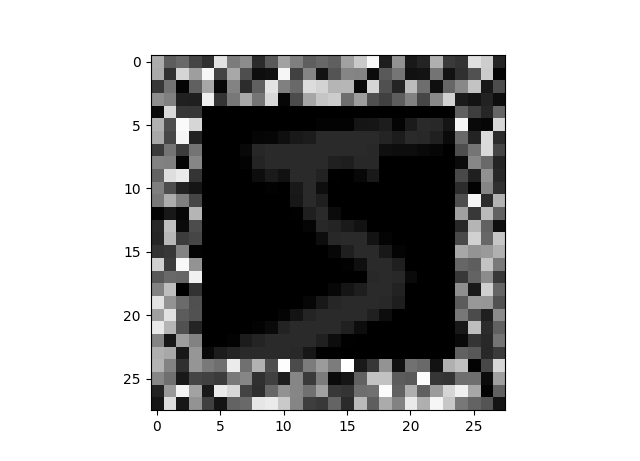}
      \caption{SNR=0.2}
      \label{fig2:sub1}
    \end{subfigure}
    \begin{subfigure}[b]{0.3\textwidth}
      \includegraphics[width=\textwidth]{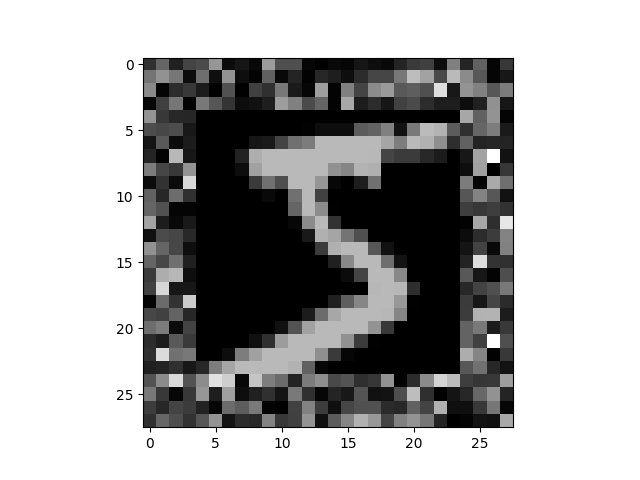}
      \caption{SNR=2.0}
      \label{fig2:sub2}
    \end{subfigure}
    \begin{subfigure}[b]{0.25\textwidth}
        \includegraphics[width=\textwidth]{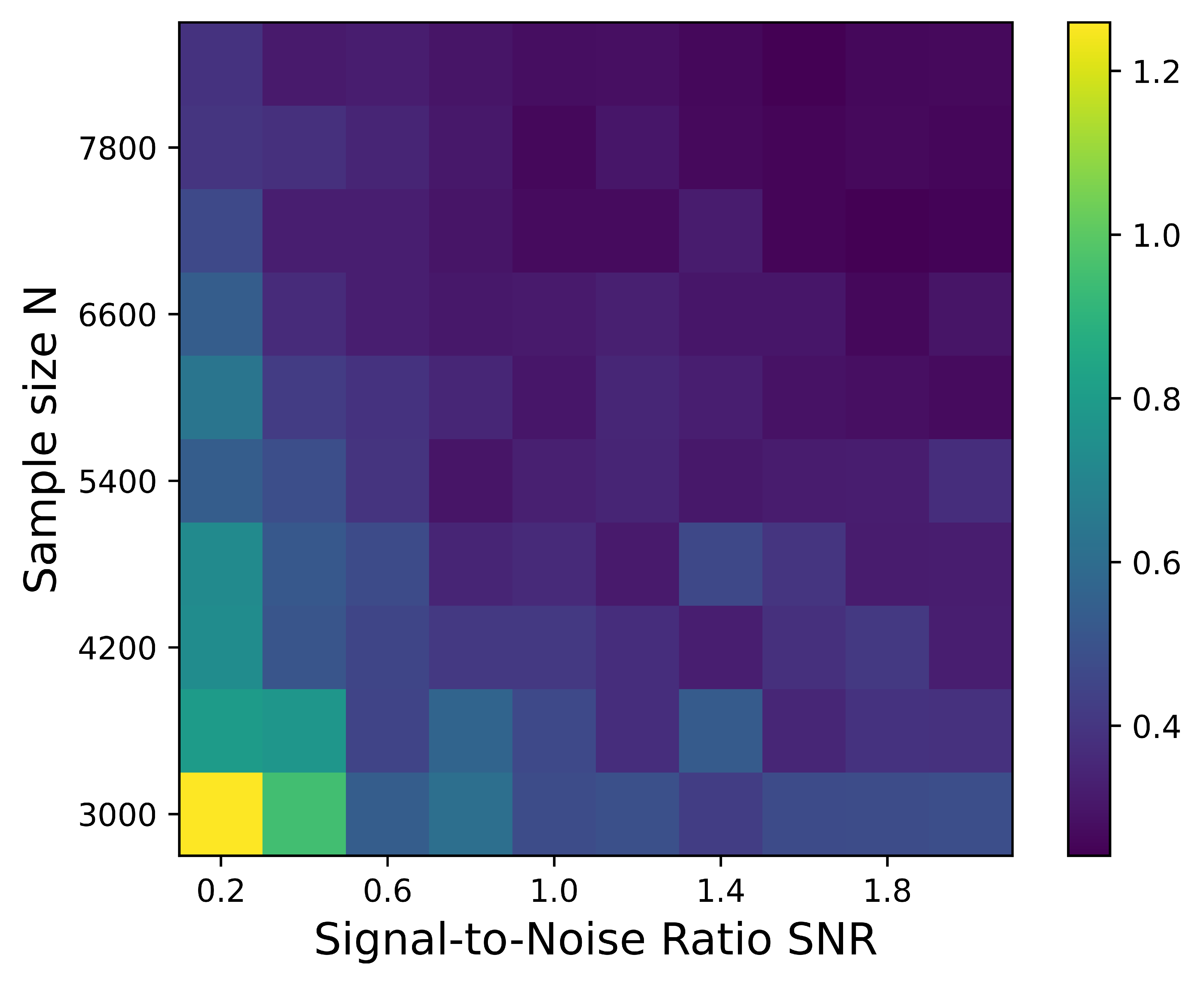}
        \caption{Test Loss Heatmap}
        \label{fig2:sub3}
    \end{subfigure}
    \caption{MNIST experiments}
    \label{fig2}
\end{figure}

\textbf{MNIST experiments setting:}
We add Gaussian random noises to the outer regions of the images with a width of 4.
The original images and the noises are multiplied by diferent factors to generate diferent dataset with specifc SNR. 
For example, in Figure 2 (a), we mutiply the original image by 1/6 and the noises by 5/6, thus SNR=0.2; 
in Figure 2 (b),we mutiply the original image by 2/3 and the noises by 1/3, thus SNR=2.0.
We consider a ViT model that consists of two attention layers, each equipped with four self-attention heads, followed by a MLP with ReLU activation. 
The training sample size N ranges from 3000 to 8400 and the SNR ranges from 0.2 to 2.0.

\textbf{MNIST experiments results:}
Figure 2 (c) shows the heatmap of test loss, which shows a transition between benign and harmful overfitting regimes. 
The larger the sample size N and signal-to-noise ratio SNR, the better the generalization performance.

\vspace{-0.5em}
\section{Conclusion}
\vspace{-0.2em}
This paper studies the training dynamics, convergence, and generalization for a two-layer Transformer in vision. By analyzing the \textit{Vectorized Q \& K} and \textit{scalarized V} in three-stage decomposition and carefully handling the \textit{softmax} function, we give the precise increasing rate of the \textit{attention} and output of the Transformer in the training process, and further analyze its generalization performance. Our theoretical results reveal a sharp condition separation between benign and harmful overfitting.
One limitation of our methodology is our proof technique relies heavily on the sparsity of feature strength, e.g., we assume the standard deviation \(\tilde{\sigma}_p\) is sufficiently larger than \(\sigma_p\), ensuring that more \textit{attention} is paid to \(\bm{\xi}_2\) rather than other tokens. 
One future direction is to generalize our analysis to study other abilities of Transformers, such as in-context learning, prompt-tuning and time series forecasting.

\section*{Acknowledgement}
We thank the reviewers for their constructive comments on both theoretical and experimental aspects.
Miao Zhang was partially sponsored by the National Natural Science Foundation of China under Grant 62306084 and U23B2051, and Shenzhen College Stability Support Plan under Grant GXWD20231128102243003.

\bibliographystyle{apalike} 
\bibliography{arxiv.bib}

\newpage


\appendix

\begin{center}
	\LARGE \bf {Appendix}
\end{center}

\etocdepthtag.toc{mtappendix}
\etocsettagdepth{mtchapter}{none}
\etocsettagdepth{mtappendix}{subsection}
\tableofcontents
\clearpage

\section{More Experimental Rresults on Training Dynamics}
\label{training_dynamics}
In this section, we follow a similar setting as the experiments in Section \ref{experimental_verification}, and add more experiments. All the experiments in this paper can be performed within hours.
\paragraph{Experiments setting:}
We focus on tracking the dynamics of QKV under benign and harmful overfitting regime, and the key parameters are as follows:
\begin{itemize}
    \item \(M = 4\)
    \item \(d = 2048\)
    \item \(d_h = 1024\)
    \item \(d_v = 256\)
    \item \(\sigma_p = 0.05\)
    \item \(\tilde{\sigma}_p = 1\)
\end{itemize}
Specifically, we set \(n = 16\), \(\Vert \bm{\mu} \Vert_2 = 25\) as benign overfitting setting, and set \(n = 10\), \(\Vert \bm{\mu} \Vert_2 = 15\) as harmful overfitting setting.

\begin{figure}[htbp]
    \centering
    \begin{subfigure}[b]{0.228\textwidth} 
        \includegraphics[width=\textwidth]{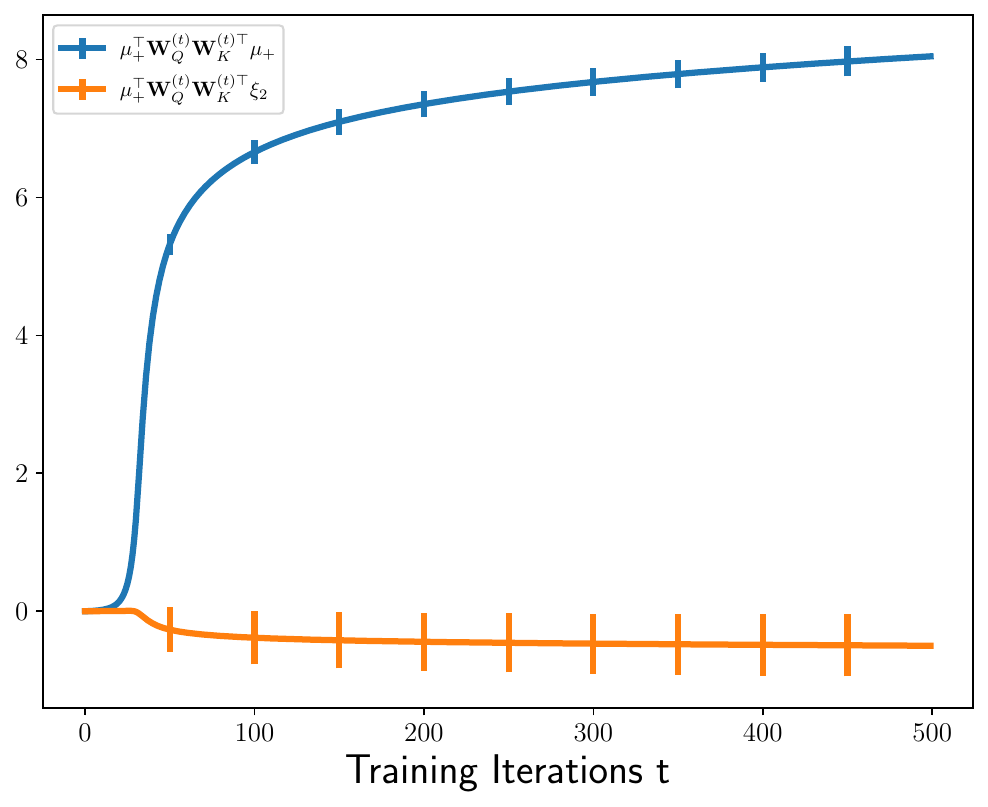}
        \label{fig:benign_QK}
    \end{subfigure}
    \begin{subfigure}[b]{0.242\textwidth}
        \includegraphics[width=\textwidth]{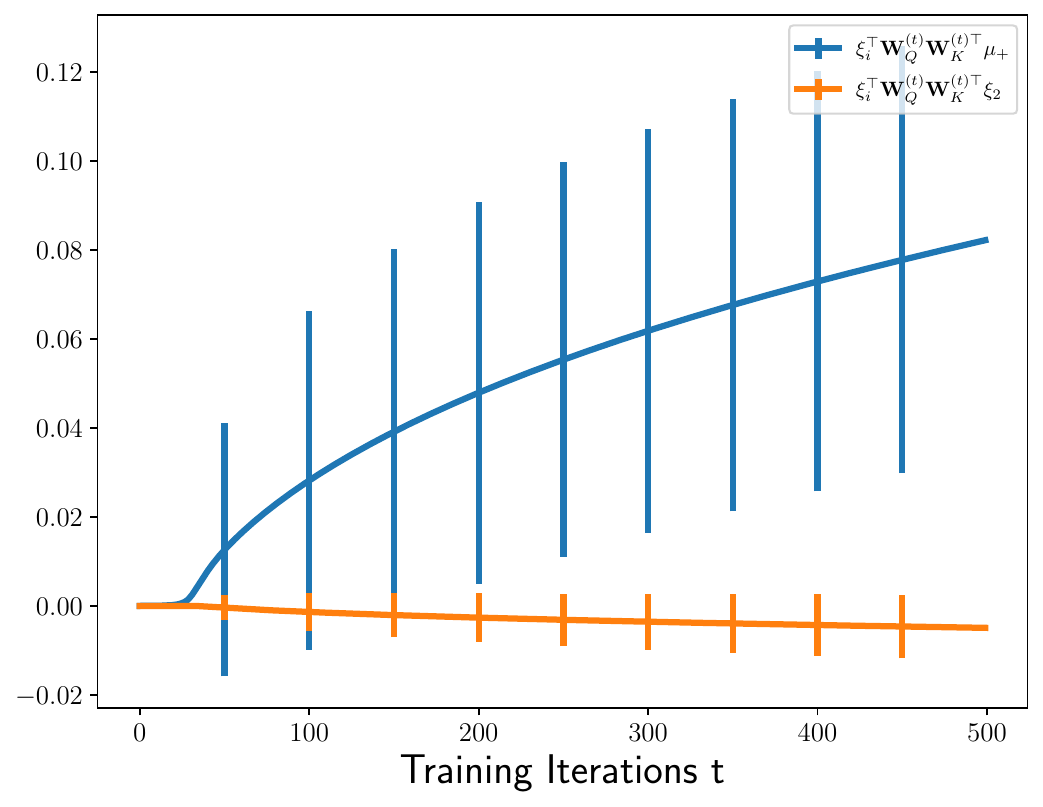}
        \label{fig:benign_QK2}
    \end{subfigure}
    \begin{subfigure}[b]{0.23\textwidth}
        \includegraphics[width=\textwidth]{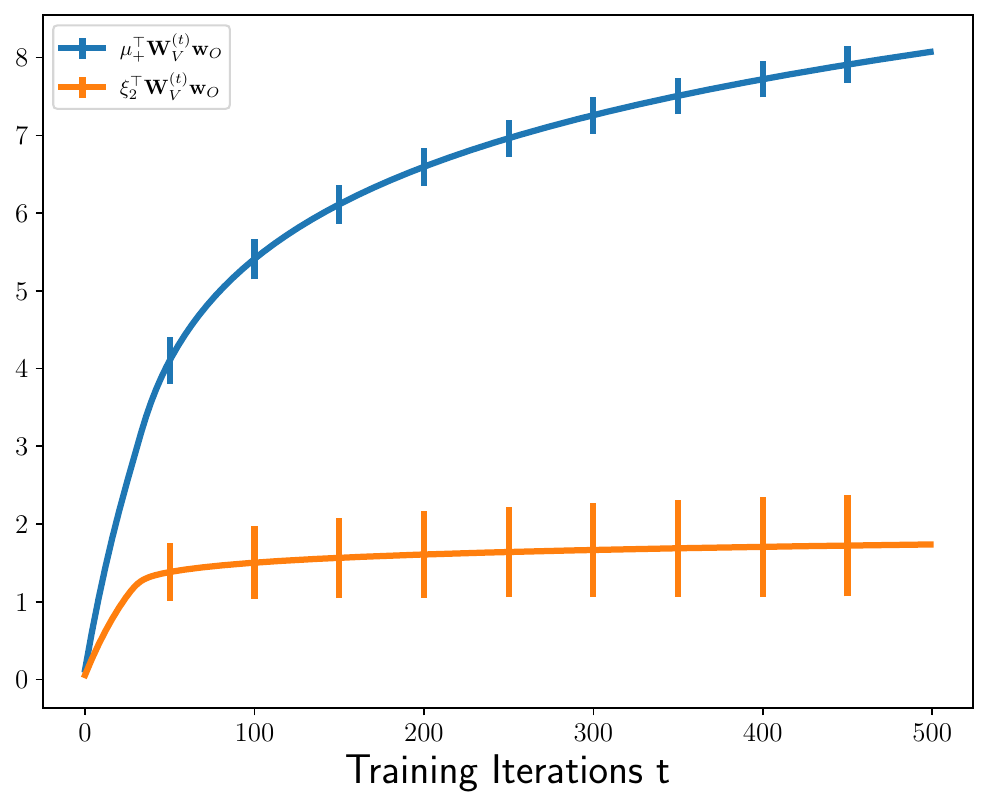}
        \label{fig:benign_V}
    \end{subfigure}
    \begin{subfigure}[b]{0.235\textwidth}
        \includegraphics[width=\textwidth]{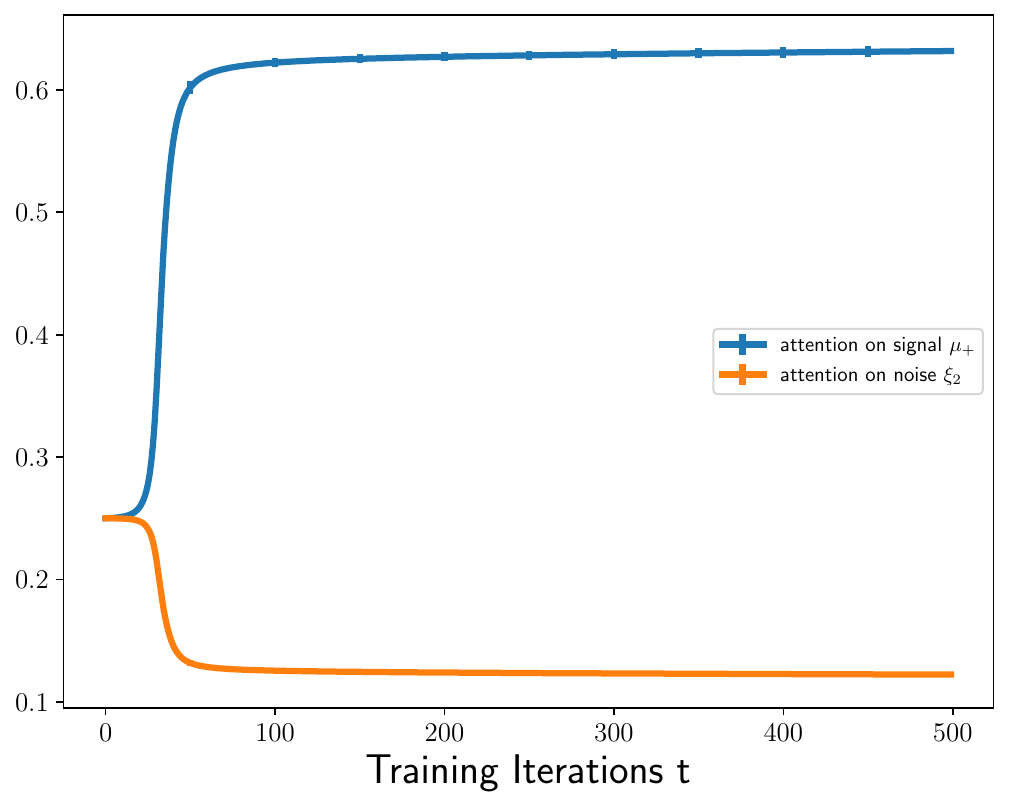}
        \label{fig:benign_attn}
    \end{subfigure}
    \caption{Training Dynamics Under Benign Overfitting Regime}
    \label{fig:benign}
\end{figure}

\begin{figure}[htbp]
    \centering
    \begin{subfigure}[b]{0.23\textwidth} 
        \includegraphics[width=\textwidth]{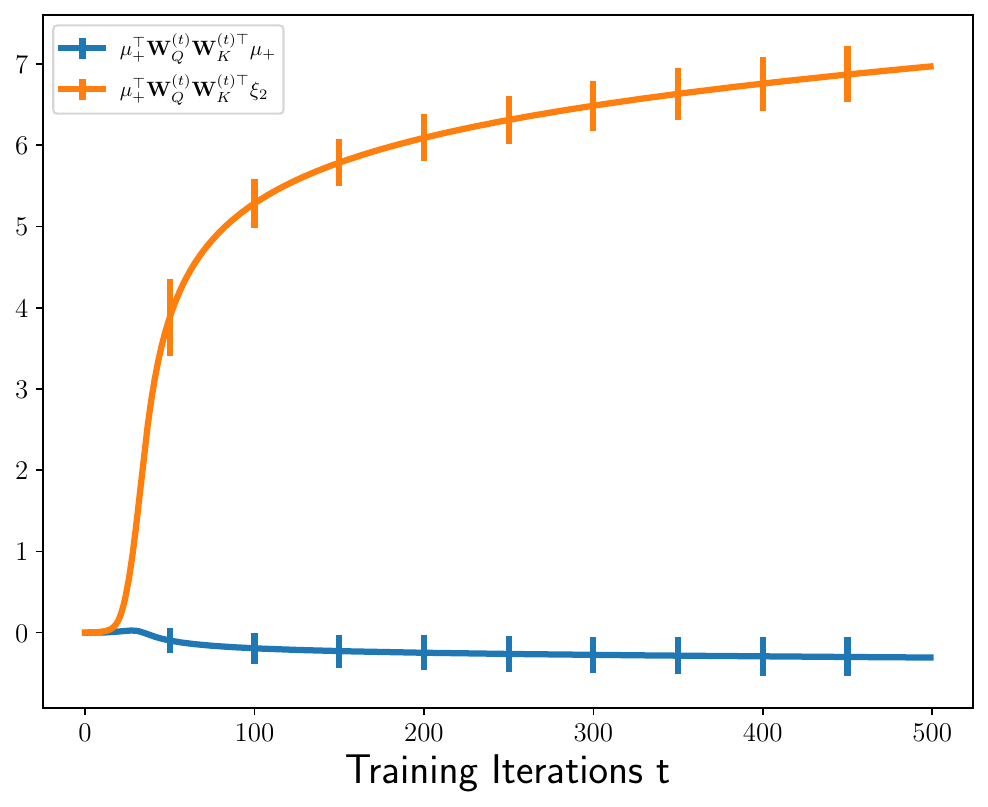}
        \label{fig:harmful_QK}
    \end{subfigure}
    \begin{subfigure}[b]{0.238\textwidth}
        \includegraphics[width=\textwidth]{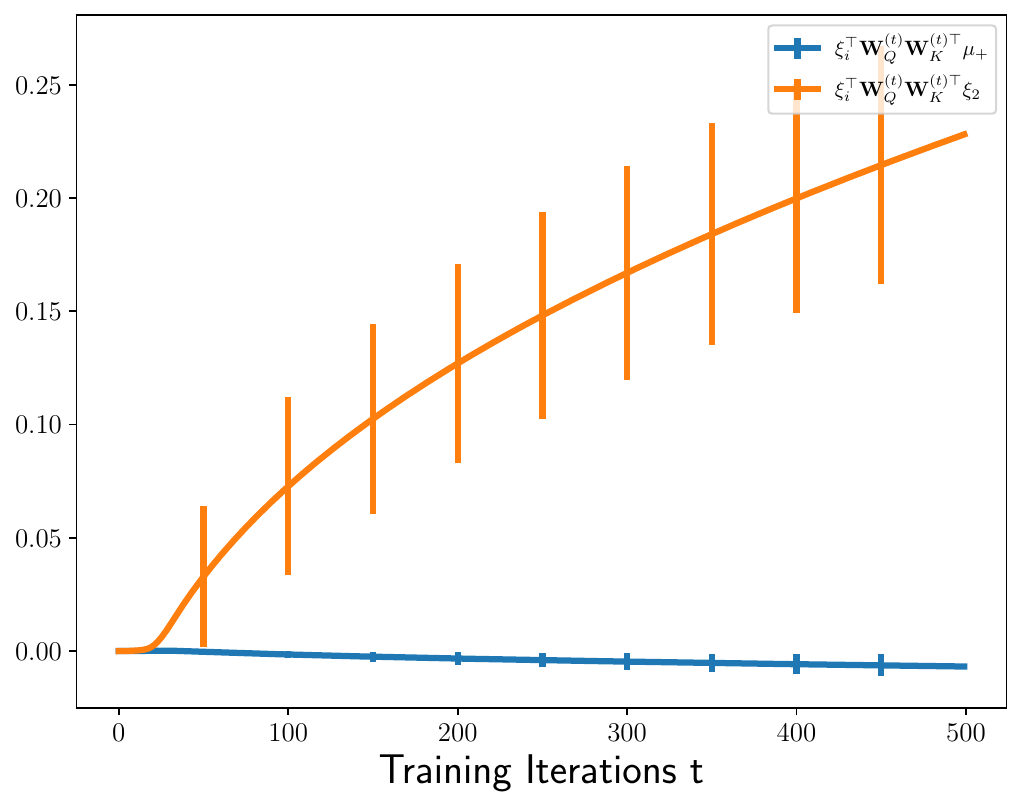}
        \label{fig:harmful_QK2}
    \end{subfigure}
    \begin{subfigure}[b]{0.23\textwidth}
        \includegraphics[width=\textwidth]{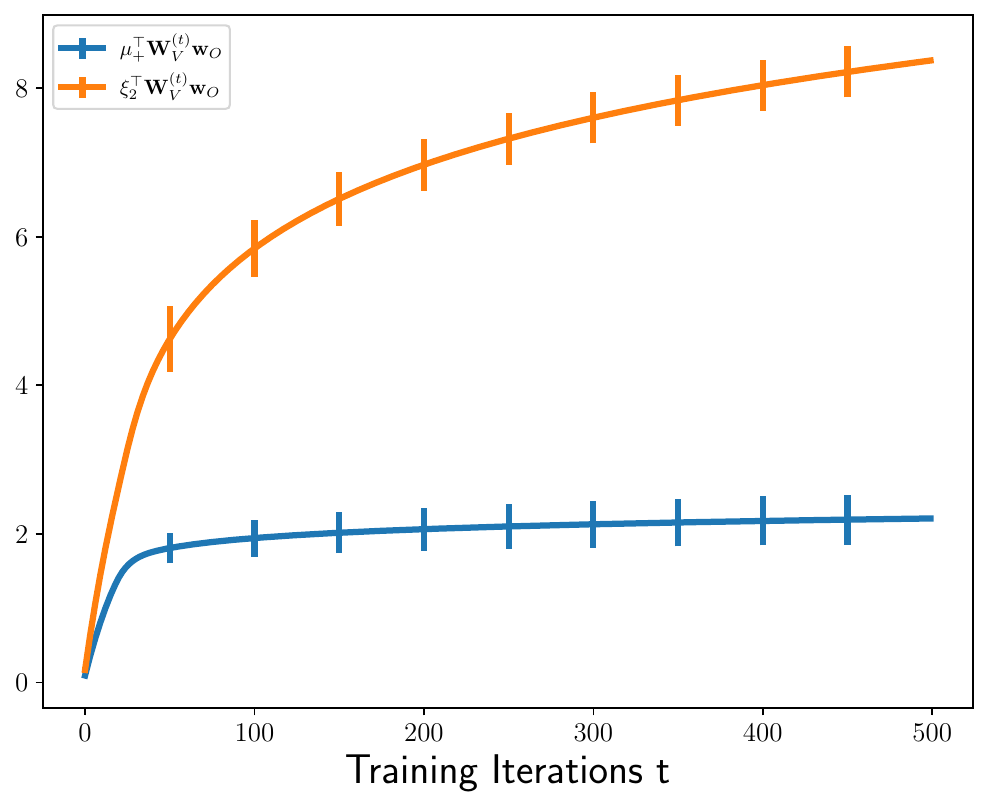}
        \label{fig:harmful_V}
    \end{subfigure}
    \begin{subfigure}[b]{0.235\textwidth}
        \includegraphics[width=\textwidth]{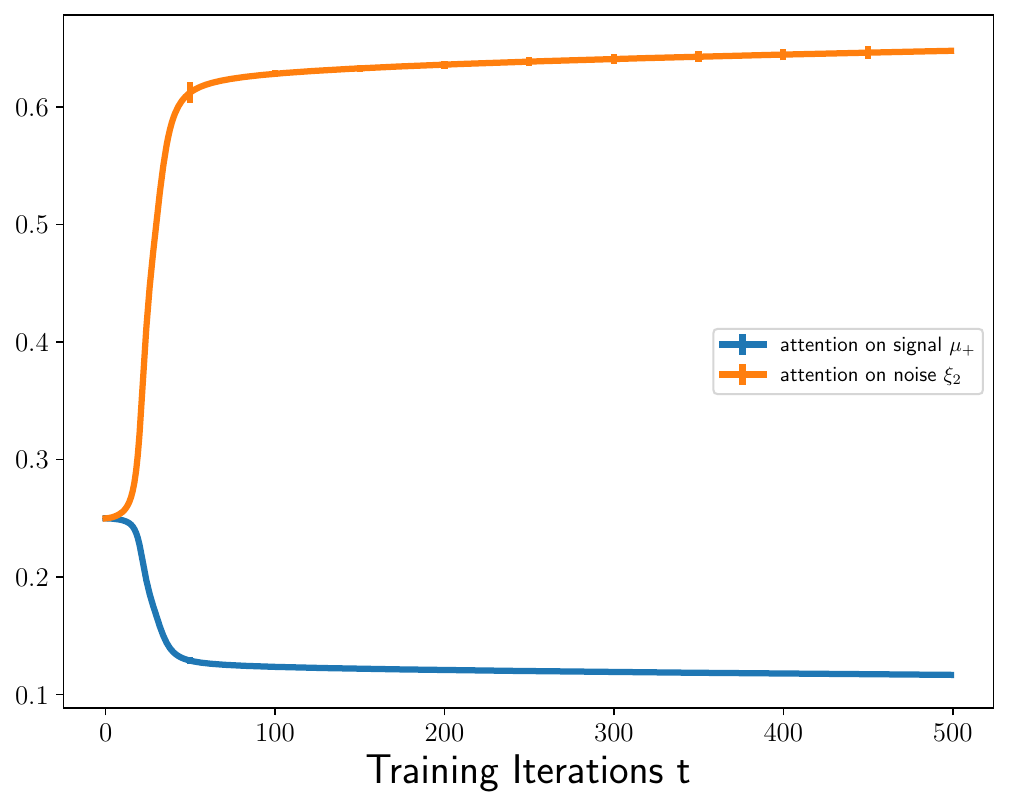}
        \label{fig:harmful_attn}
    \end{subfigure}
    \caption{Training Dynamics Under Harmful Overfitting Regime}
    \label{fig:harmful}
\end{figure}
\paragraph{Experiments results:} Figure \ref{fig:benign} shows the training dynamics under benign overfitting setting, and Figure \ref{fig:harmful} shows the training dynamics under harmful overfitting setting.
In Figure \ref{fig:benign}, \( \bm{x}^\top_l \bm{W}_{Q} \bm{W}_{K}^{\top} \bm{\mu}_+\) grow faster than \(\bm{x}^\top_l \bm{W}_{Q} \bm{W}_{K}^{\top} \bm{\xi}_2 \), 
and more and more \textit{attention} are paid on signal \(\bm{\mu}_+\). Meanwhile, \( \bm{\mu}_+ \bm{W}_V \bm{w}_O \) grows faster than \( \bm{\xi}_2 \bm{W}_V \bm{w}_O \).
In Figure \ref{fig:harmful}, \(\bm{x}^\top_l \bm{W}_{Q} \bm{W}_{K}^{\top} \bm{\xi}_2 \) grow faster than \( \bm{x}^\top_l \bm{W}_{Q} \bm{W}_{K}^{\top} \bm{\mu}_+\), 
and more and more \textit{attention} are paid on noise \(\bm{\xi}_2\). Meanwhile, \( \bm{\xi}_2 \bm{W}_V \bm{w}_O \) grows faster than \( \bm{\mu}_+ \bm{W}_V \bm{w}_O \).
In other words, the ViT prefers to learn the signals rather than memorize the noises under benign overfitting setting, while it prefers to memorize the noises rather than learn the signals under harmful overfitting setting.

\section{Basic Calculation}

In this section, we introduce some notations for this paper, and give expressions for the exact gradient of loss function \(L_S(\theta)\) with respect to \(\bm{W}_Q, \bm{W}_K\) and \(\bm{W}_V\).

\subsection{Notations}

\begin{table}
\footnotesize
    \centering
    \caption{Key notations in this paper}
    \label{tab:mytable}
    \begin{tabular}{ll}
        \toprule
        \textbf{Symbols} & \textbf{Definitions} \\
        \midrule
        \(\bm{x}_{n, i}\) & the i-th token in the n-th training sample \\ 
                          & if \(i \in [M] \backslash \{1\}\), \(\bm{x}_{n, i} = \bm{\xi}_{n, i}\).  \\
        \midrule
        \(\bm{\varphi}_{n, i}^{(t)}\) & the i-th row of \textit{attention} for the n-th sample, i.e., \(\bm{\varphi}_{n, i}^{(t)} := \bm{\varphi} (\bm{x}_{n, i}^\top \bm{W}_{Q}^{(t)} \bm{W}_{K}^{(t)\top} \bm{X}_n^\top ) \) \\
        \midrule
        \(S_+, S_-\) & the training samples with +1 labels and -1 labels, \\
                     &i.e. ,\(S_+ := \{ n \in [N] : y_n = 1 \} \), \(S_- := \{ n \in [N] : y_n = -1 \} \)\\
        \midrule
        \(\bm{q}_+^{(t)}, \bm{q}_-^{(t)}, \bm{q}_{n, i}^{(t)}\) & vectorized Q, defined as \(\bm{q}_+^{(t)} = \bm{\mu}_+^\top \bm{W}_Q^{(t)}, \bm{q}_-^{(t)} = \bm{\mu}_-^\top \bm{W}_Q^{(t)}, \bm{q}_{n, i}^{(t)} = \bm{\xi}_{n, i}^{\top} \bm{W}_Q^{(t)}\) \\
        \midrule
        \(\bm{k}_+^{(t)}, \bm{k}_-^{(t)}, \bm{k}_{n, i}^{(t)}\) & vectorized K, defined as \(\bm{k}_+^{(t)} = \bm{\mu}_+^\top \bm{W}_K^{(t)}, \bm{k}_-^{(t)} = \bm{\mu}_-^\top \bm{W}_K^{(t)}, \bm{k}_{n, i}^{(t)} = \bm{\xi}_{n, i}^{\top} \bm{W}_K^{(t)}\) \\
        \midrule
        \(V_+^{(t)}, V_-^{(t)}, V_{n, i}^{(t)}\) & scalarized V, defined as  \( V_+^{(t)} := \bm{\mu}_+^\top \bm{W}_V^{(t)} \bm{w}_O \), \( V_-^{(t)} := \bm{\mu}_-^\top \bm{W}_V^{(t)} \bm{w}_O \), \( V_{n, i}^{(t)} := \bm{\xi}_{n, i}^\top \bm{W}_V^{(t)} \bm{w}_O \)\\
        \midrule
        \(\alpha_{\pm, \pm}^{(t)}, \alpha_{n, \pm, i}^{(t)}\) & linear combinations coefficients for the dynamics of \(\bm{q}_+^{(t)}\) and \(\bm{q}_-^{(t)}\), \\
                                                              & i.e., \(\bm{q}_\pm^{(t + 1)} - \bm{q}_\pm^{(t)} = \alpha_{\pm, \pm}^{(t)} \bm{k}_\pm^{(t)} + \sideset{}{_{n \in S_\pm}}{\sum} \sideset{}{_{i=2}^M}{\sum} \alpha_{n, \pm, i}^{(t)} \bm{k}_{n, i}^{(t)}\) \\
        \midrule
        \(\alpha_{n, i, \pm}^{(t)}, \alpha_{n, i, n^\prime, i^\prime}^{(t)}\) & linear combinations coefficients for the dynamics of \(\bm{q}_{n, i}^{(t)}\), \\
                                                                              & i.e., \(\bm{q}_{n, i}^{(t + 1)} - \bm{q}_{n, i}^{(t)} = \alpha_{n, i, +}^{(t)} \bm{k}_+^{(t)} + \alpha_{n, i, -}^{(t)} \bm{k}_-^{(t)} + \sideset{}{_{n^\prime=1}^N}{\sum} \sideset{}{_{i^\prime=2}^M}{\sum} \alpha_{n, i, n^\prime, i^\prime}^{(t)} \bm{k}_{n^\prime, i^\prime}^{(t)}\) \\
        \midrule
        \(\beta_{\pm, \pm}^{(t)}, \beta_{n, \pm, i}^{(t)}\) & linear combinations coefficients for the dynamics of \(\bm{k}_+^{(t)}\) and \(\bm{k}_-^{(t)}\), \\
                                                            & i.e., \(\bm{k}_\pm^{(t + 1)} - \bm{k}_\pm^{(t)} = \beta_{\pm, \pm}^{(t)} \bm{q}_\pm^{(t)} + \sideset{}{_{n \in S_\pm}}{\sum} \sideset{}{_{i=2}^M}{\sum} \beta_{n, \pm, i}^{(t)} \bm{q}_{n, i}^{(t)}\) \\
        \midrule
        \(\beta_{n, i, \pm}^{(t)}, \beta_{n, i, n^\prime, i^\prime}^{(t)}\) & linear combinations coefficients for the dynamics of \(\bm{q}_{n, i}^{(t)}\), \\
                                                                            & i.e., \(\bm{k}_{n, i}^{(t + 1)} - \bm{k}_{n, i}^{(t)} = \beta_{n, i, +}^{(t)} \bm{q}_+^{(t)} + \beta_{n, i, -}^{(t)} \bm{q}_-^{(t)} + \sideset{}{_{n^\prime=1}^N}{\sum} \sideset{}{_{i^\prime=2}^M}{\sum} \beta_{n, i, n^\prime, i^\prime}^{(t)} \bm{q}_{n^\prime, i^\prime}^{(t)}\) \\
        \midrule
        \(softmax(\langle \bm{q}_\pm^{(t)}, \bm{k}_\pm^{(t)} \rangle)\) & a general references to \(\frac{ \exp ( \langle \bm{q}_+^{(t)}, \bm{k}_+^{(t)} \rangle )}{  \exp ( \langle \bm{q}_+^{(t)}, \bm{k}_+^{(t)} \rangle ) + \sum\limits_{k=2}^M  \exp ( \langle \bm{q}_+^{(t)}, \bm{k}_{n, k}^{(t)} \rangle )}\) for \(n \in S_+\), \\
                                                                        & and \(\frac{ \exp ( \langle \bm{q}_-^{(t)}, \bm{k}_-^{(t)} \rangle )}{  \exp ( \langle \bm{q}_-^{(t)}, \bm{k}_-^{(t)} \rangle ) + \sum\limits_{k=2}^M  \exp ( \langle \bm{q}_-^{(t)}, \bm{k}_{n, k}^{(t)} \rangle )}\) for \(n \in S_-\) \\
        \midrule
        \(softmax(\langle \bm{q}_{n, i}^{(t)}, \bm{k}_\pm^{(t)} \rangle)\) & a general references to \(\frac{ \exp ( \langle \bm{q}_{n, i}^{(t)}, \bm{k}_+^{(t)} \rangle )}{  \exp ( \langle \bm{q}_{n, i}^{(t)}, \bm{k}_+^{(t)} \rangle ) + \sum\limits_{k=2}^M  \exp ( \langle \bm{q}_{n, i}^{(t)}, \bm{k}_{n, k}^{(t)} \rangle )}\) for \(n \in S_+, i \in [M] \backslash \{1\}\), \\
                                                                           & and \(\frac{ \exp ( \langle \bm{q}_{n, i}^{(t)}, \bm{k}_-^{(t)} \rangle )}{  \exp ( \langle \bm{q}_{n, i}^{(t)}, \bm{k}_-^{(t)} \rangle ) + \sum\limits_{k=2}^M  \exp ( \langle \bm{q}_{n, i}^{(t)}, \bm{k}_{n, k}^{(t)} \rangle )}\) for \(n \in S_-, i \in [M] \backslash \{1\}\) \\
        \midrule
        \(softmax(\langle \bm{q}_\pm^{(t)}, \bm{k}_{n, j}^{(t)} \rangle)\) & a general references to \(\frac{ \exp ( \langle \bm{q}_+^{(t)}, \bm{k}_{n, j}^{(t)} \rangle )}{  \exp ( \langle \bm{q}_+^{(t)}, \bm{k}_+^{(t)} \rangle ) + \sum\limits_{k=2}^M  \exp ( \langle \bm{q}_+^{(t)}, \bm{k}_{n, k}^{(t)} \rangle )}\) for \(n \in S_+, i \in [M] \backslash \{1\}\), \\
                                                                           & and \(\frac{ \exp ( \langle \bm{q}_-^{(t)}, \bm{k}_{n, j}^{(t)} \rangle )}{  \exp ( \langle \bm{q}_-^{(t)}, \bm{k}_-^{(t)} \rangle ) + \sum\limits_{k=2}^M  \exp ( \langle \bm{q}_-^{(t)}, \bm{k}_{n, k}^{(t)} \rangle )}\) for \(n \in S_-, i \in [M] \backslash \{1\}\) \\
        \midrule
        \(softmax(\langle \bm{q}_{n, i}^{(t)}, \bm{k}_{n, j}^{(t)} \rangle)\) & a general references to \(\frac{ \exp ( \langle \bm{q}_{n, i}^{(t)}, \bm{k}_{n, j}^{(t)} \rangle )}{ \exp ( \langle \bm{q}_{n, i}^{(t)}, \bm{k}_+^{(t)} \rangle ) + \sum\limits_{k=2}^M  \exp ( \langle \bm{q}_{n, i}^{(t)}, \bm{k}_{n, k}^{(t)} \rangle )}\) for \(n \in S_+, i, j \in [M] \backslash \{1\}\), \\
                                                                              & and \(\frac{ \exp ( \langle \bm{q}_{n, i}^{(t)}, \bm{k}_{n, j}^{(t)} \rangle )}{ \exp ( \langle \bm{q}_{n, i}^{(t)}, \bm{k}_-^{(t)} \rangle ) + \sum\limits_{k=2}^M  \exp ( \langle \bm{q}_{n, i}^{(t)}, \bm{k}_{n, k}^{(t)} \rangle )}\) for \(n \in S_-, i, j \in [M] \backslash \{1\}\) \\
        \midrule
        \(\Lambda_{n, \pm, j}^{(t)}, \Lambda_{n, i, \pm, j}^{(t)}\) & \(\Lambda_{n, \pm, j}^{(t)} := \langle \bm{q}_\pm^{(t)}, \bm{k}_\pm^{(t)} \rangle - \langle \bm{q}_\pm^{(t)}, \bm{k}_{n, j}^{(t)} \rangle, \quad \Lambda_{n, i, \pm, j}^{(t)} := \langle \bm{q}_{n, i}^{(t)}, \bm{k}_\pm^{(t)} \rangle - \langle \bm{q}_{n, i}^{(t)}, \bm{k}_{n, j}^{(t)} \rangle \) \\
        \midrule
        \(\Psi_{\pm}^{(t)}, \Psi_{n, \pm, j}^{(t)}, \Psi_{n, i, \pm}^{(t)}, \Psi_{n, i, j}^{(t)}\) & \(\Psi_{\pm}^{(t)} := \langle \bm{q}_\pm^{(t)}, \bm{k}_{n, 2}^{(t)} \rangle - \langle \bm{q}_\pm^{(t)}, \bm{k}_\pm^{(t)} \rangle, \quad \Psi_{n, \pm, j}^{(t)} := \langle \bm{q}_\pm^{(t)}, \bm{k}_{n, 2}^{(t)} \rangle - \langle \bm{q}_\pm^{(t)}, \bm{k}_{n, j}^{(t)} \rangle \), \\
                                                                                                   & \(\Psi_{n, i, \pm}^{(t)} := \langle \bm{q}_{n, i}^{(t)}, \bm{k}_{n, 2}^{(t)} \rangle - \langle \bm{q}_{n, i}^{(t)}, \bm{k}_\pm^{(t)} \rangle, \quad \Psi_{n, i, j}^{(t)} := \langle \bm{q}_{n, i}^{(t)}, \bm{k}_{n, 2}^{(t)} \rangle - \langle \bm{q}_{n, i}^{(t)}, \bm{k}_{n, j}^{(t)} \rangle \) \\
        \bottomrule
    \end{tabular}
\end{table}

\FloatBarrier

\clearpage

\subsection{Gradient Calculation}

\textbf{Gradient of Softmax}. Before computing the gradient of QKV, we solve for the gradient of Softmax function. Suppose there is an input vector \(\bm{a} = (a_1, a_2, \dots, a_m)\), then the softmax function can be defined as follows: 

\[
    softmax(a_i) = \frac{\exp(a_i)}{ \sum\limits_{j=1}^{m} \exp(a_j) }  
\]

Then we calculate the gradient of \(softmax(a_i)\) with respect to \(a_i\) and \(a_j\) for \(i,j \in [m], i \ne j\).

\begin{equation}
\begin{split}
    \label{softmax_1}
    \frac{\partial \ softmax(a_i)}{\partial \ a_i} &= \frac{\exp(a_i) \sum\limits_{j=1}^{m} \exp(a_j) - \exp(a_i) \exp(a_i)}{ (\sum\limits_{j=1}^{m} \exp(a_j))^2 } \\
    &= \frac{\exp(a_i)}{ \sum\limits_{j=1}^{m} \exp(a_j) } \Big( 1 - \frac{\exp(a_i)}{ \sum\limits_{j=1}^{m} \exp(a_j) } \Big) \\
    &= softmax(a_i) \cdot \Big( 1 - softmax(a_i) \Big)
\end{split}
\end{equation}

\begin{equation}
\begin{split}
    \label{softmax_2}
    \frac{\partial \ softmax(a_i)}{\partial \ a_j} &= - \frac{ \exp(a_i) \exp(a_j) }{ (\sum\limits_{j=1}^{m} \exp(a_j))^2 } \\
    &= softmax(a_i) \cdot softmax(a_j)
\end{split}
\end{equation}

\eqref{softmax_1} and \eqref{softmax_2} show that the gradient of a Softmax function can be represented as a Jacobi matrix where the elements on the diagonal are \(softmax(a_i) \cdot \Big( 1 - softmax(a_i) \Big)\) and the elements on the off-diagonal are \(softmax(a_i) \cdot softmax(a_j)\).
With these properties, the gradient of our attention vector \(\bm{\varphi}_{n, i}\) (\(i \in [M], n \in [N]\)) can be expressed as follows

\begin{equation}
    \label{softmax_3}
    \frac{\partial \bm{\varphi}_{n, i} }{\partial \big( \bm{x}_{n, i}^\top \bm{W}_{Q} \bm{W}_{K}^{\top} \bm{X}_n^\top \big) } = diag(\bm{\varphi}_{n, i}) - \bm{\varphi}_{n, i}^\top \bm{\varphi}_{n, i} 
\end{equation}

\begin{lemma}
    \label{A1}
    the gradients of loss function \(L_S(\theta)\) with respect to \(\bm{W}_Q, \bm{W}_K\) and \(\bm{W}_V\) are given by
    \begin{equation*}
    \begin{split}
        \nabla_{\bm{W}_Q}L_S(\theta)
        &= \frac{1}{NM} \sum\limits_{n \in S_+} \ell_n^{\prime}(\theta) ( \bm{\mu}_+ \bm{w}_O^\top \bm{W}_V^\top \bm{X}_n^\top (diag(\bm{\varphi}_{n, 1}) - \bm{\varphi}_{n, 1}^\top \bm{\varphi}_{n, 1}) \\
        &+ \sum\limits_{i=2}^M \bm{\xi}_{n, i} \bm{w}_O^\top \bm{W}_V^\top \bm{X}_n^\top (diag(\bm{\varphi}_{n, i}) - \bm{\varphi}_{n, i}^\top \bm{\varphi}_{n, i}) ) \bm{X}_n \bm{W}_K \\
        &- \frac{1}{NM} \sum\limits_{n \in S_-} \ell_n^{\prime}(\theta) ( \bm{\mu}_- \bm{w}_O^\top \bm{W}_V^\top \bm{X}_n^\top (diag(\bm{\varphi}_{n, 1}) - \bm{\varphi}_{n, 1}^\top \bm{\varphi}_{n, 1}) \\
        &+ \sum\limits_{i=2}^M \bm{\xi}_{n, i} \bm{w}_O^\top \bm{W}_V^\top \bm{X}_n^\top (diag(\bm{\varphi}_{n, i}) - \bm{\varphi}_{n, i}^\top \bm{\varphi}_{n, i}) ) \bm{X}_n \bm{W}_K ,
    \end{split}
    \end{equation*}

    \begin{equation*}
    \begin{split}
        \nabla_{\bm{W}_K}L_S(\theta)
        &= \frac{1}{NM} \sum\limits_{n \in S_+} \ell_n^{\prime}(\theta) ( \bm{X}_n^\top (diag(\bm{\varphi}_{n, 1}) - \bm{\varphi}_{n, 1}^\top \bm{\varphi}_{n, 1}) \bm{X}_n \bm{W}_V \bm{w}_O \bm{\mu}_+^\top \\
        &+ \sum\limits_{i=2}^M \bm{X}_n^\top (diag(\bm{\varphi}_{n, i}) - \bm{\varphi}_{n, i}^\top \bm{\varphi}_{n, i}) \bm{X}_n \bm{W}_V \bm{w}_O \bm{\xi}_{n, i}^\top ) \bm{W}_Q \\
        &- \frac{1}{NM} \sum\limits_{n \in S_-} \ell_n^{\prime}(\theta) ( \bm{X}_n^\top (diag(\bm{\varphi}_{n, 1}) - \bm{\varphi}_{n, 1}^\top \bm{\varphi}_{n, 1}) \bm{X}_n \bm{W}_V \bm{w}_O \bm{\mu}_-^\top \\
        &+ \sum\limits_{i=2}^M \bm{X}_n^\top (diag(\bm{\varphi}_{n, i}) - \bm{\varphi}_{n, i}^\top \bm{\varphi}_{n, i}) \bm{X}_n \bm{W}_V \bm{w}_O \bm{\xi}_{n, i}^\top ) \bm{W}_Q ,
    \end{split}
    \end{equation*}

    \begin{equation*}
    \begin{split}
        \nabla_{\bm{W}_V}L_S(\theta) = \frac{1}{NM} \sum\limits_{n=1}^{N} y_n \ell_n^{\prime}(\theta) [\bm{w}_O 
        \sum\limits_{l=1}^M  \bm{\varphi}(\bm{x}_{n, l} \bm{W}_Q \bm{W}_K^\top \bm{X}_n^\top) \bm{X}_n ]^{\top}
    \end{split}
    \end{equation*}

\end{lemma}

\it{Proof of Lemma} \ref{A1}. \rm
Considering that in the vector \( \bm{x}_{n, i}^\top \bm{W}_{Q} \bm{W}_{K}^{\top} \bm{X}_n^\top \), the \(j\)-th element of \( \bm{x}_{n, i}^\top \), i.e. \( (\bm{x}_{n, i})_{[j]} \), is only used to multiply \( \bm{W}_{Q[j, :]} \), then we have
\[
    \frac{\partial (\bm{x}_{n, i}^\top \bm{W}_{Q} \bm{W}_{K}^{\top} \bm{X}_n^\top)}{\partial \bm{W}_{Q[j, :]}} = (\bm{x}_{n, i})_{[j]} \cdot (\bm{X}_n \bm{W}_{K})
\]
And then we get
\begin{equation*}
\begin{split}
    &\nabla_{\bm{W}_{Q[j,:]}} \big( \bm{\varphi} ( \bm{x}_{n, i}^\top \bm{W}_{Q} \bm{W}_{K}^{\top} \bm{X}_n^\top ) \bm{X}_n \bm{W}_V \bm{w}_O \big) \\
    &= (\bm{x}_{n, i})_{[j]} \big[ \frac{ \partial ( \bm{\varphi}_{n, i} \bm{X}_n \bm{W}_V \bm{w}_O ) }{\partial ( \bm{x}_{n, i}^\top \bm{W}_{Q} \bm{W}_{K}^{\top} \bm{X}_n^\top ) } \big] (\bm{X}_n \bm{W}_{K})
\end{split}
\end{equation*}
Considering all \(j \in [d]\), we get

\begin{equation}
\begin{split}
    \label{Q_1}
    &\nabla_{\bm{W}_{Q}} \big( \bm{\varphi} ( \bm{x}_{n, i}^\top \bm{W}_{Q} \bm{W}_{K}^{\top} \bm{X}_n^\top ) \bm{X}_n \bm{W}_V \bm{w}_O \big) \\
    &= \bm{x}_{n, i} \big[ \frac{ \partial ( \bm{\varphi}_{n, i} \bm{X}_n \bm{W}_V \bm{w}_O ) }{\partial ( \bm{x}_{n, i}^\top \bm{W}_{Q} \bm{W}_{K}^{\top} \bm{X}_n^\top ) } \big] \bm{X}_n \bm{W}_K
\end{split}
\end{equation}

\begin{equation}
\begin{split}
    \label{GD_Q}
    \nabla_{\bm{W}_Q}L_S(\theta)
    &= \frac{1}{N} \sum\limits_{n=1}^{N} y_n \ell^{\prime} (y_n f(\bm{X}_n, \theta)) \nabla_{\bm{W}_Q}f(\bm{X}_n, \theta) \\
    &= \frac{1}{NM} \sum\limits_{n=1}^{N} y_n \ell_n^{\prime}(\theta) \sum\limits_{i=1}^M \nabla_{\bm{W}_Q} \big( \bm{\varphi} ( \bm{x}_{n, i}^\top \bm{W}_{Q} \bm{W}_{K}^{\top} \bm{X}_n^\top ) \bm{X}_n \bm{W}_V \bm{w}_O \big)\\
    &= \frac{1}{NM} \sum\limits_{n=1}^{N} y_n \ell_n^{\prime}(\theta) \sum\limits_{i=1}^M \bm{x}_{n, i} \big[ \frac{ \partial ( \bm{\varphi}_{n, i} \bm{X}_n \bm{W}_V \bm{w}_O ) }{\partial ( \bm{x}_{n, i}^\top \bm{W}_{Q} \bm{W}_{K}^{\top} \bm{X}_n^\top ) } \big] \bm{X}_n \bm{W}_K \\
    &= \frac{1}{NM} \sum\limits_{n=1}^{N} y_n \ell_n^{\prime}(\theta) \sum\limits_{i=1}^M \bm{x}_{n, i} \big[ (\bm{X}_n \bm{W}_V \bm{w}_O)^\top \frac{\partial \bm{\varphi}_{n, i}}{\partial ( \bm{x}_{n, i}^\top \bm{W}_{Q} \bm{W}_{K}^{\top} \bm{X}_n^\top ) } \big] \bm{X}_n \bm{W}_K \\
    &= \frac{1}{NM} \sum\limits_{n=1}^{N} y_n \ell_n^{\prime}(\theta) \sum\limits_{i=1}^M \bm{x}_{n, i} \bm{w}_O^\top \bm{W}_V^\top \bm{X}_n^\top (diag(\bm{\varphi}_{n, i}) - \bm{\varphi}_{n, i}^\top \bm{\varphi}_{n, i}) \bm{X}_n \bm{W}_K \\
    &= \frac{1}{NM} \sum\limits_{n \in S_+} \ell_n^{\prime}(\theta) ( \bm{\mu}_+ \bm{w}_O^\top \bm{W}_V^\top \bm{X}_n^\top (diag(\bm{\varphi}_{n, 1}) - \bm{\varphi}_{n, 1}^\top \bm{\varphi}_{n, 1}) \\
    &+ \sum\limits_{i=2}^M \bm{\xi}_{n, i} \bm{w}_O^\top \bm{W}_V^\top \bm{X}_n^\top (diag(\bm{\varphi}_{n, i}) - \bm{\varphi}_{n, i}^\top \bm{\varphi}_{n, i}) ) \bm{X}_n \bm{W}_K \\
    &- \frac{1}{NM} \sum\limits_{n \in S_-} \ell_n^{\prime}(\theta) ( \bm{\mu}_- \bm{w}_O^\top \bm{W}_V^\top \bm{X}_n^\top (diag(\bm{\varphi}_{n, 1}) - \bm{\varphi}_{n, 1}^\top \bm{\varphi}_{n, 1}) \\
    &+ \sum\limits_{i=2}^M \bm{\xi}_{n, i} \bm{w}_O^\top \bm{W}_V^\top \bm{X}_n^\top (diag(\bm{\varphi}_{n, i}) - \bm{\varphi}_{n, i}^\top \bm{\varphi}_{n, i}) ) \bm{X}_n \bm{W}_K \\
\end{split}
\end{equation}
where the third equality is by \eqref{Q_1}, the fourth equality is by chain rule, the fifth equality is by \eqref{softmax_3}, For the last equality, we expand the equality by materializing all the \( \bm{x}_{n, i} \) (e.g., \( \bm{x}_{n, 1} = \bm{\mu}_+ \ for \ n \in S_+  \)).

Using the similar method, we obtain the gradient of \( \bm{W}_K \).
\begin{equation}
\begin{split}
    \label{GD_K}
    \nabla_{\bm{W}_K}L_S(\theta)
    &= \frac{1}{N} \sum\limits_{n=1}^{N} y_n \ell^{\prime} (y_n f(\bm{X}_n, \theta)) \nabla_{\bm{W}_K}f(\bm{X}_n, \theta) \\
    &= \frac{1}{NM} \sum\limits_{n=1}^{N} y_n \ell_n^{\prime}(\theta) \sum\limits_{i=1}^M \nabla_{\bm{W}_K} \big( \bm{\varphi} ( \bm{x}_{n, i}^\top \bm{W}_{Q} \bm{W}_{K}^{\top} \bm{X}_n^\top ) \bm{X}_n \bm{W}_V \bm{w}_O \big) \\
    &= \frac{1}{NM} \sum\limits_{n=1}^{N} y_n \ell_n^{\prime}(\theta) \sum\limits_{i=1}^M \bm{X}_n^\top \big[ \frac{ \partial ( \bm{\varphi}_{n, i} \bm{X}_n \bm{W}_V \bm{w}_O ) }{\partial ( \bm{x}_{n, i}^\top \bm{W}_{Q} \bm{W}_{K}^{\top} \bm{X}_n^\top ) } \big]^\top \bm{x}_{n, i}^\top \bm{W}_Q \\
    &= \frac{1}{NM} \sum\limits_{n=1}^{N} y_n \ell_n^{\prime}(\theta) \sum\limits_{i=1}^M \bm{X}_n^\top \big[ \frac{\partial \bm{\varphi}_{n, i}}{\partial ( \bm{x}_{n, i}^\top \bm{W}_{Q} \bm{W}_{K}^{\top} \bm{X}_n^\top ) } (\bm{X}_n \bm{W}_V \bm{w}_O) \big] \bm{x}_{n, i}^\top \bm{W}_Q \\
    &= \frac{1}{NM} \sum\limits_{n \in S_+} \ell_n^{\prime}(\theta) ( \bm{X}_n^\top (diag(\bm{\varphi}_{n, 1}) - \bm{\varphi}_{n, 1}^\top \bm{\varphi}_{n, 1}) \bm{X}_n \bm{W}_V \bm{w}_O \bm{\mu}_+^\top \\
    &+ \sum\limits_{i=2}^M \bm{X}_n^\top (diag(\bm{\varphi}_{n, i}) - \bm{\varphi}_{n, i}^\top \bm{\varphi}_{n, i}) \bm{X}_n \bm{W}_V \bm{w}_O \bm{\xi}_{n, i}^\top ) \bm{W}_Q \\
    &- \frac{1}{NM} \sum\limits_{n \in S_-} \ell_n^{\prime}(\theta) ( \bm{X}_n^\top (diag(\bm{\varphi}_{n, 1}) - \bm{\varphi}_{n, 1}^\top \bm{\varphi}_{n, 1}) \bm{X}_n \bm{W}_V \bm{w}_O \bm{\mu}_-^\top \\
    &+ \sum\limits_{i=2}^M \bm{X}_n^\top (diag(\bm{\varphi}_{n, i}) - \bm{\varphi}_{n, i}^\top \bm{\varphi}_{n, i}) \bm{X}_n \bm{W}_V \bm{w}_O \bm{\xi}_{n, i}^\top ) \bm{W}_Q
\end{split}
\end{equation}

The gradient of \( \bm{W}_V \) can be obtained using the chain rule as follows
\begin{equation}
\begin{split}
    \label{nabla_V}
    \nabla_{\bm{W}_V}L_S(\theta) 
    &= \frac{1}{N} \sum\limits_{n=1}^{N} y_n \ell^{\prime} (y_n f(\bm{X}_n, \theta)) \nabla_{\bm{W}_V}f(\bm{X}_n, \theta) \\
    &= \frac{1}{NM} \sum\limits_{n=1}^{N} y_n \ell_n^{\prime}(\theta) [\bm{w}_O 
    \sum\limits_{l=1}^M  \bm{\varphi}(\bm{x}_{n, l} \bm{W}_Q \bm{W}_K^\top (\bm{X}_n)^\top) \bm{X}_n ]^{\top}
\end{split}
\end{equation}
\subsection{Update Rules}

\begin{definition}[Scalarized V]
    \label{def1}
    Let \(\bm{W}_V^{(t)}\) be the V matrix of the ViT at the t-th iteration of gradient descent. Then there exist coefficients \( \gamma_{V, +}^{(t)} \), \( \gamma_{V, -}^{(t)} \), \( \rho_{V, n, i}^{(t)} \) such that
    \[
        \bm{\mu}_+^\top \bm{W}_V^{(t)} \bm{w}_O = \bm{\mu}_+^\top \bm{W}_V^{(0)} \bm{w}_O + \gamma_{V, +}^{(t)} \Vert \bm{w}_O \Vert_2^2,
    \]
    \[
        \bm{\mu}_-^\top \bm{W}_V^{(t)} \bm{w}_O = \bm{\mu}_-^\top \bm{W}_V^{(0)} \bm{w}_O + \gamma_{V, -}^{(t)} \Vert \bm{w}_O \Vert_2^2,
    \]
    \[
        \bm{\xi}_{n, i}^\top \bm{W}_V^{(t)} \bm{w}_O = \bm{\xi}_{n, i}^\top \bm{W}_V^{(0)} \bm{w}_O + \rho_{V, n, i}^{(t)} \Vert \bm{w}_O \Vert_2^2
    \]
    for \( i \in [M] \backslash \{1\}, n \in [N] \).
    We further denote \( V_+^{(t)} := \bm{\mu}_+^\top \bm{W}_V^{(t)} \bm{w}_O \), \( V_-^{(t)} := \bm{\mu}_-^\top \bm{W}_V^{(t)} \bm{w}_O \) and \( V_{n, i}^{(t)} := \bm{\xi}_{n, i}^\top \bm{W}_V^{(t)} \bm{w}_O \). We refer to it as scalarized V.

\end{definition}

\begin{definition}[Vectorized Q \& K]
    \label{def2}
    Let \(\bm{W}_Q^{(t)}\) and \(\bm{W}_K^{(t)}\) be the QK matrices of the ViT at the t-th iteration of gradient descent. Then we define the vectorized Q and vectorized K as follows
    \[ \bm{q}_+^{(t)} = \bm{\mu}_+^\top \bm{W}_Q^{(t)},  \qquad \bm{q}_-^{(t)} = \bm{\mu}_-^\top \bm{W}_Q^{(t)}, \qquad \bm{q}_{n, i}^{(t)} = \bm{\xi}_{n, i}^\top \bm{W}_Q^{(t)}, \]
    \[ \bm{k}_+^{(t)} = \bm{\mu}_+^\top \bm{W}_K^{(t)},  \qquad \bm{k}_-^{(t)} = \bm{\mu}_-^\top \bm{W}_K^{(t)}, \qquad \bm{k}_{n, i}^{(t)} = \bm{\xi}_{n, i}^\top \bm{W}_K^{(t)} \]
    for \( i \in [M] \backslash \{1\}, n \in [N] \).
\end{definition} 

\begin{definition}[Gradient Decomposition]
    \label{def3}
    There exist coefficients \( \alpha_{+, +}^{(t)} \), \( \alpha_{n, +, i}^{(t)} \), \( \alpha_{-, -}^{(t)} \), \( \alpha_{n, -, i}^{(t)} \), \( \alpha_{n, i, +}^{(t)} \), \( \alpha_{n, i, -}^{(t)} \), \( \alpha_{n, i, n^\prime, i^\prime}^{(t)} \), 
    \( \beta_{+, +}^{(t)} \), \( \beta_{n, +, i}^{(t)} \), \( \beta_{-, -}^{(t)} \), \( \beta_{n, -, i}^{(t)} \), \( \beta_{n, i, +}^{(t)} \), \( \beta_{n, i, -}^{(t)} \), \( \beta_{n, i, n^\prime, i^\prime}^{(t)} \) such that
    \[
        \Delta \bm{q}_+^{(t)} := \bm{q}_+^{(t + 1)} - \bm{q}_+^{(t)} = \alpha_{+, +}^{(t)} \bm{k}_+^{(t)} + \sum\limits_{n \in S_+} \sum\limits_{i=2}^M \alpha_{n, +, i}^{(t)} \bm{k}_{n, i}^{(t)},  
    \]
    \[
        \Delta \bm{q}_-^{(t)} := \bm{q}_-^{(t + 1)} - \bm{q}_-^{(t)} = \alpha_{-, -}^{(t)} \bm{k}_-^{(t)} + \sum\limits_{n \in S_-} \sum\limits_{i=2}^M \alpha_{n, -, i}^{(t)} \bm{k}_{n, i}^{(t)},  
    \]
    \[
        \Delta \bm{q}_{n, i}^{(t)} := \bm{q}_{n, i}^{(t + 1)} - \bm{q}_{n, i}^{(t)} = \alpha_{n, i, +}^{(t)} \bm{k}_+^{(t)} + \alpha_{n, i, -}^{(t)} \bm{k}_-^{(t)} + \sum\limits_{n^\prime=1}^N \sum\limits_{i^\prime=2}^M \alpha_{n, i, n^\prime, i^\prime}^{(t)} \bm{k}_{n^\prime, i^\prime}^{(t)},  
    \]
    \[
        \Delta \bm{k}_+^{(t)} := \bm{k}_+^{(t + 1)} - \bm{k}_+^{(t)} = \beta_{+, +}^{(t)} \bm{q}_+^{(t)} + \sum\limits_{n \in S_+} \sum\limits_{i=2}^M \beta_{n, +, i}^{(t)} \bm{q}_{n, i}^{(t)},  
    \]
    \[
        \Delta \bm{k}_-^{(t)} := \bm{k}_-^{(t + 1)} - \bm{k}_-^{(t)} = \beta_{-, -}^{(t)} \bm{q}_-^{(t)} + \sum\limits_{n \in S_-} \sum\limits_{i=2}^M \beta_{n, -, i}^{(t)} \bm{q}_{n, i}^{(t)},  
    \]
    \[
        \Delta \bm{k}_{n, i}^{(t)} := \bm{k}_{n, i}^{(t + 1)} - \bm{k}_{n, i}^{(t)} = \beta_{n, i, +}^{(t)} \bm{q}_+^{(t)} + \beta_{n, i, -}^{(t)} \bm{q}_-^{(t)} + \sum\limits_{n^\prime=1}^N \sum\limits_{i^\prime=2}^M \beta_{n, i, n^\prime, i^\prime}^{(t)} \bm{q}_{n^\prime, i^\prime}^{(t)}  
    \]
    for \( i, i^\prime \in [M] \backslash \{1\} \) and \( n, n^\prime \in [N] \).

\end{definition}
\textbf{Remark.} With the scalarized V in \ref{def1}, we can portray the learning process of \( \bm{W}_V \) on signals and noises by analyzing the dynamics of \( \gamma_{V, +}^{(t)} \), \( \gamma_{V, -}^{(t)} \) and \( \rho_{V, n, i}^{(t)} \) (or \( V_+^{(t)} \), \( V_-^{(t)} \) and \( V_{n, i}^{(t)} \)) . Also, as scalars, the effect of \( \bm{W}_V \) on the learning process of \( \bm{W}_Q \) and \( \bm{W}_K \) is more conveniently to describe. 
With the Vectorized Q \& K in \ref{def2}, we can decompose \( \bm{x}^\top \bm{W}_Q \bm{W}_K \bm{X}^\top \) into the inner product of vectors \( \bm{q} \) (a generalized reference to \(\bm{q}_+\), \(\bm{q}_-\) and \(\bm{q}_{n, i}\)) and \( \bm{k} \) (a generalized reference to \(\bm{k}_+\), \(\bm{k}_-\) and \(\bm{k}_{n, i}\)), e.g., \( \langle \bm{q}_+, \bm{k}_+ \rangle = \bm{\mu}_+^\top \bm{W}_Q \bm{W}_K \bm{\mu}_+ \). Then the dynamics of \( \bm{x}^\top \bm{W}_Q \bm{W}_K \bm{X}^\top \) can be characterized by analyzing the dynamics of the inner products of \( \bm{q} \) and \( \bm{k} \), and then further get the dynamics of attentions.
The gradient decomposition in \ref{def3} comes from \eqref{GD_Q} and \eqref{GD_K}. Note that \eqref{GD_Q} ends in \( \bm{X}_n \bm{W}_K \) and \eqref{GD_K} ends in \( \bm{x}_{n, i}^\top \bm{W}_Q \). This suggests that the dynamics of \( \bm{q}_\pm^{(t)} \) and \( \bm{q}_{n, i}^{(t)} \) can be decomposed into linear combinations of \( \bm{k}_\pm^{(t)} \) and \( \bm{k}_{n, i}^{(t)} \), and similarly, the gradient of \( \bm{k}_\pm^{(t)} \) and \( \bm{k}_{n, i}^{(t)} \) can be decomposed into linear combinations of \( \bm{q}_\pm^{(t)} \) and \( \bm{q}_{n, i}^{(t)} \). 
We calculate \(\alpha\) (a generalized reference to \(\alpha_{+, +}\), \(\alpha_+^{n, i}\), \(\alpha_{-, -}\), etc.) and \(\beta\) (a generalized reference to \(\beta_{+, +}\), \(\beta_{n, +, i}\), \(\beta_{-, -}\), etc.) in detail in Appendix \ref{specific_calculation_process}.

\begin{lemma}[Update Rule for V]
    \label{update_rule4V}
    The coefficients \( \gamma_{V, +}^{(t)} \), \( \gamma_{V, -}^{(t)} \), \( \rho_{V, n, i}^{(t)} \) defined in Definition \ref{def1} satisfy the following iterative equations:
    \begin{equation*}
    \begin{split}
        & \gamma_{V, +}^{(t + 1)} = \gamma_{V, +}^{(t)} \\
        &- \frac{\eta \Vert \bm{\mu} \Vert_2^2}{NM} \sum\limits_{n \in S_+} \ell_n^{\prime(t)} \big( \frac{ \exp ( \langle \bm{q}_+^{(t)}, \bm{k}_+^{(t)} \rangle )}{  \exp ( \langle \bm{q}_+^{(t)}, \bm{k}_+^{(t)} \rangle ) + \sum\limits_{k=2}^M  \exp ( \langle \bm{q}_+^{(t)}, \bm{k}_{n, k}^{(t)} \rangle )}  \\
        &+ \sum\limits_{j=2}^M \frac{ \exp ( \langle \bm{q}_{n, j}^{(t)}, \bm{k}_+^{(t)} \rangle )}{  \exp ( \langle \bm{q}_{n, j}^{(t)}, \bm{k}_+^{(t)} \rangle ) + \sum\limits_{k=2}^M  \exp ( \langle \bm{q}_{n, j}^{(t)}, \bm{k}_{n, k}^{(t)} \rangle )} \big),
    \end{split}
    \end{equation*}
    \begin{equation*}
    \begin{split}
        & \gamma_{V, -}^{(t + 1)} = \gamma_{V, -}^{(t)} \\
        &+ \frac{\eta \Vert \bm{\mu} \Vert_2^2}{NM} \sum\limits_{n \in S_-} \ell_n^{\prime(t)} \big( \frac{ \exp ( \langle \bm{q}_-^{(t)}, \bm{k}_-^{(t)} \rangle )}{  \exp ( \langle \bm{q}_-^{(t)}, \bm{k}_-^{(t)} \rangle ) + \sum\limits_{k=2}^M  \exp ( \langle \bm{q}_-^{(t)}, \bm{k}_{n, k}^{(t)} \rangle )}  \\
        &+ \sum\limits_{j=2}^M \frac{ \exp ( \langle \bm{q}_{n, j}^{(t)}, \bm{k}_-^{(t)} \rangle )}{  \exp ( \langle \bm{q}_{n, j}^{(t)}, \bm{k}_-^{(t)} \rangle ) + \sum\limits_{k=2}^M  \exp ( \langle \bm{q}_{n, j}^{(t)}, \bm{k}_{n, k}^{(t)} \rangle )} \big),
    \end{split}
    \end{equation*}
    \begin{equation*}
    \begin{split}
        \rho_{V, n, i}^{(t+1)} &= \rho_{V, n, i}^{(t)} \\
        &- \frac{\eta}{NM} \sum\limits_{n^\prime \in S_+} \ell_{n^\prime}^{\prime(t)} \big( 
        \sum\limits_{i^\prime=2}^M ( \langle \bm{\xi}_{n, i}, \bm{\xi}_{n^\prime, i^\prime} \rangle \frac{ \exp ( \langle \bm{q}_+^{(t)}, \bm{k}_{n^\prime, i^\prime}^{(t)} \rangle )}{  \exp ( \langle \bm{q}_+^{(t)}, \bm{k}_+^{(t)} \rangle ) + \sum\limits_{k=2}^M  \exp ( \langle \bm{q}_+^{(t)}, \bm{k}_{n^\prime, k}^{(t)} \rangle )} \\
        &+ \sum\limits_{j=2}^M \langle \bm{\xi}_{n, i}, \bm{\xi}_{n^\prime, i^\prime} \rangle \frac{ \exp ( \langle \bm{q}_{n^\prime, j}^{(t)}, \bm{k}_{n^\prime, i^\prime}^{(t)} \rangle  )}{ \exp ( \langle \bm{q}_{n^\prime, j}^{(t)}, \bm{k}_+^{(t)} \rangle ) + \sum\limits_{k=2}^M  \exp ( \langle \bm{q}_{n^\prime, j}^{(t)}, \bm{k}_{n^\prime, k}^{(t)} \rangle )} ) \big) \\
        &+ \frac{\eta}{NM} \sum\limits_{n^\prime \in S_-} \ell_{n^\prime}^{\prime(t)} \big(
        \sum\limits_{i^\prime=2}^M ( \langle \bm{\xi}_{n, i}, \bm{\xi}_{n^\prime, i^\prime} \rangle \frac{ \exp ( \langle \bm{q}_-^{(t)}, \bm{k}_{n^\prime, i^\prime}^{(t)} \rangle )}{  \exp ( \langle \bm{q}_-^{(t)}, \bm{k}_-^{(t)} \rangle ) + \sum\limits_{k=2}^M  \exp ( \langle \bm{q}_-^{(t)}, \bm{k}_{n^\prime, k}^{(t)} \rangle )} \\
        &+ \sum\limits_{j=2}^M \langle \bm{\xi}_{n, i}, \bm{\xi}_{n^\prime, i^\prime} \rangle \frac{ \exp ( \langle \bm{q}_{n^\prime, j}^{(t)}, \bm{k}_{n^\prime, i^\prime}^{(t)} \rangle  )}{ \exp ( \langle \bm{q}_{n^\prime, j}^{(t)}, \bm{k}_-^{(t)} \rangle ) + \sum\limits_{k=2}^M  \exp ( \langle \bm{q}_{n^\prime, j}^{(t)}, \bm{k}_{n^\prime, k}^{(t)} \rangle )} ) \big) \\
    \end{split}
    \end{equation*}
    for \(i \in [M] \backslash \{1\}, n \in [N]\).
\end{lemma}
\it{Proof of Lemma} \ref{update_rule4V}. \rm Base on \eqref{nabla_V}, we have

\begin{equation}
\begin{split}
    \bm{x}^\top \nabla_{\bm{W}_V} L_S(\theta) \bm{w}_O &= \bm{x}^\top \frac{1}{NM} \sum\limits_{n=1}^{N} y_n \ell_n^{\prime}(\theta) [\bm{w}_O \sum\limits_{l=1}^M  \bm{\varphi}(\bm{x}_{n, l} \bm{W}_Q \bm{W}_K^\top (\bm{X}_n)^\top) \bm{X}_n ]^{\top} \bm{w}_O \\
    &= \frac{1}{NM} \sum\limits_{n=1}^N y_n \ell_n^\prime (\theta) \sum\limits_{l=1}^M \bm{x}^\top \bm{X}_n^\top \bm{\varphi}(\bm{x}_{n, l} \bm{W}_Q \bm{W}_K^\top (\bm{X}_n)^\top)^\top \Vert \bm{w}_O \Vert_2^2 \\
    &= \frac{1}{NM} \sum\limits_{n \in S_+} \ell_n^\prime (\theta) \sum\limits_{l=1}^M \big( \langle \bm{x}, \bm{\mu}_+ \rangle \frac{ \exp (\bm{x}_{n, l}^\top \bm{W}_Q \bm{W}_K^{\top} \bm{\mu}_+)}{ \exp  (\bm{x}_{n, l}^\top \bm{W}_Q \bm{W}_K^{\top} \bm{\mu}_+) + \sum\limits_{j=2}^M  \exp  ( \bm{x}_{n, l}^\top \bm{W}_Q \bm{W}_K^{\top} \bm{\xi}_{n, j} ) } \\
    &+ \sum\limits_{i=2}^M ( \langle \bm{x}, \bm{\xi}_{n, i} \rangle \frac{ \exp  ( \bm{x}_{n, l}^\top \bm{W}_Q \bm{W}_K^{\top} \bm{\xi}_{n, i} )}{ \exp  (\bm{x}_{n, l}^\top \bm{W}_Q \bm{W}_K^{\top} \bm{\mu}_+) + \sum\limits_{j=2}^M  \exp  ( \bm{x}_{n, l}^\top \bm{W}_Q \bm{W}_K^{\top} \bm{\xi}_{n, j} ) } ) \big) \Vert \bm{w}_O \Vert_2^2 \\
    &- \frac{1}{NM} \sum\limits_{n \in S_-} \ell_n^\prime (\theta) \sum\limits_{l=1}^M \big( \langle \bm{x}, \bm{\mu}_- \rangle \frac{ \exp (\bm{x}_{n, l}^\top \bm{W}_Q \bm{W}_K^{\top} \bm{\mu}_-)}{ \exp  (\bm{x}_{n, l}^\top \bm{W}_Q \bm{W}_K^{\top} \bm{\mu}_- ) + \sum\limits_{j=2}^M  \exp  ( \bm{x}_{n, l}^\top \bm{W}_Q \bm{W}_K^{\top} \bm{\xi}_{n, j} ) } \\
    &+ \sum\limits_{i=2}^M ( \langle \bm{x}, \bm{\xi}_{n, i} \rangle \frac{ \exp  ( \bm{x}_{n, l}^\top \bm{W}_Q \bm{W}_K^{\top} \bm{\xi}_{n, i} )}{ \exp  (\bm{x}_{n, l}^\top \bm{W}_Q \bm{W}_K^{\top} \bm{\mu}_- ) + \sum\limits_{j=2}^M  \exp  ( \bm{x}_{n, l}^\top \bm{W}_Q \bm{W}_K^{\top} \bm{\xi}_{n, j} ) } ) \big) \Vert \bm{w}_O \Vert_2^2 \\
    &= \frac{1}{NM} \sum\limits_{n \in S_+} \ell_n^\prime (\theta) \big( ( \langle \bm{x}, \bm{\mu}_+ \rangle \frac{ \exp ( \bm{\mu}_+^\top \bm{W}_Q \bm{W}_K^{\top} \bm{\mu}_+ )}{  \exp ( \bm{\mu}_+^\top \bm{W}_Q \bm{W}_K^{\top} \bm{\mu}_+ ) + \sum\limits_{k=2}^M  \exp ( \bm{\mu}_+^\top \bm{W}_Q \bm{W}_K^{\top} \bm{\xi}_{n, k} )}  \\
    &+ \sum\limits_{j=2}^M \langle \bm{x}, \bm{\mu}_+ \rangle \frac{ \exp ( \bm{\xi}_{n, j}^{\top} \bm{W}_Q \bm{W}_K^{\top} \bm{\mu}_+ )}{  \exp ( \bm{\xi}_{n, j}^{\top} \bm{W}_Q \bm{W}_K^{\top} \bm{\mu}_+ ) + \sum\limits_{k=2}^M  \exp ( \bm{\xi}_{n, j}^{\top} \bm{W}_Q \bm{W}_K^{\top} \bm{\xi}_{n, k} )} ) \\
    &+ \sum\limits_{i=2}^M ( \langle \bm{x}, \bm{\xi}_{n, i} \rangle \frac{ \exp ( \bm{\mu}_+ \bm{W}_Q \bm{W}_K^{\top} \bm{\xi}_{n, i} )}{  \exp ( \bm{\mu}_+^\top \bm{W}_Q \bm{W}_K^{\top} \bm{\mu}_+ ) + \sum\limits_{k=2}^M  \exp ( \bm{\mu}_+^\top \bm{W}_Q \bm{W}_K^{\top} \bm{\xi}_{n, k} )} \\
    &+ \sum\limits_{j=2}^M \langle \bm{x}, \bm{\xi}_{n, i} \rangle \frac{ \exp ( \bm{\xi}_{n, j} \bm{W}_Q \bm{W}_K^{\top} \bm{\xi}_{n, i}  )}{  \exp ( \bm{\xi}_{n, j}^{\top} \bm{W}_Q \bm{W}_K^{\top} \bm{\mu}_+ ) + \sum\limits_{k=2}^M  \exp ( \bm{\xi}_{n, j}^{\top} \bm{W}_Q \bm{W}_K^{\top} \bm{\xi}_{n, k} )} ) \big) \Vert \bm{w}_O \Vert_2^2 \\
    &- \frac{1}{NM} \sum\limits_{n \in S_-} \ell_n^\prime (\theta) \big( ( \langle \bm{x}, \bm{\mu}_- \rangle \frac{ \exp ( \bm{\mu}_-^\top \bm{W}_Q \bm{W}_K^{\top} \bm{\mu}_- )}{  \exp ( \bm{\mu}_-^\top \bm{W}_Q \bm{W}_K^{\top} \bm{\mu}_- ) + \sum\limits_{k=2}^M  \exp ( \bm{\mu}_-^\top \bm{W}_Q \bm{W}_K^{\top} \bm{\xi}_{n, k} )}  \\
    &+ \sum\limits_{j=2}^M \langle \bm{x}, \bm{\mu}_- \rangle \frac{ \exp ( \bm{\xi}_{n, j}^{\top} \bm{W}_Q \bm{W}_K^{\top} \bm{\mu}_- )}{  \exp ( \bm{\xi}_{n, j}^{\top} \bm{W}_Q \bm{W}_K^{\top} \bm{\mu}_- ) + \sum\limits_{k=2}^M  \exp ( \bm{\xi}_{n, j}^{\top} \bm{W}_Q \bm{W}_K^{\top} \bm{\xi}_{n, k} )} ) \\
    &+ \sum\limits_{i=2}^M ( \langle \bm{x}, \bm{\xi}_{n, i} \rangle \frac{ \exp ( \bm{\mu}_- \bm{W}_Q \bm{W}_K^{\top} \bm{\xi}_{n, i} )}{  \exp ( \bm{\mu}_-^\top \bm{W}_Q \bm{W}_K^{\top} \bm{\mu}_- ) + \sum\limits_{k=2}^M  \exp ( \bm{\mu}_-^\top \bm{W}_Q \bm{W}_K^{\top} \bm{\xi}_{n, k} )} \\
    &+ \sum\limits_{j=2}^M \langle \bm{x}, \bm{\xi}_{n, i} \rangle \frac{ \exp ( \bm{\xi}_{n, j} \bm{W}_Q \bm{W}_K^{\top} \bm{\xi}_{n, i}  )}{  \exp ( \bm{\xi}_{n, j}^{\top} \bm{W}_Q \bm{W}_K^{\top} \bm{\mu}_- ) + \sum\limits_{k=2}^M  \exp ( \bm{\xi}_{n, j}^{\top} \bm{W}_Q \bm{W}_K^{\top} \bm{\xi}_{n, k} )} ) \big) \Vert \bm{w}_O \Vert_2^2
\end{split}
\end{equation}
where the second equality we expand \( \bm{X}_n \) into vectors and make inner products with \( x \), the third equality we materializing all the \( \bm{x}_{n, l} \) (e.g., \( \bm{x}_{n, 1} = \bm{\mu}_+ \) for \( n \in S_+ \)).
Note the orthogonality between \(\bm{\mu}\) and \(\bm{\xi}_{n, i}\), we can remove many of the terms in this equation. Take \( \bm{x} = \bm{\mu}_+ \) as an example, we have

\begin{equation}
\begin{split}
    &\bm{\mu}_+^\top \nabla_{\bm{W}_V} L_S(\theta) \bm{w}_O \\
    &= \frac{\Vert \bm{\mu} \Vert_2^2}{NM} \sum\limits_{n \in S_+} \ell_n^\prime (\theta) \big( \frac{ \exp ( \bm{\mu}_+^\top \bm{W}_Q \bm{W}_K^{\top} \bm{\mu}_+ )}{  \exp ( \bm{\mu}_+^\top \bm{W}_Q \bm{W}_K^{\top} \bm{\mu}_+ ) + \sum\limits_{k=2}^M  \exp ( \bm{\mu}_+^\top \bm{W}_Q \bm{W}_K^{\top} \bm{\xi}_{n, k} )}  \\
    &+ \sum\limits_{j=2}^M \frac{ \exp ( \bm{\xi}_{n, j}^{\top} \bm{W}_Q \bm{W}_K^{\top} \bm{\mu}_+ )}{  \exp ( \bm{\xi}_{n, j}^{\top} \bm{W}_Q \bm{W}_K^{\top} \bm{\mu}_+ ) + \sum\limits_{k=2}^M  \exp ( \bm{\xi}_{n, j}^{\top} \bm{W}_Q \bm{W}_K^{\top} \bm{\xi}_{n, k} )} \big) \Vert \bm{w}_O \Vert_2^2 \\
\end{split}
\end{equation}
Then we have
\begin{equation}
\begin{split}
    &\bm{\mu}_+^\top \bm{W}_V^{(t + 1)} \bm{w}_O - \bm{\mu}_+^\top \bm{W}_V^{(t)} \bm{w}_O = \bm{\mu}_+^\top \big( - \eta \nabla_{\bm{W}_V} L_S(\theta(t)) \big) \bm{w}_O \\
    &- \frac{\eta \Vert \bm{\mu} \Vert_2^2}{NM} \sum\limits_{n \in S_+} \ell_n^{\prime(t)} \big( \frac{ \exp ( \langle \bm{q}_+^{(t)}, \bm{k}_+^{(t)} \rangle )}{  \exp ( \langle \bm{q}_+^{(t)}, \bm{k}_+^{(t)} \rangle ) + \sum\limits_{k=2}^M  \exp ( \langle \bm{q}_+^{(t)}, \bm{k}_{n, k}^{(t)} \rangle )}  \\
    &+ \sum\limits_{j=2}^M \frac{ \exp ( \langle \bm{q}_{n, j}^{(t)}, \bm{k}_+^{(t)} \rangle )}{  \exp ( \langle \bm{q}_{n, j}^{(t)}, \bm{k}_+^{(t)} \rangle ) + \sum\limits_{k=2}^M  \exp ( \langle \bm{q}_{n, j}^{(t)}, \bm{k}_{n, k}^{(t)} \rangle )} \big) \Vert \bm{w}_O \Vert_2^2 \\
\end{split}
\end{equation}
Dividing by \( \Vert \bm{w}_O \Vert_2^2 \) we get
\begin{equation}
\begin{split}
    & \gamma_{V, +}^{(t + 1)} = \gamma_{V, +}^{(t)} \\
    &- \frac{\eta \Vert \bm{\mu} \Vert_2^2}{NM} \sum\limits_{n \in S_+} \ell_n^{\prime(t)} \big( \frac{ \exp ( \langle \bm{q}_+^{(t)}, \bm{k}_+^{(t)} \rangle )}{  \exp ( \langle \bm{q}_+^{(t)}, \bm{k}_+^{(t)} \rangle ) + \sum\limits_{k=2}^M  \exp ( \langle \bm{q}_+^{(t)}, \bm{k}_{n, k}^{(t)} \rangle )}  \\
    &+ \sum\limits_{j=2}^M \frac{ \exp ( \langle \bm{q}_{n, j}^{(t)}, \bm{k}_+^{(t)} \rangle )}{  \exp ( \langle \bm{q}_{n, j}^{(t)}, \bm{k}_+^{(t)} \rangle ) + \sum\limits_{k=2}^M  \exp ( \langle \bm{q}_{n, j}^{(t)}, \bm{k}_{n, k}^{(t)} \rangle )} \big) \\
\end{split}
\end{equation}
This proves the update rule for \( \gamma_{V, +}^{(t)} \), The proof for \( \gamma_{V, +}^{(t)} \) and \( \rho_{V, n, i}^{(t)} \) is similar to it.

\begin{lemma}[Update Rule for QK]
    \label{update_rule4QK}
    The dynamics of \( \bm{x}^\top \bm{W}_Q \bm{W}_K \bm{x} \) can be characterized as follows

    \begin{equation*}
    \begin{split}
        &\langle \bm{q}_+^{(t+1)}, \bm{k}_+^{(t+1)} \rangle - \langle \bm{q}_+^{(t)}, \bm{k}_+^{(t)} \rangle \\
        &= \alpha_{+, +}^{(t)} \Vert \bm{k}_+^{(t)} \Vert_2^2 + \sum\limits_{n \in S_+} \sum\limits_{i=2}^M \alpha_{n, +, i}^{(t)} \langle \bm{k}_+^{(t)}, \bm{k}_{n, i}^{(t)} \rangle \\
        &+ \beta_{+, +}^{(t)} \Vert \bm{q}_+^{(t)} \Vert_2^2 + \sum\limits_{n \in S_+} \sum\limits_{i=2}^M \beta_{n, +, i}^{(t)} \langle \bm{q}_+^{(t)}, \bm{q}_{n, i}^{(t)} \rangle \\
        &+ \Big( \alpha_{+, +}^{(t)} \bm{k}_+^{(t)} + \sum\limits_{n \in S_+} \sum\limits_{i=2}^M \alpha_{n, +, i}^{(t)} \bm{k}_{n, i}^{(t)} \Big) \\
        &\cdot \Big( \beta_{+, +}^{(t)} \bm{q}_+^{(t)\top} + \sum\limits_{n \in S_+} \sum\limits_{i=2}^M \beta_{n, +, i}^{(t)} \bm{q}_{n, i}^{(t)\top} \Big),
    \end{split}
    \end{equation*}

    \begin{equation*}
    \begin{split}
        &\langle \bm{q}_-^{(t+1)}, \bm{k}_-^{(t+1)} \rangle - \langle \bm{q}_-^{(t)}, \bm{k}_-^{(t)} \rangle \\
        &= \alpha_{-, -}^{(t)} \Vert \bm{k}_-^{(t)} \Vert_2^2 + \sum\limits_{n \in S_-} \sum\limits_{i=2}^M \alpha_{n, -, i}^{(t)} \langle \bm{k}_-^{(t)}, \bm{k}_{n, i}^{(t)} \rangle \\
        &+ \beta_{-, -}^{(t)} \Vert \bm{q}_-^{(t)} \Vert_2^2 + \sum\limits_{n \in S_-} \sum\limits_{i=2}^M \beta_{n, -, i}^{(t)} \langle \bm{q}_-^{(t)}, \bm{q}_{n, i}^{(t)} \rangle \\
        &+ \Big( \alpha_{-, -}^{(t)} \bm{k}_-^{(t)} + \sum\limits_{n \in S_-} \sum\limits_{i=2}^M \alpha_{n, -, i}^{(t)} \bm{k}_{n, i}^{(t)} \Big) \\
        &\cdot \Big( \beta_{-, -}^{(t)} \bm{q}_-^{(t)\top} + \sum\limits_{n \in S_-} \sum\limits_{i=2}^M \beta_{n, -, i}^{(t)} \bm{q}_{n, i}^{(t)\top} \Big),
    \end{split}
    \end{equation*}
    
    \begin{equation*}
    \begin{split}
        &\langle \bm{q}_{n, i}^{(t+1)}, \bm{k}_+^{(t+1)} \rangle - \langle \bm{q}_{n, i}^{(t)}, \bm{k}_+^{(t)} \rangle \\
        &= \alpha_{n, i, +}^{(t)} \Vert \bm{k}_+^{(t)} \Vert_2^2 + \alpha_{n, i, -}^{(t)} \langle \bm{k}_+^{(t)}, \bm{k}_-^{(t)} \rangle + \sum\limits_{n^\prime =1}^N \sum\limits_{l=2}^M \alpha_{n, i, n^\prime, l}^{(t)} \langle \bm{k}_+^{(t)}, \bm{k}_{n^\prime, l}^{(t)} \rangle \\
        &+ \beta_{+, +}^{(t)} \langle \bm{q}_+^{(t)}, \bm{q}_{n, i}^{(t)} \rangle + \sum\limits_{n^\prime \in S_+} \sum\limits_{l=2}^M \beta_{n^\prime, +, l}^{(t)} \langle \bm{q}_{n, i}^{(t)}, \bm{q}_{n^\prime, l}^{(t)} \rangle \\
        &+ \Big( \alpha_{n, i, +}^{(t)} \bm{k}_+^{(t)} + \alpha_{n, i, -}^{(t)} \bm{k}_-^{(t)} + \sum\limits_{n^\prime =1}^N \sum\limits_{l=2}^M \alpha_{n, i, n^\prime, l}^{(t)} \bm{k}_{n^\prime, l}^{(t)} \Big) \\
        &\cdot \Big( \beta_{+, +}^{(t)} \bm{q}_+^{(t)\top} + \sum\limits_{n^\prime \in S_+} \sum\limits_{l=2}^M \beta_{n^\prime, +, l}^{(t)} \bm{q}_{n^\prime, l}^{(t)\top} \Big),
    \end{split}
    \end{equation*}

    \begin{equation*}
    \begin{split}
        &\langle \bm{q}_{n, i}^{(t+1)}, \bm{k}_-^{(t+1)} \rangle - \langle \bm{q}_{n, i}^{(t)}, \bm{k}_-^{(t)} \rangle \\
        &= \alpha_{n, i, -}^{(t)} \Vert \bm{k}_-^{(t)} \Vert_2^2 + \alpha_{n, i, +}^{(t)} \langle \bm{k}_+^{(t)}, \bm{k}_-^{(t)} \rangle + \sum\limits_{n^\prime =1}^N \sum\limits_{l=2}^M \alpha_{n, i, n^\prime, l}^{(t)} \langle \bm{k}_-^{(t)}, \bm{k}_{n^\prime, l}^{(t)} \rangle \\
        &+ \beta_{-, -}^{(t)} \langle \bm{q}_-^{(t)}, \bm{q}_{n, i}^{(t)} \rangle + \sum\limits_{n^\prime \in S_-} \sum\limits_{l=2}^M \beta_{n^\prime, -, l}^{(t)} \langle \bm{q}_{n, i}^{(t)}, \bm{q}_{n^\prime, l}^{(t)} \rangle \\
        &+ \Big( \alpha_{n, i, +}^{(t)} \bm{k}_+^{(t)} + \alpha_{n, i, -}^{(t)} \bm{k}_-^{(t)} + \sum\limits_{n^\prime =1}^N \sum\limits_{l=2}^M \alpha_{n, i, n^\prime, l}^{(t)} \bm{k}_{n^\prime, l}^{(t)} \Big) \\
        &\cdot \Big( \beta_{-, -}^{(t)} \bm{q}_-^{(t)\top} + \sum\limits_{n^\prime \in S_-} \sum\limits_{l=2}^M \beta_{n^\prime, -, l}^{(t)} \bm{q}_{n^\prime, l}^{(t)\top} \Big),
    \end{split}
    \end{equation*}
    
    \begin{equation*}
    \begin{split}
        &\langle \bm{q}_+^{(t+1)}, \bm{k}_{n, j}^{(t+1)} \rangle - \langle \bm{q}_+^{(t)}, \bm{k}_{n, j}^{(t)} \rangle \\
        &= \alpha_{+, +}^{(t)} \langle \bm{k}_+^{(t)}, \bm{k}_{n, j}^{(t)} \rangle + \sum\limits_{n^\prime \in S_+} \sum\limits_{l=2}^M \alpha_{n^\prime, +, l}^{(t)} \langle \bm{k}_{n, j}^{(t)}, \bm{k}_{n^\prime, l}^{(t)} \rangle \\
        &+ \beta_{n, j, +}^{(t)} \Vert \bm{q}_+^{(t)} \Vert_2^2 + \beta_{n, j, -}^{(t)} \langle \bm{q}_+^{(t)}, \bm{q}_-^{(t)} \rangle + \sum\limits_{n^\prime = 1}^N \sum\limits_{l=2}^M \beta_{n, j, n^\prime, l}^{(t)} \langle \bm{q}_+^{(t)}, \bm{q}_{n^\prime, l}^{(t)} \rangle \\
        &+ \Big( \alpha_{+, +}^{(t)} \bm{k}_+^{(t)} + \sum\limits_{n^\prime \in S_+} \sum\limits_{l=2}^M \alpha_{n^\prime, +, l}^{(t)} \bm{k}_{n^\prime, l}^{(t)} \Big) \\
        &\cdot \Big( \beta_{n, j, +}^{(t)} \bm{q}_+^{(t)\top} + \beta_{n, j, -}^{(t)} \bm{q}_-^{(t)\top} + \sum\limits_{n^\prime = 1}^N \sum\limits_{l=2}^M \beta_{n, j, n^\prime, l}^{(t)} \bm{q}_{n^\prime, l}^{(t)\top} \Big),
    \end{split}
    \end{equation*}

    \begin{equation*}
    \begin{split}
        &\langle \bm{q}_-^{(t+1)}, \bm{k}_{n, j}^{(t+1)} \rangle - \langle \bm{q}_-^{(t)}, \bm{k}_{n, j}^{(t)} \rangle \\
        &= \alpha_{-, -}^{(t)} \langle \bm{k}_-^{(t)}, \bm{k}_{n, j}^{(t)} \rangle + \sum\limits_{n^\prime \in S_-} \sum\limits_{l=2}^M \alpha_{n^\prime, -, l}^{(t)} \langle \bm{k}_{n, j}^{(t)}, \bm{k}_{n^\prime, l}^{(t)} \rangle \\
        &+ \beta_{n, j, -}^{(t)} \Vert \bm{q}_-^{(t)} \Vert_2^2 + \beta_{n, j, +}^{(t)} \langle \bm{q}_+^{(t)}, \bm{q}_-^{(t)} \rangle + \sum\limits_{n^\prime = 1}^N \sum\limits_{l=2}^M \beta_{n, j, n^\prime, l}^{(t)} \langle \bm{q}_-^{(t)}, \bm{q}_{n^\prime, l}^{(t)} \rangle \\
        &+ \Big( \alpha_{-, -}^{(t)} \bm{k}_-^{(t)} + \sum\limits_{n^\prime \in S_-} \sum\limits_{l=2}^M \alpha_{n^\prime, -, l}^{(t)} \bm{k}_{n^\prime, l}^{(t)} \Big) \\
        &\cdot \Big( \beta_{n, j, +}^{(t)} \bm{q}_+^{(t)\top} + \beta_{n, j, -}^{(t)} \bm{q}_-^{(t)\top} + \sum\limits_{n^\prime = 1}^N \sum\limits_{l=2}^M \beta_{n, j, n^\prime, l}^{(t)} \bm{q}_{n^\prime, l}^{(t)\top} \Big),
    \end{split}
    \end{equation*}
    
    \begin{equation*}
    \begin{split}
        &\langle \bm{q}_{n, i}^{(t+1)}, \bm{k}_{n, j}^{(t+1)} \rangle - \langle \bm{q}_{n, i}^{(t)}, \bm{k}_{n, j}^{(t)} \rangle \\
        &= \alpha_{n, i, +}^{(t)} \langle \bm{k}_+^{(t)}, \bm{k}_{n, j}^{(t)} \rangle + \alpha_{n, i, -}^{(t)} \langle \bm{k}_-^{(t)}, \bm{k}_{n, j}^{(t)} \rangle + \sum\limits_{n^\prime =1}^N \sum\limits_{l=2}^M \alpha_{n, i, n^\prime, l}^{(t)} \langle \bm{k}_{n^\prime, l}^{(t)}, \bm{k}_{n, j}^{(t)} \rangle \\
        &+ \beta_{n, j, +}^{(t)} \langle \bm{q}_+^{(t)}, \bm{q}_{n, i}^{(t)} \rangle + \beta_{n, j, -}^{(t)} \langle \bm{q}_-^{(t)}, \bm{q}_{n, i}^{(t)} \rangle + \sum\limits_{n^\prime = 1}^N \sum\limits_{l=2}^M \beta_{n, j, n^\prime, l}^{(t)} \langle \bm{q}_{n^\prime, l}^{(t)}, \bm{q}_{n, i}^{(t)} \rangle \\
        &+ \Big( \alpha_{n, i, +}^{(t)} \bm{k}_+^{(t)} + \alpha_{n, i, -}^{(t)} \bm{k}_-^{(t)} + \sum\limits_{n^\prime =1}^N \sum\limits_{l=2}^M \alpha_{n, i, n^\prime, l}^{(t)} \bm{k}_{n^\prime, l}^{(t)} \Big) \\
        &\cdot \Big( \beta_{n, j, +}^{(t)} \bm{q}_+^{(t)\top} + \beta_{n, j, -}^{(t)} \bm{q}_-^{(t)\top} + \sum\limits_{n^\prime = 1}^N \sum\limits_{l=2}^M \beta_{n, j, n^\prime, l}^{(t)} \bm{q}_{n^\prime, l}^{(t)\top} \Big),
    \end{split}
    \end{equation*}
    for \(i, j \in [M] \backslash \{1\}, n \in [N]\).
\end{lemma}
\it{Proof of Lemma} \ref{update_rule4QK}. \rm Based on Definition \ref{def3}, we have
\begin{equation}
    \label{q+_dec}
    \Delta \bm{q}_+^{(t)} := \bm{q}_+^{(t + 1)} - \bm{q}_+^{(t)} = \alpha_{+, +}^{(t)} \bm{k}_+^{(t)} + \sum\limits_{n \in S_+} \sum\limits_{i=2}^M \alpha_{n, +, i}^{(t)} \bm{k}_{n, i}^{(t)}
\end{equation}
\begin{equation}
    \label{k+_dec}
    \Delta \bm{k}_+^{(t)} := \bm{k}_+^{(t + 1)} - \bm{k}_+^{(t)} = \beta_{+, +}^{(t)} \bm{q}_+^{(t)} + \sum\limits_{n \in S_+} \sum\limits_{i=2}^M \beta_{n, +, i}^{(t)} \bm{q}_{n, i}^{(t)} 
\end{equation}
Then we get
\begin{equation}
\begin{split}
    \label{qk_dynamic}
    &\langle \bm{q}_+^{(t+1)}, \bm{k}_+^{(t+1)} \rangle - \langle \bm{q}_+^{(t)}, \bm{k}_+^{(t)} \rangle \\
    &= \bm{\mu}_+^\top (\bm{W}_Q^{(t)} + \Delta\bm{W}_Q^{(t)}) (\bm{W}_K^{(t)\top} + \Delta\bm{W}_K^{(t)\top}) \bm{\mu}_+ - \langle \bm{q}_+^{(t)}, \bm{k}_+^{(t)} \rangle \\
    &= \langle \Delta \bm{q}_+^{(t)}, \bm{k}_+^{(t)} \rangle + \langle \bm{q}_+^{(t)}, \Delta \bm{k}_+^{(t)} \rangle + \langle \Delta \bm{q}_+^{(t)}, \Delta \bm{k}_+^{(t)} \rangle \\
    &= \alpha_{+, +}^{(t)} \Vert \bm{k}_+^{(t)} \Vert_2^2 + \sum\limits_{n \in S_+} \sum\limits_{i=2}^M \alpha_{n, +, i}^{(t)} \langle \bm{k}_+^{(t)}, \bm{k}_{n, i}^{(t)} \rangle \\
    &+ \beta_{+, +}^{(t)} \Vert \bm{q}_+^{(t)} \Vert_2^2 + \sum\limits_{n \in S_+} \sum\limits_{i=2}^M \beta_{n, +, i}^{(t)} \langle \bm{q}_+^{(t)}, \bm{q}_{n, i}^{(t)} \rangle \\
    &+ \Big( \alpha_{+, +}^{(t)} \bm{k}_+^{(t)} + \sum\limits_{n \in S_+} \sum\limits_{i=2}^M \alpha_{n, +, i}^{(t)} \bm{k}_{n, i}^{(t)} \Big) \\
    &\cdot \Big( \beta_{+, +}^{(t)} \bm{q}_+^{(t)\top} + \sum\limits_{n \in S_+} \sum\limits_{i=2}^M \beta_{n, +, i}^{(t)} \bm{q}_{n, i}^{(t)\top} \Big)
\end{split}
\end{equation}
where \( \Delta\bm{W}_Q^{(t)}:= \bm{W}_Q^{(t + 1)} - \bm{W}_Q^{(t)} \), \( \Delta\bm{W}_K^{(t)}:= \bm{W}_K^{(t + 1)} - \bm{W}_K^{(t)} \), the second equality is by \( \Delta \bm{q}_+^{(t)} = \bm{\mu}_+^\top \Delta\bm{W}_Q^{(t)} \) and \( \Delta \bm{k}_+^{(t)} = \bm{\mu}_+^\top \Delta\bm{W}_K^{(t)} \), 
the third equality is by plugging \eqref{q+_dec} and \eqref{k+_dec}.
This proves the update rule for \( \langle \bm{q}_+^{(t)}, \bm{k}_+^{(t)} \rangle \), the proof for \( \langle \bm{q}_{n, i}^{(t)}, \bm{k}_+^{(t)} \rangle \), \( \langle \bm{q}_+^{(t)}, \bm{k}_{n, j}^{(t)} \rangle \), \( \langle \bm{q}_{n, i}^{(t)}, \bm{k}_{n, j}^{(t)} \rangle \) are exactly the same.

\textbf{Remark.} In \eqref{qk_dynamic}, note that the magnitude of \( \langle \bm{k}_+^{(t)}, \bm{k}_{n, i}^{(t)} \rangle, \langle \bm{q}_+^{(t)}, \bm{q}_{n, i}^{(t)} \rangle \) is much smaller than that of \( \Vert \bm{k}_+^{(t)} \Vert_2^2, \Vert \bm{q}_+^{(t)} \Vert_2^2 \), and that the magnitude of \( \alpha_{n, +, i}^{(t)}, \beta_{n, +, i}^{(t)} \) will not be greater than that of \( \alpha_{+, +}^{(t)}, \beta_{+, +}^{(t)} \) (we will prove it later). Also, since the learning rate \(\eta\) is sufficiently small, the last term of \eqref{qk_dynamic} will be very small. Then \eqref{qk_dynamic} can be expressed in the following form
\begin{equation*}
\begin{split}
    &\langle \bm{q}_+^{(t+1)}, \bm{k}_+^{(t+1)} \rangle - \langle \bm{q}_+^{(t)}, \bm{k}_+^{(t)} \rangle \\
    &= \alpha_{+, +}^{(t)} \Vert \bm{k}_+^{(t)} \Vert_2^2 + \beta_{+, +}^{(t)} \Vert \bm{q}_+^{(t)} \Vert_2^2 + \{lower \ order \ term\}
\end{split}
\end{equation*}
The dynamics of \( \langle \bm{q}_{n, i}^{(t)}, \bm{k}_+^{(t)} \rangle, \langle \bm{q}_+^{(t)}, \bm{k}_{n, j}^{(t)} \rangle, \langle \bm{q}_{n, i}^{(t)}, \bm{k}_{n, j}^{(t)} \rangle \) can be expressed in a similar way.
Later we will rigorously explain the so-call \{lower order term\}.

\section{Concentration Inequalities}
\label{concentration_inequalities}
In this section, we will give some concentration inequalities that show some important properties of the data and the ViT parameters at random initialization.

\begin{lemma}[Lemma B.1 in \cite{cao2022benign}]
    \label{dataset_size}
    Suppose that \(\delta > 0\) and \(n \ge 8\log(4/\delta)\). Then with probability at least \(1 - \delta\),
    \[
        \frac{N}{4} \le \vert \{n \in [N] : y_n = 1\} \vert, \vert \{n \in [N] : y_n = -1\} \vert \le \frac{3N}{4}.
    \]
\end{lemma}

\begin{lemma}[Initialization of V]
    \label{initialization_of_V}
    Suppose that \(\delta > 0\). Then with probability at least \(1 - \delta\),
    \[
        \vert V_\pm^{(0)} \vert \le d_h^{-\frac{1}{4}},
    \]
    \[
        \vert V_{n, i}^{(0)} \vert \le d_h^{-\frac{1}{4}}
    \]
    for \( i \in [M] \backslash \{1\}, n \in [N] \).
\end{lemma}

\it{Proof of Lemma} \ref{initialization_of_V}. \rm It is clear that \(\bm{\mu}_\pm^\top \bm{W}_V^{(0)} \bm{w}_O\) is a random variable with mean zero and variance \(\sigma_V^2 \Vert \bm{w}_O \Vert_2^2 \Vert \bm{\mu} \Vert_2^2\).
Therefore, by Gaussian tail bound and the condition that \(\sigma_V \le \widetilde{O} \big( \Vert \bm{w}_O \Vert_2^{-1} \cdot \min \{ \Vert \bm{\mu} \Vert_2^{-1}, ( \sigma_p \sqrt{d} )^{-1} \} \cdot d_h^{-\frac{1}{4}} \big)\), with probability at least \(1 - \delta / NM\),
\[
    \vert \bm{\mu}_\pm^\top \bm{W}_V^{(0)} \bm{w}_O \vert \le \sigma_V \Vert \bm{w}_O \Vert_2 \Vert \bm{\mu} \Vert_2 \sqrt{\log (2NM / \delta)} \le d_h^{-\frac{1}{4}}
\]
Moreover, each element of vector \(\bm{w}_O^\top \bm{W}_V^{(0)}\) is a random variable with mean zero and variance \(\sigma_V^2 \Vert \bm{w}_O \Vert_2^2\). Therefore, by Bernstein's inequality and the condition of \(\sigma_V\), with probability at least \(1 - \delta / NM\)
\[
    \vert \bm{\xi}_{n, i}^\top \bm{W}_V^{(0)} \bm{w}_O \vert \le 2 \sigma_V \sigma_p \Vert \bm{w}_O \Vert_2 \sqrt{d \log (2NM / \delta)} \le d_h^{-\frac{1}{4}}
\]
Applying a union bound completes the proof.

\begin{lemma}[Initialization of QK]
    \label{initialization_QK}
    Suppose that \(\delta > 0\). Then with probability at least \(1 - \delta\),
    \[
        \frac{\Vert \bm{\mu} \Vert_2^2 \sigma_h^2 d_h}{2} \le \Vert \bm{q}_\pm^{(0)} \Vert_2^2 \le \frac{3 \Vert \bm{\mu} \Vert_2^2 \sigma_h^2 d_h}{2},
    \]
    \[
        \frac{\tilde{\sigma}_p^2 \sigma_h^2 d d_h}{2} \le \Vert \bm{q}_{n, 2}^{(0)} \Vert_2^2 \le \frac{3 \tilde{\sigma_p}^2 \sigma_h^2 d d_h}{2},
    \]
    \[
        \frac{\sigma_p^2 \sigma_h^2 d d_h}{2} \le \Vert \bm{q}_{n, i^\prime}^{(0)} \Vert_2^2 \le \frac{3 \sigma_p^2 \sigma_h^2 d d_h}{2},
    \]
    \[
        \frac{\Vert \bm{\mu} \Vert_2^2 \sigma_h^2 d_h}{2} \le \Vert \bm{k}_\pm^{(0)} \Vert_2^2 \le \frac{3 \Vert \bm{\mu} \Vert_2^2 \sigma_h^2 d_h}{2},
    \]
    \[
        \frac{\tilde{\sigma}_p^2 \sigma_h^2 d d_h}{2} \le \Vert \bm{k}_{n, 2}^{(0)} \Vert_2^2 \le \frac{3 \tilde{\sigma_p}^2 \sigma_h^2 d d_h}{2},
    \]
    \[
        \frac{\sigma_p^2 \sigma_h^2 d d_h}{2} \le \Vert \bm{k}_{n, i^\prime}^{(0)} \Vert_2^2 \le \frac{3 \sigma_p^2 \sigma_h^2 d d_h}{2},
    \]
    \[
        \vert \langle \bm{q}_+^{(0)}, \bm{q}_-^{(0)} \rangle \vert \le 2 \Vert \bm{\mu} \Vert_2^2 \sigma_h^2 \cdot \sqrt{d_h \log (6N^2M^2/\delta)},
    \]
    \[
        \vert \langle \bm{q}_\pm^{(0)}, \bm{q}_{n, i}^{(0)} \rangle \vert \le 2 \Vert \bm{\mu} \Vert_2 \tilde{\sigma}_p \sigma_h^2 d^{\frac{1}{2}} \cdot \sqrt{ d_h \log (6N^2M^2/\delta)},
    \]
    \[
        \vert \langle \bm{k}_+^{(0)}, \bm{k}_-^{(0)} \rangle \vert \le 2 \Vert \bm{\mu} \Vert_2^2 \sigma_h^2 \cdot \sqrt{d_h \log (6N^2M^2/\delta)},
    \]
    \[
        \vert \langle \bm{q}_\pm^{(0)}, \bm{k}_\pm^{(0)} \rangle \vert \le 2 \Vert \bm{\mu} \Vert_2^2 \sigma_h^2 \cdot \sqrt{d_h \log (6N^2M^2/\delta)},
    \]
    \[
        \vert \langle \bm{q}_\pm^{(0)}, \bm{k}_\mp^{(0)} \rangle \vert \le 2 \Vert \bm{\mu} \Vert_2^2 \sigma_h^2 \cdot \sqrt{d_h \log (6N^2M^2/\delta)},
    \]
    \[
        \vert \langle \bm{q}_{n, i}^{(0)}, \bm{k}_\pm^{(0)} \rangle \vert \le 2 \Vert \bm{\mu} \Vert_2 \tilde{\sigma}_p \sigma_h^2 d^{\frac{1}{2}} \cdot \sqrt{ d_h \log (6N^2M^2/\delta)},
    \]
    \[
        \vert \langle \bm{q}_{n, i}^{(0)}, \bm{q}_{n^\prime, j}^{(0)} \rangle \vert \le 2 \tilde{\sigma}_p^2 \sigma_h^2 d \cdot \sqrt{d_h \log (6N^2M^2/\delta)},
    \]
    \[
        \vert \langle \bm{k}_{n, i}^{(0)}, \bm{k}_{n^\prime, j}^{(0)} \rangle \vert \le 2 \tilde{\sigma}_p^2 \sigma_h^2 d \cdot \sqrt{d_h \log (6N^2M^2/\delta)},
    \]
    \[
        \vert \langle \bm{k}_\pm^{(0)}, \bm{k}_{n, i}^{(0)} \rangle \vert \le 2 \Vert \bm{\mu} \Vert_2 \tilde{\sigma}_p \sigma_h^2 d^{\frac{1}{2}} \cdot \sqrt{ d_h \log (6N^2M^2/\delta)},
    \]
    \[
        \vert \langle \bm{q}_\pm^{(0)}, \bm{k}_{n, i}^{(0)} \rangle \vert \le 2 \Vert \bm{\mu} \Vert_2 \tilde{\sigma}_p \sigma_h^2 d^{\frac{1}{2}} \cdot \sqrt{ d_h \log (6N^2M^2/\delta)},
    \]
    \[
        \vert \langle \bm{q}_{n, i}^{(0)}, \bm{k}_{n^\prime, j}^{(0)} \rangle \vert \le 2 \tilde{\sigma}_p^2 \sigma_h^2 d \cdot \sqrt{d_h \log (6N^2M^2/\delta)}
    \]
    for \(i, j \in [M] \backslash \{1\}\), \(i^\prime \in [M] \backslash \{1, 2\}\) and \(n, n^\prime \in [N]\).
\end{lemma}

\it{Proof of Lemma} \ref{initialization_QK}. \rm It is clear that each element of vector \(\bm{q}_+^{(0)}\) is a random variable with mean zero and variance \(\Vert \bm{\mu} \Vert_2^2 \sigma_h^2\). 
By Bernstein's inequality, with probability at least \(1 - \delta / 3N^2M^2\), we have
\[
    \vert \Vert \bm{q}_+^{(0)} \Vert_2^2 - \Vert \bm{\mu} \Vert_2^2 \sigma_h^2 d_h \vert = O(\Vert \bm{\mu} \Vert_2^2 \sigma_h^2 \sqrt{d_h \log (6N^2M^2/\delta)}).
\]
Therefore, as long as \(d_h = \Omega(\log (6N^2M^2/\delta))\), we have
\[
    \frac{\Vert \bm{\mu} \Vert_2^2 \sigma_h^2 d_h}{2} \le \Vert \bm{q}_+^{(0)} \Vert_2^2 \le \frac{3 \Vert \bm{\mu} \Vert_2^2 \sigma_h^2 d_h}{2},
\]
Similarly, we have
\[
    \frac{\Vert \bm{\mu} \Vert_2^2 \sigma_h^2 d_h}{2} \le \Vert \bm{q}_-^{(0)} \Vert_2^2 \le \frac{3 \Vert \bm{\mu} \Vert_2^2 \sigma_h^2 d_h}{2},
\]
\[
    \frac{\tilde{\sigma}_p^2 \sigma_h^2 d d_h}{2} \le \Vert \bm{q}_{n, 2}^{(0)} \Vert_2^2 \le \frac{3 \tilde{\sigma_p}^2 \sigma_h^2 d d_h}{2},
\]
\[
    \frac{\sigma_p^2 \sigma_h^2 d d_h}{2} \le \Vert \bm{q}_{n, i^\prime}^{(0)} \Vert_2^2 \le \frac{3 \sigma_p^2 \sigma_h^2 d d_h}{2},
\]
\[
    \frac{\Vert \bm{\mu} \Vert_2^2 \sigma_h^2 d_h}{2} \le \Vert \bm{k}_\pm^{(0)} \Vert_2^2 \le \frac{3 \Vert \bm{\mu} \Vert_2^2 \sigma_h^2 d_h}{2},
\]
\[
    \frac{\tilde{\sigma}_p^2 \sigma_h^2 d d_h}{2} \le \Vert \bm{k}_{n, 2}^{(0)} \Vert_2^2 \le \frac{3 \tilde{\sigma_p}^2 \sigma_h^2 d d_h}{2},
\]
\[
    \frac{\sigma_p^2 \sigma_h^2 d d_h}{2} \le \Vert \bm{k}_{n, i^\prime}^{(0)} \Vert_2^2 \le \frac{3 \sigma_p^2 \sigma_h^2 d d_h}{2},
\]
Moreover, \(\langle \bm{q}_+^{(0)}, \bm{q}_-^{(0)} \rangle\) has mean zero. By Bernstein's inequality, with probability at least \(1 - \delta /3N^2M^2\), we have
\[
    \vert \langle \bm{q}_+^{(0)}, \bm{q}_-^{(0)} \rangle \vert \le 2 \Vert \bm{\mu} \Vert_2^2 \sigma_h^2 \cdot \sqrt{d_h \log (6N^2M^2/\delta)}.
\]
We can prove the rest of this lemma in a similar way.

\begin{lemma} [Lemma B.2 in \cite{cao2022benign}]
    \label{caoyuan}
    Suppose that \(\delta > 0\) and d = \(\Omega( \log (4NM/\delta))\). Then with probability at least \(1 - \delta\)
    \[
        \tilde{\sigma}_p^2d / 2 \le \Vert \bm{\xi}_{n, 2} \Vert_2^2 \le 3\tilde{\sigma}_p^2d / 2,
    \]
    \[
        \sigma_p^2d / 2 \le \Vert \bm{\xi}_{n, i} \Vert_2^2 \le 3\sigma_p^2d / 2,
    \]
    \[
        \vert \langle \bm{\xi}_{n, i} , \bm{\xi}_{n^\prime, i^\prime} \rangle \vert \le 2 \tilde{\sigma}_p^2 \cdot \sqrt{d \log (4N^2M^2/\delta)}
    \]
    for \(i, i^\prime \in [M] \backslash \{1\}, n, n^\prime \in [N], i \ne i^\prime \ or \ n \ne n^\prime\).
\end{lemma}

\section{Benign Overfitting}
In this section, we consider the benign overfitting regime under the condition that \(N \cdot \mathrm{SNR}^2 = \Omega(1)\).
We analyze the dynamics of \(V_\pm\), \(V_{n, i}\), the inner product of \(\bm{q}_\pm\), \(\bm{q}_{n, i}\) and \(\bm{k}_\pm\), \(\bm{k}_{n, i}\) during gradient descent training, and further give the upper bound for population loss. The proofs in this section are based on the results in Section \ref{concentration_inequalities}, which hold with high probability.

\subsection{Stage \uppercase\expandafter{\romannumeral1}}
In Stage \uppercase\expandafter{\romannumeral1}, \( V_\pm^{(t)} \), \( V_{n, i}^{(t)} \) begin to pull apart until \( \vert V_\pm^{(t)} \vert \) is sufficiently larger than \( \vert V_{n, i}^{(t)} \vert \). At the same time, the inner products of \(\bm{q}\) and \(\bm{k}\) maintain their magnitude.

\begin{lemma}[Gradient of Loss]
    \label{gradient_of_loss}
    As long as \( \max \{ \vert V_+^{(t)} \vert , \vert V_-^{(t)} \vert, \vert V_{n, i}^{(t)} \vert \} = o (1) \), we have \( - \ell^\prime (y_n f(\bm{X}_n, \theta (t))) \) remains \( 1/2 \pm o(1) \).
\end{lemma}
\it Proof of Lemma \ref{gradient_of_loss}. \rm Note that \( \ell (z) = \log(1 + \exp(-z)) \) and \( - \ell^\prime = \exp(-z) / (1 + \exp(-z)) \), without loss of generality, we assume \( y_n = 1 \), we have
\[
    - \ell^\prime (f(\bm{X}_n, \theta (t))) = \frac{1}{1 + \exp(\frac{1}{M} \sum\limits_{l=1}^M \bm{\varphi} (\bm{x}_{n, l}^\top \bm{W}_{Q}^{(t)} \bm{W}_{K}^{(t)\top} \bm{X}_n^\top ) \bm{X}_n \bm{W}_V^{(t)} \bm{w}_O)}.
\]
Note that
\[
    - \max \{ \vert V_+^{(t)} \vert , \vert V_-^{(t)} \vert, \vert V_{n, i}^{(t)} \vert \} \le \frac{1}{M} \sum\limits_{l=1}^M \bm{\varphi} (\bm{x}_{n, l}^\top \bm{W}_{Q}^{(t)} \bm{W}_{K}^{(t)\top} \bm{X}_n^\top ) \bm{X}_n \bm{W}_V^{(t)} \bm{w}_O \le \max \{ \vert V_+^{(t)} \vert , \vert V_-^{(t)} \vert, \vert V_{n, i}^{(t)} \vert \}.
\]
Then we have
\[
    - \ell^\prime (f(\bm{X}_n, \theta (t))) \ge \frac{1}{1 + \exp(0 + o(1))} \ge \frac{1}{2 + o(1)} \ge \frac{1}{2} - o(1),
\]
\[
    - \ell^\prime (f(\bm{X}_n, \theta (t))) \le \frac{\exp(0 + o(1))}{1 + \exp(0 + o(1))} \le \frac{1 + o(1)}{1 + 1 + o(1)} \le \frac{1}{2} + o(1).
\]

\begin{lemma}[Bound of Attention]
    \label{bound_of_attention}
    As long as \(\vert \langle \bm{q}_\pm^{(t)}, \bm{k}_\pm^{(t)} \rangle \vert, \vert \langle \bm{q}_{n, i}^{(t)}, \bm{k}_\pm^{(t)} \rangle \vert, \vert \langle \bm{q}_\pm^{(t)}, \bm{k}_{n, j}^{(t)} \rangle \vert, \vert \langle \bm{q}_{n, i}^{(t)}, \bm{k}_{n, j}^{(t)} \rangle \vert = o(1)\), we have
    \[
        \frac{1}{M} - o(1) \le softmax(\langle \bm{q}_\pm^{(t)}, \bm{k}_\pm^{(t)} \rangle) \le \frac{1}{M} + o(1),
    \]
    \[
        \frac{1}{M} - o(1) \le  softmax(\langle \bm{q}_{n, i}^{(t)}, \bm{k}_\pm^{(t)} \rangle) \le \frac{1}{M} + o(1),
    \]
    \[
        \frac{1}{M} - o(1) \le  softmax(\langle \bm{q}_\pm^{(t)}, \bm{k}_{n, j}^{(t)} \rangle) \le \frac{1}{M} + o(1),
    \]
    \[
        \frac{1}{M} - o(1) \le  softmax(\langle \bm{q}_{n, i}^{(t)}, \bm{k}_{n, j}^{(t)} \rangle) \le \frac{1}{M} + o(1).
    \]
\end{lemma}
\it Proof of Lemma \ref{bound_of_attention}. \rm It is clear that \(\exp (o(1)) = 1 + o(1)\). Therefore, as long as \(\vert \langle \bm{q}_\pm^{(t)}, \bm{k}_\pm^{(t)} \rangle \vert = o(1)\), we have
\begin{equation*}
\begin{split}
    &\frac{1}{M} - o(1) = \frac{1}{1 + (M - 1) + (M - 1) o(1)} = \frac{1}{1 + (M - 1) \exp (o(1))} =\\
    &\frac{\exp (-o(1))}{\exp (-o(1)) + (M - 1) \exp (o(1))} \le  softmax(\langle \bm{q}_\pm^{(t)}, \bm{k}_\pm^{(t)} \rangle) \le \frac{\exp (o(1))}{\exp (o(1)) + (M - 1) \exp (-o(1))} \\
    &= \frac{\exp (o(1))}{\exp (o(1)) + (M - 1)} = \frac{1 + o(1)}{1 + o(1) + (M - 1)} = \frac{1}{M} + o(1)
\end{split}
\end{equation*}
Similarly, we have
    \[
        \frac{1}{M} - o(1) \le  softmax(\langle \bm{q}_{n, i}^{(t)}, \bm{k}_\pm^{(t)} \rangle) \le \frac{1}{M} + o(1),
    \]
    \[
        \frac{1}{M} - o(1) \le  softmax(\langle \bm{q}_\pm^{(t)}, \bm{k}_{n, j}^{(t)} \rangle) \le \frac{1}{M} + o(1),
    \]
    \[
        \frac{1}{M} - o(1) \le  softmax(\langle \bm{q}_{n, i}^{(t)}, \bm{k}_{n, j}^{(t)} \rangle) \le \frac{1}{M} + o(1).
    \]

\begin{lemma}[Upper bound of V]
    \label{upper_bound_of_V}
    Let \(T_0 = O(\frac{1}{\eta d_h^{\frac{1}{4}} \Vert \bm{\mu} \Vert_2^2 \Vert \bm{w}_O \Vert_2^2})\). Then under the same conditions as Theorem \ref{benign}, we have
    \[
        \vert V_+^{(t)} \vert, \vert V_-^{(t)} \vert, \vert V_{n, i}^{(t)} \vert = O (d_h^{-\frac{1}{4}})
    \]
    for \( t \in [0, T_0] \).
\end{lemma}
\it Proof of Lemma \ref{upper_bound_of_V}. \rm By Lemma \ref{update_rule4V}, we have
\begin{equation*}
\begin{split}
    &\vert \gamma_{V, +}^{(t + 1)} - \gamma_{V, +}^{(t)} \vert \\
    &\le - \frac{\eta \Vert \bm{\mu} \Vert_2^2}{NM} \sum\limits_{n \in S_+} \ell_n^{\prime(t)} \big( \frac{ \exp ( \langle \bm{q}_+^{(t)}, \bm{k}_+^{(t)} \rangle )}{  \exp ( \langle \bm{q}_+^{(t)}, \bm{k}_+^{(t)} \rangle ) + \sum\limits_{k=2}^M  \exp ( \langle \bm{q}_+^{(t)}, \bm{k}_{n, k}^{(t)} \rangle )}  \\
    &+ \sum\limits_{j=2}^M \frac{ \exp ( \langle \bm{q}_{n, j}^{(t)}, \bm{k}_+^{(t)} \rangle )}{  \exp ( \langle \bm{q}_{n, j}^{(t)}, \bm{k}_+^{(t)} \rangle ) + \sum\limits_{k=2}^M  \exp ( \langle \bm{q}_{n, j}^{(t)}, \bm{k}_{n, k}^{(t)} \rangle )} \big) \\
    &\le \frac{\eta \Vert \bm{\mu} \Vert_2^2}{NM} \cdot \frac{3N}{4} \big( 1 + (M - 1) \big) \\
    &\le \frac{3\eta \Vert \bm{\mu} \Vert_2^2}{4},
\end{split}
\end{equation*}
where the second inequality is by Lemma \ref{dataset_size} and \( - \ell_n^{\prime(t)} \le 1 \). Similarly, we have
\begin{equation*}
\begin{split}
    \vert \gamma_{V, -}^{(t + 1)} - \gamma_{V, -}^{(t)} \vert \le \frac{3\eta \Vert \bm{\mu} \Vert_2^2}{4}.
\end{split}
\end{equation*}
By Definition \ref{def1}, we have
\begin{equation*}
\begin{split}
    \vert V_+^{(t)} \vert &= \big\vert V_+^{(0)}  + \sum\limits_{s = 0}^{t - 1} ( \gamma_{V, +}^{(s + 1)} - \gamma_{V, +}^{(s)} ) \Vert \bm{w}_O \Vert_2^2 \big\vert \\
    &\le \vert V_+^{(0)} \vert + \sum\limits_{s = 0}^{t - 1} \vert \gamma_{V, +}^{(s + 1)} - \gamma_{V, +}^{(s)} \vert \cdot \Vert \bm{w}_O \Vert_2^2 \\
    &\le d_h^{-\frac{1}{4}} + \frac{3\eta \Vert \bm{\mu} \Vert_2^2}{4} \cdot \Vert \bm{w}_O \Vert_2^2 \cdot O(\frac{1}{\eta d_h^{\frac{1}{4}} \Vert \bm{\mu} \Vert_2^2 \Vert \bm{w}_O \Vert_2^2}) \\
    &= O(d_h^{-\frac{1}{4}}),
\end{split}
\end{equation*}
where the first inequality is by triangle inequality, the second inequality is by Lemma \ref{initialization_of_V}. Similarly, we have \( \vert V_-^{(t)} \vert = O(d_h^{-\frac{1}{4}}) \).

By Lemma \ref{update_rule4V}, we have
\begin{equation}
\begin{split}
    \label{rho_bound}
    &\vert \rho_{V, n, i}^{(t+1)} - \rho_{V, n, i}^{(t)} \vert \\
    &\le \Big\vert - \frac{\eta}{NM} \sum\limits_{n^\prime \in S_+} \ell_{n^\prime}^{\prime(t)} \big( 
    \sum\limits_{i=2}^M ( \langle \bm{\xi}_{n, i}, \bm{\xi}_{n^\prime, i^\prime} \rangle \frac{ \exp ( \langle \bm{q}_+^{(t)}, \bm{k}_{n^\prime, i^\prime}^{(t)} \rangle )}{  \exp ( \langle \bm{q}_+^{(t)}, \bm{k}_+^{(t)} \rangle ) + \sum\limits_{k=2}^M  \exp ( \langle \bm{q}_+^{(t)}, \bm{k}_{n^\prime, k}^{(t)} \rangle )} \\
    &+ \sum\limits_{j=2}^M \langle \bm{\xi}_{n, i}, \bm{\xi}_{n^\prime, i^\prime} \rangle \frac{ \exp ( \langle \bm{q}_{n^\prime, j}^{(t)}, \bm{k}_{n^\prime, i^\prime}^{(t)} \rangle  )}{ \exp ( \langle \bm{q}_{n^\prime, j}^{(t)}, \bm{k}_+^{(t)} \rangle ) + \sum\limits_{k=2}^M  \exp ( \langle \bm{q}_{n^\prime, j}^{(t)}, \bm{k}_{n^\prime, k}^{(t)} \rangle )} ) \big) \\
    &+ \frac{\eta}{NM} \sum\limits_{n^\prime \in S_-} \ell_{n^\prime}^{\prime(t)} \big(
    \sum\limits_{i=2}^M ( \langle \bm{\xi}_{n, i}, \bm{\xi}_{n^\prime, i^\prime} \rangle \frac{ \exp ( \langle \bm{q}_-^{(t)}, \bm{k}_{n^\prime, i^\prime}^{(t)} \rangle )}{  \exp ( \langle \bm{q}_-^{(t)}, \bm{k}_-^{(t)} \rangle ) + \sum\limits_{k=2}^M  \exp ( \langle \bm{q}_-^{(t)}, \bm{k}_{n^\prime, k}^{(t)} \rangle )} \\
    &+ \sum\limits_{j=2}^M \langle \bm{\xi}_{n, i}, \bm{\xi}_{n^\prime, i^\prime} \rangle \frac{ \exp ( \langle \bm{q}_{n^\prime, j}^{(t)}, \bm{k}_{n^\prime, i^\prime}^{(t)} \rangle  )}{ \exp ( \langle \bm{q}_{n^\prime, j}^{(t)}, \bm{k}_-^{(t)} \rangle ) + \sum\limits_{k=2}^M  \exp ( \langle \bm{q}_{n^\prime, j}^{(t)}, \bm{k}_{n^\prime, k}^{(t)} \rangle )} ) \big) \Big\vert \\
    &\le \frac{3 \eta \tilde{\sigma}_p^2 d}{2NM} \cdot M + \frac{\eta}{NM} \cdot MN \cdot 2 \tilde{\sigma}_p^2 \cdot \sqrt{d \log (4N^2M^2/\delta)} \\
    &\le \frac{2 \eta \tilde{\sigma}_p^2 d}{N} \\
    &= O ( \eta \Vert \bm{\mu} \Vert_2^2 )
\end{split}
\end{equation}
where the second inequality is by Lemma \ref{caoyuan} and \( - \ell_n^{\prime(t)} \le 1 \), the third inequality is by \(d = \widetilde{\Omega} \Big( \epsilon^{-2} N^2 d_h \Big) \ge 4N\sqrt{\log (4N^2M^2/\delta)} \), the last inequality is by \( N \cdot \mathrm{SNR}^2 = \Omega(1) \). Then by Definition \ref{def1}, we have
\begin{equation*}
\begin{split}
    \vert V_{n, i}^{(t)} \vert &= \big\vert V_{n, i}^{(0)}  + \sum\limits_{s = 0}^{t - 1} ( \rho_{V, n, i}^{(s + 1)} - \rho_{V, n, i}^{(s)} ) \Vert \bm{w}_O \Vert_2^2 \big\vert \\
    &\le \vert V_{n, i}^{(0)} \vert + \sum\limits_{s = 0}^{t - 1} \vert \rho_{V, n, i}^{(s + 1)} - \rho_{V, n, i}^{(s)} \vert \cdot \Vert \bm{w}_O \Vert_2^2 \\
    &\le d_h^{-\frac{1}{4}} + O ( \eta \Vert \bm{\mu} \Vert_2^2 ) \cdot \Vert \bm{w}_O \Vert_2^2 \cdot O(\frac{1}{\eta d_h^{\frac{1}{4}} \Vert \bm{\mu} \Vert_2^2 \Vert \bm{w}_O \Vert_2^2}) \\
    &= O(d^{-\frac{1}{4}}),
\end{split}
\end{equation*}
where the first inequality is by triangle inequality, the second inequality is by Lemma \ref{initialization_of_V}, which completes the proof.

\begin{lemma}[Inner Products Hold Magnitude]
    \label{inner_products_hold_magnitude}
    Let \(T_0 = O(\frac{1}{\eta d_h^{\frac{1}{4}} \Vert \bm{\mu} \Vert_2^2 \Vert \bm{w}_O \Vert_2^2})\). Then under the same conditions as Theorem \ref{benign}, we have
    \begin{align*}
        &\vert \langle \bm{q}_\pm^{(t)}, \bm{k}_\pm^{(t)} \rangle \vert, \vert \langle \bm{q}_{n, i}^{(t)}, \bm{k}_\pm^{(t)} \rangle \vert, \vert \langle \bm{q}_\pm^{(t)}, \bm{k}_{n, j}^{(t)} \rangle \vert, \vert \langle \bm{q}_{n, i}^{(t)}, \bm{k}_{n^\prime, j}^{(t)} \rangle \vert \\
        &= O \Big( \max \{ \Vert \bm{\mu} \Vert_2^2, \sigma_p^2 d \} \cdot \sigma_h^2 \cdot \sqrt{d_h \log (6N^2M^2/\delta)} \Big),
    \end{align*}
    \begin{align*}
        &\vert \langle \bm{q}_\pm^{(t)}, \bm{q}_\mp^{(t)} \rangle \vert, \vert \langle \bm{q}_{n, i}^{(t)}, \bm{q}_\pm^{(t)} \rangle \vert, \vert \langle \bm{q}_{n, i}^{(t)}, \bm{q}_{n^\prime, j}^{(t)} \rangle \vert \\
        &= O \Big( \max \{ \Vert \bm{\mu} \Vert_2^2, \sigma_p^2 d \} \cdot \sigma_h^2 \cdot \sqrt{d_h \log (6N^2M^2/\delta)} \Big),
    \end{align*}
    \begin{align*}
        &\vert \langle \bm{k}_\pm^{(t)}, \bm{k}_\mp^{(t)} \rangle \vert, \vert \langle \bm{k}_{n, i}^{(t)}, \bm{k}_\pm^{(t)} \rangle \vert, \vert \langle \bm{k}_{n, i}^{(t)}, \bm{k}_{n^\prime, j}^{(t)} \rangle \vert \\
        &= O \Big( \max \{ \Vert \bm{\mu} \Vert_2^2, \sigma_p^2 d \} \cdot \sigma_h^2 \cdot \sqrt{d_h \log (6N^2M^2/\delta)} \Big),
    \end{align*}
    \[
        \Vert \bm{q}_\pm^{(t)} \Vert_2^2, \Vert \bm{k}_\pm^{(t)} \Vert_2^2 = \Theta (\Vert \bm{\mu} \Vert_2^2 \sigma_h^2 d_h),
    \]
    \[
        \Vert \bm{q}_{n, i}^{(t)} \Vert_2^2, \Vert \bm{k}_{n, i}^{(t)} \Vert_2^2 = \Theta ( \sigma_p^2 \sigma_h^2 d d_h )
    \]
    for \( i, j \in [M] \backslash \{1\} \), \( n, n^\prime \in [N] \) and \( t \in [0, T_0] \).
\end{lemma}
The proof for Lemma \ref{inner_products_hold_magnitude} is in Section \ref{proof4lemmaC4}. Note that \(\sigma_h^2 \le \min \{ \Vert \bm{\mu} \Vert_2^{-2}, ( \sigma_p^2 d )^{-1} \} \cdot d_h^{-\frac{1}{2}} \cdot \big( \log ( 6N^2M^2 / \delta ) \big)^{-\frac{3}{2}}\), thus \(O \Big( \max \{ \Vert \bm{\mu} \Vert_2^2, \sigma_p^2 d \} \cdot \sigma_h^2 \cdot \sqrt{d_h \log (6N^2M^2/\delta)} \Big) = o(1)\).
\begin{lemma}[V's Beginning of Learning Signals]
    \label{stage1}
    Under the same conditions as Theorem \ref{benign}, there exist \(T_1 = \frac{10M(3M + 1) N}{\eta d_h^{\frac{1}{4}} (N \Vert \bm{\mu} \Vert_2^2 - 60M^2 C_p^2 \sigma_p^2 d) \Vert \bm{w}_O \Vert_2^2} \) such that 
    the first element of vector \( \bm{X}_n \bm{W}_V^{(t)} \bm{w}_O \) dominate its other elements, that is, \( V_+^{(t)} \ge 3 M \cdot \vert V_{n, i}^{(t)} \vert \) for all \( n \in S_+ \), \( i \in [M] \backslash \{1\} \) and
    \( V_-^{(t)} \le - 3 M \cdot \vert V_{n, i}^{(t)} \vert \) for all \( n \in S_- \), \( i \in [M] \backslash \{1\} \).
\end{lemma}
\it Proof of Lemma \ref{stage1}. \rm
Let \(C\) be a constant larger than \(10M(3M + 1)\), then as long as \(N \cdot \mathrm{SNR}^2 = \frac{N \Vert \bm{\mu} \Vert_2^2}{\sigma_p^2 d} \ge \frac{60C M^2 C_p^2}{C - 10M(3M + 1)}\), we have \( N \Vert \bm{\mu} \Vert_2^2 - 60M^2 C_p^2 \sigma_p^2 d \ge \frac{10M(3M + 1) N \Vert \bm{\mu} \Vert_2^2}{C} \),
and further get
\[
    T_1 = \frac{10M(3M + 1) N}{\eta d_h^{\frac{1}{4}} (N \Vert \bm{\mu} \Vert_2^2 - 60M^2 C_p^2 \sigma_p^2 d) \Vert \bm{w}_O \Vert_2^2} \le \frac{C}{\eta d_h^{\frac{1}{4}} \Vert \bm{\mu} \Vert_2^2 \Vert \bm{w}_O \Vert_2^2} = O \Big( \frac{1}{\eta d_h^{\frac{1}{4}} \Vert \bm{\mu} \Vert_2^2 \Vert \bm{w}_O \Vert_2^2} \Big),
\]
which satisfies the time condition in Lemma \ref{upper_bound_of_V} and Lemma \ref{inner_products_hold_magnitude}. Then by Lemma \ref{gradient_of_loss} and Lemma \ref{bound_of_attention} we have
\[
    - \ell_n^{\prime(t)} = \frac{1}{2} \pm o(1),  
\]
\[
    \frac{1}{M} - o(1) \le softmax(\langle \bm{q}_\pm^{(t)}, \bm{k}_\pm^{(t)} \rangle) \le \frac{1}{M} + o(1),
\]
\[
    \frac{1}{M} - o(1) \le  softmax(\langle \bm{q}_{n, i}^{(t)}, \bm{k}_\pm^{(t)} \rangle) \le \frac{1}{M} + o(1),
\]
\[
    \frac{1}{M} - o(1) \le  softmax(\langle \bm{q}_\pm^{(t)}, \bm{k}_{n, j}^{(t)} \rangle) \le \frac{1}{M} + o(1),
\]
\[
    \frac{1}{M} - o(1) \le  softmax(\langle \bm{q}_{n, i}^{(t)}, \bm{k}_{n, j}^{(t)} \rangle) \le \frac{1}{M} + o(1)
\]
for \( i, j \in [M] \backslash \{1\} \), \( n \in [N] \) and \( t \in [0, T_1] \). Plugging them in the update rule for \(\gamma_{V, +}^{(t)}\) showed in Lemma \ref{update_rule4V} and we have
\begin{equation}
\begin{split}
    & \gamma_{V, +}^{(t + 1)} - \gamma_{V, +}^{(t)} \\
    &= - \frac{\eta \Vert \bm{\mu} \Vert_2^2}{NM} \sum\limits_{n \in S_+} \ell_n^{\prime(t)} \big( \frac{ \exp ( \langle \bm{q}_+^{(t)}, \bm{k}_+^{(t)} \rangle )}{  \exp ( \langle \bm{q}_+^{(t)}, \bm{k}_+^{(t)} \rangle ) + \sum\limits_{k=2}^M  \exp ( \langle \bm{q}_+^{(t)}, \bm{k}_{n, k}^{(t)} \rangle )}  \\
    &+ \sum\limits_{j=2}^M \frac{ \exp ( \langle \bm{q}_{n, j}^{(t)}, \bm{k}_+^{(t)} \rangle )}{  \exp ( \langle \bm{q}_{n, j}^{(t)}, \bm{k}_+^{(t)} \rangle ) + \sum\limits_{k=2}^M  \exp ( \langle \bm{q}_{n, j}^{(t)}, \bm{k}_{n, k}^{(t)} \rangle )} \big) \\
    &\ge \frac{\eta \Vert \bm{\mu} \Vert_2^2}{NM} \cdot \frac{N}{4} \cdot (\frac{1}{2} \pm o(1)) \cdot M (\frac{1}{M} \pm o(1)) \\
    &\ge \frac{\eta \Vert \bm{\mu} \Vert_2^2}{10 M}.
\end{split}
\end{equation}
Then by Definition \ref{def1} and taking a summation, we have
\begin{equation}
\begin{split}
    \label{V_p_bound}
    V_+^{(T_1)} &\ge - \vert V_+^{(0)} \vert + T_1 \frac{\eta \Vert \bm{\mu} \Vert_2^2}{10 M} \Vert \bm{w}_O \Vert_2^2 \\
    &= -d_h^{-\frac{1}{4}} + \frac{10M(3M + 1) N}{\eta d_h^{\frac{1}{4}} (N \Vert \bm{\mu} \Vert_2^2 - 60M^2 C_p^2 \sigma_p^2 d) \Vert \bm{w}_O \Vert_2^2} \frac{\eta \Vert \bm{\mu} \Vert_2^2}{10 M} \Vert \bm{w}_O \Vert_2^2 \\
    &= -d_h^{-\frac{1}{4}} + \frac{(3 M + 1) N \Vert \bm{\mu} \Vert_2^2}{ d_h^{\frac{1}{4}} ( N \Vert \bm{\mu} \Vert_2^2 - 60M^2 C_p^2 \sigma_p^2 d)}.
\end{split}
\end{equation}
Similarly, we have
\begin{equation}
\begin{split}
    \label{V_m_bound}
    V_-^{(T_1)} \le d_h^{-\frac{1}{4}} - \frac{(3 M + 1) N \Vert \bm{\mu} \Vert_2^2}{ d_h^{\frac{1}{4}} ( N \Vert \bm{\mu} \Vert_2^2 - 60M^2 C_p^2 \sigma_p^2 d)}.
\end{split}
\end{equation}
Similarly, by \(\vert \rho_{V, n, i}^{(t+1)} - \rho_{V, n, i}^{(t)} \vert \le \frac{2 \eta \tilde{\sigma}_p^2 d}{N} \) in \eqref{rho_bound}, we have
\begin{equation}
\begin{split}
    \label{V_i_bound}
    \vert V_{n, i}^{(T_1)} \vert &\le \vert V_{n, i}^{(0)} \vert + T_1 \frac{2 \eta \tilde{\sigma}_p^2 d}{N} \Vert \bm{w}_O \Vert_2^2 \\
    &= d_h^{-\frac{1}{4}} + \frac{10M(3M + 1) N}{\eta d_h^{\frac{1}{4}} (N \Vert \bm{\mu} \Vert_2^2 - 60M^2 C_p^2 \sigma_p^2 d) \Vert \bm{w}_O \Vert_2^2} \frac{2 \eta \tilde{\sigma}_p^2 d}{N} \Vert \bm{w}_O \Vert_2^2 \\
    &= d_h^{-\frac{1}{4}} + \frac{20 M C_p^2 (3 M + 1) \sigma_p^2 d}{ d_h^{\frac{1}{4}} ( N \Vert \bm{\mu} \Vert_2^2 - 60M^2 C_p^2 \sigma_p^2 d)}.
\end{split}
\end{equation}
According to \eqref{V_p_bound}, \eqref{V_m_bound} and \eqref{V_i_bound}, it is easy to verify that \( V_+^{(T_1)} - 3 M \cdot \vert V_{n, i}^{(T_1)} \vert \ge 0 \) and \( V_-^{(T_1)} + 3 M \cdot \vert V_{n, i}^{(T_1)} \vert \le 0 \), which completes the proof.

\subsection{Stage \uppercase\expandafter{\romannumeral2}}
\label{stage2}

In stage \uppercase\expandafter{\romannumeral2}, \(\langle \bm{q}_+, \bm{k}_+ \rangle\), \(\langle \bm{q}_{n, i}, \bm{k}_+ \rangle\) grows while \(\langle \bm{q}_+, \bm{k}_{n, j} \rangle\), \(\langle \bm{q}_{n, i}, \bm{k}_{n, j} \rangle\) decreases, resulting in attention focusing more and more on the signals and less on the noises. 
By the results of stage \uppercase\expandafter{\romannumeral1}, we have the following conditions at the beginning of stage \uppercase\expandafter{\romannumeral2}

\[
    V_+^{(T_1)} \ge 3 M \cdot \vert V_{n, i}^{(T_1)} \vert,
\]
\[
    V_-^{(T_1)} \le - 3 M \cdot \vert V_{n, i}^{(T_1)} \vert,
\]
\[
    \vert V_+^{(T_1)} \vert, \vert V_-^{(T_1)} \vert, \vert V_{n, i}^{(T_1)} \vert = O (d_h^{-\frac{1}{4}}),
\]
\begin{align*}
    &\vert \langle \bm{q}_\pm^{(T_1)}, \bm{k}_\pm^{(T_1)} \rangle \vert, \vert \langle \bm{q}_{n, i}^{(T_1)}, \bm{k}_\pm^{(T_1)} \rangle \vert, \vert \langle \bm{q}_\pm^{(T_1)}, \bm{k}_{n, j}^{(T_1)} \rangle \vert, \vert \langle \bm{q}_{n, i}^{(T_1)}, \bm{k}_{n^\prime, j}^{(T_1)} \rangle \vert \\
    &= O \Big( \max \{ \Vert \bm{\mu} \Vert_2^2, \sigma_p^2 d \} \cdot \sigma_h^2 \cdot \sqrt{d_h \log (6N^2M^2/\delta)} \Big),
\end{align*}
\begin{align*}
    &\vert \langle \bm{q}_\pm^{(T_1)}, \bm{q}_\mp^{(T_1)} \rangle \vert, \vert \langle \bm{q}_{n, i}^{(T_1)}, \bm{q}_\pm^{(T_1)} \rangle \vert, \vert \langle \bm{q}_{n, i}^{(T_1)}, \bm{q}_{n^\prime, j}^{(T_1)} \rangle \vert \\
    &= O \Big( \max \{ \Vert \bm{\mu} \Vert_2^2, \sigma_p^2 d \} \cdot \sigma_h^2 \cdot \sqrt{d_h \log (6N^2M^2/\delta)} \Big),
\end{align*}
\begin{align*}
    &\vert \langle \bm{k}_\pm^{(T_1)}, \bm{k}_\mp^{(T_1)} \rangle \vert, \vert \langle \bm{k}_{n, i}^{(T_1)}, \bm{k}_\pm^{(T_1)} \rangle \vert, \vert \langle \bm{k}_{n, i}^{(T_1)}, \bm{k}_{n^\prime, j}^{(T_1)} \rangle \vert \\
    &= O \Big( \max \{ \Vert \bm{\mu} \Vert_2^2, \sigma_p^2 d \} \cdot \sigma_h^2 \cdot \sqrt{d_h \log (6N^2M^2/\delta)} \Big),
\end{align*}
\[
    \Vert \bm{q}_\pm^{(T_1)} \Vert_2^2, \Vert \bm{k}_\pm^{(T_1)} \Vert_2^2 = \Theta (\Vert \bm{\mu} \Vert_2^2 \sigma_h^2 d_h),
\]
\[
    \Vert \bm{q}_{n, i}^{(T_1)} \Vert_2^2, \Vert \bm{k}_{n, i}^{(T_1)} \Vert_2^2 = \Theta ( \sigma_p^2 \sigma_h^2 d d_h )
\]
for \( i, j \in [M] \backslash \{1\} \), \( n, n^\prime \in [N] \).

Some of the proofs at this stage are based on the above conditions.

\textbf{Notations.} To better characterize the gap between different inner products, we define the following notations:
\begin{itemize}
    \item denote \( \Lambda_{n, +, j}^{(t)} = \langle \bm{q}_+^{(t)}, \bm{k}_+^{(t)} \rangle - \langle \bm{q}_+^{(t)}, \bm{k}_{n, j}^{(t)} \rangle, \quad n \in S_+ \).
    \item denote \( \Lambda_{n, -, j}^{(t)} = \langle \bm{q}_-^{(t)}, \bm{k}_-^{(t)} \rangle - \langle \bm{q}_-^{(t)}, \bm{k}_{n, j}^{(t)} \rangle, \quad n \in S_- \).
    \item denote \( \Lambda_{n, i, +, j}^{(t)} = \langle \bm{q}_{n, i}^{(t)}, \bm{k}_+^{(t)} \rangle - \langle \bm{q}_{n, i}^{(t)}, \bm{k}_{n, j}^{(t)} \rangle, \quad n \in S_+ \).
    \item denote \( \Lambda_{n, i, -, j}^{(t)} = \langle \bm{q}_{n, i}^{(t)}, \bm{k}_-^{(t)} \rangle - \langle \bm{q}_{n, i}^{(t)}, \bm{k}_{n, j}^{(t)} \rangle, \quad n \in S_- \).
\end{itemize}
\begin{lemma}[Upper bound of V]
    \label{upper_bound_of_V2}
    Let \(T_0 = O \Big( \frac{1}{\eta \Vert \bm{\mu} \Vert_2^2 \Vert \bm{w}_O \Vert_2^2 \log (6N^2M^2 / \delta)} \Big)\). Then under the same conditions as Theorem \ref{benign}, we have
    \[
        \vert V_+^{(t)} \vert, \vert V_-^{(t)} \vert, \vert V_{n, i}^{(t)} \vert = o (1)
    \]
    for \( t \in [0, T_0] \).
\end{lemma}

The proof of Lemma \ref{upper_bound_of_V2} is similar to that of Lemma \ref{upper_bound_of_V}, except that the time \(T_0\) is changed.

Let \(T_2 = \Theta \Big( \frac{1}{\eta \Vert \bm{\mu} \Vert_2^2 \Vert \bm{w}_O \Vert_2^2 \log (6N^2M^2 / \delta)} \Big)\), then by Lemma \ref{upper_bound_of_V2} and Lemma \ref{gradient_of_loss} we have \(\frac{1}{2} - o(1) \le - \ell_n^{\prime(t)} \le \frac{1}{2} + o(1) \) for \(n \in [N], t \in [T_1, T_2] \), 
which can simplify the calculations of \(\alpha\) and \(\beta\) defined in Definition \ref{def3} by replacing \( - \ell_n^{\prime(t)} \) by their bounds. 
Next we prove the following four propositions \( \mathcal{B}(t) \), \( \mathcal{C}(t) \), \( \mathcal{D}(t) \), \( \mathcal{E}(t) \) by induction on t for \(t \in [T_1, T_2]\):

\begin{itemize}
    \item \( \mathcal{B}(t): \)
    \[
        V_+^{(t)} \ge \eta C_3 \Vert \bm{\mu} \Vert_2^2 \Vert \bm{w}_O \Vert_2^2 (t - T_1)
    \]
    \[
        V_-^{(t)} \le - \eta C_3 \Vert \bm{\mu} \Vert_2^2 \Vert \bm{w}_O \Vert_2^2 (t - T_1)
    \]
    \[
        V_+^{(t)} \ge 3 M \cdot \vert V_{n, i}^{(t)} \vert,
    \]
    \[
        V_-^{(t)} \le - 3 M \cdot \vert V_{n, i}^{(t)} \vert,
    \]
    \[
        \vert V_\pm^{(t)} \vert \le O(d_h^{-\frac{1}{4}}) + \eta C_4 \Vert \bm{\mu} \Vert_2^2 \Vert \bm{w}_O \Vert_2^2 (t - T_1)
    \]
    \[
        \vert V_{n, i}^{(t)} \vert \le O(d_h^{-\frac{1}{4}}) + \eta C_4 \Vert \bm{\mu} \Vert_2^2 \Vert \bm{w}_O \Vert_2^2 (t - T_1)
    \]
    for \(i \in [M] \backslash \{1\}, n \in [N]\).

    \item \( \mathcal{C}(t): \)
    \[
        \Vert \bm{q}_\pm^{(t)} \Vert_2^2, \Vert \bm{k}_\pm^{(t)} \Vert_2^2 = \Theta \Big( \Vert \bm{\mu} \Vert_2^2 \sigma_h^2 d_h \Big),
    \]
    \[
        \Vert \bm{q}_{n, i}^{(t)} \Vert_2^2, \Vert \bm{k}_{n, i}^{(t)} \Vert_2^2 = \Theta \Big( \sigma_p^2 \sigma_h^2 d d_h \Big),
    \]
    \[
        \vert \langle \bm{q}_+^{(t)}, \bm{q}_-^{(t)} \rangle \vert, \vert \langle \bm{q}_\pm^{(t)}, \bm{q}_{n, i}^{(t)} \rangle \vert, \vert \langle \bm{q}_{n, i}^{(t)}, \bm{q}_{n^\prime, j}^{(t)} \rangle \vert = o(1),
    \]
    \[
        \vert \langle \bm{k}_+^{(t)}, \bm{k}_-^{(t)} \rangle \vert, \vert \langle \bm{k}_\pm^{(t)}, \bm{k}_{n, i}^{(t)} \rangle \vert, \vert \langle \bm{k}_{n, i}^{(t)}, \bm{k}_{n^\prime, j}^{(t)} \rangle \vert = o(1),
    \]
    for \(i, j \in [M] \backslash \{1\}, n, n^\prime \in [N], i \ne j \ or \ n \ne n^\prime \).

    \item \( \mathcal{D}(t): \)
    \[
        \langle \bm{q}_\pm^{(t + 1)}, \bm{k}_\pm^{(t + 1)} \rangle \ge \langle \bm{q}_\pm^{(t)}, \bm{k}_\pm^{(t)} \rangle
    \]
    \[
        \langle \bm{q}_{n, i}^{(t + 1)}, \bm{k}_\pm^{(t + 1)} \rangle \ge \langle \bm{q}_{n, i}^{(t)}, \bm{k}_\pm^{(t)} \rangle
    \]
    \[
        \langle \bm{q}_\pm^{(t + 1)}, \bm{k}_{n, j}^{(t + 1)} \rangle \le \langle \bm{q}_\pm^{(t)}, \bm{k}_{n, j}^{(t)} \rangle
    \]
    \[
        \langle \bm{q}_{n, i}^{(t + 1)}, \bm{k}_{n, j}^{(t + 1)} \rangle \le \langle \bm{q}_{n, i}^{(t)}, \bm{k}_{n, j}^{(t)} \rangle
    \]
    \[
        \Lambda_{n, \pm, j}^{(t + 1)} \ge \log \Big( \exp(\Lambda_{n, \pm, j}^{(T_1)}) + \frac{\eta^2 C_8 \Vert \bm{\mu} \Vert_2^4 \Vert \bm{w}_O \Vert_2^2 d_h^{\frac{1}{2}}}{N \big( \log (6N^2M^2 / \delta) \big)^{2} } \cdot (t - T_1)(t - T_1 + 1) \Big)
    \]
    \[
        \Lambda_{n, i, \pm, j}^{(t + 1)}  \ge \log \Big( \exp(\Lambda_{n, i, \pm, j}^{(T_1)}) + \frac{\eta^2 C_8 \sigma_p^2 d \Vert \bm{\mu} \Vert_2^2 \Vert \bm{w}_O \Vert_2^2 d_h^{\frac{1}{2}} }{N \big( \log (6N^2M^2 / \delta) \big)^{2}} \cdot (t - T_1)(t - T_1 + 1) \Big)
    \]
    for \(i, j \in [M] \backslash \{1\}, n \in [N] \).

    \item \( \mathcal{E}(t): \)
    \[
        \vert \langle \bm{q}_\pm^{(t)}, \bm{k}_\pm^{(t)} \rangle \vert, \vert \langle \bm{q}_\pm^{(t)}, \bm{k}_{n, j}^{(t)} \rangle \vert, \vert \langle \bm{q}_{n, i}^{(t)}, \bm{k}_\pm^{(t)} \rangle \vert, \vert \langle \bm{q}_{n, i}^{(t)}, \bm{k}_{n, j}^{(t)} \rangle \vert \le \log(d_h^{\frac{1}{2}})
    \]
    \[
        \vert \langle \bm{q}_\pm^{(t)}, \bm{k}_\mp^{(t)} \rangle \vert, \vert \langle \bm{q}_{n, i}^{(t)}, \bm{k}_{\overline{n}, j}^{(t)} \rangle \vert = o(1)
    \]
    for \(i, j \in [M] \backslash \{1\}, n, \overline{n} \in [N], n \ne \overline{n} \).
\end{itemize}

By the results of Stage \uppercase\expandafter{\romannumeral1}, we know that \( \mathcal{B}(T_1) \), \( \mathcal{C}(T_1) \), \( \mathcal{E}(T_1) \) are true. To prove that \( \mathcal{B}(t) \), \( \mathcal{C}(t) \), \( \mathcal{D}(t) \) and \( \mathcal{E}(t) \) are true in stage 2, we will prove the following claims holds for \(t \in [T_1, T_2]\):
\begin{claim}
    \label{claim1}
    \( \mathcal{D}(T_1), \dots, \mathcal{D}(t-1) \Longrightarrow \mathcal{B}(t + 1) \) 
\end{claim}
\begin{claim}
    \label{claim2}
    \( \mathcal{B}(T_1), \dots, \mathcal{B}(t), \mathcal{C}(T_1), \dots, \mathcal{C}(t), \mathcal{D}(T_1), \dots, \mathcal{D}(t - 1) \Longrightarrow \mathcal{D}(t) \) 
\end{claim}
\begin{claim}
    \label{claim3}
    \( \mathcal{B}(T_1), \dots, \mathcal{B}(t), \mathcal{D}(T_1), \dots, \mathcal{D}(t - 1), \mathcal{E}(T_1), \dots, \mathcal{E}(t) \Longrightarrow \mathcal{C}(t + 1) \) 
\end{claim}
\begin{claim}
    \label{claim4}
    \( \mathcal{B}(T_1), \dots, \mathcal{B}(t), \mathcal{C}(T_1), \dots, \mathcal{C}(t), \mathcal{D}(T_1), \dots, \mathcal{D}(t - 1) \Longrightarrow \mathcal{E}(t + 1) \)
\end{claim}

\subsubsection{Proof of Claim \ref{claim1}}
\label{proof_claim1}

By the results of Stage \uppercase\expandafter{\romannumeral1}, we have
\[
    \vert \langle \bm{q}_\pm^{(T_1)}, \bm{k}_\pm^{(T_1)} \rangle \vert, \vert \langle \bm{q}_\pm^{(T_1)}, \bm{k}_{n, j}^{(T_1)} \rangle \vert, \vert \langle \bm{q}_{n, i}^{(T_1)}, \bm{k}_\pm^{(T_1)} \rangle \vert, \vert \langle \bm{q}_{n, i}^{(T_1)}, \bm{k}_{n, j}^{(T_1)} \rangle \vert = o(1)
\]
Assume that \( \mathcal{D}(T_1), \dots, \mathcal{D}(t-1) \) (\(t \in [T_1, T_2]\))  are true, then \( \langle \bm{q}_\pm^{(s)}, \bm{k}_\pm^{(s)} \rangle \), \( \langle \bm{q}_{n, i}^{(s)}, \bm{k}_\pm^{(s)} \rangle \) are monotonically non-decreasing and \( \langle \bm{q}_\pm^{(s)}, \bm{k}_{n, j}^{(s)} \rangle \), \( \langle \bm{q}_{n, i}^{(s)}, \bm{k}_{n, j}^{(s)} \rangle \) are monotonically non-increasing for \(s \in [T_1, t-1]\), 
so we have
\[
    \langle \bm{q}_\pm^{(s)}, \bm{k}_\pm^{(s)} \rangle, \langle \bm{q}_{n, i}^{(s)}, \bm{k}_\pm^{(s)} \rangle \ge - o(1),
\]
\[
    \langle \bm{q}_\pm^{(s)}, \bm{k}_{n, j}^{(s)} \rangle, \langle \bm{q}_{n, i}^{(s)}, \bm{k}_{n, j}^{(s)} \rangle \le o(1),
\]
for \(s \in [T_1, t]\). Further we have the lower bounds for the attention on signal \(\bm{\mu}_\pm\) as follows for \( s \in [T_1, t] \):
\begin{equation}
\begin{split}
    softmax(\langle \bm{q}_\pm^{(s)}, \bm{k}_\pm^{(s)} \rangle) &\ge \frac{\exp (-o(1))}{\exp (-o(1)) + (M - 1) \exp (o(1))} \\
    &= \frac{1}{1 + (M - 1) \exp (o(1))} \\
    &= \frac{1}{1 + (M - 1) + (M - 1) o(1)} \\
    &= \frac{1}{M} - o(1),
\end{split}
\end{equation}
where the second equality is by \( \exp(o(1)) = 1 + o(1) \). Similarly, we have \( softmax(\langle \bm{q}_{n, i}^{(s)}, \bm{k}_\pm^{(s)} \rangle) \ge \frac{1}{M} - o(1) \).  Plugging them in the update rule for \(\gamma_{V, +}^{(s)}\) showed in Lemma \ref{update_rule4V} and we have
\begin{equation}
\begin{split}
    & \gamma_{V, +}^{(s + 1)} - \gamma_{V, +}^{(s)} \\
    &= - \frac{\eta \Vert \bm{\mu} \Vert_2^2}{NM} \sum\limits_{n \in S_+} \ell_n^{\prime(t)} \big( \frac{ \exp ( \langle \bm{q}_+^{(s)}, \bm{k}_+^{(s)} \rangle )}{  \exp ( \langle \bm{q}_+^{(s)}, \bm{k}_+^{(s)} \rangle ) + \sum\limits_{k=2}^M  \exp ( \langle \bm{q}_+^{(s)}, \bm{k}_{n, k}^{(s)} \rangle )}  \\
    &+ \sum\limits_{j=2}^M \frac{ \exp ( \langle \bm{q}_{n, j}^{(s)}, \bm{k}_+^{(s)} \rangle )}{  \exp ( \langle \bm{q}_{n, j}^{(s)}, \bm{k}_+^{(s)} \rangle ) + \sum\limits_{k=2}^M  \exp ( \langle \bm{q}_{n, j}^{(s)}, \bm{k}_{n, k}^{(s)} \rangle )} \big) \\
    &= \frac{\eta \Vert \bm{\mu} \Vert_2^2}{NM} \cdot \frac{N}{4} \cdot (\frac{1}{2} - o(1)) \cdot M (\frac{1}{M} - o(1)) \\
    &\ge \frac{\eta \Vert \bm{\mu} \Vert_2^2}{10 M},
\end{split}
\end{equation}
for \(s \in [T_1, t]\).
Then by Definition \ref{def1} and taking a summation, we have
\begin{equation}
\begin{split}
    \label{V_p_bound2}
    V_+^{(t + 1)} &\ge V_+^{(T_1)} + (t - T_1 + 1) \frac{\eta \Vert \bm{\mu} \Vert_2^2}{10 M} \Vert \bm{w}_O \Vert_2^2 \\
    &\ge \eta C_3 \Vert \bm{\mu} \Vert_2^2 \Vert \bm{w}_O \Vert_2^2 (t - T_1 + 1),
\end{split}
\end{equation}
where the last inequality is by \(V_+^{(T_1)} \ge 0\) and \( M = \Theta(1) \). Similarly, we have 
\[
    V_-^{(t + 1)} \le - \eta C_3 \Vert \bm{\mu} \Vert_2^2 \Vert \bm{w}_O \Vert_2^2 (t - T_1 + 1).
\]

By \(\vert \rho_{V, n, i}^{(t+1)} - \rho_{V, n, i}^{(t)} \vert \le \frac{2 \eta \tilde{\sigma}_p^2 d}{N} \) in \eqref{rho_bound} and taking a summation, we have
\begin{equation}
\begin{split}
    \label{V_i_bound2}
    \vert V_{n, i}^{(t + 1)} \vert &\le \vert V_{n, i}^{(T_1)} \vert + (t - T_1 + 1) \frac{2 \eta \tilde{\sigma}_p^2 d}{N} \Vert \bm{w}_O \Vert_2^2 \le \vert V_{n, i}^{(T_1)} \vert + (t - T_1 + 1) \frac{2 \eta C_p^2 \sigma_p^2 d}{N} \Vert \bm{w}_O \Vert_2^2.
\end{split}
\end{equation}
Combining \eqref{V_p_bound2} and \eqref{V_i_bound2} we have
\begin{equation}
\begin{split}
    &V_+^{(t + 1)} - 3M \cdot \vert V_{n, i}^{(t + 1)} \vert \\
    &\ge V_+^{(T_1)} + (t - T_1 + 1) \frac{\eta \Vert \bm{\mu} \Vert_2^2}{10 M} \Vert \bm{w}_O \Vert_2^2 - 3M \cdot \big( \vert V_{n, i}^{(T_1)} \vert + (t - T_1 + 1) \frac{2 \eta C_p^2 \sigma_p^2 d}{N} \Vert \bm{w}_O \Vert_2^2 \big) \\
    &\ge V_+^{(T_1)} - 3M \cdot \vert V_{n, i}^{(T_1)} \vert + (t - T_1 + 1) \frac{\eta \big( N \Vert \bm{\mu} \Vert_2^2 - 60 M^2 C_p^2 \sigma_p^2 d \big)}{10NM} \\
    &\ge 0,
\end{split}
\end{equation}
where the last inequality is by \(V_+^{(T_1)} \ge 3M \cdot \vert V_{n, i}^{(T_1)} \vert\) and requires \(N \cdot \mathrm{SNR}^2 \ge 60M^2 C_p^2\). The proof for \(V_-^{(t + 1)} \le - 3 M \cdot \vert V_{n, i}^{(t + 1)} \vert\) is the same.

Next, we prove the upper bound for \(V_\pm\) and \(V_{n, i}\). Based on the upper bound of attention(<1) and \( - \ell_n^\prime \le 1 \), we have
\begin{equation}
\begin{split}
    \gamma_{V,+}^{(s+1)} &\le \gamma_{V,+}^{(s)} - \frac{\eta}{NM} \sum\limits_{n \in S_+} \ell_n^\prime (\theta (s)) ( \Vert \bm{\mu} \Vert_2^2 + \sum\limits_{j=2}^M \Vert \bm{\mu} \Vert_2^2 ) \\
    &\le \gamma_{V,+}^{(s)} + \frac{3\eta \Vert \bm{\mu} \Vert_2^2}{4} \\
    &\le \gamma_{V,+}^{(s)} + \eta C_4 \Vert \bm{\mu} \Vert_2^2
\end{split}
\end{equation}
Then we can get that
\begin{equation}
\begin{split}
    \vert V_+^{(t + 1)} \vert &\le V_+^{(T_1)} + (\gamma_{V,+}^{(t + 1)} - \gamma_{V,+}^{(T_1)}) \Vert \bm{w}_O \Vert_2^2 \\
    &\le V_+^{(T_1)} + \sum\limits_{s=T_1}^{t} \eta C_4 \Vert \bm{\mu} \Vert_2^2 \Vert \bm{w}_O \Vert_2^2 \\
    &\le O(d_h^{-\frac{1}{4}}) + \eta C_4 \Vert \bm{\mu} \Vert_2^2 \Vert \bm{w}_O \Vert_2^2 (t - T_1 + 1)
\end{split}
\end{equation}

where the first inequality is by the monotonicity of $\gamma_{V,+}$ and the definition of $V_+$, the last inequality is by the result of stage 1 where $V_+^{(T_1)} = O(d^{-1})$. Similarly, we have
\begin{equation}
    \vert V_-^{(t + 1)} \vert \le O(d_h^{-\frac{1}{4}}) + \eta C_4 \Vert \bm{\mu} \Vert_2^2 \Vert \bm{w}_O \Vert_2^2 (t - T_1 + 1)
\end{equation}
which completes the proof for the upper bound of $V_\pm$.

Expanding \eqref{V_i_bound2} yields
\begin{equation}
\begin{split}
    \vert V_{n, i}^{(t + 1)} \vert &\le \vert V_{n, i}^{(T_1)} \vert + \frac{2 \eta C_p^2 \sigma_p^2 d}{N} \cdot \Vert \bm{w}_O \Vert_2^2 (t - T_1 + 1) \\
    &\le O(d_h^{-\frac{1}{4}}) + \eta C_4 \Vert \bm{\mu} \Vert_2^2 \Vert \bm{w}_O \Vert_2^2 (t - T_1 + 1)
\end{split}
\end{equation}
where the last inequality is by the result of phase 1 where $\vert V_{n, i}^{(T_1)} \vert = O(d_h^{-\frac{1}{4}})$ and the condition that $N \cdot \mathrm{SNR}^2 > \Omega(1)$.

\subsubsection{Proof of Claim \ref{claim2}}
\label{claim2_proof}
By the results of \ref{lower_bound_of_qk}, we have the dynamic of \(\langle \bm{q}, \bm{k} \rangle\) as follows
\begin{equation}
\begin{split}
    \label{q_p_k_p_dy}
    &\langle \bm{q}_+^{(s+1)}, \bm{k}_+^{(s+1)} \rangle - \langle \bm{q}_+^{(s)}, \bm{k}_+^{(s)} \rangle \\
    &\ge \frac{\eta^2 C_6 \Vert \bm{\mu} \Vert_2^6 \Vert \bm{w}_O \Vert_2^2 \sigma_h^2 d_h (s - T_1)}{N} \cdot \frac{1}{\exp(\Lambda_{n, +, j}^{(s)})},
\end{split}
\end{equation}
\begin{equation}
\begin{split}
    &\langle \bm{q}_-^{(s+1)}, \bm{k}_-^{(s+1)} \rangle - \langle \bm{q}_-^{(s)}, \bm{k}_-^{(s)} \rangle \\
    &\ge \frac{\eta^2 C_6 \Vert \bm{\mu} \Vert_2^6 \Vert \bm{w}_O \Vert_2^2 \sigma_h^2 d_h (s - T_1)}{N} \cdot \frac{1}{\exp(\Lambda_{n, -, j}^{(s)})},
\end{split}
\end{equation}
\begin{equation}
\begin{split}
    \label{q_p_k_j_dy}
    &\langle \bm{q}_+^{(s+1)}, \bm{k}_{n, j}^{(s+1)} \rangle - \langle \bm{q}_+^{(s)}, \bm{k}_{n, j}^{(s)} \rangle \\
    &\le - \frac{\eta^2 C_6 \sigma_p^2 d \Vert \bm{\mu} \Vert_2^4 \Vert \bm{w}_O \Vert_2^2 \sigma_h^2 d_h (s - T_1)}{N} \cdot \frac{1}{\exp(\Lambda_{n, +, j}^{(s)})},
\end{split}
\end{equation}
\begin{equation}
\begin{split}
    &\langle \bm{q}_-^{(s+1)}, \bm{k}_{n, j}^{(s+1)} \rangle - \langle \bm{q}_-^{(s)}, \bm{k}_{n, j}^{(s)} \rangle \\
    &\le - \frac{\eta^2 C_6 \sigma_p^2 d \Vert \bm{\mu} \Vert_2^4 \Vert \bm{w}_O \Vert_2^2 \sigma_h^2 d_h (s - T_1)}{N} \cdot \frac{1}{\exp(\Lambda_{n, -, j}^{(s)})},
\end{split}
\end{equation}
\begin{equation}
\begin{split}
    \label{q_i_k_p_dy}
    &\langle \bm{q}_{n, i}^{(s+1)}, \bm{k}_+^{(s+1)} \rangle - \langle \bm{q}_{n, i}^{(s)}, \bm{k}_+^{(s)} \rangle \\
    &\ge \frac{\eta^2 C_6 \sigma_p^2 d \Vert \bm{\mu} \Vert_2^4 \Vert \bm{w}_O \Vert_2^2 \sigma_h^2 d_h (s - T_1)}{N} \cdot \frac{1}{\exp(\Lambda_{n, i, +, j}^{(s)})},
\end{split}
\end{equation}
\begin{equation}
\begin{split}
    &\langle \bm{q}_{n, i}^{(s+1)}, \bm{k}_-^{(s+1)} \rangle - \langle \bm{q}_{n, i}^{(s)}, \bm{k}_-^{(s)} \rangle \\
    &\ge \frac{\eta^2 C_6 \sigma_p^2 d \Vert \bm{\mu} \Vert_2^4 \Vert \bm{w}_O \Vert_2^2 \sigma_h^2 d_h (s - T_1)}{N} \cdot \frac{1}{\exp(\Lambda_{n, i, -, j}^{(s)})},
\end{split}
\end{equation}
\begin{equation}
\begin{split}
    \label{q_i_k_j_dy}
    &\langle \bm{q}_{n, i}^{(s+1)}, \bm{k}_{n, j}^{(s+1)} \rangle - \langle \bm{q}_{n, i}^{(s)}, \bm{k}_{n, j}^{(s)} \rangle \\
    &\le - \frac{\eta^2 C_6 \sigma_p^4 d^2 \Vert \bm{\mu} \Vert_2^2 \Vert \bm{w}_O \Vert_2^2 \sigma_h^2 d_h (s - T_1) }{N} \cdot \frac{1}{\exp(\Lambda_{n, i, \pm, j}^{(s)})}
\end{split}
\end{equation}
for \(s \in [T_1, t]\).
The seven equations above show that \(\langle \bm{q}_\pm^{(s)}, \bm{k}_\pm^{(s)} \rangle\), \(\langle \bm{q}_{n, i}^{(s)}, \bm{k}_\pm^{(s)} \rangle\) are monotonically increasing and \(\langle \bm{q}_\pm^{(s)}, \bm{k}_{n, j}^{(s)} \rangle\), \(\langle \bm{q}_{n, i}^{(s)}, \bm{k}_{n, j}^{(s)} \rangle\) are monotonically decreasing. 
Next, we provide the logarithmic increasing lower bounds of \(\Lambda_{n, \pm, j}^{(s + 1)}\) and \(\Lambda_{n, i, \pm, j}^{(s + 1)}\).

By \eqref{q_p_k_p_dy} and \eqref{q_p_k_j_dy}, we have
\begin{equation}
\begin{split}
    \Lambda_{n, +, j}^{(s+1)} - \Lambda_{n, +, j}^{(s)} &= (\langle \bm{q}_+^{(s+1)}, \bm{k}_+^{(s+1)} \rangle - \langle \bm{q}_+^{(s)}, \bm{k}_+^{(s)} \rangle) - (\langle \bm{q}_+^{(s+1)}, \bm{k}_{n, j}^{(s+1)} \rangle - \langle \bm{q}_+^{(s)}, \bm{k}_{n, j}^{(s)} \rangle) \\
    &\ge \frac{\eta^2 C_6 \Vert \bm{\mu} \Vert_2^6 \Vert \bm{w}_O \Vert_2^2 \sigma_h^2 d_h (s - T_1)}{N} \cdot \frac{1}{\exp(\Lambda_{n, +, j}^{(s)})} \\
    &+ \frac{\eta^2 C_6 \sigma_p^2 d \Vert \bm{\mu} \Vert_2^4 \Vert \bm{w}_O \Vert_2^2 \sigma_h^2 d_h (s - T_1)}{N} \cdot \frac{1}{\exp(\Lambda_{n, +, j}^{(s)})} \\
    &\ge \frac{\eta^2 C_7 \max \{ \Vert \bm{\mu} \Vert_2^2, \sigma_p^2 d \} \Vert \bm{\mu} \Vert_2^4 \Vert \bm{w}_O \Vert_2^2 \sigma_h^2 d_h (s - T_1)}{N} \cdot \frac{1}{\exp(\Lambda_{n, +, j}^{(s)})} \\
    &\ge \frac{\eta^2 C_7 \Vert \bm{\mu} \Vert_2^4 \Vert \bm{w}_O \Vert_2^2 d_h^{\frac{1}{2}} (s - T_1)}{N \big( \log (6N^2M^2 / \delta) \big)^{2} } \cdot \frac{1}{\exp(\Lambda_{n, +, j}^{(s)})},
\end{split}
\end{equation}
where the last inequality is by \( \sigma_h^2 \ge \min \{ \Vert \bm{\mu} \Vert_2^{-2}, ( \sigma_p^2 d )^{-1} \} d_h^{-\frac{1}{2}} (\log (6N^2M^2 / \delta))^{-2} \). 
Multiply both sides simultaneously by \(\exp(\Lambda_{n, +, j}^{(s)})\) and get
\begin{equation}
    \exp(\Lambda_{n, +, j}^{(s)}) \Big( \Lambda_{n, +, j}^{(s+1)} - \Lambda_{n, +, j}^{(s)} \Big) \ge \frac{\eta^2 C_7 \Vert \bm{\mu} \Vert_2^4 \Vert \bm{w}_O \Vert_2^2 d_h^{\frac{1}{2}} (s - T_1)}{N \big( \log (6N^2M^2 / \delta) \big)^{2} }.
\end{equation}
Taking a summation from \(T_1\) to \(t\) and get
\begin{equation}
\begin{split}
    &\sum\limits_{s = T_1}^t \exp(\Lambda_{n, +, j}^{(s)}) \Big( \Lambda_{n, +, j}^{(s+1)} - \Lambda_{n, +, j}^{(s)} \Big) \\
    &\ge \sum\limits_{s = T_1}^t \frac{\eta^2 C_7 \Vert \bm{\mu} \Vert_2^4 \Vert \bm{w}_O \Vert_2^2 d_h^{\frac{1}{2}} (s - T_1)}{N \big( \log (6N^2M^2 / \delta) \big)^{2} } \\
    &\ge \frac{\eta^2 C_8 \Vert \bm{\mu} \Vert_2^4 \Vert \bm{w}_O \Vert_2^2 d_h^{\frac{1}{2}}}{N \big( \log (6N^2M^2 / \delta) \big)^{2} } \cdot (t - T_1)(t - T_1 + 1).
\end{split}
\end{equation}
By the property that \(\Lambda_{n, +, j}^{(t)}\) is monotonically increasing, we have
\begin{equation}
\begin{split}
    &\int_{\Lambda_{n, +, j}^{(T_1)}}^{\Lambda_{n, +, j}^{(t + 1)}} \exp(x) dx \ge \sum\limits_{s = T_1}^t \exp(\Lambda_{n, +, j}^{(s)}) \Big( \Lambda_{n, +, j}^{(s+1)} - \Lambda_{n, +, j}^{(s)} \Big) \\
    &\ge \frac{\eta^2 C_8 \Vert \bm{\mu} \Vert_2^4 \Vert \bm{w}_O \Vert_2^2 d_h^{\frac{1}{2}}}{N \big( \log (6N^2M^2 / \delta) \big)^{2} } \cdot (t - T_1)(t - T_1 + 1).
\end{split}
\end{equation}
By \(\int_{\Lambda_{n, +, j}^{(T_1)}}^{\Lambda_{n, +, j}^{(t + 1)}} \exp(x) dx = \exp(\Lambda_{n, +, j}^{(t + 1)}) - \exp(\Lambda_{n, +, j}^{(T_1)}) \) we get
\begin{equation}
\begin{split}
    \Lambda_{n, +, j}^{(t + 1)} \ge \log\Big( \exp(\Lambda_{n, +, j}^{(T_1)}) + \frac{\eta^2 C_8 \Vert \bm{\mu} \Vert_2^4 \Vert \bm{w}_O \Vert_2^2 d_h^{\frac{1}{2}}}{N \big( \log (6N^2M^2 / \delta) \big)^{2} } \cdot (t - T_1)(t - T_1 + 1) \Big).
\end{split}
\end{equation}
Similarly, we have 
\begin{equation}
\begin{split}
    \Lambda_{n, -, j}^{(t + 1)} \ge \log\Big( \exp(\Lambda_{n, -, j}^{(T_1)}) + \frac{\eta^2 C_8 \Vert \bm{\mu} \Vert_2^4 \Vert \bm{w}_O \Vert_2^2 d_h^{\frac{1}{2}}}{N \big( \log (6N^2M^2 / \delta) \big)^{2} } \cdot (t - T_1)(t - T_1 + 1) \Big).
\end{split}
\end{equation}
By \eqref{q_i_k_p_dy} and \eqref{q_i_k_j_dy}, we have
\begin{equation}
\begin{split}
    \Lambda_{n, i, +, j}^{(s + 1)} - \Lambda_{n, i, +, j}^{(s)} &= (\langle \bm{q}_{n, i}^{(s+1)}, \bm{k}_+^{(s+1)} \rangle - \langle \bm{q}_{n, i}^{(s)}, \bm{k}_+^{(s)} \rangle) - (\langle \bm{q}_{n, i}^{(s+1)}, \bm{k}_{n, j}^{(s+1)} \rangle - \langle \bm{q}_{n, i}^{(s)}, \bm{k}_{n, j}^{(s)} \rangle) \\
    &\ge \frac{\eta^2 C_6 \sigma_p^2 d \Vert \bm{\mu} \Vert_2^4 \Vert \bm{w}_O \Vert_2^2 \sigma_h^2 d_h (s - T_1)}{N} \cdot \frac{1}{\exp(\Lambda_{n, i, +, j}^{(s)})} \\
    &+ \frac{\eta^2 C_6 \sigma_p^4 d^2 \Vert \bm{\mu} \Vert_2^2 \Vert \bm{w}_O \Vert_2^2 \sigma_h^2 d_h (s - T_1) }{N} \cdot \frac{1}{\exp(\Lambda_{n, i, +, j}^{(s)})} \\
    &\ge \frac{\eta^2 C_7 \max \{ \Vert \bm{\mu} \Vert_2^2, \sigma_p^2 d \} \sigma_p^2 d \Vert \bm{\mu} \Vert_2^2 \Vert \bm{w}_O \Vert_2^2 \sigma_h^2 d_h (s - T_1) }{N} \cdot \frac{1}{\exp(\Lambda_{n, i, +, j}^{(s)})} \\
    &\ge \frac{\eta^2 C_7 \sigma_p^2 d \Vert \bm{\mu} \Vert_2^2 \Vert \bm{w}_O \Vert_2^2 d_h^{\frac{1}{2}} (s - T_1) }{N \big( \log (6N^2M^2 / \delta) \big)^{2}} \cdot \frac{1}{\exp(\Lambda_{n, i, +, j}^{(s)})}.
\end{split}
\end{equation}
Then using the similar method as for \(\Lambda_{n, \pm, j}^{(t)}\), we get
\begin{equation}
    \Lambda_{n, i, \pm, j}^{(t + 1)} \ge \log\Big( \exp(\Lambda_{n, i, \pm, j}^{(T_1)}) + \frac{\eta^2 C_8 \sigma_p^2 d \Vert \bm{\mu} \Vert_2^2 \Vert \bm{w}_O \Vert_2^2 d_h^{\frac{1}{2}} }{N \big( \log (6N^2M^2 / \delta) \big)^{2}} \cdot (t - T_1)(t - T_1 + 1) \Big),
\end{equation}
which complete the proof. The proof for Claim \ref{claim3} is in Section \ref{proof_claim3}

\subsubsection{Proof of Claim \ref{claim4}}

By the results of \ref{upper_bound_of_qk}, we have

\begin{equation}
\begin{split}
    \label{q_p_k_p_dy_upper_bound}
    \langle \bm{q}_+^{(s+1)}, \bm{k}_+^{(s+1)} \rangle - \langle \bm{q}_+^{(s)}, \bm{k}_+^{(s)} \rangle \le \frac{\eta C_{10} \Vert \bm{\mu} \Vert_2^4 \sigma_h^2 d_h}{\exp ( \langle \bm{q}_+^{(s)}, \bm{k}_+^{(s)} \rangle)}
\end{split}
\end{equation}
for \(s \in [T_1, t]\).
Further we have
\begin{equation}
\begin{split}
    \exp(\langle \bm{q}_+^{(s+1)}, \bm{k}_+^{(s+1)} \rangle) &\le \exp \Big( \langle \bm{q}_+^{(s)}, \bm{k}_+^{(s)} \rangle + \frac{\eta C_{10} \Vert \bm{\mu} \Vert_2^4 \sigma_h^2 d_h}{\exp ( \langle \bm{q}_+^{(s)}, \bm{k}_+^{(s)} \rangle)} \Big) \\
    &= \exp \Big( \langle \bm{q}_+^{(s)}, \bm{k}_+^{(s)} \rangle \Big) \cdot \exp \Big( \frac{\eta C_{10} \Vert \bm{\mu} \Vert_2^4 \sigma_h^2 d_h}{\exp ( \langle \bm{q}_+^{(s)}, \bm{k}_+^{(s)} \rangle)} \Big) \\
    &\le C_{11} \exp \Big( \langle \bm{q}_+^{(s)}, \bm{k}_+^{(s)} \rangle \Big).
\end{split}
\end{equation}
For the last inequality, by \(\eta \le \widetilde{O} ( \min \{ \bm{\mu} \Vert_2^{-2}, ( \sigma_p^2 d )^{-1} \} \cdot d_h^{-\frac{1}{2}} )\), \(\sigma_h^2 \le \min \{ \Vert \bm{\mu} \Vert_2^{-2}, ( \sigma_p^2 d )^{-1} \} d_h^{-\frac{1}{2}} (\log (6N^2M^2 / \delta))^{-\frac{3}{2}}\), \(\langle \bm{q}_+^{(T_1)}, \bm{k}_+^{(T_1)} \rangle = o(1)\) and the monotonicity of \(\langle \bm{q}_+^{(s)}, \bm{k}_+^{(s)} \rangle\) for \(s \in [T_1, t]\), we have \(\exp \Big( \frac{\eta C_{10} \Vert \bm{\mu} \Vert_2^4 \sigma_h^2 d_h}{\exp ( \langle \bm{q}_+^{(t)}, \bm{k}_+^{(t)} \rangle)} \Big) \le \exp(o(1)) \le C_{11}\). Multiplying both sides by \(\Big( \langle \bm{q}_+^{(s+1)}, \bm{k}_+^{(s+1)} \rangle - \langle \bm{q}_+^{(s)}, \bm{k}_+^{(s)} \rangle \Big)\) simultaneously gives
\begin{equation}
\begin{split}
    &\exp(\langle \bm{q}_+^{(s+1)}, \bm{k}_+^{(s+1)} \rangle) \Big( \langle \bm{q}_+^{(s+1)}, \bm{k}_+^{(s+1)} \rangle - \langle \bm{q}_+^{(s)}, \bm{k}_+^{(s)} \rangle \Big) \\
    &\le C_{11} \exp \Big( \langle \bm{q}_+^{(s)}, \bm{k}_+^{(s)} \rangle \Big) \cdot \Big( \langle \bm{q}_+^{(s+1)}, \bm{k}_+^{(s+1)} \rangle - \langle \bm{q}_+^{(s)}, \bm{k}_+^{(s)} \rangle \Big) \\
    &\le \eta C_{12} \Vert \bm{\mu} \Vert_2^4 \sigma_h^2 d_h,
\end{split}
\end{equation}
where the last inequality is by plugging \eqref{q_p_k_p_dy_upper_bound}. Taking a summation we have
\begin{equation}
\begin{split}
    &\int_{\langle \bm{q}_+^{(T_1)}, \bm{k}_+^{(T_1)} \rangle}^{\langle \bm{q}_+^{(t + 1)}, \bm{k}_+^{(t + 1)} \rangle} \exp(x) dx \\
    &\le \sum\limits_{s = T_1}^{t} \exp(\langle \bm{q}_+^{(s+1)}, \bm{k}_+^{(s+1)} \rangle) \Big( \langle \bm{q}_+^{(s+1)}, \bm{k}_+^{(s+1)} \rangle - \langle \bm{q}_+^{(s)}, \bm{k}_+^{(s)} \rangle \Big) \\
    &\le \sum\limits_{s = T_1}^{t} \eta C_{12} \Vert \bm{\mu} \Vert_2^4 \sigma_h^2 d_h \\
    &\le T_2 \cdot \eta C_{12} \Vert \bm{\mu} \Vert_2^4 \sigma_h^2 d_h \\
    &\le \frac{d_h^{\frac{1}{2}}}{\log (6N^2M^2 / \delta)^2},
\end{split}
\end{equation}
where the first inequality is due to \(\langle \bm{q}_+^{(s)}, \bm{k}_+^{(s)} \rangle\) is monotone increasing, the last inequality is by \(T_2 = \Theta(\eta^{-1} \Vert \bm{\mu} \Vert_2^{-2} \Vert \bm{w}_O \Vert_2^{-2} \log (6N^2M^2 / \delta)^{-1} ) \) and \( \sigma_h^2 \le \min \{ \Vert \bm{\mu} \Vert_2^{-2}, ( \sigma_p^2 d )^{-1} \} d_h^{-\frac{1}{2}} (\log (6N^2M^2 / \delta))^{-\frac{3}{2}}\). By \( \int_{\langle \bm{q}_+^{(T_1)}, \bm{k}_+^{(T_1)} \rangle}^{\langle \bm{q}_+^{(t + 1)}, \bm{k}_+^{(t + 1)} \rangle} \exp(x) dx = \exp(\langle \bm{q}_+^{(t + 1)}, \bm{k}_+^{(t + 1)} \rangle) - \exp(\langle \bm{q}_+^{(T_1)}, \bm{k}_+^{(T_1)} \rangle) \), we have
\begin{equation}
\begin{split}
    \langle \bm{q}_+^{(t + 1)}, \bm{k}_+^{(t + 1)} \rangle \le \log \Big( \langle \bm{q}_+^{(T_1)}, \bm{k}_+^{(T_1)} \rangle + \frac{d_h^{\frac{1}{2}}}{\log (6N^2M^2 / \delta)} \Big) \le \log \Big( d_h^{\frac{1}{2}} \Big),
\end{split}
\end{equation}
By the results of \ref{upper_bound_of_qk}, we also have

\begin{equation}
\begin{split}
    &\langle \bm{q}_-^{(s+1)}, \bm{k}_-^{(s+1)} \rangle - \langle \bm{q}_-^{(s)}, \bm{k}_-^{(s)} \rangle \le \frac{\eta C_{10} \Vert \bm{\mu} \Vert_2^4 \sigma_h^2 d_h}{\exp ( \langle \bm{q}_-^{(s)}, \bm{k}_-^{(s)} \rangle)}.
\end{split}
\end{equation}

\begin{equation}
\begin{split}
    &\langle \bm{q}_\pm^{(s+1)}, \bm{k}_{n, j}^{(s+1)} \rangle - \langle \bm{q}_\pm^{(s)}, \bm{k}_{n, j}^{(s)} \rangle \ge - \frac{\eta C_{10} \sigma_p^2 d \Vert \bm{\mu} \Vert_2^2 \sigma_h^2 d_h }{N} \cdot \exp (\langle \bm{q}_\pm^{(s)}, \bm{k}_{n, j}^{(s)} \rangle).
\end{split}
\end{equation}

\begin{equation}
\begin{split}
    &\langle \bm{q}_{n, i}^{(s+1)}, \bm{k}_\pm^{(s+1)} \rangle - \langle \bm{q}_{n, i}^{(s)}, \bm{k}_\pm^{(s)} \rangle \le \frac{\eta C_{10} \sigma_p^2 d \Vert \bm{\mu} \Vert_2^2 \sigma_h^2 d_h}{N \exp (\langle \bm{q}_{n, i}^{(s)}, \bm{k}_\pm^{(s)} \rangle)}.
\end{split}
\end{equation}

\begin{equation}
\begin{split}
    &\langle \bm{q}_{n, i}^{(s+1)}, \bm{k}_{n, j}^{(s+1)} \rangle - \langle \bm{q}_{n, i}^{(s)}, \bm{k}_{n, j}^{(s)} \rangle \ge - \frac{\eta C_{10} \sigma_p^4 d^2 \sigma_h^2 d_h }{N} \cdot \exp (\langle \bm{q}_{n, i}^{(s)}, \bm{k}_{n, j}^{(s)} \rangle).
\end{split}
\end{equation}

Then using the similar method as for \(\langle \bm{q}_+^{(t + 1)}, \bm{k}_+^{(t + 1)} \rangle\), we get
\[
    \langle \bm{q}_-^{(t + 1)}, \bm{k}_-^{(t + 1)} \rangle \le \log \Big( d_h^{\frac{1}{2}} \Big),
\]
\[
    \langle \bm{q}_\pm^{(t + 1)}, \bm{k}_{n, j}^{(t + 1)} \rangle \ge - \log \Big( d_h^{\frac{1}{2}} \Big),
\]
\[
    \langle \bm{q}_{n, i}^{(t + 1)}, \bm{k}_\pm^{(t + 1)} \rangle \le \log \Big( d_h^{\frac{1}{2}} \Big),
\]
\[
    \langle \bm{q}_{n, i}^{(t + 1)}, \bm{k}_{n, j}^{(t + 1)} \rangle \ge - \log \Big( d_h^{\frac{1}{2}} \Big).
\]

Next we provide the upper bound for \(\vert \langle \bm{q}_\pm^{(t + 1)}, \bm{k}_\mp^{(t + 1)} \rangle \vert, \vert \langle \bm{q}_{n, i}^{(t + 1)}, \bm{k}_{n^\prime, j}^{(t + 1)} \rangle \vert\). By the results of \ref{sum_alpha_beta}, we have
\begin{equation}
\begin{split}
    \sum\limits_{s = T_1}^{t}\vert \alpha_{+, +}^{(s)} \vert, \sum\limits_{s = T_1}^{t}\vert \alpha_{-, -}^{(s)} \vert, \sum\limits_{s = T_1}^{t}\vert \beta_{+, +}^{(s)} \vert, \sum\limits_{s = T_1}^{t}\vert \beta_{-, -}^{(s)} \vert, \sum\limits_{s = T_1}^{t}\vert \beta_{n, i, +}^{(s)} \vert, \sum\limits_{s = T_1}^{t}\vert \beta_{n, i, -}^{(s)} \vert = O \Big( N^{\frac{1}{2}} d_h^{-\frac{1}{4}} \Big),
\end{split}
\end{equation}
for \(i \in [M] \backslash \{1\}, n \in S_\pm\).
\begin{equation}
\begin{split}
    \sum\limits_{s = T_1}^{t}\vert \alpha_{n, +, i}^{(s)} \vert, \sum\limits_{s = T_1}^{t}\vert \alpha_{n, -, i}^{(s)} \vert = O \Big( N^{-\frac{1}{2}} d_h^{-\frac{1}{4}} \Big),
\end{split}
\end{equation}
for \(i \in [M] \backslash \{1\}, n \in S_\pm\).
\begin{equation}
\begin{split}
    &\sum\limits_{s = T_1}^{t}\vert \beta_{n, +, i}^{(s)} \vert, \sum\limits_{s = T_1}^{t}\vert \beta_{n, -, i}^{(s)} \vert = O \Big(\mathrm{SNR} \cdot N^{-\frac{1}{2}} d_h^{-\frac{1}{4}} \Big)
\end{split}
\end{equation}
for \(i \in [M] \backslash \{1\}, n \in S_\pm\).
\begin{equation}
\begin{split}
    &\sum\limits_{s = T_1}^{t}\vert \alpha_{n, i, +}^{(s)} \vert, \sum\limits_{s = T_1}^{t}\vert \alpha_{n, i, -}^{(s)} \vert, \sum\limits_{s = T_1}^{t}\vert \alpha_{n, i, n, j}^{(s)} \vert, \sum\limits_{s = T_1}^{t}\vert \beta_{n, j, n, i}^{(s)} \vert = O \Big( d_h^{-\frac{1}{4}} \Big)
\end{split}
\end{equation}
for \(i, j \in [M] \backslash \{1\}, n \in S_\pm\).
\begin{equation}
\begin{split}
    \sum\limits_{s = T_1}^{t}\vert \alpha_{n, i, n^\prime, j}^{(s)} \vert, \sum\limits_{s = T_1}^{t}\vert \beta_{n, j, n^\prime, i}^{(s)} \vert = O \Big( d^{-\frac{1}{2}} d_h^{-\frac{1}{4}} \log (6N^2M^2 / \delta) \Big)
\end{split}
\end{equation}
for \(i, j \in [M] \backslash \{1\}, n, n^\prime \in [N], n \ne n^\prime\).
Plugging these into the update rule of \( \langle \bm{q}_\pm^{(t)}, \bm{k}_\mp^{(t)} \rangle,  \langle \bm{q}_{n, i}^{(t)}, \bm{k}_{\overline{n}, j}^{(t)} \rangle \) and assume that propositions \( \mathcal{C}(T_1), \dots, \mathcal{C}(t) \) hold, we have
\begin{equation}
\begin{split}
    &\vert \langle \bm{q}_+^{(t + 1)}, \bm{k}_-^{(t + 1)} \rangle \vert \le \vert \langle \bm{q}_+^{(T_1)}, \bm{k}_-^{(T_1)} \rangle \vert + \sum\limits_{s = T_1}^{t} \vert \langle \bm{q}_+^{(s+1)}, \bm{k}_-^{(s+1)} \rangle - \langle \bm{q}_+^{(s)}, \bm{k}_-^{(s)} \rangle \vert \\
    &\le \vert \langle \bm{q}_+^{(T_1)}, \bm{k}_-^{(T_1)} \rangle \vert \\
    &+ \sum\limits_{s = T_1}^{t} \Big\vert \alpha_{+, +}^{(s)} \langle \bm{k}_+^{(s)}, \bm{k}_-^{(s)} \rangle + \sum\limits_{n \in S_+} \sum\limits_{i=2}^M \alpha_{n, +, i}^{(s)} \langle \bm{k}_{n, i}^{(s)}, \bm{k}_-^{(s)} \rangle \\
    &+ \beta_{-, -}^{(s)} \langle \bm{q}_+^{(s)}, \bm{q}_-^{(s)} \rangle + \sum\limits_{n \in S_-} \sum\limits_{i=2}^M \beta_{n, -, i}^{(s)} \langle \bm{q}_{n, i}^{(s)}, \bm{q}_+^{(s)} \rangle \\
    &+ \Big( \alpha_{+, +}^{(s)} \bm{k}_+^{(s)} + \sum\limits_{n \in S_+} \sum\limits_{i=2}^M \alpha_{n, +, i}^{(s)} \bm{k}_{n, i}^{(s)} \Big) \\
    &\cdot \Big( \beta_{-, -}^{(s)} \bm{q}_-^{(s)\top} + \sum\limits_{n \in S_-} \sum\limits_{i=2}^M \beta_{n, -, i}^{(s)} \bm{q}_{n, i}^{(s)\top} \Big) \Big\vert \\
    &\le \vert \langle \bm{q}_+^{(T_1)}, \bm{k}_-^{(T_1)} \rangle \vert \\
    &+ \sum\limits_{s = T_1}^{t} \vert \alpha_{+, +}^{(s)} \vert \vert \langle \bm{k}_+^{(s)}, \bm{k}_-^{(s)} \rangle \vert + \sum\limits_{n \in S_+} \sum\limits_{i=2}^M \sum\limits_{s = T_1}^{t} \vert \alpha_{n, +, i}^{(s)} \vert \vert \langle \bm{k}_{n, i}^{(s)}, \bm{k}_-^{(s)} \rangle \vert \\
    &+ \sum\limits_{s = T_1}^{t} \vert \beta_{-, -}^{(s)} \vert \vert \langle \bm{q}_+^{(s)}, \bm{q}_-^{(s)} \rangle \vert + \sum\limits_{n \in S_-} \sum\limits_{i=2}^M \sum\limits_{s = T_1}^{t} \vert \beta_{n, -, i}^{(s)} \vert \vert \langle \bm{q}_{n, i}^{(s)}, \bm{q}_+^{(s)} \rangle \vert \\
    &+ \{lower \ order \ term\} \\
    &= \vert \langle \bm{q}_+^{(T_1)}, \bm{k}_-^{(T_1)} \rangle \vert \\
    &+ O \Big( N^{\frac{1}{2}} d_h^{-\frac{1}{4}} \Big) \cdot o(1) + N \cdot M \cdot O \Big( N^{-\frac{1}{2}} d_h^{-\frac{1}{4}} \Big) \cdot o(1) \\
    &+ O \Big( N^{\frac{1}{2}} d_h^{-\frac{1}{4}} \Big) \cdot o(1) + N \cdot M \cdot O \Big(\mathrm{SNR} \cdot N^{-\frac{1}{2}} d_h^{-\frac{1}{4}} \Big) \cdot o(1) \\
    &=\vert \langle \bm{q}_+^{(T_1)}, \bm{k}_-^{(T_1)} \rangle \vert + o \Big( N^{\frac{1}{2}} d_h^{-\frac{1}{4}} \Big) + o \Big(\mathrm{SNR} \cdot N^{\frac{1}{2}} d_h^{-\frac{1}{4}} \Big) \\
    &= o(1),
\end{split}
\end{equation}
where the first inequality is by triangle inequality, the second inequality is by \eqref{q_p_k_m_dy}, the last equality is by \(\vert \langle \bm{q}_+^{(T_1)}, \bm{k}_-^{(T_1)} \rangle \vert = o(1)\) and \(d_h = \widetilde{\Omega} \Big( \max \{\mathrm{SNR}^4, \mathrm{SNR}^{-4}\} N^2 \epsilon^{-2} \Big)\). Similarly we have \(\vert \langle \bm{q}_-^{(t + 1)}, \bm{k}_+^{(t + 1)} \rangle \vert = o(1)\).
\begin{equation}
\begin{split}
    &\vert \langle \bm{q}_{n, i}^{(t+1)}, \bm{k}_{\overline{n}, j}^{(t+1)} \rangle \vert \le \vert \langle \bm{q}_{n, i}^{(T_1)}, \bm{k}_{\overline{n}, j}^{(T_1)} \rangle \vert + \sum\limits_{s = T_1}^{t} \vert \langle \bm{q}_{n, i}^{(s+1)}, \bm{k}_{\overline{n}, j}^{(s+1)} \rangle - \langle \bm{q}_{n, i}^{(s)}, \bm{k}_{\overline{n}, j}^{(s)} \rangle \vert \\
    &\le \vert \langle \bm{q}_{n, i}^{(T_1)}, \bm{k}_{\overline{n}, j}^{(T_1)} \rangle \vert \\
    &+ \sum\limits_{s = T_1}^{t} \Big\vert \alpha_{n, i, +}^{(s)} \langle \bm{k}_+^{(s)}, \bm{k}_{\overline{n}, j}^{(s)} \rangle + \alpha_{n, i, -}^{(s)} \langle \bm{k}_-^{(s)}, \bm{k}_{\overline{n}, j}^{(s)} \rangle + \sum\limits_{n^\prime =1}^N \sum\limits_{l=2}^M \alpha_{n, i, n^\prime, l}^{(s)} \langle \bm{k}_{n^\prime, l}^{(s)}, \bm{k}_{\overline{n}, j}^{(s)} \rangle \\
    &+ \beta_{\overline{n}, j, +}^{(s)} \langle \bm{q}_+^{(s)}, \bm{q}_{n, i}^{(s)} \rangle + \beta_{\overline{n}, j, -}^{(s)} \langle \bm{q}_-^{(s)}, \bm{q}_{n, i}^{(s)} \rangle + \sum\limits_{n^\prime = 1}^N \sum\limits_{l=2}^M \beta_{\overline{n}, j, n^\prime, l}^{(s)} \langle \bm{q}_{n^\prime, l}^{(s)}, \bm{q}_{n, i}^{(s)} \rangle \\
    &+ \Big( \alpha_{n, i, +}^{(s)} \bm{k}_+^{(s)} + \alpha_{n, i, -}^{(s)} \bm{k}_-^{(s)} + \sum\limits_{n^\prime =1}^N \sum\limits_{l=2}^M \alpha_{n, i, n^\prime, l}^{(s)} \bm{k}_{n^\prime, l}^{(s)} \Big) \\
    &\cdot \Big( \beta_{\overline{n}, j, +}^{(s)} \bm{q}_+^{(s)\top} + \beta_{\overline{n}, j, -}^{(s)} \bm{q}_-^{(s)\top} + \sum\limits_{n^\prime = 1}^N \sum\limits_{l=2}^M \beta_{\overline{n}, j, n^\prime, l}^{(s)} \bm{q}_{n^\prime, l}^{(s)\top} \Big) \Big\vert \\
    &\le \vert \langle \bm{q}_{n, i}^{(T_1)}, \bm{k}_{\overline{n}, j}^{(T_1)} \rangle \vert \\
    &+ \sum\limits_{s = T_1}^{t} \vert \alpha_{n, i, +}^{(s)} \vert \vert \langle \bm{k}_+^{(s)}, \bm{k}_{\overline{n}, j}^{(s)} \rangle \vert + \sum\limits_{s = T_1}^{t} \vert \alpha_{n, i, -}^{(s)} \vert \vert \langle \bm{k}_-^{(s)}, \bm{k}_{\overline{n}, j}^{(s)} \rangle \vert \\
    &+ \sum\limits_{s = T_1}^{t} \vert \alpha_{n, i, \overline{n}, j}^{(s)} \vert \Vert \bm{k}_{\overline{n}, j}^{(s)} \Vert_2^2 + \sum\limits_{l=2}^M \sum\limits_{s = T_1}^{t} \vert \alpha_{n, i, n, l}^{(s)} \vert \vert \langle \bm{k}_{n, l}^{(s)}, \bm{k}_{\overline{n}, j}^{(s)} \rangle \vert \\
    &+ \sum\limits_{n^\prime \ne n \wedge (l \ne j \vee n^\prime \ne \overline{n})} \sum\limits_{s = T_1}^{t} \vert \alpha_{n, i, n^\prime, l}^{(s)} \vert \vert \langle \bm{k}_{n^\prime, l}^{(s)}, \bm{k}_{\overline{n}, j}^{(s)} \rangle \vert \\
    &+ \sum\limits_{s = T_1}^{t} \vert \beta_{\overline{n}, j, +}^{(s)} \vert \vert \langle \bm{q}_+^{(s)}, \bm{q}_{n, i}^{(s)} \rangle \vert + \sum\limits_{s = T_1}^{t} \vert \beta_{\overline{n}, j, -}^{(s)} \vert \vert \langle \bm{q}_-^{(s)}, \bm{q}_{n, i}^{(s)} \rangle \vert \\
    &+ \sum\limits_{s = T_1}^{t} \vert \beta_{\overline{n}, j, n, i}^{(s)} \vert \Vert \bm{q}_{n, i}^{(s)} \Vert_2^2 + \sum\limits_{l=2}^M \sum\limits_{s = T_1}^{t} \vert \beta_{\overline{n}, j, \overline{n}, l}^{(s)} \vert \vert \langle \bm{q}_{\overline{n}, l}^{(s)}, \bm{q}_{n, i}^{(s)} \rangle \vert \\
    &+ \sum\limits_{n^\prime \ne \overline{n} \wedge (l \ne i \vee n^\prime \ne n)} \sum\limits_{s = T_1}^{t} \vert \beta_{\overline{n}, j, n^\prime, l}^{(s)} \vert \vert \langle \bm{q}_{n^\prime, l}^{(s)}, \bm{q}_{n, i}^{(s)} \rangle \vert \\
    &+ \{lower \ order \ term\} \\
    &= \vert \langle \bm{q}_{n, i}^{(T_1)}, \bm{k}_{\overline{n}, j}^{(T_1)} \rangle \vert \\
    &+ O \Big( d_h^{-\frac{1}{4}} \Big) \cdot o(1) + O \Big( d^{-\frac{1}{2}} d_h^{-\frac{1}{4}} \log (6N^2M^2 / \delta) \Big) \cdot \Theta ( \sigma_p^2 \sigma_h^2 d d_h ) \\
    &+ M \cdot O \Big( d_h^{-\frac{1}{4}} \Big) \cdot o(1) + N \cdot M \cdot O \Big( d^{-\frac{1}{2}} d_h^{-\frac{1}{4}} \log (6N^2M^2 / \delta) \Big) \cdot o(1) \\
    &+ O \Big( N^{\frac{1}{2}} d_h^{-\frac{1}{4}} \Big) \cdot o(1) \\
    &= \vert \langle \bm{q}_{n, i}^{(T_1)}, \bm{k}_{\overline{n}, j}^{(T_1)} \rangle \vert + o \Big( d_h^{-\frac{1}{4}} \Big) \\
    &+ O \Big( d^{-\frac{1}{2}} d_h^{\frac{1}{4}} \Big) + o \Big( N d^{-\frac{1}{2}} d_h^{-\frac{1}{4}} \log (6N^2M^2 / \delta) \Big) \\
    &= o(1),
\end{split}
\end{equation}
where the first inequality is by triangle inequality, the second inequality is by \eqref{q_i_k_j_ne_dy}, the second equality is by \( \sigma_h^2 \le \min \{ \Vert \bm{\mu} \Vert_2^{-2}, ( \sigma_p^2 d )^{-1} \} \cdot d_h^{-\frac{1}{2}} \cdot \big( \log ( 6N^2M^2 / \delta ) \big)^{-\frac{3}{2}}\), the last equality is by \(\vert \langle \bm{q}_{n, i}^{(T_1)}, \bm{k}_{\overline{n}, j}^{(T_1)} \rangle \vert = o(1)\), \(d = \widetilde{\Omega} \Big( \epsilon^{-2} N^2 d_h \Big) \) and \(d_h = \widetilde{\Omega} \Big( \max \{\mathrm{SNR}^4, \mathrm{SNR}^{-4}\} N^2 \epsilon^{-2} \Big)\).

\subsection{Stage \uppercase\expandafter{\romannumeral3}}
\label{stage3}
In Stage \uppercase\expandafter{\romannumeral3}, the outputs of ViT grow up and the loss derivatives are no longer at \(o(1)\). We will carefully compute the growth rate of \(V_\pm\) and \(V_{n, i}\) while keeping monitoring the monotonicity of \(\langle \bm{q}, \bm{k} \rangle\). 
By substituting \(t = T_2 = \Theta \Big( \frac{1}{\eta \Vert \bm{\mu} \Vert_2^2 \Vert \bm{w}_O \Vert_2^2} \Big) \) into propositions \( \mathcal{B}(t) \), \( \mathcal{C}(t) \), \( \mathcal{D}(t) \), \( \mathcal{E}(t) \) in Stage \uppercase\expandafter{\romannumeral2}, we have the following conditions at the beginning of stage \uppercase\expandafter{\romannumeral3}
\[
    \vert V_+^{(T_2)} \vert, \vert V_-^{(T_2)} \vert, \vert V_{n, i}^{(T_2)} \vert = o (1),
\]
\[
    V_+^{(T_2)} \ge 3 M \cdot \vert V_{n, i}^{(T_2)} \vert,
\]
\[
    V_-^{(T_2)} \le - 3 M \cdot \vert V_{n, i}^{(T_2)} \vert,
\]
\[
    \Vert \bm{q}_\pm^{(T_2)} \Vert_2^2, \Vert \bm{k}_\pm^{(T_2)} \Vert_2^2 = \Theta \Big( \Vert \bm{\mu} \Vert_2^2 \sigma_h^2 d_h \Big),
\]
\[
    \Vert \bm{q}_{n, i}^{(T_2)} \Vert_2^2, \Vert \bm{k}_{n, i}^{(T_2)} \Vert_2^2 = \Theta \Big( \sigma_p^2 \sigma_h^2 d d_h \Big),
\]
\[
    \vert \langle \bm{q}_+^{(T_2)}, \bm{q}_-^{(T_2)} \rangle \vert, \vert \langle \bm{q}_\pm^{(T_2)}, \bm{q}_{n, i}^{(T_2)} \rangle \vert, \vert \langle \bm{q}_{n, i}^{(T_2)}, \bm{q}_{n^\prime, j}^{(T_2)} \rangle \vert = o(1),
\]
\[
    \vert \langle \bm{k}_+^{(T_2)}, \bm{k}_-^{(T_2)} \rangle \vert, \vert \langle \bm{k}_\pm^{(T_2)}, \bm{k}_{n, i}^{(T_2)} \rangle \vert, \vert \langle \bm{k}_{n, i}^{(T_2)}, \bm{k}_{n^\prime, j}^{(T_2)} \rangle \vert = o(1),
\]
for \(i, j \in [M] \backslash \{1\}, n, n^\prime \in [N], i \ne j \ or \ n \ne n^\prime \).

\[
    \Lambda_{n, \pm, j}^{(T_2)} \ge \log \Big( \exp(\Lambda_{n, \pm, j}^{(T_1)}) + \Theta \Big( \frac{ d_h^{\frac{1}{2}}}{N \big( \log (6N^2M^2 / \delta) \big)^{3} } \Big) \Big)
\]
\[
    \Lambda_{n, i, \pm, j}^{(T_2)}  \ge \log \Big( \exp(\Lambda_{n, i, \pm, j}^{(T_1)}) + \Theta \Big( \frac{ \sigma_p^2 d d_h^{\frac{1}{2}} }{N \Vert \bm{\mu} \Vert_2^2 \big( \log (6N^2M^2 / \delta) \big)^{3}} \Big) \Big)
\]
\[
    \vert \langle \bm{q}_\pm^{(T_2)}, \bm{k}_\pm^{(T_2)} \rangle \vert, \vert \langle \bm{q}_\pm^{(T_2)}, \bm{k}_{n, j}^{(T_2)} \rangle \vert, \vert \langle \bm{q}_{n, i}^{(T_2)}, \bm{k}_\pm^{(T_2)} \rangle \vert, \vert \langle \bm{q}_{n, i}^{(T_2)}, \bm{k}_{n, j}^{(T_2)} \rangle \vert \le \log(d_h^{\frac{1}{2}})
\]
\[
    \vert \langle \bm{q}_\pm^{(T_2)}, \bm{k}_\mp^{(T_2)} \rangle \vert, \vert \langle \bm{q}_{n, i}^{(T_2)}, \bm{k}_{\overline{n}, j}^{(T_2)} \rangle \vert = o(1)
\]
for \(i, j \in [M] \backslash \{1\}, n, \overline{n} \in [N], n \ne \overline{n} \).

Let \(T_3 = \Theta \Big( \frac{1}{\eta \epsilon \Vert \bm{\mu} \Vert_2^2 \Vert \bm{w}_O \Vert_2^2} \Big)\), 
Next we prove the following four propositions \( \mathcal{F}(t) \), \( \mathcal{G}(t) \), \( \mathcal{H}(t) \), \( \mathcal{I}(t) \) by induction on t for \(t \in [T_2, T_3]\):
\begin{itemize}
    \item \( \mathcal{F}(t): \)
    \[
        V_+^{(t)} \ge 3 M \cdot \vert V_{n, i}^{(t)} \vert,
    \]
    \[
        V_-^{(t)} \le - 3 M \cdot \vert V_{n, i}^{(t)} \vert,
    \]
    \[
        \vert V_{n, i}^{(t)} \vert = o (1), 
    \]
    \[
        \log \Big( \exp ( V_{+}^{(T_2)}) + \eta C_{17} \Vert \bm{\mu} \Vert_2^2 \Vert \bm{w}_O \Vert_2^2 (t - T_2) \Big) \le V_+^{(t)} \le 2 \log \big( O(\frac{1}{\epsilon}) \big), 
    \]
    \[
        - 2 \log \big( O(\frac{1}{\epsilon}) \big) \le V_-^{(t)} \le - \log \Big( \exp ( - V_{-}^{(T_2)}) + \eta C_{17} \Vert \bm{\mu} \Vert_2^2 \Vert \bm{w}_O \Vert_2^2 (t - T_2) \Big)
    \]
    for \(i \in [M] \backslash \{1\}, n \in [N]\).

    \item \( \mathcal{G}(t): \)
    \[
        \Vert \bm{q}_\pm^{(t)} \Vert_2^2, \Vert \bm{k}_\pm^{(t)} \Vert_2^2 = \Theta (\Vert \bm{\mu} \Vert_2^2 \sigma_h^2 d_h),
    \]
    \[
        \Vert \bm{q}_{n, i}^{(t)} \Vert_2^2, \Vert \bm{k}_{n, i}^{(t)} \Vert_2^2 = \Theta \Big( \sigma_p^2 \sigma_h^2 d d_h \Big),
    \]
    \[
        \vert \langle \bm{q}_+^{(t)}, \bm{q}_-^{(t)} \rangle \vert, \vert \langle \bm{q}_\pm^{(t)}, \bm{q}_{n, i}^{(t)} \rangle \vert, \vert \langle \bm{q}_{n, i}^{(t)}, \bm{q}_{n^\prime, j}^{(t)} \rangle \vert = o(1),
    \]
    \[
        \vert \langle \bm{k}_+^{(t)}, \bm{k}_-^{(t)} \rangle \vert, \vert \langle \bm{k}_\pm^{(t)}, \bm{k}_{n, i}^{(t)} \rangle \vert, \vert \langle \bm{k}_{n, i}^{(t)}, \bm{k}_{n^\prime, j}^{(t)} \rangle \vert = o(1)
    \]
    for \(i, j \in [M] \backslash \{1\}, n, n^\prime \in [N], i \ne j \ or \ n \ne n^\prime \).

    \item \( \mathcal{H}(t): \)
    \[
        \langle \bm{q}_\pm^{(t + 1)}, \bm{k}_\pm^{(t + 1)} \rangle \ge \langle \bm{q}_\pm^{(t)}, \bm{k}_\pm^{(t)} \rangle,
    \]
    \[
        \langle \bm{q}_{n, i}^{(t + 1)}, \bm{k}_\pm^{(t + 1)} \rangle \ge \langle \bm{q}_{n, i}^{(t)}, \bm{k}_\pm^{(t)} \rangle,
    \]
    \[
        \langle \bm{q}_\pm^{(t + 1)}, \bm{k}_{n, j}^{(t + 1)} \rangle \le \langle \bm{q}_\pm^{(t)}, \bm{k}_{n, j}^{(t)} \rangle,
    \]
    \[
        \langle \bm{q}_{n, i}^{(t + 1)}, \bm{k}_{n, j}^{(t + 1)} \rangle \le \langle \bm{q}_{n, i}^{(t)}, \bm{k}_{n, j}^{(t)} \rangle
    \]
    for \(i, j \in [M] \backslash \{1\}, n \in [N] \).

    \item \( \mathcal{I}(t): \)
    \[
        \vert \langle \bm{q}_\pm^{(t)}, \bm{k}_\pm^{(t)} \rangle \vert, \vert \langle \bm{q}_\pm^{(t)}, \bm{k}_{n, j}^{(t)} \rangle \vert, \vert \langle \bm{q}_{n, i}^{(t)}, \bm{k}_\pm^{(t)} \rangle \vert, \vert \langle \bm{q}_{n, i}^{(t)}, \bm{k}_{n, j}^{(t)} \rangle \vert \le \log ( \epsilon^{-1} d_h^{\frac{1}{2}} ),
    \]
    \[
        \vert \langle \bm{q}_\pm^{(t)}, \bm{k}_\mp^{(t)} \rangle \vert, \vert \langle \bm{q}_{n, i}^{(t)}, \bm{k}_{\overline{n}, j}^{(t)} \rangle \vert = o(1)
    \]
    for \(i, j \in [M] \backslash \{1\}, n, n^\prime \in [N], n \ne \overline{n} \).
\end{itemize}
By the results of Stage \uppercase\expandafter{\romannumeral2}, we know that \( \mathcal{F}(T_1) \), \( \mathcal{G}(T_2) \), \( \mathcal{I}(T_2) \) are true. To prove that \( \mathcal{F}(t) \), \( \mathcal{G}(t) \), \( \mathcal{H}(t) \) and \( \mathcal{I}(t) \) are true in stage 3, we will prove the following claims holds for \(t \in [T_2, T_3]\):
\begin{claim}
    \label{claim5}
    \( \mathcal{H}(T_2), \dots, \mathcal{H}(t - 1) \Longrightarrow \mathcal{F}(t + 1) \) 
\end{claim}
\begin{claim}
    \label{claim6}
    \( \mathcal{F}(t), \mathcal{G}(t), \mathcal{H}(T_2), \dots, \mathcal{H}(t - 1) \Longrightarrow \mathcal{H}(t) \) 
\end{claim}
\begin{claim}
    \label{claim7}
    \( \mathcal{F}(T_2), \dots, \mathcal{F}(t), \mathcal{H}(T_2), \dots, \mathcal{H}(t - 1), \mathcal{I}(T_2), \dots, \mathcal{I}(t) \Longrightarrow \mathcal{G}(t + 1) \) 
\end{claim}
\begin{claim}
    \label{claim8}
    \( \mathcal{F}(T_2), \dots, \mathcal{F}(t), \mathcal{G}(T_2), \dots, \mathcal{G}(t), \mathcal{H}(T_2), \dots, \mathcal{H}(t - 1) \Longrightarrow \mathcal{I}(t + 1) \)
\end{claim}

\subsubsection{Proof of Claim \ref{claim5}}
\label{proof_claim5}

The proofs for \(V_+^{(t)} \ge 3 M \cdot \vert V_{n, i}^{(t)} \vert\) and \(V_-^{(t)} \le - 3 M \cdot \vert V_{n, i}^{(t)} \vert\) are the same as for \ref{proof_claim1}. Based on \(\mathcal{H}(T_2), \dots, \mathcal{H}(t)\) where \(\langle \bm{q}_\pm^{(s)}, \bm{k}_\pm^{(s)} \rangle\) and \(\langle \bm{q}_{n, i}^{(s)}, \bm{k}_\pm^{(s)} \rangle\) are monotonically non-decreasing and \( \max\limits_j \langle \bm{q}_\pm^{(s)}, \bm{k}_{n, j}^{(s)} \rangle \), \( \max\limits_j \langle \bm{q}_{n, i}^{(s)}, \bm{k}_{n, j}^{(s)} \rangle \) are monotonically non-increasing for \(s \in [T_2, t-1]\), we have
\begin{equation}
    \label{lambda_pm_nj}
    \Lambda_{n, \pm, j}^{(s)} \ge \Lambda_{n, \pm, j}^{(T_2)} \ge \log \Big( \exp(\Lambda_{n, \pm, j}^{(T_1)}) + \Theta \Big( \frac{ d_h^{\frac{1}{2}}}{N \big( \log (6N^2M^2 / \delta) \big)^{3} } \Big) \Big),
\end{equation}
\begin{equation}
    \label{lambda_nij_pm}
    \Lambda_{n, i, \pm, j}^{(s)} \ge \Lambda_{n, i, \pm, j}^{(T_2)} \ge \log \Big( \exp(\Lambda_{n, i, \pm, j}^{(T_1)}) + \Theta \Big( \frac{ \sigma_p^2 d d_h^{\frac{1}{2}} }{N \Vert \bm{\mu} \Vert_2^2 \big( \log (6N^2M^2 / \delta) \big)^{3}} \Big) \Big)
\end{equation}
for \(i, j \in [M] \backslash \{1\}, n \in [N], s \in [T_2, t] \). We further get 
\begin{equation}
\begin{split}
    \label{q_p_k_j_upper_bound_stage3}
    &\frac{ \exp (\langle \bm{q}_\pm^{(s)}, \bm{k}_{n, j}^{(s)} \rangle)}{ \exp  (\langle \bm{q}_\pm^{(s)}, \bm{k}_\pm^{(s)} \rangle) + \sum\limits_{j^\prime=2}^M  \exp  ( \langle \bm{q}_\pm^{(s)}, \bm{k}_{n, j^\prime}^{(s)} \rangle ) } \\
    &\le  \frac{ \exp (\langle \bm{q}_\pm^{(s)}, \bm{k}_{n, j}^{(s)} \rangle)}{ C \exp (\langle \bm{q}_\pm^{(s)}, \bm{k}_\pm^{(s)} \rangle) } \\
    &= \frac{1}{ C \exp( \Lambda_{n, \pm, j}^{(s)} ) } \\
    &\le \frac{1}{ C\exp(\Lambda_{n, \pm, j}^{(T_1)}) + \Theta \Big( \frac{ d_h^{\frac{1}{2}}}{N \big( \log (6N^2M^2 / \delta) \big)^{3} } \Big) } \\
    &\le \frac{1}{ \Theta \Big( \frac{ d_h^{\frac{1}{2}}}{N \big( \log (6N^2M^2 / \delta) \big)^{3} } \Big) } \\
    &= O \Big( \frac{N \big( \log (6N^2M^2 / \delta) \big)^{3}}{d_h^{\frac{1}{2}}} \Big).
\end{split}
\end{equation}
For the first inequality, by \eqref{lambda_pm_nj} and the monotonicity of \(\langle \bm{q}^{(s)}, \bm{k}^{(s)} \rangle\) (\(\langle \bm{q}_\pm^{(s)}, \bm{k}_\pm^{(s)} \rangle\) is increasing and \(\langle \bm{q}_\pm^{(s)}, \bm{k}_{n, j}^{(s)} \rangle\) is decreasing), there exist a constant \(C\) such that \( C \exp  (\langle \bm{q}_\pm^{(s)}, \bm{k}_\pm^{(s)} \rangle) \ge \exp(\langle \bm{q}_\pm^{(s)}, \bm{k}_\pm^{(s)} \rangle) + \sum\limits_{j^\prime=2}^M  \exp  ( \langle \bm{q}_\pm^{(s)}, \bm{k}_{n, j^\prime}^{(s)} \rangle ) \).
The second inequality is by plugging \eqref{lambda_pm_nj}.
Similarly, we have
\begin{equation}
\begin{split}
    \label{q_i_k_j_upper_bound_stage3}
    &\frac{ \exp (\langle \bm{q}_{n, i}^{(s)}, \bm{k}_{n, j}^{(s)} \rangle)}{ \exp  (\langle \bm{q}_{n, i}^{(s)}, \bm{k}_\pm^{(s)} \rangle) + \sum\limits_{j^\prime=2}^M  \exp  ( \langle \bm{q}_{n, i}^{(s)}, \bm{k}_{n, j^\prime}^{(s)} \rangle ) } \\
    &\le \frac{1}{ C \exp( \Lambda_{n, i, \pm, j}^{(s)} ) } \\
    &= O \Big( \frac{N \Vert \bm{\mu} \Vert_2^2 \big( \log (6N^2M^2 / \delta) \big)^{3}}{ \sigma_p^2 d d_h^{\frac{1}{2}} } \Big).
\end{split}
\end{equation}
Plugging \eqref{q_p_k_j_upper_bound_stage3} and \eqref{q_i_k_j_upper_bound_stage3} into the update rule of \(\rho_{V, n, i}\) in Lemma \ref{update_rule4V} and get
\begin{equation}
\begin{split}
    &\vert \rho_{V, n, i}^{(s+1)} - \rho_{V, n, i}^{(s)} \vert \\
    &\le \frac{\eta}{NM} \vert \ell_n^{\prime(s)} \vert \cdot \Vert \bm{\xi}_{n, i} \Vert_2^2 \cdot \Big( O \Big( \frac{N \big( \log (6N^2M^2 / \delta) \big)^{3}}{d_h^{\frac{1}{2}}} \Big) + O \Big( \frac{N \Vert \bm{\mu} \Vert_2^2 \big( \log (6N^2M^2 / \delta) \big)^{3}}{ \sigma_p^2 d d_h^{\frac{1}{2}} } \Big) \Big) \\
    &+ \frac{\eta}{NM} \sum\limits_{n^\prime \ne n \vee i \ne i^\prime} \vert \ell_{n^\prime}^{\prime(t)} \vert \cdot \langle \bm{\xi}_{n, i}, \bm{\xi}_{n^\prime, i^\prime} \rangle \cdot \Big( O \Big( \frac{N \big( \log (6N^2M^2 / \delta) \big)^{3}}{d_h^{\frac{1}{2}}} \Big) + O \Big( \frac{N \Vert \bm{\mu} \Vert_2^2 \big( \log (6N^2M^2 / \delta) \big)^{3}}{ \sigma_p^2 d d_h^{\frac{1}{2}} } \Big) \Big) \\
    &\le \frac{3\eta \tilde{\sigma}_p^2 d}{2NM} \cdot \Big( O \Big( \frac{N \big( \log (6N^2M^2 / \delta) \big)^{3}}{d_h^{\frac{1}{2}}} \Big) + O \Big( \frac{N \Vert \bm{\mu} \Vert_2^2 \big( \log (6N^2M^2 / \delta) \big)^{3}}{ \sigma_p^2 d d_h^{\frac{1}{2}} } \Big) \Big) \\
    &+ \frac{\eta}{NM} N \cdot M \cdot 2 \tilde{\sigma}_p^2 \sqrt{d \log (4N^2M^2/\delta)}  \sqrt{d} \cdot \Big( O \Big( \frac{N \big( \log (6N^2M^2 / \delta) \big)^{3}}{d_h^{\frac{1}{2}}} \Big) + O \Big( \frac{N \Vert \bm{\mu} \Vert_2^2 \big( \log (6N^2M^2 / \delta) \big)^{3}}{ \sigma_p^2 d d_h^{\frac{1}{2}} } \Big) \Big) \\
    &\le \frac{2\eta}{NM} \cdot \Big( O \Big( \frac{N \sigma_p^2 d \big( \log (6N^2M^2 / \delta) \big)^{3}}{d_h^{\frac{1}{2}}} \Big) + O \Big( \frac{N \Vert \bm{\mu} \Vert_2^2 \big( \log (6N^2M^2 / \delta) \big)^{3}}{ d_h^{\frac{1}{2}} } \Big) \Big) \\
    &= O \Big( \frac{\eta \sigma_p^2 d \big( \log (6N^2M^2 / \delta) \big)^{3}}{d_h^{\frac{1}{2}}} + \frac{\eta \Vert \bm{\mu} \Vert_2^2 \big( \log (6N^2M^2 / \delta) \big)^{3}}{ d_h^{\frac{1}{2}} } \Big)
\end{split}
\end{equation}
where the second inequality is by Lemma \ref{caoyuan} and \(\vert \ell_n^{\prime(t)} \vert \le 1\). For the last inequality, since \(d = \widetilde{\Omega} \Big( \epsilon^{-2} N^2 d_h \Big) \), we have \(N \cdot M \cdot 2 \tilde{\sigma}_p^2 \sqrt{d \log (4N^2M^2/\delta)} \le \frac{1}{2} \tilde{\sigma}_p^2 d\). By Definition \ref{def1} and taking a summation we have
\begin{equation}
\begin{split}
    &\vert V_{n, i}^{(t+1)} \vert \le \vert V_{n, i}^{(T_2)} \vert + \sum\limits_{s = T_2}^{t} \vert \rho_{V, n, i}^{(s+1)} - \rho_{V, n, i}^{(s)} \vert \cdot \Vert \bm{w}_O \Vert_2^2 \\
    &\le \vert V_{n, i}^{(T_2)} \vert + T_3 \cdot \vert \rho_{V, n, i}^{(t+1)} - \rho_{V, n, i}^{(t)} \vert \cdot \Vert \bm{w}_O \Vert_2^2 \\
    &\le o(1) + \Theta \Big( \frac{1}{\eta \epsilon \Vert \bm{\mu} \Vert_2^2 \Vert \bm{w}_O \Vert_2^2} \Big) \cdot O \Big( \frac{\eta \sigma_p^2 d \big( \log (6N^2M^2 / \delta) \big)^{3}}{d_h^{\frac{1}{2}}} + \frac{\eta \Vert \bm{\mu} \Vert_2^2 \big( \log (6N^2M^2 / \delta) \big)^{3}}{ d_h^{\frac{1}{2}} } \Big) \cdot \Vert \bm{w}_O \Vert_2^2 \\
    &= o(1) + O \Big( \frac{ \sigma_p^2 d \big( \log (6N^2M^2 / \delta) \big)^{3}}{ \epsilon \Vert \bm{\mu} \Vert_2^2 d_h^{\frac{1}{2}}} + \frac{ \big( \log (6N^2M^2 / \delta) \big)^{3}}{ \epsilon d_h^{\frac{1}{2}} } \Big) \\
    &= o(1) + o(1) \\
    &= o(1),
\end{split}
\end{equation}
where the first equality is by \(N \cdot \mathrm{SNR}^2 \ge \Omega(1)\), the second equality is by \(d_h = \widetilde{\Omega} \Big( \max \{\mathrm{SNR}^4, \mathrm{SNR}^{-4}\} N^2 \epsilon^{-2} \Big)\). Then we have a constant upper bound for the sum of \(V_{n, i}\) as follows:
\[
    \sum\limits_{i \in [M] \backslash \{1\}} \vert V_{n, i}^{(s)} \vert = (M - 1) \cdot o(1) \le C_{15}
\]
for \(n \in [N], s \in [T_2, t]\).

Expanding \eqref{q_p_k_j_upper_bound_stage3} and \eqref{q_i_k_j_upper_bound_stage3} we have
\begin{equation}
    \label{q_pm_k_j_attn}
    \frac{ \exp (\langle \bm{q}_\pm^{(s)}, \bm{k}_{n, j}^{(s)} \rangle)}{ \exp  (\langle \bm{q}_\pm^{(s)}, \bm{k}_+^{(s)} \rangle) + \sum\limits_{j^\prime=2}^M  \exp  ( \langle \bm{q}_\pm^{(s)}, \bm{k}_{n, j^\prime}^{(s)} \rangle ) }
    = O \Big( \frac{N \big( \log (6N^2M^2 / \delta) \big)^{3}}{d_h^{\frac{1}{2}}} \Big)
    = o(1),
\end{equation}
where the equality is by \(d_h = \widetilde{\Omega} \Big( \max \{\mathrm{SNR}^4, \mathrm{SNR}^{-4}\} N^2 \epsilon^{-2} \Big)\).
\begin{equation}
    \label{q_i_k_j_attn}
    \frac{ \exp (\langle \bm{q}_{n, i}^{(s)}, \bm{k}_{n, j}^{(s)} \rangle)}{ \exp  (\langle \bm{q}_{n, i}^{(s)}, \bm{k}_+^{(s)} \rangle) + \sum\limits_{j^\prime=2}^M  \exp  ( \langle \bm{q}_{n, i}^{(s)}, \bm{k}_{n, j^\prime}^{(s)} \rangle ) }
    = O \Big( \frac{N \Vert \bm{\mu} \Vert_2^2 \big( \log (6N^2M^2 / \delta) \big)^{3}}{ \sigma_p^2 d d_h^{\frac{1}{2}} } \Big)
    = o(1),
\end{equation}
where the equality is by \(d_h = \widetilde{\Omega} \Big( \max \{\mathrm{SNR}^4, \mathrm{SNR}^{-4}\} N^2 \epsilon^{-2} \Big)\). Then we have
\begin{equation}
    \label{q_pm_k_pm_attn}
    softmax(\langle \bm{q}_\pm^{(s)}, \bm{k}_\pm^{(s)} \rangle) \ge 1 - (M - 1) \cdot o(1) \ge 1 - o(1),
\end{equation}
\begin{equation}
    \label{q_i_k_pm_attn}
    softmax(\langle \bm{q}_{n, i}^{(s)}, \bm{k}_\pm^{(s)} \rangle) \ge 1 - (M - 1) \cdot o(1) \ge 1 - o(1)
\end{equation}
for \(i \in [M] \backslash \{1\}, n \in [N], s \in [T_2, t]\). Next we provide the bounds for \(- \ell_n^{\prime(s)}\).
Note that \( \ell (z) = \log(1 + \exp(-z)) \) and \( - \ell^\prime = \exp(-z) / (1 + \exp(-z)) \), without loss of generality, we assume \( y_n = 1 \), we have
\begin{equation}
    \label{loss_gradient}
    - \ell^\prime (f(\bm{X}_n, \theta (s))) = \frac{1}{1 + \exp(\frac{1}{M} \sum\limits_{l=1}^M \bm{\varphi} (\bm{x}_{n, l}^\top \bm{W}_{Q}^{(s)} \bm{W}_{K}^{(s)\top} (\bm{X}_n)^\top ) \bm{X}_n \bm{W}_V^{(s)} \bm{w}_O)}
\end{equation}
where \(\frac{1}{M} \sum\limits_{l=1}^M \bm{\varphi} (\bm{x}_{n, l}^\top \bm{W}_{Q}^{(s)} \bm{W}_{K}^{(s)\top} (\bm{X}_n)^\top ) \bm{X}_n \bm{W}_V^{(s)} \bm{w}_O\) can be bounded as follows:
\begin{equation}
\begin{split}
    \label{output_lower_bound}
    &\frac{1}{M} \sum\limits_{l=1}^M \bm{\varphi} (\bm{x}_{n, l}^\top \bm{W}_{Q}^{(s)} \bm{W}_{K}^{(s)\top} (\bm{X}_n)^\top ) \bm{X}_n \bm{W}_V^{(s)} \bm{w}_O \\
    &= \frac{1}{M} \Big( \big( softmax(\langle \bm{q}_\pm^{(s)}, \bm{k}_\pm^{(s)} \rangle) + \sum\limits_{l=2}^M softmax(\langle \bm{q}_{n, l}^{(s)}, \bm{k}_\pm^{(s)} \rangle) \big) \cdot \bm{\mu}_+^\top \bm{W}_V^{(s)} \bm{w}_O \\
    &+ \sum\limits_{j \in [M] \backslash \{1\}} \big( softmax(\langle \bm{q}_\pm^{(s)}, \bm{k}_{n, j}^{(s)} \rangle) + \sum\limits_{l=2}^M softmax(\langle \bm{q}_{n, l}^{(s)}, \bm{k}_{n, j}^{(s)} \rangle) \big) \cdot \bm{\xi}_{n, j}^{\top} \bm{W}_V^{(s)} \bm{w}_O \Big) \\
    &= \frac{1}{M} \Big( M \cdot \big( 1 - o(1) \big) \cdot V_{+}^{(s)} +  M \cdot o(1) \cdot \sum\limits_{j \in [M] \backslash \{1\}} V_{n, i}^{(s)} \Big) \\
    &\ge \frac{V_{+}^{(s)}}{2}
\end{split}
\end{equation}
for \(s \in [T_2, t]\). The second equality is by plugging \eqref{q_pm_k_j_attn}, \eqref{q_i_k_j_attn}, \eqref{q_pm_k_pm_attn} and \eqref{q_i_k_pm_attn}, the inequality is by \(V_+^{(s)} \ge 3M \cdot \vert V_{n, i}^{(s)} \vert \) for \(s \in [T_2, t]\). Similarly, we have
\begin{equation}
\begin{split}
    \label{output_upper_bound}
    &\frac{1}{M} \sum\limits_{l=1}^M \bm{\varphi} (\bm{x}_{n, l}^\top \bm{W}_{Q}^{(s)} \bm{W}_{K}^{(s)\top} (\bm{X}_n)^\top ) \bm{X}_n \bm{W}_V^{(s)} \bm{w}_O \\
    &\le \max\limits_{i \in [M] \backslash \{1\}} \{ V_{+}^{(s)}, V_{n, i}^{(s)} \} \\
    &= V_{+}^{(s)}
\end{split}
\end{equation}
Plugging \eqref{output_lower_bound} and \eqref{output_upper_bound} into \eqref{loss_gradient} we have
\begin{equation}
\begin{split}
    \label{loss_gradient_lower_bound}
    - \ell^\prime (f(\bm{X}_n, \theta (s))) &= \frac{1}{1 + \exp(\frac{1}{M} \sum\limits_{l=1}^M \bm{\varphi} (\bm{x}_{n, l}^\top \bm{W}_{Q}^{(s)} \bm{W}_{K}^{(s)\top} (\bm{X}_n)^\top ) \bm{X}_n \bm{W}_V^{(s)} \bm{w}_O)} \\
    &\ge \frac{1}{1 + \exp (V_{+}^{(s)})} \\
    &\ge \frac{C_{16}}{\exp (V_{+}^{(s)})}
\end{split}
\end{equation}
where the first inequality is by plugging \eqref{output_upper_bound}. For the last inequality, note that \(V_{+}^{(T_2)} \ge 0\) and \(V_{+}^{(s)}\) is monotonically increasing, so there exist a constant \(C_{16}\) such that \(\frac{1}{1 + \exp (V_{+}^{(s)})} \ge \frac{C_{16}}{\exp (V_{+}^{(s)})}\).
We also have the upper bound
\begin{equation}
\begin{split}
    \label{loss_gradient_upper_bound}
    - \ell^\prime (f(\bm{X}_n, \theta (s))) &= \frac{1}{1 + \exp(\frac{1}{M} \sum\limits_{l=1}^M \bm{\varphi} (\bm{x}_{n, l}^\top \bm{W}_{Q}^{(s)} \bm{W}_{K}^{(s)\top} (\bm{X}_n)^\top ) \bm{X}_n \bm{W}_V^{(s)} \bm{w}_O)} \\
    &\le \frac{1}{1 + \exp ( V_{+}^{(s)} / 2)} \\
    &\le \frac{1}{\exp ( V_{+}^{(s)} / 2)}
\end{split}
\end{equation}
Plugging \eqref{q_pm_k_pm_attn}, \eqref{q_i_k_pm_attn} and \eqref{loss_gradient_lower_bound} into the update rule of \(\gamma_{V, +}^{(t)}\) and in Lemma \ref{update_rule4V} and get
\begin{equation}
\begin{split}
    & \gamma_{V, +}^{(s + 1)} - \gamma_{V, +}^{(s)} \\
    &= - \frac{\eta \Vert \bm{\mu} \Vert_2^2}{NM} \sum\limits_{n \in S_+} \ell_n^{\prime(s)} \big( \frac{ \exp ( \langle \bm{q}_+^{(s)}, \bm{k}_+^{(s)} \rangle )}{  \exp ( \langle \bm{q}_+^{(s)}, \bm{k}_+^{(s)} \rangle ) + \sum\limits_{k=2}^M  \exp ( \langle \bm{q}_+^{(s)}, \bm{k}_{n, k}^{(s)} \rangle )}  \\
    &+ \sum\limits_{j=2}^M \frac{ \exp ( \langle \bm{q}_{n, j}^{(s)}, \bm{k}_+^{(s)} \rangle )}{  \exp ( \langle \bm{q}_{n, j}^{(s)}, \bm{k}_+^{(s)} \rangle ) + \sum\limits_{k=2}^M  \exp ( \langle \bm{q}_{n, j}^{(s)}, \bm{k}_{n, k}^{(s)} \rangle )} \big) \\
    &\ge - \frac{\eta \Vert \bm{\mu} \Vert_2^2}{NM} \sum\limits_{n \in S_+} \ell_n^{\prime(s)} (M \cdot (1 - o(1))) \\
    &\ge \frac{\eta \Vert \bm{\mu} \Vert_2^2}{N} \cdot \frac{N}{4} \cdot (1 - o(1)) \cdot \frac{C_{16}}{\exp (V_{+}^{(s)})} \\
    &\ge \frac{\eta C_{17} \Vert \bm{\mu} \Vert_2^2}{\exp (V_{+}^{(s)})}
\end{split}
\end{equation}
where the second inequality is by \eqref{loss_gradient_lower_bound}. Then by definition \ref{def1}, we get
\[
    V_{+}^{(s+1)} - V_{+}^{(s)} = \big( \gamma_{V, +}^{(s + 1)} - \gamma_{V, +}^{(s)} \big) \Vert \bm{w}_O \Vert_2^2 \ge \frac{\eta C_{17} \Vert \bm{\mu} \Vert_2^2 \Vert \bm{w}_O \Vert_2^2}{\exp (V_{+}^{(s)})}
\]
Multiply both sides simultaneously by \(\exp (V_{+}^{(s)})\) and get
\[
    \exp (V_{+}^{(s)}) \big( V_{+}^{(s+1)} - V_{+}^{(s)} \big) \ge \eta C_{17} \Vert \bm{\mu} \Vert_2^2 \Vert \bm{w}_O \Vert_2^2
\]
Taking a summation from \(T_2\) to \(t\) and get
\begin{equation}
\begin{split}
    &\sum\limits_{s = T_2}^{t} \exp (V_{+}^{(s)}) \big(V_{+}^{(s+1)} - V_{+}^{(s)} \big) \\
    &\ge \sum\limits_{s = T_2}^{t} \eta C_{17} \Vert \bm{\mu} \Vert_2^2 \Vert \bm{w}_O \Vert_2^2 \\
    &\ge \eta C_{17} \Vert \bm{\mu} \Vert_2^2 \Vert \bm{w}_O \Vert_2^2 (t - T_2 + 1)
\end{split}
\end{equation}
By the property that \(V_{+}^{(s)}\) is monotonically increasing, we have
\begin{equation}
\begin{split}
    \int_{V_{+}^{(T_2)}}^{V_{+}^{(t+1)}} \exp(x) dx &\ge \sum\limits_{s = T_2}^{t} \exp ( V_{+}^{(s)}) \big( V_{+}^{(s+1)} - V_{+}^{(s)} \big) \\
    &\ge \eta C_{17} \Vert \bm{\mu} \Vert_2^2 \Vert \bm{w}_O \Vert_2^2 (t - T_2 + 1)
\end{split}
\end{equation}
By \(\int_{V_{+}^{(T_2)}}^{V_{+}^{(t+1)}} \exp(x) dx = \exp ( V_{+}^{(t+1)}) - \exp ( V_{+}^{(T_2)})\) we get
\begin{equation}
\begin{split}
    V_{+}^{(t+1)} \ge \log \Big( \exp ( V_{+}^{(T_2)}) + \eta C_{17} \Vert \bm{\mu} \Vert_2^2 \Vert \bm{w}_O \Vert_2^2 (t - T_2 + 1) \Big)
\end{split}
\end{equation}
Similarly, we have
\begin{equation}
\begin{split}
    V_{-}^{(t+1)} \le - \log \Big( \exp ( -V_{-}^{(T_2)}) + \eta C_{17} \Vert \bm{\mu} \Vert_2^2 \Vert \bm{w}_O \Vert_2^2 (t - T_2 + 1) \Big)
\end{split}
\end{equation}
Next we provide upper bounds for \(V_{+}^{(t+1)}\) and \(V_{-}^{(t+1)}\). By the update rule of \(\gamma_{V, +}^{(t)}\) and in Lemma \ref{update_rule4V} we have
\begin{equation*}
\begin{split}
    & \gamma_{V, +}^{(s + 1)} = \gamma_{V, +}^{(s)} \\
    &- \frac{\eta \Vert \bm{\mu} \Vert_2^2}{NM} \sum\limits_{n \in S_+} \ell_n^{\prime(s)} \big( \frac{ \exp ( \langle \bm{q}_+^{(s)}, \bm{k}_+^{(s)} \rangle )}{  \exp ( \langle \bm{q}_+^{(s)}, \bm{k}_+^{(s)} \rangle ) + \sum\limits_{k=2}^M  \exp ( \langle \bm{q}_+^{(s)}, \bm{k}_{n, k}^{(s)} \rangle )}  \\
    &+ \sum\limits_{j=2}^M \frac{ \exp ( \langle \bm{q}_{n, j}^{(s)}, \bm{k}_+^{(s)} \rangle )}{  \exp ( \langle \bm{q}_{n, j}^{(s)}, \bm{k}_+^{(s)} \rangle ) + \sum\limits_{k=2}^M  \exp ( \langle \bm{q}_{n, j}^{(s)}, \bm{k}_{n, k}^{(s)} \rangle )} \big) \\
    &\le \frac{\eta \Vert \bm{\mu} \Vert_2^2}{NM} \sum\limits_{n \in S_+} - \ell_n^{\prime(s)} \cdot M\\
    &\le \frac{\eta \Vert \bm{\mu} \Vert_2^2}{N} \cdot \frac{3N}{4} \cdot \frac{1}{\exp ( V_{+}^{(s)} / 2)} \\
    &= \frac{3\eta \Vert \bm{\mu} \Vert_2^2}{4\exp ( V_{+}^{(s)} / 2)}
\end{split}
\end{equation*}
where the second inequality is by \eqref{loss_gradient_upper_bound}. Then by definition \ref{def1}, we get
\begin{equation}
    \label{V_p_dy}
    V_{+}^{(s+1)} - V_{+}^{(s)} = \big( \gamma_{V, +}^{(s + 1)} - \gamma_{V, +}^{(s)} \big) \Vert \bm{w}_O \Vert_2^2 \le \frac{3\eta \Vert \bm{\mu} \Vert_2^2 \Vert \bm{w}_O \Vert_2^2}{4\exp ( V_{+}^{(s)} / 2)}
\end{equation}
Further we have
\begin{equation}
\begin{split}
    \exp(V_{+}^{(s+1)} / 2) &\le \exp(V_{+}^{(s)}/2 + \frac{3\eta \Vert \bm{\mu} \Vert_2^2 \Vert \bm{w}_O \Vert_2^2}{8\exp ( V_{+}^{(s)} / 2)}) \\
    &= \exp(V_{+}^{(s)} / 2) \cdot \exp(\frac{3\eta \Vert \bm{\mu} \Vert_2^2 \Vert \bm{w}_O \Vert_2^2}{8\exp ( V_{+}^{(s)} / 2)}) \\
    &\le C_{18} \exp(V_{+}^{(s)} / 2).
\end{split}
\end{equation}
For the last inequality, by \(\eta \le \widetilde{O} ( \min \{ \bm{\mu} \Vert_2^{-2}, ( \sigma_p^2 d )^{-1} \} \cdot d_h^{-\frac{1}{2}} )\), \(V_{+}^{(T_2)} =\Theta(1) \) and the monotonicity of \(V_{+}^{(s)}\), we have \(\exp ( \frac{3\eta \Vert \bm{\mu} \Vert_2^2 \Vert \bm{w}_O \Vert_2^2}{8\exp ( V_{+}^{(s)} / 2)} ) \le \exp(o(1)) \le C_{18}\). 
Multiplying both sides by \(\Big( V_{+}^{(s+1)}/2 - V_{+}^{(s)}/2 \Big)\) simultaneously gives
\begin{equation}
\begin{split}
    \exp(V_{+}^{(s+1)} / 2) \Big( V_{+}^{(s+1)} / 2 - V_{+}^{(s)}/2 \Big) &\le C_{18} \exp(V_{+}^{(s)} / 2) \Big( V_{+}^{(s+1)}/2 - V_{+}^{(s)}/2 \Big) \\
    &\le \frac{3\eta C_{18} \Vert \bm{\mu} \Vert_2^2 \Vert \bm{w}_O \Vert_2^2}{8}
\end{split}
\end{equation}
where the last inequality is by plugging \eqref{V_p_dy}. Taking a summation we have
\begin{equation}
\begin{split}
    &\int_{V_{+}^{(T_2)} / 2}^{V_{+}^{(t+1)} / 2} \exp(x) dx \\
    &\le \sum\limits_{s = T_2}^{t} \exp(V_{+}^{(s+1)} / 2) \Big( V_{+}^{(s+1)} / 2 - V_{+}^{(s)}/2 \Big) \\
    &\le \sum\limits_{s = T_2}^{T_3} \frac{3\eta C_{18} \Vert \bm{\mu} \Vert_2^2 \Vert \bm{w}_O \Vert_2^2}{8} \\
    &\le \Theta \Big( \frac{1}{\eta \epsilon \Vert \bm{\mu} \Vert_2^2 \Vert \bm{w}_O \Vert_2^2} \Big) \cdot \frac{3\eta C_{18} \Vert \bm{\mu} \Vert_2^2 \Vert \bm{w}_O \Vert_2^2}{8} \\
    &= O(\frac{1}{\epsilon})
\end{split}
\end{equation}
By \(\int_{V_{+}^{(T_2)} / 2}^{V_{+}^{(t+1)} / 2} \exp(x) dx = \exp(V_{+}^{(t+1)} / 2) - \exp(V_{+}^{(T_2)} / 2)\) we have
\[
    V_{+}^{(t+1)} \le 2 \log \big( \exp(V_{+}^{(T_2)} / 2) + O(\frac{1}{\epsilon}) \big) = 2 \log \big( O(\frac{1}{\epsilon}) \big)
\]
Similarly, we have
\[
    V_{-}^{(t+1)} \ge - 2 \log \big( O(\frac{1}{\epsilon}) \big)
\]

\subsubsection{Proof of Claim \ref{claim6}}
\label{proof_claim6}
By \( \mathcal{H}(T_2), \dots, \mathcal{H}(t - 1) \), we have \( softmax(\langle \bm{q}_\pm^{(t)}, \bm{k}_\pm^{(t)} \rangle), softmax(\langle \bm{q}_{n, i}^{(t)}, \bm{k}_\pm^{(t)} \rangle) = 1 - o(1) \) and \( softmax( \langle \bm{q}_\pm^{(t)}, \bm{k}_{n, j}^{(t)} \rangle ) , softmax( \langle \bm{q}_{n, i}^{(t)}, \bm{k}_{n, j}^{(t)} \rangle ) = o(1) \), which have been proved in \ref{proof_claim5}. 
By the results of \ref{bound_of_alpha_beta}, we have the signs of \(\alpha\) and \(\beta\) as follows:
\[
    \alpha_{+, +}^{(t)}, \alpha_{-, -}^{(t)}, \beta_{+, +}^{(t)}, \beta_{-, -}^{(t)}, \alpha_{n, i, +}^{(t)}, \alpha_{n, i, -}^{(t)}, \beta_{n, +, i}^{(t)}, \beta_{n, -, i}^{(t)} \ge 0,
\]
\[
    \alpha_{n, +, i}^{(t)}, \alpha_{n, -, i}^{(t)}, \alpha_{n, i, n, j}^{(t)}, \beta_{n, i, +}^{(t)}, \beta_{n, i, -}^{(t)}, \beta_{n, j, n, i}^{(t)} \le 0.
\]
Then combined with \( \mathcal{G}(T) \) and we have the dynamics of \(\langle \bm{q}, \bm{k} \rangle\) as follows:
\begin{equation}
\begin{split}
    &\langle \bm{q}_+^{(t+1)}, \bm{k}_+^{(t+1)} \rangle - \langle \bm{q}_+^{(t)}, \bm{k}_+^{(t)} \rangle \\
    &= \alpha_{+, +}^{(t)} \Vert \bm{k}_+^{(t)} \Vert_2^2 + \sum\limits_{n \in S_+} \sum\limits_{i=2}^M \alpha_{n, +, i}^{(t)} \langle \bm{k}_+^{(t)}, \bm{k}_{n, i}^{(t)} \rangle \\
    &+ \beta_{+, +}^{(t)} \Vert \bm{q}_+^{(t)} \Vert_2^2 + \sum\limits_{n \in S_+} \sum\limits_{i=2}^M \beta_{n, +, i}^{(t)} \langle \bm{q}_+^{(t)}, \bm{q}_{n, i}^{(t)} \rangle \\
    &+ \Big( \alpha_{+, +}^{(t)} \bm{k}_+^{(t)} + \sum\limits_{n \in S_+} \sum\limits_{i=2}^M \alpha_{n, +, i}^{(t)} \bm{k}_{n, i}^{(t)} \Big) \\
    &\cdot \Big( \beta_{+, +}^{(t)} \bm{q}_+^{(t)\top} + \sum\limits_{n \in S_+} \sum\limits_{i=2}^M \beta_{n, +, i}^{(t)} \bm{q}_{n, i}^{(t)\top} \Big) \\
    &= \alpha_{+, +}^{(t)} \Vert \bm{k}_+^{(t)} \Vert_2^2 + \beta_{+, +}^{(t)} \Vert \bm{q}_+^{(t)} \Vert_2^2 + \{ lower \ order \ term \} \\
    &\ge 0
\end{split}
\end{equation}
Similarly, we have
\begin{equation}
\begin{split}
    \langle \bm{q}_-^{(t+1)}, \bm{k}_-^{(t+1)} \rangle - \langle \bm{q}_-^{(t)}, \bm{k}_-^{(t)} \rangle \ge 0.
\end{split}
\end{equation}

\begin{equation}
\begin{split}
    &\langle \bm{q}_{n, i}^{(t+1)}, \bm{k}_+^{(t+1)} \rangle - \langle \bm{q}_{n, i}^{(t)}, \bm{k}_+^{(t)} \rangle \\
    &= \alpha_{n, i, +}^{(t)} \Vert \bm{k}_+^{(t)} \Vert_2^2 + \alpha_{n, i, -}^{(t)} \langle \bm{k}_+^{(t)}, \bm{k}_-^{(t)} \rangle + \sum\limits_{n^\prime =1}^N \sum\limits_{l=2}^M \alpha_{n, i, n^\prime, l}^{(t)} \langle \bm{k}_+^{(t)}, \bm{k}_{n^\prime, l}^{(t)} \rangle \\
    &+ \beta_{+, +}^{(t)} \langle \bm{q}_+^{(t)}, \bm{q}_{n, i}^{(t)} \rangle + \sum\limits_{n^\prime \in S_+} \sum\limits_{l=2}^M \beta_{n^\prime, +, l}^{(t)} \langle \bm{q}_{n, i}^{(t)}, \bm{q}_{n^\prime, l}^{(t)} \rangle \\
    &+ \Big( \alpha_{n, i, +}^{(t)} \bm{k}_+^{(t)} + \alpha_{n, i, -}^{(t)} \bm{k}_-^{(t)} + \sum\limits_{n^\prime =1}^N \sum\limits_{l=2}^M \alpha_{n, i, n^\prime, l}^{(t)} \bm{k}_{n^\prime, l}^{(t)} \Big) \\
    &\cdot \Big( \beta_{+, +}^{(t)} \bm{q}_+^{(t)\top} + \sum\limits_{n^\prime \in S_+} \sum\limits_{l=2}^M \beta_{n^\prime, +, l}^{(t)} \bm{q}_{n^\prime, l}^{(t)\top} \Big) \\
    &= \alpha_{n, i, +}^{(t)} \Vert \bm{k}_+^{(t)} \Vert_2^2 + \beta_{n, +, i}^{(t)} \Vert \bm{q}_{n, i}^{(t)} \Vert_2^2 + \{ lower \ order \ term \} \\
    &\ge 0.
\end{split}
\end{equation}
Similarly, we have
\begin{equation}
\begin{split}
    \langle \bm{q}_{n, i}^{(t+1)}, \bm{k}_-^{(t+1)} \rangle - \langle \bm{q}_{n, i}^{(t)}, \bm{k}_-^{(t)} \rangle \ge 0.
\end{split}
\end{equation}

\begin{equation}
\begin{split}
    &\langle \bm{q}_+^{(t+1)}, \bm{k}_{n, j}^{(t+1)} \rangle - \langle \bm{q}_+^{(t)}, \bm{k}_{n, j}^{(t)} \rangle \\
    &= \alpha_{+, +}^{(t)} \langle \bm{k}_+^{(t)}, \bm{k}_{n, j}^{(t)} \rangle + \sum\limits_{n^\prime \in S_+} \sum\limits_{l=2}^M \alpha_{n^\prime, +, l}^{(t)} \langle \bm{k}_{n, j}^{(t)}, \bm{k}_{n^\prime, l}^{(t)} \rangle \\
    &+ \beta_{n, j, +}^{(t)} \Vert \bm{q}_+^{(t)} \Vert_2^2 + \beta_{n, j, -}^{(t)} \langle \bm{q}_+^{(t)}, \bm{q}_-^{(t)} \rangle + \sum\limits_{n^\prime = 1}^N \sum\limits_{l=2}^M \beta_{n, j, n^\prime, l}^{(t)} \langle \bm{q}_+^{(t)}, \bm{q}_{n^\prime, l}^{(t)} \rangle \\
    &+ \Big( \alpha_{+, +}^{(t)} \bm{k}_+^{(t)} + \sum\limits_{n^\prime \in S_+} \sum\limits_{l=2}^M \alpha_{n^\prime, +, l}^{(t)} \bm{k}_{n^\prime, l}^{(t)} \Big) \\
    &\cdot \Big( \beta_{n, j, +}^{(t)} \bm{q}_+^{(t)\top} + \beta_{n, j, -}^{(t)} \bm{q}_-^{(t)\top} + \sum\limits_{n^\prime = 1}^N \sum\limits_{l=2}^M \beta_{n, j, n^\prime, l}^{(t)} \bm{q}_{n^\prime, l}^{(t)\top} \Big) \\
    &= \alpha_{n, +, j}^{(t)} \Vert \bm{k}_{n, j}^{(t)} \Vert_2^2 + \beta_{n, j, +}^{(t)} \Vert \bm{q}_+^{(t)} \Vert_2^2 + \{ lower \ order \ term \} \\
    &\le 0
\end{split}
\end{equation}
Similarly, we have
\begin{equation}
\begin{split}
    \langle \bm{q}_-^{(t+1)}, \bm{k}_{n, j}^{(t+1)} \rangle - \langle \bm{q}_-^{(t)}, \bm{k}_{n, j}^{(t)} \rangle \le 0.
\end{split}
\end{equation}
\begin{equation}
\begin{split}
    &\langle \bm{q}_{n, i}^{(t+1)}, \bm{k}_{n, j}^{(t+1)} \rangle - \langle \bm{q}_{n, i}^{(t)}, \bm{k}_{n, j}^{(t)} \rangle \\
    &= \alpha_{n, i, +}^{(t)} \langle \bm{k}_+^{(t)}, \bm{k}_{n, j}^{(t)} \rangle + \alpha_{n, i, -}^{(t)} \langle \bm{k}_-^{(t)}, \bm{k}_{n, j}^{(t)} \rangle + \sum\limits_{n^\prime =1}^N \sum\limits_{l=2}^M \alpha_{n, i, n^\prime, l}^{(t)} \langle \bm{k}_{n^\prime, l}^{(t)}, \bm{k}_{n, j}^{(t)} \rangle \\
    &+ \beta_{n, j, +}^{(t)} \langle \bm{q}_+^{(t)}, \bm{q}_{n, i}^{(t)} \rangle + \beta_{n, j, -}^{(t)} \langle \bm{q}_-^{(t)}, \bm{q}_{n, i}^{(t)} \rangle + \sum\limits_{n^\prime = 1}^N \sum\limits_{l=2}^M \beta_{n, j, n^\prime, l}^{(t)} \langle \bm{q}_{n^\prime, l}^{(t)}, \bm{q}_{n, i}^{(t)} \rangle \\
    &+ \Big( \alpha_{n, i, +}^{(t)} \bm{k}_+^{(t)} + \alpha_{n, i, -}^{(t)} \bm{k}_-^{(t)} + \sum\limits_{n^\prime =1}^N \sum\limits_{l=2}^M \alpha_{n, i, n^\prime, l}^{(t)} \bm{k}_{n^\prime, l}^{(t)} \Big) \\
    &\cdot \Big( \beta_{n, j, +}^{(t)} \bm{q}_+^{(t)\top} + \beta_{n, j, -}^{(t)} \bm{q}_-^{(t)\top} + \sum\limits_{n^\prime = 1}^N \sum\limits_{l=2}^M \beta_{n, j, n^\prime, l}^{(t)} \bm{q}_{n^\prime, l}^{(t)\top} \Big) \\
    &= \alpha_{n, i, n, j}^{(t)} \Vert \bm{k}_{n, j}^{(t)} \Vert_2^2 + \beta_{n, j, n, i}^{(t)} \Vert \bm{q}_{n, i}^{(t)} \Vert_2^2 \\
    &+ \{ lower \ order \ term \} \\
    &\le 0,
\end{split}
\end{equation}
which completes the proof.
The proof for Claim \ref{claim7} is in Section \ref{proof_claim7}

\subsubsection{Proof of Claim \ref{claim8}}
\label{proof_claim8}
By the results of \ref{upper_bound_of_qk2}, we have

\begin{equation}
\begin{split}
    \label{q_p_k_p_dy_upper_bound2}
    \langle \bm{q}_+^{(t+1)}, \bm{k}_+^{(t+1)} \rangle - \langle \bm{q}_+^{(t)}, \bm{k}_+^{(t)} \rangle \le \frac{\eta C_{10} \Vert \bm{\mu} \Vert_2^4 \sigma_h^2 d_h \log \big( O(\frac{1}{\epsilon}) \big)}{\exp ( \langle \bm{q}_+^{(t)}, \bm{k}_+^{(t)} \rangle)},
\end{split}
\end{equation}
Further we have
\begin{equation}
\begin{split}
    \exp(\langle \bm{q}_+^{(t+1)}, \bm{k}_+^{(t+1)} \rangle) &\le \exp \Big( \langle \bm{q}_+^{(t)}, \bm{k}_+^{(t)} \rangle + \frac{\eta C_{10} \Vert \bm{\mu} \Vert_2^4 \sigma_h^2 d_h \log \big( O(\frac{1}{\epsilon}) \big)}{\exp ( \langle \bm{q}_+^{(t)}, \bm{k}_+^{(t)} \rangle)} \Big) \\
    &= \exp \Big( \langle \bm{q}_+^{(t)}, \bm{k}_+^{(t)} \rangle \Big) \cdot \exp \Big( \frac{\eta C_{10} \Vert \bm{\mu} \Vert_2^4 \sigma_h^2 d_h \log \big( O(\frac{1}{\epsilon}) \big)}{\exp ( \langle \bm{q}_+^{(t)}, \bm{k}_+^{(t)} \rangle)} \Big) \\
    &\le C_{11} \exp \Big( \langle \bm{q}_+^{(t)}, \bm{k}_+^{(t)} \rangle \Big).
\end{split}
\end{equation}
For the last inequality, by \(\eta \le \widetilde{O} ( \min \{ \bm{\mu} \Vert_2^{-2}, ( \sigma_p^2 d )^{-1} \} \cdot d_h^{-\frac{1}{2}} )\), \(\sigma_h^2 \le \min \{ \Vert \bm{\mu} \Vert_2^{-2}, ( \sigma_p^2 d )^{-1} \} \cdot d_h^{-\frac{1}{2}} \cdot \big( \log ( 6N^2M^2 / \delta ) \big)^{-\frac{3}{2}}\), \(\langle \bm{q}_+^{(T_1)}, \bm{k}_+^{(T_1)} \rangle = o(1)\) and the monotonicity of \(\langle \bm{q}_+^{(s)}, \bm{k}_+^{(s)} \rangle\) for \(s \in [T_1, t]\), we have \(\exp \Big( \frac{\eta C_{10} \Vert \bm{\mu} \Vert_2^4 \sigma_h^2 d_h \log \big( O(\frac{1}{\epsilon}) \big)}{\exp ( \langle \bm{q}_+^{(t)}, \bm{k}_+^{(t)} \rangle)} \Big) \le \exp(o(1)) \le C_{11}\). Multiplying both sides by \(\Big( \langle \bm{q}_+^{(t+1)}, \bm{k}_+^{(t+1)} \rangle - \langle \bm{q}_+^{(t)}, \bm{k}_+^{(t)} \rangle \Big)\) simultaneously gives
\begin{equation}
\begin{split}
    &\exp(\langle \bm{q}_+^{(t+1)}, \bm{k}_+^{(t+1)} \rangle) \Big( \langle \bm{q}_+^{(t+1)}, \bm{k}_+^{(t+1)} \rangle - \langle \bm{q}_+^{(t)}, \bm{k}_+^{(t)} \rangle \Big) \\
    &\le C_{11} \exp \Big( \langle \bm{q}_+^{(t)}, \bm{k}_+^{(t)} \rangle \Big) \cdot \Big( \langle \bm{q}_+^{(t+1)}, \bm{k}_+^{(t+1)} \rangle - \langle \bm{q}_+^{(t)}, \bm{k}_+^{(t)} \rangle \Big) \\
    &\le \eta C_{12} \Vert \bm{\mu} \Vert_2^4 \sigma_h^2 d_h \log \big( O(\frac{1}{\epsilon}) \big),
\end{split}
\end{equation}
where the last inequality is by plugging \eqref{q_p_k_p_dy_upper_bound2}. Taking a summation we have
\begin{equation}
\begin{split}
    &\int_{\langle \bm{q}_+^{(T_2)}, \bm{k}_+^{(T_2)} \rangle}^{\langle \bm{q}_+^{(t+1)}, \bm{k}_+^{(t+1)} \rangle} \exp(x) dx \\
    &\le \sum\limits_{s = T_2}^{t} \exp(\langle \bm{q}_+^{(s+1)}, \bm{k}_+^{(s+1)} \rangle) \Big( \langle \bm{q}_+^{(s+1)}, \bm{k}_+^{(s+1)} \rangle - \langle \bm{q}_+^{(s)}, \bm{k}_+^{(s)} \rangle \Big) \\
    &\le \sum\limits_{s = T_2}^{t} \eta C_{12} \Vert \bm{\mu} \Vert_2^4 \sigma_h^2 d_h \log \big( O(\frac{1}{\epsilon}) \big) \\
    &\le T_3 \cdot \eta C_{12} \Vert \bm{\mu} \Vert_2^4 \sigma_h^2 d_h \log \big( O(\frac{1}{\epsilon}) \big) \\
    &=O \Big( \frac{d_h^{\frac{1}{2}} \log \big( O(\frac{1}{\epsilon}) \big)}{\epsilon ( \log (6N^2M^2 / \delta) )^{\frac{3}{2}} } \Big)
\end{split}
\end{equation}
where the first inequality is due to \(\langle \bm{q}_+^{(s)}, \bm{k}_+^{(s)} \rangle\) is monotone increasing, the last equality is by \(T_3 = \Theta(\eta^{-1} \epsilon^{-1} \Vert \bm{\mu} \Vert_2^{-2} \Vert \bm{w}_O \Vert_2^{-2}) \), \(\Vert \bm{w}_O \Vert_2^{2} = \Theta(1)\) and \( \sigma_h^2 \le \min \{ \Vert \bm{\mu} \Vert_2^{-2}, ( \sigma_p^2 d )^{-1} \} \cdot d_h^{-\frac{1}{2}} \cdot \big( \log ( 6N^2M^2 / \delta ) \big)^{-\frac{3}{2}} \). By \( \int_{\langle \bm{q}_+^{(T_2)}, \bm{k}_+^{(T_2)} \rangle}^{\langle \bm{q}_+^{(t+1)}, \bm{k}_+^{(t+1)} \rangle} \exp(x) dx = \exp(\langle \bm{q}_+^{(t+1)}, \bm{k}_+^{(t+1)} \rangle) - \exp(\langle \bm{q}_+^{(T_2)}, \bm{k}_+^{(T_2)} \rangle) \), we have
\begin{equation}
\begin{split}
    \langle \bm{q}_+^{(t+1)}, \bm{k}_+^{(t+1)} \rangle \le \log \Big( \exp(\langle \bm{q}_+^{(T_2)}, \bm{k}_+^{(T_2)} \rangle) + O \Big( \frac{d_h^{\frac{1}{2}} \log \big( O(\frac{1}{\epsilon}) \big) }{ \epsilon ( \log (6N^2M^2 / \delta) )^{\frac{3}{2}} } \Big) \Big) \le \log ( \epsilon^{-1} d_h^{\frac{1}{2}} ),
\end{split}
\end{equation}
where the last inequality is by \(\langle \bm{q}_+^{(T_2)}, \bm{k}_+^{(T_2)} \rangle \le \log \big( d_h^{\frac{1}{2}} \big) \). By the results of \ref{upper_bound_of_qk2}, we also have

\begin{equation}
\begin{split}
    &\langle \bm{q}_-^{(t+1)}, \bm{k}_-^{(t+1)} \rangle - \langle \bm{q}_-^{(t)}, \bm{k}_-^{(t)} \rangle \le \frac{\eta C_{10} \Vert \bm{\mu} \Vert_2^4 \sigma_h^2 d_h \log \big( O(\frac{1}{\epsilon}) \big)}{\exp ( \langle \bm{q}_-^{(t)}, \bm{k}_-^{(t)} \rangle)}.
\end{split}
\end{equation}

\begin{equation}
\begin{split}
    &\langle \bm{q}_\pm^{(t+1)}, \bm{k}_{n, j}^{(t+1)} \rangle - \langle \bm{q}_\pm^{(t)}, \bm{k}_{n, j}^{(t)} \rangle \ge - \frac{\eta C_{10} \sigma_p^2 d \Vert \bm{\mu} \Vert_2^2 \sigma_h^2 d_h \log \big( O(\frac{1}{\epsilon}) \big) }{N} \cdot \exp (\langle \bm{q}_\pm^{(t)}, \bm{k}_{n, j}^{(t)} \rangle).
\end{split}
\end{equation}

\begin{equation}
\begin{split}
    &\langle \bm{q}_{n, i}^{(t+1)}, \bm{k}_\pm^{(t+1)} \rangle - \langle \bm{q}_{n, i}^{(t)}, \bm{k}_\pm^{(t)} \rangle \le \frac{\eta C_{10} \sigma_p^2 d \Vert \bm{\mu} \Vert_2^2 \sigma_h^2 d_h \log \big( O(\frac{1}{\epsilon}) \big)}{N \exp (\langle \bm{q}_{n, i}^{(t)}, \bm{k}_\pm^{(t)} \rangle)}.
\end{split}
\end{equation}

\begin{equation}
\begin{split}
    &\langle \bm{q}_{n, i}^{(t+1)}, \bm{k}_{n, j}^{(t+1)} \rangle - \langle \bm{q}_{n, i}^{(t)}, \bm{k}_{n, j}^{(t)} \rangle \ge - \frac{\eta C_{10} \sigma_p^4 d^2 \sigma_h^2 d_h \log \big( O(\frac{1}{\epsilon}) \big) }{N} \cdot \exp (\langle \bm{q}_{n, i}^{(t)}, \bm{k}_{n, j}^{(t)} \rangle).
\end{split}
\end{equation}

Then using the similar method as for \(\langle \bm{q}_+^{(t+1)}, \bm{k}_+^{(t+1)} \rangle\), we get
\[
    \langle \bm{q}_-^{(t+1)}, \bm{k}_-^{(t+1)} \rangle \le \log ( \epsilon^{-1} d_h^{\frac{1}{2}} ),
\]
\[
    \langle \bm{q}_\pm^{(t+1)}, \bm{k}_{n, j}^{(t+1)} \rangle \ge - \log ( \epsilon^{-1} d_h^{\frac{1}{2}} ),
\]
\[
    \langle \bm{q}_{n, i}^{(t+1)}, \bm{k}_\pm^{(t+1)} \rangle \le \log ( \epsilon^{-1} d_h^{\frac{1}{2}} ),
\]
\[
    \langle \bm{q}_{n, i}^{(t+1)}, \bm{k}_{n, j}^{(t+1)} \rangle \ge - \log ( \epsilon^{-1} d_h^{\frac{1}{2}} ),
\]
Next we provide the upper bound for \(\vert \langle \bm{q}_\pm^{(t+1)}, \bm{k}_\mp^{(t+1)} \rangle \vert, \vert \langle \bm{q}_{n, i}^{(t+1)}, \bm{k}_{n^\prime, j}^{(t+1)} \rangle \vert\). By the results of \ref{sum_alpha_beta2}, we have
\begin{equation}
\begin{split}
    &\sum\limits_{s = T_2}^{t}\vert \beta_{n, +, i}^{(t)} \vert, \sum\limits_{s = T_2}^{t}\vert \beta_{n, -, i}^{(t)} \vert = O \Big( \frac{\mathrm{SNR}^2 \big( \log (6N^2M^2 / \delta) \big)^{3} \log \big( O(\frac{1}{\epsilon}) \big) }{ \epsilon d_h^{\frac{1}{2}} } \Big),
\end{split}
\end{equation}
for \(i \in [M] \backslash \{1\}, n \in S_\pm\).
\begin{equation}
\begin{split}
    &\sum\limits_{s = T_2}^{t}\vert \alpha_{+, +}^{(t)} \vert, \sum\limits_{s = T_2}^{t}\vert \alpha_{-, -}^{(t)} \vert, \sum\limits_{s = T_2}^{t}\vert \beta_{+, +}^{(t)} \vert, \sum\limits_{s = T_2}^{t}\vert \beta_{-, -}^{(t)} \vert, \sum\limits_{s = T_2}^{t}\vert \beta_{n, i, +}^{(t)} \vert, \sum\limits_{s = T_2}^{t}\vert \beta_{n, i, -}^{(t)} \vert \\
    &= O \Big( \frac{ N \big( \log (6N^2M^2 / \delta) \big)^{3} \log \big( O(\frac{1}{\epsilon}) \big) }{\epsilon d_h^{\frac{1}{2}}} \Big),
\end{split}
\end{equation}
for \(i \in [M] \backslash \{1\}, n \in S_\pm\).
\begin{equation}
\begin{split}
    &\sum\limits_{s = T_2}^{t}\vert \alpha_{n, +, i}^{(t)} \vert, \sum\limits_{s = T_2}^{t}\vert \alpha_{n, -, i}^{(t)} \vert, \sum\limits_{s = T_2}^{t}\vert \alpha_{n, i, +}^{(t)} \vert, \sum\limits_{s = T_2}^{t}\vert \alpha_{n, i, -}^{(t)} \vert, \sum\limits_{s = T_2}^{t}\vert \alpha_{n, i, n, j}^{(t)} \vert, \sum\limits_{s = T_2}^{t}\vert \beta_{n, j, n, i}^{(t)} \vert \\
    &= O \Big( \frac{ \big( \log (6N^2M^2 / \delta) \big)^{3} \log \big( O(\frac{1}{\epsilon}) \big) }{\epsilon d_h^{\frac{1}{2}}} \Big),
\end{split}
\end{equation}
for \(i, j \in [M] \backslash \{1\}, n \in S_\pm\).
\begin{equation}
\begin{split}
    \sum\limits_{s = T_2}^{t}\vert \alpha_{n, i, n^\prime, j}^{(t)} \vert, \sum\limits_{s = T_2}^{t}\vert \beta_{n, j, n^\prime, i}^{(t)} \vert = O \Big( \frac{ \big( \log (6N^2M^2 / \delta) \big)^{4} \log \big( O(\frac{1}{\epsilon}) \big) }{\epsilon d^{\frac{1}{2}} d_h^{\frac{1}{2}}} \Big)
\end{split}
\end{equation}
for \(i, j \in [M] \backslash \{1\}, n, n^\prime \in [N], n \ne n^\prime\).
Plugging these and proposition \( \mathcal{G}(t) \) into the update rule of \(\vert \langle \bm{q}_\pm^{(t)}, \bm{k}_\mp^{(t)} \rangle \vert, \vert \langle \bm{q}_{n, i}^{(t)}, \bm{k}_{\overline{n}, j}^{(t)} \rangle \vert\) and get
\begin{equation}
\begin{split}
    &\vert \langle \bm{q}_+^{(t + 1)}, \bm{k}_-^{(t + 1)} \rangle \vert \le \vert \langle \bm{q}_+^{(T_2)}, \bm{k}_-^{(T_2)} \rangle \vert + \sum\limits_{s = T_2}^{t} \vert \langle \bm{q}_+^{(s+1)}, \bm{k}_-^{(s+1)} \rangle - \langle \bm{q}_+^{(s)}, \bm{k}_-^{(s)} \rangle \vert \\
    &\le \vert \langle \bm{q}_+^{(T_2)}, \bm{k}_-^{(T_2)} \rangle \vert \\
    &+ \sum\limits_{s = T_2}^{t} \Big\vert \alpha_{+, +}^{(s)} \langle \bm{k}_+^{(s)}, \bm{k}_-^{(s)} \rangle + \sum\limits_{n \in S_+} \sum\limits_{i=2}^M \alpha_{n, +, i}^{(s)} \langle \bm{k}_{n, i}^{(s)}, \bm{k}_-^{(s)} \rangle \\
    &+ \beta_{-, -}^{(s)} \langle \bm{q}_+^{(s)}, \bm{q}_-^{(s)} \rangle + \sum\limits_{n \in S_-} \sum\limits_{i=2}^M \beta_{n, -, i}^{(s)} \langle \bm{q}_{n, i}^{(s)}, \bm{q}_+^{(s)} \rangle \\
    &+ \Big( \alpha_{+, +}^{(s)} \bm{k}_+^{(s)} + \sum\limits_{n \in S_+} \sum\limits_{i=2}^M \alpha_{n, +, i}^{(s)} \bm{k}_{n, i}^{(s)} \Big) \\
    &\cdot \Big( \beta_{-, -}^{(s)} \bm{q}_-^{(s)\top} + \sum\limits_{n \in S_-} \sum\limits_{i=2}^M \beta_{n, -, i}^{(s)} \bm{q}_{n, i}^{(s)\top} \Big) \Big\vert \\
    &\le \vert \langle \bm{q}_+^{(T_2)}, \bm{k}_-^{(T_2)} \rangle \vert \\
    &+ \sum\limits_{s = T_2}^{t} \vert \alpha_{+, +}^{(t)} \vert \vert \langle \bm{k}_+^{(t)}, \bm{k}_-^{(t)} \rangle \vert + \sum\limits_{n \in S_+} \sum\limits_{i=2}^M \sum\limits_{s = T_2}^{t} \vert \alpha_{n, +, i}^{(t)} \vert \vert \langle \bm{k}_{n, i}^{(t)}, \bm{k}_-^{(t)} \rangle \vert \\
    &+ \sum\limits_{s = T_2}^{t} \vert \beta_{-, -}^{(t)} \vert \vert \langle \bm{q}_+^{(t)}, \bm{q}_-^{(t)} \rangle \vert + \sum\limits_{n \in S_-} \sum\limits_{i=2}^M \sum\limits_{s = T_2}^{t} \vert \beta_{n, -, i}^{(t)} \vert \vert \langle \bm{q}_{n, i}^{(t)}, \bm{q}_+^{(t)} \rangle \vert \\
    &+ \{lower \ order \ term\} \\
    &= \vert \langle \bm{q}_+^{(T_2)}, \bm{k}_-^{(T_2)} \rangle \vert \\
    &+ O \Big( \frac{ N \big( \log (6N^2M^2 / \delta) \big)^{3} \log \big( O(\frac{1}{\epsilon}) \big) }{\epsilon d_h^{\frac{1}{2}}} \Big) \cdot o(1) + N \cdot M \cdot O \Big( \frac{ \big( \log (6N^2M^2 / \delta) \big)^{3} \log \big( O(\frac{1}{\epsilon}) \big) }{\epsilon d_h^{\frac{1}{2}}} \Big) \cdot o(1) \\
    &+ O \Big( \frac{ N \big( \log (6N^2M^2 / \delta) \big)^{3} \log \big( O(\frac{1}{\epsilon}) \big) }{\epsilon d_h^{\frac{1}{2}}} \Big) \cdot o(1) + N \cdot M \cdot O \Big( \frac{\mathrm{SNR}^2 \big( \log (6N^2M^2 / \delta) \big)^{3} \log \big( O(\frac{1}{\epsilon}) \big) }{ \epsilon d_h^{\frac{1}{2}} } \Big) \cdot o(1) \\
    &=\vert \langle \bm{q}_+^{(T_2)}, \bm{k}_-^{(T_2)} \rangle \vert + o \Big( \frac{ N \big( \log (6N^2M^2 / \delta) \big)^{3} \log \big( O(\frac{1}{\epsilon}) \big) }{\epsilon d_h^{\frac{1}{2}}} \Big) + o \Big( \frac{N \cdot \mathrm{SNR}^2 \big( \log (6N^2M^2 / \delta) \big)^{3} \log \big( O(\frac{1}{\epsilon}) \big) }{ \epsilon d_h^{\frac{1}{2}} } \Big) \\
    &= o(1),
\end{split}
\end{equation}
where the first inequality is by triangle inequality, the second inequality is by \eqref{q_p_k_m_dy}, the last equality is by \(\vert \langle \bm{q}_+^{(T_2)}, \bm{k}_-^{(T_2)} \rangle \vert = o(1)\) and \(d_h = \widetilde{\Omega} \Big( \max \{\mathrm{SNR}^4, \mathrm{SNR}^{-4}\} N^2 \epsilon^{-2} \Big)\). Similarly we have \(\vert \langle \bm{q}_-^{(t + 1)}, \bm{k}_+^{(t + 1)} \rangle \vert = o(1)\).
\begin{equation}
\begin{split}
    &\vert \langle \bm{q}_{n, i}^{(t+1)}, \bm{k}_{\overline{n}, j}^{(t+1)} \rangle \vert \le \vert \langle \bm{q}_{n, i}^{(T_2)}, \bm{k}_{\overline{n}, j}^{(T_2)} \rangle \vert + \sum\limits_{s = T_2}^{t} \vert \langle \bm{q}_{n, i}^{(s+1)}, \bm{k}_{\overline{n}, j}^{(s+1)} \rangle - \langle \bm{q}_{n, i}^{(s)}, \bm{k}_{\overline{n}, j}^{(s)} \rangle \vert \\
    &\le \vert \langle \bm{q}_{n, i}^{(T_2)}, \bm{k}_{\overline{n}, j}^{(T_2)} \rangle \vert \\
    &+ \sum\limits_{s = T_2}^{t} \Big\vert \alpha_{n, i, +}^{(s)} \langle \bm{k}_+^{(s)}, \bm{k}_{\overline{n}, j}^{(s)} \rangle + \alpha_{n, i, -}^{(s)} \langle \bm{k}_-^{(s)}, \bm{k}_{\overline{n}, j}^{(s)} \rangle + \sum\limits_{n^\prime =1}^N \sum\limits_{l=2}^M \alpha_{n, i, n^\prime, l}^{(s)} \langle \bm{k}_{n^\prime, l}^{(s)}, \bm{k}_{\overline{n}, j}^{(s)} \rangle \\
    &+ \beta_{\overline{n}, j, +}^{(s)} \langle \bm{q}_+^{(s)}, \bm{q}_{n, i}^{(s)} \rangle + \beta_{\overline{n}, j, -}^{(s)} \langle \bm{q}_-^{(s)}, \bm{q}_{n, i}^{(s)} \rangle + \sum\limits_{n^\prime = 1}^N \sum\limits_{l=2}^M \beta_{\overline{n}, j, n^\prime, l}^{(s)} \langle \bm{q}_{n^\prime, l}^{(s)}, \bm{q}_{n, i}^{(s)} \rangle \\
    &+ \Big( \alpha_{n, i, +}^{(s)} \bm{k}_+^{(s)} + \alpha_{n, i, -}^{(s)} \bm{k}_-^{(s)} + \sum\limits_{n^\prime =1}^N \sum\limits_{l=2}^M \alpha_{n, i, n^\prime, l}^{(s)} \bm{k}_{n^\prime, l}^{(s)} \Big) \\
    &\cdot \Big( \beta_{\overline{n}, j, +}^{(s)} \bm{q}_+^{(s)\top} + \beta_{\overline{n}, j, -}^{(s)} \bm{q}_-^{(s)\top} + \sum\limits_{n^\prime = 1}^N \sum\limits_{l=2}^M \beta_{\overline{n}, j, n^\prime, l}^{(s)} \bm{q}_{n^\prime, l}^{(s)\top} \Big) \Big\vert \\
    &\le \vert \langle \bm{q}_{n, i}^{(T_2)}, \bm{k}_{\overline{n}, j}^{(T_2)} \rangle \vert \\
    &+ \sum\limits_{s = T_2}^{t} \vert \alpha_{n, i, +}^{(t)} \vert \vert \langle \bm{k}_+^{(t)}, \bm{k}_{\overline{n}, j}^{(t)} \rangle \vert + \sum\limits_{s = T_2}^{t} \vert \alpha_{n, i, -}^{(t)} \vert \vert \langle \bm{k}_-^{(t)}, \bm{k}_{\overline{n}, j}^{(t)} \rangle \vert \\
    &+ \sum\limits_{s = T_2}^{t} \vert \alpha_{n, i, \overline{n}, j}^{(t)} \vert \Vert \bm{k}_{\overline{n}, j}^{(t)} \Vert_2^2 + \sum\limits_{l=2}^M \sum\limits_{s = T_2}^{t} \vert \alpha_{n, i, n, l}^{(t)} \vert \vert \langle \bm{k}_{n, l}^{(t)}, \bm{k}_{\overline{n}, j}^{(t)} \rangle \vert \\
    &+ \sum\limits_{n^\prime \ne n \wedge (l \ne j \vee n^\prime \ne \overline{n})} \sum\limits_{t = T_2}^{T_3} \vert \alpha_{n, i, n^\prime, l}^{(t)} \vert \vert \langle \bm{k}_{n^\prime, l}^{(t)}, \bm{k}_{\overline{n}, j}^{(t)} \rangle \vert \\
    &+ \sum\limits_{s = T_2}^{t} \vert \beta_{\overline{n}, j, +}^{(t)} \vert \vert \langle \bm{q}_+^{(t)}, \bm{q}_{n, i}^{(t)} \rangle \vert + \sum\limits_{s = T_2}^{t} \vert \beta_{\overline{n}, j, -}^{(t)} \vert \vert \langle \bm{q}_-^{(t)}, \bm{q}_{n, i}^{(t)} \rangle \vert \\
    &+ \sum\limits_{s = T_2}^{t} \vert \beta_{\overline{n}, j, n, i}^{(t)} \vert \Vert \bm{q}_{n, i}^{(t)} \Vert_2^2 + \sum\limits_{l=2}^M \sum\limits_{s = T_2}^{t} \vert \beta_{\overline{n}, j, \overline{n}, l}^{(t)} \vert \vert \langle \bm{q}_{\overline{n}, l}^{(t)}, \bm{q}_{n, i}^{(t)} \rangle \vert \\
    &+ \sum\limits_{n^\prime \ne \overline{n} \wedge (l \ne i \vee n^\prime \ne n)} \sum\limits_{s = T_2}^{t} \vert \beta_{\overline{n}, j, n^\prime, l}^{(t)} \vert \vert \langle \bm{q}_{n^\prime, l}^{(t)}, \bm{q}_{n, i}^{(t)} \rangle \vert \\
    &+ \{lower \ order \ term\} \\
    &= \vert \langle \bm{q}_{n, i}^{(T_2)}, \bm{k}_{\overline{n}, j}^{(T_2)} \rangle \vert \\
    &+ O \Big( \frac{ \big( \log (6N^2M^2 / \delta) \big)^{3} \log \big( O(\frac{1}{\epsilon}) \big) }{\epsilon d_h^{\frac{1}{2}}} \Big) \cdot o(1) + O \Big( \frac{ \big( \log (6N^2M^2 / \delta) \big)^{4} \log \big( O(\frac{1}{\epsilon}) \big) }{\epsilon d^{\frac{1}{2}} d_h^{\frac{1}{2}}} \Big) \cdot \Theta ( \sigma_p^2 \sigma_h^2 d d_h ) \\
    &+ M \cdot O \Big( \frac{ \big( \log (6N^2M^2 / \delta) \big)^{3} \log \big( O(\frac{1}{\epsilon}) \big) }{\epsilon d_h^{\frac{1}{2}}} \Big) \cdot o(1) + N \cdot M \cdot O \Big( \frac{ \big( \log (6N^2M^2 / \delta) \big)^{4} \log \big( O(\frac{1}{\epsilon}) \big) }{\epsilon d^{\frac{1}{2}} d_h^{\frac{1}{2}}} \Big) \cdot o(1) \\
    &+ O \Big( \frac{ N \big( \log (6N^2M^2 / \delta) \big)^{3} \log \big( O(\frac{1}{\epsilon}) \big) }{\epsilon d_h^{\frac{1}{2}}} \Big) \cdot o(1) \\
    &= \vert \langle \bm{q}_{n, i}^{(T_2)}, \bm{k}_{\overline{n}, j}^{(T_2)} \rangle \vert + o \Big( N \frac{ \big( \log (6N^2M^2 / \delta) \big)^{3} \log \big( O(\frac{1}{\epsilon}) \big) }{\epsilon d_h^{\frac{1}{2}}} \Big) \\
    &+ O \Big( \frac{ \big( \log (6N^2M^2 / \delta) \big)^{3} \log \big( O(\frac{1}{\epsilon}) \big) }{\epsilon d^{\frac{1}{2}}} \Big) + o \Big( \frac{ N \big( \log (6N^2M^2 / \delta) \big)^{4} \log \big( O(\frac{1}{\epsilon}) \big) }{\epsilon d^{\frac{1}{2}} d_h^{\frac{1}{2}}} \Big) \\
    &= o(1),
\end{split}
\end{equation}
where the first inequality is by triangle inequality, the second inequality is by \eqref{q_i_k_j_ne_dy}, the second equality is by \( \sigma_h^2 \le \min \{ \Vert \bm{\mu} \Vert_2^{-2}, ( \sigma_p^2 d )^{-1} \} \cdot d_h^{-\frac{1}{2}} \cdot \big( \log ( 6N^2M^2 / \delta ) \big)^{-\frac{3}{2}}\), the last equality is by \(\vert \langle \bm{q}_{n, i}^{(T_2)}, \bm{k}_{\overline{n}, j}^{(T_2)} \rangle \vert = o(1)\), \(d = \widetilde{\Omega} \Big( \epsilon^{-2} N^2 d_h \Big) \) and \(d_h = \widetilde{\Omega} \Big( \max \{\mathrm{SNR}^4, \mathrm{SNR}^{-4}\} N^2 \epsilon^{-2} \Big)\).

\begin{lemma}[Convergence of Training Loss]
    \label{stage3_loss}
    There exist \(T = \frac{C_{19}}{\eta \epsilon \Vert \bm{\mu} \Vert_2^2 \Vert \bm{w}_O \Vert_2^2}\) such that
    \[
        L_S(\theta(T)) \le \epsilon
    \]
\end{lemma}
\it Proof of Lemma \ref{stage3_loss}. \rm
Recall \eqref{q_pm_k_j_attn}, \eqref{q_i_k_j_attn}, \eqref{q_pm_k_pm_attn} and \eqref{q_pm_k_j_attn}, we have
\begin{equation}
    \label{q_pm_k_j_attn2}
    softmax(\langle \bm{q}_\pm^{(T)}, \bm{k}_{n, j}^{(t)} \rangle) = O \Big( \frac{N \big( \log (6N^2M^2 / \delta) \big)^{3}}{d_h^{\frac{1}{2}}} \Big),
\end{equation}
\begin{equation}
    \label{q_i_k_j_attn2}
    softmax(\langle \bm{q}_{n, i}^{(t)}, \bm{k}_{n, j}^{(t)} \rangle) = O \Big( \frac{N \Vert \bm{\mu} \Vert_2^2 \big( \log (6N^2M^2 / \delta) \big)^{3}}{ \sigma_p^2 d d_h^{\frac{1}{2}} } \Big) = O \Big( \frac{N \cdot \mathrm{SNR}^2 \cdot \big( \log (6N^2M^2 / \delta) \big)^{3}}{ d_h^{\frac{1}{2}} } \Big),
\end{equation}
\begin{equation}
    \label{q_pm_k_pm_attn2}
    softmax(\langle \bm{q}_\pm^{(t)}, \bm{k}_\pm^{(t)} \rangle) = 1 - O \Big( \frac{N \big( \log (6N^2M^2 / \delta) \big)^{3}}{d_h^{\frac{1}{2}}} \Big),
\end{equation}
\begin{equation}
    \label{q_i_k_pm_attn2}
    softmax(\langle \bm{q}_{n, i}^{(t)}, \bm{k}_\pm^{(t)} \rangle) = 1 - O \Big( \frac{N \cdot \mathrm{SNR}^2 \cdot \big( \log (6N^2M^2 / \delta) \big)^{3}}{ d_h^{\frac{1}{2}} } \Big)
\end{equation}

Substituting \(t = T = \frac{C_{19}}{\eta \epsilon \Vert \bm{\mu} \Vert_2^2 \Vert \bm{w}_O \Vert_2^2} \) into propositions \( \mathcal{F}(t) \) and get
\begin{equation}
\begin{split}
    \label{V_p_bound3}
    V_+^{(t)} &\ge \log \Big( \exp ( V_{+}^{(T_2)}) + \eta C_{17} \Vert \bm{\mu} \Vert_2^2 \Vert \bm{w}_O \Vert_2^2 (t - T_2) \Big) \\
    &\ge \log \Big( \exp ( V_{+}^{(T_2)}) + \frac{C_{20}}{\epsilon} \Big) \\
    &\ge \log \Big( \frac{C_{20}}{\epsilon} \Big),
\end{split}
\end{equation}
\begin{equation}
    \label{V_i_bound3}
    \vert V_{n, i}^{(t)} \vert = O (1).
\end{equation}
For \(n \in S_+\), we bound \(f(\bm{X}_n, \theta (t))\) as follows
\begin{equation}
\begin{split}
    f(\bm{X}_n, \theta (t)) &= \frac{1}{M} \sum\limits_{l=1}^M \bm{\varphi} (\bm{x}_{n, l}^\top \bm{W}_{Q} \bm{W}_{K}^{\top} (\bm{X}_n)^\top ) \bm{X}_n \bm{W}_V \bm{w}_O \\
    &\frac{1}{M} \cdot \Big( \big( \frac{ \exp (\langle \bm{q}_+^{(t)}, \bm{k}_+^{(t)} \rangle)}{ \exp  (\langle \bm{q}_+^{(t)}, \bm{k}_+^{(t)} \rangle) + \sum\limits_{j=2}^M  \exp  ( \langle \bm{q}_+^{(t)}, \bm{k}_{n, j}^{(t)} \rangle ) } \\
    &+ \sum\limits_{i=2}^{M} \frac{ \exp (\langle \bm{q}_{n, i}^{(t)}, \bm{k}_+^{(t)} \rangle)}{ \exp  (\langle \bm{q}_{n, i}^{(t)}, \bm{k}_+^{(t)} \rangle) + \sum\limits_{j=2}^M  \exp  ( \langle \bm{q}_{n, i}^{(t)}, \bm{k}_{n, j}^{(t)} \rangle ) } \big) \cdot V_+^{(T)} \\
    &+ \sum\limits_{i=2}^{M} \big( \frac{ \exp (\langle \bm{q}_+^{(t)}, \bm{k}_{n, i}^{(t)} \rangle)}{ \exp  (\langle \bm{q}_+^{(t)}, \bm{k}_+^{(t)} \rangle) + \sum\limits_{j=2}^M  \exp  ( \langle \bm{q}_+^{(t)}, \bm{k}_{n, j}^{(t)} \rangle ) } \\
    &+ \sum\limits_{l=2}^{M} \frac{\exp (\langle \bm{q}_{n, l}^{(t)}, \bm{k}_{n, i}^{(t)} \rangle)}{\exp (\langle \bm{q}_{n, l}^{(t)}, \bm{k}_+^{(t)} \rangle) + \sum\limits_{j=2}^M \exp ( \langle \bm{q}_{n, l}^{(t)}, \bm{k}_{n, j}^{(t)} \rangle )} \big) \cdot V_{n, i}^{(t)} \Big) \\
    &\ge \frac{1}{M} \cdot \Big( 1 - O \Big( \frac{N \big( \log (6N^2M^2 / \delta) \big)^{3}}{d_h^{\frac{1}{2}}} \Big) \\
    &+ (M - 1) \cdot \Big( 1 - O \big( \frac{N \cdot \mathrm{SNR}^2 \cdot \big( \log (6N^2M^2 / \delta) \big)^{3}}{ d_h^{\frac{1}{2}} } \big) \Big) \Big) \cdot \log \Big( \frac{C_{20}}{\epsilon} \Big) \\
    &- \frac{1}{M} \cdot (M - 1) \cdot \Big( O \Big( \frac{N \big( \log (6N^2M^2 / \delta) \big)^{3}}{d_h^{\frac{1}{2}}} \Big) \\
    &+ (M - 1) \cdot O \big( \frac{N \cdot \mathrm{SNR}^2 \cdot \big( \log (6N^2M^2 / \delta) \big)^{3}}{ d_h^{\frac{1}{2}} } \big) \Big) \cdot O(1) \\
    &= \log \Big( \frac{C_{20}}{\epsilon} \Big) - O \Big( \frac{N \cdot \mathrm{SNR}^2 \cdot \big( \log (6N^2M^2 / \delta) \big)^{3} \log \Big( \frac{C_{20}}{\epsilon} \Big)}{ d_h^{\frac{1}{2}} } \Big) \\
    &\ge \log \Big( \frac{C_{20}}{\epsilon} \Big) - \log(C_{20}) \\
    &\ge \log \Big( \frac{1}{\epsilon} \Big),
\end{split}
\end{equation}
where the first inequality is by plugging \eqref{q_pm_k_j_attn2}, \eqref{q_i_k_j_attn2}, \eqref{q_pm_k_pm_attn2} and \eqref{q_i_k_pm_attn2}. For the second inequality, by \(d_h = \widetilde{\Omega} \Big( \max \{\mathrm{SNR}^4, \mathrm{SNR}^{-4}\} N^2 \epsilon^{-2} \Big)\), we have \(O \Big( \frac{N \cdot \mathrm{SNR}^2 \cdot \big( \log (6N^2M^2 / \delta) \big)^{3} \log \Big( \frac{C_{20}}{\epsilon} \Big)}{ d_h^{\frac{1}{2}} } \Big) = o(1) = \log(1 + o(1)) \le \log(C_{20})\) as long as \(C_{20}\) is sufficiently large.
Then we have
\begin{equation}
\begin{split}
    \ell_n^{(t)} &= \log \Big( 1 + \exp(- f(\bm{X}_n, \theta (t))) \Big) \\
    &\le \exp(- f(\bm{X}_n, \theta (t))) \\
    &\le \exp\Big(- \log \Big( \frac{1}{\epsilon} \Big) \Big) \\
    &\le \epsilon.
\end{split}
\end{equation}
Similarly, we have \(\ell_n^{(t)} \le \epsilon\) for \(n \in S_-\). Therefore, we have \(L_S(\theta(T)) = \frac{1}{N}\sum\limits_{n = 1}^{N} \ell_n^{(t)} \le \epsilon\).

\subsection{Population Loss}
\label{population_loss_section}
Consider a new data point \((\bm{X}, y)\) generated from the distribution defined in Definition \ref{data_def}. Without loss of generality, we suppose that the signal token is \(\bm{\mu}_+\) and the label is 1, i.e. \(\bm{X} = ( \bm{\mu}_+, \bm{\xi}_2, \dots, \bm{\xi}_M) \), \(y = 1\). 
We will first give the bounds of attention, then the bounds of the output of ViT, and finally the bound of the expectation of loss.
\begin{lemma}[Bound of Attention]
    \label{attn}
    Under the same conditions as Theorem \ref{benign}, with probability at least \(1 - \delta / 2N^2M\), we have that 
    \[
        \frac{ \exp ( \bm{\mu}_+^{\top} \bm{W}_Q^{(T_3)} \bm{W}_K^{(T_3)\top} \bm{\mu}_+ )}{ \exp ( \bm{\mu}_+^{\top} \bm{W}_Q^{(T_3)} \bm{W}_K^{(T_3)\top} \bm{\mu}_+ ) + \sum\limits_{i=2}^M  \exp ( \bm{\mu}_+^{\top} \bm{W}_Q^{(T_3)} \bm{W}_K^{(T_3)\top} \bm{\xi}_i )} \ge \frac{1}{2M}.
    \]
\end{lemma}
\it{Proof of Lemma} \ref{attn}. \rm By the result of \ref{stage2} and \ref{stage3}, \(\bm{\mu}_+^{\top} \bm{W}_Q^{(T_1)} \bm{W}_K^{(T_1)\top} \bm{\mu}_+ = o(1)\), and \(\bm{\mu}_+^{\top} \bm{W}_Q^{(t)} \bm{W}_K^{(t)\top} \bm{\mu}_+\) is monotonically non-decreasing for \(t \in [T_1, T_3]\), we have \(\bm{\mu}_+^{\top} \bm{W}_Q^{(T_3)} \bm{W}_K^{(T_3)\top} \bm{\mu}_+ \ge - o(1)\).

By proposition \( \mathcal{G}(t) \) in \ref{stage3}, we have \(\Vert \bm{\mu}_+^{\top} \bm{W}_Q^{(T_3)} \Vert_2^2 = \Theta \big( \Vert \bm{\mu} \Vert_2^2 \sigma_h^2 d_h \big)\). In order to give bounds for \(\bm{\mu}_+^{\top} \bm{W}_Q^{(T_3)} \bm{W}_K^{(T_3)\top} \bm{\xi}_i\), we provide a bound for the F-norm of \(\bm{W}_K^{(T_3)} - \bm{W}_K^{(0)}\).
According to the properties of F-norm, we have 
\begin{equation}
    \label{delta_K}
    \Vert \bm{W}_K^{(T_3)} - \bm{W}_K^{(0)} \Vert_F \le \Vert \bm{W}_K^{(T_1)} - \bm{W}_K^{(0)} \Vert_F + \Vert \bm{W}_K^{(T_2)} - \bm{W}_K^{(T_1)} \Vert_F + \Vert \bm{W}_K^{(T_3)} - \bm{W}_K^{(T_2)} \Vert_F
\end{equation}
Next, we will provide the bounds of \(\Vert \bm{W}_K^{(T_1)} - \bm{W}_K^{(0)} \Vert_F\), \(\Vert \bm{W}_K^{(T_2)} - \bm{W}_K^{(T_1)} \Vert_F\), and \(\Vert \bm{W}_K^{(T_3)} - \bm{W}_K^{(T_2)} \Vert_F\) respectively.
\\
\textbf{Bounding \(\Vert \bm{W}_K^{(T_1)} - \bm{W}_K^{(0)} \Vert_F\):} Recall that
\begin{equation*}
\begin{split}
    \nabla_{\bm{W}_K}L_S(\theta)
    &= \frac{1}{NM} \sum\limits_{n \in S_+} \ell_n^{\prime}(\theta) ( \bm{X}_n^\top (diag(\bm{\varphi}_{n, 1}) - \bm{\varphi}_{n, 1}^\top \bm{\varphi}_{n, 1}) \bm{X}_n \bm{W}_V \bm{w}_O \bm{\mu}_+^\top \\
    &+ \sum\limits_{i=2}^M \bm{X}_n^\top (diag(\bm{\varphi}_{n, i}) - \bm{\varphi}_{n, i}^\top \bm{\varphi}_{n, i}) \bm{X}_n \bm{W}_V \bm{w}_O \bm{\xi}_{n, i}^\top ) \bm{W}_Q \\
    &- \frac{1}{NM} \sum\limits_{n \in S_-} \ell_n^{\prime}(\theta) ( \bm{X}_n^\top (diag(\bm{\varphi}_{n, 1}) - \bm{\varphi}_{n, 1}^\top \bm{\varphi}_{n, 1}) \bm{X}_n \bm{W}_V \bm{w}_O \bm{\mu}_-^\top \\
    &+ \sum\limits_{i=2}^M \bm{X}_n^\top (diag(\bm{\varphi}_{n, i}) - \bm{\varphi}_{n, i}^\top \bm{\varphi}_{n, i}) \bm{X}_n \bm{W}_V \bm{w}_O \bm{\xi}_{n, i}^\top ) \bm{W}_Q ,
\end{split}
\end{equation*}
so we have
\begin{equation}
\begin{split}
    \label{F_norm_dy}
    &\Vert \bm{W}_K^{(t+1)} - \bm{W}_K^{(t)} \Vert_F = \Vert \eta \nabla_{\bm{W}_K}L_S(\theta(t)) \Vert_F \\
    &\le \frac{\eta}{NM} \sum\limits_{n \in S_+} \vert \ell_n^{\prime(t)} \vert \big( \Vert \bm{X}_n^{(t)\top} \Vert_F \Vert (diag(\bm{\varphi}_{n, 1}^{(t)}) - \bm{\varphi}_{n, 1}^{(t)\top} \bm{\varphi}_{n, 1}^{(t)}) \Vert_F \Vert \bm{X}_n \bm{W}_V^{(t)} \bm{w}_O \Vert_F \Vert \bm{\mu}_+^\top \bm{W}_Q^{(t)} \Vert_F \\
    &+ \sum\limits_{i=2}^M \Vert \bm{X}_n^\top \Vert_F \Vert (diag(\bm{\varphi}_{n, i}^{(t)}) - \bm{\varphi}_{n, i}^{(t)\top} \bm{\varphi}_{n, i}^{(t)}) \Vert_F \Vert \bm{X}_n \bm{W}_V^{(t)} \bm{w}_O \Vert_F \Vert \bm{\xi}_{n, i}^\top \bm{W}_Q^{(t)} \Vert_F \big) \\
    &+ \frac{\eta}{NM} \sum\limits_{n \in S_-} \vert \ell_n^{\prime(t)} \vert \big( \Vert \bm{X}_n^\top \Vert_F \Vert (diag(\bm{\varphi}_{n, 1}^{(t)}) - \bm{\varphi}_{n, 1}^{(t)\top} \bm{\varphi}_{n, 1}^{(t)}) \Vert_F \Vert \bm{X}_n \bm{W}_V^{(t)} \bm{w}_O \Vert_F \Vert \bm{\mu}_-^\top \bm{W}_Q^{(t)} \Vert_F \\
    &+ \sum\limits_{i=2}^M \Vert \bm{X}_n^\top \Vert_F \Vert (diag(\bm{\varphi}_{n, i}^{(t)}) - \bm{\varphi}_{n, i}^{(t)\top} \bm{\varphi}_{n, i}^{(t)}) \Vert_F \Vert \bm{X}_n \bm{W}_V^{(t)} \bm{w}_O \Vert_F \Vert \bm{\xi}_{n, i}^\top \bm{W}_Q^{(t)} \Vert_F \big)
\end{split}
\end{equation}
We bound \(\vert \ell_n^{\prime(t)} \vert\), \(\Vert \bm{X}_n \Vert_F\), \(\Vert (diag(\bm{\varphi}_{n, i}^{(t)}) - \bm{\varphi}_{n, i}^{(t)\top} \bm{\varphi}_{n, i}^{(t)}) \Vert_F\), \(\Vert \bm{X}_n \bm{W}_V^{(t)} \bm{w}_O \Vert_F\), \(\Vert \bm{\mu}_\pm^\top \bm{W}_Q^{(t)} \Vert_F\) and \(\Vert \bm{\xi}_{n, i}^\top \bm{W}_Q^{(t)} \Vert_F\) respectively.
\begin{itemize}
    \item \(\vert \ell_n^{\prime(t)} \vert \le 1\)
    \item By Lemma \ref{caoyuan}, \(\Vert \bm{X}_n \Vert_F \le \sqrt{\Vert \bm{\mu} \Vert_2^2 + (M - 1) \frac{3\sigma_p^2 d}{2}} = O \big( \max\{\Vert \bm{\mu} \Vert_2, \sigma_p \sqrt{d}\} \big) \)
    \item Note that each element of vector \(\bm{\varphi}_{n, i}\) has an upper bound 1, and the number of tokens has a constant upper bound (\(M = \Theta(1)\)), then we have \(\Vert (diag(\bm{\varphi}_{n, i}^{(t)}) - \bm{\varphi}_{n, i}^{(t)\top} \bm{\varphi}_{n, i}^{(t)}) \Vert_F = O(1)\).
    \item By Lemma \ref{upper_bound_of_V}, we have \(\vert V_+^{(t)} \vert, \vert V_-^{(t)} \vert, \vert V_{n, i}^{(t)} \vert = O (d_h^{-\frac{1}{4}})\), so we have 
    \[
        \Vert \bm{X}_n \bm{W}_V \bm{w}_O \Vert_F = \sqrt{\big( V_+^{(t)} \big)^2 + \sum\limits_{i = 2}^M \big( V_{n, i}^{(t)} \big)^2 } = \sqrt{M \cdot \big( d_h^{-\frac{1}{4}} \big)^2} = O \big( d_h^{-\frac{1}{4}} \big)
    \]
    \item By Lemma \ref{inner_products_hold_magnitude}, we have \(\Vert \bm{\mu}_\pm^\top \bm{W}_Q^{(t)} \Vert_F = \Theta \big( \Vert \bm{\mu} \Vert_2 \sigma_h d_h^{\frac{1}{2}} \big)\), \(\Vert \bm{\xi}_{n, i}^\top \bm{W}_Q^{(t)} \Vert_F = \Theta \big( \sigma_p \sigma_h d^{\frac{1}{2}} d_h^{\frac{1}{2}} \big)\)
\end{itemize}
Plugging them into \eqref{F_norm_dy} and get
\begin{equation}
\begin{split}
    \label{F_norm_dy1}
    \Vert \bm{W}_K^{(t+1)} - \bm{W}_K^{(t)} \Vert_F &\le \frac{\eta}{NM} \cdot NM \cdot O \big( \max\{\Vert \bm{\mu} \Vert_2, \sigma_p \sqrt{d}\} \big) \cdot O \big( d_h^{-\frac{1}{4}} \big) \cdot O \big( \max\{\Vert \bm{\mu} \Vert_2, \sigma_p \sqrt{d}\} \sigma_h d_h^{\frac{1}{2}} \big) \\
    &= O \big( \eta \cdot \max\{\Vert \bm{\mu} \Vert_2^2, \sigma_p^2 d\} \cdot \sigma_h d_h^{\frac{1}{4}} \big)
\end{split}
\end{equation}
Taking a summation we have
\begin{equation}
\begin{split}
    \label{delta_K_1}
    \Vert \bm{W}_K^{(T_1)} - \bm{W}_K^{(0)} \Vert_F &\le \sum\limits_{s=0}^{T_1 - 1} \Vert \bm{W}_K^{(t+1)} - \bm{W}_K^{(t)} \Vert_F \\
    &\le O \big( \frac{1}{\eta d_h^{\frac{1}{4}} \Vert \bm{\mu} \Vert_2^2 \Vert \bm{w}_O \Vert_2^2} \big) \cdot O \big( \eta \cdot \max\{\Vert \bm{\mu} \Vert_2^2, \sigma_p^2 d\} \cdot \sigma_h d_h^{\frac{1}{4}} \big) \\
    &= O \Big( \frac{\max\{\Vert \bm{\mu} \Vert_2^2, \sigma_p^2 d\}}{\Vert \bm{\mu} \Vert_2^2} \cdot \sigma_h \Big) \\
    &\le O \Big( N \sigma_h \Big),
\end{split}
\end{equation}
where the last inequality is by \(N \cdot \mathrm{SNR}^2 = \Omega(1)\).
\\
\textbf{Bounding \(\Vert \bm{W}_K^{(T_2)} - \bm{W}_K^{(T_1)} \Vert_F\):} By \ref{sum_alpha_beta}, we know that each element of matrix \((diag(\bm{\varphi}_{n, 1}^{(t)}) - \bm{\varphi}_{n, 1}^{(t)\top} \bm{\varphi}_{n, 1}^{(t)})\) have upper bound \( O \Big( \frac{1}{ 1 + \frac{\eta^2 \Vert \bm{\mu} \Vert_2^4 \Vert \bm{w}_O \Vert_2^2 d_h^{\frac{1}{2}}}{N \big( \log (6N^2M^2 / \delta) \big)^{2} } \cdot (t - T_1)(t - T_1 - 1) } \Big)\), and each element of matrix \((diag(\bm{\varphi}_{n, i}^{(t)}) - \bm{\varphi}_{n, i}^{(t)\top} \bm{\varphi}_{n, i}^{(t)})\) have upper bound \( O \Big( \frac{1}{ 1 + \frac{\eta^2 \sigma_p^2 d \Vert \bm{\mu} \Vert_2^2 \Vert \bm{w}_O \Vert_2^2 d_h^{\frac{1}{2}} }{N \big( \log (6N^2M^2 / \delta) \big)^{2}} \cdot (t - T_1)(t - T_1 - 1) } \Big)\). 
Considering that the number of tokens has a constant upper bound (\(M = \Theta(1)\)), we have

\begin{equation}
\begin{split}
    \Vert (diag(\bm{\varphi}_{n, 1}^{(t)}) - \bm{\varphi}_{n, 1}^{(t)\top} \bm{\varphi}_{n, 1}^{(t)}) \Vert_F = O \Big( \frac{1}{ 1 + \frac{\eta^2 \Vert \bm{\mu} \Vert_2^4 \Vert \bm{w}_O \Vert_2^2 d_h^{\frac{1}{2}}}{N \big( \log (6N^2M^2 / \delta) \big)^{2} } \cdot (t - T_1)(t - T_1 - 1) } \Big)
\end{split}
\end{equation}
\begin{equation}
\begin{split}
    \Vert (diag(\bm{\varphi}_{n, i}^{(t)}) - \bm{\varphi}_{n, i}^{(t)\top} \bm{\varphi}_{n, i}^{(t)}) \Vert_F = O \Big( \frac{1}{ 1 + \frac{\eta^2 \sigma_p^2 d \Vert \bm{\mu} \Vert_2^2 \Vert \bm{w}_O \Vert_2^2 d_h^{\frac{1}{2}} }{N \big( \log (6N^2M^2 / \delta) \big)^{2}} \cdot (t - T_1)(t - T_1 - 1) } \Big)
\end{split}
\end{equation}
By \( \mathcal{B}(t) \) in \ref{stage2}, we have \(\vert V_\pm^{(t)} \vert, \vert V_{n, i}^{(t)} \vert \le O(d_h^{-\frac{1}{4}}) + \eta C_4 \Vert \bm{\mu} \Vert_2^2 \Vert \bm{w}_O \Vert_2^2 (t - T_1)\) for \(t \in [T_1, T_2]\). We further have
\[
    \Vert \bm{X}_n \bm{W}_V^{(t)} \bm{w}_O \Vert_F = O \Big( d_h^{-\frac{1}{4}} + \eta \Vert \bm{\mu} \Vert_2^2 \Vert \bm{w}_O \Vert_2^2 (t - T_1) \Big).
\]
Then we have
\begin{equation}
\begin{split}
    &\Vert (diag(\bm{\varphi}_{n, 1}^{(t)}) - \bm{\varphi}_{n, 1}^{(t)\top} \bm{\varphi}_{n, 1}^{(t)}) \Vert_F \cdot \Vert \bm{X}_n \bm{W}_V^{(t)} \bm{w}_O \Vert_F \\
    &= O \Big( \frac{1}{ 1 + \frac{\eta^2 \Vert \bm{\mu} \Vert_2^4 \Vert \bm{w}_O \Vert_2^2 d_h^{\frac{1}{2}}}{N \big( \log (6N^2M^2 / \delta) \big)^{2} } \cdot (t - T_1)(t - T_1 - 1) } \Big) \cdot O \Big( d_h^{-\frac{1}{4}} + \eta \Vert \bm{\mu} \Vert_2^2 \Vert \bm{w}_O \Vert_2^2 (t - T_1) \Big) \\
    &= O \Big( N^{\frac{1}{2}} d_h^{-\frac{1}{4}} \log (6N^2M^2 / \delta) \Big),
\end{split}
\end{equation}
The reason for the last equation is similar to \eqref{V_attn_upper_bound}. We also have
\begin{equation}
\begin{split}
    &\Vert (diag(\bm{\varphi}_{n, i}^{(t)}) - \bm{\varphi}_{n, i}^{(t)\top} \bm{\varphi}_{n, i}^{(t)}) \Vert_F \cdot \Vert \bm{X}_n \bm{W}_V^{(t)} \bm{w}_O \Vert_F \\
    &= O \Big( \frac{1}{ 1 + \frac{\eta^2 \sigma_p^2 d \Vert \bm{\mu} \Vert_2^2 \Vert \bm{w}_O \Vert_2^2 d_h^{\frac{1}{2}} }{N \big( \log (6N^2M^2 / \delta) \big)^{2}} \cdot (t - T_1)(t - T_1 - 1) } \Big) \cdot O \Big( d_h^{-\frac{1}{4}} + \eta \Vert \bm{\mu} \Vert_2^2 \Vert \bm{w}_O \Vert_2^2 (t - T_1) \Big) \\
    &= O \Big( \frac{\Vert \bm{\mu} \Vert_2}{\sigma_p \sqrt{d}} N^{\frac{1}{2}} d_h^{-\frac{1}{4}} \log (6N^2M^2 / \delta) \Big).
\end{split}
\end{equation}
Plugging them into \eqref{F_norm_dy} and get
\begin{equation}
\begin{split}
    \label{F_norm_dy2}
    \Vert \bm{W}_K^{(t+1)} - \bm{W}_K^{(t)} \Vert_F &\le \frac{\eta}{NM} \cdot N \cdot O \big( \max\{\Vert \bm{\mu} \Vert_2, \sigma_p \sqrt{d}\} \big) \cdot \Big( O \Big( N^{\frac{1}{2}} d_h^{-\frac{1}{4}} \log (6N^2M^2 / \delta) \Big) \cdot O \big( \Vert \bm{\mu} \Vert_2 \sigma_h d_h^{\frac{1}{2}} \big) \\
    &+ (M - 1) \cdot O \Big( \frac{\Vert \bm{\mu} \Vert_2}{\sigma_p \sqrt{d}} N^{\frac{1}{2}} d_h^{-\frac{1}{4}} \log (6N^2M^2 / \delta) \Big) \cdot O \big( \sigma_p \sqrt{d} \sigma_h d_h^{\frac{1}{2}} \big) \Big) \\
    &= O \Big( \eta \cdot \max\{\Vert \bm{\mu} \Vert_2, \sigma_p \sqrt{d}\} N^{\frac{1}{2}} \Vert \bm{\mu} \Vert_2 \sigma_h d_h^{\frac{1}{4}} \log (6N^2M^2 / \delta) \Big)
\end{split}
\end{equation}
Taking a summation we have
\begin{equation}
\begin{split}
    \label{delta_K_2}
    \Vert \bm{W}_K^{(T_2)} - \bm{W}_K^{(T_1)} \Vert_F &\le \sum\limits_{s=T_1}^{T_2 - 1} \Vert \bm{W}_K^{(t+1)} - \bm{W}_K^{(t)} \Vert_F \\
    &\le O \big( \frac{1}{\eta \Vert \bm{\mu} \Vert_2^2 \Vert \bm{w}_O \Vert_2^2} \big) \cdot O \Big( \eta \max\{\Vert \bm{\mu} \Vert_2, \sigma_p \sqrt{d}\} N^{\frac{1}{2}} \Vert \bm{\mu} \Vert_2 \sigma_h d_h^{\frac{1}{4}} \log (6N^2M^2 / \delta) \Big) \\
    &= O \Big( \frac{\max\{\Vert \bm{\mu} \Vert_2, \sigma_p \sqrt{d}\}}{\Vert \bm{\mu} \Vert_2} \cdot N^{\frac{1}{2}} \sigma_h d_h^{\frac{1}{4}} \log (6N^2M^2 / \delta) \Big) \\
    &\le O \Big( N^{\frac{1}{2}}(1 + \mathrm{SNR}^{-1}) \sigma_h d_h^{\frac{1}{4}} \log (6N^2M^2 / \delta) \Big),
\end{split}
\end{equation}
\\
\textbf{Bounding \(\Vert \bm{W}_K^{(T_3)} - \bm{W}_K^{(T_2)} \Vert_F\):} Recall that (\eqref{q_p_k_j_upper_bound_stage3}, \eqref{q_i_k_j_upper_bound_stage3})
\[
    \frac{ \exp (\langle \bm{q}_\pm^{(t)}, \bm{k}_{n, j}^{(t)} \rangle)}{ \exp  (\langle \bm{q}_\pm^{(t)}, \bm{k}_\pm^{(t)} \rangle) + \sum\limits_{j^\prime=2}^M  \exp  ( \langle \bm{q}_\pm^{(t)}, \bm{k}_{n, j^\prime}^{(t)} \rangle ) } = O \Big( \frac{N \big( \log (6N^2M^2 / \delta) \big)^{3} }{d_h^{\frac{1}{2}}} \Big),
\]
\[
    \frac{ \exp (\langle \bm{q}_{n, i}^{(t)}, \bm{k}_{n, j}^{(t)} \rangle)}{ \exp  (\langle \bm{q}_{n, i}^{(t)}, \bm{k}_+^{(t)} \rangle) + \sum\limits_{j^\prime=2}^M  \exp  ( \langle \bm{q}_{n, i}^{(t)}, \bm{k}_{n, j^\prime}^{(t)} \rangle ) } = O \Big( \frac{N \Vert \bm{\mu} \Vert_2^2 \big( \log (6N^2M^2 / \delta) \big)^{3}}{ \sigma_p^2 d d_h^{\frac{1}{2}} } \Big)
\]
for \(i, j \in [M] \backslash \{1\}, n \in S_\pm, t \in [T_2, T_3]\). We have
\begin{equation}
\begin{split}
    \Vert (diag(\bm{\varphi}_{n, 1}^{(t)}) - \bm{\varphi}_{n, 1}^{(t)\top} \bm{\varphi}_{n, 1}^{(t)}) \Vert_F = O \Big( \frac{N \big( \log (6N^2M^2 / \delta) \big)^{3} }{d_h^{\frac{1}{2}}} \Big),
\end{split}
\end{equation}
\begin{equation}
\begin{split}
    \Vert (diag(\bm{\varphi}_{n, i}^{(t)}) - \bm{\varphi}_{n, i}^{(t)\top} \bm{\varphi}_{n, i}^{(t)}) \Vert_F = O \Big( \frac{N \Vert \bm{\mu} \Vert_2^2 \big( \log (6N^2M^2 / \delta) \big)^{3}}{ \sigma_p^2 d d_h^{\frac{1}{2}} } \Big).
\end{split}
\end{equation}
By \( \mathcal{F}(t) \) in \ref{stage3}, we have
\[
    \vert V_\pm^{(t)} \vert \le 2 \log \big( O(\frac{1}{\epsilon}) \big)
\]
\[
    \vert V_{n, i}^{(t)} \vert = O(1)
\]
we further have
\[
    \Vert \bm{X}_n \bm{W}_V^{(t)} \bm{w}_O \Vert_F = O \Big( \log O\big(\frac{1}{\epsilon} \big) \Big).
\]
Plugging them into \eqref{F_norm_dy} and get
\begin{equation}
\begin{split}
    \label{F_norm_dy3}
    \Vert \bm{W}_K^{(t+1)} - \bm{W}_K^{(t)} \Vert_F &\le \frac{\eta}{NM} \cdot N \cdot O \big( \max\{\Vert \bm{\mu} \Vert_2, \sigma_p \sqrt{d}\} \big) \cdot \Big( O \Big( \frac{N \big( \log (6N^2M^2 / \delta) \big)^{3} }{d_h^{\frac{1}{2}}} \Big) \cdot O \big( \Vert \bm{\mu} \Vert_2 \sigma_h d_h^{\frac{1}{2}} \big) \\
    &+ (M - 1) \cdot O \Big( \frac{N \Vert \bm{\mu} \Vert_2^2 \big( \log (6N^2M^2 / \delta) \big)^{3}}{ \sigma_p^2 d d_h^{\frac{1}{2}} } \Big) \cdot O \big( \sigma_p \sqrt{d} \sigma_h d_h^{\frac{1}{2}} \big) \Big) \cdot O \Big( \log \big( O(\frac{1}{\epsilon}) \big) \Big) \\
    &= O \Big( \eta \cdot \max\{\Vert \bm{\mu} \Vert_2^2, \sigma_p^2 d\} \cdot \frac{N \Vert \bm{\mu} \Vert_2 \big( \log (6N^2M^2 / \delta) \big)^{3} \log \big( O(\frac{1}{\epsilon}) \big) \sigma_h }{ \sigma_p d^{\frac{1}{2}} } \Big)
\end{split}
\end{equation}
Taking a summation we have
\begin{equation}
\begin{split}
    \label{delta_K_3}
    &\Vert \bm{W}_K^{(T_3)} - \bm{W}_K^{(T_2)} \Vert_F \le \sum\limits_{s=T_2}^{T_3 - 1} \Vert \bm{W}_K^{(t+1)} - \bm{W}_K^{(t)} \Vert_F \\
    &\le O \big( \frac{1}{\eta \epsilon \Vert \bm{\mu} \Vert_2^2 \Vert \bm{w}_O \Vert_2^2} \big) \cdot O \Big( \eta \cdot \max\{\Vert \bm{\mu} \Vert_2^2, \sigma_p^2 d\} \cdot \frac{N \Vert \bm{\mu} \Vert_2 \big( \log (6N^2M^2 / \delta) \big)^{3} \log \big( O(\frac{1}{\epsilon}) \big) \sigma_h d_h^{\frac{1}{2}} }{ \sigma_p d^{\frac{1}{2}} d_h^{\frac{1}{2}} } \Big) \\
    &= O \Big( \max\{\Vert \bm{\mu} \Vert_2^2, \sigma_p^2 d\} \cdot \frac{N \big( \log (6N^2M^2 / \delta) \big)^{3} \log \big( O(\frac{1}{\epsilon}) \big) \sigma_h }{ \epsilon \Vert \bm{\mu} \Vert_2 \sigma_p d^{\frac{1}{2}} } \Big).
\end{split}
\end{equation}
\\
Combining \eqref{delta_K}, \eqref{delta_K_1}, \eqref{delta_K_2} and \eqref{delta_K_3} and get
\begin{equation}
\begin{split}
    \label{K_dy_F_norm}
    &\Vert \bm{W}_K^{(T_3)} - \bm{W}_K^{(0)} \Vert_F \le \Vert \bm{W}_K^{(T_1)} - \bm{W}_K^{(0)} \Vert_F + \Vert \bm{W}_K^{(T_2)} - \bm{W}_K^{(T_1)} \Vert_F + \Vert \bm{W}_K^{(T_3)} - \bm{W}_K^{(T_2)} \Vert_F \\
    &= O \Big( N \sigma_h \Big) + O \Big( N^{\frac{1}{2}} (1 + \mathrm{SNR}^{-1}) \sigma_h d_h^{\frac{1}{4}} \log (6N^2M^2 / \delta) \Big) \\
    &+ O \Big( \max\{\Vert \bm{\mu} \Vert_2^2, \sigma_p^2 d\} \cdot \frac{N \big( \log (6N^2M^2 / \delta) \big)^{3} \log \big( O(\frac{1}{\epsilon}) \big) \sigma_h }{ \epsilon \Vert \bm{\mu} \Vert_2 \sigma_p d^{\frac{1}{2}} } \Big) \\
    &= O \Big( \sigma_h d_h^{\frac{1}{2}} \Big),
\end{split}
\end{equation}
where the last equality is by \(d_h = \widetilde{\Omega} \Big( \max \{\mathrm{SNR}^4, \mathrm{SNR}^{-4}\} N^2 \epsilon^{-2} \Big)\).
\\
\textbf{Bounding \(\bm{\mu}_+^{\top} \bm{W}_Q^{(T_3)} \bm{W}_K^{(T_3)\top} \bm{\xi}_i\):} We decompose \(\bm{W}_K^{(T_3)}\) into \(\bm{W}_K^{(0)}\) + \( \Big( \bm{W}_K^{(T_3)} - \bm{W}_K^{(0)} \Big) \), then we have
\[
    \bm{\mu}_+^{\top} \bm{W}_Q^{(T_3)} \bm{W}_K^{(T_3)\top} \bm{\xi}_i = \bm{\mu}_+^{\top} \bm{W}_Q^{(T_3)} \bm{W}_K^{(0)\top} \bm{\xi}_i + \bm{\mu}_+^{\top} \bm{W}_Q^{(T_3)} \Big( \bm{W}_K^{(T_3)\top} - \bm{W}_K^{(0)\top} \Big) \bm{\xi}_i
\]
Now we bound \(\bm{\mu}_+^{\top} \bm{W}_Q^{(T_3)} \bm{W}_K^{(0)\top} \bm{\xi}_i\) and \(\bm{\mu}_+^{\top} \bm{W}_Q^{(T_3)} \Big( \bm{W}_K^{(T_3)\top} - \bm{W}_K^{(0)\top} \Big) \bm{\xi}_i\) respectively.
\\
\textbf{Bounding \(\bm{\mu}_+^{\top} \bm{W}_Q^{(T_3)} \bm{W}_K^{(0)\top} \bm{\xi}_i\):}
Note that \(\Vert \bm{\mu}_+^{\top} \bm{W}_Q^{(T_3)} \Vert_2^2 = \Theta \big( \Vert \bm{\mu} \Vert_2^2 \sigma_h^2 d_h \big)\) and each element of  \(\bm{W}_K^{(0)\top}\) is sampled from a Gaussian distribution \(\mathcal{N}(0, \sigma_h^2)\), we have that each element of \(\bm{\mu}_+^{\top} \bm{W}_Q^{(T_3)} \bm{W}_K^{(0)\top}\) is a random variable with mean zero and standard deviation smaller than \( C_{21} \Vert \bm{\mu} \Vert_2 \sigma_h^2 d_h^{\frac{1}{2}}\). 
Therefore, for any \(i \in [M] \backslash \{1\}\) by Bernstein's inequality, with probability at least \(1 - \delta / 3N^2M^2\),
\[
    \bm{\mu}_+^{\top} \bm{W}_Q^{(T_3)} \bm{W}_K^{(0)\top} \bm{\xi}_i = O(\Vert \bm{\mu} \Vert_2 \sigma_p \sigma_h^2 d_h^{\frac{1}{2}} \sqrt{ d \log (6N^2M^2/\delta)}) = o(1),
\]
where the last equality is by \(\sigma_h^2 \le \min \{ \Vert \bm{\mu} \Vert_2^{-2}, ( \sigma_p^2 d )^{-1} \} \cdot d_h^{-\frac{1}{2}} \cdot \big( \log ( 6N^2M^2 / \delta ) \big)^{-\frac{3}{2}}\).
\\
\textbf{Bounding \(\bm{\mu}_+^{\top} \bm{W}_Q^{(T_3)} \Big( \bm{W}_K^{(T_3)\top} - \bm{W}_K^{(0)\top} \Big) \bm{\xi}_i\):} By \eqref{K_dy_F_norm}, we have
\begin{equation}
\begin{split}
    &\Big\Vert \bm{\mu}_+^{\top} \bm{W}_Q^{(T_3)} \Big( \bm{W}_K^{(T_3)\top} - \bm{W}_K^{(0)\top} \Big) \Big\Vert_2 \le \Vert \bm{\mu}_+^{\top} \bm{W}_Q^{(T_3)} \Vert_2 \Vert \bm{W}_K^{(T_3)\top} - \bm{W}_K^{(0)\top} \Vert_F \\
    &\le C_{22} \Vert \bm{\mu} \Vert_2 \sigma_h^2 d_h.
\end{split}
\end{equation}
Therefore \(\bm{\mu}_+^{\top} \bm{W}_Q^{(T_3)} \Big( \bm{W}_K^{(T_3)\top} - \bm{W}_K^{(0)\top} \Big) \bm{\xi}_i\) is a Gaussian distribution with mean zero and standard deviation smaller than \(C_{22} \Vert \bm{\mu} \Vert_2 \sigma_p \sigma_h^2 d_h\).
By Gaussian tail bound, for any \(i \in [M] \backslash \{1\}\), with probability at least \(1 - \delta / 6N^2M^2\),
\[
    \bm{\mu}_+^{\top} \bm{W}_Q^{(T_3)} \Big( \bm{W}_K^{(T_3)\top} - \bm{W}_K^{(0)\top} \Big) \bm{\xi}_i \le C_{22} \Vert \bm{\mu} \Vert_2 \sigma_p \sigma_h^2 d_h \sqrt{2 \log \big(12N^2M^2/\delta)} = o(1),
\]
where the last equality is by \(d = \widetilde{\Omega} \Big( \epsilon^{-2} N^2 d_h \Big) \) and \(\sigma_h^2 \le \min \{ \Vert \bm{\mu} \Vert_2^{-2}, ( \sigma_p^2 d )^{-1} \} \cdot d_h^{-\frac{1}{2}} \cdot \big( \log ( 6N^2M^2 / \delta ) \big)^{-\frac{3}{2}}\).
Applying a union bound, with probability at least \(1 - \delta / 2N^2M\),
\begin{align*}
    \bm{\mu}_+^{\top} \bm{W}_Q^{(T_3)} \bm{W}_K^{(T_3)\top} \bm{\xi}_i &= \bm{\mu}_+^{\top} \bm{W}_Q^{(T_3)} \bm{W}_K^{(0)\top} \bm{\xi}_i + \bm{\mu}_+^{\top} \bm{W}_Q^{(T_3)} \Big( \bm{W}_K^{(T_3)\top} - \bm{W}_K^{(0)\top} \Big) \bm{\xi}_i \\
    &= o(1) + o(1) \\
    &= o(1).
\end{align*}
Therefore, we have
\begin{equation*}
\begin{split}
    &\frac{ \exp ( \bm{\mu}_+^{\top} \bm{W}_Q^{(T_3)} \bm{W}_K^{(T_3)\top} \bm{\mu}_+ )}{ \exp ( \bm{\mu}_+^{\top} \bm{W}_Q^{(T_3)} \bm{W}_K^{(T_3)\top} \bm{\mu}_+ ) + \sum\limits_{i=2}^M  \exp ( \bm{\mu}_+^{\top} \bm{W}_Q^{(T_3)} \bm{W}_K^{(T_3)\top} \bm{\xi}_i )} \ge \frac{\exp(-o(1))}{\exp(-o(1)) + (M-1)\exp(o(1))} \\
    &\ge \frac{1}{1 + (M - 1) \exp (o(1))} \\
    &\ge \frac{1}{1 + (M - 1) + (M - 1) o(1)} \\
    &\ge \frac{1}{M} - o(1) \\
    &\ge \frac{1}{2M},
\end{split}
\end{equation*}
which complete the proof.
\begin{lemma}[Bound of V]
    \label{bound_output}
    Under the same conditions as Theorem \ref{benign}, with probability at least \(1 - \delta / 2N^2M\), we have that 
    \[
        \vert \bm{\xi}_i^{\top} \bm{W}_V^{(T_3)} \bm{w}_O \vert \le 1
    \]
\end{lemma}
\it{Proof of Lemma} \ref{bound_output}. \rm
By \eqref{nabla_V}, we have
\begin{equation}
\begin{split}
    \bm{W}_V^{(t+1)} \bm{w}_O - \bm{W}_V^{(t)} \bm{w}_O &= - \frac{\eta}{NM} \sum\limits_{n=1}^{N} y_n \ell_n^{\prime(t)} [\bm{w}_O \sum\limits_{l=1}^M  \bm{\varphi}(\bm{x}_{n, l} \bm{W}_Q^{(t)} \bm{W}_K^{(t)\top} \bm{X}_n^\top) \bm{X}_n ]^{\top} \bm{w}_O \\
    &= - \frac{\eta}{NM} \sum\limits_{n=1}^N y_n \ell_n^{\prime(t)} \sum\limits_{l=1}^M \bm{X}_n^\top \bm{\varphi}(\bm{x}_{n, l} \bm{W}_Q^{(t)} \bm{W}_K^{(t)\top} \bm{X}_n^\top)^\top \Vert \bm{w}_O \Vert_2^2.
\end{split}
\end{equation}
Then we have
\begin{equation}
\begin{split}
    \label{VO_dy}
    \Vert \bm{W}_V^{(t+1)} \bm{w}_O - \bm{W}_V^{(t)} \bm{w}_O \Vert_2 &\le \frac{\eta}{NM} \sum\limits_{n=1}^N \big\vert y_n \ell_n^{\prime(t)} \big\vert \sum\limits_{l=1}^M \big\Vert \bm{X}_n \big\Vert_F \big\Vert \bm{\varphi}(\bm{x}_{n, l} \bm{W}_Q^{(t)} \bm{W}_K^{(t)\top} \bm{X}_n^\top) \big\Vert_2 \Vert \bm{w}_O \Vert_2^2 \\
    &\le \frac{\eta}{NM} \cdot NM \cdot O \big( \max\{\Vert \bm{\mu} \Vert_2, \sigma_p \sqrt{d}\} \big) \cdot O(1) \cdot \Vert \bm{w}_O \Vert_2^2 \\
    &= O \Big( \eta \cdot \max\{\Vert \bm{\mu} \Vert_2, \sigma_p \sqrt{d}\} \cdot \Vert \bm{w}_O \Vert_2^2 \Big),
\end{split}
\end{equation}
where the second inequality is by \(\big\vert y_n \ell_n^{\prime(t)} \big\vert \le 1\), \(\Vert \bm{X}_n \Vert_F = O \big( \max\{\Vert \bm{\mu} \Vert_2, \sigma_p \sqrt{d}\} \big)\) and \(\big\Vert \bm{\varphi}(\bm{x}_{n, l} \bm{W}_Q^{(t)} \bm{W}_K^{(t)\top} (\bm{X}_n)^\top) \big\Vert_2 = O(1)\).
Taking a summation and get
\begin{equation}
\begin{split}
    \label{VO3}
    \Vert \bm{W}_V^{(T_3)} \bm{w}_O \Vert_2 &\le \Vert \bm{W}_V^{(0)} \bm{w}_O \Vert_2 + \sum\limits_{t = 0}^{T_3 - 1} \Vert \bm{W}_V^{(t+1)} \bm{w}_O - \bm{W}_V^{(t)} \bm{w}_O \Vert_2 \\
    &= \Vert \bm{W}_V^{(0)} \bm{w}_O \Vert_2 + O \big( \frac{1}{\eta \epsilon \Vert \bm{\mu} \Vert_2^2 \Vert \bm{w}_O \Vert_2^2} \big) \cdot O \Big( \eta \cdot \max\{\Vert \bm{\mu} \Vert_2, \sigma_p \sqrt{d}\} \cdot \Vert \bm{w}_O \Vert_2^2 \Big) \\
    &= O \Big( \sigma_V \Vert \bm{w}_O \Vert_2 \sqrt{d} \Big) + O \Big( \frac{\max\{\Vert \bm{\mu} \Vert_2, \sigma_p \sqrt{d}\}}{\epsilon \Vert \bm{\mu} \Vert_2^2} \Big) \\
    &= O \Big( \sigma_V \Vert \bm{w}_O \Vert_2 \sqrt{d} + \frac{\max\{\Vert \bm{\mu} \Vert_2, \sigma_p \sqrt{d}\}}{\epsilon \Vert \bm{\mu} \Vert_2^2} \Big).
\end{split}
\end{equation}
Then \(\bm{\xi}_i^{\top} \bm{W}_V^{(T_3)} \bm{w}_O\) is a Gaussian random variable with mean zero and standard deviation smaller than \(O \Big( \sigma_V \Vert \bm{w}_O \Vert_2 \sigma_p \sqrt{d} + \frac{\max\{\Vert \bm{\mu} \Vert_2, \sigma_p \sqrt{d}\} \cdot \sigma_p}{\epsilon \Vert \bm{\mu} \Vert_2^2} \Big)\).
By Gaussian tail bound, for any \(i \in [M] \backslash \{1\}\), with probability at least \(1 - \delta / 2N^2M^2\),
\begin{equation*}
\begin{split}
    \vert \bm{\xi}_i^{\top} \bm{W}_V^{(T_3)} \bm{w}_O \vert &\le O \Big( \sigma_V \Vert \bm{w}_O \Vert_2 \sigma_p \sqrt{d} + \frac{\max\{\Vert \bm{\mu} \Vert_2, \sigma_p \sqrt{d}\} \cdot \sigma_p}{\epsilon \Vert \bm{\mu} \Vert_2^2} \Big) \cdot \sqrt{2 \log \big(4N^2M^2/\delta)} \\
    &\le 1,
\end{split}
\end{equation*}
where the last inequality is by \(\sigma_V \le \widetilde{O} \big( \Vert \bm{w}_O \Vert_2^{-1} \cdot \min \{ \Vert \bm{\mu} \Vert_2^{-1}, ( \sigma_p \sqrt{d} )^{-1} \} \cdot d_h^{-\frac{1}{4}} \big)\), \(N \cdot \mathrm{SNR}^2 = \Omega(1)\) and \(d = \widetilde{\Omega} \Big( \epsilon^{-2} N^2 d_h \Big) \) and \(d_h = \widetilde{\Omega} \Big( \max \{\mathrm{SNR}^4, \mathrm{SNR}^{-4}\} N^2 \epsilon^{-2} \Big) \). Applying a union bound completes the proof.

\begin{lemma}[Population Loss]
    \label{population_loss}
    Under the same conditions as Theorem \ref{benign}, we have
    \[
        L_D \big( \theta(T_3) \big) = o(1) 
    \]
\end{lemma}
\it{Proof of Lemma} \ref{population_loss}. \rm Let event A to be the event that Lemma \ref{attn} holds, and event B to be the event that Lemma \ref{bound_output} holds. By a union bound, \(\mathbb{P}(A \wedge B) \ge 1 - \delta / N^2M \).
We decompose \(L_D \big( \theta(T_3) \big)\) into the following form:
\begin{equation}
    \label{loss}
    L_D \big( \theta(T_3) \big) = \mathbb{E} [ \mathbbm{1}(A \wedge B) \ell (y f(\bm{X}, \theta(T_3))) ] + \mathbb{E} [ \mathbbm{1}((A \wedge B)^c) \ell (y f(\bm{X}, \theta(T_3))) ]
\end{equation}
Then we bound \(\mathbb{E} [ \mathbbm{1}(A \wedge B) \ell (y f(\bm{X}, \theta(T_3))) ]\) and \(\mathbb{E} [ \mathbbm{1}((A \wedge B)^c) \ell (y f(\bm{X}, \theta(T_3))) ]\) respectively.
\\
\textbf{Bounding \(\mathbb{E} [ \mathbbm{1}(A \wedge B) \ell (y f(\bm{X}, \theta(T_3))) ]\):} When event A and event B hold, we have
\begin{equation}
\begin{split}
    y (f(\bm{X}, \theta (T_3))) &= \frac{1}{M} \sum\limits_{l=1}^M \bm{\varphi} (\bm{x}_l^{\top} \bm{W}_{Q}^{(T_3)} \bm{W}_{K}^{(T_3)\top} \bm{X}^\top ) \bm{X} \bm{W}_V^{(T_3)} \bm{w}_O \\
    &\ge \frac{1}{2M^2} \big\vert \bm{\mu}_+^{\top} \bm{W}_V^{(T_3)} \bm{w}_O \big\vert - \big( 1 - \frac{1}{2M^2} \big) \max \{ \big\vert \bm{\xi}_i^{\top} \bm{W}_V^{(T_3)} \bm{w}_O \big\vert \} \\
    &\ge \frac{1}{2M^2} \log \Big( \frac{C_{20}}{\epsilon} \Big) - 1 \\
    &\ge \frac{\log \Big( \frac{C_{20}}{\epsilon} \Big)}{C_{21}} - 1,
\end{split}
\end{equation}
where the first inequality is by Lemma \ref{attn}, the second inequality is by \eqref{V_p_bound3} and Lemma \ref{bound_output}, the last inequality is by \(M = \Theta(1)\). 
Therefore,

\begin{equation}
\begin{split}
    \label{term1}
    \mathbb{E} [ \mathbbm{1}(A \wedge B) \ell (y f(\bm{X}, \theta(T_3))) ] &\le \ell ( y f(\bm{X}, \theta(T_3)) ) \\
    &\le \log (1 + \exp(-y f(\bm{X}, \theta(T_3)))) \\
    &\le \exp(-y f(\bm{X}, \theta(T_3))) \\
    &\le \exp \Big( 1 - \frac{\log \Big( \frac{C_{20}}{\epsilon} \Big)}{C_{21}} \Big) \\
    &\le 3 \exp \Big(\frac{\log \Big( \frac{\epsilon}{C_{20}} \Big)}{C_{21}} \Big)
\end{split}
\end{equation}

\textbf{Bounding \(\mathbb{E} [ \mathbbm{1}((A \wedge B)^c) \ell (y f(\bm{X}, \theta(T_3))) ]\):}
\begin{equation}
\begin{split}
    \mathbb{E} [ \mathbbm{1}((A \wedge B)^c) \ell (y f(\bm{X}, \theta(T_3))) ] &\le \sqrt{\mathbb{E}[\mathbbm{1}((A \wedge B)^c)]} \cdot \sqrt{\mathbb{E}[ \ell (y f(\bm{X}, \theta(T_3)))^2 ]} \\
    &\le \sqrt{\mathbb{P}((A \wedge B)^c)} \cdot \sqrt{\mathbb{E}[ \exp(-2y f(\bm{X}, \theta(T_3))) ]} \\
    &\le \frac{\delta^{\frac{1}{2}}}{N} \cdot \sqrt{\mathbb{E}[ \exp(-2 f(\bm{X}, \theta(T_3))) ]},
\end{split}
\end{equation}
where the first inequality is by Cauchy-Schwartz inequality, the second inequality is by \(\log(1 + \exp(-z)) \le \exp(-z)\).

Next we provide a bound for \(\mathbb{E}[ \exp(-2 f(\bm{X}, \theta(T_3))) ]\). Note that 
\[
    y (f(\bm{X}, \theta (T_3))) = \frac{1}{M} \sum\limits_{l=1}^M \bm{\varphi} (\bm{x}_l^{\top} \bm{W}_{Q}^{(T_3)} \bm{W}_{K}^{(T_3)\top} \bm{X}^\top ) \bm{X} \bm{W}_V^{(T_3)} \bm{w}_O  
\]
and \(\bm{\mu}_+ \bm{W}_V^{T_3} \bm{w}_O \ge 0\), we further have
\[
    y (f(\bm{X}, \theta (T_3))) \ge -\sum\limits_{i = 2}^{M} \vert \bm{\xi}_i \bm{W}_V^{(T_3)} \bm{w}_O \vert
\]
Therefore,
\begin{equation}
\begin{split}
    \mathbb{E}[ \exp(-2 f(\bm{X}, \theta(T_3))) ] &\le \mathbb{E}[ \exp(2 \sum\limits_{i = 2}^{M} \vert \bm{\xi}_i \bm{W}_V^{(T_3)} \bm{w}_O \vert) ] \\
    &\le \mathbb{E}[ \prod\limits_{i = 2}^{M} \exp(2 \cdot \vert \bm{\xi}_i \bm{W}_V^{(T_3)} \bm{w}_O \vert) ] \\
    &\le \mathbb{E}[ \prod\limits_{i = 2}^{M} \Big( \exp(2 \bm{\xi}_i \bm{W}_V^{(T_3)} \bm{w}_O ) + \exp(-2 \bm{\xi}_i \bm{W}_V^{(T_3)} \bm{w}_O ) \Big) ] \\
    &\le \prod\limits_{i = 2}^{M} \mathbb{E}[ \Big( \exp(2 \bm{\xi}_i \bm{W}_V^{(T_3)} \bm{w}_O ) + \exp(-2 \bm{\xi}_i \bm{W}_V^{(T_3)} \bm{w}_O ) \Big) ] \\
    &\le 2^M \prod\limits_{i = 2}^{M} \mathbb{E}[ \exp(2 \bm{\xi}_i \bm{W}_V^{(T_3)} \bm{w}_O ) ]
\end{split}
\end{equation}
where the third inequality is by \(\vert x \vert \le \max\{x, -x\}\), The fourth inequality is by the independence of \(\bm{\xi}_i \bm{W}_V^{(T_3)} \bm{w}_O\). 

We denote \(\sigma = \Vert \bm{W}_V^{(T_3)} \bm{w}_O \Vert_2\), then \( \bm{\xi}_i \bm{W}_V^{(T_3)} \bm{w}_O \) is a Gaussian distribution with mean zero and variance smaller than \( C_p^2 \sigma^2 \sigma_p^2\). 
Therefore, 
\begin{equation}
\begin{split}
    \mathbb{E}[ \exp(2 \bm{\xi}_i \bm{W}_V^{(T_3)} \bm{w}_O ) ] &\le \exp(2C_p^2 \sigma^2 \sigma_p^2) \\
    &= \exp \Big( 2C_p^2 \sigma_p^2 \cdot O \Big( \sigma_V^2 \Vert \bm{w}_O \Vert_2^2 d + \frac{\max\{\Vert \bm{\mu} \Vert_2^2, \sigma_p^2 d\}}{\epsilon^2 \Vert \bm{\mu} \Vert_2^4} \Big) \Big) \\
    &= \exp \Big( O \Big( \sigma_V^2 \Vert \bm{w}_O \Vert_2^2 \sigma_p^2 d + \frac{\max\{\Vert \bm{\mu} \Vert_2^2, \sigma_p^2 d\} \cdot \sigma_p^2}{\epsilon^2 \Vert \bm{\mu} \Vert_2^4} \Big) \Big) \\
    &= \exp \Big( o(1) \Big) \\
    &= 1 + o(1),
\end{split}
\end{equation}
where the first equality is by \eqref{VO3}, the third equality is by \(\sigma_V \le \widetilde{O} \big( \Vert \bm{w}_O \Vert_2^{-1} \cdot \min \{ \Vert \bm{\mu} \Vert_2^{-1}, ( \sigma_p \sqrt{d} )^{-1} \} \cdot d_h^{-\frac{1}{4}} \big)\), \(N \cdot \mathrm{SNR}^2 \ge \Omega(1)\) and \(d = \widetilde{\Omega} \Big( \epsilon^{-2} N^2 d_h \Big) \). Then we have
\begin{equation}
\begin{split}
    \mathbb{E}[ \exp(-2 f(\bm{X}, \theta(T_3))) ] &\le 2^M \prod\limits_{i = 2}^{M} \mathbb{E}[ \exp(2 \bm{\xi}_i \bm{W}_V^{(T_3)} \bm{w}_O ) ] \\
    &\le 2^M \cdot (1 + o(1))^M \\
    &= O(1).
\end{split}
\end{equation}
We further get
\begin{equation}
\begin{split}
    \label{term2}
    \mathbb{E} [ \mathbbm{1}((A \wedge B)^c) \ell (y f(\bm{X}, \theta(T_3))) ] &\le \frac{\delta^{\frac{1}{2}}}{N} \cdot \sqrt{\mathbb{E}[ \exp(-2 f(\bm{X}, \theta(T_3))) ]} \le \frac{C_{24}\delta^{\frac{1}{2}}}{N}
\end{split}
\end{equation}

Plugging \eqref{term1} and \eqref{term2} into \eqref{loss}, we have
\[
    L_D \big( \theta(T_3) \big) \le 3 \exp \Big(\frac{\log \Big( \frac{\epsilon}{C_{20}} \Big)}{C_{21}} \Big) + \frac{C_{24}\delta^{\frac{1}{2}}}{N} = o(1).
\]

\section{Harmful Overfitting}
In this section, we consider harmful overfitting case under the condition that \(N^{-1} \cdot \mathrm{SNR}^{-2} \ge \Omega(1)\). The proofs in this section are based on the results in Section \ref{concentration_inequalities}, which hold with high probability.

\subsection{Stage \uppercase\expandafter{\romannumeral1}}
In Stage \uppercase\expandafter{\romannumeral1}, \( V_\pm^{(t)} \), \( V_{n, i}^{(t)} \) begin to pull apart until \( \vert V_\pm^{(t)} \vert \) is sufficiently larger than \( \vert V_{n, i}^{(t)} \vert \). At the same time, the inner products of \(\bm{q}\) and \(\bm{k}\) maintain their magnitude.

\begin{lemma}[Upper bound of V]
    \label{upper_bound_of_V_harmful}
    Let \(T_0 = O(\frac{N}{\eta d_h^{\frac{1}{4}} \sigma_p^2 d \Vert \bm{w}_O \Vert_2^2})\). Then under the same conditions as Theorem \ref{harmful}, we have
    \[
        \vert V_+^{(t)} \vert, \vert V_-^{(t)} \vert, \vert V_{n, i}^{(t)} \vert = O (d_h^{-\frac{1}{4}})
    \]
    for \( t \in [0, T_0] \).
\end{lemma}
\it Proof of Lemma \ref{upper_bound_of_V_harmful}. \rm By Lemma \ref{update_rule4V}, we have
\begin{equation*}
\begin{split}
    &\vert \gamma_{V, +}^{(t + 1)} - \gamma_{V, +}^{(t)} \vert \\
    &\le - \frac{\eta \Vert \bm{\mu} \Vert_2^2}{NM} \sum\limits_{n \in S_+} \ell_n^{\prime(t)} \big( \frac{ \exp ( \langle \bm{q}_+^{(t)}, \bm{k}_+^{(t)} \rangle )}{  \exp ( \langle \bm{q}_+^{(t)}, \bm{k}_+^{(t)} \rangle ) + \sum\limits_{k=2}^M  \exp ( \langle \bm{q}_+^{(t)}, \bm{k}_{n, k}^{(t)} \rangle )}  \\
    &+ \sum\limits_{j=2}^M \frac{ \exp ( \langle \bm{q}_{n, j}^{(t)}, \bm{k}_+^{(t)} \rangle )}{  \exp ( \langle \bm{q}_{n, j}^{(t)}, \bm{k}_+^{(t)} \rangle ) + \sum\limits_{k=2}^M  \exp ( \langle \bm{q}_{n, j}^{(t)}, \bm{k}_{n, k}^{(t)} \rangle )} \big) \\
    &\le \frac{\eta \Vert \bm{\mu} \Vert_2^2}{NM} \cdot \frac{3N}{4} \big( 1 + (M - 1) \big) \\
    &\le \frac{3\eta \Vert \bm{\mu} \Vert_2^2}{4},
\end{split}
\end{equation*}
where the second inequality is by Lemma \ref{dataset_size} and \( - \ell_n^{\prime(t)} \le 1 \). Similarly, we have
\begin{equation*}
\begin{split}
    \vert \gamma_{V, -}^{(t + 1)} - \gamma_{V, -}^{(t)} \vert \le \frac{3\eta \Vert \bm{\mu} \Vert_2^2}{4}.
\end{split}
\end{equation*}
By Definition \ref{def1}, we have
\begin{equation*}
\begin{split}
    \vert V_+^{(t)} \vert &= \big\vert V_+^{(0)}  + \sum\limits_{s = 0}^{t - 1} ( \gamma_{V, +}^{(s + 1)} - \gamma_{V, +}^{(s)} ) \Vert \bm{w}_O \Vert_2^2 \big\vert \\
    &\le \vert V_+^{(0)} \vert + \sum\limits_{s = 0}^{T_0 - 1} \vert \gamma_{V, +}^{(s + 1)} - \gamma_{V, +}^{(s)} \vert \cdot \Vert \bm{w}_O \Vert_2^2 \\
    &\le d_h^{-\frac{1}{4}} + \frac{3\eta \Vert \bm{\mu} \Vert_2^2}{4} \cdot \Vert \bm{w}_O \Vert_2^2 \cdot O(\frac{N}{\eta d_h^{\frac{1}{4}} \sigma_p^2 d \Vert \bm{w}_O \Vert_2^2}) \\
    &= d_h^{-\frac{1}{4}} + O(\frac{N \Vert \bm{\mu} \Vert_2^2}{ d_h^{\frac{1}{4}} \sigma_p^2 d}) \\
    &= O(d_h^{-\frac{1}{4}}),
\end{split}
\end{equation*}
where the first inequality is by triangle inequality and \( t \le T_0 \), the second inequality is by Lemma \ref{initialization_of_V}, the last equality is by \(N^{-1} \cdot \mathrm{SNR}^{-2} = \Omega(1)\). Similarly, we have \( \vert V_-^{(t)} \vert = O(d_h^{-\frac{1}{4}}) \).

By Lemma \ref{update_rule4V}, we have
\begin{equation}
\begin{split}
    \label{rho_bound2}
    &\vert \rho_{V, n, i}^{(t+1)} - \rho_{V, n, i}^{(t)} \vert \\
    &\le \Big\vert - \frac{\eta}{NM} \sum\limits_{n^\prime \in S_+} \ell_{n^\prime}^{\prime(t)} \big( 
    \sum\limits_{i=2}^M ( \langle \bm{\xi}_{n, i}, \bm{\xi}_{n^\prime, i^\prime} \rangle \frac{ \exp ( \langle \bm{q}_+^{(t)}, \bm{k}_{n^\prime, i^\prime}^{(t)} \rangle )}{  \exp ( \langle \bm{q}_+^{(t)}, \bm{k}_+^{(t)} \rangle ) + \sum\limits_{k=2}^M  \exp ( \langle \bm{q}_+^{(t)}, \bm{k}_{n^\prime, k}^{(t)} \rangle )} \\
    &+ \sum\limits_{j=2}^M \langle \bm{\xi}_{n, i}, \bm{\xi}_{n^\prime, i^\prime} \rangle \frac{ \exp ( \langle \bm{q}_{n^\prime, j}^{(t)}, \bm{k}_{n^\prime, i^\prime}^{(t)} \rangle  )}{ \exp ( \langle \bm{q}_{n^\prime, j}^{(t)}, \bm{k}_+^{(t)} \rangle ) + \sum\limits_{k=2}^M  \exp ( \langle \bm{q}_{n^\prime, j}^{(t)}, \bm{k}_{n^\prime, k}^{(t)} \rangle )} ) \big) \\
    &+ \frac{\eta}{NM} \sum\limits_{n^\prime \in S_-} \ell_{n^\prime}^{\prime(t)} \big(
    \sum\limits_{i=2}^M ( \langle \bm{\xi}_{n, i}, \bm{\xi}_{n^\prime, i^\prime} \rangle \frac{ \exp ( \langle \bm{q}_-^{(t)}, \bm{k}_{n^\prime, i^\prime}^{(t)} \rangle )}{  \exp ( \langle \bm{q}_-^{(t)}, \bm{k}_-^{(t)} \rangle ) + \sum\limits_{k=2}^M  \exp ( \langle \bm{q}_-^{(t)}, \bm{k}_{n^\prime, k}^{(t)} \rangle )} \\
    &+ \sum\limits_{j=2}^M \langle \bm{\xi}_{n, i}, \bm{\xi}_{n^\prime, i^\prime} \rangle \frac{ \exp ( \langle \bm{q}_{n^\prime, j}^{(t)}, \bm{k}_{n^\prime, i^\prime}^{(t)} \rangle  )}{ \exp ( \langle \bm{q}_{n^\prime, j}^{(t)}, \bm{k}_-^{(t)} \rangle ) + \sum\limits_{k=2}^M  \exp ( \langle \bm{q}_{n^\prime, j}^{(t)}, \bm{k}_{n^\prime, k}^{(t)} \rangle )} ) \big) \Big\vert \\
    &\le \frac{3 \eta \tilde{\sigma}_p^2 d}{2NM} \cdot M + \frac{\eta}{NM} \cdot MN \cdot 2 \tilde{\sigma}_p^2 \cdot \sqrt{d \log (4N^2M^2/\delta)} \\
    &\le \frac{2 \eta \tilde{\sigma}_p^2 d}{N} \\
    &= \frac{2 \eta C_p^2 \sigma_p^2 d}{N}
\end{split}
\end{equation}
where the second inequality is by Lemma \ref{caoyuan} and \( - \ell_n^{\prime(t)} \le 1 \), the third inequality is by \(d = \widetilde{\Omega} \Big( \epsilon^{-2} N^2 d_h \Big) \), the last inequality is by \( N \cdot \mathrm{SNR}^2 = \Omega(1) \). Then by Definition \ref{def1}, we have
\begin{equation*}
\begin{split}
    \vert V_{n, i}^{(t)} \vert &= \big\vert V_{n, i}^{(0)}  + \sum\limits_{s = 0}^{t - 1} ( \gamma_{V, n, i}^{(s + 1)} - \gamma_{V, n, i}^{(s)} ) \Vert \bm{w}_O \Vert_2^2 \big\vert \\
    &\le \vert V_{n, i}^{(0)} \vert + \sum\limits_{s = 0}^{t - 1} \vert \gamma_{V, n, i}^{(s + 1)} - \gamma_{V, n, i}^{(s)} \vert \cdot \Vert \bm{w}_O \Vert_2^2 \\
    &\le d_h^{-\frac{1}{4}} + \frac{2 \eta C_p^2 \sigma_p^2 d}{N} \cdot \Vert \bm{w}_O \Vert_2^2 \cdot O(\frac{N}{\eta d_h^{\frac{1}{4}} \sigma_p^2 d \Vert \bm{w}_O \Vert_2^2}) \\
    &= O(d^{-\frac{1}{4}}),
\end{split}
\end{equation*}
where the first inequality is by triangle inequality, the second inequality is by Lemma \ref{initialization_of_V}, which completes the proof.

\begin{lemma}[Inner Products Hold Magnitude]
    \label{inner_products_hold_magnitude_harmful}
    Let \(T_0 = O(\frac{N}{\eta d_h^{\frac{1}{4}} \sigma_p^2 d \Vert \bm{w}_O \Vert_2^2})\). Then under the same conditions as Theorem \ref{harmful}, we have
    \begin{align*}
        &\vert \langle \bm{q}_\pm^{(t)}, \bm{k}_\pm^{(t)} \rangle \vert, \vert \langle \bm{q}_{n, i}^{(t)}, \bm{k}_\pm^{(t)} \rangle \vert, \vert \langle \bm{q}_\pm^{(t)}, \bm{k}_{n, j}^{(t)} \rangle \vert, \vert \langle \bm{q}_{n, i}^{(t)}, \bm{k}_{n^\prime, j}^{(t)} \rangle \vert \\
        &= O \Big( \max \{ \Vert \bm{\mu} \Vert_2^2, \sigma_p^2 d \} \cdot \sigma_h^2 \cdot \sqrt{d_h \log (6N^2M^2/\delta)} \Big),
    \end{align*}
    \begin{align*}
        &\vert \langle \bm{q}_\pm^{(t)}, \bm{q}_\mp^{(t)} \rangle \vert, \vert \langle \bm{q}_{n, i}^{(t)}, \bm{q}_\pm^{(t)} \rangle \vert, \vert \langle \bm{q}_{n, i}^{(t)}, \bm{q}_{n^\prime, j}^{(t)} \rangle \vert \\
        &= O \Big( \max \{ \Vert \bm{\mu} \Vert_2^2, \sigma_p^2 d \} \cdot \sigma_h^2 \cdot \sqrt{d_h \log (6N^2M^2/\delta)} \Big),
    \end{align*}
    \begin{align*}
        &\vert \langle \bm{k}_\pm^{(t)}, \bm{k}_\mp^{(t)} \rangle \vert, \vert \langle \bm{k}_{n, i}^{(t)}, \bm{k}_\pm^{(t)} \rangle \vert, \vert \langle \bm{k}_{n, i}^{(t)}, \bm{k}_{n^\prime, j}^{(t)} \rangle \vert \\
        &= O \Big( \max \{ \Vert \bm{\mu} \Vert_2^2, \sigma_p^2 d \} \cdot \sigma_h^2 \cdot \sqrt{d_h \log (6N^2M^2/\delta)} \Big),
    \end{align*}
    \[
        \Vert \bm{q}_\pm^{(t)} \Vert_2^2, \Vert \bm{k}_\pm^{(t)} \Vert_2^2 = \Theta (\Vert \bm{\mu} \Vert_2^2 \sigma_h^2 d_h),
    \]
    \[
        \Vert \bm{q}_{n, i}^{(t)} \Vert_2^2, \Vert \bm{k}_{n, i}^{(t)} \Vert_2^2 = \Theta ( \sigma_p^2 \sigma_h^2 d d_h )
    \]
    for \( i, j \in [M] \backslash \{1\} \), \( n, n^\prime \in [N] \) and \( t \in [0, T_0] \).
\end{lemma}
The proof for Lemma \ref{inner_products_hold_magnitude_harmful} is similar to \ref{inner_products_hold_magnitude}.
\begin{lemma}[V's Beginning of Learning Signals]
    \label{stage1_harmful}
    There exist \(T_1 = \frac{4M(3M + 1) N}{\eta d_h^{\frac{1}{4}} (C_p^2 - 24M) \sigma_p^2 d \Vert \bm{w}_O \Vert_2^2} \) such that 
    the second element of vector \( \bm{X}_n \bm{W}_V^{(T_1)} \bm{w}_O \) dominate its other elements, which means \( V_{n, 2}^{(T_1)} \ge 3 M \cdot \max \{ \vert V_+^{(T_1)} \vert, \vert V_{n, i}^{(T_1)} \vert \}\) for all \( n \in S_+ \), \( i \in [M] \backslash \{1, 2\} \) and
    \( V_{n, 2}^{(T_1)} \le - 3 M \cdot \max \{ \vert V_-^{(T_1)} \vert, \vert V_{n, i}^{(T_1)} \vert \} \) for all \( n \in S_- \), \( i \in [M] \backslash \{1, 2\} \).
\end{lemma}
\it Proof of Lemma \ref{stage1_harmful}. \rm
According to Condition \ref{cond} where \( C_p^2 = 25M \), we have
Then we have
\[
    T_1 = \frac{4M(3M + 1) N}{\eta d_h^{\frac{1}{4}} (25M - 24M) \sigma_p^2 d \Vert \bm{w}_O \Vert_2^2} = \frac{4(3M + 1) N}{\eta d_h^{\frac{1}{4}} \sigma_p^2 d \Vert \bm{w}_O \Vert_2^2} = O \Big( \frac{N}{\eta d_h^{\frac{1}{4}} \sigma_p^2 d \Vert \bm{w}_O \Vert_2^2} \Big),
\]
which satisfies the time condition in Lemma \ref{upper_bound_of_V_harmful} and Lemma \ref{inner_products_hold_magnitude_harmful}. Then by Lemma \ref{gradient_of_loss} and Lemma \ref{bound_of_attention} we have
\[
    - \ell_n^{\prime(t)} = \frac{1}{2} \pm o(1),  
\]
\[
    \frac{1}{M} - o(1) \le softmax(\langle \bm{q}_\pm^{(t)}, \bm{k}_\pm^{(t)} \rangle) \le \frac{1}{M} + o(1),
\]
\[
    \frac{1}{M} - o(1) \le  softmax(\langle \bm{q}_{n, i}^{(t)}, \bm{k}_\pm^{(t)} \rangle) \le \frac{1}{M} + o(1),
\]
\[
    \frac{1}{M} - o(1) \le  softmax(\langle \bm{q}_\pm^{(t)}, \bm{k}_{n, j}^{(t)} \rangle) \le \frac{1}{M} + o(1),
\]
\[
    \frac{1}{M} - o(1) \le  softmax(\langle \bm{q}_{n, i}^{(t)}, \bm{k}_{n, j}^{(t)} \rangle) \le \frac{1}{M} + o(1)
\]
for \( i, j \in [M] \backslash \{1\} \), \( n \in [N] \) and \( t \in [0, T_1] \). Plugging them in the update rule for \(\gamma_{V, +}^{(t)}\) showed in Lemma \ref{update_rule4V} and we have
\begin{equation}
\begin{split}
    &\vert \gamma_{V, +}^{(t + 1)} - \gamma_{V, +}^{(t)} \vert \\
    &= \Big\vert - \frac{\eta \Vert \bm{\mu} \Vert_2^2}{NM} \sum\limits_{n \in S_+} \ell_n^{\prime(t)} \big( \frac{ \exp ( \langle \bm{q}_+^{(t)}, \bm{k}_+^{(t)} \rangle )}{  \exp ( \langle \bm{q}_+^{(t)}, \bm{k}_+^{(t)} \rangle ) + \sum\limits_{k=2}^M  \exp ( \langle \bm{q}_+^{(t)}, \bm{k}_{n, k}^{(t)} \rangle )}  \\
    &+ \sum\limits_{j=2}^M \frac{ \exp ( \langle \bm{q}_{n, j}^{(t)}, \bm{k}_+^{(t)} \rangle )}{  \exp ( \langle \bm{q}_{n, j}^{(t)}, \bm{k}_+^{(t)} \rangle ) + \sum\limits_{k=2}^M  \exp ( \langle \bm{q}_{n, j}^{(t)}, \bm{k}_{n, k}^{(t)} \rangle )} \big) \Big\vert \\
    &\le \frac{\eta \Vert \bm{\mu} \Vert_2^2}{NM} \cdot \frac{3N}{4} \cdot (\frac{1}{2} \pm o(1)) \cdot M (\frac{1}{M} \pm o(1)) \\
    &\le \frac{\eta \Vert \bm{\mu} \Vert_2^2}{2 M} \\
    &\le \frac{2 \eta \sigma_p^2 d}{NM}
\end{split}
\end{equation}
where the last inequality is by \(N^{-1} \cdot \mathrm{SNR}^{-2} = \Omega(1)\).Then by Definition \ref{def1} and taking a summation, we have
\begin{equation}
\begin{split}
    \label{V_p_bound_harmful}
    \vert V_+^{(T_1)} \vert &\le \vert V_+^{(0)} \vert + T_1 \frac{2 \eta \sigma_p^2 d}{NM} \Vert \bm{w}_O \Vert_2^2 \\
    &= d_h^{-\frac{1}{4}} + \frac{4M(3M + 1) N}{\eta d_h^{\frac{1}{4}} (C_p^2 - 24M) \sigma_p^2 d \Vert \bm{w}_O \Vert_2^2} \cdot \frac{2 \eta \sigma_p^2 d}{NM} \Vert \bm{w}_O \Vert_2^2 \\
    &= d_h^{-\frac{1}{4}} + \frac{8(3M + 1)}{d_h^{\frac{1}{4}} (C_p^2 - 24M)}.
\end{split}
\end{equation}

Similarly, we have
\begin{equation}
\begin{split}
    \label{V_m_bound_harmful}
    \vert V_-^{(T_1)} \vert \le d_h^{-\frac{1}{4}} + \frac{8(3M + 1)}{d_h^{\frac{1}{4}} (C_p^2 - 24M)}.
\end{split}
\end{equation}

For \(n \in S_+ \), by Lemma \ref{update_rule4V}, we have
\begin{equation}
\begin{split}
    \label{rho_bound_harmful}
    &\rho_{V, n, 2}^{(t+1)} - \rho_{V, n, 2}^{(t)} \\
    &= - \frac{\eta}{NM} \sum\limits_{n^\prime \in S_+} \ell_{n^\prime}^{\prime(t)} \big( 
    \sum\limits_{i^\prime=2}^M ( \langle \bm{\xi}_{n, 2}, \bm{\xi}_{n^\prime, i^\prime} \rangle \frac{ \exp ( \langle \bm{q}_+^{(t)}, \bm{k}_{n^\prime, i^\prime}^{(t)} \rangle )}{  \exp ( \langle \bm{q}_+^{(t)}, \bm{k}_+^{(t)} \rangle ) + \sum\limits_{k=2}^M  \exp ( \langle \bm{q}_+^{(t)}, \bm{k}_{n^\prime, k}^{(t)} \rangle )} \\
    &+ \sum\limits_{j=2}^M \langle \bm{\xi}_{n, 2}, \bm{\xi}_{n^\prime, i^\prime} \rangle \frac{ \exp ( \langle \bm{q}_{n^\prime, j}^{(t)}, \bm{k}_{n^\prime, i^\prime}^{(t)} \rangle  )}{ \exp ( \langle \bm{q}_{n^\prime, j}^{(t)}, \bm{k}_+^{(t)} \rangle ) + \sum\limits_{k=2}^M  \exp ( \langle \bm{q}_{n^\prime, j}^{(t)}, \bm{k}_{n^\prime, k}^{(t)} \rangle )} ) \big) \\
    &+ \frac{\eta}{NM} \sum\limits_{n^\prime \in S_-} \ell_{n^\prime}^{\prime(t)} \big(
    \sum\limits_{i^\prime=2}^M ( \langle \bm{\xi}_{n, 2}, \bm{\xi}_{n^\prime, i^\prime} \rangle \frac{ \exp ( \langle \bm{q}_-^{(t)}, \bm{k}_{n^\prime, i^\prime}^{(t)} \rangle )}{  \exp ( \langle \bm{q}_-^{(t)}, \bm{k}_-^{(t)} \rangle ) + \sum\limits_{k=2}^M  \exp ( \langle \bm{q}_-^{(t)}, \bm{k}_{n^\prime, k}^{(t)} \rangle )} \\
    &+ \sum\limits_{j=2}^M \langle \bm{\xi}_{n, 2}, \bm{\xi}_{n^\prime, i^\prime} \rangle \frac{ \exp ( \langle \bm{q}_{n^\prime, j}^{(t)}, \bm{k}_{n^\prime, i^\prime}^{(t)} \rangle  )}{ \exp ( \langle \bm{q}_{n^\prime, j}^{(t)}, \bm{k}_-^{(t)} \rangle ) + \sum\limits_{k=2}^M  \exp ( \langle \bm{q}_{n^\prime, j}^{(t)}, \bm{k}_{n^\prime, k}^{(t)} \rangle )} ) \big) \\
    &\ge \frac{\eta \tilde{\sigma}_p^2 d}{2NM} \cdot M(\frac{1}{M} - o(1)) - \frac{\eta}{NM} \cdot MN \cdot 2 \tilde{\sigma}_p^2 \cdot \sqrt{d \log (4N^2M^2/\delta)} \\
    &\ge \frac{\eta \tilde{\sigma}_p^2 d}{4NM} \\
    &\ge \frac{\eta C_p^2 \sigma_p^2 d}{4NM}
\end{split}
\end{equation}
where the second inequality is by Lemma \ref{caoyuan} and \( - \ell_n^{\prime(t)} \le 1 \), the third inequality is by \(d = \widetilde{\Omega} \Big( \epsilon^{-2} N^2 d_h \Big) \). Then by Definition \ref{def1}, we have
\begin{equation}
\begin{split}
    \label{V_2_p_bound}
    V_{n, 2}^{(T_1)} &= V_{n, 2}^{(0)}  + \sum\limits_{s = 0}^{T_1 - 1} ( \rho_{V, n, 2}^{(s + 1)} - \rho_{V, n, 2}^{(s)} ) \Vert \bm{w}_O \Vert_2^2 \\
    &\ge -d_h^{-\frac{1}{4}} + \frac{4M(3M + 1) N}{\eta d_h^{\frac{1}{4}} (C_p^2 - 24M) \sigma_p^2 d \Vert \bm{w}_O \Vert_2^2} \cdot \frac{\eta C_p^2 \sigma_p^2 d}{4NM} \Vert \bm{w}_O \Vert_2^2 \\
    &\ge -d_h^{-\frac{1}{4}} + \frac{C_p^2(3M + 1)}{d_h^{\frac{1}{4}} (C_p^2 - 24M)}
\end{split}
\end{equation}
for \(n \in S_+\). Similarly, for \(n \in S_-\), we have
\begin{equation}
\begin{split}
    \label{V_2_m_bound}
    V_{n, 2}^{(T_1)} \le d_h^{-\frac{1}{4}} - \frac{C_p^2(3M + 1)}{d_h^{\frac{1}{4}} (C_p^2 - 24M)}.
\end{split}
\end{equation}

By Lemma \ref{update_rule4V}, we have
\begin{equation}
\begin{split}
    &\vert \rho_{V, n, i}^{(t+1)} - \rho_{V, n, i}^{(t)} \vert \\
    &\le \Big\vert - \frac{\eta}{NM} \sum\limits_{n^\prime \in S_+} \ell_{n^\prime}^{\prime(t)} \big( 
    \sum\limits_{i=2}^M ( \langle \bm{\xi}_{n, i}, \bm{\xi}_{n^\prime, i^\prime} \rangle \frac{ \exp ( \langle \bm{q}_+^{(t)}, \bm{k}_{n^\prime, i^\prime}^{(t)} \rangle )}{  \exp ( \langle \bm{q}_+^{(t)}, \bm{k}_+^{(t)} \rangle ) + \sum\limits_{k=2}^M  \exp ( \langle \bm{q}_+^{(t)}, \bm{k}_{n^\prime, k}^{(t)} \rangle )} \\
    &+ \sum\limits_{j=2}^M \langle \bm{\xi}_{n, i}, \bm{\xi}_{n^\prime, i^\prime} \rangle \frac{ \exp ( \langle \bm{q}_{n^\prime, j}^{(t)}, \bm{k}_{n^\prime, i^\prime}^{(t)} \rangle  )}{ \exp ( \langle \bm{q}_{n^\prime, j}^{(t)}, \bm{k}_+^{(t)} \rangle ) + \sum\limits_{k=2}^M  \exp ( \langle \bm{q}_{n^\prime, j}^{(t)}, \bm{k}_{n^\prime, k}^{(t)} \rangle )} ) \big) \\
    &+ \frac{\eta}{NM} \sum\limits_{n^\prime \in S_-} \ell_{n^\prime}^{\prime(t)} \big(
    \sum\limits_{i=2}^M ( \langle \bm{\xi}_{n, i}, \bm{\xi}_{n^\prime, i^\prime} \rangle \frac{ \exp ( \langle \bm{q}_-^{(t)}, \bm{k}_{n^\prime, i^\prime}^{(t)} \rangle )}{  \exp ( \langle \bm{q}_-^{(t)}, \bm{k}_-^{(t)} \rangle ) + \sum\limits_{k=2}^M  \exp ( \langle \bm{q}_-^{(t)}, \bm{k}_{n^\prime, k}^{(t)} \rangle )} \\
    &+ \sum\limits_{j=2}^M \langle \bm{\xi}_{n, i}, \bm{\xi}_{n^\prime, i^\prime} \rangle \frac{ \exp ( \langle \bm{q}_{n^\prime, j}^{(t)}, \bm{k}_{n^\prime, i^\prime}^{(t)} \rangle  )}{ \exp ( \langle \bm{q}_{n^\prime, j}^{(t)}, \bm{k}_-^{(t)} \rangle ) + \sum\limits_{k=2}^M  \exp ( \langle \bm{q}_{n^\prime, j}^{(t)}, \bm{k}_{n^\prime, k}^{(t)} \rangle )} ) \big) \Big\vert \\
    &\le \frac{3 \eta \sigma_p^2 d}{2NM} \cdot M(\frac{1}{M} + o(1)) + \frac{\eta}{NM} \cdot MN \cdot 2 \tilde{\sigma}_p^2 \cdot \sqrt{d \log (4N^2M^2/\delta)} \\
    &\le \frac{2 \eta \sigma_p^2 d}{NM}
\end{split}
\end{equation}
for \(i \in [M] \backslash \{1, 2\}\), where the second inequality is by Lemma \ref{caoyuan} and \( - \ell_n^{\prime(t)} \le 1 \), the last inequality is by \(d = \widetilde{\Omega} \Big( \epsilon^{-2} N^2 d_h \Big) \). Then by Definition \ref{def1}, we have
\begin{equation}
\begin{split}
    \label{V_i_bound_harmful}
    \vert V_{n, i}^{(t)} \vert &= \big\vert V_{n, i}^{(0)}  + \sum\limits_{s = 0}^{t - 1} ( \gamma_{V, n, i}^{(s + 1)} - \gamma_{V, n, i}^{(s)} ) \Vert \bm{w}_O \Vert_2^2 \big\vert \\
    &\le \vert V_{n, i}^{(0)} \vert + \sum\limits_{s = 0}^{t - 1} \vert \gamma_{V, n, i}^{(s + 1)} - \gamma_{V, n, i}^{(s)} \vert \cdot \Vert \bm{w}_O \Vert_2^2 \\
    &\le d_h^{-\frac{1}{4}} + \frac{4M(3M + 1) N}{\eta d_h^{\frac{1}{4}} (C_p^2 - 24M) \sigma_p^2 d \Vert \bm{w}_O \Vert_2^2} \cdot \frac{2 \eta \sigma_p^2 d}{NM} \cdot \Vert \bm{w}_O \Vert_2^2 \\
    &\le d_h^{-\frac{1}{4}} + \frac{8(3M + 1)}{d_h^{\frac{1}{4}} (C_p^2 - 24M)}
\end{split}
\end{equation}
for \(i \in [M] \backslash \{1, 2\}\), where the first inequality is by triangle inequality, the second inequality is by Lemma \ref{initialization_of_V}.

According to \eqref{V_p_bound_harmful}, \eqref{V_m_bound_harmful}, \eqref{V_2_p_bound}, \eqref{V_2_m_bound} and \eqref{V_i_bound_harmful}, it is easy to verify that \( V_{n, 2}^{(T_1)} \ge 3 M \cdot \max \{ \vert V_+^{(T_1)} \vert, \vert V_{n, i}^{(T_1)} \vert \}\) for all \( n \in S_+ \), \( i \in [M] \backslash \{1, 2\} \) and
\( V_{n, 2}^{(T_1)} \le - 3 M \cdot \max \{ \vert V_-^{(T_1)} \vert, \vert V_{n, i}^{(T_1)} \vert \} \) for all \( n \in S_- \), \( i \in [M] \backslash \{1, 2\} \), which completes the proof.

\subsection{Stage \uppercase\expandafter{\romannumeral2}}
\label{stage2_harmful}

In stage \uppercase\expandafter{\romannumeral2}, \(\langle \bm{q}_+, \bm{k}_+ \rangle\), \(\langle \bm{q}_{n, i}, \bm{k}_+ \rangle\) grows while \(\langle \bm{q}_+, \bm{k}_{n, j} \rangle\), \(\langle \bm{q}_{n, i}, \bm{k}_{n, j} \rangle\) decreases, resulting in attention focusing more and more on the signal and less on the noise. 
By the results of stage \uppercase\expandafter{\romannumeral1}, we have the following condition at the beginning of stage \uppercase\expandafter{\romannumeral2}

\[ V_{n, 2}^{(T_1)} \ge 3 M \cdot \max \{ \vert V_+^{(T_1)} \vert, \vert V_{n, i}^{(T_1)} \vert \}\]
for all \( n \in S_+ \), \( i \in [M] \backslash \{1, 2\} \).
\[ V_{n, 2}^{(T_1)} \le - 3 M \cdot \max \{ \vert V_-^{(T_1)} \vert, \vert V_{n, i}^{(T_1)} \vert \} \]
for all \( n \in S_- \), \( i \in [M] \backslash \{1, 2\} \).
\[
    \vert V_+^{(T_1)} \vert, \vert V_-^{(T_1)} \vert, \vert V_{n, i}^{(T_1)} \vert = O (d_h^{-\frac{1}{4}}),
\]
\begin{align*}
    &\vert \langle \bm{q}_\pm^{(T_1)}, \bm{k}_\pm^{(T_1)} \rangle \vert, \vert \langle \bm{q}_{n, i}^{(T_1)}, \bm{k}_\pm^{(T_1)} \rangle \vert, \vert \langle \bm{q}_\pm^{(T_1)}, \bm{k}_{n, j}^{(T_1)} \rangle \vert, \vert \langle \bm{q}_{n, i}^{(T_1)}, \bm{k}_{n^\prime, j}^{(T_1)} \rangle \vert \\
    &= O \Big( \max \{ \Vert \bm{\mu} \Vert_2^2, \sigma_p^2 d \} \cdot \sigma_h^2 \cdot \sqrt{d_h \log (6N^2M^2/\delta)} \Big),
\end{align*}
\begin{align*}
    &\vert \langle \bm{q}_\pm^{(T_1)}, \bm{q}_\mp^{(T_1)} \rangle \vert, \vert \langle \bm{q}_{n, i}^{(T_1)}, \bm{q}_\pm^{(T_1)} \rangle \vert, \vert \langle \bm{q}_{n, i}^{(T_1)}, \bm{q}_{n^\prime, j}^{(T_1)} \rangle \vert \\
    &= O \Big( \max \{ \Vert \bm{\mu} \Vert_2^2, \sigma_p^2 d \} \cdot \sigma_h^2 \cdot \sqrt{d_h \log (6N^2M^2/\delta)} \Big),
\end{align*}
\begin{align*}
    &\vert \langle \bm{k}_\pm^{(T_1)}, \bm{k}_\mp^{(T_1)} \rangle \vert, \vert \langle \bm{k}_{n, i}^{(T_1)}, \bm{k}_\pm^{(T_1)} \rangle \vert, \vert \langle \bm{k}_{n, i}^{(T_1)}, \bm{k}_{n^\prime, j}^{(T_1)} \rangle \vert \\
    &= O \Big( \max \{ \Vert \bm{\mu} \Vert_2^2, \sigma_p^2 d \} \cdot \sigma_h^2 \cdot \sqrt{d_h \log (6N^2M^2/\delta)} \Big),
\end{align*}
\[
    \Vert \bm{q}_\pm^{(T_1)} \Vert_2^2, \Vert \bm{k}_\pm^{(T_1)} \Vert_2^2 = \Theta (\Vert \bm{\mu} \Vert_2^2 \sigma_h^2 d_h),
\]
\[
    \Vert \bm{q}_{n, i}^{(T_1)} \Vert_2^2, \Vert \bm{k}_{n, i}^{(T_1)} \Vert_2^2 = \Theta ( \sigma_p^2 \sigma_h^2 d d_h )
\]
for \( i, j \in [M] \backslash \{1\} \), \( n, n^\prime \in [N] \).

Some of the proofs at this stage are based on the above conditions.

\textbf{Notations.} To better characterize the gap between different inner products, we define the following notations:
\begin{itemize}
    \item denote \( \Psi_{+}^{(t)} = \langle \bm{q}_+^{(t)}, \bm{k}_{n, 2}^{(t)} \rangle - \langle \bm{q}_+^{(t)}, \bm{k}_+^{(t)} \rangle, \quad n \in S_+ \).
    \item denote \( \Psi_{n, +, j}^{(t)} = \langle \bm{q}_+^{(t)}, \bm{k}_{n, 2}^{(t)} \rangle - \langle \bm{q}_+^{(t)}, \bm{k}_{n, j}^{(t)} \rangle, \quad n \in S_+, j \in [M] \backslash \{1, 2\} \).
    \item denote \( \Psi_{-}^{(t)} = \langle \bm{q}_-^{(t)}, \bm{k}_{n, 2}^{(t)} \rangle - \langle \bm{q}_-^{(t)}, \bm{k}_-^{(t)} \rangle, \quad n \in S_- \).
    \item denote \( \Psi_{n, -, j}^{(t)} = \langle \bm{q}_-^{(t)}, \bm{k}_{n, 2}^{(t)} \rangle - \langle \bm{q}_-^{(t)}, \bm{k}_{n, j}^{(t)} \rangle, \quad n \in S_-, j \in [M] \backslash \{1, 2\} \).
    \item denote \( \Psi_{n, i, +}^{(t)} = \langle \bm{q}_{n, i}^{(t)}, \bm{k}_{n, 2}^{(t)} \rangle - \langle \bm{q}_{n, i}^{(t)}, \bm{k}_+^{(t)} \rangle, \quad n \in S_+, i \in [M] \backslash \{1\} \).
    \item denote \( \Psi_{n, i, -}^{(t)} = \langle \bm{q}_{n, i}^{(t)}, \bm{k}_{n, 2}^{(t)} \rangle - \langle \bm{q}_{n, i}^{(t)}, \bm{k}_-^{(t)} \rangle, \quad n \in S_-, i \in [M] \backslash \{1\} \).
    \item denote \( \Psi_{n, i, j}^{(t)} = \langle \bm{q}_{n, i}^{(t)}, \bm{k}_{n, 2}^{(t)} \rangle - \langle \bm{q}_{n, i}^{(t)}, \bm{k}_{n, j}^{(t)} \rangle, i \in [M] \backslash \{1\}, j \in [M] \backslash \{1, 2\} \).
\end{itemize}

\begin{lemma}[Upper bound of V]
    \label{upper_bound_of_V2_harmful}
    Let \(T_0 = O \Big( \frac{N}{\eta \sigma_p^2 d \Vert \bm{w}_O \Vert_2^2 \log (6N^2M^2 / \delta)} \Big)\). Then under the same conditions as Theorem \ref{harmful}, we have
    \[
        \vert V_+^{(t)} \vert, \vert V_-^{(t)} \vert, \vert V_{n, i}^{(t)} \vert = o (1)
    \]
    for \( t \in [0, T_0] \).
\end{lemma}

The proof of Lemma \ref{upper_bound_of_V2_harmful} is similar to that of Lemma \ref{upper_bound_of_V_harmful}, except that the time \(T_0\) is changed.

Let \(T_2 = \Theta \Big( \frac{N}{\eta \sigma_p^2 d \Vert \bm{w}_O \Vert_2^2 \log (6N^2M^2 / \delta)} \Big)\), then by Lemma \ref{upper_bound_of_V2} and Lemma \ref{gradient_of_loss} we have \(\frac{1}{2} - o(1) \le - \ell_n^{\prime(t)} \le \frac{1}{2} + o(1) \) for \(n \in [N], t \in [T_1, T_2] \), 
which can simplify the calculations of \(\alpha\) and \(\beta\) defined in Definition \ref{def3} by replacing \( - \ell_n^{\prime(t)} \) by their bounds. 
Next we prove the following four propositions \( \mathcal{J}(t) \), \( \mathcal{K}(t) \), \( \mathcal{L}(t) \), \( \mathcal{M}(t) \) by induction on t for \(t \in [T_1, T_2]\):

\begin{itemize}
    \item \( \mathcal{J}(t): \)
    \[
        V_{n, 2}^{(t)} \ge \frac{\eta C_3 \sigma_p^2 d \Vert \bm{w}_O \Vert_2^2 (t - T_1)}{N}
    \]
    \[
        V_{n, 2}^{(t)} \le - \frac{\eta C_3 \sigma_p^2 d \Vert \bm{w}_O \Vert_2^2 (t - T_1)}{N}
    \]
    \[ 
        V_{n, 2}^{(t)} \ge 3 M \cdot \max \{ \vert V_+^{(T_1)} \vert, \vert V_{n, i}^{(T_1)} \vert \}
    \]
    for all \( n \in S_+ \), \( i \in [M] \backslash \{1, 2\} \).
    \[ 
        V_{n, 2}^{(t)} \le - 3 M \cdot \max \{ \vert V_-^{(T_1)} \vert, \vert V_{n, i}^{(T_1)} \vert \} 
    \]
    for all \( n \in S_- \), \( i \in [M] \backslash \{1, 2\} \).
    \[
        \vert V_\pm^{(t)} \vert \le O(d_h^{-\frac{1}{4}}) + \frac{\eta C_4 \sigma_p^2 d \Vert \bm{w}_O \Vert_2^2 (t - T_1)}{N}
    \]
    \[
        \vert V_{n, i}^{(t)} \vert \le O(d_h^{-\frac{1}{4}}) + \frac{\eta C_4 \sigma_p^2 d \Vert \bm{w}_O \Vert_2^2 (t - T_1)}{N}
    \]
    for \(i \in [M] \backslash \{1\}, n \in [N]\).

    \item \( \mathcal{K}(t): \)
    \[
        \Vert \bm{q}_\pm^{(t)} \Vert_2^2, \Vert \bm{k}_\pm^{(t)} \Vert_2^2 = \Theta \Big( \Vert \bm{\mu} \Vert_2^2 \sigma_h^2 d_h \Big),
    \]
    \[
        \Vert \bm{q}_{n, i}^{(t)} \Vert_2^2, \Vert \bm{k}_{n, i}^{(t)} \Vert_2^2 = \Theta \Big( \sigma_p^2 \sigma_h^2 d d_h \Big),
    \]
    \[
        \vert \langle \bm{q}_+^{(t)}, \bm{q}_-^{(t)} \rangle \vert, \vert \langle \bm{q}_\pm^{(t)}, \bm{q}_{n, i}^{(t)} \rangle \vert, \vert \langle \bm{q}_{n, i}^{(t)}, \bm{q}_{n^\prime, j}^{(t)} \rangle \vert = o(1),
    \]
    \[
        \vert \langle \bm{k}_+^{(t)}, \bm{k}_-^{(t)} \rangle \vert, \vert \langle \bm{k}_\pm^{(t)}, \bm{k}_{n, i}^{(t)} \rangle \vert, \vert \langle \bm{k}_{n, i}^{(t)}, \bm{k}_{n^\prime, j}^{(t)} \rangle \vert = o(1),
    \]
    for \(i, j \in [M] \backslash \{1\}, n, n^\prime \in [N], i \ne j \ or \ n \ne n^\prime \).

    \item \( \mathcal{L}(t): \)
    \[
        \langle \bm{q}_\pm^{(t + 1)}, \bm{k}_{n, 2}^{(t + 1)} \rangle - \langle \bm{q}_\pm^{(t)}, \bm{k}_{n, 2}^{(t)} \rangle \ge 2 \big( \langle \bm{q}_\pm^{(t + 1)}, \bm{k}_\pm^{(t + 1)} \rangle - \langle \bm{q}_\pm^{(t)}, \bm{k}_\pm^{(t)} \rangle \big)
    \]
    \[
        \langle \bm{q}_\pm^{(t + 1)}, \bm{k}_{n, 2}^{(t + 1)} \rangle - \langle \bm{q}_\pm^{(t)}, \bm{k}_{n, 2}^{(t)} \rangle \ge 2 \big( \langle \bm{q}_\pm^{(t + 1)}, \bm{k}_{n, j}^{(t + 1)} \rangle - \langle \bm{q}_\pm^{(t)}, \bm{k}_{n, j}^{(t)} \rangle \big)
    \]
    \[
        \langle \bm{q}_{n, i}^{(t + 1)}, \bm{k}_{n, 2}^{(t + 1)} \rangle - \langle \bm{q}_{n, i}^{(t)}, \bm{k}_{n, 2}^{(t)} \rangle \ge 2 \big( \langle \bm{q}_{n, i}^{(t + 1)}, \bm{k}_\pm^{(t + 1)} \rangle - \langle \bm{q}_{n, i}^{(t)}, \bm{k}_\pm^{(t)} \rangle \big)
    \]
    \[
        \langle \bm{q}_{n, i}^{(t + 1)}, \bm{k}_{n, 2}^{(t + 1)} \rangle - \langle \bm{q}_{n, i}^{(t)}, \bm{k}_{n, 2}^{(t)} \rangle \ge 2 \big( \langle \bm{q}_{n, i}^{(t + 1)}, \bm{k}_{n, j}^{(t + 1)} \rangle - \langle \bm{q}_{n, i}^{(t)}, \bm{k}_{n, j}^{(t)} \rangle \big)
    \]
    \[
        \Psi_{\pm}^{(t + 1)} \ge \log\Big( \exp(\Psi_{\pm}^{(T_1)}) + \frac{\eta^2 C_7 \Vert \bm{\mu} \Vert_2^2 \sigma_p^2 d \Vert \bm{w}_O \Vert_2^2 d_h^{\frac{1}{2}}}{N^2 \big( \log (6N^2M^2 / \delta) \big)^{2} } \cdot (t - T_1)(t - T_1 + 1) \Big)
    \]
    \[
        \Psi_{n, \pm, j}^{(t + 1)} \ge \log\Big( \exp(\Psi_{n, \pm, j}^{(T_1)}) + \frac{\eta^2 C_7 \Vert \bm{\mu} \Vert_2^2 \sigma_p^2 d \Vert \bm{w}_O \Vert_2^2 d_h^{\frac{1}{2}}}{N^2 \big( \log (6N^2M^2 / \delta) \big)^{2} } \cdot (t - T_1)(t - T_1 + 1) \Big)
    \]
    \[
        \Psi_{n, i, \pm}^{(t + 1)} \ge \log\Big( \exp(\Psi_{n, i, \pm}^{(T_1)}) + \frac{\eta^2 C_7 \sigma_p^4 d^2 \Vert \bm{w}_O \Vert_2^2 d_h^{\frac{1}{2}}}{N^2 \big( \log (6N^2M^2 / \delta) \big)^{2} } \cdot (t - T_1)(t - T_1 + 1) \Big)
    \]
    \[
        \Psi_{n, i, j}^{(t + 1)} \ge \log\Big( \exp(\Psi_{n, i, j}^{(T_1)}) + \frac{\eta^2 C_7 \sigma_p^4 d^2 \Vert \bm{w}_O \Vert_2^2 d_h^{\frac{1}{2}}}{N^2 \big( \log (6N^2M^2 / \delta) \big)^{2} } \cdot (t - T_1)(t - T_1 + 1) \Big)
    \]
    for \(i \in [M] \backslash \{1\}, j \in [M] \backslash \{1, 2\}, n \in S_\pm \).

    \item \( \mathcal{M}(t): \)
    \[
        \vert \Psi_{\pm}^{(t)} \vert, \vert \Psi_{n, \pm, j^\prime}^{(t)} \vert, \vert \Psi_{n, i, \pm}^{(t)} \vert, \vert \Psi_{n, i, j^\prime}^{(t)} \vert \le \log(d_h^{\frac{1}{2}})
    \]
    \[
        \vert \langle \bm{q}_\pm^{(t)}, \bm{k}_\pm^{(t)} \rangle \vert, \vert \langle \bm{q}_\pm^{(t)}, \bm{k}_{n, j}^{(t)} \rangle \vert, \vert \langle \bm{q}_{n, i}^{(t)}, \bm{k}_\pm^{(t)} \rangle \vert, \vert \langle \bm{q}_{n, i}^{(t)}, \bm{k}_{n, j}^{(t)} \rangle \vert \le 2\log(d_h^{\frac{1}{2}})
    \]
    \[
        \vert \langle \bm{q}_\pm^{(t)}, \bm{k}_\mp^{(t)} \rangle \vert, \vert \langle \bm{q}_{n, i}^{(t)}, \bm{k}_{\overline{n}, j}^{(t)} \rangle \vert = o(1)
    \]
    for \(i, j \in [M] \backslash \{1\}, j^\prime \in [M] \backslash \{1, 2\}, n, \overline{n} \in [N], n \ne \overline{n} \).
\end{itemize}

By the results of Stage \uppercase\expandafter{\romannumeral1}, we know that \( \mathcal{J}(T_1) \), \( \mathcal{K}(T_1) \), \( \mathcal{M}(T_1) \) are true. To prove that \( \mathcal{J}(t) \), \( \mathcal{K}(t) \), \( \mathcal{L}(t) \) and \( \mathcal{M}(t) \) are true in stage 2, we give the following claims holds for \(t \in [T_1, T_2]\):
\begin{claim}
    \label{claim9}
    \( \mathcal{L}(T_1), \dots, \mathcal{L}(t-1) \Longrightarrow \mathcal{J}(t + 1) \) 
\end{claim}
\begin{claim}
    \label{claim10}
    \( \mathcal{J}(T_1), \dots, \mathcal{J}(t), \mathcal{K}(T_1), \dots, \mathcal{K}(t), \mathcal{L}(T_1), \dots, \mathcal{L}(t - 1) \Longrightarrow \mathcal{L}(t) \) 
\end{claim}
\begin{claim}
    \label{claim11}
    \( \mathcal{J}(T_1), \dots, \mathcal{J}(t), \mathcal{L}(T_1), \dots, \mathcal{L}(t - 1), \mathcal{M}(T_1), \dots, \mathcal{M}(t) \Longrightarrow \mathcal{K}(t + 1) \) 
\end{claim}
\begin{claim}
    \label{claim12}
    \( \mathcal{J}(T_1), \dots, \mathcal{J}(t), \mathcal{K}(T_1), \dots, \mathcal{K}(t), \mathcal{L}(T_1), \dots, \mathcal{L}(t - 1) \Longrightarrow \mathcal{M}(t + 1) \)
\end{claim}

The proofs of Claim \ref{claim9}-\ref{claim12} are similar to the proofs of Claim \ref{claim5}-\ref{claim8}.
Next, we show some important procedures for proving Claim \ref{claim10}.

\textbf{Proof of Claim \ref{claim10}:}

Similar to \ref{claim2_proof}, we have the dynamic of \(\Psi\) as follows
\begin{equation}
\begin{split}
    \Psi_{\pm}^{(s+1)} - \Psi_{\pm}^{(s)} \ge \frac{\eta^2 C_7 \Vert \bm{\mu} \Vert_2^2 \sigma_p^2 d \Vert \bm{w}_O \Vert_2^2 d_h^{\frac{1}{2}} (s - T_1)}{N^2 \big( \log (6N^2M^2 / \delta) \big)^{2} } \cdot \frac{1}{\exp(\Psi_{\pm}^{(s)})},
\end{split}
\end{equation}
\begin{equation}
\begin{split}
    \Psi_{n, \pm, j}^{(s+1)} - \Psi_{n, \pm, j}^{(s)} \ge \frac{\eta^2 C_7 \Vert \bm{\mu} \Vert_2^2 \sigma_p^2 d \Vert \bm{w}_O \Vert_2^2 d_h^{\frac{1}{2}} (s - T_1)}{N^2 \big( \log (6N^2M^2 / \delta) \big)^{2} } \cdot \frac{1}{\exp(\Psi_{n, \pm, j}^{(s)})},
\end{split}
\end{equation}
\begin{equation}
\begin{split}
    \Psi_{n, i, \pm}^{(s+1)} - \Psi_{n, i, \pm}^{(s)} \ge \frac{\eta^2 C_7 \sigma_p^4 d^2 \Vert \bm{w}_O \Vert_2^2 d_h^{\frac{1}{2}} (s - T_1)}{N^2 \big( \log (6N^2M^2 / \delta) \big)^{2} } \cdot \frac{1}{\exp(\Psi_{n, i, \pm}^{(s)})},
\end{split}
\end{equation}
\begin{equation}
\begin{split}
    \Psi_{n, i, j}^{(s+1)} - \Psi_{n, i, j}^{(s)} \ge \frac{\eta^2 C_7 \sigma_p^4 d^2 \Vert \bm{w}_O \Vert_2^2 d_h^{\frac{1}{2}} (s - T_1)}{N^2 \big( \log (6N^2M^2 / \delta) \big)^{2} } \cdot \frac{1}{\exp(\Psi_{n, i, j}^{(s)})},
\end{split}
\end{equation}
for \(n \in [N], i \in [M] \backslash \{1\}, j \in [M] \backslash \{1, 2\}, s \in [T_1, t]\).
Next, we provide the logarithmic increasing lower bounds of \(\Psi\).
Recall that
\begin{equation}
\begin{split}
    \Psi_{+}^{(s+1)} - \Psi_{+}^{(s)} \ge \frac{\eta^2 C_7 \Vert \bm{\mu} \Vert_2^2 \sigma_p^2 d \Vert \bm{w}_O \Vert_2^2 d_h^{\frac{1}{2}} (s - T_1)}{N^2 \big( \log (6N^2M^2 / \delta) \big)^{2} } \cdot \frac{1}{\exp(\Psi_{+}^{(s)})},
\end{split}
\end{equation}
Multiply both sides simultaneously by \(\exp(\Psi_{+}^{(s)})\) and get
\begin{equation}
    \exp(\Psi_{+}^{(s)}) \Big( \Psi_{+}^{(s+1)} - \Psi_{+}^{(s)} \Big) \ge \frac{\eta^2 C_7 \Vert \bm{\mu} \Vert_2^2 \sigma_p^2 d \Vert \bm{w}_O \Vert_2^2 d_h^{\frac{1}{2}} (s - T_1)}{N^2 \big( \log (6N^2M^2 / \delta) \big)^{2} }.
\end{equation}
Taking a summation from \(T_1\) to \(t\) and get
\begin{equation}
\begin{split}
    &\sum\limits_{s = T_1}^t \exp(\Psi_{+}^{(s)}) \Big( \Psi_{+}^{(s+1)} - \Psi_{+}^{(s)} \Big) \\
    &\ge \sum\limits_{s = T_1}^t \frac{\eta^2 C_7 \Vert \bm{\mu} \Vert_2^2 \sigma_p^2 d \Vert \bm{w}_O \Vert_2^2 d_h^{\frac{1}{2}} (s - T_1)}{N^2 \big( \log (6N^2M^2 / \delta) \big)^{2} } \\
    &\ge \frac{\eta^2 C_7 \Vert \bm{\mu} \Vert_2^2 \sigma_p^2 d \Vert \bm{w}_O \Vert_2^2 d_h^{\frac{1}{2}}}{N^2 \big( \log (6N^2M^2 / \delta) \big)^{2} } \cdot (t - T_1)(t - T_1 + 1).
\end{split}
\end{equation}
By the property that \(\Psi_{+}^{(s)}\) is monotonically increasing, we have
\begin{equation}
\begin{split}
    &\int_{\Psi_{+}^{(T_1)}}^{\Psi_{+}^{(t + 1)}} \exp(x) dx \ge \sum\limits_{s = T_1}^t \exp(\Psi_{+}^{(s)}) \Big( \Psi_{+}^{(s+1)} - \Psi_{+}^{(s)} \Big) \\
    &\ge \frac{\eta^2 C_7 \Vert \bm{\mu} \Vert_2^2 \sigma_p^2 d \Vert \bm{w}_O \Vert_2^2 d_h^{\frac{1}{2}}}{N^2 \big( \log (6N^2M^2 / \delta) \big)^{2} } \cdot (t - T_1)(t - T_1 + 1).
\end{split}
\end{equation}
By \(\int_{\Psi_{+}^{(T_1)}}^{\Psi_{+}^{(t + 1)}} \exp(x) dx = \exp(\Psi_{+}^{(t + 1)}) - \exp(\Psi_{+}^{(T_1)}) \) we get
\begin{equation}
\begin{split}
    \Psi_{+}^{(t + 1)} \ge \log\Big( \exp(\Psi_{+}^{(T_1)}) + \frac{\eta^2 C_7 \Vert \bm{\mu} \Vert_2^2 \sigma_p^2 d \Vert \bm{w}_O \Vert_2^2 d_h^{\frac{1}{2}}}{N^2 \big( \log (6N^2M^2 / \delta) \big)^{2} } \cdot (t - T_1)(t - T_1 + 1) \Big).
\end{split}
\end{equation}
Similarly, we have 
\begin{equation}
\begin{split}
    \Psi_{-}^{(t + 1)} \ge \log\Big( \exp(\Psi_{-}^{(T_1)}) + \frac{\eta^2 C_7 \Vert \bm{\mu} \Vert_2^2 \sigma_p^2 d \Vert \bm{w}_O \Vert_2^2 d_h^{\frac{1}{2}}}{N^2 \big( \log (6N^2M^2 / \delta) \big)^{2} } \cdot (t - T_1)(t - T_1 + 1) \Big).
\end{split}
\end{equation}

\begin{equation}
\begin{split}
    \Psi_{n, \pm, j}^{(t + 1)} \ge \log\Big( \exp(\Psi_{n, \pm, j}^{(T_1)}) + \frac{\eta^2 C_7 \Vert \bm{\mu} \Vert_2^2 \sigma_p^2 d \Vert \bm{w}_O \Vert_2^2 d_h^{\frac{1}{2}}}{N^2 \big( \log (6N^2M^2 / \delta) \big)^{2} } \cdot (t - T_1)(t - T_1 + 1) \Big).
\end{split}
\end{equation}

\begin{equation}
\begin{split}
    \Psi_{n, i, \pm}^{(t + 1)} \ge \log\Big( \exp(\Psi_{n, i, \pm}^{(T_1)}) + \frac{\eta^2 C_7 \sigma_p^4 d^2 \Vert \bm{w}_O \Vert_2^2 d_h^{\frac{1}{2}}}{N^2 \big( \log (6N^2M^2 / \delta) \big)^{2} } \cdot (t - T_1)(t - T_1 + 1) \Big).
\end{split}
\end{equation}

\begin{equation}
\begin{split}
    \Psi_{n, i, j}^{(t + 1)} \ge \log\Big( \exp(\Psi_{n, i, j}^{(T_1)}) + \frac{\eta^2 C_7 \sigma_p^4 d^2 \Vert \bm{w}_O \Vert_2^2 d_h^{\frac{1}{2}}}{N^2 \big( \log (6N^2M^2 / \delta) \big)^{2} } \cdot (t - T_1)(t - T_1 + 1) \Big).
\end{split}
\end{equation}
for \(n \in [N], i \in [M] \backslash \{1\}, j \in [M] \backslash \{1, 2\}\).

\subsection{Stage \uppercase\expandafter{\romannumeral3}}
\label{stage3_harmful}
In Stage \uppercase\expandafter{\romannumeral3}, the outputs of ViT grow up and the loss derivatives are no longer at \(o(1)\). We will carefully compute the growth rate of \(V_\pm\) and \(V_{n, i}\) while keeping monitoring the monotonicity of \(\langle \bm{q}, \bm{k} \rangle\). 
By substituting \(t = T_2 = \Theta \Big( \frac{N}{\eta \sigma_p^2 d \Vert \bm{w}_O \Vert_2^2} \Big) \) into propositions \( \mathcal{J}(t) \), \( \mathcal{K}(t) \), \( \mathcal{L}(t) \), \( \mathcal{M}(t) \) in Stage \uppercase\expandafter{\romannumeral2}, we have the following conditions at the beginning of stage \uppercase\expandafter{\romannumeral3}
\[
    \vert V_+^{(T_2)} \vert, \vert V_-^{(T_2)} \vert, \vert V_{n, i}^{(T_2)} \vert = o (1)
\]
for all \( n \in [N] \), \( i \in [M] \backslash \{1\} \).
\[ 
    V_{n, 2}^{(T_1)} \ge 3 M \cdot \max \{ \vert V_+^{(T_1)} \vert, \vert V_{n, i}^{(T_1)} \vert \}
\]
for all \( n \in S_+ \), \( i \in [M] \backslash \{1, 2\} \).
\[ 
    V_{n, 2}^{(T_1)} \le - 3 M \cdot \max \{ \vert V_-^{(T_1)} \vert, \vert V_{n, i}^{(T_1)} \vert \} 
\]
for all \( n \in S_- \), \( i \in [M] \backslash \{1, 2\} \).
\[
    \Vert \bm{q}_\pm^{(T_2)} \Vert_2^2, \Vert \bm{k}_\pm^{(T_2)} \Vert_2^2 = \Theta \Big( \Vert \bm{\mu} \Vert_2^2 \sigma_h^2 d_h \Big),
\]
\[
    \Vert \bm{q}_{n, i}^{(T_2)} \Vert_2^2, \Vert \bm{k}_{n, i}^{(T_2)} \Vert_2^2 = \Theta \Big( \sigma_p^2 \sigma_h^2 d d_h \Big),
\]
\[
    \vert \langle \bm{q}_+^{(T_2)}, \bm{q}_-^{(T_2)} \rangle \vert, \vert \langle \bm{q}_\pm^{(T_2)}, \bm{q}_{n, i}^{(T_2)} \rangle \vert, \vert \langle \bm{q}_{n, i}^{(T_2)}, \bm{q}_{n^\prime, j}^{(T_2)} \rangle \vert = o(1),
\]
\[
    \vert \langle \bm{k}_+^{(T_2)}, \bm{k}_-^{(T_2)} \rangle \vert, \vert \langle \bm{k}_\pm^{(T_2)}, \bm{k}_{n, i}^{(T_2)} \rangle \vert, \vert \langle \bm{k}_{n, i}^{(T_2)}, \bm{k}_{n^\prime, j}^{(T_2)} \rangle \vert = o(1),
\]
for \(i, j \in [M] \backslash \{1\}, n, n^\prime \in [N], i \ne j \ or \ n \ne n^\prime \).
\[
    \Psi_{\pm}^{(T_2)} \ge \log\Big( \exp(\Psi_{\pm}^{(T_1)}) + \frac{C_7 \Vert \bm{\mu} \Vert_2^2 \Vert \bm{w}_O \Vert_2^2 d_h^{\frac{1}{2}}}{\sigma_p^2 d \big( \log (6N^2M^2 / \delta) \big)^{3} } \Big)
\]
\[
    \Psi_{n, \pm, j}^{(T_2)} \ge \log\Big( \exp(\Psi_{n, \pm, j}^{(T_1)}) + \frac{ C_7 \Vert \bm{\mu} \Vert_2^2 \Vert \bm{w}_O \Vert_2^2 d_h^{\frac{1}{2}}}{\sigma_p^2 d \big( \log (6N^2M^2 / \delta) \big)^{3} } \Big)
\]
\[
    \Psi_{n, i, \pm}^{(T_2)} \ge \log\Big( \exp(\Psi_{n, i, \pm}^{(T_1)}) + \frac{ C_7 \Vert \bm{w}_O \Vert_2^2 d_h^{\frac{1}{2}}}{\big( \log (6N^2M^2 / \delta) \big)^{3} } \Big)
\]
\[
    \Psi_{n, i, j}^{(T_2)} \ge \log\Big( \exp(\Psi_{n, i, j}^{(T_1)}) + \frac{ C_7 \Vert \bm{w}_O \Vert_2^2 d_h^{\frac{1}{2}}}{ \big( \log (6N^2M^2 / \delta) \big)^{3} } \Big)
\]
\[
    \vert \Psi_{\pm}^{(t)} \vert, \vert \Psi_{n, \pm, j^\prime}^{(t)} \vert, \vert \Psi_{n, i, \pm}^{(t)} \vert, \vert \Psi_{n, i, j^\prime}^{(t)} \vert \le \log(d_h^{\frac{1}{2}})
\]
\[
    \vert \langle \bm{q}_\pm^{(T_2)}, \bm{k}_\pm^{(T_2)} \rangle \vert, \vert \langle \bm{q}_\pm^{(T_2)}, \bm{k}_{n, j}^{(T_2)} \rangle \vert, \vert \langle \bm{q}_{n, i}^{(T_2)}, \bm{k}_\pm^{(T_2)} \rangle \vert, \vert \langle \bm{q}_{n, i}^{(T_2)}, \bm{k}_{n, j}^{(T_2)} \rangle \vert \le 2\log(d_h^{\frac{1}{2}})
\]
\[
    \vert \langle \bm{q}_\pm^{(T_2)}, \bm{k}_\mp^{(T_2)} \rangle \vert, \vert \langle \bm{q}_{n, i}^{(T_2)}, \bm{k}_{\overline{n}, j}^{(T_2)} \rangle \vert = o(1)
\]
for \(i, j \in [M] \backslash \{1\}, j^\prime \in [M] \backslash \{1, 2\}, n, \overline{n} \in [N], n \ne \overline{n} \).

Let \(T_3 = \Theta \Big( \frac{N}{\eta \epsilon \sigma_p^2 d \Vert \bm{w}_O \Vert_2^2} \Big)\), 
Next we prove the following four propositions \( \mathcal{N}(t) \), \( \mathcal{O}(t) \), \( \mathcal{P}(t) \), \( \mathcal{Q}(t) \) by induction on t for \(t \in [T_2, T_3]\):
\begin{itemize}
    \item \( \mathcal{N}(t): \)
    \[ 
        V_{n, 2}^{(t)} \ge 3 M \cdot \max \{ \vert V_+^{(T_1)} \vert, \vert V_{n, i}^{(T_1)} \vert \}
    \]
    for all \( n \in S_+ \), \( i \in [M] \backslash \{1, 2\} \).
    \[ 
        V_{n, 2}^{(t)} \le - 3 M \cdot \max \{ \vert V_-^{(T_1)} \vert, \vert V_{n, i}^{(T_1)} \vert \} 
    \]
    for all \( n \in S_- \), \( i \in [M] \backslash \{1, 2\} \).
    \[
        \vert V_\pm^{(t)} \vert, \vert V_{n, i}^{(t)} \vert = o (1),  
    \]
    for \(i \in [M] \backslash \{1, 2\}, n \in [N]\).
    \[
        \log \Big( \exp ( V_{n, 2}^{(T_2)}) + \frac{\eta C_{17} \sigma_p^2 d \Vert \bm{w}_O \Vert_2^2 (t - T_2)}{N} \Big) \le V_{n, 2}^{(t)} \le 2 \log \big( O(\frac{1}{\epsilon}) \big), 
    \]
    for all \( n \in S_+ \).
    \[
        - 2 \log \big( O(\frac{1}{\epsilon}) \big) \le V_{n, 2}^{(t)} \le - \log \Big( \exp ( - V_{n, 2}^{(T_2)}) + \frac{\eta C_{17} \sigma_p^2 d \Vert \bm{w}_O \Vert_2^2 (t - T_2)}{N} \Big)
    \]
    for all \( n \in S_- \).

    \item \( \mathcal{O}(t): \)
    \[
        \Vert \bm{q}_\pm^{(t)} \Vert_2^2, \Vert \bm{k}_\pm^{(t)} \Vert_2^2 = \Theta (\Vert \bm{\mu} \Vert_2^2 \sigma_h^2 d_h),
    \]
    \[
        \Vert \bm{q}_{n, i}^{(t)} \Vert_2^2, \Vert \bm{k}_{n, i}^{(t)} \Vert_2^2 = \Theta \Big( \sigma_p^2 \sigma_h^2 d d_h \Big),
    \]
    \[
        \vert \langle \bm{q}_+^{(t)}, \bm{q}_-^{(t)} \rangle \vert, \vert \langle \bm{q}_\pm^{(t)}, \bm{q}_{n, i}^{(t)} \rangle \vert, \vert \langle \bm{q}_{n, i}^{(t)}, \bm{q}_{n^\prime, j}^{(t)} \rangle \vert = o(1),
    \]
    \[
        \vert \langle \bm{k}_+^{(t)}, \bm{k}_-^{(t)} \rangle \vert, \vert \langle \bm{k}_\pm^{(t)}, \bm{k}_{n, i}^{(t)} \rangle \vert, \vert \langle \bm{k}_{n, i}^{(t)}, \bm{k}_{n^\prime, j}^{(t)} \rangle \vert = o(1)
    \]
    for \(i, j \in [M] \backslash \{1\}, n, n^\prime \in [N], i \ne j \ or \ n \ne n^\prime \).

    \item \( \mathcal{P}(t): \)
    \[
        \langle \bm{q}_\pm^{(t + 1)}, \bm{k}_{n, 2}^{(t + 1)} \rangle - \langle \bm{q}_\pm^{(t)}, \bm{k}_{n, 2}^{(t)} \rangle \ge 2 \big( \langle \bm{q}_\pm^{(t + 1)}, \bm{k}_\pm^{(t + 1)} \rangle - \langle \bm{q}_\pm^{(t)}, \bm{k}_\pm^{(t)} \rangle \big)
    \]
    \[
        \langle \bm{q}_\pm^{(t + 1)}, \bm{k}_{n, 2}^{(t + 1)} \rangle - \langle \bm{q}_\pm^{(t)}, \bm{k}_{n, 2}^{(t)} \rangle \ge 2 \big( \langle \bm{q}_\pm^{(t + 1)}, \bm{k}_{n, j}^{(t + 1)} \rangle - \langle \bm{q}_\pm^{(t)}, \bm{k}_{n, j}^{(t)} \rangle \big)
    \]
    \[
        \langle \bm{q}_{n, i}^{(t + 1)}, \bm{k}_{n, 2}^{(t + 1)} \rangle - \langle \bm{q}_{n, i}^{(t)}, \bm{k}_{n, 2}^{(t)} \rangle \ge 2 \big( \langle \bm{q}_{n, i}^{(t + 1)}, \bm{k}_\pm^{(t + 1)} \rangle - \langle \bm{q}_{n, i}^{(t)}, \bm{k}_\pm^{(t)} \rangle \big)
    \]
    \[
        \langle \bm{q}_{n, i}^{(t + 1)}, \bm{k}_{n, 2}^{(t + 1)} \rangle - \langle \bm{q}_{n, i}^{(t)}, \bm{k}_{n, 2}^{(t)} \rangle \ge 2 \big( \langle \bm{q}_{n, i}^{(t + 1)}, \bm{k}_{n, j}^{(t + 1)} \rangle - \langle \bm{q}_{n, i}^{(t)}, \bm{k}_{n, j}^{(t)} \rangle \big)
    \]
    \[
        \Psi_{\pm}^{(t + 1)} \ge \Psi_{\pm}^{(t)}
    \]
    \[
        \Psi_{n, \pm, j}^{(t + 1)} \ge \Psi_{n, \pm, j}^{(t)}
    \]
    \[
        \Psi_{n, i, \pm}^{(t + 1)} \ge \Psi_{n, i, \pm}^{(t)}
    \]
    \[
        \Psi_{n, i, j}^{(t + 1)} \ge \Psi_{n, i, j}^{(t)}
    \]
    for \(i \in [M] \backslash \{1\}, j \in [M] \backslash \{1, 2\}, n \in S_\pm \).

    \item \( \mathcal{Q}(t): \)
    \[
        \vert \Psi_{\pm}^{(t)} \vert, \vert \Psi_{n, \pm, j^\prime}^{(t)} \vert, \vert \Psi_{n, i, \pm}^{(t)} \vert, \vert \Psi_{n, i, j^\prime}^{(t)} \vert \le \log ( \epsilon^{-1} d_h^{\frac{1}{2}} )
    \]
    \[
        \vert \langle \bm{q}_\pm^{(t)}, \bm{k}_\pm^{(t)} \rangle \vert, \vert \langle \bm{q}_\pm^{(t)}, \bm{k}_{n, j}^{(t)} \rangle \vert, \vert \langle \bm{q}_{n, i}^{(t)}, \bm{k}_\pm^{(t)} \rangle \vert, \vert \langle \bm{q}_{n, i}^{(t)}, \bm{k}_{n, j}^{(t)} \rangle \vert \le 2\log ( \epsilon^{-1} d_h^{\frac{1}{2}} )
    \]
    \[
        \vert \langle \bm{q}_\pm^{(t)}, \bm{k}_\mp^{(t)} \rangle \vert, \vert \langle \bm{q}_{n, i}^{(t)}, \bm{k}_{\overline{n}, j}^{(t)} \rangle \vert = o(1)
    \]
    for \(i, j \in [M] \backslash \{1\}, j^\prime \in [M] \backslash \{1, 2\}, n, \overline{n} \in [N], n \ne \overline{n} \).
\end{itemize}
By the results of Stage \uppercase\expandafter{\romannumeral2}, we know that \( \mathcal{N}(T_1) \), \( \mathcal{O}(T_2) \), \( \mathcal{Q}(T_2) \) are true. To prove that \( \mathcal{N}(t) \), \( \mathcal{O}(t) \), \( \mathcal{P}(t) \) and \( \mathcal{Q}(t) \) are true in Stage \uppercase\expandafter{\romannumeral3}, we will prove the following claims holds for \(t \in [T_2, T_3]\):
\begin{claim}
    \label{claim13}
    \( \mathcal{P}(T_2), \dots, \mathcal{P}(t - 1) \Longrightarrow \mathcal{N}(t + 1) \) 
\end{claim}
\begin{claim}
    \label{claim14}
    \( \mathcal{N}(t), \mathcal{O}(t), \mathcal{P}(T_2), \dots, \mathcal{P}(t - 1) \Longrightarrow \mathcal{P}(t) \) 
\end{claim}
\begin{claim}
    \label{claim15}
    \( \mathcal{N}(T_2), \dots, \mathcal{N}(t), \mathcal{P}(T_2), \dots, \mathcal{P}(t - 1), \mathcal{Q}(T_2), \dots, \mathcal{Q}(t) \Longrightarrow \mathcal{O}(t + 1) \) 
\end{claim}
\begin{claim}
    \label{claim16}
    \( \mathcal{N}(T_2), \dots, \mathcal{N}(t), \mathcal{O}(T_2), \dots, \mathcal{O}(t), \mathcal{P}(T_2), \dots, \mathcal{P}(t - 1) \Longrightarrow \mathcal{Q}(t + 1) \)
\end{claim}

\begin{lemma}[Convergence of Training Loss]
    \label{stage3_loss_harmful}
    There exist \(T = \frac{C_{19} N}{\eta \epsilon \sigma_p^2 d \Vert \bm{w}_O \Vert_2^2}\) such that
    \[
        L_S(\theta(T)) \ge 0.1
    \]
\end{lemma}
\it Proof of Lemma \ref{stage3_loss_harmful}. \rm
Similar to \ref{stage3_loss}, we have

\begin{equation}
    softmax(\langle \bm{q}_\pm^{(T)}, \bm{k}_\pm^{(T)} \rangle) = O \Big( \frac{ \sigma_p^2 d \big( \log (6N^2M^2 / \delta) \big)^{3} }{ \Vert \bm{\mu} \Vert_2^2 \Vert \bm{w}_O \Vert_2^2 d_h^{\frac{1}{2}} } \Big),
\end{equation}
\begin{equation}
    softmax(\langle \bm{q}_\pm^{(T)}, \bm{k}_{n, j}^{(t)} \rangle) = O \Big( \frac{ \sigma_p^2 d \big( \log (6N^2M^2 / \delta) \big)^{3} }{ \Vert \bm{\mu} \Vert_2^2 \Vert \bm{w}_O \Vert_2^2 d_h^{\frac{1}{2}} } \Big),
\end{equation}
\begin{equation}
    softmax(\langle \bm{q}_{n, i}^{(t)}, \bm{k}_\pm^{(T)} \rangle) = O \Big( \frac{ \big( \log (6N^2M^2 / \delta) \big)^{3} }{ \Vert \bm{w}_O \Vert_2^2 d_h^{\frac{1}{2}} } \Big),
\end{equation}
\begin{equation}
    softmax(\langle \bm{q}_{n, i}^{(t)}, \bm{k}_{n, j}^{(t)} \rangle) = O \Big( \frac{ \big( \log (6N^2M^2 / \delta) \big)^{3} }{ \Vert \bm{w}_O \Vert_2^2 d_h^{\frac{1}{2}} } \Big),
\end{equation}
\begin{equation}
    softmax(\langle \bm{q}_\pm^{(t)}, \bm{k}_{n, 2}^{(t)} \rangle) = 1 - O \Big( \frac{ \sigma_p^2 d \big( \log (6N^2M^2 / \delta) \big)^{3} }{ \Vert \bm{\mu} \Vert_2^2 \Vert \bm{w}_O \Vert_2^2 d_h^{\frac{1}{2}} } \Big),
\end{equation}
\begin{equation}
    softmax(\langle \bm{q}_{n, i}^{(t)}, \bm{k}_{n, 2}^{(t)} \rangle) = 1 - O \Big( \frac{ \big( \log (6N^2M^2 / \delta) \big)^{3} }{ \Vert \bm{w}_O \Vert_2^2 d_h^{\frac{1}{2}} } \Big)
\end{equation}
for \(i \in [M] \backslash \{1\}, j \in [M] \backslash \{1, 2\}, n \in [N]\).

For \(n \in S_+\), substituting \(t = T = \frac{C_{19} N}{\eta \epsilon \sigma_p^2 d \Vert \bm{w}_O \Vert_2^2} \) into propositions \( \mathcal{N}(t) \) and get
\begin{equation}
\begin{split}
    V_{n, 2}^{(t)} &\ge \log \Big( \exp ( V_{n, 2}^{(T_2)}) + \frac{\eta C_{17} \sigma_p^2 d \Vert \bm{w}_O \Vert_2^2}{N} (t - T_2) \Big) \\
    &\ge \log \Big( \exp ( V_{n, 2}^{(T_2)}) + \frac{C_{20}}{\epsilon} \Big) \\
    &\ge \log \Big( \frac{C_{20}}{\epsilon} \Big),
\end{split}
\end{equation}
\begin{equation}
    \vert V_+^{(t)} \vert, \vert V_{n, i}^{(t)} \vert = o (1)
\end{equation}
 we bound \(f(\bm{X}_n, \theta (t))\) as follows
\begin{equation}
\begin{split}
    f(\bm{X}_n, \theta (t)) &= \frac{1}{M} \sum\limits_{l=1}^M \bm{\varphi} (\bm{x}_{n, l}^\top \bm{W}_{Q}^{(t)} \bm{W}_{K}^{(t)\top} \bm{X}_n^\top ) \bm{X}_n \bm{W}_V^{(t)} \bm{w}_O \\
    &\frac{1}{M} \cdot \Big( \big( \frac{ \exp (\langle \bm{q}_+^{(t)}, \bm{k}_+^{(t)} \rangle)}{ \exp  (\langle \bm{q}_+^{(t)}, \bm{k}_+^{(t)} \rangle) + \sum\limits_{j=2}^M  \exp  ( \langle \bm{q}_+^{(t)}, \bm{k}_{n, j}^{(t)} \rangle ) } \\
    &+ \sum\limits_{i=2}^{M} \frac{ \exp (\langle \bm{q}_{n, i}^{(t)}, \bm{k}_+^{(t)} \rangle)}{ \exp  (\langle \bm{q}_{n, i}^{(t)}, \bm{k}_+^{(t)} \rangle) + \sum\limits_{j=2}^M  \exp  ( \langle \bm{q}_{n, i}^{(t)}, \bm{k}_{n, j}^{(t)} \rangle ) } \big) \cdot V_+^{(T)} \\
    &+ \big( \frac{ \exp (\langle \bm{q}_+^{(t)}, \bm{k}_{n, 2}^{(t)} \rangle)}{ \exp  (\langle \bm{q}_+^{(t)}, \bm{k}_+^{(t)} \rangle) + \sum\limits_{j=2}^M  \exp  ( \langle \bm{q}_+^{(t)}, \bm{k}_{n, j}^{(t)} \rangle ) } \\
    &+ \sum\limits_{l=2}^{M} \frac{\exp (\langle \bm{q}_{n, l}^{(t)}, \bm{k}_{n, 2}^{(t)} \rangle)}{\exp (\langle \bm{q}_{n, l}^{(t)}, \bm{k}_+^{(t)} \rangle) + \sum\limits_{j=2}^M \exp ( \langle \bm{q}_{n, l}^{(t)}, \bm{k}_{n, j}^{(t)} \rangle )} \big) \cdot V_{n, 2}^{(t)} \\
    &+ \sum\limits_{i=2}^{M} \big( \frac{ \exp (\langle \bm{q}_+^{(t)}, \bm{k}_{n, i}^{(t)} \rangle)}{ \exp  (\langle \bm{q}_+^{(t)}, \bm{k}_+^{(t)} \rangle) + \sum\limits_{j=2}^M  \exp  ( \langle \bm{q}_+^{(t)}, \bm{k}_{n, j}^{(t)} \rangle ) } \\
    &+ \sum\limits_{l=2}^{M} \frac{\exp (\langle \bm{q}_{n, l}^{(t)}, \bm{k}_{n, i}^{(t)} \rangle)}{\exp (\langle \bm{q}_{n, l}^{(t)}, \bm{k}_+^{(t)} \rangle) + \sum\limits_{j=2}^M \exp ( \langle \bm{q}_{n, l}^{(t)}, \bm{k}_{n, j}^{(t)} \rangle )} \big) \cdot V_{n, i}^{(t)} \Big) \\
    &\ge \frac{1}{M} \cdot \Big( 1 - O \Big( \frac{ \sigma_p^2 d \big( \log (6N^2M^2 / \delta) \big)^{3} }{ \Vert \bm{\mu} \Vert_2^2 \Vert \bm{w}_O \Vert_2^2 d_h^{\frac{1}{2}} } \Big) \\
    &+ (M - 1) \cdot \Big( 1 - O \Big( \frac{ \big( \log (6N^2M^2 / \delta) \big)^{3} }{ \Vert \bm{w}_O \Vert_2^2 d_h^{\frac{1}{2}} } \Big) \Big) \Big) \cdot \log \Big( \frac{C_{20}}{\epsilon} \Big) \\
    &- \frac{1}{M} \cdot (M - 1) \cdot \Big( O \Big( \frac{ \sigma_p^2 d \big( \log (6N^2M^2 / \delta) \big)^{3} }{ \Vert \bm{\mu} \Vert_2^2 \Vert \bm{w}_O \Vert_2^2 d_h^{\frac{1}{2}} } \Big) \\
    &+ (M - 1) \cdot O \Big( \frac{ \big( \log (6N^2M^2 / \delta) \big)^{3} }{ \Vert \bm{w}_O \Vert_2^2 d_h^{\frac{1}{2}} } \Big) \Big) \cdot O(1) \\
    &= \log \Big( \frac{C_{20}}{\epsilon} \Big) - O \Big( \frac{ \sigma_p^2 d \big( \log (6N^2M^2 / \delta) \big)^{3} }{ \Vert \bm{\mu} \Vert_2^2 \Vert \bm{w}_O \Vert_2^2 d_h^{\frac{1}{2}} } \Big) \\
    &\ge \log \Big( \frac{C_{20}}{\epsilon} \Big) - \log(C_{20}) \\
    &\ge \log \Big( \frac{1}{\epsilon} \Big).
\end{split}
\end{equation}
For the second inequality, by \(d_h = \widetilde{\Omega} \Big( \max \{\mathrm{SNR}^4, \mathrm{SNR}^{-4}\} N^2 \epsilon^{-2} \Big)\), we have \(O \Big( \frac{ \sigma_p^2 d \big( \log (6N^2M^2 / \delta) \big)^{3} }{ \Vert \bm{\mu} \Vert_2^2 \Vert \bm{w}_O \Vert_2^2 d_h^{\frac{1}{2}} } \Big) = o(1) = \log(1 + o(1)) \le \log(C_{20})\) as long as \(C_{20}\) is sufficiently large.
Then we have
\begin{equation}
\begin{split}
    \ell_n^{(t)} &= \log \Big( 1 + \exp(- f(\bm{X}_n, \theta (t))) \Big) \\
    &\le \exp(- f(\bm{X}_n, \theta (t))) \\
    &\le \exp\Big(- \log \Big( \frac{1}{\epsilon} \Big) \Big) \\
    &\le \epsilon.
\end{split}
\end{equation}
Similarly, we have \(\ell_n^{(t)} \le \epsilon\) for \(n \in S_-\). Therefore, we have \(L_S(\theta(T)) = \frac{1}{N}\sum\limits_{n = 1}^{N} \ell_n^{(t)} \le \epsilon\).

\subsection{Population Loss}

\begin{lemma}[Population Loss]
    \label{population_loss_harmful}
    Under the same conditions as Theorem \ref{harmful}, let \(T_3 = \Theta \Big( \frac{N}{\eta \epsilon \sigma_p^2 d \Vert \bm{w}_O \Vert_2^2} \Big)\), we have
    \[
        L_D \big( \theta(T_3) \big) \ge 0.1
    \]
\end{lemma}
\it{Proof of Lemma} \ref{population_loss_harmful}. \rm Similar to \eqref{VO_dy}
\begin{equation}
\begin{split}
    \Vert \bm{W}_V^{(t+1)} \bm{w}_O - \bm{W}_V^{(t)} \bm{w}_O \Vert_2 &\le \frac{\eta}{NM} \sum\limits_{n=1}^N \big\vert y_n \ell_n^{\prime(t)} \big\vert \sum\limits_{l=1}^M \big\Vert \bm{X}_n \big\Vert_F \big\Vert \bm{\varphi}(\bm{x}_{n, l} \bm{W}_Q^{(t)} \bm{W}_K^{(t)\top} (\bm{X}_n)^\top) \big\Vert_2 \Vert \bm{w}_O \Vert_2^2 \\
    &\le \frac{\eta}{NM} \cdot NM \cdot O \big( \max\{\Vert \bm{\mu} \Vert_2, \sigma_p \sqrt{d}\} \big) \cdot O(1) \cdot \Vert \bm{w}_O \Vert_2^2 \\
    &= O \Big( \eta \cdot \sigma_p \sqrt{d} \cdot \Vert \bm{w}_O \Vert_2^2 \Big).
\end{split}
\end{equation}
Taking a summation, we have
\begin{equation}
\begin{split}
    \Vert \bm{W}_V^{(T_3)} \bm{w}_O \Vert_2 &\le \Vert \bm{W}_V^{(0)} \bm{w}_O \Vert_2 + \sum\limits_{t = 0}^{T_3 - 1} \Vert \bm{W}_V^{(t+1)} \bm{w}_O - \bm{W}_V^{(t)} \bm{w}_O \Vert_2 \\
    &= \Vert \bm{W}_V^{(0)} \bm{w}_O \Vert_2 + O \big( \frac{N}{\eta \epsilon \sigma_p^2 d \Vert \bm{w}_O \Vert_2^2} \big) \cdot O \Big( \eta \cdot \sigma_p \sqrt{d} \cdot \Vert \bm{w}_O \Vert_2^2 \Big) \\
    &= O \Big( \sigma_V \Vert \bm{w}_O \Vert_2 \sqrt{d} \Big) + O \Big( \frac{N}{\epsilon \sigma_p \sqrt{d}} \Big).
\end{split}
\end{equation}
Then \(\bm{\xi}_i^{\top} \bm{W}_V^{(T_3)} \bm{w}_O\) is a Gaussian random variable with mean zero and standard deviation smaller than \(O \Big( \sigma_V \Vert \bm{w}_O \Vert_2 \sigma_p \sqrt{d} + \frac{N}{\epsilon \sqrt{d}} \Big)\).
By Gaussian tail bound, for any \(i \in [M] \backslash \{1\}\), with probability at least \(1 - 1 / 2 M\),
\begin{equation*}
\begin{split}
    \vert \bm{\xi}_i^{\top} \bm{W}_V^{(T_3)} \bm{w}_O \vert &\le O \Big( \sigma_V \Vert \bm{w}_O \Vert_2 \sigma_p \sqrt{d} + \frac{N}{\epsilon \sqrt{d}} \Big) \cdot \sqrt{2 \log \big(4M)} \le 1/2,
\end{split}
\end{equation*}
where the last inequality is by \(\sigma_V \le \widetilde{O} \big( \Vert \bm{w}_O \Vert_2^{-1} \cdot \min \{ \Vert \bm{\mu} \Vert_2^{-1}, ( \sigma_p \sqrt{d} )^{-1} \} \cdot d_h^{-\frac{1}{4}} \big)\) and \(d = \widetilde{\Omega} \Big( \epsilon^{-2} N^2 d_h \Big) \). Applying a union bound, with probability at least \(1 - 1 / 2\), \(\vert \bm{\xi}_i^{\top} \bm{W}_V^{(T_3)} \bm{w}_O \vert \le 1/2\).
Recall that \(V_\pm^{(T_3)} = o(1)\), with probability at least \(1 - 1 / 2\), we have
\begin{equation}
\begin{split}
    y (f(\bm{X}, \theta (T_3))) &= \frac{1}{M} \sum\limits_{l=1}^M \bm{\varphi} (\bm{x}_l^{\top} \bm{W}_{Q}^{(T_3)} \bm{W}_{K}^{(T_3)\top} \bm{X}^\top ) \bm{X} \bm{W}_V^{(T_3)} \bm{w}_O \\
    &\ge \log (1 + e^{-1/2}).
\end{split}
\end{equation}
Thus, \(L_{D}(\theta(t)) \ge \log (1 + e^{-1/2}) \cdot 0.5 \ge 0.1\).

\section{Complete Calculation Process For Benign Overfitting}
\label{specific_calculation_process}
In this section, we show more calculation process under benign overfitting regime. The calculactions for harmful overfitting is similar.

\subsection{Calculactions for \(\alpha\) and \(\beta\)}
\label{alpha_and_beta}
In this subsection, we give the calculactions for \(\alpha\) and \(\beta\) defined in Definition \ref{def3}.

\textbf{Restatement of Lemma \ref{A1}.} the gradients of loss function \(L_S(\theta)\) with respect to \(\bm{W}_Q, \bm{W}_K\) and \(\bm{W}_V\) are given by
\begin{equation}
\begin{split}
    \label{nabla_Q}
    \nabla_{\bm{W}_Q}L_S(\theta)
    &= \frac{1}{NM} \sum\limits_{n \in S_+} \ell_n^{\prime}(\theta) ( \bm{\mu}_+ \bm{w}_O^\top \bm{W}_V^\top \bm{X}_n^\top (diag(\bm{\varphi}_{n, 1}) - \bm{\varphi}_{n, 1}^\top \bm{\varphi}_{n, 1}) \\
    &+ \sum\limits_{i=2}^M \bm{\xi}_{n, i} \bm{w}_O^\top \bm{W}_V^\top \bm{X}_n^\top (diag(\bm{\varphi}_{n, i}) - \bm{\varphi}_{n, i}^\top \bm{\varphi}_{n, i}) ) \bm{X}_n \bm{W}_K \\
    &- \frac{1}{NM} \sum\limits_{n \in S_-} \ell_n^{\prime}(\theta) ( \bm{\mu}_- \bm{w}_O^\top \bm{W}_V^\top \bm{X}_n^\top (diag(\bm{\varphi}_{n, 1}) - \bm{\varphi}_{n, 1}^\top \bm{\varphi}_{n, 1}) \\
    &+ \sum\limits_{i=2}^M \bm{\xi}_{n, i} \bm{w}_O^\top \bm{W}_V^\top \bm{X}_n^\top (diag(\bm{\varphi}_{n, i}) - \bm{\varphi}_{n, i}^\top \bm{\varphi}_{n, i}) ) \bm{X}_n \bm{W}_K ,
\end{split}
\end{equation}

\begin{equation}
\begin{split}
    \label{nabla_K}
    \nabla_{\bm{W}_K}L_S(\theta)
    &= \frac{1}{NM} \sum\limits_{n \in S_+} \ell_n^{\prime}(\theta) ( \bm{X}_n^\top (diag(\bm{\varphi}_{n, 1}) - \bm{\varphi}_{n, 1}^\top \bm{\varphi}_{n, 1}) \bm{X}_n \bm{W}_V \bm{w}_O \bm{\mu}_+^\top \\
    &+ \sum\limits_{i=2}^M \bm{X}_n^\top (diag(\bm{\varphi}_{n, i}) - \bm{\varphi}_{n, i}^\top \bm{\varphi}_{n, i}) \bm{X}_n \bm{W}_V \bm{w}_O \bm{\xi}_{n, i}^\top ) \bm{W}_Q \\
    &- \frac{1}{NM} \sum\limits_{n \in S_-} \ell_n^{\prime}(\theta) ( \bm{X}_n^\top (diag(\bm{\varphi}_{n, 1}) - \bm{\varphi}_{n, 1}^\top \bm{\varphi}_{n, 1}) \bm{X}_n \bm{W}_V \bm{w}_O \bm{\mu}_-^\top \\
    &+ \sum\limits_{i=2}^M \bm{X}_n^\top (diag(\bm{\varphi}_{n, i}) - \bm{\varphi}_{n, i}^\top \bm{\varphi}_{n, i}) \bm{X}_n \bm{W}_V \bm{w}_O \bm{\xi}_{n, i}^\top ) \bm{W}_Q .
\end{split}
\end{equation}
We will give expressions for \(\alpha\) and \(\beta\) defined in Definition \ref{def3} based on these two equations above.

By \eqref{nabla_Q} and the orthogonal relation between \(\bm{\mu}\) and \(\bm{\xi}\), we have
\begin{equation}
\begin{split}
    \Delta \bm{q}_+^{(t)} = \bm{\mu}_+^\top \Delta\bm{W}_Q^{(t)}
    &= \frac{\eta}{NM} \sum\limits_{n \in S_+} - \ell_n^{\prime(t)} \Vert \bm{\mu} \Vert_2^2 \bm{w}_O^\top \bm{W}_V^{(t)\top} \bm{X}_n^\top (diag(\bm{\varphi}_{n, 1}) - \bm{\varphi}_{n, 1}^\top \bm{\varphi}_{n, 1}) \bm{X}_n \bm{W}_K^{(t)} \\
    &= \alpha_{+, +}^{(t)} \bm{k}_+^{(t)} + \sum\limits_{n \in S_+} \sum\limits_{i=2}^M \alpha_{n, +, i}^{(t)} \bm{k}_{n, i}^{(t)}
\end{split}
\end{equation}
where \(\bm{w}_O^\top \bm{W}_V^{(t)\top} \bm{X}_n^\top\) and \( \bm{X}_n \bm{W}_K^{(t)} \) can be viewed in the following forms
\[
    \bm{w}_O^\top \bm{W}_V^{(t)\top} \bm{X}_n^\top = \big( V_+^{(t)}, V_{n, 2}^{(t)}, \dots, V_{n, M}^{(t)} \big),
\]
\[
    \bm{X}_n \bm{W}_K^{(t)} = \big( \bm{k}_+^{(t)\top}, \bm{k}_{n, 2}^{(t)\top}, \dots, \bm{k}_{n, M}^{(t)\top} \big)^\top.
\]
Then we can express \( \alpha_{+, +}^{(t)} \) and \( \alpha_{n, +, i}^{(t)} \) as follows
\begin{equation}
\begin{split}
    &\alpha_{+, +}^{(t)} = \frac{\eta}{NM} \sum\limits_{n \in S_+} - \ell_n^{\prime(t)} \Vert \bm{\mu} \Vert_2^2 \\
    &\cdot \Big( V_+^{(t)} \big( \frac{ \exp (\langle \bm{q}_+^{(t)}, \bm{k}_+^{(t)} \rangle)}{ \exp  (\langle \bm{q}_+^{(t)}, \bm{k}_+^{(t)} \rangle) + \sum\limits_{j=2}^M  \exp  ( \langle \bm{q}_+^{(t)}, \bm{k}_{n, j}^{(t)} \rangle ) } \\
    & - (\frac{ \exp (\langle \bm{q}_+^{(t)}, \bm{k}_+^{(t)} \rangle)}{ \exp  (\langle \bm{q}_+^{(t)}, \bm{k}_+^{(t)} \rangle) + \sum\limits_{j=2}^M  \exp  ( \langle \bm{q}_+^{(t)}, \bm{k}_{n, j}^{(t)} \rangle ) })^2 \big) \\
    & - \sum\limits_{i=2}^M \big( V_{n, i}^{(t)} \cdot \frac{ \exp (\langle \bm{q}_+^{(t)}, \bm{k}_+^{(t)} \rangle)}{ \exp  (\langle \bm{q}_+^{(t)}, \bm{k}_+^{(t)} \rangle) + \sum\limits_{j=2}^M  \exp  ( \langle \bm{q}_+^{(t)}, \bm{k}_{n, j}^{(t)} \rangle ) } \\
    & \cdot \frac{ \exp (\langle \bm{q}_+^{(t)}, \bm{k}_{n, i}^{(t)} \rangle)}{ \exp  (\langle \bm{q}_+^{(t)}, \bm{k}_+^{(t)} \rangle) + \sum\limits_{j=2}^M  \exp  ( \langle \bm{q}_+^{(t)}, \bm{k}_{n, j}^{(t)} \rangle ) } \big) \Big),
\end{split}
\end{equation}

\begin{equation}
\begin{split}
    &\alpha_{n, +, i}^{(t)} = - \frac{\eta}{NM} \ell_n^{\prime(t)} \Vert \bm{\mu} \Vert_2^2 \\
    &\cdot \Big( - V_+^{(t)} \cdot \frac{ \exp (\langle \bm{q}_+^{(t)}, \bm{k}_+^{(t)} \rangle)}{ \exp  (\langle \bm{q}_+^{(t)}, \bm{k}_+^{(t)} \rangle) + \sum\limits_{j=2}^M  \exp  ( \langle \bm{q}_+^{(t)}, \bm{k}_{n, j}^{(t)} \rangle ) } \\
    &\cdot \frac{ \exp (\langle \bm{q}_+^{(t)}, \bm{k}_{n, i}^{(t)} \rangle)}{ \exp  (\langle \bm{q}_+^{(t)}, \bm{k}_+^{(t)} \rangle) + \sum\limits_{j=2}^M  \exp  ( \langle \bm{q}_+^{(t)}, \bm{k}_{n, j}^{(t)} \rangle ) } \\
    &+ V_{n, i}^{(t)} \big( \frac{ \exp (\langle \bm{q}_+^{(t)}, \bm{k}_{n, i}^{(t)} \rangle)}{ \exp  (\langle \bm{q}_+^{(t)}, \bm{k}_+^{(t)} \rangle) + \sum\limits_{j=2}^M  \exp  ( \langle \bm{q}_+^{(t)}, \bm{k}_{n, j}^{(t)} \rangle ) } \\
    &- (\frac{ \exp (\langle \bm{q}_+^{(t)}, \bm{k}_{n, i}^{(t)} \rangle)}{ \exp  (\langle \bm{q}_+^{(t)}, \bm{k}_+^{(t)} \rangle) + \sum\limits_{j=2}^M  \exp  ( \langle \bm{q}_+^{(t)}, \bm{k}_{n, j}^{(t)} \rangle ) })^2 \big) \\
    &- \sum\limits_{k \ne i} \big( V_{n, k}^{(t)} \cdot \frac{ \exp (\langle \bm{q}_+^{(t)}, \bm{k}_{n, i}^{(t)} \rangle)}{ \exp  (\langle \bm{q}_+^{(t)}, \bm{k}_+^{(t)} \rangle) + \sum\limits_{j=2}^M  \exp  ( \langle \bm{q}_+^{(t)}, \bm{k}_{n, j}^{(t)} \rangle ) } \\
    & \cdot \frac{ \exp (\langle \bm{q}_+^{(t)}, \bm{k}_{n, k}^{(t)} \rangle)}{ \exp  (\langle \bm{q}_+^{(t)}, \bm{k}_+^{(t)} \rangle) + \sum\limits_{j=2}^M  \exp  ( \langle \bm{q}_+^{(t)}, \bm{k}_{n, j}^{(t)} \rangle ) } \big) \Big).
\end{split}
\end{equation}

Using the similar method as for \( \Delta \bm{q}_+^{(t)} \), we get the other \(\alpha\) and \(\beta\) as follows

\begin{equation}
\begin{split}
    &\alpha_{-, -}^{(t)} = \frac{\eta}{NM} \sum\limits_{n \in S_-} \ell_n^{\prime(t)} \Vert \bm{\mu} \Vert_2^2 \\
    &\cdot \Big( V_-^{(t)} \big( \frac{ \exp (\langle \bm{q}_-^{(t)}, \bm{k}_-^{(t)} \rangle)}{ \exp  (\langle \bm{q}_-^{(t)}, \bm{k}_-^{(t)} \rangle) + \sum\limits_{j=2}^M  \exp  ( \langle \bm{q}_-^{(t)}, \bm{k}_{n, j}^{(t)} \rangle ) } \\
    & - (\frac{ \exp (\langle \bm{q}_-^{(t)}, \bm{k}_-^{(t)} \rangle)}{ \exp  (\langle \bm{q}_-^{(t)}, \bm{k}_-^{(t)} \rangle) + \sum\limits_{j=2}^M  \exp  ( \langle \bm{q}_-^{(t)}, \bm{k}_{n, j}^{(t)} \rangle ) })^2 \big) \\
    & - \sum\limits_{i=2}^M \big( V_{n, i}^{(t)} \cdot \frac{ \exp (\langle \bm{q}_-^{(t)}, \bm{k}_-^{(t)} \rangle)}{ \exp  (\langle \bm{q}_-^{(t)}, \bm{k}_-^{(t)} \rangle) + \sum\limits_{j=2}^M  \exp  ( \langle \bm{q}_-^{(t)}, \bm{k}_{n, j}^{(t)} \rangle ) } \\
    & \cdot \frac{ \exp (\langle \bm{q}_-^{(t)}, \bm{k}_{n, i}^{(t)} \rangle)}{ \exp  (\langle \bm{q}_-^{(t)}, \bm{k}_-^{(t)} \rangle) + \sum\limits_{j=2}^M  \exp  ( \langle \bm{q}_-^{(t)}, \bm{k}_{n, j}^{(t)} \rangle ) } \big) \Big),
\end{split}
\end{equation}

\begin{equation}
\begin{split}
    &\alpha_{n, -, i}^{(t)} = \frac{\eta}{NM} \ell_n^{\prime(t)} \Vert \bm{\mu} \Vert_2^2 \\
    &\cdot \Big( - V_-^{(t)} \cdot \frac{ \exp (\langle \bm{q}_-^{(t)}, \bm{k}_-^{(t)} \rangle)}{ \exp  (\langle \bm{q}_-^{(t)}, \bm{k}_-^{(t)} \rangle) + \sum\limits_{j=2}^M  \exp  ( \langle \bm{q}_-^{(t)}, \bm{k}_{n, j}^{(t)} \rangle ) } \\
    &\cdot \frac{ \exp (\langle \bm{q}_-^{(t)}, \bm{k}_{n, i}^{(t)} \rangle)}{ \exp  (\langle \bm{q}_-^{(t)}, \bm{k}_-^{(t)} \rangle) + \sum\limits_{j=2}^M  \exp  ( \langle \bm{q}_-^{(t)}, \bm{k}_{n, j}^{(t)} \rangle ) } \\
    &+ V_{n, i}^{(t)} \big( \frac{ \exp (\langle \bm{q}_-^{(t)}, \bm{k}_{n, i}^{(t)} \rangle)}{ \exp  (\langle \bm{q}_-^{(t)}, \bm{k}_-^{(t)} \rangle) + \sum\limits_{j=2}^M  \exp  ( \langle \bm{q}_-^{(t)}, \bm{k}_{n, j}^{(t)} \rangle ) } \\
    &- (\frac{ \exp (\langle \bm{q}_-^{(t)}, \bm{k}_{n, i}^{(t)} \rangle)}{ \exp  (\langle \bm{q}_-^{(t)}, \bm{k}_-^{(t)} \rangle) + \sum\limits_{j=2}^M  \exp  ( \langle \bm{q}_-^{(t)}, \bm{k}_{n, j}^{(t)} \rangle ) })^2 \big) \\
    &- \sum\limits_{k \ne i} \big( V_{n, k}^{(t)} \cdot \frac{ \exp (\langle \bm{q}_-^{(t)}, \bm{k}_{n, i}^{(t)} \rangle)}{ \exp  (\langle \bm{q}_-^{(t)}, \bm{k}_-^{(t)} \rangle) + \sum\limits_{j=2}^M  \exp  ( \langle \bm{q}_-^{(t)}, \bm{k}_{n, j}^{(t)} \rangle ) } \\
    & \cdot \frac{ \exp (\langle \bm{q}_-^{(t)}, \bm{k}_{n, k}^{(t)} \rangle)}{ \exp  (\langle \bm{q}_-^{(t)}, \bm{k}_-^{(t)} \rangle) + \sum\limits_{j=2}^M  \exp  ( \langle \bm{q}_-^{(t)}, \bm{k}_{n, j}^{(t)} \rangle ) } \big) \Big),
\end{split}
\end{equation}

\begin{equation}
\begin{split}
    &\alpha_{n^\prime, i^\prime, +}^{(t)} = \frac{\eta}{NM} \sum\limits_{n \in S_+} - \ell_n^{\prime(t)} \sum\limits_{i=2}^M \langle \bm{\xi}_{n^\prime, i^\prime}, \bm{\xi}_{n, i} \rangle \\
    &\cdot \Big( V_+^{(t)} \big( \frac{ \exp (\langle \bm{q}_{n, i}^{(t)}, \bm{k}_+^{(t)} \rangle)}{ \exp  (\langle \bm{q}_{n, i}^{(t)}, \bm{k}_+^{(t)} \rangle) + \sum\limits_{j=2}^M  \exp  ( \langle \bm{q}_{n, i}^{(t)}, \bm{k}_{n, j}^{(t)} \rangle ) } \\
    & - (\frac{ \exp (\langle \bm{q}_{n, i}^{(t)}, \bm{k}_+^{(t)} \rangle)}{ \exp  (\langle \bm{q}_{n, i}^{(t)}, \bm{k}_+^{(t)} \rangle) + \sum\limits_{j=2}^M  \exp  ( \langle \bm{q}_{n, i}^{(t)}, \bm{k}_{n, j}^{(t)} \rangle ) })^2 \big) \\
    & -\sum\limits_{k=2}^M \big( V_{n, i}^{(t)} \cdot \frac{ \exp (\langle \bm{q}_{n, i}^{(t)}, \bm{k}_+^{(t)} \rangle)}{ \exp  (\langle \bm{q}_{n, i}^{(t)}, \bm{k}_+^{(t)} \rangle) + \sum\limits_{j=2}^M  \exp  ( \langle \bm{q}_{n, i}^{(t)}, \bm{k}_{n, j}^{(t)} \rangle ) } \\
    & \cdot \frac{ \exp (\langle \bm{q}_{n, i}^{(t)}, \bm{k}_{n, k}^{(t)} \rangle)}{ \exp  (\langle \bm{q}_{n, i}^{(t)}, \bm{k}_+^{(t)} \rangle) + \sum\limits_{j=2}^M  \exp  ( \langle \bm{q}_{n, i}^{(t)}, \bm{k}_{n, j}^{(t)} \rangle ) } \big) \Big),
\end{split}
\end{equation}

\begin{equation}
\begin{split}
    &\alpha_{n^\prime, i^\prime, -}^{(t)} = \frac{\eta}{NM} \sum\limits_{n \in S_-} \ell_n^{\prime(t)} \sum\limits_{i=2}^M \langle \bm{\xi}_{n^\prime, i^\prime}, \bm{\xi}_{n, i} \rangle \\
    &\cdot \Big( V_-^{(t)} \big( \frac{ \exp (\langle \bm{q}_{n, i}^{(t)}, \bm{k}_-^{(t)} \rangle)}{ \exp  (\langle \bm{q}_{n, i}^{(t)}, \bm{k}_-^{(t)} \rangle) + \sum\limits_{j=2}^M  \exp  ( \langle \bm{q}_{n, i}^{(t)}, \bm{k}_{n, j}^{(t)} \rangle ) } \\
    & - (\frac{ \exp (\langle \bm{q}_{n, i}^{(t)}, \bm{k}_-^{(t)} \rangle)}{ \exp  (\langle \bm{q}_{n, i}^{(t)}, \bm{k}_-^{(t)} \rangle) + \sum\limits_{j=2}^M  \exp  ( \langle \bm{q}_{n, i}^{(t)}, \bm{k}_{n, j}^{(t)} \rangle ) })^2 \big) \\
    & -\sum\limits_{k=2}^M \big( V_{n, i}^{(t)} \cdot \frac{ \exp (\langle \bm{q}_{n, i}^{(t)}, \bm{k}_-^{(t)} \rangle)}{ \exp  (\langle \bm{q}_{n, i}^{(t)}, \bm{k}_-^{(t)} \rangle) + \sum\limits_{j=2}^M  \exp  ( \langle \bm{q}_{n, i}^{(t)}, \bm{k}_{n, j}^{(t)} \rangle ) } \\
    & \cdot \frac{ \exp (\langle \bm{q}_{n, i}^{(t)}, \bm{k}_{n, k}^{(t)} \rangle)}{ \exp  (\langle \bm{q}_{n, i}^{(t)}, \bm{k}_-^{(t)} \rangle) + \sum\limits_{j=2}^M  \exp  ( \langle \bm{q}_{n, i}^{(t)}, \bm{k}_{n, j}^{(t)} \rangle ) } \big) \Big),
\end{split}
\end{equation}

\begin{equation}
\begin{split}
    &\alpha_{n^\prime, i^\prime, n, i}^{(t)} = \frac{\eta}{NM} - \ell_n^{\prime(t)} \sum\limits_{k=2}^M \langle \bm{\xi}_{n^\prime, i^\prime}, \bm{\xi}_{n, k} \rangle \\
    &\cdot \Big( - V_+^{(t)} \cdot \frac{ \exp (\langle \bm{q}_{n, k}^{(t)}, \bm{k}_+^{(t)} \rangle)}{ \exp (\langle \bm{q}_{n, k}^{(t)}, \bm{k}_+^{(t)} \rangle) + \sum\limits_{j=2}^M \exp ( \langle \bm{q}_{n, k}^{(t)}, \bm{k}_{n, j}^{(t)} \rangle ) } \\
    &\cdot \frac{ \exp (\langle \bm{q}_{n, k}^{(t)}, \bm{k}_{n, i}^{(t)} \rangle)}{ \exp  (\langle \bm{q}_{n, k}^{(t)}, \bm{k}_+^{(t)} \rangle) + \sum\limits_{j=2}^M  \exp  ( \langle \bm{q}_{n, k}^{(t)}, \bm{k}_{n, j}^{(t)} \rangle ) } \\
    &+V_{n, i}^{(t)} \big( \frac{ \exp (\langle \bm{q}_{n, k}^{(t)}, \bm{k}_{n, i}^{(t)} \rangle)}{ \exp  (\langle \bm{q}_{n, k}^{(t)}, \bm{k}_+^{(t)} \rangle) + \sum\limits_{j=2}^M  \exp  ( \langle \bm{q}_{n, k}^{(t)}, \bm{k}_{n, j}^{(t)} \rangle ) } \\
    &- (\frac{ \exp (\langle \bm{q}_{n, k}^{(t)}, \bm{k}_{n, i}^{(t)} \rangle)}{ \exp  (\langle \bm{q}_{n, k}^{(t)}, \bm{k}_+^{(t)} \rangle) + \sum\limits_{j=2}^M  \exp  ( \langle \bm{q}_{n, k}^{(t)}, \bm{k}_{n, j}^{(t)} \rangle ) })^2 \big) \\
    &- \sum\limits_{l \ne i} \big( V_{n, l}^{(t)} \cdot \frac{ \exp (\langle \bm{q}_{n, k}^{(t)}, \bm{k}_{n, i}^{(t)} \rangle)}{ \exp  (\langle \bm{q}_{n, k}^{(t)}, \bm{k}_+^{(t)} \rangle) + \sum\limits_{j=2}^M  \exp  ( \langle \bm{q}_{n, k}^{(t)}, \bm{k}_{n, j}^{(t)} \rangle ) } \\
    & \cdot \frac{ \exp (\langle \bm{q}_{n, k}^{(t)}, \bm{k}_{n, l}^{(t)} \rangle)}{ \exp  (\langle \bm{q}_{n, k}^{(t)}, \bm{k}_+^{(t)} \rangle) + \sum\limits_{j=2}^M  \exp  ( \langle \bm{q}_{n, k}^{(t)}, \bm{k}_{n, j}^{(t)} \rangle ) } \big) \Big)
\end{split}
\end{equation}
for \(n \in S_+\),

\begin{equation}
\begin{split}
    &\alpha_{n^\prime, i^\prime, n, i}^{(t)} = \frac{\eta}{NM} \ell_n^{\prime(t)} \sum\limits_{k=2}^M \langle \bm{\xi}_{n^\prime, i^\prime}, \bm{\xi}_{n, k} \rangle \\
    &\cdot \Big( - V_-^{(t)} \cdot \frac{ \exp (\langle \bm{q}_{n, k}^{(t)}, \bm{k}_-^{(t)} \rangle)}{ \exp (\langle \bm{q}_{n, k}^{(t)}, \bm{k}_-^{(t)} \rangle) + \sum\limits_{j=2}^M \exp ( \langle \bm{q}_{n, k}^{(t)}, \bm{k}_{n, j}^{(t)} \rangle ) } \\
    &\cdot \frac{ \exp (\langle \bm{q}_{n, k}^{(t)}, \bm{k}_{n, i}^{(t)} \rangle)}{ \exp  (\langle \bm{q}_{n, k}^{(t)}, \bm{k}_-^{(t)} \rangle) + \sum\limits_{j=2}^M  \exp  ( \langle \bm{q}_{n, k}^{(t)}, \bm{k}_{n, j}^{(t)} \rangle ) } \\
    &+V_{n, i}^{(t)} \big( \frac{ \exp (\langle \bm{q}_{n, k}^{(t)}, \bm{k}_{n, i}^{(t)} \rangle)}{ \exp  (\langle \bm{q}_{n, k}^{(t)}, \bm{k}_-^{(t)} \rangle) + \sum\limits_{j=2}^M  \exp  ( \langle \bm{q}_{n, k}^{(t)}, \bm{k}_{n, j}^{(t)} \rangle ) } \\
    &- (\frac{ \exp (\langle \bm{q}_{n, k}^{(t)}, \bm{k}_{n, i}^{(t)} \rangle)}{ \exp  (\langle \bm{q}_{n, k}^{(t)}, \bm{k}_-^{(t)} \rangle) + \sum\limits_{j=2}^M  \exp  ( \langle \bm{q}_{n, k}^{(t)}, \bm{k}_{n, j}^{(t)} \rangle ) })^2 \big) \\
    &- \sum\limits_{l \ne i} \big( V_{n, l}^{(t)} \cdot \frac{ \exp (\langle \bm{q}_{n, k}^{(t)}, \bm{k}_{n, i}^{(t)} \rangle)}{ \exp  (\langle \bm{q}_{n, k}^{(t)}, \bm{k}_-^{(t)} \rangle) + \sum\limits_{j=2}^M  \exp  ( \langle \bm{q}_{n, k}^{(t)}, \bm{k}_{n, j}^{(t)} \rangle ) } \\
    & \cdot \frac{ \exp (\langle \bm{q}_{n, k}^{(t)}, \bm{k}_{n, l}^{(t)} \rangle)}{ \exp  (\langle \bm{q}_{n, k}^{(t)}, \bm{k}_-^{(t)} \rangle) + \sum\limits_{j=2}^M  \exp  ( \langle \bm{q}_{n, k}^{(t)}, \bm{k}_{n, j}^{(t)} \rangle ) } \big) \Big)
\end{split}
\end{equation}
for \(n \in S_-\),

\begin{equation}
\begin{split}
    &\beta_{-, -}^{(t)} = \frac{\eta \Vert \bm{\mu} \Vert_2^2}{NM} \sum\limits_{n \in S_-} \ell_n^{\prime(t)} \\
    &\cdot \Big( V_-^{(t)} \big( \frac{ \exp (\langle \bm{q}_-^{(t)}, \bm{k}_-^{(t)} \rangle)}{ \exp  (\langle \bm{q}_-^{(t)}, \bm{k}_-^{(t)} \rangle) + \sum\limits_{j=2}^M  \exp  ( \langle \bm{q}_-^{(t)}, \bm{k}_{n, j}^{(t)} \rangle ) } \\
    & - (\frac{ \exp (\langle \bm{q}_-^{(t)}, \bm{k}_-^{(t)} \rangle)}{ \exp  (\langle \bm{q}_-^{(t)}, \bm{k}_-^{(t)} \rangle) + \sum\limits_{j=2}^M  \exp  ( \langle \bm{q}_-^{(t)}, \bm{k}_{n, j}^{(t)} \rangle ) })^2 \big) \\
    & -\sum\limits_{i=2}^M \big( V_{n, i}^{(t)} \cdot \frac{ \exp (\langle \bm{q}_-^{(t)}, \bm{k}_-^{(t)} \rangle)}{ \exp  (\langle \bm{q}_-^{(t)}, \bm{k}_-^{(t)} \rangle) + \sum\limits_{j=2}^M  \exp  ( \langle \bm{q}_-^{(t)}, \bm{k}_{n, j}^{(t)} \rangle ) } \\
    & \cdot \frac{ \exp (\langle \bm{q}_-^{(t)}, \bm{k}_{n, i}^{(t)} \rangle)}{ \exp  (\langle \bm{q}_-^{(t)}, \bm{k}_-^{(t)} \rangle) + \sum\limits_{j=2}^M  \exp  ( \langle \bm{q}_-^{(t)}, \bm{k}_{n, j}^{(t)} \rangle ) } \big) \Big),
\end{split}
\end{equation}

\begin{equation}
\begin{split}
    &\beta_{n, +, i}^{(t)} = - \frac{\eta \Vert \bm{\mu} \Vert_2^2}{NM} \ell_n^{\prime(t)} \\
    &\cdot \Big( V_+^{(t)} \big( \frac{ \exp (\langle \bm{q}_{n, i}^{(t)}, \bm{k}_+^{(t)} \rangle)}{ \exp  (\langle \bm{q}_{n, i}^{(t)}, \bm{k}_+^{(t)} \rangle) + \sum\limits_{j=2}^M  \exp  ( \langle \bm{q}_{n, i}^{(t)}, \bm{k}_{n, j}^{(t)} \rangle ) } \\
    & - (\frac{ \exp (\langle \bm{q}_{n, i}^{(t)}, \bm{k}_+^{(t)} \rangle)}{ \exp  (\langle \bm{q}_{n, i}^{(t)}, \bm{k}_+^{(t)} \rangle) + \sum\limits_{j=2}^M  \exp  ( \langle \bm{q}_{n, i}^{(t)}, \bm{k}_{n, j}^{(t)} \rangle ) })^2 \big) \\
    & -\sum\limits_{k=2}^M \big( V_{n, i}^{(t)} \cdot \frac{ \exp (\langle \bm{q}_{n, i}^{(t)}, \bm{k}_+^{(t)} \rangle)}{ \exp  (\langle \bm{q}_{n, i}^{(t)}, \bm{k}_+^{(t)} \rangle) + \sum\limits_{j=2}^M  \exp  ( \langle \bm{q}_{n, i}^{(t)}, \bm{k}_{n, j}^{(t)} \rangle ) } \\
    & \cdot \frac{ \exp (\langle \bm{q}_{n, i}^{(t)}, \bm{k}_{n, k}^{(t)} \rangle)}{ \exp  (\langle \bm{q}_{n, i}^{(t)}, \bm{k}_+^{(t)} \rangle) + \sum\limits_{j=2}^M  \exp  ( \langle \bm{q}_{n, i}^{(t)}, \bm{k}_{n, j}^{(t)} \rangle ) } \big) \Big),
\end{split}
\end{equation}

\begin{equation}
\begin{split}
    &\beta_{n, -, i}^{(t)} = \frac{\eta \Vert \bm{\mu} \Vert_2^2}{NM} \ell_n^{\prime(t)} \\
    &\cdot \Big( V_-^{(t)} \big( \frac{ \exp (\langle \bm{q}_{n, i}^{(t)}, \bm{k}_-^{(t)} \rangle)}{ \exp  (\langle \bm{q}_{n, i}^{(t)}, \bm{k}_-^{(t)} \rangle) + \sum\limits_{j=2}^M  \exp  ( \langle \bm{q}_{n, i}^{(t)}, \bm{k}_{n, j}^{(t)} \rangle ) } \\
    & - (\frac{ \exp (\langle \bm{q}_{n, i}^{(t)}, \bm{k}_-^{(t)} \rangle)}{ \exp  (\langle \bm{q}_{n, i}^{(t)}, \bm{k}_-^{(t)} \rangle) + \sum\limits_{j=2}^M  \exp  ( \langle \bm{q}_{n, i}^{(t)}, \bm{k}_{n, j}^{(t)} \rangle ) })^2 \big) \\
    & -\sum\limits_{k=2}^M \big( V_{n, i}^{(t)} \cdot \frac{ \exp (\langle \bm{q}_{n, i}^{(t)}, \bm{k}_-^{(t)} \rangle)}{ \exp  (\langle \bm{q}_{n, i}^{(t)}, \bm{k}_-^{(t)} \rangle) + \sum\limits_{j=2}^M  \exp  ( \langle \bm{q}_{n, i}^{(t)}, \bm{k}_{n, j}^{(t)} \rangle ) } \\
    & \cdot \frac{ \exp (\langle \bm{q}_{n, i}^{(t)}, \bm{k}_{n, k}^{(t)} \rangle)}{ \exp  (\langle \bm{q}_{n, i}^{(t)}, \bm{k}_-^{(t)} \rangle) + \sum\limits_{j=2}^M  \exp  ( \langle \bm{q}_{n, i}^{(t)}, \bm{k}_{n, j}^{(t)} \rangle ) } \big) \Big),
\end{split}
\end{equation}

\begin{equation}
\begin{split}
    &\beta_{n^\prime, i^\prime, +}^{(t)} = \frac{\eta}{NM} \sum\limits_{n \in S_+} - \ell_n^{\prime(t)} \\
    &\Big( - V_+^{(t)} \sum\limits_{i=2}^M \big( \langle \bm{\xi}_{n^\prime, i^\prime}, \bm{\xi}_{n, i} \rangle \frac{\exp (\langle \bm{q}_+^{(t)}, \bm{k}_+^{(t)} \rangle)}{\exp (\langle \bm{q}_+^{(t)}, \bm{k}_+^{(t)} \rangle) + \sum\limits_{j=2}^M \exp ( \langle \bm{q}_+^{(t)}, \bm{k}_{n, j}^{(t)} \rangle )} \\
    &\cdot \frac{\exp (\langle \bm{q}_+^{(t)}, \bm{k}_{n, i}^{(t)} \rangle)}{\exp (\langle \bm{q}_+^{(t)}, \bm{k}_+^{(t)} \rangle) + \sum\limits_{j=2}^M \exp ( \langle \bm{q}_+^{(t)}, \bm{k}_{n, j}^{(t)} \rangle )} \big) \\
    &+ \sum\limits_{k=2}^M V_{n, k}^{(t)} \big( \langle \bm{\xi}_{n^\prime, i^\prime}, \bm{\xi}_{n, k} \rangle \frac{\exp (\langle \bm{q}_+^{(t)}, \bm{k}_{n, k}^{(t)} \rangle)}{\exp (\langle \bm{q}_+^{(t)}, \bm{k}_+^{(t)} \rangle) + \sum\limits_{j=2}^M \exp ( \langle \bm{q}_+^{(t)}, \bm{k}_{n, j}^{(t)} \rangle )}\\
    &- \sum\limits_{i=2}^M \big( \langle \bm{\xi}_{n^\prime, i^\prime}, \bm{\xi}_{n, i} \rangle \frac{\exp (\langle \bm{q}_+^{(t)}, \bm{k}_{n, k}^{(t)} \rangle)}{\exp (\langle \bm{q}_+^{(t)}, \bm{k}_+^{(t)} \rangle) + \sum\limits_{j=2}^M \exp ( \langle \bm{q}_+^{(t)}, \bm{k}_{n, j}^{(t)} \rangle )} \\
    &\cdot \frac{\exp (\langle \bm{q}_+^{(t)}, \bm{k}_{n, i}^{(t)} \rangle)}{\exp (\langle \bm{q}_+^{(t)}, \bm{k}_+^{(t)} \rangle) + \sum\limits_{j=2}^M \exp ( \langle \bm{q}_+^{(t)}, \bm{k}_{n, j}^{(t)} \rangle )} \big) \big) \Big),
\end{split}
\end{equation}

\begin{equation}
\begin{split}
    &\beta_{n^\prime, i^\prime, -}^{(t)} = \frac{\eta}{NM} \sum\limits_{n \in S_-} \ell_n^{\prime(t)} \\
    &\Big( - V_-^{(t)} \sum\limits_{i=2}^M \big( \langle \bm{\xi}_{n^\prime, i^\prime}, \bm{\xi}_{n, i} \rangle \frac{\exp (\langle \bm{q}_-^{(t)}, \bm{k}_-^{(t)} \rangle)}{\exp (\langle \bm{q}_-^{(t)}, \bm{k}_-^{(t)} \rangle) + \sum\limits_{j=2}^M \exp ( \langle \bm{q}_-^{(t)}, \bm{k}_{n, j}^{(t)} \rangle )} \\
    &\cdot \frac{\exp (\langle \bm{q}_-^{(t)}, \bm{k}_{n, i}^{(t)} \rangle)}{\exp (\langle \bm{q}_-^{(t)}, \bm{k}_-^{(t)} \rangle) + \sum\limits_{j=2}^M \exp ( \langle \bm{q}_-^{(t)}, \bm{k}_{n, j}^{(t)} \rangle )} \big) \\
    &+ \sum\limits_{k=2}^M V_{n, k}^{(t)} \big( \langle \bm{\xi}_{n^\prime, i^\prime}, \bm{\xi}_{n, k} \rangle \frac{\exp (\langle \bm{q}_-^{(t)}, \bm{k}_{n, k}^{(t)} \rangle)}{\exp (\langle \bm{q}_-^{(t)}, \bm{k}_-^{(t)} \rangle) + \sum\limits_{j=2}^M \exp ( \langle \bm{q}_-^{(t)}, \bm{k}_{n, j}^{(t)} \rangle )}\\
    &- \sum\limits_{i=2}^M \big( \langle \bm{\xi}_{n^\prime, i^\prime}, \bm{\xi}_{n, i} \rangle \frac{\exp (\langle \bm{q}_-^{(t)}, \bm{k}_{n, k}^{(t)} \rangle)}{\exp (\langle \bm{q}_-^{(t)}, \bm{k}_-^{(t)} \rangle) + \sum\limits_{j=2}^M \exp ( \langle \bm{q}_-^{(t)}, \bm{k}_{n, j}^{(t)} \rangle )} \\
    &\cdot \frac{\exp (\langle \bm{q}_-^{(t)}, \bm{k}_{n, i}^{(t)} \rangle)}{\exp (\langle \bm{q}_-^{(t)}, \bm{k}_-^{(t)} \rangle) + \sum\limits_{j=2}^M \exp ( \langle \bm{q}_-^{(t)}, \bm{k}_{n, j}^{(t)} \rangle )} \big) \big) \Big),
\end{split}
\end{equation}

\begin{equation}
\begin{split}
    &\beta_{n^\prime, i^\prime, n, i}^{(t)} = \frac{\eta}{NM} - \ell_n^{\prime(t)} \\
    &\Big( - V_+^{(t)} \sum\limits_{k=2}^M \big( \langle \bm{\xi}_{n^\prime, i^\prime}, \bm{\xi}_{n, k} \rangle \frac{\exp (\langle \bm{q}_{n, i}^{(t)}, \bm{k}_+^{(t)} \rangle)}{\exp (\langle \bm{q}_{n, i}^{(t)}, \bm{k}_+^{(t)} \rangle) + \sum\limits_{j=2}^M \exp ( \langle \bm{q}_{n, i}^{(t)}, \bm{k}_{n, j}^{(t)} \rangle )} \\
    &\cdot \frac{\exp (\langle \bm{q}_{n, i}^{(t)}, \bm{k}_{n, k}^{(t)} \rangle)}{\exp (\langle \bm{q}_{n, i}^{(t)}, \bm{k}_+^{(t)} \rangle) + \sum\limits_{j=2}^M \exp ( \langle \bm{q}_{n, i}^{(t)}, \bm{k}_{n, j}^{(t)} \rangle )} \big) \\
    &+ \sum\limits_{k=2}^M V_{n, k}^{(t)} \big( \langle \bm{\xi}_{n^\prime, i^\prime}, \bm{\xi}_{n, k} \rangle \frac{\exp (\langle \bm{q}_{n, i}^{(t)}, \bm{k}_{n, k}^{(t)} \rangle)}{\exp (\langle \bm{q}_{n, i}^{(t)}, \bm{k}_+^{(t)} \rangle) + \sum\limits_{j=2}^M \exp ( \langle \bm{q}_{n, i}^{(t)}, \bm{k}_{n, j}^{(t)} \rangle )}\\
    &- \sum\limits_{l=2}^M \big( \langle \bm{\xi}_{n^\prime, i^\prime}, \bm{\xi}_{n, l} \rangle \frac{\exp (\langle \bm{q}_{n, i}^{(t)}, \bm{k}_{n, k}^{(t)} \rangle)}{\exp (\langle \bm{q}_{n, i}^{(t)}, \bm{k}_+^{(t)} \rangle) + \sum\limits_{j=2}^M \exp ( \langle \bm{q}_{n, i}^{(t)}, \bm{k}_{n, j}^{(t)} \rangle )} \\
    &\cdot \frac{\exp (\langle \bm{q}_{n, i}^{(t)}, \bm{k}_{n, l}^{(t)} \rangle)}{\exp (\langle \bm{q}_{n, i}^{(t)}, \bm{k}_+^{(t)} \rangle) + \sum\limits_{j=2}^M \exp ( \langle \bm{q}_{n, i}^{(t)}, \bm{k}_{n, j}^{(t)} \rangle )} \big) \big) \Big)
\end{split}
\end{equation}
for \(n \in S_+\),

\begin{equation}
\begin{split}
    &\beta_{n^\prime, i^\prime, n, i}^{(t)} = \frac{\eta}{NM} \ell_n^{\prime(t)} \\
    &\Big( - V_-^{(t)} \sum\limits_{k=2}^M \big( \langle \bm{\xi}_{n^\prime, i^\prime}, \bm{\xi}_{n, k} \rangle \frac{\exp (\langle \bm{q}_{n, i}^{(t)}, \bm{k}_-^{(t)} \rangle)}{\exp (\langle \bm{q}_{n, i}^{(t)}, \bm{k}_-^{(t)} \rangle) + \sum\limits_{j=2}^M \exp ( \langle \bm{q}_{n, i}^{(t)}, \bm{k}_{n, j}^{(t)} \rangle )} \\
    &\cdot \frac{\exp (\langle \bm{q}_{n, i}^{(t)}, \bm{k}_{n, k}^{(t)} \rangle)}{\exp (\langle \bm{q}_{n, i}^{(t)}, \bm{k}_-^{(t)} \rangle) + \sum\limits_{j=2}^M \exp ( \langle \bm{q}_{n, i}^{(t)}, \bm{k}_{n, j}^{(t)} \rangle )} \big) \\
    &+ \sum\limits_{k=2}^M V_{n, k}^{(t)} \big( \langle \bm{\xi}_{n^\prime, i^\prime}, \bm{\xi}_{n, k} \rangle \frac{\exp (\langle \bm{q}_{n, i}^{(t)}, \bm{k}_{n, k}^{(t)} \rangle)}{\exp (\langle \bm{q}_{n, i}^{(t)}, \bm{k}_-^{(t)} \rangle) + \sum\limits_{j=2}^M \exp ( \langle \bm{q}_{n, i}^{(t)}, \bm{k}_{n, j}^{(t)} \rangle )}\\
    &- \sum\limits_{l=2}^M \big( \langle \bm{\xi}_{n^\prime, i^\prime}, \bm{\xi}_{n, l} \rangle \frac{\exp (\langle \bm{q}_{n, i}^{(t)}, \bm{k}_{n, k}^{(t)} \rangle)}{\exp (\langle \bm{q}_{n, i}^{(t)}, \bm{k}_-^{(t)} \rangle) + \sum\limits_{j=2}^M \exp ( \langle \bm{q}_{n, i}^{(t)}, \bm{k}_{n, j}^{(t)} \rangle )} \\
    &\cdot \frac{\exp (\langle \bm{q}_{n, i}^{(t)}, \bm{k}_{n, l}^{(t)} \rangle)}{\exp (\langle \bm{q}_{n, i}^{(t)}, \bm{k}_-^{(t)} \rangle) + \sum\limits_{j=2}^M \exp ( \langle \bm{q}_{n, i}^{(t)}, \bm{k}_{n, j}^{(t)} \rangle )} \big) \big) \Big) \\
\end{split}
\end{equation}
for \(n \in S_-\).

\subsection{Update Rules for Inner Products}
In this subsection, we give the update rules for the inner products of \(\bm{q}\) and \(\bm{k}\).

\textbf{Restatement of Lemma \ref{update_rule4QK}.} The dynamics of \( \bm{x} \bm{W}_Q \bm{W}_K \bm{X}^\top \) can be characterized as follows

\begin{equation*}
\begin{split}
    &\langle \bm{q}_+^{(t+1)}, \bm{k}_+^{(t+1)} \rangle - \langle \bm{q}_+^{(t)}, \bm{k}_+^{(t)} \rangle \\
    &= \alpha_{+, +}^{(t)} \Vert \bm{k}_+^{(t)} \Vert_2^2 + \sum\limits_{n \in S_+} \sum\limits_{i=2}^M \alpha_{n, +, i}^{(t)} \langle \bm{k}_+^{(t)}, \bm{k}_{n, i}^{(t)} \rangle \\
    &+ \beta_{+, +}^{(t)} \Vert \bm{q}_+^{(t)} \Vert_2^2 + \sum\limits_{n \in S_+} \sum\limits_{i=2}^M \beta_{n, +, i}^{(t)} \langle \bm{q}_+^{(t)}, \bm{q}_{n, i}^{(t)} \rangle \\
    &+ \Big( \alpha_{+, +}^{(t)} \bm{k}_+^{(t)} + \sum\limits_{n \in S_+} \sum\limits_{i=2}^M \alpha_{n, +, i}^{(t)} \bm{k}_{n, i}^{(t)} \Big) \\
    &\cdot \Big( \beta_{+, +}^{(t)} \bm{q}_+^{(t)\top} + \sum\limits_{n \in S_+} \sum\limits_{i=2}^M \beta_{n, +, i}^{(t)} \bm{q}_{n, i}^{(t)\top} \Big),
\end{split}
\end{equation*}

\begin{equation*}
\begin{split}
    &\langle \bm{q}_-^{(t+1)}, \bm{k}_-^{(t+1)} \rangle - \langle \bm{q}_-^{(t)}, \bm{k}_-^{(t)} \rangle \\
    &= \alpha_{-, -}^{(t)} \Vert \bm{k}_-^{(t)} \Vert_2^2 + \sum\limits_{n \in S_-} \sum\limits_{i=2}^M \alpha_{n, -, i}^{(t)} \langle \bm{k}_-^{(t)}, \bm{k}_{n, i}^{(t)} \rangle \\
    &+ \beta_{-, -}^{(t)} \Vert \bm{q}_-^{(t)} \Vert_2^2 + \sum\limits_{n \in S_-} \sum\limits_{i=2}^M \beta_{n, -, i}^{(t)} \langle \bm{q}_+^{(t)}, \bm{q}_{n, i}^{(t)} \rangle \\
    &+ \Big( \alpha_{-, -}^{(t)} \bm{k}_-^{(t)} + \sum\limits_{n \in S_-} \sum\limits_{i=2}^M \alpha_{n, -, i}^{(t)} \bm{k}_{n, i}^{(t)} \Big) \\
    &\cdot \Big( \beta_{-, -}^{(t)} \bm{q}_-^{(t)\top} + \sum\limits_{n \in S_-} \sum\limits_{i=2}^M \beta_{n, -, i}^{(t)} \bm{q}_{n, i}^{(t)\top} \Big),
\end{split}
\end{equation*}

\begin{equation*}
\begin{split}
    &\langle \bm{q}_{n, i}^{(t+1)}, \bm{k}_+^{(t+1)} \rangle - \langle \bm{q}_{n, i}^{(t)}, \bm{k}_+^{(t)} \rangle \\
    &= \alpha_{n, i, +}^{(t)} \Vert \bm{k}_+^{(t)} \Vert_2^2 + \alpha_{n, i, -}^{(t)} \langle \bm{k}_+^{(t)}, \bm{k}_-^{(t)} \rangle + \sum\limits_{n^\prime =1}^N \sum\limits_{l=2}^M \alpha_{n, i, n^\prime, l}^{(t)} \langle \bm{k}_+^{(t)}, \bm{k}_{n^\prime, l}^{(t)} \rangle \\
    &+ \beta_{+, +}^{(t)} \langle \bm{q}_+^{(t)}, \bm{q}_{n, i}^{(t)} \rangle + \sum\limits_{n^\prime \in S_+} \sum\limits_{l=2}^M \beta_{n^\prime, +, l}^{(t)} \langle \bm{q}_{n, i}^{(t)}, \bm{q}_{n^\prime, l}^{(t)} \rangle \\
    &+ \Big( \alpha_{n, i, +}^{(t)} \bm{k}_+^{(t)} + \alpha_{n, i, -}^{(t)} \bm{k}_-^{(t)} + \sum\limits_{n^\prime =1}^N \sum\limits_{l=2}^M \alpha_{n, i, n^\prime, l}^{(t)} \bm{k}_{n^\prime, l}^{(t)} \Big) \\
    &\cdot \Big( \beta_{+, +}^{(t)} \bm{q}_+^{(t)\top} + \sum\limits_{n^\prime \in S_+} \sum\limits_{l=2}^M \beta_{n^\prime, +, l}^{(t)} \bm{q}_{n^\prime, l}^{(t)\top} \Big),
\end{split}
\end{equation*}

\begin{equation*}
\begin{split}
    &\langle \bm{q}_{n, i}^{(t+1)}, \bm{k}_-^{(t+1)} \rangle - \langle \bm{q}_{n, i}^{(t)}, \bm{k}_-^{(t)} \rangle \\
    &= \alpha_{n, i, -}^{(t)} \Vert \bm{k}_-^{(t)} \Vert_2^2 + \alpha_{n, i, +}^{(t)} \langle \bm{k}_+^{(t)}, \bm{k}_-^{(t)} \rangle + \sum\limits_{n^\prime =1}^N \sum\limits_{l=2}^M \alpha_{n, i, n^\prime, l}^{(t)} \langle \bm{k}_-^{(t)}, \bm{k}_{n^\prime, l}^{(t)} \rangle \\
    &+ \beta_{-, -}^{(t)} \langle \bm{q}_-^{(t)}, \bm{q}_{n, i}^{(t)} \rangle + \sum\limits_{n^\prime \in S_-} \sum\limits_{l=2}^M \beta_{n^\prime, -, l}^{(t)} \langle \bm{q}_{n, i}^{(t)}, \bm{q}_{n^\prime, l}^{(t)} \rangle \\
    &+ \Big( \alpha_{n, i, +}^{(t)} \bm{k}_+^{(t)} + \alpha_{n, i, -}^{(t)} \bm{k}_-^{(t)} + \sum\limits_{n^\prime =1}^N \sum\limits_{l=2}^M \alpha_{n, i, n^\prime, l}^{(t)} \bm{k}_{n^\prime, l}^{(t)} \Big) \\
    &\cdot \Big( \beta_{-, -}^{(t)} \bm{q}_-^{(t)\top} + \sum\limits_{n^\prime \in S_-} \sum\limits_{l=2}^M \beta_{n^\prime, -, l}^{(t)} \bm{q}_{n^\prime, l}^{(t)\top} \Big),
\end{split}
\end{equation*}

\begin{equation*}
\begin{split}
    &\langle \bm{q}_+^{(t+1)}, \bm{k}_{n, j}^{(t+1)} \rangle - \langle \bm{q}_+^{(t)}, \bm{k}_{n, j}^{(t)} \rangle \\
    &= \alpha_{+, +}^{(t)} \langle \bm{k}_+^{(t)}, \bm{k}_{n, j}^{(t)} \rangle + \sum\limits_{n^\prime \in S_+} \sum\limits_{l=2}^M \alpha_{n^\prime, +, l}^{(t)} \langle \bm{k}_{n, j}^{(t)}, \bm{k}_{n^\prime, l}^{(t)} \rangle \\
    &+ \beta_{n, j, +}^{(t)} \Vert \bm{q}_+^{(t)} \Vert_2^2 + \beta_{n, j, -}^{(t)} \langle \bm{q}_+^{(t)}, \bm{q}_-^{(t)} \rangle + \sum\limits_{n^\prime = 1}^N \sum\limits_{l=2}^M \beta_{n, j, n^\prime, l}^{(t)} \langle \bm{q}_+^{(t)}, \bm{q}_{n^\prime, l}^{(t)} \rangle \\
    &+ \Big( \alpha_{+, +}^{(t)} \bm{k}_+^{(t)} + \sum\limits_{n^\prime \in S_+} \sum\limits_{l=2}^M \alpha_{n^\prime, +, l}^{(t)} \bm{k}_{n^\prime, l}^{(t)} \Big) \\
    &\cdot \Big( \beta_{n, j, +}^{(t)} \bm{q}_+^{(t)\top} + \beta_{n, j, -}^{(t)} \bm{q}_-^{(t)\top} + \sum\limits_{n^\prime = 1}^N \sum\limits_{l=2}^M \beta_{n, j, n^\prime, l}^{(t)} \bm{q}_{n^\prime, l}^{(t)\top} \Big),
\end{split}
\end{equation*}

\begin{equation*}
\begin{split}
    &\langle \bm{q}_-^{(t+1)}, \bm{k}_{n, j}^{(t+1)} \rangle - \langle \bm{q}_-^{(t)}, \bm{k}_{n, j}^{(t)} \rangle \\
    &= \alpha_{-, -}^{(t)} \langle \bm{k}_-^{(t)}, \bm{k}_{n, j}^{(t)} \rangle + \sum\limits_{n^\prime \in S_-} \sum\limits_{l=2}^M \alpha_{n^\prime, -, l}^{(t)} \langle \bm{k}_{n, j}^{(t)}, \bm{k}_{n^\prime, l}^{(t)} \rangle \\
    &+ \beta_{n, j, -}^{(t)} \Vert \bm{q}_-^{(t)} \Vert_2^2 + \beta_{n, j, +}^{(t)} \langle \bm{q}_+^{(t)}, \bm{q}_-^{(t)} \rangle + \sum\limits_{n^\prime = 1}^N \sum\limits_{l=2}^M \beta_{n, j, n^\prime, l}^{(t)} \langle \bm{q}_-^{(t)}, \bm{q}_{n^\prime, l}^{(t)} \rangle \\
    &+ \Big( \alpha_{-, -}^{(t)} \bm{k}_-^{(t)} + \sum\limits_{n^\prime \in S_-} \sum\limits_{l=2}^M \alpha_{n^\prime, -, l}^{(t)} \bm{k}_{n^\prime, l}^{(t)} \Big) \\
    &\cdot \Big( \beta_{n, j, +}^{(t)} \bm{q}_+^{(t)\top} + \beta_{n, j, -}^{(t)} \bm{q}_-^{(t)\top} + \sum\limits_{n^\prime = 1}^N \sum\limits_{l=2}^M \beta_{n, j, n^\prime, l}^{(t)} \bm{q}_{n^\prime, l}^{(t)\top} \Big),
\end{split}
\end{equation*}

\begin{equation*}
\begin{split}
    &\langle \bm{q}_{n, i}^{(t+1)}, \bm{k}_{n, j}^{(t+1)} \rangle - \langle \bm{q}_{n, i}^{(t)}, \bm{k}_{n, j}^{(t)} \rangle \\
    &= \alpha_{n, i, +}^{(t)} \langle \bm{k}_+^{(t)}, \bm{k}_{n, j}^{(t)} \rangle + \alpha_{n, i, -}^{(t)} \langle \bm{k}_-^{(t)}, \bm{k}_{n, j}^{(t)} \rangle + \sum\limits_{n^\prime =1}^N \sum\limits_{l=2}^M \alpha_{n, i, n^\prime, l}^{(t)} \langle \bm{k}_{n^\prime, l}^{(t)}, \bm{k}_{n, j}^{(t)} \rangle \\
    &+ \beta_{n, j, +}^{(t)} \langle \bm{q}_+^{(t)}, \bm{q}_{n, i}^{(t)} \rangle + \beta_{n, j, -}^{(t)} \langle \bm{q}_-^{(t)}, \bm{q}_{n, i}^{(t)} \rangle + \sum\limits_{n^\prime = 1}^N \sum\limits_{l=2}^M \beta_{n, j, n^\prime, l}^{(t)} \langle \bm{q}_{n^\prime, l}^{(t)}, \bm{q}_{n, i}^{(t)} \rangle \\
    &+ \Big( \alpha_{n, i, +}^{(t)} \bm{k}_+^{(t)} + \alpha_{n, i, -}^{(t)} \bm{k}_-^{(t)} + \sum\limits_{n^\prime =1}^N \sum\limits_{l=2}^M \alpha_{n, i, n^\prime, l}^{(t)} \bm{k}_{n^\prime, l}^{(t)} \Big) \\
    &\cdot \Big( \beta_{n, j, +}^{(t)} \bm{q}_+^{(t)\top} + \beta_{n, j, -}^{(t)} \bm{q}_-^{(t)\top} + \sum\limits_{n^\prime = 1}^N \sum\limits_{l=2}^M \beta_{n, j, n^\prime, l}^{(t)} \bm{q}_{n^\prime, l}^{(t)\top} \Big),
\end{split}
\end{equation*}
for \(i, j \in [M] \backslash \{1\}, n \in [N]\).

In addition to these equations above, we give the complete inner product update rule as follows

\begin{equation}
\begin{split}
    \label{q_i_k_j_ne_dy}
    &\langle \bm{q}_{n, i}^{(t+1)}, \bm{k}_{\overline{n}, j}^{(t+1)} \rangle - \langle \bm{q}_{n, i}^{(t)}, \bm{k}_{\overline{n}, j}^{(t)} \rangle \\
    &= \bm{\xi}_{n, j}^{\top} (\bm{W}_Q^{(t)} + \Delta\bm{W}_Q^{(t)}) (\bm{W}_K^{(t)\top} + \Delta\bm{W}_K^{(t)\top}) \bm{\xi}_{\overline{n}, k} - \langle \bm{q}_{n, i}^{(t)}, \bm{k}_{\overline{n}, j}^{(t)} \rangle \\
    &= \langle \Delta \bm{q}_{n, i}^{(t)}, \bm{k}_{\overline{n}, j}^{(t)} \rangle + \langle \bm{q}_{n, i}^{(t)}, \Delta \bm{k}_{\overline{n}, j}^{(t)} \rangle + \langle \Delta \bm{q}_{n, i}^{(t)}, \Delta \bm{k}_{\overline{n}, j}^{(t)} \rangle \\
    &= \alpha_{n, i, +}^{(t)} \langle \bm{k}_+^{(t)}, \bm{k}_{\overline{n}, j}^{(t)} \rangle + \alpha_{n, i, -}^{(t)} \langle \bm{k}_-^{(t)}, \bm{k}_{\overline{n}, j}^{(t)} \rangle + \sum\limits_{n^\prime =1}^N \sum\limits_{l=2}^M \alpha_{n, i, n^\prime, l}^{(t)} \langle \bm{k}_{n^\prime, l}^{(t)}, \bm{k}_{\overline{n}, j}^{(t)} \rangle \\
    &+ \beta_{\overline{n}, j, +}^{(t)} \langle \bm{q}_+^{(t)}, \bm{q}_{n, i}^{(t)} \rangle + \beta_{\overline{n}, j, -}^{(t)} \langle \bm{q}_-^{(t)}, \bm{q}_{n, i}^{(t)} \rangle + \sum\limits_{n^\prime = 1}^N \sum\limits_{l=2}^M \beta_{\overline{n}, j, n^\prime, l}^{(t)} \langle \bm{q}_{n^\prime, l}^{(t)}, \bm{q}_{n, i}^{(t)} \rangle \\
    &+ \Big( \alpha_{n, i, +}^{(t)} \bm{k}_+^{(t)} + \alpha_{n, i, -}^{(t)} \bm{k}_-^{(t)} + \sum\limits_{n^\prime =1}^N \sum\limits_{l=2}^M \alpha_{n, i, n^\prime, l}^{(t)} \bm{k}_{n^\prime, l}^{(t)} \Big) \\
    &\cdot \Big( \beta_{\overline{n}, j, +}^{(t)} \bm{q}_+^{(t)\top} + \beta_{\overline{n}, j, -}^{(t)} \bm{q}_-^{(t)\top} + \sum\limits_{n^\prime = 1}^N \sum\limits_{l=2}^M \beta_{\overline{n}, j, n^\prime, l}^{(t)} \bm{q}_{n^\prime, l}^{(t)\top} \Big)
\end{split}
\end{equation}
for \(n \ne \overline{n}\),

\begin{equation}
\begin{split}
    &\langle \bm{q}_-^{(t+1)}, \bm{k}_+^{(t+1)} \rangle - \langle \bm{q}_-^{(t)}, \bm{k}_+^{(t)} \rangle \\
    &= \langle \Delta \bm{q}_-^{(t)}, \bm{k}_+^{(t)} \rangle + \langle \bm{q}_-^{(t)}, \Delta \bm{k}_+^{(t)} \rangle + \langle \Delta \bm{q}_-^{(t)}, \Delta \bm{k}_+^{(t)} \rangle \\
    &= \alpha_{-, -}^{(t)} \langle \bm{k}_+^{(t)}, \bm{k}_-^{(t)} \rangle + \sum\limits_{n \in S_-} \sum\limits_{i=2}^M \alpha_{n, -, i}^{(t)} \langle \bm{k}_{n, i}^{(t)}, \bm{k}_+^{(t)} \rangle \\
    &+ \beta_{+, +}^{(t)} \langle \bm{q}_+^{(t)}, \bm{q}_-^{(t)} \rangle + \sum\limits_{n \in S_+} \sum\limits_{i=2}^M \beta_{n, +, i}^{(t)} \langle \bm{q}_{n, i}^{(t)}, \bm{q}_-^{(t)} \rangle \\
    &+ \Big( \alpha_{-, -}^{(t)} \bm{k}_-^{(t)} + \sum\limits_{n \in S_-} \sum\limits_{i=2}^M \alpha_{n, -, i}^{(t)} \bm{k}_{n, i}^{(t)} \Big) \\
    &\cdot \Big( \beta_{+, +}^{(t)} \bm{q}_+^{(t)\top} + \sum\limits_{n \in S_+} \sum\limits_{i=2}^M \beta_{n, +, i}^{(t)} \bm{q}_{n, i}^{(t)\top} \Big),
\end{split}
\end{equation}

\begin{equation}
\begin{split}
    \label{q_p_q_m_dy}
    &\langle \bm{q}_+^{(t+1)}, \bm{q}_-^{(t+1)} \rangle - \langle \bm{q}_+^{(t)}, \bm{q}_-^{(t)} \rangle \\
    &= \langle \Delta \bm{q}_+^{(t)}, \bm{q}_-^{(t)} \rangle + \langle \bm{q}_+^{(t)}, \Delta \bm{q}_-^{(t)} \rangle + \langle \Delta \bm{q}_+^{(t)}, \Delta \bm{q}_-^{(t)} \rangle \\
    &= \alpha_{+, +}^{(t)} \langle \bm{q}_-^{(t)}, \bm{k}_+^{(t)} \rangle + \sum\limits_{n \in S_+} \sum\limits_{i=2}^M \alpha_{n, +, i}^{(t)} \langle \bm{q}_-^{(t)}, \bm{k}_{n, i}^{(t)} \rangle \\
    &+ \alpha_{-, -}^{(t)} \langle \bm{q}_+^{(t)}, \bm{k}_-^{(t)} \rangle + \sum\limits_{n \in S_-} \sum\limits_{i=2}^M \alpha_{n, -, i}^{(t)} \langle \bm{q}_+^{(t)}, \bm{k}_{n, i}^{(t)} \rangle \\
    &+ \Big( \alpha_{+, +}^{(t)} \bm{k}_+^{(t)} + \sum\limits_{n \in S_+} \sum\limits_{i=2}^M \alpha_{n, +, i}^{(t)} \bm{k}_{n, i}^{(t)} \Big) \\
    &\cdot \Big( \alpha_{-, -}^{(t)} \bm{k}_-^{(t)\top} + \sum\limits_{n \in S_-} \sum\limits_{i=2}^M \alpha_{n, -, i}^{(t)} \bm{k}_{n, i}^{(t)\top} \Big),
\end{split}
\end{equation}

\begin{equation}
\begin{split}
    \label{q_p_sq_dy}
    &\Vert \bm{q}_+^{(t+1)} \Vert_2^2 - \Vert \bm{q}_+^{(t)} \Vert_2^2 \\
    &= 2 \langle \Delta \bm{q}_+^{(t)}, \bm{q}_+^{(t)} \rangle + \langle \Delta \bm{q}_+^{(t)}, \Delta \bm{q}_+^{(t)} \rangle \\
    &= 2 \alpha_{+, +}^{(t)} \langle \bm{q}_+^{(t)}, \bm{k}_+^{(t)} \rangle + 2 \sum\limits_{n \in S_+} \sum\limits_{i=2}^M \alpha_{n, +, i}^{(t)} \langle \bm{q}_+^{(t)}, \bm{k}_{n, i}^{(t)} \rangle \\
    &+ \Big( \alpha_{+, +}^{(t)} \bm{k}_+^{(t)} + \sum\limits_{n \in S_+} \sum\limits_{i=2}^M \alpha_{n, +, i}^{(t)} \bm{k}_{n, i}^{(t)} \Big) \\
    &\cdot \Big( \alpha_{+, +}^{(t)} \bm{k}_+^{(t)\top} + \sum\limits_{n \in S_+} \sum\limits_{i=2}^M \alpha_{n, +, i}^{(t)} \bm{k}_{n, i}^{(t)\top} \Big),
\end{split}
\end{equation}

\begin{equation}
\begin{split}
    \label{q_n_sq_dy}
    &\Vert \bm{q}_-^{(t+1)} \Vert_2^2 - \Vert \bm{q}_-^{(t)} \Vert_2^2 \\
    &= 2 \langle \Delta \bm{q}_-^{(t)}, \bm{q}_-^{(t)} \rangle + \langle \Delta \bm{q}_-^{(t)}, \Delta \bm{q}_-^{(t)} \rangle \\
    &= 2 \alpha_{-, -}^{(t)} \langle \bm{q}_-^{(t)}, \bm{k}_-^{(t)} \rangle + 2 \sum\limits_{n \in S_-} \sum\limits_{i=2}^M \alpha_{n, -, i}^{(t)} \langle \bm{q}_-^{(t)}, \bm{k}_{n, i}^{(t)} \rangle \\
    &+ \Big( \alpha_{-, -}^{(t)} \bm{k}_-^{(t)} + \sum\limits_{n \in S_-} \sum\limits_{i=2}^M \alpha_{n, -, i}^{(t)} \bm{k}_{n, i}^{(t)} \Big) \\
    &\cdot \Big( \alpha_{-, -}^{(t)} \bm{k}_-^{(t)\top} + \sum\limits_{n \in S_-} \sum\limits_{i=2}^M \alpha_{n, -, i}^{(t)} \bm{k}_{n, i}^{(t)\top} \Big),
\end{split}
\end{equation}

\begin{equation}
\begin{split}
    &\Vert \bm{q}_{n, i}^{(t+1)} \Vert_2^2 - \Vert \bm{q}_{n, i}^{(t)} \Vert_2^2 \\
    &= 2 \langle \Delta \bm{q}_{n, i}^{(t)}, \bm{q}_{n, i}^{(t)} \rangle + \langle \Delta \bm{q}_{n, i}^{(t)}, \Delta \bm{q}_{n, i}^{(t)} \rangle \\
    &= 2 \alpha_{n, i, +}^{(t)} \langle \bm{q}_{n, i}^{(t)}, \bm{k}_+^{(t)} \rangle + 2 \alpha_{n, i, -}^{(t)} \langle \bm{q}_{n, i}^{(t)}, \bm{k}_-^{(t)} \rangle + 2 \sum\limits_{n^\prime = 1}^N \sum\limits_{l=2}^M \alpha_{n, i, n^\prime, l}^{(t)} \langle \bm{q}_{n, i}^{(t)}, \bm{k}_{n^\prime, l}^{(t)} \rangle \\
    &+ \Big( \alpha_{n, i, +}^{(t)} \bm{k}_+^{(t)} + \alpha_{n, i, -}^{(t)} \bm{k}_-^{(t)} + \sum\limits_{n^\prime = 1}^N \sum\limits_{l=2}^M \alpha_{n, i, n^\prime, l}^{(t)} \bm{k}_{n^\prime, l}^{(t)} \Big) \\
    &\cdot \Big( \alpha_{n, i, +}^{(t)} \bm{k}_+^{(t)\top} + \alpha_{n, i, -}^{(t)} \bm{k}_-^{(t)\top} + \sum\limits_{n^\prime = 1}^N \sum\limits_{l=2}^M \alpha_{n, i, n^\prime, l}^{(t)} \bm{k}_{n^\prime, l}^{(t)\top} \Big),
\end{split}
\end{equation}

\begin{equation}
\begin{split}
    \label{q_p_q_i_dy}
    &\langle \bm{q}_+^{(t+1)}, \bm{q}_{n, i}^{(t+1)} \rangle - \langle \bm{q}_+^{(t)}, \bm{q}_{n, i}^{(t)} \rangle \\
    &= \langle \Delta \bm{q}_+^{(t)}, \bm{q}_{n, i}^{(t)} \rangle + \langle \bm{q}_+^{(t)}, \Delta \bm{q}_{n, i}^{(t)} \rangle + \langle \Delta \bm{q}_+^{(t)}, \Delta \bm{q}_{n, i}^{(t)} \rangle \\
    &= \alpha_{+, +}^{(t)} \langle \bm{q}_{n, i}^{(t)}, \bm{k}_+^{(t)} \rangle + \sum\limits_{n^\prime \in S_+} \sum\limits_{l=2}^M \alpha_{n^\prime, +, l}^{(t)} \langle \bm{q}_{n, i}^{(t)}, \bm{k}_{n^\prime, l}^{(t)} \rangle \\
    &+ \alpha_{n, i, +}^{(t)} \langle \bm{q}_+^{(t)}, \bm{k}_+^{(t)} \rangle + \alpha_{n, i, -}^{(t)} \langle \bm{q}_+^{(t)}, \bm{k}_-^{(t)} \rangle + \sum\limits_{n^\prime = 1}^N \sum\limits_{l=2}^M \alpha_{n, i, n^\prime, l}^{(t)} \langle \bm{q}_+^{(t)}, \bm{k}_{n^\prime, l}^{(t)} \rangle \\
    &+ \Big( \alpha_{+, +}^{(t)} \bm{k}_+^{(t)} + \sum\limits_{n \in S_+} \sum\limits_{i=2}^M \alpha_{n, +, i}^{(t)} \bm{k}_{n, i}^{(t)} \Big) \\
    &\cdot \Big( \alpha_{n, i, +}^{(t)} \bm{k}_+^{(t)\top} + \alpha_{n, i, -}^{(t)} \bm{k}_-^{(t)\top} + \sum\limits_{n^\prime=1}^N \sum\limits_{l=2}^M \alpha_{n, i, n^\prime, l}^{(t)} \bm{k}_{n^\prime, l}^{(t)\top} \Big),
\end{split}
\end{equation}

\begin{equation}
\begin{split}
    \label{q_n_q_i_dy}
    &\langle \bm{q}_-^{(t+1)}, \bm{q}_{n, i}^{(t+1)} \rangle - \langle \bm{q}_-^{(t)}, \bm{q}_{n, i}^{(t)} \rangle \\
    &= \langle \Delta \bm{q}_-^{(t)}, \bm{q}_{n, i}^{(t)} \rangle + \langle \bm{q}_-^{(t)}, \Delta \bm{q}_{n, i}^{(t)} \rangle + \langle \Delta \bm{q}_-^{(t)}, \Delta \bm{q}_{n, i}^{(t)} \rangle \\
    &= \alpha_{-, -}^{(t)} \langle \bm{q}_{n, i}^{(t)}, \bm{k}_-^{(t)} \rangle + \sum\limits_{n^\prime \in S_-} \sum\limits_{l=2}^M \alpha_{n^\prime, -, l}^{(t)} \langle \bm{q}_{n, i}^{(t)}, \bm{k}_{n^\prime, l}^{(t)} \rangle \\
    &+ \alpha_{n, i, +}^{(t)} \langle \bm{q}_-^{(t)}, \bm{k}_+^{(t)} \rangle + \alpha_{n, i, -}^{(t)} \langle \bm{q}_-^{(t)}, \bm{k}_-^{(t)} \rangle + \sum\limits_{n^\prime = 1}^N \sum\limits_{l=2}^M \alpha_{n, i, n^\prime, l}^{(t)} \langle \bm{q}_-^{(t)}, \bm{k}_{n^\prime, l}^{(t)} \rangle \\
    &+ \Big( \alpha_{-, -}^{(t)} \bm{k}_-^{(t)} + \sum\limits_{n \in S_-} \sum\limits_{i=2}^M \alpha_{n, -, i}^{(t)} \bm{k}_{n, i}^{(t)} \Big) \\
    &\cdot \Big( \alpha_{n, i, +}^{(t)} \bm{k}_+^{(t)\top} + \alpha_{n, i, -}^{(t)} \bm{k}_-^{(t)\top} + \sum\limits_{n^\prime=1}^N \sum\limits_{l=2}^M \alpha_{n, i, n^\prime, l}^{(t)} \bm{k}_{n^\prime, l}^{(t)\top} \Big),
\end{split}
\end{equation}
    
\begin{equation}
\begin{split}
    \label{q_i_q_j_dy}
    &\langle \bm{q}_{n, i}^{(t+1)}, \bm{q}_{n, j}^{(t+1)} \rangle - \langle \bm{q}_{n, i}^{(t)}, \bm{q}_{n, j}^{(t)} \rangle \\
    &= \langle \Delta \bm{q}_{n, i}^{(t)}, \bm{q}_{n, j}^{(t)} \rangle + \langle \bm{q}_{n, i}^{(t)}, \Delta \bm{q}_{n, j}^{(t)} \rangle + \langle \Delta \bm{q}_{n, i}^{(t)}, \Delta \bm{q}_{n, j}^{(t)} \rangle \\
    &= \alpha_{n, i, +}^{(t)} \langle \bm{q}_{n, j}^{(t)}, \bm{k}_+^{(t)} \rangle + \alpha_{n, i, -}^{(t)} \langle \bm{q}_{n, j}^{(t)}, \bm{k}_-^{(t)} \rangle + \sum\limits_{n^\prime = 1}^N \sum\limits_{l=2}^M \alpha_{n, i, n^\prime, l}^{(t)} \langle \bm{q}_{n, j}^{(t)}, \bm{k}_{n^\prime, l}^{(t)} \rangle \\
    &+ \alpha_{n, j, +}^{(t)} \langle \bm{q}_{n, i}^{(t)}, \bm{k}_+^{(t)} \rangle + \alpha_{n, j, -}^{(t)} \langle \bm{q}_{n, i}^{(t)}, \bm{k}_-^{(t)} \rangle + \sum\limits_{n^\prime = 1}^N \sum\limits_{l=2}^M \alpha_{n, j, n^\prime, l}^{(t)} \langle \bm{q}_{n, i}^{(t)}, \bm{k}_{n^\prime, l}^{(t)} \rangle \\
    &+ \Big( \alpha_{n, i, +}^{(t)} \bm{k}_+^{(t)} + \alpha_{n, i, -}^{(t)} \bm{k}_-^{(t)} + \sum\limits_{n^\prime = 1}^N \sum\limits_{l=2}^M \alpha_{n, i, n^\prime, l}^{(t)} \bm{k}_{n^\prime, l}^{(t)} \Big) \\
    &\cdot \Big( \alpha_{n, j, +}^{(t)} \bm{k}_+^{(t)\top} + \alpha_{n, j, -}^{(t)} \bm{k}_-^{(t)\top} + \sum\limits_{n^\prime = 1}^N \sum\limits_{l=2}^M \alpha_{n, j, n^\prime, l}^{(t)} \bm{k}_{n^\prime, l}^{(t)\top} \Big),
\end{split}
\end{equation}

\begin{equation}
\begin{split}
    \label{q_i_q_j_n_dy}
    &\langle \bm{q}_{n, i}^{(t+1)}, \bm{q}_{\overline{n}, j}^{(t+1)} \rangle - \langle \bm{q}_{n, i}^{(t)}, \bm{q}_{\overline{n}, j}^{(t)} \rangle \\
    &= \langle \Delta \bm{q}_{n, i}^{(t)}, \bm{q}_{\overline{n}, j}^{(t)} \rangle + \langle \bm{q}_{n, i}^{(t)}, \Delta \bm{q}_{\overline{n}, j}^{(t)} \rangle + \langle \Delta \bm{q}_{n, i}^{(t)}, \Delta \bm{q}_{\overline{n}, j}^{(t)} \rangle \\
    &= \alpha_{n, i, +}^{(t)} \langle \bm{q}_{\overline{n}, j}^{(t)}, \bm{k}_+^{(t)} \rangle + \alpha_{n, i, -}^{(t)} \langle \bm{q}_{\overline{n}, j}^{(t)}, \bm{k}_-^{(t)} \rangle + \sum\limits_{n^\prime = 1}^N \sum\limits_{l=2}^M \alpha_{n, i, n^\prime, l}^{(t)} \langle \bm{q}_{\overline{n}, j}^{(t)}, \bm{k}_{n^\prime, l}^{(t)} \rangle \\
    &+ \alpha_{\overline{n}, j, +}^{(t)} \langle \bm{q}_{n, i}^{(t)}, \bm{k}_+^{(t)} \rangle + \alpha_{\overline{n}, j, -}^{(t)} \langle \bm{q}_{n, i}^{(t)}, \bm{k}_-^{(t)} \rangle + \sum\limits_{n^\prime = 1}^N \sum\limits_{l=2}^M \alpha_{\overline{n}, j, n^\prime, l}^{(t)} \langle \bm{q}_{n, i}^{(t)}, \bm{k}_{n^\prime, l}^{(t)} \rangle \\
    &+ \Big( \alpha_{n, i, +}^{(t)} \bm{k}_+^{(t)} + \alpha_{n, i, -}^{(t)} \bm{k}_-^{(t)} + \sum\limits_{n^\prime = 1}^N \sum\limits_{l=2}^M \alpha_{n, i, n^\prime, l}^{(t)} \bm{k}_{n^\prime, l}^{(t)} \Big) \\
    &\cdot \Big( \alpha_{\overline{n}, j, +}^{(t)} \bm{k}_+^{(t)\top} + \alpha_{\overline{n}, j, -}^{(t)} \bm{k}_-^{(t)\top} + \sum\limits_{n^\prime = 1}^N \sum\limits_{l=2}^M \alpha_{\overline{n}, j, n^\prime, l}^{(t)} \bm{k}_{n^\prime, l}^{(t)\top} \Big)
\end{split}
\end{equation}
for \(n \ne \overline{n}\),

\begin{equation}
\begin{split}
    &\Vert \bm{k}_+^{(t+1)} \Vert_2^2 - \Vert \bm{k}_+^{(t)} \Vert_2^2 \\
    &= 2 \langle \Delta \bm{k}_+^{(t)}, \bm{k}_+^{(t)} \rangle + \langle \Delta \bm{k}_+^{(t)}, \Delta \bm{k}_+^{(t)} \rangle \\
    &= 2 \beta_{+, +}^{(t)} \langle \bm{q}_+^{(t)}, \bm{k}_+^{(t)} \rangle + 2 \sum\limits_{n \in S_+} \sum\limits_{i=2}^M \beta_{n, +, i}^{(t)} \langle \bm{q}_{n, i}^{(t)}, \bm{k}_+^{(t)} \rangle \\
    &+ \Big( \beta_{+, +}^{(t)} \bm{q}_+^{(t)} + \sum\limits_{n \in S_+} \sum\limits_{i=2}^M \beta_{n, +, i}^{(t)} \bm{q}_{n, i}^{(t)} \Big) \\
    &\cdot \Big( \beta_{+, +}^{(t)} \bm{q}_+^{(t)\top} + \sum\limits_{n \in S_+} \sum\limits_{i=2}^M \beta_{n, +, i}^{(t)} \bm{q}_{n, i}^{(t)\top} \Big),
\end{split}
\end{equation}

\begin{equation}
\begin{split}
    &\Vert \bm{k}_-^{(t+1)} \Vert_2^2 - \Vert \bm{k}_-^{(t)} \Vert_2^2 \\
    &= 2 \langle \Delta \bm{k}_-^{(t)}, \bm{k}_-^{(t)} \rangle + \langle \Delta \bm{k}_-^{(t)}, \Delta \bm{k}_-^{(t)} \rangle \\
    &= 2 \beta_{-, -}^{(t)} \langle \bm{q}_-^{(t)}, \bm{k}_-^{(t)} \rangle + 2 \sum\limits_{n \in S_-} \sum\limits_{i=2}^M \beta_{n, -, i}^{(t)} \langle \bm{q}_{n, i}^{(t)}, \bm{k}_-^{(t)} \rangle \\
    &+ \Big( \beta_{-, -}^{(t)} \bm{q}_-^{(t)} + \sum\limits_{n \in S_-} \sum\limits_{i=2}^M \beta_{n, -, i}^{(t)} \bm{q}_{n, i}^{(t)} \Big) \\
    &\cdot \Big( \beta_{-, -}^{(t)} \bm{q}_-^{(t)\top} + \sum\limits_{n \in S_-} \sum\limits_{i=2}^M \beta_{n, -, i}^{(t)} \bm{q}_{n, i}^{(t)\top} \Big),
\end{split}
\end{equation}

\begin{equation}
\begin{split}
    &\Vert \bm{k}_{n, i}^{(t+1)} \Vert_2^2 - \Vert \bm{k}_{n, i}^{(t)} \Vert_2^2 \\
    &= 2 \langle \Delta \bm{k}_{n, i}^{(t)}, \bm{k}_{n, i}^{(t)} \rangle + \langle \Delta \bm{k}_{n, i}^{(t)}, \Delta \bm{k}_{n, i}^{(t)} \rangle \\
    &= 2 \beta_{n, i, +}^{(t)} \langle \bm{q}_+^{(t)}, \bm{k}_{n, i}^{(t)} \rangle + 2 \beta_{n, i, -}^{(t)} \langle \bm{q}_-^{(t)}, \bm{k}_{n, i}^{(t)} \rangle + 2 \sum\limits_{n^\prime = 1}^N \sum\limits_{l=2}^M \beta_{n, i, n^\prime, l}^{(t)} \langle \bm{q}_{n^\prime, l}^{(t)}, \bm{k}_{n, i}^{(t)} \rangle \\
    &+ \Big( \beta_{n, i, +}^{(t)} \bm{q}_+^{(t)} + \beta_{n, i, -}^{(t)} \bm{q}_-^{(t)} + \sum\limits_{n^\prime = 1}^N \sum\limits_{l=2}^M \beta_{n, i, n^\prime, l}^{(t)} \bm{q}_{n^\prime, l}^{(t)} \Big) \\
    &\cdot \Big( \beta_{n, i, +}^{(t)} \bm{q}_+^{(t)\top} + \beta_{n, i, -}^{(t)} \bm{q}_-^{(t)\top} + \sum\limits_{n^\prime = 1}^N \sum\limits_{l=2}^M \beta_{n, i, n^\prime, l}^{(t)} \bm{q}_{n^\prime, l}^{(t)\top} \Big),
\end{split}
\end{equation}

\begin{equation}
\begin{split}
    \label{k_p_k_i_dy}
    &\langle \bm{k}_+^{(t+1)}, \bm{k}_{n, i}^{(t+1)} \rangle - \langle \bm{k}_+^{(t)}, \bm{k}_{n, i}^{(t)} \rangle \\
    &= \langle \Delta \bm{k}_+^{(t)}, \bm{k}_{n, i}^{(t)} \rangle + \langle \bm{k}_+^{(t)}, \Delta \bm{k}_{n, i}^{(t)} \rangle + \langle \Delta \bm{k}_+^{(t)}, \Delta \bm{k}_{n, i}^{(t)} \rangle \\
    &= \beta_{+, +}^{(t)} \langle \bm{q}_+^{(t)}, \bm{k}_{n, i}^{(t)} \rangle + \sum\limits_{n^\prime \in S_+} \sum\limits_{l=2}^M \beta_{n^\prime, +, l}^{(t)} \langle \bm{q}_{n^\prime, l}^{(t)}, \bm{k}_{n, i}^{(t)} \rangle \\
    &+ \beta_{n, i, +}^{(t)} \langle \bm{q}_+^{(t)}, \bm{k}_+^{(t)} \rangle + \beta_{n, i, -}^{(t)} \langle \bm{q}_-^{(t)}, \bm{k}_+^{(t)} \rangle + \sum\limits_{n^\prime = 1}^N \sum\limits_{l=2}^M \beta_{n, i, n^\prime, l}^{(t)} \langle \bm{q}_{n^\prime, l}^{(t)}, \bm{k}_+^{(t)} \rangle \\
    &+ \Big( \beta_{+, +}^{(t)} \bm{q}_+^{(t)} + \sum\limits_{n^\prime \in S_+} \sum\limits_{l=2}^M \beta_{n^\prime, +, l}^{(t)} \bm{q}_{n^\prime, l}^{(t)} \Big) \\
    &\cdot \Big( \beta_{n, i, +}^{(t)} \bm{q}_+^{(t)\top} + \beta_{n, i, -}^{(t)} \bm{q}_-^{(t)\top} + \sum\limits_{n^\prime = 1}^N \sum\limits_{l=2}^M \beta_{n, i, n^\prime, l}^{(t)} \bm{q}_{n^\prime, l}^{(t)\top} \Big),
\end{split}
\end{equation}

\begin{equation}
\begin{split}
    \label{k_n_k_i_dy}
    &\langle \bm{k}_-^{(t+1)}, \bm{k}_{n, i}^{(t+1)} \rangle - \langle \bm{k}_-^{(t)}, \bm{k}_{n, i}^{(t)} \rangle \\
    &= \langle \Delta \bm{k}_-^{(t)}, \bm{k}_{n, i}^{(t)} \rangle + \langle \bm{k}_-^{(t)}, \Delta \bm{k}_{n, i}^{(t)} \rangle + \langle \Delta \bm{k}_-^{(t)}, \Delta \bm{k}_{n, i}^{(t)} \rangle \\
    &= \beta_{-, -}^{(t)} \langle \bm{q}_-^{(t)}, \bm{k}_{n, i}^{(t)} \rangle + \sum\limits_{n^\prime \in S_-} \sum\limits_{l=2}^M \beta_{n^\prime, -, l}^{(t)} \langle \bm{q}_{n^\prime, l}^{(t)}, \bm{k}_{n, i}^{(t)} \rangle \\
    &+ \beta_{n, i, +}^{(t)} \langle \bm{q}_+^{(t)}, \bm{k}_-^{(t)} \rangle + \beta_{n, i, -}^{(t)} \langle \bm{q}_-^{(t)}, \bm{k}_-^{(t)} \rangle + \sum\limits_{n^\prime = 1}^N \sum\limits_{l=2}^M \beta_{n, i, n^\prime, l}^{(t)} \langle \bm{q}_{n^\prime, l}^{(t)}, \bm{k}_+^{(t)} \rangle \\
    &+ \Big( \beta_{-, -}^{(t)} \bm{q}_-^{(t)} + \sum\limits_{n^\prime \in S_-} \sum\limits_{l=2}^M \beta_{n^\prime, -, l}^{(t)} \bm{q}_{n^\prime, l}^{(t)} \Big) \\
    &\cdot \Big( \beta_{n, i, +}^{(t)} \bm{q}_+^{(t)\top} + \beta_{n, i, -}^{(t)} \bm{q}_-^{(t)\top} + \sum\limits_{n^\prime = 1}^N \sum\limits_{l=2}^M \beta_{n, i, n^\prime, l}^{(t)} \bm{q}_{n^\prime, l}^{(t)\top} \Big),
\end{split}
\end{equation}

\begin{equation}
\begin{split}
    \label{k_i_k_j_dy}
    &\langle \bm{k}_{n, i}^{(t+1)}, \bm{k}_{n, j}^{(t+1)} \rangle - \langle \bm{k}_{n, i}^{(t)}, \bm{k}_{n, j}^{(t)} \rangle \\
    &= \langle \Delta \bm{k}_{n, i}^{(t)}, \bm{k}_{n, j}^{(t)} \rangle + \langle \bm{k}_{n, i}^{(t)}, \Delta \bm{k}_{n, j}^{(t)} \rangle + \langle \Delta \bm{k}_{n, i}^{(t)}, \Delta \bm{k}_{n, j}^{(t)} \rangle \\
    &= \beta_{n, i, +}^{(t)} \langle \bm{q}_+^{(t)}, \bm{k}_{n, j}^{(t)} \rangle + \beta_{n, i, -}^{(t)} \langle \bm{q}_-^{(t)}, \bm{k}_{n, j}^{(t)} \rangle + \sum\limits_{n^\prime = 1}^N \sum\limits_{l=2}^M \beta_{n, i, n^\prime, l}^{(t)} \langle \bm{q}_{n^\prime, l}^{(t)}, \bm{k}_{n, j}^{(t)} \rangle \\
    &+ \beta_{n, j, +}^{(t)} \langle \bm{q}_+^{(t)}, \bm{k}_{n, i}^{(t)} \rangle + \beta_{n, j, -}^{(t)} \langle \bm{q}_-^{(t)}, \bm{k}_{n, i}^{(t)} \rangle + \sum\limits_{n^\prime = 1}^N \sum\limits_{l=2}^M \beta_{n, j, n^\prime, l}^{(t)} \langle \bm{q}_{n^\prime, l}^{(t)}, \bm{k}_{n, i}^{(t)} \rangle \\
    &+ \Big( \beta_{n, i, +}^{(t)} \bm{q}_+^{(t)} + \beta_{n, i, -}^{(t)} \bm{q}_-^{(t)} + \sum\limits_{n^\prime = 1}^N \sum\limits_{l=2}^M \beta_{n, i, n^\prime, l}^{(t)} \bm{q}_{n^\prime, l}^{(t)} \Big) \\
    &\cdot \Big( \beta_{n, j, +}^{(t)} \bm{q}_+^{(t)\top} + \beta_{n, j, -}^{(t)} \bm{q}_-^{(t)\top} + \sum\limits_{n^\prime = 1}^N \sum\limits_{l=2}^M \beta_{n, j, n^\prime, l}^{(t)} \bm{q}_{n^\prime, l}^{(t)\top} \Big),
\end{split}
\end{equation}

\begin{equation}
\begin{split}
    \label{k_i_k_j_n_dy}
    &\langle \bm{k}_{n, i}^{(t+1)}, \bm{k}_{\overline{n}, j}^{(t+1)} \rangle - \langle \bm{k}_{n, i}^{(t)}, \bm{k}_{\overline{n}, j}^{(t)} \rangle \\
    &= \langle \Delta \bm{k}_{n, i}^{(t)}, \bm{k}_{\overline{n}, j}^{(t)} \rangle + \langle \bm{k}_{n, i}^{(t)}, \Delta \bm{k}_{\overline{n}, j}^{(t)} \rangle + \langle \Delta \bm{k}_{n, i}^{(t)}, \Delta \bm{k}_{\overline{n}, j}^{(t)} \rangle \\
    &= \beta_{n, i, +}^{(t)} \langle \bm{q}_+^{(t)}, \bm{k}_{\overline{n}, j}^{(t)} \rangle + \beta_{n, i, -}^{(t)} \langle \bm{q}_-^{(t)}, \bm{k}_{\overline{n}, j}^{(t)} \rangle + \sum\limits_{n^\prime = 1}^N \sum\limits_{l=2}^M \beta_{n, i, n^\prime, l}^{(t)} \langle \bm{q}_{n^\prime, l}^{(t)}, \bm{k}_{\overline{n}, j}^{(t)} \rangle \\
    &+ \beta_{\overline{n}, j, +}^{(t)} \langle \bm{q}_+^{(t)}, \bm{k}_{n, i}^{(t)} \rangle + \beta_{\overline{n}, j, -}^{(t)} \langle \bm{q}_-^{(t)}, \bm{k}_{n, i}^{(t)} \rangle + \sum\limits_{n^\prime = 1}^N \sum\limits_{l=2}^M \beta_{\overline{n}, j, n^\prime, l}^{(t)} \langle \bm{q}_{n^\prime, l}^{(t)}, \bm{k}_{n, i}^{(t)} \rangle \\
    &+ \Big( \beta_{n, i, +}^{(t)} \bm{q}_+^{(t)} + \beta_{n, i, -}^{(t)} \bm{q}_-^{(t)} + \sum\limits_{n^\prime = 1}^N \sum\limits_{l=2}^M \beta_{n, i, n^\prime, l}^{(t)} \bm{q}_{n^\prime, l}^{(t)} \Big) \\
    &\cdot \Big( \beta_{\overline{n}, j, +}^{(t)} \bm{q}_+^{(t)\top} + \beta_{\overline{n}, j, -}^{(t)} \bm{q}_-^{(t)\top} + \sum\limits_{n^\prime = 1}^N \sum\limits_{l=2}^M \beta_{\overline{n}, j, n^\prime, l}^{(t)} \bm{q}_{n^\prime, l}^{(t)\top} \Big) \\
\end{split}
\end{equation}
for \(n \ne \overline{n}\),

\begin{equation}
\begin{split}
    \label{q_p_k_m_dy}
    &\langle \bm{q}_+^{(t+1)}, \bm{k}_-^{(t+1)} \rangle - \langle \bm{q}_+^{(t)}, \bm{k}_-^{(t)} \rangle \\
    &= \langle \Delta \bm{q}_+^{(t)}, \bm{k}_-^{(t)} \rangle + \langle \bm{q}_+^{(t)}, \Delta \bm{k}_-^{(t)} \rangle + \langle \Delta \bm{q}_+^{(t)}, \Delta \bm{k}_-^{(t)} \rangle \\
    &= \alpha_{+, +}^{(t)} \langle \bm{k}_+^{(t)}, \bm{k}_-^{(t)} \rangle + \sum\limits_{n \in S_+} \sum\limits_{i=2}^M \alpha_{n, +, i}^{(t)} \langle \bm{k}_{n, i}^{(t)}, \bm{k}_-^{(t)} \rangle \\
    &+ \beta_{-, -}^{(t)} \langle \bm{q}_+^{(t)}, \bm{q}_-^{(t)} \rangle + \sum\limits_{n \in S_-} \sum\limits_{i=2}^M \beta_{n, -, i}^{(t)} \langle \bm{q}_{n, i}^{(t)}, \bm{q}_+^{(t)} \rangle \\
    &+ \Big( \alpha_{+, +}^{(t)} \bm{k}_+^{(t)} + \sum\limits_{n \in S_+} \sum\limits_{i=2}^M \alpha_{n, +, i}^{(t)} \bm{k}_{n, i}^{(t)} \Big) \\
    &\cdot \Big( \beta_{-, -}^{(t)} \bm{q}_-^{(t)\top} + \sum\limits_{n \in S_-} \sum\limits_{i=2}^M \beta_{n, -, i}^{(t)} \bm{q}_{n, i}^{(t)\top} \Big),
\end{split}
\end{equation}

\begin{equation}
\begin{split}
    \label{k_p_k_m_dy}
    &\langle \bm{k}_+^{(t+1)}, \bm{k}_-^{(t+1)} \rangle - \langle \bm{k}_+^{(t)}, \bm{k}_-^{(t)} \rangle \\
    &= \langle \Delta \bm{k}_+^{(t)}, \bm{k}_-^{(t)} \rangle + \langle \bm{k}_+^{(t)}, \Delta \bm{k}_-^{(t)} \rangle + \langle \Delta \bm{k}_+^{(t)}, \Delta \bm{k}_-^{(t)} \rangle \\
    &= \beta_{+, +}^{(t)} \langle \bm{q}_+^{(t)}, \bm{k}_-^{(t)} \rangle + \sum\limits_{n \in S_+} \sum\limits_{i=2}^M \beta_{n, +, i}^{(t)} \langle \bm{q}_{n, i}^{(t)}, \bm{k}_-^{(t)} \rangle \\
    &+ \beta_{-, -}^{(t)} \langle \bm{q}_-^{(t)}, \bm{k}_+^{(t)} \rangle + \sum\limits_{n \in S_-} \sum\limits_{i=2}^M \beta_{n, -, i}^{(t)} \langle \bm{q}_{n, i}^{(t)}, \bm{k}_+^{(t)} \rangle \\
    &+ \Big( \beta_{+, +}^{(t)} \bm{q}_+^{(t)} + \sum\limits_{n \in S_+} \sum\limits_{i=2}^M \beta_{n, +, i}^{(t)} \bm{q}_{n, i}^{(t)} \Big) \\
    &\cdot \Big( \beta_{-, -}^{(t)} \bm{q}_-^{(t)\top} + \sum\limits_{n \in S_-} \sum\limits_{i=2}^M \beta_{n, -, i}^{(t)} \bm{q}_{n, i}^{(t)\top} \Big).
\end{split}
\end{equation}

\subsection{Proof of Lemma \ref{inner_products_hold_magnitude}}
\label{proof4lemmaC4}
Let \(T_0 = O(\frac{1}{\eta d_h^{\frac{1}{4}} \Vert \bm{\mu} \Vert_2^2 \Vert \bm{w}_O \Vert_2^2})\). By Lemma \ref{upper_bound_of_V}, we have \( \vert V_+^{(t)} \vert, \vert V_-^{(t)} \vert, \vert V_{n, i}^{(t)} \vert = O (d_h^{-\frac{1}{4}}) \) for \(t \in [0, T_0]\) by Lemma \ref{upper_bound_of_V}. Plugging this into the expression for \(\alpha\) and \(\beta\) gives

\begin{align*}
    &\vert \alpha_{+, +}^{(t)} \vert = \Big\vert \frac{\eta}{NM} \sum\limits_{n \in S_+} - \ell_n^{\prime(t)} \Vert \bm{\mu} \Vert_2^2 \\
    &\cdot \Big( V_+^{(t)} \big( \frac{ \exp (\langle \bm{q}_+^{(t)}, \bm{k}_+^{(t)} \rangle)}{ \exp  (\langle \bm{q}_+^{(t)}, \bm{k}_+^{(t)} \rangle) + \sum\limits_{j=2}^M  \exp  ( \langle \bm{q}_+^{(t)}, \bm{k}_{n, j}^{(t)} \rangle ) } \\
    & - (\frac{ \exp (\langle \bm{q}_+^{(t)}, \bm{k}_+^{(t)} \rangle)}{ \exp  (\langle \bm{q}_+^{(t)}, \bm{k}_+^{(t)} \rangle) + \sum\limits_{j=2}^M  \exp  ( \langle \bm{q}_+^{(t)}, \bm{k}_{n, j}^{(t)} \rangle ) })^2 \big) \\
    & - \sum\limits_{i=2}^M \big( V_{n, i}^{(t)} \cdot \frac{ \exp (\langle \bm{q}_+^{(t)}, \bm{k}_+^{(t)} \rangle)}{ \exp  (\langle \bm{q}_+^{(t)}, \bm{k}_+^{(t)} \rangle) + \sum\limits_{j=2}^M  \exp  ( \langle \bm{q}_+^{(t)}, \bm{k}_{n, j}^{(t)} \rangle ) } \\
    & \cdot \frac{ \exp (\langle \bm{q}_+^{(t)}, \bm{k}_{n, i}^{(t)} \rangle)}{ \exp  (\langle \bm{q}_+^{(t)}, \bm{k}_+^{(t)} \rangle) + \sum\limits_{j=2}^M  \exp  ( \langle \bm{q}_+^{(t)}, \bm{k}_{n, j}^{(t)} \rangle ) } \big) \Big) \Big\vert \\
    &\le \frac{\eta \Vert \bm{\mu} \Vert_2^2}{NM} \cdot \frac{3NM}{4} \cdot O(d_h^{-\frac{1}{4}}) \\
    &= O(\frac{\eta \Vert \bm{\mu} \Vert_2^2}{d_h^{\frac{1}{4}}}),
\end{align*}
where the inequality is by \( - \ell_n^{\prime(t)} \le 1 \) and the property that attention is smaller than 1 ( e.g. \( \frac{ \exp (\langle \bm{q}_+^{(t)}, \bm{k}_+^{(t)} \rangle)}{ \exp  (\langle \bm{q}_+^{(t)}, \bm{k}_+^{(t)} \rangle) + \sum\limits_{j=2}^M  \exp  ( \langle \bm{q}_+^{(t)}, \bm{k}_{n, j}^{(t)} \rangle ) } \le 1 \) ). We also have

\begin{align*}
    &\vert \alpha_{n, +, i}^{(t)} \vert = \Big\vert - \frac{\eta}{NM} \ell_n^{\prime(t)} \Vert \bm{\mu} \Vert_2^2 \\
    &\cdot \Big( - V_+^{(t)} \cdot \frac{ \exp (\langle \bm{q}_+^{(t)}, \bm{k}_+^{(t)} \rangle)}{ \exp  (\langle \bm{q}_+^{(t)}, \bm{k}_+^{(t)} \rangle) + \sum\limits_{j=2}^M  \exp  ( \langle \bm{q}_+^{(t)}, \bm{k}_{n, j}^{(t)} \rangle ) } \\
    &\cdot \frac{ \exp (\langle \bm{q}_+^{(t)}, \bm{k}_{n, i}^{(t)} \rangle)}{ \exp  (\langle \bm{q}_+^{(t)}, \bm{k}_+^{(t)} \rangle) + \sum\limits_{j=2}^M  \exp  ( \langle \bm{q}_+^{(t)}, \bm{k}_{n, j}^{(t)} \rangle ) } \\
    &+ V_{n, i}^{(t)} \big( \frac{ \exp (\langle \bm{q}_+^{(t)}, \bm{k}_{n, i}^{(t)} \rangle)}{ \exp  (\langle \bm{q}_+^{(t)}, \bm{k}_+^{(t)} \rangle) + \sum\limits_{j=2}^M  \exp  ( \langle \bm{q}_+^{(t)}, \bm{k}_{n, j}^{(t)} \rangle ) } \\
    &- (\frac{ \exp (\langle \bm{q}_+^{(t)}, \bm{k}_{n, i}^{(t)} \rangle)}{ \exp  (\langle \bm{q}_+^{(t)}, \bm{k}_+^{(t)} \rangle) + \sum\limits_{j=2}^M  \exp  ( \langle \bm{q}_+^{(t)}, \bm{k}_{n, j}^{(t)} \rangle ) })^2 \big) \\
    &- \sum\limits_{k \ne i} \big( V_{n, k}^{(t)} \cdot \frac{ \exp (\langle \bm{q}_+^{(t)}, \bm{k}_{n, i}^{(t)} \rangle)}{ \exp  (\langle \bm{q}_+^{(t)}, \bm{k}_+^{(t)} \rangle) + \sum\limits_{j=2}^M  \exp  ( \langle \bm{q}_+^{(t)}, \bm{k}_{n, j}^{(t)} \rangle ) } \\
    &\cdot \frac{ \exp (\langle \bm{q}_+^{(t)}, \bm{k}_{n, k}^{(t)} \rangle)}{ \exp  (\langle \bm{q}_+^{(t)}, \bm{k}_+^{(t)} \rangle) + \sum\limits_{j=2}^M  \exp  ( \langle \bm{q}_+^{(t)}, \bm{k}_{n, j}^{(t)} \rangle ) } \big) \Big) \Big\vert \\
    &\le \frac{\eta \Vert \bm{\mu} \Vert_2^2}{NM} \cdot M \cdot O(d_h^{-\frac{1}{4}}) \\
    &= O(\frac{\eta \Vert \bm{\mu} \Vert_2^2}{d_h^{\frac{1}{4}} N}),
\end{align*}
where the inequality is by \( - \ell_n^{\prime(t)} \le 1 \) and the property that attention is smaller than 1. We also have

\begin{align*}
    &\vert \alpha_{n^\prime, i^\prime, +}^{(t)} \vert = \Big\vert \frac{\eta}{NM} \sum\limits_{n \in S_+} - \ell_n^{\prime(t)} \sum\limits_{i=2}^M \langle \bm{\xi}_{n^\prime, i^\prime}, \bm{\xi}_{n, i} \rangle \\
    &\cdot \Big( V_+^{(t)} \big( \frac{ \exp (\langle \bm{q}_{n, i}^{(t)}, \bm{k}_+^{(t)} \rangle)}{ \exp  (\langle \bm{q}_{n, i}^{(t)}, \bm{k}_+^{(t)} \rangle) + \sum\limits_{j=2}^M  \exp  ( \langle \bm{q}_{n, i}^{(t)}, \bm{k}_{n, j}^{(t)} \rangle ) } \\
    & - (\frac{ \exp (\langle \bm{q}_{n, i}^{(t)}, \bm{k}_+^{(t)} \rangle)}{ \exp  (\langle \bm{q}_{n, i}^{(t)}, \bm{k}_+^{(t)} \rangle) + \sum\limits_{j=2}^M  \exp  ( \langle \bm{q}_{n, i}^{(t)}, \bm{k}_{n, j}^{(t)} \rangle ) })^2 \big) \\
    & -\sum\limits_{k=2}^M \big( V_{n, i}^{(t)} \cdot \frac{ \exp (\langle \bm{q}_{n, i}^{(t)}, \bm{k}_+^{(t)} \rangle)}{ \exp  (\langle \bm{q}_{n, i}^{(t)}, \bm{k}_+^{(t)} \rangle) + \sum\limits_{j=2}^M  \exp  ( \langle \bm{q}_{n, i}^{(t)}, \bm{k}_{n, j}^{(t)} \rangle ) } \\
    & \cdot \frac{ \exp (\langle \bm{q}_{n, i}^{(t)}, \bm{k}_{n, k}^{(t)} \rangle)}{ \exp  (\langle \bm{q}_{n, i}^{(t)}, \bm{k}_+^{(t)} \rangle) + \sum\limits_{j=2}^M  \exp  ( \langle \bm{q}_{n, i}^{(t)}, \bm{k}_{n, j}^{(t)} \rangle ) } \big) \Big) \Big\vert \\
    &\le \frac{\eta}{NM} \Big( \frac{3 \tilde{\sigma}_p^2 d}{2} + \frac{3N}{4} \cdot \tilde{\sigma}_p^2 \cdot \sqrt{d \log (4N^2M^2/\delta)} \Big) \cdot M \cdot O(d_h^{-\frac{1}{4}}) \\
    &= O( \frac{\eta \sigma_p^2 d}{d_h^{\frac{1}{4}}N} ),
\end{align*}
where the inequality is by \( - \ell_n^{\prime(t)} \le 1 \), Lemma \ref{caoyuan} and the property that attention is smaller than 1.

Similarly, we have
\[
    \vert \alpha_{-, -}^{(t)} \vert, \vert \beta_{+, +}^{(t)} \vert, \vert \beta_{-, -}^{(t)} \vert = O \Big( \frac{\eta \Vert \bm{\mu} \Vert_2^2}{d_h^{\frac{1}{4}}} \Big),
\]
\[
    \vert \alpha_{n, -, l}^{(t)} \vert, \vert \beta_{n, +, l}^{(t)} \vert, \vert \beta_{n, -, l}^{(t)} \vert = O \Big( \frac{\eta \Vert \bm{\mu} \Vert_2^2}{d_h^{\frac{1}{4}}N} \Big),
\]
\[
    \vert \alpha_{n, l, -}^{(t)} \vert, \vert \beta_{n, l, +}^{(t)} \vert, \vert \beta_{n, l, -}^{(t)} \vert, \vert \alpha_{n, l, n^\prime, l^\prime}^{(t)} \vert, \vert \beta_{n, l, n^\prime, l^\prime}^{(t)} \vert = O \Big( \frac{\eta \sigma_p^2 d}{d_h^{\frac{1}{4}}N} \Big)
\]
for \(t \in [0, T_0]\).

Next we use induction to show that the following proposition \(\mathcal{A}(t)\) holds for \(t \in [0, T_0]\)

\(\mathcal{A}(t):\)
\begin{align*}
    &\vert \langle \bm{q}_\pm^{(t)}, \bm{k}_\pm^{(t)} \rangle \vert, \vert \langle \bm{q}_{n, i}^{(t)}, \bm{k}_\pm^{(t)} \rangle \vert, \vert \langle \bm{q}_\pm^{(t)}, \bm{k}_{n, j}^{(t)} \rangle \vert, \vert \langle \bm{q}_{n, i}^{(t)}, \bm{k}_{n^\prime, j}^{(t)} \rangle \vert \\
    &= O \Big( \max \{ \Vert \bm{\mu} \Vert_2^2, \sigma_p^2 d \} \cdot \sigma_h^2 \cdot \sqrt{d_h \log (6N^2M^2/\delta)} \Big),
\end{align*}
\begin{align*}
    &\vert \langle \bm{q}_\pm^{(t)}, \bm{q}_\mp^{(t)} \rangle \vert, \vert \langle \bm{q}_{n, i}^{(t)}, \bm{q}_\pm^{(t)} \rangle \vert, \vert \langle \bm{q}_{n, i}^{(t)}, \bm{q}_{n^\prime, j}^{(t)} \rangle \vert \\
    &= O \Big( \max \{ \Vert \bm{\mu} \Vert_2^2, \sigma_p^2 d \} \cdot \sigma_h^2 \cdot \sqrt{d_h \log (6N^2M^2/\delta)} \Big),
\end{align*}
\begin{align*}
    &\vert \langle \bm{k}_\pm^{(t)}, \bm{k}_\mp^{(t)} \rangle \vert, \vert \langle \bm{k}_{n, i}^{(t)}, \bm{k}_\pm^{(t)} \rangle \vert, \vert \langle \bm{k}_{n, i}^{(t)}, \bm{k}_{n^\prime, j}^{(t)} \rangle \vert \\
    &= O \Big( \max \{ \Vert \bm{\mu} \Vert_2^2, \sigma_p^2 d \} \cdot \sigma_h^2 \cdot \sqrt{d_h \log (6N^2M^2/\delta)} \Big),
\end{align*}
\[
    \Vert \bm{q}_\pm^{(t)} \Vert_2^2, \Vert \bm{k}_\pm^{(t)} \Vert_2^2 = \Theta (\Vert \bm{\mu} \Vert_2^2 \sigma_h^2 d_h)
\]
\[
    \Vert \bm{q}_{n, i}^{(t)} \Vert_2^2, \Vert \bm{k}_{n, i}^{(t)} \Vert_2^2 = \Theta ( \sigma_p^2 \sigma_h^2 d d_h )
\]
for \( i, j \in [M] \backslash \{1\} \), \( n, n^\prime \in [N] \).

By Lemma \ref{initialization_QK} we know that \(\mathcal{A}(0)\) is true. Now we assume \(\mathcal{A}(0), \dots, \mathcal{A}(T)\) is true, then we need to proof that \(\mathcal{A}(T+1)\) is true. We first proof \(\vert \langle \bm{q}_+^{(T+1)}, \bm{k}_+^{(T+1)} \rangle \vert = O \Big( \max \{ \Vert \bm{\mu} \Vert_2^2, \sigma_p^2 d \} \cdot \sigma_h^2 \cdot \sqrt{d_h \log (6N^2M^2/\delta)} \Big) \), as an example.

\begin{equation}
\begin{split}
    &\big\vert \langle \bm{q}_+^{(t+1)}, \bm{k}_+^{(t+1)} \rangle - \langle \bm{q}_+^{(t)}, \bm{k}_+^{(t)} \rangle \big\vert \\
    &= \Big\vert \alpha_{+, +}^{(t)} \Vert \bm{k}_+^{(t)} \Vert_2^2 + \sum\limits_{n \in S_+} \sum\limits_{i=2}^M \alpha_{n, +, i}^{(t)} \langle \bm{k}_+^{(t)}, \bm{k}_{n, i}^{(t)} \rangle \\
    &+ \beta_{+, +}^{(t)} \Vert \bm{q}_+^{(t)} \Vert_2^2 + \sum\limits_{n \in S_+} \sum\limits_{i=2}^M \beta_{n, +, i}^{(t)} \langle \bm{q}_+^{(t)}, \bm{q}_{n, i}^{(t)} \rangle \\
    &+ \Big( \alpha_{+, +}^{(t)} \bm{k}_+^{(t)} + \sum\limits_{n \in S_+} \sum\limits_{i=2}^M \alpha_{n, +, i}^{(t)} \bm{k}_{n, i}^{(t)} \Big) \\
    &\cdot \Big( \beta_{+, +}^{(t)} \bm{q}_+^{(t)\top} + \sum\limits_{n \in S_+} \sum\limits_{i=2}^M \beta_{n, +, i}^{(t)} \bm{q}_{n, i}^{(t)\top} \Big) \Big\vert \\
    &\le O \Big( \frac{\eta \Vert \bm{\mu} \Vert_2^2}{d_h^{\frac{1}{4}}} \Big) \cdot \Theta (\Vert \bm{\mu} \Vert_2^2 \sigma_h^2 d_h) \\
    &+ NM \cdot O \Big( \frac{\eta \Vert \bm{\mu} \Vert_2^2}{d_h^{\frac{1}{4}}N} \Big) \cdot O \Big( \max \{ \Vert \bm{\mu} \Vert_2^2, \sigma_p^2 d \} \cdot \sigma_h^2 \cdot \sqrt{d_h \log (6N^2M^2/\delta)} \Big) \\
    &+ \{ lower \ order \ term \} \\
    &= O \Big( \eta \Vert \bm{\mu} \Vert_2^4 \sigma_h^2 d_h^{\frac{3}{4}} \Big)
\end{split}
\end{equation}

Taking a summation, we obtain that

\begin{equation}
\begin{split}
    &\vert \langle \bm{q}_+^{(T+1)}, \bm{k}_+^{(T+1)} \rangle \vert \le \vert \langle \bm{q}_+^{(0)}, \bm{k}_+^{(0)} \rangle \vert + \sum\limits_{t = 0}^T \big\vert \langle \bm{q}_+^{(t+1)}, \bm{k}_+^{(t+1)} \rangle - \langle \bm{q}_+^{(t)}, \bm{k}_+^{(t)} \rangle \big\vert \\
    &\le \vert \langle \bm{q}_+^{(0)}, \bm{k}_+^{(0)} \rangle \vert + \sum\limits_{t = 0}^{T_0 - 1} \big\vert \langle \bm{q}_+^{(t+1)}, \bm{k}_+^{(t+1)} \rangle - \langle \bm{q}_+^{(t)}, \bm{k}_+^{(t)} \rangle \big\vert \\
    &\le O \Big( \max \{ \Vert \bm{\mu} \Vert_2^2, \sigma_p^2 d \} \cdot \sigma_h^2 \cdot \sqrt{d_h \log (6N^2M^2/\delta)} \Big) + O \Big( \frac{1}{\eta d_h^{\frac{1}{4}} \Vert \bm{\mu} \Vert_2^2 \Vert \bm{w}_O \Vert_2^2} \Big) \cdot O \Big( \eta \Vert \bm{\mu} \Vert_2^4 \sigma_h^2 d_h^{\frac{3}{4}} \Big) \\
    &= O \Big( \max \{ \Vert \bm{\mu} \Vert_2^2, \sigma_p^2 d \} \cdot \sigma_h^2 \cdot \sqrt{d_h \log (6N^2M^2/\delta)} \Big) + O \Big( \Vert \bm{\mu} \Vert_2^2 \sigma_h^2 d_h^{\frac{1}{2}} \Big) \\
    &= O \Big( \max \{ \Vert \bm{\mu} \Vert_2^2, \sigma_p^2 d \} \cdot \sigma_h^2 \cdot \sqrt{d_h \log (6N^2M^2/\delta)} \Big)
\end{split}
\end{equation}

Similarly to \(\langle \bm{q}_+^{(t)}, \bm{k}_+^{(t)} \rangle\), it is easy to know that the inner product does not change by a magnitude more than the product of \(\max\{\alpha, \beta\}\) and \(\max\{ \langle \bm{q}, \bm{q} \rangle, \langle \bm{k}, \bm{k} \rangle \}\) in a single iteration, which can be expressed as follows

\begin{equation}
\begin{split}
    &\big\vert \langle \bm{q}^{(t+1)}, \bm{k}^{(t+1)} \rangle - \langle \bm{q}^{(t)}, \bm{k}^{(t)} \rangle \big\vert \\
    &= O \Big( \max \{ \frac{\eta \Vert \bm{\mu} \Vert_2^2}{d_h^{\frac{1}{4}}}, \frac{\eta \sigma_p^2 d}{d_h^{\frac{1}{4}}N} \} \Big) \cdot \Theta ( \max \{ \Vert \bm{\mu} \Vert_2^2 \sigma_h^2 d_h, \sigma_p^2 \sigma_h^2 d d_h \}) \\
    &= O \Big( \frac{\eta \Vert \bm{\mu} \Vert_2^2}{d_h^{\frac{1}{4}}} \Big) \cdot \Theta ( \max \{ \Vert \bm{\mu} \Vert_2^2 \sigma_h^2 d_h, \sigma_p^2 \sigma_h^2 d d_h \}) \\
    &= O \Big( \eta \Vert \bm{\mu} \Vert_2^2 \sigma_h^2 d_h^{\frac{3}{4}} \cdot \max \{ \Vert \bm{\mu} \Vert_2^2, \sigma_p^2 d \} \Big)
\end{split}
\end{equation}
where the second equality is by the condition that \(N \cdot \mathrm{SNR}^2 = \Omega(1)\).

Taking a summation, we obtain that
\begin{equation}
\begin{split}
    &\big\vert \langle \bm{q}^{(T+1)}, \bm{k}^{(T+1)} \rangle - \langle \bm{q}^{(0)}, \bm{k}^{(0)} \rangle \big\vert \le \sum\limits_{t = 0}^{T - 1} \big\vert \langle \bm{q}^{(t+1)}, \bm{k}^{(t+1)} \rangle - \langle \bm{q}^{(t)}, \bm{k}^{(t)} \rangle \big\vert \\
    &\le \sum\limits_{t = 0}^{T_0 - 1} O \Big( \eta \Vert \bm{\mu} \Vert_2^2 \sigma_h^2 d_h^{\frac{3}{4}} \cdot \max \{ \Vert \bm{\mu} \Vert_2^2, \sigma_p^2 d \} \Big) \\
    &= O \Big( \frac{1}{\eta d_h^{\frac{1}{4}} \Vert \bm{\mu} \Vert_2^2 \Vert \bm{w}_O \Vert_2^2} \Big) \cdot O \Big( \eta \Vert \bm{\mu} \Vert_2^2 \sigma_h^2 d_h^{\frac{3}{4}} \cdot \max \{ \Vert \bm{\mu} \Vert_2^2, \sigma_p^2 d \} \Big) \\
    &= O \Big( \max \{ \Vert \bm{\mu} \Vert_2^2, \sigma_p^2 d \} \cdot \sigma_h^2 d_h^{\frac{1}{2}} \Big).
\end{split}
\end{equation}
It is clear that the magnitude of \(\langle \bm{q}^{(T+1)}, \bm{k}^{(T+1)} \rangle - \langle \bm{q}^{(0)}, \bm{k}^{(0)} \rangle\) is smaller than \( \max \{ \Vert \bm{\mu} \Vert_2^2, \sigma_p^2 d \} \cdot \sigma_h^2 \cdot \sqrt{d_h \log (6N^2M^2/\delta)} \), 
Thus the magnitude of the bound for \(\langle \bm{q}^{(T+1)}, \bm{k}^{(T+1)} \rangle\) is the same as that of \(\langle \bm{q}^{(T)}, \bm{k}^{(T)} \rangle\). The proof for \(\langle \bm{q}^{(T + 1)}, \bm{q}^{(T + 1)} \rangle\) and \(\langle \bm{k}^{(T + 1)}, \bm{k}^{(T + 1)} \rangle\) is exactly the same, and we can conclude the proof by an induction.

\subsection{Lower Bounds of \(\alpha\) and \(\beta\)}
\label{bound_of_alpha_beta}
In this subsection, we present some bounds for \(\alpha\) and \(\beta\) which can be used in \ref{stage2} and \ref{stage3}. All the calculations in this subsection are based on the precise expression for \(\alpha\) and \(\beta\) in \ref{alpha_and_beta} and assume that \( \mathcal{B}(T_1), \dots, \mathcal{B}(s), \mathcal{D}(T_1), \dots, \mathcal{D}(s - 1) \) hold (\(s \in [T_1, t]\)). Then the following propositions hold:
\[
    V_+^{(s)} \ge 3 M \cdot \vert V_{n, i}^{(s)} \vert,
\]
\[
    V_-^{(s)} \le - 3 M \cdot \vert V_{n, i}^{(s)} \vert,
\]
\[
    softmax(\langle \bm{q}_\pm^{(s)}, \bm{k}_\pm^{(s)} \rangle), softmax(\langle \bm{q}_{n, i}^{(s)}, \bm{k}_\pm^{(s)} \rangle) \ge \frac{1}{M} - o(1),
\]
\[
    softmax( \langle \bm{q}_\pm^{(s)}, \bm{k}_{n, j}^{(s)} \rangle ) , softmax( \langle \bm{q}_{n, i}^{(s)}, \bm{k}_{n, j}^{(s)} \rangle ) \le \frac{1}{M} + o(1).
\]
Now we give the bounds respectively for \( \alpha_{+, +}^{(s)} \), \( \alpha_{n, +, i}^{(s)} \), \( \alpha_{-, -}^{(s)} \), \( \alpha_{n, -, i}^{(s)} \), \( \alpha_{n, i, +}^{(s)} \), \( \alpha_{n, i, -}^{(s)} \), \( \alpha_{n, i, n^\prime, i^\prime}^{(s)} \), 
\( \beta_{+, +}^{(s)} \), \( \beta_{n, +, i}^{(s)} \), \( \beta_{-, -}^{(s)} \), \( \beta_{n, -, i}^{(s)} \), \( \beta_{n, i, +}^{(s)} \), \( \beta_{n, i, -}^{(s)} \), \( \beta_{n, i, n^\prime, i^\prime}^{(s)} \).

\begin{equation}
\begin{split}
    &\alpha_{+, +}^{(s)} 
    = \frac{\eta}{NM} \sum\limits_{n \in S_+} - \ell_n^{\prime(s)} \Vert \bm{\mu} \Vert_2^2 \cdot \frac{ \exp (\langle \bm{q}_+^{(s)}, \bm{k}_+^{(s)} \rangle)}{ \exp  (\langle \bm{q}_+^{(s)}, \bm{k}_+^{(s)} \rangle) + \sum\limits_{j=2}^M  \exp  ( \langle \bm{q}_+^{(s)}, \bm{k}_{n, j}^{(s)} \rangle ) } \\
    &\cdot \Big( V_+^{(s)} \big( 1 - \frac{ \exp (\langle \bm{q}_+^{(s)}, \bm{k}_+^{(s)} \rangle)}{ \exp  (\langle \bm{q}_+^{(s)}, \bm{k}_+^{(s)} \rangle) + \sum\limits_{j=2}^M  \exp  ( \langle \bm{q}_+^{(s)}, \bm{k}_{n, j}^{(s)} \rangle ) } \big) \\
    & - \sum\limits_{i=2}^M \big( V_{n, i}^{(s)} \cdot \frac{ \exp (\langle \bm{q}_+^{(s)}, \bm{k}_{n, i}^{(s)} \rangle)}{ \exp  (\langle \bm{q}_+^{(s)}, \bm{k}_+^{(s)} \rangle) + \sum\limits_{j=2}^M  \exp  ( \langle \bm{q}_+^{(s)}, \bm{k}_{n, j}^{(s)} \rangle ) } \big) \Big) \\
    &\ge \frac{\eta}{NM} \sum\limits_{n \in S_+} - \ell_n^{\prime(s)} \Vert \bm{\mu} \Vert_2^2 \cdot \frac{ \exp (\langle \bm{q}_+^{(s)}, \bm{k}_+^{(s)} \rangle)}{ \exp  (\langle \bm{q}_+^{(s)}, \bm{k}_+^{(s)} \rangle) + \sum\limits_{j=2}^M  \exp  ( \langle \bm{q}_+^{(s)}, \bm{k}_{n, j}^{(s)} \rangle ) } \\
    &\cdot \Big( V_+^{(s)} \big( 1 - \frac{ \exp (\langle \bm{q}_+^{(s)}, \bm{k}_+^{(s)} \rangle)}{ \exp  (\langle \bm{q}_+^{(s)}, \bm{k}_+^{(s)} \rangle) + \sum\limits_{j=2}^M  \exp  ( \langle \bm{q}_+^{(s)}, \bm{k}_{n, j}^{(s)} \rangle ) } \big) \\
    & - \frac{1}{2} \cdot V_+^{(s)} \sum\limits_{i=2}^M \cdot \frac{ \exp (\langle \bm{q}_+^{(s)}, \bm{k}_{n, i}^{(s)} \rangle)}{ \exp  (\langle \bm{q}_+^{(s)}, \bm{k}_+^{(s)} \rangle) + \sum\limits_{j=2}^M  \exp  ( \langle \bm{q}_+^{(s)}, \bm{k}_{n, j}^{(s)} \rangle ) } \Big) \\
    &\ge \frac{\eta}{2NM} \sum\limits_{n \in S_+} - \ell_n^{\prime(s)} \Vert \bm{\mu} \Vert_2^2 V_+^{(s)} \cdot \frac{ \exp (\langle \bm{q}_+^{(s)}, \bm{k}_+^{(s)} \rangle)}{ \exp  (\langle \bm{q}_+^{(s)}, \bm{k}_+^{(s)} \rangle) + \sum\limits_{j=2}^M  \exp  ( \langle \bm{q}_+^{(s)}, \bm{k}_{n, j}^{(s)} \rangle ) } \\
    &\cdot \big( 1 - \frac{ \exp (\langle \bm{q}_+^{(s)}, \bm{k}_+^{(s)} \rangle)}{ \exp  (\langle \bm{q}_+^{(s)}, \bm{k}_+^{(s)} \rangle) + \sum\limits_{j=2}^M  \exp  ( \langle \bm{q}_+^{(s)}, \bm{k}_{n, j}^{(s)} \rangle ) } \big) \\
\end{split}
\end{equation}
where the first inequality is by \( V_+^{(s)} \ge 3M \cdot \vert V_{n, i}^{(s)} \vert \), the second inequality is by the fact that the sum of attention equal to 1.
Similarly, we have
\begin{equation}
\begin{split}
    &\beta_{+, +}^{(s)} \ge \\
    &\frac{\eta}{2NM} \sum\limits_{n \in S_+} - \ell_n^{\prime(s)} \Vert \bm{\mu} \Vert_2^2 V_+^{(s)} \cdot \frac{ \exp (\langle \bm{q}_+^{(s)}, \bm{k}_+^{(s)} \rangle)}{ \exp  (\langle \bm{q}_+^{(s)}, \bm{k}_+^{(s)} \rangle) + (M - 1)  \exp  ( \max\limits_{j} \{ \langle \bm{q}_+^{(s)}, \bm{k}_{n, j}^{(s)} \rangle \} ) } \\
    &\cdot \frac{ \exp ( \max\limits_{j} \{ \langle \bm{q}_+^{(s)}, \bm{k}_{n, j}^{(s)} \rangle \} ) }{ \exp  (\langle \bm{q}_+^{(s)}, \bm{k}_+^{(s)} \rangle) + (M - 1) \exp  ( \max\limits_{j} \{ \langle \bm{q}_+^{(s)}, \bm{k}_{n, j}^{(s)} \rangle \} ) },
\end{split}
\end{equation}
\begin{equation}
\begin{split}
    &\alpha_{-, -}^{(s)} \ge \\
    &\frac{\eta}{2NM} \sum\limits_{n \in S_-} \ell_n^{\prime(s)} \Vert \bm{\mu} \Vert_2^2 V_-^{(s)} \cdot \frac{ \exp (\langle \bm{q}_-^{(s)}, \bm{k}_-^{(s)} \rangle)}{ \exp  (\langle \bm{q}_-^{(s)}, \bm{k}_-^{(s)} \rangle) + (M - 1)  \exp  ( \max\limits_{j} \{ \langle \bm{q}_-^{(s)}, \bm{k}_{n, j}^{(s)} \rangle \} ) } \\
    &\cdot \frac{ \exp ( \max\limits_{j} \{ \langle \bm{q}_-^{(s)}, \bm{k}_{n, j}^{(s)} \rangle \} ) }{ \exp  (\langle \bm{q}_-^{(s)}, \bm{k}_-^{(s)} \rangle) + (M - 1) \exp  ( \max\limits_{j} \{ \langle \bm{q}_-^{(s)}, \bm{k}_{n, j}^{(s)} \rangle \} ) },
\end{split}
\end{equation}
\begin{equation}
\begin{split}
    &\beta_{-, -}^{(s)} \ge \\
    &\frac{\eta}{2NM} \sum\limits_{n \in S_-} \ell_n^{\prime(s)} \Vert \bm{\mu} \Vert_2^2 V_-^{(s)} \cdot \frac{ \exp (\langle \bm{q}_-^{(s)}, \bm{k}_-^{(s)} \rangle)}{ \exp  (\langle \bm{q}_-^{(s)}, \bm{k}_-^{(s)} \rangle) + (M - 1)  \exp  ( \max\limits_{j} \{ \langle \bm{q}_-^{(s)}, \bm{k}_{n, j}^{(s)} \rangle \} ) } \\
    &\cdot \frac{ \exp ( \max\limits_{j} \{ \langle \bm{q}_-^{(s)}, \bm{k}_{n, j}^{(s)} \rangle \} ) }{ \exp  (\langle \bm{q}_-^{(s)}, \bm{k}_-^{(s)} \rangle) + (M - 1) \exp  ( \max\limits_{j} \{ \langle \bm{q}_-^{(s)}, \bm{k}_{n, j}^{(s)} \rangle \} ) }.
\end{split}
\end{equation}

\begin{align*}
    &\alpha_{n, +, i}^{(s)} = - \frac{\eta}{NM} \ell_n^{\prime(s)} \Vert \bm{\mu} \Vert_2^2 \cdot \frac{ \exp (\langle \bm{q}_+^{(s)}, \bm{k}_{n, i}^{(s)} \rangle)}{ \exp  (\langle \bm{q}_+^{(s)}, \bm{k}_+^{(s)} \rangle) + \sum\limits_{j=2}^M  \exp  ( \langle \bm{q}_+^{(s)}, \bm{k}_{n, j}^{(s)} \rangle ) } \\
    &\cdot \Big( - V_+^{(s)} \cdot \frac{ \exp (\langle \bm{q}_+^{(s)}, \bm{k}_+^{(s)} \rangle)}{ \exp  (\langle \bm{q}_+^{(s)}, \bm{k}_+^{(s)} \rangle) + \sum\limits_{j=2}^M  \exp  ( \langle \bm{q}_+^{(s)}, \bm{k}_{n, j}^{(s)} \rangle ) } \\
    &+ V_{n, i}^{(s)} \big( 1 - \frac{ \exp (\langle \bm{q}_+^{(s)}, \bm{k}_{n, i}^{(s)} \rangle)}{ \exp  (\langle \bm{q}_+^{(s)}, \bm{k}_+^{(s)} \rangle) + \sum\limits_{j=2}^M  \exp  ( \langle \bm{q}_+^{(s)}, \bm{k}_{n, j}^{(s)} \rangle ) } \big) \\
    &- \sum\limits_{k \ne i} \big( V_{n, k}^{(s)} \cdot \frac{ \exp (\langle \bm{q}_+^{(s)}, \bm{k}_{n, k}^{(s)} \rangle)}{ \exp  (\langle \bm{q}_+^{(s)}, \bm{k}_+^{(s)} \rangle) + \sum\limits_{j=2}^M  \exp  ( \langle \bm{q}_+^{(s)}, \bm{k}_{n, j}^{(s)} \rangle ) } \big) \Big) \\
    &\le - \frac{\eta}{NM} \ell_n^{\prime(s)} \Vert \bm{\mu} \Vert_2^2 \cdot \frac{ \exp (\langle \bm{q}_+^{(s)}, \bm{k}_{n, i}^{(s)} \rangle)}{ \exp  (\langle \bm{q}_+^{(s)}, \bm{k}_+^{(s)} \rangle) + \sum\limits_{j=2}^M  \exp  ( \langle \bm{q}_+^{(s)}, \bm{k}_{n, j}^{(s)} \rangle ) } \\
    &\cdot \Big( - V_+^{(s)} \cdot \frac{ \exp (\langle \bm{q}_+^{(s)}, \bm{k}_+^{(s)} \rangle)}{ \exp  (\langle \bm{q}_+^{(s)}, \bm{k}_+^{(s)} \rangle) + \sum\limits_{j=2}^M  \exp  ( \langle \bm{q}_+^{(s)}, \bm{k}_{n, j}^{(s)} \rangle ) } \\
    &+ \vert V_{n, i}^{(s)} \vert + \sum\limits_{k \ne i} \big( \vert V_{n, k}^{(s)} \vert \cdot (\frac{1}{M} + o(1)) \big) \Big) \\
    &\le - \frac{\eta}{NM} \ell_n^{\prime(s)} \Vert \bm{\mu} \Vert_2^2 \cdot \frac{ \exp (\langle \bm{q}_+^{(s)}, \bm{k}_{n, i}^{(s)} \rangle)}{ \exp  (\langle \bm{q}_+^{(s)}, \bm{k}_+^{(s)} \rangle) + \sum\limits_{j=2}^M  \exp  ( \langle \bm{q}_+^{(s)}, \bm{k}_{n, j}^{(s)} \rangle ) } \\
    &\cdot \Big( - V_+^{(s)} \cdot \frac{ \exp (\langle \bm{q}_+^{(s)}, \bm{k}_+^{(s)} \rangle)}{ \exp  (\langle \bm{q}_+^{(s)}, \bm{k}_+^{(s)} \rangle) + \sum\limits_{j=2}^M  \exp  ( \langle \bm{q}_+^{(s)}, \bm{k}_{n, j}^{(s)} \rangle ) } + 2 M \max\limits_l \vert V_{n, l}^{(s)} \vert \Big) \\ 
    &\le - \frac{\eta}{NM} \ell_n^{\prime(s)} \Vert \bm{\mu} \Vert_2^2 \cdot \frac{ \exp (\langle \bm{q}_+^{(s)}, \bm{k}_{n, i}^{(s)} \rangle)}{ \exp  (\langle \bm{q}_+^{(s)}, \bm{k}_+^{(s)} \rangle) + \sum\limits_{j=2}^M  \exp  ( \langle \bm{q}_+^{(s)}, \bm{k}_{n, j}^{(s)} \rangle ) } \\
    &\cdot \Big( - V_+^{(s)} \cdot \frac{ \exp (\langle \bm{q}_+^{(s)}, \bm{k}_+^{(s)} \rangle)}{ \exp  (\langle \bm{q}_+^{(s)}, \bm{k}_+^{(s)} \rangle) + \sum\limits_{j=2}^M  \exp  ( \langle \bm{q}_+^{(s)}, \bm{k}_{n, j}^{(s)} \rangle ) } \\
    &+ \frac{3}{4} V_+^{(s)} \frac{ \exp (\langle \bm{q}_+^{(s)}, \bm{k}_+^{(s)} \rangle)}{ \exp  (\langle \bm{q}_+^{(s)}, \bm{k}_+^{(s)} \rangle) + \sum\limits_{j=2}^M  \exp  ( \langle \bm{q}_+^{(s)}, \bm{k}_{n, j}^{(s)} \rangle ) } \Big) \\ 
    &= \frac{\eta}{4NM} \ell_n^{\prime(s)} \Vert \bm{\mu} \Vert_2^2 V_+^{(s)} \cdot \frac{ \exp (\langle \bm{q}_+^{(s)}, \bm{k}_{n, i}^{(s)} \rangle)}{ \exp  (\langle \bm{q}_+^{(s)}, \bm{k}_+^{(s)} \rangle) + \sum\limits_{j=2}^M  \exp  ( \langle \bm{q}_+^{(s)}, \bm{k}_{n, j}^{(s)} \rangle ) } \\
    &\cdot \frac{ \exp (\langle \bm{q}_+^{(s)}, \bm{k}_+^{(s)} \rangle)}{ \exp  (\langle \bm{q}_+^{(s)}, \bm{k}_+^{(s)} \rangle) + \sum\limits_{j=2}^M  \exp  ( \langle \bm{q}_+^{(s)}, \bm{k}_{n, j}^{(s)} \rangle ) },
\end{align*}
where the third inequality is by \( V_+^{(s)} \ge 3M \cdot \vert V_{n, i}^{(s)} \vert \) and \( softmax(\langle \bm{q}_+^{(s)}, \bm{k}_+^{(s)} \rangle) \ge \frac{1}{M} - o(1) \). Similarly, we have
\begin{align*}
    &\alpha_{n, -, i}^{(s)} \le \\
    &- \frac{\eta}{4NM} \ell_n^{\prime(s)} \Vert \bm{\mu} \Vert_2^2 V_-^{(s)} \cdot \frac{ \exp (\langle \bm{q}_-^{(s)}, \bm{k}_{n, i}^{(s)} \rangle)}{ \exp  (\langle \bm{q}_-^{(s)}, \bm{k}_-^{(s)} \rangle) + \sum\limits_{j=2}^M  \exp  ( \langle \bm{q}_-^{(s)}, \bm{k}_{n, j}^{(s)} \rangle ) } \\
    &\cdot \frac{ \exp (\langle \bm{q}_-^{(s)}, \bm{k}_-^{(s)} \rangle)}{ \exp  (\langle \bm{q}_-^{(s)}, \bm{k}_-^{(s)} \rangle) + \sum\limits_{j=2}^M  \exp  ( \langle \bm{q}_-^{(s)}, \bm{k}_{n, j}^{(s)} \rangle ) }.
\end{align*}

\begin{align*}
    &\beta_{n, +, i}^{(s)} = - \frac{\eta \Vert \bm{\mu} \Vert_2^2}{NM} \ell_n^{\prime(s)} \\
    &\cdot \Big( V_+^{(s)} \big( \frac{ \exp (\langle \bm{q}_{n, i}^{(s)}, \bm{k}_+^{(s)} \rangle)}{ \exp  (\langle \bm{q}_{n, i}^{(s)}, \bm{k}_+^{(s)} \rangle) + \sum\limits_{j=2}^M  \exp  ( \langle \bm{q}_{n, i}^{(s)}, \bm{k}_{n, j}^{(s)} \rangle ) } \\
    & - (\frac{ \exp (\langle \bm{q}_{n, i}^{(s)}, \bm{k}_+^{(s)} \rangle)}{ \exp  (\langle \bm{q}_{n, i}^{(s)}, \bm{k}_+^{(s)} \rangle) + \sum\limits_{j=2}^M  \exp  ( \langle \bm{q}_{n, i}^{(s)}, \bm{k}_{n, j}^{(s)} \rangle ) })^2 \big) \\
    & -\sum\limits_{k=2}^M \big( V_{n, i}^{(s)} \cdot \frac{ \exp (\langle \bm{q}_{n, i}^{(s)}, \bm{k}_+^{(s)} \rangle)}{ \exp  (\langle \bm{q}_{n, i}^{(s)}, \bm{k}_+^{(s)} \rangle) + \sum\limits_{j=2}^M  \exp  ( \langle \bm{q}_{n, i}^{(s)}, \bm{k}_{n, j}^{(s)} \rangle ) } \\
    & \cdot \frac{ \exp (\langle \bm{q}_{n, i}^{(s)}, \bm{k}_{n, k}^{(s)} \rangle)}{ \exp  (\langle \bm{q}_{n, i}^{(s)}, \bm{k}_+^{(s)} \rangle) + \sum\limits_{j=2}^M  \exp  ( \langle \bm{q}_{n, i}^{(s)}, \bm{k}_{n, j}^{(s)} \rangle ) } \big) \Big) \\
    &=  - \frac{\eta \Vert \bm{\mu} \Vert_2^2}{NM} \ell_n^{\prime(s)} \frac{ \exp (\langle \bm{q}_{n, i}^{(s)}, \bm{k}_+^{(s)} \rangle)}{ \exp  (\langle \bm{q}_{n, i}^{(s)}, \bm{k}_+^{(s)} \rangle) + \sum\limits_{j=2}^M \exp ( \langle \bm{q}_{n, i}^{(s)}, \bm{k}_{n, j}^{(s)} \rangle ) } \\
    &\cdot \Big( V_+^{(s)} \big( 1 - \frac{ \exp (\langle \bm{q}_{n, i}^{(s)}, \bm{k}_+^{(s)} \rangle)}{ \exp  (\langle \bm{q}_{n, i}^{(s)}, \bm{k}_+^{(s)} \rangle) + \sum\limits_{j=2}^M  \exp  ( \langle \bm{q}_{n, i}^{(s)}, \bm{k}_{n, j}^{(s)} \rangle ) } \big) \\
    & -\sum\limits_{k=2}^M \big( V_{n, i}^{(s)} \cdot \frac{ \exp (\langle \bm{q}_{n, i}^{(s)}, \bm{k}_{n, k}^{(s)} \rangle)}{ \exp  (\langle \bm{q}_{n, i}^{(s)}, \bm{k}_+^{(s)} \rangle) + \sum\limits_{j=2}^M  \exp  ( \langle \bm{q}_{n, i}^{(s)}, \bm{k}_{n, j}^{(s)} \rangle ) } \big) \Big) \\
    &\ge  - \frac{\eta \Vert \bm{\mu} \Vert_2^2}{NM} \ell_n^{\prime(s)} \frac{ \exp (\langle \bm{q}_{n, i}^{(s)}, \bm{k}_+^{(s)} \rangle)}{ \exp  (\langle \bm{q}_{n, i}^{(s)}, \bm{k}_+^{(s)} \rangle) + \sum\limits_{j=2}^M  \exp  ( \langle \bm{q}_{n, i}^{(s)}, \bm{k}_{n, j}^{(s)} \rangle ) } \\
    &\cdot \Big( V_+^{(s)} \big( 1 - \frac{ \exp (\langle \bm{q}_{n, i}^{(s)}, \bm{k}_+^{(s)} \rangle)}{ \exp  (\langle \bm{q}_{n, i}^{(s)}, \bm{k}_+^{(s)} \rangle) + \sum\limits_{j=2}^M  \exp  ( \langle \bm{q}_{n, i}^{(s)}, \bm{k}_{n, j}^{(s)} \rangle ) } \big) \\
    & - \frac{1}{2} V_+^{(s)} \sum\limits_{k=2}^M \cdot \frac{ \exp (\langle \bm{q}_{n, i}^{(s)}, \bm{k}_{n, k}^{(s)} \rangle)}{ \exp  (\langle \bm{q}_{n, i}^{(s)}, \bm{k}_+^{(s)} \rangle) + \sum\limits_{j=2}^M  \exp  ( \langle \bm{q}_{n, i}^{(s)}, \bm{k}_{n, j}^{(s)} \rangle ) } \Big) \\
    &\ge - \frac{\eta \Vert \bm{\mu} \Vert_2^2}{2NM} \ell_n^{\prime(s)} V_+^{(s)} \frac{ \exp (\langle \bm{q}_{n, i}^{(s)}, \bm{k}_+^{(s)} \rangle)}{ \exp  (\langle \bm{q}_{n, i}^{(s)}, \bm{k}_+^{(s)} \rangle) + \sum\limits_{j=2}^M \exp  ( \langle \bm{q}_{n, i}^{(s)}, \bm{k}_{n, j}^{(s)} \rangle ) } \\
    &\cdot \big( 1 - \frac{ \exp (\langle \bm{q}_{n, i}^{(s)}, \bm{k}_+^{(s)} \rangle)}{ \exp  (\langle \bm{q}_{n, i}^{(s)}, \bm{k}_+^{(s)} \rangle) + \sum\limits_{j=2}^M  \exp  ( \langle \bm{q}_{n, i}^{(s)}, \bm{k}_{n, j}^{(s)} \rangle ) } \big),
\end{align*}
where the first inequality is by \( V_+^{(s)} \ge 3M \cdot \vert V_{n, i}^{(s)} \vert \), the second inequality is by \( \sum\limits_{k=2}^M softmax(\langle \bm{q}_{n, i}^{(s)}, \bm{k}_{n, k}^{(s)} \rangle) = \big( 1 - softmax(\langle \bm{q}_{n, i}^{(s)}, \bm{k}_+^{(s)} \rangle) \big) \). Similarly, we have
\begin{align*}
    &\beta_{n, -, i}^{(s)} \ge \\
    &\frac{\eta \Vert \bm{\mu} \Vert_2^2}{2NM} \ell_n^{\prime(s)} V_-^{(s)} \frac{ \exp (\langle \bm{q}_{n, i}^{(s)}, \bm{k}_-^{(s)} \rangle)}{ \exp  (\langle \bm{q}_{n, i}^{(s)}, \bm{k}_-^{(s)} \rangle) + \sum\limits_{j=2}^M \exp  ( \langle \bm{q}_{n, i}^{(s)}, \bm{k}_{n, j}^{(s)} \rangle ) } \\
    &\cdot \big( 1 - \frac{ \exp (\langle \bm{q}_{n, i}^{(s)}, \bm{k}_-^{(s)} \rangle)}{ \exp  (\langle \bm{q}_{n, i}^{(s)}, \bm{k}_-^{(s)} \rangle) + \sum\limits_{j=2}^M  \exp  ( \langle \bm{q}_{n, i}^{(s)}, \bm{k}_{n, j}^{(s)} \rangle ) } \big).
\end{align*}

\begin{align*}
    &\alpha_{n^\prime, i^\prime, +}^{(s)} = \frac{\eta \Vert \bm{\xi}_{n^\prime, i^\prime} \Vert_2^2}{NM} - \ell_{n^\prime}^{\prime}(\theta (s)) \\
    &\cdot \Big( V_+^{(s)} \big( \frac{ \exp (\langle \bm{q}_{n^\prime, i^\prime}^{(s)}, \bm{k}_+^{(s)} \rangle)}{ \exp  (\langle \bm{q}_{n^\prime, i^\prime}^{(s)}, \bm{k}_+^{(s)} \rangle) + \sum\limits_{j=2}^M  \exp  ( \langle \bm{q}_{n^\prime, i^\prime}^{(s)}, \bm{k}_{n^\prime, j}^{(s)} \rangle ) } \\
    & - (\frac{ \exp (\langle \bm{q}_{n^\prime, i^\prime}^{(s)}, \bm{k}_+^{(s)} \rangle)}{ \exp  (\langle \bm{q}_{n^\prime, i^\prime}^{(s)}, \bm{k}_+^{(s)} \rangle) + \sum\limits_{j=2}^M  \exp  ( \langle \bm{q}_{n^\prime, i^\prime}^{(s)}, \bm{k}_{n^\prime, j}^{(s)} \rangle ) })^2 \big) \\
    & -\sum\limits_{k=2}^M \big( V_{n^\prime, i^\prime}^{(s)} \cdot \frac{ \exp (\langle \bm{q}_{n^\prime, i^\prime}^{(s)}, \bm{k}_+^{(s)} \rangle)}{ \exp  (\langle \bm{q}_{n^\prime, i^\prime}^{(s)}, \bm{k}_+^{(s)} \rangle) + \sum\limits_{j=2}^M  \exp  ( \langle \bm{q}_{n^\prime, i^\prime}^{(s)}, \bm{k}_{n^\prime, j}^{(s)} \rangle ) } \\
    & \cdot \frac{ \exp (\langle \bm{q}_{n^\prime, i^\prime}^{(s)}, \bm{k}_{n^\prime, k}^{(s)} \rangle)}{ \exp  (\langle \bm{q}_{n^\prime, i^\prime}^{(s)}, \bm{k}_+^{(s)} \rangle) + \sum\limits_{j=2}^M  \exp  ( \langle \bm{q}_{n^\prime, i^\prime}^{(s)}, \bm{k}_{n^\prime, j}^{(s)} \rangle ) } \big) \Big) \\
    &+ \{ lower \ order \ term \} \\
    &= \frac{\eta \Vert \bm{\xi}_{n^\prime, i^\prime} \Vert_2^2}{NM} - \ell_{n^\prime}^{\prime}(\theta (s)) \frac{ \exp (\langle \bm{q}_{n^\prime, i^\prime}^{(s)}, \bm{k}_+^{(s)} \rangle)}{ \exp  (\langle \bm{q}_{n^\prime, i^\prime}^{(s)}, \bm{k}_+^{(s)} \rangle) + \sum\limits_{j=2}^M  \exp  ( \langle \bm{q}_{n^\prime, i^\prime}^{(s)}, \bm{k}_{n^\prime, j}^{(s)} \rangle ) } \\
    &\cdot \Big( V_+^{(s)} \big( 1 - \frac{ \exp (\langle \bm{q}_{n^\prime, i^\prime}^{(s)}, \bm{k}_+^{(s)} \rangle)}{ \exp  (\langle \bm{q}_{n^\prime, i^\prime}^{(s)}, \bm{k}_+^{(s)} \rangle) + \sum\limits_{j=2}^M  \exp  ( \langle \bm{q}_{n^\prime, i^\prime}^{(s)}, \bm{k}_{n^\prime, j}^{(s)} \rangle ) } \big) \\
    & -\sum\limits_{k=2}^M \big( V_{n^\prime, i^\prime}^{(s)} \cdot \frac{ \exp (\langle \bm{q}_{n^\prime, i^\prime}^{(s)}, \bm{k}_{n^\prime, k}^{(s)} \rangle)}{ \exp  (\langle \bm{q}_{n^\prime, i^\prime}^{(s)}, \bm{k}_+^{(s)} \rangle) + \sum\limits_{j=2}^M  \exp  ( \langle \bm{q}_{n^\prime, i^\prime}^{(s)}, \bm{k}_{n^\prime, j}^{(s)} \rangle ) } \big) \Big) \\
    &+ \{ lower \ order \ term \} \\
    &\ge \frac{\eta \Vert \bm{\xi}_{n^\prime, i^\prime} \Vert_2^2}{NM} - \ell_{n^\prime}^{\prime}(\theta (s)) \frac{ \exp (\langle \bm{q}_{n^\prime, i^\prime}^{(s)}, \bm{k}_+^{(s)} \rangle)}{ \exp  (\langle \bm{q}_{n^\prime, i^\prime}^{(s)}, \bm{k}_+^{(s)} \rangle) + \sum\limits_{j=2}^M  \exp  ( \langle \bm{q}_{n^\prime, i^\prime}^{(s)}, \bm{k}_{n, j}^{(s)} \rangle ) } \\
    &\cdot \Big( V_+^{(s)} \big( 1 - \frac{ \exp (\langle \bm{q}_{n^\prime, i^\prime}^{(s)}, \bm{k}_+^{(s)} \rangle)}{ \exp  (\langle \bm{q}_{n^\prime, i^\prime}^{(s)}, \bm{k}_+^{(s)} \rangle) + \sum\limits_{j=2}^M  \exp  ( \langle \bm{q}_{n^\prime, i^\prime}^{(s)}, \bm{k}_{n^\prime, j}^{(s)} \rangle ) } \big) \\
    & - \frac{1}{2} V_+^{(s)} \sum\limits_{k=2}^M \big( \frac{ \exp (\langle \bm{q}_{n^\prime, i^\prime}^{(s)}, \bm{k}_{n^\prime, k}^{(s)} \rangle)}{ \exp  (\langle \bm{q}_{n^\prime, i^\prime}^{(s)}, \bm{k}_+^{(s)} \rangle) + \sum\limits_{j=2}^M  \exp  ( \langle \bm{q}_{n^\prime, i^\prime}^{(s)}, \bm{k}_{n^\prime, j}^{(s)} \rangle ) } \big) \Big) \\
    &+ \{ lower \ order \ term \} \\
    &= \frac{\eta \Vert \bm{\xi}_{n^\prime, i^\prime} \Vert_2^2}{2NM} - \ell_{n^\prime}^{\prime}(\theta (s)) V_+^{(s)} \frac{ \exp (\langle \bm{q}_{n^\prime, i^\prime}^{(s)}, \bm{k}_+^{(s)} \rangle)}{ \exp  (\langle \bm{q}_{n^\prime, i^\prime}^{(s)}, \bm{k}_+^{(s)} \rangle) + \sum\limits_{j=2}^M \exp ( \langle \bm{q}_{n^\prime, i^\prime}^{(s)}, \bm{k}_{n^\prime, j}^{(s)} \rangle ) } \\
    &\cdot \big( 1 - \frac{ \exp (\langle \bm{q}_{n^\prime, i^\prime}^{(s)}, \bm{k}_+^{(s)} \rangle)}{ \exp  (\langle \bm{q}_{n^\prime, i^\prime}^{(s)}, \bm{k}_+^{(s)} \rangle) + \sum\limits_{j=2}^M  \exp  ( \langle \bm{q}_{n^\prime, i^\prime}^{(s)}, \bm{k}_{n^\prime, j}^{(s)} \rangle ) } \big) \\
    &+ \{ lower \ order \ term \} \\
    &\ge \frac{\eta \sigma_p^2 d}{5NM} - \ell_{n^\prime}^{\prime}(\theta (s)) V_+^{(s)} \frac{ \exp (\langle \bm{q}_{n^\prime, i^\prime}^{(s)}, \bm{k}_+^{(s)} \rangle)}{ \exp  (\langle \bm{q}_{n^\prime, i^\prime}^{(s)}, \bm{k}_+^{(s)} \rangle) + \sum\limits_{j=2}^M \exp ( \langle \bm{q}_{n^\prime, i^\prime}^{(s)}, \bm{k}_{n^\prime, j}^{(s)} \rangle ) } \\
    &\cdot \big( 1 - \frac{ \exp (\langle \bm{q}_{n^\prime, i^\prime}^{(s)}, \bm{k}_+^{(s)} \rangle)}{ \exp  (\langle \bm{q}_{n^\prime, i^\prime}^{(s)}, \bm{k}_+^{(s)} \rangle) + \sum\limits_{j=2}^M  \exp  ( \langle \bm{q}_{n^\prime, i^\prime}^{(s)}, \bm{k}_{n^\prime, j}^{(s)} \rangle ) } \big),
\end{align*}
where the \( \{ lower \ order \ term \} \) is by the property that \( \Vert \bm{\xi}_{n^\prime, i^\prime} \Vert_2^2 \) is much larger than \( \langle \bm{\xi}_{n^\prime, i^\prime}, \bm{\xi}_{n, i} \rangle \) in Lemma \ref{caoyuan} and the condition \(d = \widetilde{\Omega} \Big( \epsilon^{-2} N^2 d_h \Big) \), the first inequality is by \( V_+^{(s)} \ge 3M \cdot \vert V_{n, i}^{(s)} \vert \), the last inequality is by \( \Vert \bm{\xi}_{n^\prime, i^\prime} \Vert_2^2 \ge \frac{\sigma_p^2 d}{2} \) and we absorb the \( \{ lower \ order \ term \} \). 
Similarly, we have
\begin{align*}
    &\alpha_{n^\prime, i^\prime, -}^{(s)} \ge \\
    &\frac{\eta \sigma_p^2 d}{5NM} \ell_{n^\prime}^{\prime}(\theta (s)) V_-^{(s)} \frac{ \exp (\langle \bm{q}_{n^\prime, i^\prime}^{(s)}, \bm{k}_-^{(s)} \rangle)}{ \exp  (\langle \bm{q}_{n^\prime, i^\prime}^{(s)}, \bm{k}_-^{(s)} \rangle) + \sum\limits_{j=2}^M \exp ( \langle \bm{q}_{n^\prime, i^\prime}^{(s)}, \bm{k}_{n^\prime, j}^{(s)} \rangle ) } \\
    &\cdot \big( 1 - \frac{ \exp (\langle \bm{q}_{n^\prime, i^\prime}^{(s)}, \bm{k}_-^{(s)} \rangle)}{ \exp  (\langle \bm{q}_{n^\prime, i^\prime}^{(s)}, \bm{k}_-^{(s)} \rangle) + \sum\limits_{j=2}^M  \exp  ( \langle \bm{q}_{n^\prime, i^\prime}^{(s)}, \bm{k}_{n^\prime, j}^{(s)} \rangle ) } \big).
\end{align*}

\begin{align*}
    &\beta_{n^\prime, i^\prime, +}^{(s)} = \frac{\eta \Vert \bm{\xi}_{n^\prime, i^\prime} \Vert_2^2}{NM} - \ell_{n^\prime}^{\prime}(\theta (s)) \\
    &\Big( - V_+^{(s)} \big( \frac{\exp (\langle \bm{q}_+^{(s)}, \bm{k}_+^{(s)} \rangle)}{\exp (\langle \bm{q}_+^{(s)}, \bm{k}_+^{(s)} \rangle) + \sum\limits_{j=2}^M \exp ( \langle \bm{q}_+^{(s)}, \bm{k}_{n^\prime, j}^{(s)} \rangle )} \\
    &\cdot \frac{\exp (\langle \bm{q}_+^{(s)}, \bm{k}_{n^\prime, i^\prime}^{(s)} \rangle)}{\exp (\langle \bm{q}_+^{(s)}, \bm{k}_+^{(s)} \rangle) + \sum\limits_{j=2}^M \exp ( \langle \bm{q}_+^{(s)}, \bm{k}_{n^\prime, j}^{(s)} \rangle )} \big) \\
    &+ V_{n^\prime, i^\prime}^{(s)} \big( \frac{\exp (\langle \bm{q}_+^{(s)}, \bm{k}_{n^\prime, i^\prime}^{(s)} \rangle)}{\exp (\langle \bm{q}_+^{(s)}, \bm{k}_+^{(s)} \rangle) + \sum\limits_{j=2}^M \exp ( \langle \bm{q}_+^{(s)}, \bm{k}_{n^\prime, j}^{(s)} \rangle )}\\
    &\cdot \big(1 - \frac{\exp (\langle \bm{q}_+^{(s)}, \bm{k}_{n^\prime, i^\prime}^{(s)} \rangle)}{\exp (\langle \bm{q}_+^{(s)}, \bm{k}_+^{(s)} \rangle) + \sum\limits_{j=2}^M \exp ( \langle \bm{q}_+^{(s)}, \bm{k}_{n^\prime, j}^{(s)} \rangle )} \big) \big) \Big) \\
    &+ \{ lower \ order \ term \} \\
    &= \frac{\eta \Vert \bm{\xi}_{n^\prime, i^\prime} \Vert_2^2}{NM} - \ell_{n^\prime}^{\prime}(\theta (s)) \cdot \frac{\exp (\langle \bm{q}_+^{(s)}, \bm{k}_{n^\prime, i^\prime}^{(s)} \rangle)}{\exp (\langle \bm{q}_+^{(s)}, \bm{k}_+^{(s)} \rangle) + \sum\limits_{j=2}^M \exp ( \langle \bm{q}_+^{(s)}, \bm{k}_{n^\prime, j}^{(s)} \rangle )} \\
    &\Big( - V_+^{(s)} \cdot \frac{\exp (\langle \bm{q}_+^{(s)}, \bm{k}_+^{(s)} \rangle)}{\exp (\langle \bm{q}_+^{(s)}, \bm{k}_+^{(s)} \rangle) + \sum\limits_{j=2}^M \exp ( \langle \bm{q}_+^{(s)}, \bm{k}_{n^\prime, j}^{(s)} \rangle )} \\
    &+ V_{n^\prime, i^\prime}^{(s)} \cdot \big(1 - \frac{\exp (\langle \bm{q}_+^{(s)}, \bm{k}_{n^\prime, i^\prime}^{(s)} \rangle)}{\exp (\langle \bm{q}_+^{(s)}, \bm{k}_+^{(s)} \rangle) + \sum\limits_{j=2}^M \exp ( \langle \bm{q}_+^{(s)}, \bm{k}_{n^\prime, j}^{(s)} \rangle )} \big) \Big) \\
    &+ \{lower \ order \ term\} \\
    &\le \frac{\eta \Vert \bm{\xi}_{n^\prime, i^\prime} \Vert_2^2}{NM} - \ell_{n^\prime}^{\prime}(\theta (s)) \cdot \frac{\exp (\langle \bm{q}_+^{(s)}, \bm{k}_{n^\prime, i^\prime}^{(s)} \rangle)}{\exp (\langle \bm{q}_+^{(s)}, \bm{k}_+^{(s)} \rangle) + \sum\limits_{j=2}^M \exp ( \langle \bm{q}_+^{(s)}, \bm{k}_{n^\prime, j}^{(s)} \rangle )} \\
    &\Big( - V_+^{(s)} \cdot \frac{\exp (\langle \bm{q}_+^{(s)}, \bm{k}_+^{(s)} \rangle)}{\exp (\langle \bm{q}_+^{(s)}, \bm{k}_+^{(s)} \rangle) + \sum\limits_{j=2}^M \exp ( \langle \bm{q}_+^{(s)}, \bm{k}_{n^\prime, j}^{(s)} \rangle )} \\
    &+ \frac{1}{2} V_+^{(s)} \cdot \frac{\exp (\langle \bm{q}_+^{(s)}, \bm{k}_+^{(s)} \rangle)}{\exp (\langle \bm{q}_+^{(s)}, \bm{k}_+^{(s)} \rangle) + \sum\limits_{j=2}^M \exp ( \langle \bm{q}_+^{(s)}, \bm{k}_{n^\prime, j}^{(s)} \rangle )} \Big) \\
    &\le \frac{\eta \Vert \bm{\xi}_{n^\prime, i^\prime} \Vert_2^2}{2NM} \ell_{n^\prime}^{\prime}(\theta (s)) V_+^{(s)} \cdot \frac{\exp (\langle \bm{q}_+^{(s)}, \bm{k}_{n^\prime, i^\prime}^{(s)} \rangle)}{\exp (\langle \bm{q}_+^{(s)}, \bm{k}_+^{(s)} \rangle) + \sum\limits_{j=2}^M \exp ( \langle \bm{q}_+^{(s)}, \bm{k}_{n^\prime, j}^{(s)} \rangle )} \\
    &\cdot \frac{\exp (\langle \bm{q}_+^{(s)}, \bm{k}_+^{(s)} \rangle)}{\exp (\langle \bm{q}_+^{(s)}, \bm{k}_+^{(s)} \rangle) + \sum\limits_{j=2}^M \exp ( \langle \bm{q}_+^{(s)}, \bm{k}_{n^\prime, j}^{(s)} \rangle )} \\
    &+ \{lower \ order \ term\} \\
    &\le \frac{\eta \sigma_p^2 d}{5NM} \ell_{n^\prime}^{\prime}(\theta (s)) V_+^{(s)} \cdot \frac{\exp (\langle \bm{q}_+^{(s)}, \bm{k}_{n^\prime, i^\prime}^{(s)} \rangle)}{\exp (\langle \bm{q}_+^{(s)}, \bm{k}_+^{(s)} \rangle) + \sum\limits_{j=2}^M \exp ( \langle \bm{q}_+^{(s)}, \bm{k}_{n^\prime, j}^{(s)} \rangle )} \\
    &\cdot \frac{\exp (\langle \bm{q}_+^{(s)}, \bm{k}_+^{(s)} \rangle)}{\exp (\langle \bm{q}_+^{(s)}, \bm{k}_+^{(s)} \rangle) + \sum\limits_{j=2}^M \exp ( \langle \bm{q}_+^{(s)}, \bm{k}_{n^\prime, j}^{(s)} \rangle )},
\end{align*}
where the \( \{ lower \ order \ term \} \) is by the property that \( \Vert \bm{\xi}_{n^\prime, i^\prime} \Vert_2^2 \) is much larger than \( \langle \bm{\xi}_{n^\prime, i^\prime}, \bm{\xi}_{n, i} \rangle \) in Lemma \ref{caoyuan} and the condition \(d = \widetilde{\Omega} \Big( \epsilon^{-2} N^2 d_h \Big) \), the first inequality is by \( V_+^{(s)} \ge 3M \cdot \vert V_{n, i}^{(s)} \vert \), the last inequality is by \( \Vert \bm{\xi}_{n^\prime, i^\prime} \Vert_2^2 \ge \frac{\sigma_p^2 d}{2} \) and we absorb the \( \{ lower \ order \ term \} \). 
Similarly, we have
\begin{align*}
    &\beta_{n^\prime, i^\prime, -}^{(s)} \le \\
    &- \frac{\eta \sigma_p^2 d}{5NM} \ell_{n^\prime}^{\prime}(\theta (s)) V_-^{(s)} \cdot \frac{\exp (\langle \bm{q}_-^{(s)}, \bm{k}_{n^\prime, i^\prime}^{(s)} \rangle)}{\exp (\langle \bm{q}_-^{(s)}, \bm{k}_-^{(s)} \rangle) + \sum\limits_{j=2}^M \exp ( \langle \bm{q}_-^{(s)}, \bm{k}_{n^\prime, j}^{(s)} \rangle )} \\
    &\cdot \frac{\exp (\langle \bm{q}_-^{(s)}, \bm{k}_-^{(s)} \rangle)}{\exp (\langle \bm{q}_-^{(s)}, \bm{k}_-^{(s)} \rangle) + \sum\limits_{j=2}^M \exp ( \langle \bm{q}_-^{(s)}, \bm{k}_{n^\prime, j}^{(s)} \rangle )}.
\end{align*}

\begin{align*}
    &\alpha_{n, i, n, j}^{(s)} = \frac{\eta}{NM} - \ell_n^{\prime(s)} \sum\limits_{k=2}^M \langle \bm{\xi}_{n, i}, \bm{\xi}_{n, k} \rangle \cdot \frac{ \exp (\langle \bm{q}_{n, k}^{(s)}, \bm{k}_{n, j}^{(s)} \rangle)}{ \exp  (\langle \bm{q}_{n, k}^{(s)}, \bm{k}_+^{(s)} \rangle) + \sum\limits_{l^\prime=2}^M  \exp  ( \langle \bm{q}_{n, k}^{(s)}, \bm{k}_{n, l^\prime}^{(s)} \rangle ) } \\
    &\cdot \Big( - V_+^{(s)} \cdot \frac{ \exp (\langle \bm{q}_{n, k}^{(s)}, \bm{k}_+^{(s)} \rangle)}{ \exp (\langle \bm{q}_{n, k}^{(s)}, \bm{k}_+^{(s)} \rangle) + \sum\limits_{l^\prime=2}^M \exp ( \langle \bm{q}_{n, k}^{(s)}, \bm{k}_{n, l^\prime}^{(s)} \rangle ) } \\
    &+ V_{n, j}^{(s)} \big( 1 - \frac{ \exp (\langle \bm{q}_{n, k}^{(s)}, \bm{k}_{n, j}^{(s)} \rangle)}{ \exp  (\langle \bm{q}_{n, k}^{(s)}, \bm{k}_+^{(s)} \rangle) + \sum\limits_{l^\prime=2}^M  \exp  ( \langle \bm{q}_{n, k}^{(s)}, \bm{k}_{n, l^\prime}^{(s)} \rangle ) } \big) \\
    &- \sum\limits_{l \ne i} \big( V_{n, l}^{(s)} \cdot \frac{ \exp (\langle \bm{q}_{n, k}^{(s)}, \bm{k}_{n, l}^{(s)} \rangle)}{ \exp  (\langle \bm{q}_{n, k}^{(s)}, \bm{k}_+^{(s)} \rangle) + \sum\limits_{l^\prime=2}^M  \exp  ( \langle \bm{q}_{n, k}^{(s)}, \bm{k}_{n, l^\prime}^{(s)} \rangle ) } \big) \Big) \\
    &= - \frac{\eta \Vert \bm{\xi}_{n, i} \Vert_2^2}{NM} \ell_n^{\prime(s)} \cdot \frac{ \exp (\langle \bm{q}_{n, i}^{(s)}, \bm{k}_{n, j}^{(s)} \rangle)}{ \exp  (\langle \bm{q}_{n, i}^{(s)}, \bm{k}_+^{(s)} \rangle) + \sum\limits_{l^\prime=2}^M  \exp  ( \langle \bm{q}_{n, i}^{(s)}, \bm{k}_{n, l^\prime}^{(s)} \rangle ) } \\
    &\cdot \Big( - V_+^{(s)} \cdot \frac{ \exp (\langle \bm{q}_{n, i}^{(s)}, \bm{k}_+^{(s)} \rangle)}{ \exp (\langle \bm{q}_{n, i}^{(s)}, \bm{k}_+^{(s)} \rangle) + \sum\limits_{l^\prime=2}^M \exp ( \langle \bm{q}_{n, i}^{(s)}, \bm{k}_{n, l^\prime}^{(s)} \rangle ) } \\
    &+V_{n, j}^{(s)} \big( 1 - \frac{ \exp (\langle \bm{q}_{n, i}^{(s)}, \bm{k}_{n, j}^{(s)} \rangle)}{ \exp  (\langle \bm{q}_{n, i}^{(s)}, \bm{k}_+^{(s)} \rangle) + \sum\limits_{l^\prime=2}^M  \exp  ( \langle \bm{q}_{n, i}^{(s)}, \bm{k}_{n, l^\prime}^{(s)} \rangle ) } \big) \\
    &- \sum\limits_{l \ne i} \big( V_{n, l}^{(s)} \cdot \frac{ \exp (\langle \bm{q}_{n, i}^{(s)}, \bm{k}_{n, l}^{(s)} \rangle)}{ \exp  (\langle \bm{q}_{n, i}^{(s)}, \bm{k}_+^{(s)} \rangle) + \sum\limits_{l^\prime=2}^M  \exp  ( \langle \bm{q}_{n, i}^{(s)}, \bm{k}_{n, l^\prime}^{(s)} \rangle ) } \big) \Big) \\
    &+ \{ lower \ order \ term \} \\
    &\le - \frac{\eta \Vert \bm{\xi}_{n, i} \Vert_2^2}{NM} \ell_n^{\prime(s)} \cdot \frac{ \exp (\langle \bm{q}_{n, i}^{(s)}, \bm{k}_{n, j}^{(s)} \rangle)}{ \exp  (\langle \bm{q}_{n, i}^{(s)}, \bm{k}_+^{(s)} \rangle) + \sum\limits_{l^\prime=2}^M  \exp  ( \langle \bm{q}_{n, i}^{(s)}, \bm{k}_{n, l^\prime}^{(s)} \rangle ) } \\
    &\cdot \Big( - V_+^{(s)} \cdot \frac{ \exp (\langle \bm{q}_{n, i}^{(s)}, \bm{k}_+^{(s)} \rangle)}{ \exp (\langle \bm{q}_{n, i}^{(s)}, \bm{k}_+^{(s)} \rangle) + \sum\limits_{l^\prime=2}^M \exp ( \langle \bm{q}_{n, i}^{(s)}, \bm{k}_{n, l^\prime}^{(s)} \rangle ) } \\
    &+ \vert V_{n, j}^{(s)} \vert + \sum\limits_{l \ne i} \big( \vert V_{n, l}^{(s)} \vert \cdot (\frac{1}{M} + o(1)) \big) \Big) \\
    &+ \{ lower \ order \ term \} \\ 
    &\le - \frac{\eta \Vert \bm{\xi}_{n, i} \Vert_2^2}{NM} \ell_n^{\prime(s)} \cdot \frac{ \exp (\langle \bm{q}_{n, i}^{(s)}, \bm{k}_{n, j}^{(s)} \rangle)}{ \exp  (\langle \bm{q}_{n, i}^{(s)}, \bm{k}_+^{(s)} \rangle) + \sum\limits_{l^\prime=2}^M  \exp  ( \langle \bm{q}_{n, i}^{(s)}, \bm{k}_{n, l^\prime}^{(s)} \rangle ) } \\
    &\cdot \Big( - V_+^{(s)} \cdot \frac{ \exp (\langle \bm{q}_{n, i}^{(s)}, \bm{k}_+^{(s)} \rangle)}{ \exp (\langle \bm{q}_{n, i}^{(s)}, \bm{k}_+^{(s)} \rangle) + \sum\limits_{l^\prime=2}^M \exp ( \langle \bm{q}_{n, i}^{(s)}, \bm{k}_{n, l^\prime}^{(s)} \rangle ) } + 2 \max\limits_l \vert V_{n, l}^{(s)} \vert \Big) \\ 
    &+ \{ lower \ order \ term \} \\ 
    &\le - \frac{\eta \Vert \bm{\xi}_{n, i} \Vert_2^2}{NM} \ell_n^{\prime(s)} \cdot \frac{ \exp (\langle \bm{q}_{n, i}^{(s)}, \bm{k}_{n, j}^{(s)} \rangle)}{ \exp  (\langle \bm{q}_{n, i}^{(s)}, \bm{k}_+^{(s)} \rangle) + \sum\limits_{l^\prime=2}^M  \exp  ( \langle \bm{q}_{n, i}^{(s)}, \bm{k}_{n, l^\prime}^{(s)} \rangle ) } \\
    &\cdot \Big( - V_+^{(s)} \cdot \frac{ \exp (\langle \bm{q}_{n, i}^{(s)}, \bm{k}_+^{(s)} \rangle)}{ \exp (\langle \bm{q}_{n, i}^{(s)}, \bm{k}_+^{(s)} \rangle) + \sum\limits_{l^\prime=2}^M \exp ( \langle \bm{q}_{n, i}^{(s)}, \bm{k}_{n, l^\prime}^{(s)} \rangle ) } \\
    &+ \frac{3}{4} V_+^{(s)} \cdot \frac{ \exp (\langle \bm{q}_{n, i}^{(s)}, \bm{k}_+^{(s)} \rangle)}{ \exp (\langle \bm{q}_{n, i}^{(s)}, \bm{k}_+^{(s)} \rangle) + \sum\limits_{l^\prime=2}^M \exp ( \langle \bm{q}_{n, i}^{(s)}, \bm{k}_{n, l^\prime}^{(s)} \rangle ) } \Big) \\ 
    &+ \{ lower \ order \ term \} \\ 
    &\le \frac{\eta \Vert \bm{\xi}_{n, i} \Vert_2^2}{4NM} \ell_n^{\prime(s)} V_+^{(s)} \cdot \frac{ \exp (\langle \bm{q}_{n, i}^{(s)}, \bm{k}_{n, j}^{(s)} \rangle)}{ \exp  (\langle \bm{q}_{n, i}^{(s)}, \bm{k}_+^{(s)} \rangle) + \sum\limits_{l^\prime=2}^M  \exp  ( \langle \bm{q}_{n, i}^{(s)}, \bm{k}_{n, l^\prime}^{(s)} \rangle ) } \\
    &\cdot \frac{ \exp (\langle \bm{q}_{n, i}^{(s)}, \bm{k}_+^{(s)} \rangle)}{ \exp (\langle \bm{q}_{n, i}^{(s)}, \bm{k}_+^{(s)} \rangle) + \sum\limits_{l^\prime=2}^M \exp ( \langle \bm{q}_{n, i}^{(s)}, \bm{k}_{n, l^\prime}^{(s)} \rangle ) } \\
    &+ \{ lower \ order \ term \} \\ 
    &\le \frac{\eta \sigma_p^2 d}{9NM} \ell_n^{\prime(s)} V_+^{(s)} \cdot \frac{ \exp (\langle \bm{q}_{n, i}^{(s)}, \bm{k}_{n, j}^{(s)} \rangle)}{ \exp  (\langle \bm{q}_{n, i}^{(s)}, \bm{k}_+^{(s)} \rangle) + \sum\limits_{l^\prime=2}^M  \exp  ( \langle \bm{q}_{n, i}^{(s)}, \bm{k}_{n, l^\prime}^{(s)} \rangle ) } \\
    &\cdot \frac{ \exp (\langle \bm{q}_{n, i}^{(s)}, \bm{k}_+^{(s)} \rangle)}{ \exp (\langle \bm{q}_{n, i}^{(s)}, \bm{k}_+^{(s)} \rangle) + \sum\limits_{l^\prime=2}^M \exp ( \langle \bm{q}_{n, i}^{(s)}, \bm{k}_{n, l^\prime}^{(s)} \rangle ) },
\end{align*}
where the \( \{ lower \ order \ term \} \) is by the property that \( \Vert \bm{\xi}_{n, i} \Vert_2^2 \) is much larger than \( \langle \bm{\xi}_{n^\prime, i^\prime}, \bm{\xi}_{n, i} \rangle \) in Lemma \ref{caoyuan} and the condition \(d = \widetilde{\Omega} \Big( \epsilon^{-2} N^2 d_h \Big) \), the third inequality is by \( V_+^{(s)} \ge 3M \cdot \vert V_{n, i}^{(s)} \vert \) and \( softmax(\langle \bm{q}_{n, i}^{(s)}, \bm{k}_+^{(s)} \rangle) \ge \frac{1}{M} - o(1) \), the last inequality is by \( \Vert \bm{\xi}_{n, i} \Vert_2^2 \ge \frac{\sigma_p^2 d}{2} \) and we absorb the \( \{ lower \ order \ term \} \). 
Similarly, we have
\begin{align*}
    &\alpha_{n, i, n, j}^{(s)} \le \\
    &- \frac{\eta \sigma_p^2 d}{9NM} \ell_n^{\prime(s)} V_-^{(s)} \cdot \frac{ \exp (\langle \bm{q}_{n, i}^{(s)}, \bm{k}_{n, j}^{(s)} \rangle)}{ \exp  (\langle \bm{q}_{n, i}^{(s)}, \bm{k}_-^{(s)} \rangle) + \sum\limits_{l^\prime=2}^M  \exp  ( \langle \bm{q}_{n, i}^{(s)}, \bm{k}_{n, l^\prime}^{(s)} \rangle ) } \\
    &\cdot \frac{ \exp (\langle \bm{q}_{n, i}^{(s)}, \bm{k}_-^{(s)} \rangle)}{ \exp (\langle \bm{q}_{n, i}^{(s)}, \bm{k}_-^{(s)} \rangle) + \sum\limits_{l^\prime=2}^M \exp ( \langle \bm{q}_{n, i}^{(s)}, \bm{k}_{n, l^\prime}^{(s)} \rangle ) }.
\end{align*}

\begin{align*}
    &\beta_{n, j, n, i}^{(s)} = \frac{\eta}{NM} - \ell_n^{\prime(s)} \\
    &\Big( - V_+^{(s)} \sum\limits_{k=2}^M \big( \langle \bm{\xi}_{n, j}, \bm{\xi}_{n, k} \rangle \frac{\exp (\langle \bm{q}_{n, i}^{(s)}, \bm{k}_+^{(s)} \rangle)}{\exp (\langle \bm{q}_{n, i}^{(s)}, \bm{k}_+^{(s)} \rangle) + \sum\limits_{l^\prime=2}^M \exp ( \langle \bm{q}_{n, i}^{(s)}, \bm{k}_{n, l^\prime}^{(s)} \rangle )} \\
    &\cdot \frac{\exp (\langle \bm{q}_{n, i}^{(s)}, \bm{k}_{n, k}^{(s)} \rangle)}{\exp (\langle \bm{q}_{n, i}^{(s)}, \bm{k}_+^{(s)} \rangle) + \sum\limits_{l^\prime=2}^M \exp ( \langle \bm{q}_{n, i}^{(s)}, \bm{k}_{n, l^\prime}^{(s)} \rangle )} \big) \\
    &+ \sum\limits_{k=2}^M V_{n, k}^{(s)} \big( \langle \bm{\xi}_{n, j}, \bm{\xi}_{n, k} \rangle \frac{\exp (\langle \bm{q}_{n, i}^{(s)}, \bm{k}_{n, k}^{(s)} \rangle)}{\exp (\langle \bm{q}_{n, i}^{(s)}, \bm{k}_+^{(s)} \rangle) + \sum\limits_{l^\prime=2}^M \exp ( \langle \bm{q}_{n, i}^{(s)}, \bm{k}_{n, l^\prime}^{(s)} \rangle )}\\
    &- \sum\limits_{l=2}^M \big( \langle \bm{\xi}_{n, j}, \bm{\xi}_{n, l} \rangle \frac{\exp (\langle \bm{q}_{n, i}^{(s)}, \bm{k}_{n, k}^{(s)} \rangle)}{\exp (\langle \bm{q}_{n, i}^{(s)}, \bm{k}_+^{(s)} \rangle) + \sum\limits_{l^\prime=2}^M \exp ( \langle \bm{q}_{n, i}^{(s)}, \bm{k}_{n, l^\prime}^{(s)} \rangle )} \\
    &\cdot \frac{\exp (\langle \bm{q}_{n, i}^{(s)}, \bm{k}_{n, l}^{(s)} \rangle)}{\exp (\langle \bm{q}_{n, i}^{(s)}, \bm{k}_+^{(s)} \rangle) + \sum\limits_{l^\prime=2}^M \exp ( \langle \bm{q}_{n, i}^{(s)}, \bm{k}_{n, l^\prime}^{(s)} \rangle )} \big) \big) \Big) \\
    &= \frac{\eta \Vert \bm{\xi}_{n, j} \Vert_2^2}{NM} - \ell_n^{\prime(s)} \cdot \frac{\exp (\langle \bm{q}_{n, i}^{(s)}, \bm{k}_{n, j}^{(s)} \rangle)}{\exp (\langle \bm{q}_{n, i}^{(s)}, \bm{k}_+^{(s)} \rangle) + \sum\limits_{l^\prime=2}^M \exp ( \langle \bm{q}_{n, i}^{(s)}, \bm{k}_{n, l^\prime}^{(s)} \rangle )}\\
    &\Big( - V_+^{(s)} \big( \frac{\exp (\langle \bm{q}_{n, i}^{(s)}, \bm{k}_+^{(s)} \rangle)}{\exp (\langle \bm{q}_{n, i}^{(s)}, \bm{k}_+^{(s)} \rangle) + \sum\limits_{l^\prime=2}^M \exp ( \langle \bm{q}_{n, i}^{(s)}, \bm{k}_{n, l^\prime}^{(s)} \rangle )} \big) \\
    &+ V_{n, j}^{(s)} \big( 1 - \frac{\exp (\langle \bm{q}_{n, i}^{(s)}, \bm{k}_{n, j}^{(s)} \rangle)}{\exp (\langle \bm{q}_{n, i}^{(s)}, \bm{k}_+^{(s)} \rangle) + \sum\limits_{l^\prime=2}^M \exp ( \langle \bm{q}_{n, i}^{(s)}, \bm{k}_{n, l^\prime}^{(s)} \rangle )} \big) \Big) \\
    &+ \{ lower \ order \ term \} \\ 
    &\le \frac{\eta \Vert \bm{\xi}_{n, j} \Vert_2^2}{NM} - \ell_n^{\prime(s)} \cdot \frac{\exp (\langle \bm{q}_{n, i}^{(s)}, \bm{k}_{n, j}^{(s)} \rangle)}{\exp (\langle \bm{q}_{n, i}^{(s)}, \bm{k}_+^{(s)} \rangle) + \sum\limits_{l^\prime=2}^M \exp ( \langle \bm{q}_{n, i}^{(s)}, \bm{k}_{n, l^\prime}^{(s)} \rangle )}\\
    &\Big( - V_+^{(s)} \frac{\exp (\langle \bm{q}_{n, i}^{(s)}, \bm{k}_+^{(s)} \rangle)}{\exp (\langle \bm{q}_{n, i}^{(s)}, \bm{k}_+^{(s)} \rangle) + \sum\limits_{l^\prime=2}^M \exp ( \langle \bm{q}_{n, i}^{(s)}, \bm{k}_{n, l^\prime}^{(s)} \rangle )} + \max\limits_{j} \vert V_{n, j}^{(s)} \vert \Big) \\
    &+ \{ lower \ order \ term \} \\
    &\le \frac{\eta \Vert \bm{\xi}_{n, j} \Vert_2^2}{NM} - \ell_n^{\prime(s)} \cdot \frac{\exp (\langle \bm{q}_{n, i}^{(s)}, \bm{k}_{n, j}^{(s)} \rangle)}{\exp (\langle \bm{q}_{n, i}^{(s)}, \bm{k}_+^{(s)} \rangle) + \sum\limits_{l^\prime=2}^M \exp ( \langle \bm{q}_{n, i}^{(s)}, \bm{k}_{n, l^\prime}^{(s)} \rangle )}\\
    &\Big( - V_+^{(s)} \frac{\exp (\langle \bm{q}_{n, i}^{(s)}, \bm{k}_+^{(s)} \rangle)}{\exp (\langle \bm{q}_{n, i}^{(s)}, \bm{k}_+^{(s)} \rangle) + \sum\limits_{l^\prime=2}^M \exp ( \langle \bm{q}_{n, i}^{(s)}, \bm{k}_{n, l^\prime}^{(s)} \rangle )} \\
    &+ \frac{1}{2} V_+^{(s)} \frac{\exp (\langle \bm{q}_{n, i}^{(s)}, \bm{k}_+^{(s)} \rangle)}{\exp (\langle \bm{q}_{n, i}^{(s)}, \bm{k}_+^{(s)} \rangle) + \sum\limits_{l^\prime=2}^M \exp ( \langle \bm{q}_{n, i}^{(s)}, \bm{k}_{n, l^\prime}^{(s)} \rangle )}  \Big) \\
    &+ \{ lower \ order \ term \} \\
    &\le \frac{\eta \Vert \bm{\xi}_{n, j} \Vert_2^2}{2NM} \ell_n^{\prime(s)} V_+^{(s)} \cdot \frac{\exp (\langle \bm{q}_{n, i}^{(s)}, \bm{k}_{n, j}^{(s)} \rangle)}{\exp (\langle \bm{q}_{n, i}^{(s)}, \bm{k}_+^{(s)} \rangle) + \sum\limits_{l^\prime=2}^M \exp ( \langle \bm{q}_{n, i}^{(s)}, \bm{k}_{n, l^\prime}^{(s)} \rangle )}\\
    &\cdot \frac{\exp (\langle \bm{q}_{n, i}^{(s)}, \bm{k}_+^{(s)} \rangle)}{\exp (\langle \bm{q}_{n, i}^{(s)}, \bm{k}_+^{(s)} \rangle) + \sum\limits_{l^\prime=2}^M \exp ( \langle \bm{q}_{n, i}^{(s)}, \bm{k}_{n, l^\prime}^{(s)} \rangle )} \\
    &+ \{ lower \ order \ term \} \\
    &\le \frac{\eta \sigma_p^2 d}{5NM} \ell_n^{\prime(s)} V_+^{(s)} \cdot \frac{\exp (\langle \bm{q}_{n, i}^{(s)}, \bm{k}_{n, j}^{(s)} \rangle)}{\exp (\langle \bm{q}_{n, i}^{(s)}, \bm{k}_+^{(s)} \rangle) + \sum\limits_{l^\prime=2}^M \exp ( \langle \bm{q}_{n, i}^{(s)}, \bm{k}_{n, l^\prime}^{(s)} \rangle )}\\
    &\cdot \frac{\exp (\langle \bm{q}_{n, i}^{(s)}, \bm{k}_+^{(s)} \rangle)}{\exp (\langle \bm{q}_{n, i}^{(s)}, \bm{k}_+^{(s)} \rangle) + \sum\limits_{l^\prime=2}^M \exp ( \langle \bm{q}_{n, i}^{(s)}, \bm{k}_{n, l^\prime}^{(s)} \rangle )},
\end{align*}
where the \( \{ lower \ order \ term \} \) is by the property that \( \Vert \bm{\xi}_{n, i} \Vert_2^2 \) is much larger than \( \langle \bm{\xi}_{n^\prime, i^\prime}, \bm{\xi}_{n, i} \rangle \) in Lemma \ref{caoyuan} and the condition \(d = \widetilde{\Omega} \Big( \epsilon^{-2} N^2 d_h \Big) \), the second inequality is by \( V_+^{(s)} \ge 3M \cdot \vert V_{n, i}^{(s)} \vert \), the last inequality is by \( \Vert \bm{\xi}_{n, j} \Vert_2^2 \ge \frac{\sigma_p^2 d}{2} \) and we absorb the \( \{ lower \ order \ term \} \). 
Similarly, we have
\begin{align*}
    &\beta_{n, j, n, i}^{(s)} \le \\
    &- \frac{\eta \sigma_p^2 d}{5NM} \ell_n^{\prime(s)} V_-^{(s)} \cdot \frac{\exp (\langle \bm{q}_{n, i}^{(s)}, \bm{k}_{n, j}^{(s)} \rangle)}{\exp (\langle \bm{q}_{n, i}^{(s)}, \bm{k}_-^{(s)} \rangle) + \sum\limits_{l^\prime=2}^M \exp ( \langle \bm{q}_{n, i}^{(s)}, \bm{k}_{n, l^\prime}^{(s)} \rangle )}\\
    &\cdot \frac{\exp (\langle \bm{q}_{n, i}^{(s)}, \bm{k}_-^{(s)} \rangle)}{\exp (\langle \bm{q}_{n, i}^{(s)}, \bm{k}_-^{(s)} \rangle) + \sum\limits_{l^\prime=2}^M \exp ( \langle \bm{q}_{n, i}^{(s)}, \bm{k}_{n, l^\prime}^{(s)} \rangle )}.
\end{align*}

Summarizing the above equations, we have the signs of \(\alpha\) and \(\beta\) as follows:

\[
    \alpha_{+, +}^{(s)}, \alpha_{-, -}^{(s)}, \beta_{+, +}^{(s)}, \beta_{-, -}^{(s)}, \alpha_{n, i, +}^{(s)}, \alpha_{n, i, -}^{(s)}, \beta_{n, +, i}^{(s)}, \beta_{n, -, i}^{(s)} \ge 0,
\]
\[
    \alpha_{n, +, i}^{(s)}, \alpha_{n, -, i}^{(s)}, \alpha_{n, i, n, j}^{(s)}, \beta_{n, i, +}^{(s)}, \beta_{n, i, -}^{(s)}, \beta_{n, j, n, i}^{(s)} \le 0.
\]

\subsection{Lower Bounds of \(\langle\) q, k \(\rangle\)}
\label{lower_bound_of_qk}
In order to give the lower bounds for \( \langle \bm{q}, \bm{k} \rangle \), we need to rewrite the bounds of \(\alpha\) and \(\beta\) in a more concise form.
We first expand the equations in \ref{bound_of_alpha_beta} under the assumption that \(\mathcal{B}(s)\) holds for \(s \in [T_1, t]\).
\begin{equation}
\begin{split}
    \label{lower_bound_alpha_p_p}
    &\alpha_{+, +}^{(s)} \\
    &\ge \frac{\eta}{2NM} \sum\limits_{n \in S_+} - \ell_n^{\prime(s)} \Vert \bm{\mu} \Vert_2^2 V_+^{(s)} \cdot \frac{ \exp (\langle \bm{q}_+^{(s)}, \bm{k}_+^{(s)} \rangle)}{ \exp  (\langle \bm{q}_+^{(s)}, \bm{k}_+^{(s)} \rangle) + \sum\limits_{j^\prime=2}^M  \exp  ( \langle \bm{q}_+^{(s)}, \bm{k}_{n, j^\prime}^{(s)} \rangle ) } \\
    &\cdot \big( 1 - \frac{ \exp (\langle \bm{q}_+^{(s)}, \bm{k}_+^{(s)} \rangle)}{ \exp  (\langle \bm{q}_+^{(s)}, \bm{k}_+^{(s)} \rangle) + \sum\limits_{j^\prime=2}^M  \exp  ( \langle \bm{q}_+^{(s)}, \bm{k}_{n, j^\prime}^{(s)} \rangle ) } \big) \\
    &= \frac{\eta}{2NM} \sum\limits_{n \in S_+} - \ell_n^{\prime(s)} \Vert \bm{\mu} \Vert_2^2 V_+^{(s)} \cdot \frac{ \exp (\langle \bm{q}_+^{(s)}, \bm{k}_+^{(s)} \rangle)}{ \exp  (\langle \bm{q}_+^{(s)}, \bm{k}_+^{(s)} \rangle) + \sum\limits_{j^\prime=2}^M  \exp  ( \langle \bm{q}_+^{(s)}, \bm{k}_{n, j^\prime}^{(s)} \rangle ) } \\
    &\cdot \frac{ \sum\limits_{j^\prime=2}^M  \exp  ( \langle \bm{q}_+^{(s)}, \bm{k}_{n, j^\prime}^{(s)} \rangle ) }{ \exp  (\langle \bm{q}_+^{(s)}, \bm{k}_+^{(s)} \rangle) + \sum\limits_{j^\prime=2}^M  \exp  ( \langle \bm{q}_+^{(s)}, \bm{k}_{n, j^\prime}^{(s)} \rangle ) } \\
    &\ge \frac{\eta}{2NM} - \ell_n^{\prime(s)} \Vert \bm{\mu} \Vert_2^2 V_+^{(s)} \cdot \frac{ \exp (\langle \bm{q}_+^{(s)}, \bm{k}_+^{(s)} \rangle)}{ \exp  (\langle \bm{q}_+^{(s)}, \bm{k}_+^{(s)} \rangle) + \sum\limits_{j^\prime=2}^M  \exp  ( \langle \bm{q}_+^{(s)}, \bm{k}_{n, j^\prime}^{(s)} \rangle ) } \\
    &\cdot \frac{ \exp ( \langle \bm{q}_+^{(s)}, \bm{k}_{n, j}^{(s)} \rangle ) }{ \exp  (\langle \bm{q}_+^{(s)}, \bm{k}_+^{(s)} \rangle) + \sum\limits_{j^\prime=2}^M  \exp  ( \langle \bm{q}_+^{(s)}, \bm{k}_{n, j^\prime}^{(s)} \rangle ) } \\
    &\ge \frac{\eta}{2NM} - \ell_n^{\prime(s)} \Vert \bm{\mu} \Vert_2^2 V_+^{(s)} \cdot \big( \frac{1}{M} - o(1) \big) \\
    &\cdot \frac{ \exp ( \langle \bm{q}_+^{(s)}, \bm{k}_{n, j}^{(s)} \rangle ) }{ \exp  (\langle \bm{q}_+^{(s)}, \bm{k}_+^{(s)} \rangle) + \sum\limits_{j^\prime=2}^M  \exp  ( \langle \bm{q}_+^{(s)}, \bm{k}_{n, j^\prime}^{(s)} \rangle ) } \\
    &\ge \frac{\eta}{2NM} - \ell_n^{\prime(s)} \Vert \bm{\mu} \Vert_2^2 V_+^{(s)} \cdot (\frac{1}{M} - o(1)) \cdot \frac{ \exp ( \langle \bm{q}_+^{(s)}, \bm{k}_{n, j}^{(s)} \rangle ) }{ C \exp  (\langle \bm{q}_+^{(s)}, \bm{k}_+^{(s)} \rangle)} \\
    &\ge \frac{\eta^2 C_5 \Vert \bm{\mu} \Vert_2^4 \Vert \bm{w}_O \Vert_2^2 (s - T_1)}{N} \cdot \frac{1}{\exp(\Lambda_{n, +, j}^{(s)})},
\end{split}
\end{equation}
where the \(\langle \bm{q}_+^{(s)}, \bm{k}_{n, j}^{(s)} \rangle\) in the second inequality is a particular choice (we will characterize the dynamic of \( \langle \bm{q}_+^{(s)}, \bm{k}_+^{(s)} \rangle - \langle \bm{q}_+^{(s)}, \bm{k}_{n, j}^{(s)} \rangle \)), 
the third inequality is by \( softmax(\langle \bm{q}_+^{(s)}, \bm{k}_+^{(s)} \rangle) \ge \big( \frac{1}{M} - o(1) \big) \).
In the fourth inequality, by \( \langle \bm{q}^{(T_1)}, \bm{k}^{(T_1)} \rangle = o(1) \) and the monotonicity of \(\langle \bm{q}^{(s)}, \bm{k}^{(s)} \rangle\) (\(\langle \bm{q}_+^{(s)}, \bm{k}_+^{(s)} \rangle\) is increasing and \(\langle \bm{q}_+^{(s)}, \bm{k}_{n, j}^{(s)} \rangle\) is decreasing), there exist a constant C such that \( C \exp  (\langle \bm{q}_+^{(s)}, \bm{k}_+^{(s)} \rangle) \ge \exp(\langle \bm{q}_+^{(s)}, \bm{k}_+^{(s)} \rangle) + \sum\limits_{j^\prime=2}^M  \exp  ( \langle \bm{q}_+^{(s)}, \bm{k}_{n, j^\prime}^{(s)} \rangle ) \).
In the last inequality, we plugging the lower bounds of \(V_+^{(s)}\) and \(- \ell_n^{\prime(s)}\) and then absorb all the constant factors. Similarly, we have
\begin{equation}
    \label{lower_bound_beta_p_p}
    \beta_{+, +}^{(s)} \ge \frac{\eta^2 C_5 \Vert \bm{\mu} \Vert_2^4 \Vert \bm{w}_O \Vert_2^2 (s - T_1)}{N} \cdot \frac{1}{\exp(\Lambda_{n, +, j}^{(s)})},
\end{equation}
\begin{equation}
    \alpha_{-, -}^{(s)} \ge \frac{\eta^2 C_5 \Vert \bm{\mu} \Vert_2^4 \Vert \bm{w}_O \Vert_2^2 (s - T_1)}{N} \cdot \frac{1}{\exp(\Lambda_{n, -, j}^{(s)})},
\end{equation}
\begin{equation}
    \beta_{-, -}^{(s)} \ge \frac{\eta^2 C_5 \Vert \bm{\mu} \Vert_2^4 \Vert \bm{w}_O \Vert_2^2 (s - T_1)}{N} \cdot \frac{1}{\exp(\Lambda_{n, -, j}^{(s)})}.
\end{equation}

\begin{equation}
\begin{split}
    \label{lower_bound_alpha_p_j}
    &\alpha_{n, +, j}^{(s)} \\
    &\le \frac{\eta}{4NM} \ell_n^{\prime(s)} \Vert \bm{\mu} \Vert_2^2 V_+^{(s)} \cdot \frac{ \exp (\langle \bm{q}_+^{(s)}, \bm{k}_{n, j}^{(s)} \rangle)}{ \exp  (\langle \bm{q}_+^{(s)}, \bm{k}_+^{(s)} \rangle) + \sum\limits_{j^\prime=2}^M  \exp  ( \langle \bm{q}_+^{(s)}, \bm{k}_{n, j^\prime}^{(s)} \rangle ) } \\
    &\cdot \frac{ \exp (\langle \bm{q}_+^{(s)}, \bm{k}_+^{(s)} \rangle)}{ \exp  (\langle \bm{q}_+^{(s)}, \bm{k}_+^{(s)} \rangle) + \sum\limits_{j^\prime=2}^M  \exp  ( \langle \bm{q}_+^{(s)}, \bm{k}_{n, j^\prime}^{(s)} \rangle ) } \\
    &\le \frac{\eta}{4NM} \ell_n^{\prime(s)} \Vert \bm{\mu} \Vert_2^2 V_+^{(s)} \cdot \big( \frac{1}{M} - o(1) \big) \\
    &\cdot \frac{ \exp ( \langle \bm{q}_+^{(s)}, \bm{k}_{n, j}^{(s)} \rangle ) }{ \exp  (\langle \bm{q}_+^{(s)}, \bm{k}_+^{(s)} \rangle) + \sum\limits_{j^\prime=2}^M  \exp  ( \langle \bm{q}_+^{(s)}, \bm{k}_{n, j^\prime}^{(s)} \rangle ) } \\
    &\le \frac{\eta}{4NM} \ell_n^{\prime(s)} \Vert \bm{\mu} \Vert_2^2 V_+^{(s)} \cdot (\frac{1}{M} - o(1)) \cdot \frac{ \exp ( \langle \bm{q}_+^{(s)}, \bm{k}_{n, j}^{(s)} \rangle ) }{ C \exp  (\langle \bm{q}_+^{(s)}, \bm{k}_+^{(s)} \rangle)} \\
    &\le - \frac{\eta^2 C_5 \Vert \bm{\mu} \Vert_2^4 \Vert \bm{w}_O \Vert_2^2 (s - T_1)}{N} \cdot \frac{1}{\exp(\Lambda_{n, +, j}^{(s)})},   
\end{split}
\end{equation}
where the inequalities is similar to \eqref{lower_bound_alpha_p_p}. Similarly, we have
\begin{align*}
    \alpha_{n, -, j}^{(s)} \le - \frac{\eta^2 C_5 \Vert \bm{\mu} \Vert_2^4 \Vert \bm{w}_O \Vert_2^2 (s - T_1)}{N} \cdot \frac{1}{\exp(\Lambda_{n, -, j}^{(s)})}.
\end{align*}

\begin{equation}
\begin{split}
    \label{lower_bound_beta_p_i}
    &\beta_{n, +, i}^{(s)} \\
    &\ge - \frac{\eta \Vert \bm{\mu} \Vert_2^2}{2NM} \ell_n^{\prime(s)} V_+^{(s)} \frac{ \exp (\langle \bm{q}_{n, i}^{(s)}, \bm{k}_+^{(s)} \rangle)}{ \exp  (\langle \bm{q}_{n, i}^{(s)}, \bm{k}_+^{(s)} \rangle) + \sum\limits_{j^\prime=2}^M \exp  ( \langle \bm{q}_{n, i}^{(s)}, \bm{k}_{n, j^\prime}^{(s)} \rangle ) } \\
    &\cdot \big( 1 - \frac{ \exp (\langle \bm{q}_{n, i}^{(s)}, \bm{k}_+^{(s)} \rangle)}{ \exp  (\langle \bm{q}_{n, i}^{(s)}, \bm{k}_+^{(s)} \rangle) + \sum\limits_{j^\prime=2}^M  \exp  ( \langle \bm{q}_{n, i}^{(s)}, \bm{k}_{n, j^\prime}^{(s)} \rangle ) } \big) \\
    &\ge - \frac{\eta \Vert \bm{\mu} \Vert_2^2}{2NM} \ell_n^{\prime(s)} V_+^{(s)} \cdot \big( \frac{1}{M} - o(1) \big) \\
    &\cdot \frac{ \sum\limits_{j^\prime=2}^M  \exp  ( \langle \bm{q}_{n, i}^{(s)}, \bm{k}_{n, j^\prime}^{(s)} \rangle ) }{ \exp  (\langle \bm{q}_{n, i}^{(s)}, \bm{k}_+^{(s)} \rangle) + \sum\limits_{j^\prime=2}^M  \exp  ( \langle \bm{q}_{n, i}^{(s)}, \bm{k}_{n, j^\prime}^{(s)} \rangle ) } \\
    &\ge - \frac{\eta \Vert \bm{\mu} \Vert_2^2}{2NM} \ell_n^{\prime(s)} V_+^{(s)} \cdot \big( \frac{1}{M} - o(1) \big) \\
    &\cdot \frac{ \exp ( \langle \bm{q}_{n, i}^{(s)}, \bm{k}_{n, j}^{(s)} \rangle ) }{ \exp  (\langle \bm{q}_{n, i}^{(s)}, \bm{k}_+^{(s)} \rangle) + \sum\limits_{j^\prime=2}^M  \exp  ( \langle \bm{q}_{n, i}^{(s)}, \bm{k}_{n, j^\prime}^{(s)} \rangle ) } \\
    &\ge - \frac{\eta \Vert \bm{\mu} \Vert_2^2}{2NM} \ell_n^{\prime(s)} V_+^{(s)} \cdot \big( \frac{1}{M} - o(1) \big) \cdot \frac{ \exp ( \langle \bm{q}_{n, i}^{(s)}, \bm{k}_{n, j}^{(s)} \rangle ) }{ C \exp  (\langle \bm{q}_{n, i}^{(s)}, \bm{k}_+^{(s)} \rangle) } \\
    &\ge \frac{\eta^2 C_5 \Vert \bm{\mu} \Vert_2^4 \Vert \bm{w}_O \Vert_2^2 (s - T_1)}{N} \cdot \frac{1}{\exp(\Lambda_{n, i, +, j}^{(s)})},
\end{split}
\end{equation}
where the inequalities is similar to \eqref{lower_bound_alpha_p_p}, and \(\langle \bm{q}_{n, i}^{(s)}, \bm{k}_{n, j}^{(s)} \rangle\) is a particular choice (we will characterize the dynamic of \( \langle \bm{q}_{n, i}^{(s)}, \bm{k}_+^{(s)} \rangle - \langle \bm{q}_{n, i}^{(s)}, \bm{k}_{n, j}^{(s)} \rangle \)). Similarly, we have
\begin{equation}
    \beta_{n, -, i}^{(s)} \ge \frac{\eta^2 C_5 \Vert \bm{\mu} \Vert_2^4 \Vert \bm{w}_O \Vert_2^2 (s - T_1)}{N} \cdot \frac{1}{\exp(\Lambda_{n, i, -, j}^{(s)})},
\end{equation}
\begin{equation}
    \label{lower_bound_alpha_i_p}
    \alpha_{n, i, +}^{(s)} \ge \frac{\eta^2 C_5 \sigma_p^2 d \Vert \bm{\mu} \Vert_2^2 \Vert \bm{w}_O \Vert_2^2 (s - T_1) }{N} \cdot \frac{1}{\exp(\Lambda_{n, i, +, j}^{(s)})},
\end{equation}
\begin{equation}
    \alpha_{n, i, -}^{(s)} \ge \frac{\eta^2 C_5 \sigma_p^2 d \Vert \bm{\mu} \Vert_2^2 \Vert \bm{w}_O \Vert_2^2 (s - T_1) }{N} \cdot \frac{1}{\exp(\Lambda_{n, i, -, j}^{(s)})},
\end{equation}
\begin{equation}
    \label{lower_bound_beta_j_p}
    \beta_{n, j, +}^{(s)} \le - \frac{\eta^2 C_5 \sigma_p^2 d \Vert \bm{\mu} \Vert_2^2 \Vert \bm{w}_O \Vert_2^2 (s - T_1) }{N} \cdot \frac{1}{\exp(\Lambda_{n, +, j}^{(s)})},
\end{equation}
\begin{equation}
    \beta_{n, j, -}^{(s)} \le - \frac{\eta^2 C_5 \sigma_p^2 d \Vert \bm{\mu} \Vert_2^2 \Vert \bm{w}_O \Vert_2^2 (s - T_1) }{N} \cdot \frac{1}{\exp(\Lambda_{n, -, j}^{(s)})},
\end{equation}
\begin{equation}
    \label{lower_bound_alpha_i_j}
    \alpha_{n, i, n, j}^{(s)} \le - \frac{\eta^2 C_5 \sigma_p^2 d \Vert \bm{\mu} \Vert_2^2 \Vert \bm{w}_O \Vert_2^2 (s - T_1) }{N} \cdot \frac{1}{\exp(\Lambda_{n, i, \pm, j}^{(s)})},
\end{equation}
\begin{equation}
    \label{lower_bound_beta_j_i}
    \beta_{n, j, n, i}^{(s)} \le - \frac{\eta^2 C_5 \sigma_p^2 d \Vert \bm{\mu} \Vert_2^2 \Vert \bm{w}_O \Vert_2^2 (s - T_1) }{N} \cdot \frac{1}{\exp(\Lambda_{n, i, \pm, j}^{(s)})}.
\end{equation}

With the concise lower bounds for \(\alpha\) and \(\beta\) above and proposition \(\mathcal{C}(s)\), we will give the lower bounds for the dynamics of \( \langle \bm{q}, \bm{k} \rangle \).

\begin{equation}
\begin{split}
    &\langle \bm{q}_+^{(s+1)}, \bm{k}_+^{(s+1)} \rangle - \langle \bm{q}_+^{(s)}, \bm{k}_+^{(s)} \rangle \\
    &= \alpha_{+, +}^{(s)} \Vert \bm{k}_+^{(s)} \Vert_2^2 + \sum\limits_{n \in S_+} \sum\limits_{i=2}^M \alpha_{n, +, i}^{(s)} \langle \bm{k}_+^{(s)}, \bm{k}_{n, i}^{(s)} \rangle \\
    &+ \beta_{+, +}^{(s)} \Vert \bm{q}_+^{(s)} \Vert_2^2 + \sum\limits_{n \in S_+} \sum\limits_{i=2}^M \beta_{n, +, i}^{(s)} \langle \bm{q}_+^{(s)}, \bm{q}_{n, i}^{(s)} \rangle \\
    &+ \Big( \alpha_{+, +}^{(s)} \bm{k}_+^{(s)} + \sum\limits_{n \in S_+} \sum\limits_{i=2}^M \alpha_{n, +, i}^{(s)} \bm{k}_{n, i}^{(s)} \Big) \\
    &\cdot \Big( \beta_{+, +}^{(s)} \bm{q}_+^{(s)\top} + \sum\limits_{n \in S_+} \sum\limits_{i=2}^M \beta_{n, +, i}^{(s)} \bm{q}_{n, i}^{(s)\top} \Big) \\
    &= \alpha_{+, +}^{(s)} \Vert \bm{k}_+^{(s)} \Vert_2^2 + \beta_{+, +}^{(s)} \Vert \bm{q}_+^{(s)} \Vert_2^2 + \{ lower \ order \ term \} \\
    &\ge \frac{2\eta^2 C_5 \Vert \bm{\mu} \Vert_2^4 \Vert \bm{w}_O \Vert_2^2 (s - T_1)}{N} \cdot \frac{1}{\exp(\Lambda_{n, +, j}^{(s)})} \cdot \Theta (\Vert \bm{\mu} \Vert_2^2 \sigma_h^2 d_h) \\
    &+ \{ lower \ order \ term \} \\
    &\ge \frac{\eta^2 C_6 \Vert \bm{\mu} \Vert_2^6 \Vert \bm{w}_O \Vert_2^2 \sigma_h^2 d_h (s - T_1)}{N} \cdot \frac{1}{\exp(\Lambda_{n, +, j}^{(s)})},
\end{split}
\end{equation}
where the first inequality is by \eqref{lower_bound_alpha_p_p}, \eqref{lower_bound_beta_p_p}, the bound of \(\Vert \bm{q}_+^{(s)} \Vert_2^2\), \(\Vert \bm{k}_+^{(s)} \Vert_2^2\) in stage \uppercase\expandafter{\romannumeral2}, the second inequality is by absorbing the \(\{ lower \ order \ term \}\) and the constant factors. Similarly, we have
\begin{equation}
\begin{split}
    &\langle \bm{q}_-^{(s+1)}, \bm{k}_-^{(s+1)} \rangle - \langle \bm{q}_-^{(s)}, \bm{k}_-^{(s)} \rangle \\
    &\ge \frac{\eta^2 C_6 \Vert \bm{\mu} \Vert_2^6 \Vert \bm{w}_O \Vert_2^2 \sigma_h^2 d_h (s - T_1)}{N} \cdot \frac{1}{\exp(\Lambda_{n, -, j}^{(s)})}.
\end{split}
\end{equation}

\begin{equation}
\begin{split}
    &\langle \bm{q}_+^{(s+1)}, \bm{k}_{n, j}^{(s+1)} \rangle - \langle \bm{q}_+^{(s)}, \bm{k}_{n, j}^{(s)} \rangle \\
    &= \alpha_{+, +}^{(s)} \langle \bm{k}_+^{(s)}, \bm{k}_{n, j}^{(s)} \rangle + \sum\limits_{n^\prime \in S_+} \sum\limits_{l=2}^M \alpha_{n^\prime, +, l}^{(s)} \langle \bm{k}_{n, j}^{(s)}, \bm{k}_{n^\prime, l}^{(s)} \rangle \\
    &+ \beta_{n, j, +}^{(s)} \Vert \bm{q}_+^{(s)} \Vert_2^2 + \beta_{n, j, -}^{(s)} \langle \bm{q}_+^{(s)}, \bm{q}_-^{(s)} \rangle + \sum\limits_{n^\prime = 1}^N \sum\limits_{l=2}^M \beta_{n, j, n^\prime, l}^{(s)} \langle \bm{q}_+^{(s)}, \bm{q}_{n^\prime, l}^{(s)} \rangle \\
    &+ \Big( \alpha_{+, +}^{(s)} \bm{k}_+^{(s)} + \sum\limits_{n^\prime \in S_+} \sum\limits_{l=2}^M \alpha_{n^\prime, +, l}^{(s)} \bm{k}_{n^\prime, l}^{(s)} \Big) \\
    &\cdot \Big( \beta_{n, j, +}^{(s)} \bm{q}_+^{(s)\top} + \beta_{n, j, -}^{(s)} \bm{q}_-^{(s)\top} + \sum\limits_{n^\prime = 1}^N \sum\limits_{l=2}^M \beta_{n, j, n^\prime, l}^{(s)} \bm{q}_{n^\prime, l}^{(s)\top} \Big) \\
    &= \alpha_{n, +, j}^{(s)} \Vert \bm{k}_{n, j}^{(s)} \Vert_2^2 + \beta_{n, j, +}^{(s)} \Vert \bm{q}_+^{(s)} \Vert_2^2 + \{ lower \ order \ term \} \\
    &\le - \frac{\eta^2 C_5 \Vert \bm{\mu} \Vert_2^4 \Vert \bm{w}_O \Vert_2^2 (s - T_1)}{N} \cdot \frac{1}{\exp(\Lambda_{n, +, j}^{(s)})} \cdot \Theta \Big( \sigma_p^2 \sigma_h^2 d d_h \Big) \\
    &- \frac{\eta^2 C_5 \sigma_p^2 d \Vert \bm{\mu} \Vert_2^2 \Vert \bm{w}_O \Vert_2^2 (s - T_1) }{N} \cdot \frac{1}{\exp(\Lambda_{n, +, j}^{(s)})} \cdot \Theta \Big( \Vert \bm{\mu} \Vert_2^2 \sigma_h^2 d_h \Big) \\
    &+ \{ lower \ order \ term \} \\
    &\le - \frac{\eta^2 C_6 \sigma_p^2 d \Vert \bm{\mu} \Vert_2^4 \Vert \bm{w}_O \Vert_2^2 \sigma_h^2 d_h (s - T_1)}{N} \cdot \frac{1}{\exp(\Lambda_{n, +, j}^{(s)})},
\end{split}
\end{equation}
where the first inequality is by \eqref{lower_bound_alpha_p_j}, \eqref{lower_bound_beta_j_p}, the bound of \(\Vert \bm{k}_{n, j}^{(s)} \Vert_2^2\), \(\Vert \bm{q}_+^{(s)} \Vert_2^2\) in stage \uppercase\expandafter{\romannumeral2}, the second inequality is by absorbing the \(\{ lower \ order \ term \}\) and the constant factors. Similarly, we have
\begin{equation}
\begin{split}
    &\langle \bm{q}_-^{(s+1)}, \bm{k}_{n, j}^{(s+1)} \rangle - \langle \bm{q}_-^{(s)}, \bm{k}_{n, j}^{(s)} \rangle \\
    &\le - \frac{\eta^2 C_6 \sigma_p^2 d \Vert \bm{\mu} \Vert_2^4 \Vert \bm{w}_O \Vert_2^2 \sigma_h^2 d_h (s - T_1)}{N} \cdot \frac{1}{\exp(\Lambda_{n, -, j}^{(s)})}.
\end{split}
\end{equation}

\begin{equation}
\begin{split}
    &\langle \bm{q}_{n, i}^{(s+1)}, \bm{k}_+^{(s+1)} \rangle - \langle \bm{q}_{n, i}^{(s)}, \bm{k}_+^{(s)} \rangle \\
    &= \alpha_{n, i, +}^{(s)} \Vert \bm{k}_+^{(s)} \Vert_2^2 + \alpha_{n, i, -}^{(s)} \langle \bm{k}_+^{(s)}, \bm{k}_-^{(s)} \rangle + \sum\limits_{n^\prime =1}^N \sum\limits_{l=2}^M \alpha_{n, i, n^\prime, l}^{(s)} \langle \bm{k}_+^{(s)}, \bm{k}_{n^\prime, l}^{(s)} \rangle \\
    &+ \beta_{+, +}^{(s)} \langle \bm{q}_+^{(s)}, \bm{q}_{n, i}^{(s)} \rangle + \sum\limits_{n^\prime \in S_+} \sum\limits_{l=2}^M \beta_{n^\prime, +, l}^{(s)} \langle \bm{q}_{n, i}^{(s)}, \bm{q}_{n^\prime, l}^{(s)} \rangle \\
    &+ \Big( \alpha_{n, i, +}^{(s)} \bm{k}_+^{(s)} + \alpha_{n, i, -}^{(s)} \bm{k}_-^{(s)} + \sum\limits_{n^\prime =1}^N \sum\limits_{l=2}^M \alpha_{n, i, n^\prime, l}^{(s)} \bm{k}_{n^\prime, l}^{(s)} \Big) \\
    &\cdot \Big( \beta_{+, +}^{(s)} \bm{q}_+^{(s)\top} + \sum\limits_{n^\prime \in S_+} \sum\limits_{l=2}^M \beta_{n^\prime, +, l}^{(s)} \bm{q}_{n^\prime, l}^{(s)\top} \Big) \\
    &= \alpha_{n, i, +}^{(s)} \Vert \bm{k}_+^{(s)} \Vert_2^2 + \beta_{n, +, i}^{(s)} \Vert \bm{q}_{n, i}^{(s)} \Vert_2^2 + \{ lower \ order \ term \} \\
    &\ge \frac{\eta^2 C_5 \sigma_p^2 d \Vert \bm{\mu} \Vert_2^2 \Vert \bm{w}_O \Vert_2^2 (s - T_1) }{N} \cdot \frac{1}{\exp(\Lambda_{n, i, +, j}^{(s)})} \cdot \Theta \Big( \Vert \bm{\mu} \Vert_2^2 \sigma_h^2 d_h \Big) \\
    &+ \frac{\eta^2 C_5 \Vert \bm{\mu} \Vert_2^4 \Vert \bm{w}_O \Vert_2^2 (s - T_1)}{N} \cdot \frac{1}{\exp(\Lambda_{n, i, +, j}^{(s)})} \cdot \Theta \Big( \sigma_p^2 \sigma_h^2 d d_h \Big) \\
    &+ \{ lower \ order \ term \} \\
    &\ge \frac{\eta^2 C_6 \sigma_p^2 d \Vert \bm{\mu} \Vert_2^4 \Vert \bm{w}_O \Vert_2^2 \sigma_h^2 d_h (s - T_1)}{N} \cdot \frac{1}{\exp(\Lambda_{n, i, +, j}^{(s)})},
\end{split}
\end{equation}
where the first inequality is by \eqref{lower_bound_alpha_i_p}, \eqref{lower_bound_beta_p_i}, the bound of \(\Vert \bm{k}_+^{(s)} \Vert_2^2\), \(\Vert \bm{q}_{n, i}^{(s)} \Vert_2^2\) in stage \uppercase\expandafter{\romannumeral2}, the second inequality is by absorbing the \(\{ lower \ order \ term \}\) and the constant factors. Similarly, we have 
\begin{equation}
\begin{split}
    &\langle \bm{q}_{n, i}^{(s+1)}, \bm{k}_-^{(s+1)} \rangle - \langle \bm{q}_{n, i}^{(s)}, \bm{k}_-^{(s)} \rangle \\
    &\ge \frac{\eta^2 C_6 \sigma_p^2 d \Vert \bm{\mu} \Vert_2^4 \Vert \bm{w}_O \Vert_2^2 \sigma_h^2 d_h (s - T_1)}{N} \cdot \frac{1}{\exp(\Lambda_{n, i, -, j}^{(s)})}.
\end{split}
\end{equation}

\begin{equation}
\begin{split}
    &\langle \bm{q}_{n, i}^{(s+1)}, \bm{k}_{n, j}^{(s+1)} \rangle - \langle \bm{q}_{n, i}^{(s)}, \bm{k}_{n, j}^{(s)} \rangle \\
    &= \alpha_{n, i, +}^{(s)} \langle \bm{k}_+^{(s)}, \bm{k}_{n, j}^{(s)} \rangle + \alpha_{n, i, -}^{(s)} \langle \bm{k}_-^{(s)}, \bm{k}_{n, j}^{(s)} \rangle + \sum\limits_{n^\prime =1}^N \sum\limits_{l=2}^M \alpha_{n, i, n^\prime, l}^{(s)} \langle \bm{k}_{n^\prime, l}^{(s)}, \bm{k}_{n, j}^{(s)} \rangle \\
    &+ \beta_{n, j, +}^{(s)} \langle \bm{q}_+^{(s)}, \bm{q}_{n, i}^{(s)} \rangle + \beta_{n, j, -}^{(s)} \langle \bm{q}_-^{(s)}, \bm{q}_{n, i}^{(s)} \rangle + \sum\limits_{n^\prime = 1}^N \sum\limits_{l=2}^M \beta_{n, j, n^\prime, l}^{(s)} \langle \bm{q}_{n^\prime, l}^{(s)}, \bm{q}_{n, i}^{(s)} \rangle \\
    &+ \Big( \alpha_{n, i, +}^{(s)} \bm{k}_+^{(s)} + \alpha_{n, i, -}^{(s)} \bm{k}_-^{(s)} + \sum\limits_{n^\prime =1}^N \sum\limits_{l=2}^M \alpha_{n, i, n^\prime, l}^{(s)} \bm{k}_{n^\prime, l}^{(s)} \Big) \\
    &\cdot \Big( \beta_{n, j, +}^{(s)} \bm{q}_+^{(s)\top} + \beta_{n, j, -}^{(s)} \bm{q}_-^{(s)\top} + \sum\limits_{n^\prime = 1}^N \sum\limits_{l=2}^M \beta_{n, j, n^\prime, l}^{(s)} \bm{q}_{n^\prime, l}^{(s)\top} \Big) \\
    &= \alpha_{n, i, n, j}^{(s)} \Vert \bm{k}_{n, j}^{(s)} \Vert_2^2 + \beta_{n, j, n, i}^{(s)} \Vert \bm{q}_{n, i}^{(s)} \Vert_2^2 \\
    &+ \{ lower \ order \ term \} \\
    &\le - \frac{2 \eta^2 C_5 \sigma_p^2 d \Vert \bm{\mu} \Vert_2^2 \Vert \bm{w}_O \Vert_2^2 \sigma_h^2 d_h (s - T_1) }{N} \cdot \frac{1}{\exp(\Lambda_{n, i, \pm, j}^{(s)})} \cdot \Theta \Big( \sigma_p^2 \sigma_h^2 d d_h \Big) \\
    &+ \{ lower \ order \ term \} \\
    &\le - \frac{\eta^2 C_6 \sigma_p^4 d^2 \Vert \bm{\mu} \Vert_2^2 \Vert \bm{w}_O \Vert_2^2 \sigma_h^2 d_h (s - T_1) }{N} \cdot \frac{1}{\exp(\Lambda_{n, i, \pm, j}^{(s)})},
\end{split}
\end{equation}
where the first inequality is by \eqref{lower_bound_alpha_i_j}, \eqref{lower_bound_beta_j_i}, the bound of \(\Vert \bm{k}_{n, j}^{(s)} \Vert_2^2\), \(\Vert \bm{q}_{n, i}^{(s)} \Vert_2^2\) in stage \uppercase\expandafter{\romannumeral2}, the second inequality is by absorbing the \(\{ lower \ order \ term \}\) and the constant factors.

\subsection{Upper Bounds of \(\langle\) q, k \(\rangle\)}
\label{upper_bound_of_qk}
In order to give the upper bounds of \( \langle \bm{q}, \bm{k} \rangle \) in stage \uppercase\expandafter{\romannumeral2}, we need to give the upper bounds of \(\alpha\) and \(\beta\) based on the equations in \ref{alpha_and_beta} under the assumption that \(\mathcal{D}(T_1), \dots, \mathcal{D}(s - 1)\) hold for \(s \in [T_1, t]\).
\begin{align*}
    &\alpha_{+, +}^{(s)} = \frac{\eta}{NM} \sum\limits_{n \in S_+} - \ell_n^{\prime(s)} \Vert \bm{\mu} \Vert_2^2 \\
    &\cdot \Big( V_+^{(s)} \big( \frac{ \exp (\langle \bm{q}_+^{(s)}, \bm{k}_+^{(s)} \rangle)}{ \exp  (\langle \bm{q}_+^{(s)}, \bm{k}_+^{(s)} \rangle) + \sum\limits_{j=2}^M  \exp  ( \langle \bm{q}_+^{(s)}, \bm{k}_{n, j}^{(s)} \rangle ) } \\
    &- (\frac{ \exp (\langle \bm{q}_+^{(s)}, \bm{k}_+^{(s)} \rangle)}{ \exp  (\langle \bm{q}_+^{(s)}, \bm{k}_+^{(s)} \rangle) + \sum\limits_{j=2}^M  \exp  ( \langle \bm{q}_+^{(s)}, \bm{k}_{n, j}^{(s)} \rangle ) })^2 \big) \\
    &- \sum\limits_{i=2}^M \big( V_{n, i}^{(s)} \cdot \frac{ \exp (\langle \bm{q}_+^{(s)}, \bm{k}_+^{(s)} \rangle)}{ \exp  (\langle \bm{q}_+^{(s)}, \bm{k}_+^{(s)} \rangle) + \sum\limits_{j=2}^M  \exp  ( \langle \bm{q}_+^{(s)}, \bm{k}_{n, j}^{(s)} \rangle ) } \\
    &\cdot \frac{ \exp (\langle \bm{q}_+^{(s)}, \bm{k}_{n, i}^{(s)} \rangle)}{ \exp  (\langle \bm{q}_+^{(s)}, \bm{k}_+^{(s)} \rangle) + \sum\limits_{j=2}^M  \exp  ( \langle \bm{q}_+^{(s)}, \bm{k}_{n, j}^{(s)} \rangle ) } \big) \Big) \\
    &= \frac{\eta}{NM} \sum\limits_{n \in S_+} - \ell_n^{\prime(s)} \Vert \bm{\mu} \Vert_2^2 \frac{ \exp (\langle \bm{q}_+^{(s)}, \bm{k}_+^{(s)} \rangle)}{ \exp  (\langle \bm{q}_+^{(s)}, \bm{k}_+^{(s)} \rangle) + \sum\limits_{j=2}^M  \exp  ( \langle \bm{q}_+^{(s)}, \bm{k}_{n, j}^{(s)} \rangle ) } \\
    &\cdot \Big( V_+^{(s)} \cdot \frac{ \sum\limits_{j=2}^M  \exp  ( \langle \bm{q}_+^{(s)}, \bm{k}_{n, j}^{(s)} \rangle ) }{ \exp  (\langle \bm{q}_+^{(s)}, \bm{k}_+^{(s)} \rangle) + \sum\limits_{j=2}^M  \exp  ( \langle \bm{q}_+^{(s)}, \bm{k}_{n, j}^{(s)} \rangle ) } \\
    &- \sum\limits_{i=2}^M V_{n, i}^{(s)} \cdot \frac{ \exp (\langle \bm{q}_+^{(s)}, \bm{k}_{n, i}^{(s)} \rangle)}{ \exp  (\langle \bm{q}_+^{(s)}, \bm{k}_+^{(s)} \rangle) + \sum\limits_{j=2}^M  \exp  ( \langle \bm{q}_+^{(s)}, \bm{k}_{n, j}^{(s)} \rangle ) } \Big) \\
    &\le \frac{\eta}{NM} \sum\limits_{n \in S_+} \Vert \bm{\mu} \Vert_2^2 \cdot \Big( V_+^{(s)} \cdot \frac{ \sum\limits_{j=2}^M  \exp  ( \langle \bm{q}_+^{(s)}, \bm{k}_{n, j}^{(s)} \rangle ) }{ \exp  (\langle \bm{q}_+^{(s)}, \bm{k}_+^{(s)} \rangle) + \sum\limits_{j=2}^M  \exp  ( \langle \bm{q}_+^{(s)}, \bm{k}_{n, j}^{(s)} \rangle ) } \\
    &+ \max\limits_{i} \vert V_{n, i}^{(s)} \vert \cdot \frac{ \sum\limits_{j=2}^M  \exp  ( \langle \bm{q}_+^{(s)}, \bm{k}_{n, j}^{(s)} \rangle ) }{ \exp  (\langle \bm{q}_+^{(s)}, \bm{k}_+^{(s)} \rangle) + \sum\limits_{j=2}^M  \exp  ( \langle \bm{q}_+^{(s)}, \bm{k}_{n, j}^{(s)} \rangle ) } \Big) \\
    &\le \frac{\eta}{NM} \cdot \frac{3N}{4} \cdot \Vert \bm{\mu} \Vert_2^2 \cdot \Big( V_+^{(s)} \cdot \frac{ C }{ \exp  (\langle \bm{q}_+^{(s)}, \bm{k}_+^{(s)} \rangle) } + \max\limits_{i} \vert V_{n, i}^{(s)} \vert \cdot \frac{ C }{ \exp  (\langle \bm{q}_+^{(s)}, \bm{k}_+^{(s)} \rangle) } \Big) \\
    &\le \frac{\eta C_9 \Vert \bm{\mu} \Vert_2^2 }{\exp ( \langle \bm{q}_+^{(s)}, \bm{k}_+^{(s)} \rangle)},
\end{align*}
where the first inequality is by \(- \ell_n^{\prime(s)} \le 1\) and \(softmax(\langle \bm{q}_+^{(s)}, \bm{k}_+^{(s)} \rangle) \le 1\). For the second inequality, we first consider \(\frac{ \sum\limits_{j=2}^M  \exp  ( \langle \bm{q}_+^{(s)}, \bm{k}_{n, j}^{(s)} \rangle ) }{ \exp  (\langle \bm{q}_+^{(s)}, \bm{k}_+^{(s)} \rangle) + \sum\limits_{j=2}^M  \exp  ( \langle \bm{q}_+^{(s)}, \bm{k}_{n, j}^{(s)} \rangle ) } \le \frac{ \sum\limits_{j=2}^M  \exp  ( \langle \bm{q}_+^{(s)}, \bm{k}_{n, j}^{(s)} \rangle ) }{ \exp  (\langle \bm{q}_+^{(s)}, \bm{k}_+^{(s)} \rangle) }\), then by the monotonicity of \(\langle \bm{q}_+^{(s)}, \bm{k}_{n, j}^{(s)} \rangle\) and \(\langle \bm{q}_+^{(T_1)}, \bm{k}_{n, j}^{(T_1)} \rangle = o(1)\) we have \(\sum\limits_{j=2}^M  \exp  ( \langle \bm{q}_+^{(s)}, \bm{k}_{n, j}^{(s)} \rangle ) \le C\) for \(s \in [T_1, t]\). The last inequality is by \(V_+^{(s)}, V_{n, i}^{(s)} = o(1)\) for \(s \in [T_1, t]\) and absorbing the constant factors. Similarly, we have
\begin{equation*}
    \alpha_{-, -}^{(s)} \le \frac{\eta C_9 \Vert \bm{\mu} \Vert_2^2 }{\exp ( \langle \bm{q}_-^{(s)}, \bm{k}_-^{(s)} \rangle)},
\end{equation*}
\begin{equation*}
    \beta_{+, +}^{(s)} \le \frac{\eta C_9 \Vert \bm{\mu} \Vert_2^2 }{\exp ( \langle \bm{q}_+^{(s)}, \bm{k}_+^{(s)} \rangle)},
\end{equation*}
\begin{equation*}
    \beta_{-, -}^{(s)} \le \frac{\eta C_9 \Vert \bm{\mu} \Vert_2^2 }{\exp ( \langle \bm{q}_-^{(s)}, \bm{k}_-^{(s)} \rangle)}.
\end{equation*}

\begin{align*}
    &\alpha_{n, +, j}^{(s)} = - \frac{\eta}{NM} \ell_n^{\prime(s)} \Vert \bm{\mu} \Vert_2^2 \\
    &\cdot \Big( - V_+^{(s)} \cdot \frac{ \exp (\langle \bm{q}_+^{(s)}, \bm{k}_+^{(s)} \rangle)}{ \exp  (\langle \bm{q}_+^{(s)}, \bm{k}_+^{(s)} \rangle) + \sum\limits_{j^\prime=2}^M  \exp  ( \langle \bm{q}_+^{(s)}, \bm{k}_{n, j^\prime}^{(s)} \rangle ) } \\
    &\cdot \frac{ \exp (\langle \bm{q}_+^{(s)}, \bm{k}_{n, j}^{(s)} \rangle)}{ \exp  (\langle \bm{q}_+^{(s)}, \bm{k}_+^{(s)} \rangle) + \sum\limits_{j^\prime=2}^M  \exp  ( \langle \bm{q}_+^{(s)}, \bm{k}_{n, j^\prime}^{(s)} \rangle ) } \\
    &+ V_{n, i}^{(s)} \big( \frac{ \exp (\langle \bm{q}_+^{(s)}, \bm{k}_{n, j}^{(s)} \rangle)}{ \exp  (\langle \bm{q}_+^{(s)}, \bm{k}_+^{(s)} \rangle) + \sum\limits_{j^\prime=2}^M  \exp  ( \langle \bm{q}_+^{(s)}, \bm{k}_{n, j^\prime}^{(s)} \rangle ) } \\
    &- (\frac{ \exp (\langle \bm{q}_+^{(s)}, \bm{k}_{n, j}^{(s)} \rangle)}{ \exp  (\langle \bm{q}_+^{(s)}, \bm{k}_+^{(s)} \rangle) + \sum\limits_{j^\prime=2}^M  \exp  ( \langle \bm{q}_+^{(s)}, \bm{k}_{n, j^\prime}^{(s)} \rangle ) })^2 \big) \\
    &- \sum\limits_{k \ne j} \big( V_{n, k}^{(s)} \cdot \frac{ \exp (\langle \bm{q}_+^{(s)}, \bm{k}_{n, j}^{(s)} \rangle)}{ \exp  (\langle \bm{q}_+^{(s)}, \bm{k}_+^{(s)} \rangle) + \sum\limits_{j^\prime=2}^M  \exp  ( \langle \bm{q}_+^{(s)}, \bm{k}_{n, j^\prime}^{(s)} \rangle ) } \\
    & \cdot \frac{ \exp (\langle \bm{q}_+^{(s)}, \bm{k}_{n, k}^{(s)} \rangle)}{ \exp  (\langle \bm{q}_+^{(s)}, \bm{k}_+^{(s)} \rangle) + \sum\limits_{j^\prime=2}^M  \exp  ( \langle \bm{q}_+^{(s)}, \bm{k}_{n, j^\prime}^{(s)} \rangle ) } \big) \Big) \\
    &\ge - \frac{\eta}{NM} \ell_n^{\prime(s)} \Vert \bm{\mu} \Vert_2^2 \cdot \frac{ \exp (\langle \bm{q}_+^{(s)}, \bm{k}_{n, j}^{(s)} \rangle)}{ \exp  (\langle \bm{q}_+^{(s)}, \bm{k}_+^{(s)} \rangle) + \sum\limits_{j^\prime=2}^M  \exp  ( \langle \bm{q}_+^{(s)}, \bm{k}_{n, j^\prime}^{(s)} \rangle ) } \\
    &\cdot \Big( - V_+^{(s)} \cdot \frac{ \exp (\langle \bm{q}_+^{(s)}, \bm{k}_+^{(s)} \rangle)}{ \exp  (\langle \bm{q}_+^{(s)}, \bm{k}_+^{(s)} \rangle) + \sum\limits_{j^\prime=2}^M  \exp  ( \langle \bm{q}_+^{(s)}, \bm{k}_{n, j^\prime}^{(s)} \rangle ) } \\
    &- \vert V_{n, i}^{(s)} \vert \big( 1 - \frac{ \exp (\langle \bm{q}_+^{(s)}, \bm{k}_{n, j}^{(s)} \rangle)}{ \exp  (\langle \bm{q}_+^{(s)}, \bm{k}_+^{(s)} \rangle) + \sum\limits_{j^\prime=2}^M  \exp  ( \langle \bm{q}_+^{(s)}, \bm{k}_{n, j^\prime}^{(s)} \rangle ) } \big) \\
    &- \max\limits_{l} \vert V_{n, l}^{(s)} \vert \cdot \frac{ \sum\limits_{k \ne j} \exp (\langle \bm{q}_+^{(s)}, \bm{k}_{n, k}^{(s)} \rangle)}{ \exp  (\langle \bm{q}_+^{(s)}, \bm{k}_+^{(s)} \rangle) + \sum\limits_{j^\prime=2}^M  \exp  ( \langle \bm{q}_+^{(s)}, \bm{k}_{n, j^\prime}^{(s)} \rangle ) } \Big) \\
    &\ge - \frac{2\eta}{NM} \Vert \bm{\mu} \Vert_2^2 V_+^{(s)} \cdot \frac{ \exp (\langle \bm{q}_+^{(s)}, \bm{k}_{n, j}^{(s)} \rangle)}{ \exp (\langle \bm{q}_+^{(s)}, \bm{k}_+^{(s)} \rangle) + \sum\limits_{j^\prime=2}^M  \exp  ( \langle \bm{q}_+^{(s)}, \bm{k}_{n, j^\prime}^{(s)} \rangle ) } \\
    &\ge - \frac{2\eta}{NM} \Vert \bm{\mu} \Vert_2^2 V_+^{(s)} \cdot \frac{ \exp (\langle \bm{q}_+^{(s)}, \bm{k}_{n, j}^{(s)} \rangle)}{ C } \\
    &\ge - \frac{\eta C_9 \Vert \bm{\mu} \Vert_2^2}{N} \cdot \exp (\langle \bm{q}_+^{(s)}, \bm{k}_{n, j}^{(s)} \rangle),
\end{align*}
where the second inequality is by \(V_+^{(s)} \ge 3 M \cdot \vert V_{n, i}^{(s)} \vert\), \(- \ell_n^{\prime(s)} \le 1\) and the property that attention \(<1\). For the third inequality, we consider \(\frac{ \exp (\langle \bm{q}_+^{(s)}, \bm{k}_{n, j}^{(s)} \rangle)}{ \exp (\langle \bm{q}_+^{(s)}, \bm{k}_+^{(s)} \rangle) + \sum\limits_{j^\prime=2}^M  \exp  ( \langle \bm{q}_+^{(s)}, \bm{k}_{n, j^\prime}^{(s)} \rangle ) } \le \frac{ \exp (\langle \bm{q}_+^{(s)}, \bm{k}_{n, j}^{(s)} \rangle)}{ \exp (\langle \bm{q}_+^{(s)}, \bm{k}_+^{(s)} \rangle) }\) first, then by the monotonicity of \(\langle \bm{q}_+^{(s)}, \bm{k}_+^{(s)} \rangle\) and \(\langle \bm{q}_+^{(T_1)}, \bm{k}_+^{(T_1)} \rangle\) we have \(\exp (\langle \bm{q}_+^{(s)}, \bm{k}_+^{(s)} \rangle) \ge C\) for \(s \in [T_1, t]\). The last inequality is by \(V_+^{(s)} = o(1)\) for \(s \in [T_1, t]\) and absorbing the constant factors. Similarly, we have
\begin{align*}
    \alpha_{n, -, j}^{(s)} \ge - \frac{\eta C_9 \Vert \bm{\mu} \Vert_2^2}{N} \cdot \exp (\langle \bm{q}_-^{(s)}, \bm{k}_{n, j}^{(s)} \rangle).
\end{align*}

\begin{align*}
    &\beta_{n, +, i}^{(s)} = - \frac{\eta \Vert \bm{\mu} \Vert_2^2}{NM} \ell_n^{\prime(s)} \\
    &\cdot \Big( V_+^{(s)} \big( \frac{ \exp (\langle \bm{q}_{n, i}^{(s)}, \bm{k}_+^{(s)} \rangle)}{ \exp  (\langle \bm{q}_{n, i}^{(s)}, \bm{k}_+^{(s)} \rangle) + \sum\limits_{j=2}^M  \exp  ( \langle \bm{q}_{n, i}^{(s)}, \bm{k}_{n, j}^{(s)} \rangle ) } \\
    & - (\frac{ \exp (\langle \bm{q}_{n, i}^{(s)}, \bm{k}_+^{(s)} \rangle)}{ \exp  (\langle \bm{q}_{n, i}^{(s)}, \bm{k}_+^{(s)} \rangle) + \sum\limits_{j=2}^M  \exp  ( \langle \bm{q}_{n, i}^{(s)}, \bm{k}_{n, j}^{(s)} \rangle ) })^2 \big) \\
    & -\sum\limits_{k=2}^M \big( V_{n, i}^{(s)} \cdot \frac{ \exp (\langle \bm{q}_{n, i}^{(s)}, \bm{k}_+^{(s)} \rangle)}{ \exp  (\langle \bm{q}_{n, i}^{(s)}, \bm{k}_+^{(s)} \rangle) + \sum\limits_{j=2}^M  \exp  ( \langle \bm{q}_{n, i}^{(s)}, \bm{k}_{n, j}^{(s)} \rangle ) } \\
    & \cdot \frac{ \exp (\langle \bm{q}_{n, i}^{(s)}, \bm{k}_{n, k}^{(s)} \rangle)}{ \exp  (\langle \bm{q}_{n, i}^{(s)}, \bm{k}_+^{(s)} \rangle) + \sum\limits_{j=2}^M  \exp  ( \langle \bm{q}_{n, i}^{(s)}, \bm{k}_{n, j}^{(s)} \rangle ) } \big) \Big) \\
    &=  - \frac{\eta \Vert \bm{\mu} \Vert_2^2}{NM} \ell_n^{\prime(s)} \frac{ \exp (\langle \bm{q}_{n, i}^{(s)}, \bm{k}_+^{(s)} \rangle)}{ \exp  (\langle \bm{q}_{n, i}^{(s)}, \bm{k}_+^{(s)} \rangle) + \sum\limits_{j=2}^M \exp ( \langle \bm{q}_{n, i}^{(s)}, \bm{k}_{n, j}^{(s)} \rangle ) } \\
    &\cdot \Big( V_+^{(s)} \big( 1 - \frac{ \exp (\langle \bm{q}_{n, i}^{(s)}, \bm{k}_+^{(s)} \rangle)}{ \exp  (\langle \bm{q}_{n, i}^{(s)}, \bm{k}_+^{(s)} \rangle) + \sum\limits_{j=2}^M  \exp  ( \langle \bm{q}_{n, i}^{(s)}, \bm{k}_{n, j}^{(s)} \rangle ) } \big) \\
    & -\sum\limits_{k=2}^M \big( V_{n, i}^{(s)} \cdot \frac{ \exp (\langle \bm{q}_{n, i}^{(s)}, \bm{k}_{n, k}^{(s)} \rangle)}{ \exp  (\langle \bm{q}_{n, i}^{(s)}, \bm{k}_+^{(s)} \rangle) + \sum\limits_{j=2}^M  \exp  ( \langle \bm{q}_{n, i}^{(s)}, \bm{k}_{n, j}^{(s)} \rangle ) } \big) \Big) \\
    &\le - \frac{\eta \Vert \bm{\mu} \Vert_2^2}{NM} \ell_n^{\prime(s)} \frac{ \exp (\langle \bm{q}_{n, i}^{(s)}, \bm{k}_+^{(s)} \rangle)}{ \exp  (\langle \bm{q}_{n, i}^{(s)}, \bm{k}_+^{(s)} \rangle) + \sum\limits_{j=2}^M \exp ( \langle \bm{q}_{n, i}^{(s)}, \bm{k}_{n, j}^{(s)} \rangle ) } \\
    &\cdot \Big( V_+^{(s)} \big( 1 - \frac{ \exp (\langle \bm{q}_{n, i}^{(s)}, \bm{k}_+^{(s)} \rangle)}{ \exp  (\langle \bm{q}_{n, i}^{(s)}, \bm{k}_+^{(s)} \rangle) + \sum\limits_{j=2}^M  \exp  ( \langle \bm{q}_{n, i}^{(s)}, \bm{k}_{n, j}^{(s)} \rangle ) } \big) \\
    &+ \vert V_{n, i}^{(s)} \vert \big( \frac{ \sum\limits_{k=2}^M \exp (\langle \bm{q}_{n, i}^{(s)}, \bm{k}_{n, k}^{(s)} \rangle)}{ \exp  (\langle \bm{q}_{n, i}^{(s)}, \bm{k}_+^{(s)} \rangle) + \sum\limits_{j=2}^M  \exp  ( \langle \bm{q}_{n, i}^{(s)}, \bm{k}_{n, j}^{(s)} \rangle ) } \big) \Big) \\
    &\le - \frac{\eta \Vert \bm{\mu} \Vert_2^2}{NM} \ell_n^{\prime(s)} \\
    &\cdot \Big( V_+^{(s)} \cdot \frac{ \sum\limits_{j=2}^M  \exp  ( \langle \bm{q}_{n, i}^{(s)}, \bm{k}_{n, j}^{(s)} \rangle ) }{ \exp  (\langle \bm{q}_{n, i}^{(s)}, \bm{k}_+^{(s)} \rangle) + \sum\limits_{j=2}^M  \exp  ( \langle \bm{q}_{n, i}^{(s)}, \bm{k}_{n, j}^{(s)} \rangle ) } \\
    &+ \vert V_{n, i}^{(s)} \vert \cdot \frac{ \sum\limits_{k=2}^M \exp (\langle \bm{q}_{n, i}^{(s)}, \bm{k}_{n, k}^{(s)} \rangle)}{ \exp  (\langle \bm{q}_{n, i}^{(s)}, \bm{k}_+^{(s)} \rangle) + \sum\limits_{j=2}^M  \exp  ( \langle \bm{q}_{n, i}^{(s)}, \bm{k}_{n, j}^{(s)} \rangle ) } \Big) \\
    &\le - \frac{\eta \Vert \bm{\mu} \Vert_2^2}{NM} \ell_n^{\prime(s)} \cdot \Big( V_+^{(s)} \cdot \frac{ C }{ \exp  (\langle \bm{q}_{n, i}^{(s)}, \bm{k}_+^{(s)} \rangle) } + \vert V_{n, i}^{(s)} \vert \cdot \frac{ C }{ \exp  (\langle \bm{q}_{n, i}^{(s)}, \bm{k}_+^{(s)} \rangle) } \Big) \\
    &\le \frac{\eta C_9 \Vert \bm{\mu} \Vert_2^2}{N \exp  (\langle \bm{q}_{n, i}^{(s)}, \bm{k}_+^{(s)} \rangle)},
\end{align*}
where the second inequality is by \(softmax(\langle \bm{q}_{n, i}^{(s)}, \bm{k}_+^{(s)} \rangle) \le 1\). For the second inequality, we first consider \(\frac{ \sum\limits_{j=2}^M  \exp  ( \langle \bm{q}_{n, i}^{(s)}, \bm{k}_{n, j}^{(s)} \rangle ) }{ \exp  (\langle \bm{q}_{n, i}^{(s)}, \bm{k}_+^{(s)} \rangle) + \sum\limits_{j=2}^M  \exp  ( \langle \bm{q}_{n, i}^{(s)}, \bm{k}_{n, j}^{(s)} \rangle ) } \le \frac{ \sum\limits_{j=2}^M  \exp  ( \langle \bm{q}_{n, i}^{(s)}, \bm{k}_{n, j}^{(s)} \rangle ) }{ \exp  (\langle \bm{q}_{n, i}^{(s)}, \bm{k}_+^{(s)} \rangle) }\), then by the monotonicity of \(\langle \bm{q}_{n, i}^{(s)}, \bm{k}_{n, j}^{(s)} \rangle\) and \(\langle \bm{q}_{n, i}^{(T_1)}, \bm{k}_{n, j}^{(T_1)} \rangle = o(1)\) we have \(\sum\limits_{j=2}^M  \exp  ( \langle \bm{q}_{n, i}^{(s)}, \bm{k}_{n, j}^{(s)} \rangle ) \le C\) for \(s \in [T_1, t]\). The last inequality is by \(V_+^{(s)}, V_{n, i}^{(s)} = o(1)\) for \(s \in [T_1, t]\) and absorbing the constant factors. Similarly, we have
\begin{align*}
    \beta_{n, -, i}^{(s)} \le \frac{\eta C_9 \Vert \bm{\mu} \Vert_2^2}{N \exp  (\langle \bm{q}_{n, i}^{(s)}, \bm{k}_-^{(s)} \rangle)},
\end{align*}

\begin{align*}
    &\alpha_{n, i, +}^{(s)} = \frac{\eta \Vert \bm{\xi}_{n, i} \Vert_2^2}{NM} - \ell_n^{\prime(s)} \\
    &\cdot \Big( V_+^{(s)} \big( \frac{ \exp (\langle \bm{q}_{n, i}^{(s)}, \bm{k}_+^{(s)} \rangle)}{ \exp  (\langle \bm{q}_{n, i}^{(s)}, \bm{k}_+^{(s)} \rangle) + \sum\limits_{j=2}^M  \exp  ( \langle \bm{q}_{n, i}^{(s)}, \bm{k}_{n, j}^{(s)} \rangle ) } \\
    & - (\frac{ \exp (\langle \bm{q}_{n, i}^{(s)}, \bm{k}_+^{(s)} \rangle)}{ \exp  (\langle \bm{q}_{n, i}^{(s)}, \bm{k}_+^{(s)} \rangle) + \sum\limits_{j=2}^M  \exp  ( \langle \bm{q}_{n, i}^{(s)}, \bm{k}_{n, j}^{(s)} \rangle ) })^2 \big) \\
    & -\sum\limits_{k=2}^M \big( V_{n, i}^{(s)} \cdot \frac{ \exp (\langle \bm{q}_{n, i}^{(s)}, \bm{k}_+^{(s)} \rangle)}{ \exp  (\langle \bm{q}_{n, i}^{(s)}, \bm{k}_+^{(s)} \rangle) + \sum\limits_{j=2}^M  \exp  ( \langle \bm{q}_{n, i}^{(s)}, \bm{k}_{n, j}^{(s)} \rangle ) } \\
    & \cdot \frac{ \exp (\langle \bm{q}_{n, i}^{(s)}, \bm{k}_{n, k}^{(s)} \rangle)}{ \exp  (\langle \bm{q}_{n, i}^{(s)}, \bm{k}_+^{(s)} \rangle) + \sum\limits_{j=2}^M  \exp  ( \langle \bm{q}_{n, i}^{(s)}, \bm{k}_{n, j}^{(s)} \rangle ) } \big) \Big) \\
    &+ \{ lower \ order \ term \} \\
    &= \frac{\eta \Vert \bm{\xi}_{n, i} \Vert_2^2}{NM} - \ell_n^{\prime(s)} \frac{ \exp (\langle \bm{q}_{n, i}^{(s)}, \bm{k}_+^{(s)} \rangle)}{ \exp  (\langle \bm{q}_{n, i}^{(s)}, \bm{k}_+^{(s)} \rangle) + \sum\limits_{j=2}^M  \exp  ( \langle \bm{q}_{n, i}^{(s)}, \bm{k}_{n, j}^{(s)} \rangle ) } \\
    &\cdot \Big( V_+^{(s)} \big( 1 - \frac{ \exp (\langle \bm{q}_{n, i}^{(s)}, \bm{k}_+^{(s)} \rangle)}{ \exp  (\langle \bm{q}_{n, i}^{(s)}, \bm{k}_+^{(s)} \rangle) + \sum\limits_{j=2}^M  \exp  ( \langle \bm{q}_{n, i}^{(s)}, \bm{k}_{n, j}^{(s)} \rangle ) } \big) \\
    & -\sum\limits_{k=2}^M \big( V_{n, i}^{(s)} \cdot \frac{ \exp (\langle \bm{q}_{n, i}^{(s)}, \bm{k}_{n, k}^{(s)} \rangle)}{ \exp  (\langle \bm{q}_{n, i}^{(s)}, \bm{k}_+^{(s)} \rangle) + \sum\limits_{j=2}^M  \exp  ( \langle \bm{q}_{n, i}^{(s)}, \bm{k}_{n, j}^{(s)} \rangle ) } \big) \Big) \\
    &+ \{ lower \ order \ term \} \\
    &\le \frac{\eta \Vert \bm{\xi}_{n, i} \Vert_2^2}{NM} \\
    &\cdot \Big( V_+^{(s)} \big( 1 - \frac{ \exp (\langle \bm{q}_{n, i}^{(s)}, \bm{k}_+^{(s)} \rangle)}{ \exp  (\langle \bm{q}_{n, i}^{(s)}, \bm{k}_+^{(s)} \rangle) + \sum\limits_{j=2}^M  \exp  ( \langle \bm{q}_{n, i}^{(s)}, \bm{k}_{n, j}^{(s)} \rangle ) } \big) \\
    & -\sum\limits_{k=2}^M \big( V_{n, i}^{(s)} \cdot \frac{ \exp (\langle \bm{q}_{n, i}^{(s)}, \bm{k}_{n, k}^{(s)} \rangle)}{ \exp  (\langle \bm{q}_{n, i}^{(s)}, \bm{k}_+^{(s)} \rangle) + \sum\limits_{j=2}^M  \exp  ( \langle \bm{q}_{n, i}^{(s)}, \bm{k}_{n, j}^{(s)} \rangle ) } \big) \Big) \\
    &+ \{ lower \ order \ term \} \\
    &\le \frac{\eta \Vert \bm{\xi}_{n, i} \Vert_2^2}{NM} \cdot \Big( V_+^{(s)} \cdot \frac{ \sum\limits_{j=2}^M  \exp  ( \langle \bm{q}_{n, i}^{(s)}, \bm{k}_{n, j}^{(s)} \rangle ) }{ \exp  (\langle \bm{q}_{n, i}^{(s)}, \bm{k}_+^{(s)} \rangle) + \sum\limits_{j=2}^M  \exp  ( \langle \bm{q}_{n, i}^{(s)}, \bm{k}_{n, j}^{(s)} \rangle ) } \\
    &+ \vert V_{n, i}^{(s)} \vert \cdot \frac{ \sum\limits_{k=2}^M \exp (\langle \bm{q}_{n, i}^{(s)}, \bm{k}_{n, k}^{(s)} \rangle)}{ \exp  (\langle \bm{q}_{n, i}^{(s)}, \bm{k}_+^{(s)} \rangle) + \sum\limits_{j=2}^M  \exp  ( \langle \bm{q}_{n, i}^{(s)}, \bm{k}_{n, j}^{(s)} \rangle ) } \Big) \\
    &+ \{ lower \ order \ term \} \\
    &\le \frac{\eta \Vert \bm{\xi}_{n, i} \Vert_2^2}{NM} \cdot \Big( V_+^{(s)} \cdot \frac{ C }{ \exp  (\langle \bm{q}_{n, i}^{(s)}, \bm{k}_+^{(s)} \rangle) } + \vert V_{n, i}^{(s)} \vert \cdot \frac{ C }{ \exp  (\langle \bm{q}_{n, i}^{(s)}, \bm{k}_+^{(s)} \rangle)} \Big) \\
    &+ \{ lower \ order \ term \} \\
    &\le \frac{\eta C_9 \sigma_p^2 d}{N \exp  (\langle \bm{q}_{n, i}^{(s)}, \bm{k}_+^{(s)} \rangle)},
\end{align*}
where most of these processes are similar to the other equations above, and the last inequality we absorb the constant factors and the \(\{ lower \ order \ term \}\). Similarly, we have
\begin{align*}
    &\alpha_{n, i, -}^{(s)} \le \frac{\eta C_9 \sigma_p^2 d}{N \exp  (\langle \bm{q}_{n, i}^{(s)}, \bm{k}_-^{(s)} \rangle)},
\end{align*}

\begin{align*}
    &\beta_{n, i, +}^{(s)} \ge - \frac{\eta C_9 \sigma_p^2 d}{N} \exp (\langle \bm{q}_+^{(s)}, \bm{k}_{n, i}^{(s)} \rangle),
\end{align*}

\begin{align*}
    &\beta_{n, i, -}^{(s)} \ge - \frac{\eta C_9 \sigma_p^2 d}{N} \exp (\langle \bm{q}_-^{(s)}, \bm{k}_{n, i}^{(s)} \rangle),
\end{align*}

\begin{align*}
    &\alpha_{n, i, n, j}^{(s)} \ge - \frac{\eta C_9 \sigma_p^2 d}{N} \exp (\langle \bm{q}_{n, i}^{(s)}, \bm{k}_{n, j}^{(s)} \rangle),
\end{align*}

\begin{align*}
    &\beta_{n, j, n, i}^{(s)} \ge - \frac{\eta C_9 \sigma_p^2 d}{N} \exp (\langle \bm{q}_{n, i}^{(s)}, \bm{k}_{n, j}^{(s)} \rangle),
\end{align*}

Similar to \ref{lower_bound_of_qk}, we apply the bounds of \(\alpha\) and \(\beta\) above to give the upper bounds for the dynamics \(\langle \bm{q}, \bm{k} \rangle\).

\begin{equation}
\begin{split}
    &\langle \bm{q}_+^{(s+1)}, \bm{k}_+^{(s+1)} \rangle - \langle \bm{q}_+^{(s)}, \bm{k}_+^{(s)} \rangle \\
    &= \alpha_{+, +}^{(s)} \Vert \bm{k}_+^{(s)} \Vert_2^2 + \sum\limits_{n \in S_+} \sum\limits_{i=2}^M \alpha_{n, +, i}^{(s)} \langle \bm{k}_+^{(s)}, \bm{k}_{n, i}^{(s)} \rangle \\
    &+ \beta_{+, +}^{(s)} \Vert \bm{q}_+^{(s)} \Vert_2^2 + \sum\limits_{n \in S_+} \sum\limits_{i=2}^M \beta_{n, +, i}^{(s)} \langle \bm{q}_+^{(s)}, \bm{q}_{n, i}^{(s)} \rangle \\
    &+ \Big( \alpha_{+, +}^{(s)} \bm{k}_+^{(s)} + \sum\limits_{n \in S_+} \sum\limits_{i=2}^M \alpha_{n, +, i}^{(s)} \bm{k}_{n, i}^{(s)} \Big) \\
    &\cdot \Big( \beta_{+, +}^{(s)} \bm{q}_+^{(s)\top} + \sum\limits_{n \in S_+} \sum\limits_{i=2}^M \beta_{n, +, i}^{(s)} \bm{q}_{n, i}^{(s)\top} \Big) \\
    &= \alpha_{+, +}^{(s)} \Vert \bm{k}_+^{(s)} \Vert_2^2 + \beta_{+, +}^{(s)} \Vert \bm{q}_+^{(s)} \Vert_2^2 + \{ lower \ order \ term \} \\
    &\le \frac{2 \eta C_9 \Vert \bm{\mu} \Vert_2^2 }{\exp ( \langle \bm{q}_+^{(s)}, \bm{k}_+^{(s)} \rangle)} \cdot \Theta (\Vert \bm{\mu} \Vert_2^2 \sigma_h^2 d_h) + \{ lower \ order \ term \} \\
    &\le \frac{\eta C_{10} \Vert \bm{\mu} \Vert_2^4 \sigma_h^2 d_h}{\exp ( \langle \bm{q}_+^{(s)}, \bm{k}_+^{(s)} \rangle)},
\end{split}
\end{equation}
similarly, we have
\begin{equation}
\begin{split}
    &\langle \bm{q}_-^{(s+1)}, \bm{k}_-^{(s+1)} \rangle - \langle \bm{q}_-^{(s)}, \bm{k}_-^{(s)} \rangle \le \frac{\eta C_{10} \Vert \bm{\mu} \Vert_2^4 \sigma_h^2 d_h}{\exp ( \langle \bm{q}_-^{(s)}, \bm{k}_-^{(s)} \rangle)}.
\end{split}
\end{equation}

\begin{equation}
\begin{split}
    &\langle \bm{q}_+^{(s+1)}, \bm{k}_{n, j}^{(s+1)} \rangle - \langle \bm{q}_+^{(s)}, \bm{k}_{n, j}^{(s)} \rangle \\
    &= \alpha_{+, +}^{(s)} \langle \bm{k}_+^{(s)}, \bm{k}_{n, j}^{(s)} \rangle + \sum\limits_{n^\prime \in S_+} \sum\limits_{l=2}^M \alpha_{n^\prime, +, l}^{(s)} \langle \bm{k}_{n, j}^{(s)}, \bm{k}_{n^\prime, l}^{(s)} \rangle \\
    &+ \beta_{n, j, +}^{(s)} \Vert \bm{q}_+^{(s)} \Vert_2^2 + \beta_{n, j, -}^{(s)} \langle \bm{q}_+^{(s)}, \bm{q}_-^{(s)} \rangle + \sum\limits_{n^\prime = 1}^N \sum\limits_{l=2}^M \beta_{n, j, n^\prime, l}^{(s)} \langle \bm{q}_+^{(s)}, \bm{q}_{n^\prime, l}^{(s)} \rangle \\
    &+ \Big( \alpha_{+, +}^{(s)} \bm{k}_+^{(s)} + \sum\limits_{n^\prime \in S_+} \sum\limits_{l=2}^M \alpha_{n^\prime, +, l}^{(s)} \bm{k}_{n^\prime, l}^{(s)} \Big) \\
    &\cdot \Big( \beta_{n, j, +}^{(s)} \bm{q}_+^{(s)\top} + \beta_{n, j, -}^{(s)} \bm{q}_-^{(s)\top} + \sum\limits_{n^\prime = 1}^N \sum\limits_{l=2}^M \beta_{n, j, n^\prime, l}^{(s)} \bm{q}_{n^\prime, l}^{(s)\top} \Big) \\
    &= \alpha_{n, +, j}^{(s)} \Vert \bm{k}_{n, j}^{(s)} \Vert_2^2 + \beta_{n, j, +}^{(s)} \Vert \bm{q}_+^{(s)} \Vert_2^2 + \{ lower \ order \ term \} \\
    &\ge - \frac{\eta C_9 \Vert \bm{\mu} \Vert_2^2}{N} \cdot \exp (\langle \bm{q}_+^{(s)}, \bm{k}_{n, j}^{(s)} \rangle) \cdot \Theta \Big( \sigma_p^2 \sigma_h^2 d d_h \Big) \\
    &- \frac{\eta C_9 \sigma_p^2 d}{N} \exp (\langle \bm{q}_+^{(s)}, \bm{k}_{n, j}^{(s)} \rangle) \cdot \Theta \Big( \Vert \bm{\mu} \Vert_2^2 \sigma_h^2 d_h \Big) \\
    &+ \{ lower \ order \ term \} \\
    &\ge - \frac{\eta C_{10} \sigma_p^2 d \Vert \bm{\mu} \Vert_2^2 \sigma_h^2 d_h }{N} \cdot \exp (\langle \bm{q}_+^{(s)}, \bm{k}_{n, j}^{(s)} \rangle),
\end{split}
\end{equation}
similarly, we have
\begin{equation}
\begin{split}
    &\langle \bm{q}_-^{(s+1)}, \bm{k}_{n, j}^{(s+1)} \rangle - \langle \bm{q}_-^{(s)}, \bm{k}_{n, j}^{(s)} \rangle \ge - \frac{\eta C_{10} \sigma_p^2 d \Vert \bm{\mu} \Vert_2^2 \sigma_h^2 d_h }{N} \cdot \exp (\langle \bm{q}_-^{(s)}, \bm{k}_{n, j}^{(s)} \rangle).
\end{split}
\end{equation}

\begin{equation}
\begin{split}
    &\langle \bm{q}_{n, i}^{(s+1)}, \bm{k}_+^{(s+1)} \rangle - \langle \bm{q}_{n, i}^{(s)}, \bm{k}_+^{(s)} \rangle \\
    &= \alpha_{n, i, +}^{(s)} \Vert \bm{k}_+^{(s)} \Vert_2^2 + \alpha_{n, i, -}^{(s)} \langle \bm{k}_+^{(s)}, \bm{k}_-^{(s)} \rangle + \sum\limits_{n^\prime =1}^N \sum\limits_{l=2}^M \alpha_{n, i, n^\prime, l}^{(s)} \langle \bm{k}_+^{(s)}, \bm{k}_{n^\prime, l}^{(s)} \rangle \\
    &+ \beta_{+, +}^{(s)} \langle \bm{q}_+^{(s)}, \bm{q}_{n, i}^{(s)} \rangle + \sum\limits_{n^\prime \in S_+} \sum\limits_{l=2}^M \beta_{n^\prime, +, l}^{(s)} \langle \bm{q}_{n, i}^{(s)}, \bm{q}_{n^\prime, l}^{(s)} \rangle \\
    &+ \Big( \alpha_{n, i, +}^{(s)} \bm{k}_+^{(s)} + \alpha_{n, i, -}^{(s)} \bm{k}_-^{(s)} + \sum\limits_{n^\prime =1}^N \sum\limits_{l=2}^M \alpha_{n, i, n^\prime, l}^{(s)} \bm{k}_{n^\prime, l}^{(s)} \Big) \\
    &\cdot \Big( \beta_{+, +}^{(s)} \bm{q}_+^{(s)\top} + \sum\limits_{n^\prime \in S_+} \sum\limits_{l=2}^M \beta_{n^\prime, +, l}^{(s)} \bm{q}_{n^\prime, l}^{(s)\top} \Big) \\
    &= \alpha_{n, i, +}^{(s)} \Vert \bm{k}_+^{(s)} \Vert_2^2 + \beta_{n, +, i}^{(s)} \Vert \bm{q}_{n, i}^{(s)} \Vert_2^2 + \{ lower \ order \ term \} \\
    &\le \frac{\eta C_9 \sigma_p^2 d}{N \exp  (\langle \bm{q}_{n, i}^{(s)}, \bm{k}_+^{(s)} \rangle)} \cdot \Theta \Big( \Vert \bm{\mu} \Vert_2^2 \sigma_h^2 d_h \Big) \\
    &+ \frac{\eta C_9 \Vert \bm{\mu} \Vert_2^2}{N \exp (\langle \bm{q}_{n, i}^{(s)}, \bm{k}_+^{(s)} \rangle)} \cdot \Theta \Big( \sigma_p^2 \sigma_h^2 d d_h \Big) \\
    &+ \{ lower \ order \ term \} \\
    &\le \frac{\eta C_{10} \sigma_p^2 d \Vert \bm{\mu} \Vert_2^2 \sigma_h^2 d_h}{N \exp (\langle \bm{q}_{n, i}^{(s)}, \bm{k}_+^{(s)} \rangle)},
\end{split}
\end{equation}
similarly, we have
\begin{equation}
\begin{split}
    &\langle \bm{q}_{n, i}^{(s+1)}, \bm{k}_-^{(s+1)} \rangle - \langle \bm{q}_{n, i}^{(s)}, \bm{k}_-^{(s)} \rangle \le \frac{\eta C_{10} \sigma_p^2 d \Vert \bm{\mu} \Vert_2^2 \sigma_h^2 d_h}{N \exp (\langle \bm{q}_{n, i}^{(s)}, \bm{k}_-^{(s)} \rangle)}.
\end{split}
\end{equation}

\begin{equation}
\begin{split}
    &\langle \bm{q}_{n, i}^{(s+1)}, \bm{k}_{n, j}^{(s+1)} \rangle - \langle \bm{q}_{n, i}^{(s)}, \bm{k}_{n, j}^{(s)} \rangle \\
    &= \alpha_{n, i, +}^{(s)} \langle \bm{k}_+^{(s)}, \bm{k}_{n, j}^{(s)} \rangle + \alpha_{n, i, -}^{(s)} \langle \bm{k}_-^{(s)}, \bm{k}_{n, j}^{(s)} \rangle + \sum\limits_{n^\prime =1}^N \sum\limits_{l=2}^M \alpha_{n, i, n^\prime, l}^{(s)} \langle \bm{k}_{n^\prime, l}^{(s)}, \bm{k}_{n, j}^{(s)} \rangle \\
    &+ \beta_{n, j, +}^{(s)} \langle \bm{q}_+^{(s)}, \bm{q}_{n, i}^{(s)} \rangle + \beta_{n, j, -}^{(s)} \langle \bm{q}_-^{(s)}, \bm{q}_{n, i}^{(s)} \rangle + \sum\limits_{n^\prime = 1}^N \sum\limits_{l=2}^M \beta_{n, j, n^\prime, l}^{(s)} \langle \bm{q}_{n^\prime, l}^{(s)}, \bm{q}_{n, i}^{(s)} \rangle \\
    &+ \Big( \alpha_{n, i, +}^{(s)} \bm{k}_+^{(s)} + \alpha_{n, i, -}^{(s)} \bm{k}_-^{(s)} + \sum\limits_{n^\prime =1}^N \sum\limits_{l=2}^M \alpha_{n, i, n^\prime, l}^{(s)} \bm{k}_{n^\prime, l}^{(s)} \Big) \\
    &\cdot \Big( \beta_{n, j, +}^{(s)} \bm{q}_+^{(s)\top} + \beta_{n, j, -}^{(s)} \bm{q}_-^{(s)\top} + \sum\limits_{n^\prime = 1}^N \sum\limits_{l=2}^M \beta_{n, j, n^\prime, l}^{(s)} \bm{q}_{n^\prime, l}^{(s)\top} \Big) \\
    &= \alpha_{n, i, n, j}^{(s)} \Vert \bm{k}_{n, j}^{(s)} \Vert_2^2 + \beta_{n, j, n, i}^{(s)} \Vert \bm{q}_{n, i}^{(s)} \Vert_2^2 \\
    &+ \{ lower \ order \ term \} \\
    &\ge - \frac{2\eta C_9 \sigma_p^2 d}{N} \exp (\langle \bm{q}_{n, i}^{(s)}, \bm{k}_{n, j}^{(s)} \rangle) \cdot \Theta \Big( \sigma_p^2 \sigma_h^2 d d_h \Big) \\
    &+ \{ lower \ order \ term \} \\
    &\ge - \frac{\eta C_{10} \sigma_p^4 d^2 \sigma_h^2 d_h }{N} \cdot \exp (\langle \bm{q}_{n, i}^{(s)}, \bm{k}_{n, j}^{(s)} \rangle).
\end{split}
\end{equation}

\subsection{Bounds for the Sum of \(\alpha\) and \(\beta\)}
\label{sum_alpha_beta}
The gradients of the inner products of \(\bm{q}\) and \(\bm{k}\) contain a lot of coefficients \(\alpha\) and \(\beta\), and in order to conveniently give the upper bounds of some lower order inner products, we will give upper bounds for the summation of \(\alpha\) and \(\beta\) (e.g. \(\sum\limits_{s = T_1}^{t} \vert \alpha_{+, +}^{(s)} \vert\)).

Note that in the Jacobi matrix of the Softmax function, the elements on the diagonal are \(softmax(a_i) \cdot \Big( 1 - softmax(a_i) \Big)\) and the elements on the off-diagonal are \(softmax(a_i) \cdot softmax(a_j)\).
In Stage \uppercase\expandafter{\romannumeral2}, the attentions on signals \(\bm{\mu}_\pm\) increase and the attentions on noises \(\bm{\xi}\) decrease, then we can consider the following cases
\begin{itemize}
    \item if \(a_i = \langle \bm{q}_+, \bm{k}_+ \rangle\) or \(a_i = \langle \bm{q}_i, \bm{k}_+ \rangle\), \(softmax(a_i)\) has a constant upper bound 1, \(\Big( 1 - softmax(a_i) \Big)\) decreases as \(softmax(a_i)\) increases. So the upper bound of \(softmax(a_i) \cdot \Big( 1 - softmax(a_i) \Big)\) decreases as \(softmax(a_i)\) increases.
    \item if \(a_i = \langle \bm{q}_+, \bm{k}_j \rangle\) or \(a_i = \langle \bm{q}_i, \bm{k}_j \rangle\), \(\Big( 1 - softmax(a_i) \Big)\) has a constant upper bound 1. So the upper bound of \(softmax(a_i) \cdot \Big( 1 - softmax(a_i) \Big)\) decreases as \(softmax(a_i)\) decreases.
    \item if \(a_j = \langle \bm{q}_+, \bm{k}_j \rangle\) or \(a_j = \langle \bm{q}_i, \bm{k}_j \rangle\), \(softmax(a_i)\) has a constant upper bound 1. So the upper bound of \(softmax(a_i) \cdot softmax(a_j)\) decreases as \(softmax(a_j)\) decreases.
\end{itemize}
Based on the above cases, we first study the bounds of the following terms
\begin{itemize}
    \item \(1 - softmax(\langle \bm{q}_+^{(s)}, \bm{k}_+^{(s)} \rangle)\)
    \item \(1 - softmax(\langle \bm{q}_{n, i}^{(s)}, \bm{k}_+^{(s)} \rangle)\)
    \item \(softmax(\langle \bm{q}_+^{(s)}, \bm{k}_{n, j}^{(s)} \rangle)\)
    \item \(softmax(\langle \bm{q}_{n, i}^{(s)}, \bm{k}_{n, j}^{(s)} \rangle)\)
\end{itemize}
Note that \(1 - softmax(\langle \bm{q}_+^{(s)}, \bm{k}_+^{(s)} \rangle)\) = \(\sum\limits_j softmax(\langle \bm{q}_+^{(s)}, \bm{k}_{n, j}^{(s)} \rangle)\) and \(1 - softmax(\langle \bm{q}_{n, i}^{(s)}, \bm{k}_+^{(s)} \rangle)\) = \(\sum\limits_j softmax(\langle \bm{q}_{n, i}^{(s)}, \bm{k}_{n, j}^{(s)} \rangle)\), 
we only need to give the upper bounds for \(softmax(\langle \bm{q}_+^{(s)}, \bm{k}_{n, j}^{(s)} \rangle)\) and \(softmax(\langle \bm{q}_{n, i}^{(s)}, \bm{k}_{n, j}^{(s)} \rangle)\).

Assume that the propositions \(\mathcal{B}(T_1), \dots, \mathcal{B}(s)\), \(\mathcal{D}(T_1), \dots, \mathcal{D}(s - 1)\) hold (\(s \in [T_1, t]\)), we have
\begin{equation}
    \label{V_upper_bound}
    \vert V_\pm^{(s)} \vert, \vert V_{n, i}^{(s)} \vert \le O(d_h^{-\frac{1}{4}}) + \eta C_4 \Vert \bm{\mu} \Vert_2^2 \Vert \bm{w}_O \Vert_2^2 (s - T_1),
\end{equation}
\begin{equation}
    \label{lambda_1}
    \Lambda_{n, \pm, j}^{(s)} \ge \log \Big( \exp(\Lambda_{n, \pm, j}^{(T_1)}) + \frac{\eta^2 C_8 \Vert \bm{\mu} \Vert_2^4 \Vert \bm{w}_O \Vert_2^2 d_h^{\frac{1}{2}}}{N \big( \log (6N^2M^2 / \delta) \big)^{2} } \cdot (s - T_1)(s - T_1 - 1) \Big),
\end{equation}
\begin{equation}
    \label{lambda_2}
    \Lambda_{n, i, \pm, j}^{(s)}  \ge \log \Big( \exp(\Lambda_{n, i, \pm, j}^{(T_1)}) + \frac{\eta^2 C_8 \sigma_p^2 d \Vert \bm{\mu} \Vert_2^2 \Vert \bm{w}_O \Vert_2^2 d_h^{\frac{1}{2}} }{N \big( \log (6N^2M^2 / \delta) \big)^{2}} \cdot (s - T_1)(s - T_1 - 1) \Big),
\end{equation}
for \(i, j \in [M] \backslash \{1\}, n \in [N] , s \in [T_1, t]\).

Then we have
\begin{equation}
\begin{split}
    \label{q_p_k_j_upper_bound}
    &\frac{ \exp (\langle \bm{q}_\pm^{(s)}, \bm{k}_{n, j}^{(s)} \rangle)}{ \exp  (\langle \bm{q}_\pm^{(s)}, \bm{k}_\pm^{(s)} \rangle) + \sum\limits_{j^\prime=2}^M  \exp  ( \langle \bm{q}_\pm^{(s)}, \bm{k}_{n, j^\prime}^{(s)} \rangle ) } \\
    &\le  \frac{ \exp (\langle \bm{q}_\pm^{(s)}, \bm{k}_{n, j}^{(s)} \rangle)}{ C \exp (\langle \bm{q}_\pm^{(s)}, \bm{k}_\pm^{(s)} \rangle) } \\
    &= \frac{1}{ C \exp( \Lambda_{n, \pm, j}^{(s)} ) } \\
    &\le \frac{1}{ C\exp(\Lambda_{n, \pm, j}^{(T_1)}) + \frac{\eta^2 C_8 C \Vert \bm{\mu} \Vert_2^4 \Vert \bm{w}_O \Vert_2^2 d_h^{\frac{1}{2}}}{N \big( \log (6N^2M^2 / \delta) \big)^{2} } \cdot (s - T_1)(s - T_1 - 1) } \\
    &\le \frac{1}{ C_{13} + \frac{\eta^2 C_{13} \Vert \bm{\mu} \Vert_2^4 \Vert \bm{w}_O \Vert_2^2 d_h^{\frac{1}{2}}}{N \big( \log (6N^2M^2 / \delta) \big)^{2} } \cdot (s - T_1)(s - T_1 - 1) }.
\end{split}
\end{equation}
For the first inequality, by \( \langle \bm{q}^{(T_1)}, \bm{k}^{(T_1)} \rangle = o(1) \) and the monotonicity of \(\langle \bm{q}^{(s)}, \bm{k}^{(s)} \rangle\) (\(\langle \bm{q}_\pm^{(s)}, \bm{k}_\pm^{(s)} \rangle\) is increasing and \(\langle \bm{q}_\pm^{(s)}, \bm{k}_{n, j}^{(s)} \rangle\) is decreasing), there exist a constant C such that \( C \exp  (\langle \bm{q}_\pm^{(s)}, \bm{k}_\pm^{(s)} \rangle) \ge \exp(\langle \bm{q}_\pm^{(s)}, \bm{k}_\pm^{(s)} \rangle) + \sum\limits_{j^\prime=2}^M  \exp  ( \langle \bm{q}_\pm^{(s)}, \bm{k}_{n, j^\prime}^{(s)} \rangle ) \).
The second inequality is by plugging \eqref{lambda_1}. For the last inequality, by \( \Lambda_{n, \pm, j}^{(T_1)} = o(1) \), there exist a constant \(C_{13}\) such that \(C_{13} \le C\exp(\Lambda_{n, \pm, j}^{(T_1)}) \) and \(C_{13} \le C_8 C\).
Similarly, we have
\begin{equation}
\begin{split}
    \label{q_i_k_j_upper_bound}
    &\frac{ \exp (\langle \bm{q}_{n, i}^{(s)}, \bm{k}_{n, j}^{(s)} \rangle)}{ \exp  (\langle \bm{q}_{n, i}^{(s)}, \bm{k}_+^{(s)} \rangle) + \sum\limits_{j^\prime=2}^M  \exp  ( \langle \bm{q}_{n, i}^{(s)}, \bm{k}_{n, j^\prime}^{(s)} \rangle ) } \\
    &\le \frac{1}{ C \exp( \Lambda_{n, i, +, j}^{(s)} ) } \\
    &\le \frac{1}{ C_{13} + \frac{\eta^2 C_{13} \sigma_p^2 d \Vert \bm{\mu} \Vert_2^2 \Vert \bm{w}_O \Vert_2^2 d_h^{\frac{1}{2}} }{N \big( \log (6N^2M^2 / \delta) \big)^{2}} \cdot (s - T_1)(s - T_1 - 1) }.
\end{split}
\end{equation}
Plugging \eqref{V_upper_bound},\eqref{q_p_k_j_upper_bound} and \eqref{q_i_k_j_upper_bound} into the expressions of \(\alpha\), \(\beta\) we have

\begin{equation}
\begin{split}
    \label{alpha_p_p_upper_bound}
    &\vert \alpha_{+, +}^{(s)} \vert = \Big\vert \frac{\eta}{NM} \sum\limits_{n \in S_+} - \ell_n^{\prime(s)} \Vert \bm{\mu} \Vert_2^2 \\
    &\cdot \Big( V_+^{(s)} \big( \frac{ \exp (\langle \bm{q}_+^{(s)}, \bm{k}_+^{(s)} \rangle)}{ \exp  (\langle \bm{q}_+^{(s)}, \bm{k}_+^{(s)} \rangle) + \sum\limits_{j=2}^M  \exp  ( \langle \bm{q}_+^{(s)}, \bm{k}_{n, j}^{(s)} \rangle ) } \\
    & - (\frac{ \exp (\langle \bm{q}_+^{(s)}, \bm{k}_+^{(s)} \rangle)}{ \exp  (\langle \bm{q}_+^{(s)}, \bm{k}_+^{(s)} \rangle) + \sum\limits_{j=2}^M  \exp  ( \langle \bm{q}_+^{(s)}, \bm{k}_{n, j}^{(s)} \rangle ) })^2 \big) \\
    & - \sum\limits_{i=2}^M \big( V_{n, i}^{(s)} \cdot \frac{ \exp (\langle \bm{q}_+^{(s)}, \bm{k}_+^{(s)} \rangle)}{ \exp  (\langle \bm{q}_+^{(s)}, \bm{k}_+^{(s)} \rangle) + \sum\limits_{j=2}^M  \exp  ( \langle \bm{q}_+^{(s)}, \bm{k}_{n, j}^{(s)} \rangle ) } \\
    & \cdot \frac{ \exp (\langle \bm{q}_+^{(s)}, \bm{k}_{n, i}^{(s)} \rangle)}{ \exp  (\langle \bm{q}_+^{(s)}, \bm{k}_+^{(s)} \rangle) + \sum\limits_{j=2}^M  \exp  ( \langle \bm{q}_+^{(s)}, \bm{k}_{n, j}^{(s)} \rangle ) } \big) \Big) \Big\vert \\
    &\le \frac{\eta \Vert \bm{\mu} \Vert_2^2}{NM} \cdot \frac{3N}{4} \cdot \Big( O(d_h^{-\frac{1}{4}}) + \eta C_4 \Vert \bm{\mu} \Vert_2^2 \Vert \bm{w}_O \Vert_2^2 (s - T_1) \Big) \\
    &\cdot O\Big( \frac{1}{ C_{13} + \frac{\eta^2 C_{13} \Vert \bm{\mu} \Vert_2^4 \Vert \bm{w}_O \Vert_2^2 d_h^{\frac{1}{2}}}{N \big( \log (6N^2M^2 / \delta) \big)^{2} } \cdot (s - T_1)(s - T_1 - 1) } \Big) \\
    &= O \Big( \frac{\eta \Vert \bm{\mu} \Vert_2^2 d_h^{-\frac{1}{4}}}{ C_{13} + \frac{\eta^2 C_{13} \Vert \bm{\mu} \Vert_2^4 \Vert \bm{w}_O \Vert_2^2 d_h^{\frac{1}{2}}}{N \big( \log (6N^2M^2 / \delta) \big)^{2} } \cdot (s - T_1)(s - T_1 - 1) } \Big) \\
    &+ O \Big( \frac{\eta^2 \Vert \bm{\mu} \Vert_2^4 \Vert \bm{w}_O \Vert_2^2 (s - T_1)}{ C_{13} + \frac{\eta^2 C_{13} \Vert \bm{\mu} \Vert_2^4 \Vert \bm{w}_O \Vert_2^2 d_h^{\frac{1}{2}}}{N \big( \log (6N^2M^2 / \delta) \big)^{2} } \cdot (s - T_1)(s - T_1 - 1) } \Big) \\
    &= O \Big( \eta \Vert \bm{\mu} \Vert_2^2 d_h^{-\frac{1}{4}} \Big) + O \Big( \frac{\eta^2 \Vert \bm{\mu} \Vert_2^4 \Vert \bm{w}_O \Vert_2^2 (s - T_1)}{ C_{13} + \frac{\eta^2 C_{13} \Vert \bm{\mu} \Vert_2^4 \Vert \bm{w}_O \Vert_2^2 d_h^{\frac{1}{2}}}{N \big( \log (6N^2M^2 / \delta) \big)^{2} } \cdot (s - T_1)(s - T_1 - 1) } \Big).
\end{split}
\end{equation}
where the third equality is by\(\frac{\eta^2 C_{13} \Vert \bm{\mu} \Vert_2^4 \Vert \bm{w}_O \Vert_2^2 d_h^{\frac{1}{2}}}{N \big( \log (6N^2M^2 / \delta) \big)^{2} } \cdot (s - T_1)(s - T_1 - 1) \ge 0\) for \(s \in [T_1, t]\). 
Next, we give an upper bound for \(\frac{\eta^2 \Vert \bm{\mu} \Vert_2^4 \Vert \bm{w}_O \Vert_2^2 (s - T_1)}{ C_{13} + \frac{\eta^2 C_{13} \Vert \bm{\mu} \Vert_2^4 \Vert \bm{w}_O \Vert_2^2 d_h^{\frac{1}{2}}}{N \big( \log (6N^2M^2 / \delta) \big)^{2} } \cdot (s - T_1)(s - T_1 - 1) }\) as follows:
\begin{equation}
\begin{split}
    \label{V_attn_upper_bound}
    &\frac{\eta^2 \Vert \bm{\mu} \Vert_2^4 \Vert \bm{w}_O \Vert_2^2 (s - T_1)}{ C_{13} + \frac{\eta^2 C_{13} \Vert \bm{\mu} \Vert_2^4 \Vert \bm{w}_O \Vert_2^2 d_h^{\frac{1}{2}}}{N \big( \log (6N^2M^2 / \delta) \big)^{2} } \cdot (s - T_1)(s - T_1 - 1) } \\
    &= \frac{\eta^2 \Vert \bm{\mu} \Vert_2^4 \Vert \bm{w}_O \Vert_2^2 }{ \frac{C_{13}}{(s - T_1)} + \frac{\eta^2 C_{13} \Vert \bm{\mu} \Vert_2^4 \Vert \bm{w}_O \Vert_2^2 d_h^{\frac{1}{2}}}{N \big( \log (6N^2M^2 / \delta) \big)^{2} } \cdot (s - T_1) - \frac{\eta^2 C_{13} \Vert \bm{\mu} \Vert_2^4 \Vert \bm{w}_O \Vert_2^2 d_h^{\frac{1}{2}}}{N \big( \log (6N^2M^2 / \delta) \big)^{2} } } \\
    &\le \frac{\eta^2 \Vert \bm{\mu} \Vert_2^4 \Vert \bm{w}_O \Vert_2^2}{2\sqrt{\frac{\eta^2 C_{13}^2 \Vert \bm{\mu} \Vert_2^4 \Vert \bm{w}_O \Vert_2^2 d_h^{\frac{1}{2}}}{N \big( \log (6N^2M^2 / \delta) \big)^{2} }} - \frac{\eta^2 C_{13} \Vert \bm{\mu} \Vert_2^4 \Vert \bm{w}_O \Vert_2^2 d_h^{\frac{1}{2}}}{N \big( \log (6N^2M^2 / \delta) \big)^{2} } } \\
    &= \frac{\eta^2 \Vert \bm{\mu} \Vert_2^4 \Vert \bm{w}_O \Vert_2^2}{\frac{2\eta C_{13} \Vert \bm{\mu} \Vert_2^2 \Vert \bm{w}_O \Vert_2 d_h^{\frac{1}{4}}}{N^{\frac{1}{2}} \big( \log (6N^2M^2 / \delta) \big) } - \frac{\eta^2 C_{13} \Vert \bm{\mu} \Vert_2^4 \Vert \bm{w}_O \Vert_2^2 d_h^{\frac{1}{2}}}{N \big( \log (6N^2M^2 / \delta) \big)^{2} } } \\
    &= \frac{\eta^2 \Vert \bm{\mu} \Vert_2^4 \Vert \bm{w}_O \Vert_2^2}{ \Theta \Big( \frac{\eta \Vert \bm{\mu} \Vert_2^2 \Vert \bm{w}_O \Vert_2 d_h^{\frac{1}{4}}}{N^{\frac{1}{2}} \log (6N^2M^2 / \delta) } \Big) } \\
    &= O \Big( \eta \Vert \bm{\mu} \Vert_2^2 N^{\frac{1}{2}} d_h^{-\frac{1}{4}} \log (6N^2M^2 / \delta) \Big),
\end{split}
\end{equation}
where the inequality is by \(ax + \frac{b}{x} \ge 2\sqrt{ab}\) for \(x > 0\), the third equality is by absorbing the lower order term \(\frac{\eta^2 C_{13} \Vert \bm{\mu} \Vert_2^4 \Vert \bm{w}_O \Vert_2^2 d_h^{\frac{1}{2}}}{N \big( \log (6N^2M^2 / \delta) \big)^{2} }\), the last equality is by \(\Vert \bm{w}_O \Vert_2 = \Theta(1)\). Plugging this into \eqref{alpha_p_p_upper_bound} and get
\begin{equation}
\begin{split}
    \vert \alpha_{+, +}^{(s)} \vert &= O \Big( \eta \Vert \bm{\mu} \Vert_2^2 d_h^{-\frac{1}{4}} \Big) + O \Big( \eta \Vert \bm{\mu} \Vert_2^2 N^{\frac{1}{2}} d_h^{-\frac{1}{4}} \log (6N^2M^2 / \delta) \Big) \\
    &= O \Big( \eta \Vert \bm{\mu} \Vert_2^2 N^{\frac{1}{2}} d_h^{-\frac{1}{4}} \log (6N^2M^2 / \delta) \Big).
\end{split}
\end{equation}
Similarly, we have
\begin{equation}
\begin{split}
    \vert \alpha_{-, -}^{(s)} \vert, \vert \beta_{+, +}^{(s)} \vert, \vert \beta_{-, -}^{(s)} \vert = O \Big( \eta \Vert \bm{\mu} \Vert_2^2 N^{\frac{1}{2}} d_h^{-\frac{1}{4}} \log (6N^2M^2 / \delta) \Big),
\end{split}
\end{equation}
\begin{equation}
\begin{split}
    \vert \alpha_{n, +, i}^{(s)} \vert, \vert \alpha_{n, -, i}^{(s)} \vert = O \Big( \eta \Vert \bm{\mu} \Vert_2^2 N^{-\frac{1}{2}} d_h^{-\frac{1}{4}} \log (6N^2M^2 / \delta) \Big),
\end{split}
\end{equation}
\begin{equation}
\begin{split}
    \vert \beta_{n, +, i}^{(s)} \vert, \vert \beta_{n, -, i}^{(s)} \vert = O \Big( \frac{ \eta \Vert \bm{\mu} \Vert_2^3  \log (6N^2M^2 / \delta) }{ \sigma_p d^{\frac{1}{2}} N^{\frac{1}{2}} d_h^{\frac{1}{4}} } \Big) = O \Big( \eta \Vert \bm{\mu} \Vert_2^2 \cdot \mathrm{SNR} \cdot N^{-\frac{1}{2}} d_h^{-\frac{1}{4}} \log (6N^2M^2 / \delta) \Big),
\end{split}
\end{equation}
for \(i \in [M] \backslash \{1\}, n \in S_\pm\).
\begin{equation}
\begin{split}
    \vert \alpha_{n, i, +}^{(s)} \vert, \vert \alpha_{n, i, -}^{(s)} \vert = O \Big( \frac{\eta \Vert \bm{\mu} \Vert_2 \sigma_p d^{\frac{1}{2}} \log (6N^2M^2 / \delta) }{ N^{\frac{1}{2}} d_h^{\frac{1}{4}} } \Big) = O \Big( \eta \Vert \bm{\mu} \Vert_2^2 d_h^{-\frac{1}{4}} \log (6N^2M^2 / \delta) \Big),
\end{split}
\end{equation}
for \(i \in [M] \backslash \{1\}, n \in S_\pm\), the last equality is by \(N \cdot \mathrm{SNR}^2 \ge \Omega(1)\).
\begin{equation}
\begin{split}
    \vert \beta_{n, i, +}^{(s)} \vert, \vert \beta_{n, i, -}^{(s)} \vert = O \Big( \frac{\eta \sigma_p^2 d \log (6N^2M^2 / \delta) }{ N^{\frac{1}{2}} d_h^{\frac{1}{4}} } \Big) = O \Big( \eta \Vert \bm{\mu} \Vert_2^2 N^{\frac{1}{2}} d_h^{-\frac{1}{4}} \log (6N^2M^2 / \delta) \Big),
\end{split}
\end{equation}
for \(i \in [M] \backslash \{1\}, n \in S_\pm\), the last equality is by \(N \cdot \mathrm{SNR}^2 \ge \Omega(1)\).
\begin{equation}
\begin{split}
    \vert \alpha_{n, i, n, j}^{(s)} \vert, \vert \beta_{n, j, n, i}^{(s)} \vert = O \Big( \frac{\eta \Vert \bm{\mu} \Vert_2 \sigma_p d^{\frac{1}{2}} \log (6N^2M^2 / \delta) }{ N^{\frac{1}{2}} d_h^{\frac{1}{4}} } \Big) = O \Big( \eta \Vert \bm{\mu} \Vert_2^2 d_h^{-\frac{1}{4}} \log (6N^2M^2 / \delta) \Big),
\end{split}
\end{equation}
for \(i, j \in [M] \backslash \{1\}, n \in [N]\), the last equality is by \(N \cdot \mathrm{SNR}^2 \ge \Omega(1)\).
\begin{equation}
\begin{split}
    \vert \alpha_{n, i, n^\prime, j}^{(s)} \vert, \vert \beta_{n, j, n^\prime, i}^{(s)} \vert = O \Big( \frac{\eta \Vert \bm{\mu} \Vert_2 \sigma_p \log (6N^2M^2 / \delta) \log (4N^2M^2/\delta) }{ N^{\frac{1}{2}} d_h^{\frac{1}{4}} } \Big) = O \Big( \eta \Vert \bm{\mu} \Vert_2^2 d^{-\frac{1}{2}} d_h^{-\frac{1}{4}} \big( \log (6N^2M^2 / \delta) \big)^2 \Big),
\end{split}
\end{equation}
for \(i, j \in [M] \backslash \{1\}, n, n^\prime \in [N], n \ne n^\prime\), the last equality is by \(N \cdot \mathrm{SNR}^2 \ge \Omega(1)\). Taking a summation we obtain that

\begin{equation}
\begin{split}
    \sum\limits_{s = T_1}^{t}\vert \alpha_{+, +}^{(s)} \vert &= O \Big( \frac{1}{\eta \Vert \bm{\mu} \Vert_2^2 \Vert \bm{w}_O \Vert_2^2 \log (6N^2M^2 / \delta) } \Big) \cdot O \Big( \eta \Vert \bm{\mu} \Vert_2^2 N^{\frac{1}{2}} d_h^{-\frac{1}{4}} \log (6N^2M^2 / \delta) \Big) \\
    &= O \Big( N^{\frac{1}{2}} d_h^{-\frac{1}{4}} \Big),
\end{split}
\end{equation}
where the last equality is by \(\Vert \bm{w}_O \Vert = \Theta (1)\). Similarly, we have
\begin{equation}
\begin{split}
    \sum\limits_{s = T_1}^{t}\vert \alpha_{-, -}^{(s)} \vert, \sum\limits_{s = T_1}^{t}\vert \beta_{+, +}^{(s)} \vert, \sum\limits_{s = T_1}^{t}\vert \beta_{-, -}^{(s)} \vert, \sum\limits_{s = T_1}^{t}\vert \beta_{n, i, +}^{(s)} \vert, \sum\limits_{s = T_1}^{t}\vert \beta_{n, i, -}^{(s)} \vert = O \Big( N^{\frac{1}{2}} d_h^{-\frac{1}{4}} \Big),
\end{split}
\end{equation}
for \(i \in [M] \backslash \{1\}, n \in S_\pm\).
\begin{equation}
\begin{split}
    \sum\limits_{s = T_1}^{t}\vert \alpha_{n, +, i}^{(s)} \vert, \sum\limits_{s = T_1}^{t}\vert \alpha_{n, -, i}^{(s)} \vert = O \Big( N^{-\frac{1}{2}} d_h^{-\frac{1}{4}} \Big),
\end{split}
\end{equation}
for \(i \in [M] \backslash \{1\}, n \in S_\pm\).
\begin{equation}
\begin{split}
    &\sum\limits_{s = T_1}^{t}\vert \beta_{n, +, i}^{(s)} \vert, \sum\limits_{s = T_1}^{t}\vert \beta_{n, -, i}^{(s)} \vert = O \Big(\mathrm{SNR} \cdot N^{-\frac{1}{2}} d_h^{-\frac{1}{4}} \Big)
\end{split}
\end{equation}
for \(i \in [M] \backslash \{1\}, n \in S_\pm\).
\begin{equation}
\begin{split}
    \sum\limits_{s = T_1}^{t}\vert \alpha_{n, i, +}^{(s)} \vert, \sum\limits_{s = T_1}^{t}\vert \alpha_{n, i, -}^{(s)} \vert, \sum\limits_{s = T_1}^{t}\vert \alpha_{n, i, n, j}^{(s)} \vert, \sum\limits_{s = T_1}^{t}\vert \beta_{n, j, n, i}^{(s)} \vert
    = O \Big( d_h^{-\frac{1}{4}} \Big)
\end{split}
\end{equation}
for \(i, j \in [M] \backslash \{1\}, n \in S_\pm\).
\begin{equation}
\begin{split}
    \sum\limits_{s = T_1}^{t}\vert \alpha_{n, i, n^\prime, j}^{(s)} \vert, \sum\limits_{s = T_1}^{t}\vert \beta_{n, j, n^\prime, i}^{(s)} \vert = O \Big( d^{-\frac{1}{2}} d_h^{-\frac{1}{4}} \log (6N^2M^2 / \delta) \Big)
\end{split}
\end{equation}
for \(i, j \in [M] \backslash \{1\}, n, n^\prime \in [N], n \ne n^\prime\).

With these sums of \(\alpha\) and \(\beta\) above, we can easily prove \textbf{Claim} \ref{claim3} and \textbf{Claim} \ref{claim4}. 

\subsection{Proof of Claim \ref{claim3}}
\label{proof_claim3}
In this subsection, we assume that \( \mathcal{E}(T_1), \dots, \mathcal{E}(t) \) hold, and then proof that \( \mathcal{C}(t+1) \) is true with the result of \ref{sum_alpha_beta}.

\begin{align*}
    &\Big\vert \Vert \bm{q}_+^{(t+1)} \Vert_2^2 - \Vert \bm{q}_+^{(T_1)} \Vert_2^2 \Big\vert \le \sum\limits_{s = T_1}^{t} \Big\vert \Vert \bm{q}_+^{(s+1)} \Vert_2^2 - \Vert \bm{q}_+^{(s)} \Vert_2^2 \Big\vert \\
    &\le \sum\limits_{s = T_1}^{t} \Big\vert 2 \alpha_{+, +}^{(s)} \langle \bm{q}_+^{(s)}, \bm{k}_+^{(s)} \rangle + 2 \sum\limits_{n \in S_+} \sum\limits_{i=2}^M \alpha_{n, +, i}^{(s)} \langle \bm{q}_+^{(s)}, \bm{k}_{n, i}^{(s)} \rangle \\
    &+ \Big( \alpha_{+, +}^{(s)} \bm{k}_+^{(s)} + \sum\limits_{n \in S_+} \sum\limits_{i=2}^M \alpha_{n, +, i}^{(s)} \bm{k}_{n, i}^{(s)} \Big) \\
    &\cdot \Big( \alpha_{+, +}^{(s)} \bm{k}_+^{(s)\top} + \sum\limits_{n \in S_+} \sum\limits_{i=2}^M \alpha_{n, +, i}^{(s)} \bm{k}_{n, i}^{(s)\top} \Big) \Big\vert \\
    &\le 2 \sum\limits_{s = T_1}^{t} \vert \alpha_{+, +}^{(s)} \vert \vert \langle \bm{q}_+^{(s)}, \bm{k}_+^{(s)} \rangle \vert + 2 \sum\limits_{n \in S_+} \sum\limits_{i=2}^M \sum\limits_{s = T_1}^{t} \vert \alpha_{n, +, i}^{(s)} \vert \vert \langle \bm{q}_+^{(s)}, \bm{k}_{n, i}^{(s)} \rangle \vert \\
    &+ \{lower \ order \ term\} \\
    &= O \Big( N^{\frac{1}{2}} d_h^{-\frac{1}{4}} \Big) \cdot O(\log (d_h^{\frac{1}{2}})) + N \cdot M \cdot O \Big( N^{-\frac{1}{2}} d_h^{-\frac{1}{4}} \Big) \cdot O(\log (d_h^{\frac{1}{2}})) \\
    &= O \Big( N^{\frac{1}{2}} d_h^{-\frac{1}{4}} \log (d_h^{\frac{1}{2}}) \Big) \\
\end{align*}
where the first inequality is by triangle inequality, the second inequality is by \ref{q_p_sq_dy}.
Since \(\sigma_h^2 \ge \big( \max\{\sigma_p^2 d, \Vert \mu \Vert_2^2\} \big)^{-1} \cdot d_h^{-\frac{1}{2}} (\log (6N^2M^2 / \delta))^{-2}\) and \(d_h = \widetilde{\Omega} \Big( \max \{\mathrm{SNR}^4, \mathrm{SNR}^{-4}\} N^2 \epsilon^{-2} \Big) \), we have \(N^{\frac{1}{2}} d_h^{-\frac{1}{4}} \log (d_h^{\frac{1}{2}}) = o (\Vert \bm{\mu} \Vert_2^2 \sigma_h^2 d_h)\), so \(\Vert \bm{q}_+^{(t+1)} \Vert_2^2 = \Vert \bm{q}_+^{(T_1)} \Vert_2^2 + o (\Vert \bm{\mu} \Vert_2^2 \sigma_h^2 d_h) = \Theta (\Vert \bm{\mu} \Vert_2^2 \sigma_h^2 d_h)\).
Similarly, we have
\[
    \Big\vert \Vert \bm{q}_-^{(t+1)} \Vert_2^2 - \Vert \bm{q}_-^{(T_1)} \Vert_2^2 \Big\vert = O \Big( N^{\frac{1}{2}} d_h^{-\frac{1}{4}} \log (d_h^{\frac{1}{2}}) \Big) = o (\Vert \bm{\mu} \Vert_2^2 \sigma_h^2 d_h),
\]
\[
    \Big\vert \Vert \bm{k}_\pm^{(t+1)} \Vert_2^2 - \Vert \bm{k}_\pm^{(T_1)} \Vert_2^2 \Big\vert = O \Big( (1 + \mathrm{SNR})N^{\frac{1}{2}} d_h^{-\frac{1}{4}} \log (d_h^{\frac{1}{2}}) \Big) = o (\Vert \bm{\mu} \Vert_2^2 \sigma_h^2 d_h),
\]
\[
    \Big\vert \Vert \bm{q}_{n, i}^{(t+1)} \Vert_2^2 - \Vert \bm{q}_{n, i}^{(T_1)} \Vert_2^2 \Big\vert = O \Big( d_h^{-\frac{1}{4}} \log (d_h^{\frac{1}{2}}) \Big) = o (\sigma_p^2 \sigma_h^2 d d_h),
\]
\[
    \Big\vert \Vert \bm{k}_{n, i}^{(t+1)} \Vert_2^2 - \Vert \bm{k}_{n, i}^{(T_1)} \Vert_2^2 \Big\vert = O \Big( N^{\frac{1}{2}} d_h^{-\frac{1}{4}} \log (d_h^{\frac{1}{2}}) \Big) = o (\sigma_p^2 \sigma_h^2 d d_h),
\]
so we have
\[
    \Vert \bm{q}_\pm^{(t+1)} \Vert_2^2, \Vert \bm{k}_\pm^{(t+1)} \Vert_2^2 = \Theta (\Vert \bm{\mu} \Vert_2^2 \sigma_h^2 d_h),
\]
\[
    \Vert \bm{q}_{n, i}^{(t+1)} \Vert_2^2, \Vert \bm{k}_{n, i}^{(t+1)} \Vert_2^2 = \Theta ( \sigma_p^2 \sigma_h^2 d d_h )
\]
for \(i \in [M] \backslash \{1\}, n \in [N]\).

\begin{align*}
    &\vert \langle \bm{q}_+^{(t+1)}, \bm{q}_-^{(t+1)} \rangle \vert \le \vert \langle \bm{q}_+^{(T_1)}, \bm{q}_-^{(T_1)} \rangle \vert + \sum\limits_{s = T_1}^{t} \Big\vert \langle \bm{q}_+^{(s+1)}, \bm{q}_-^{(s+1)} \rangle - \langle \bm{q}_+^{(s)}, \bm{q}_-^{(s)} \rangle \Big\vert \\
    &\le \vert \langle \bm{q}_+^{(T_1)}, \bm{q}_-^{(T_1)} \rangle \vert \\
    &+ \sum\limits_{s = T_1}^{t} \Big\vert \alpha_{+, +}^{(s)} \langle \bm{q}_-^{(s)}, \bm{k}_+^{(s)} \rangle + \sum\limits_{n \in S_+} \sum\limits_{i=2}^M \alpha_{n, +, i}^{(s)} \langle \bm{q}_-^{(s)}, \bm{k}_{n, i}^{(s)} \rangle \\
    &+ \alpha_{-, -}^{(s)} \langle \bm{q}_+^{(s)}, \bm{k}_-^{(s)} \rangle + \sum\limits_{n \in S_-} \sum\limits_{i=2}^M \alpha_{n, -, i}^{(s)} \langle \bm{q}_+^{(s)}, \bm{k}_{n, i}^{(s)} \rangle \\
    &+ \Big( \alpha_{+, +}^{(s)} \bm{k}_+^{(s)} + \sum\limits_{n \in S_+} \sum\limits_{i=2}^M \alpha_{n, +, i}^{(s)} \bm{k}_{n, i}^{(s)} \Big) \\
    &\cdot \Big( \alpha_{-, -}^{(s)} \bm{k}_-^{(s)\top} + \sum\limits_{n \in S_-} \sum\limits_{i=2}^M \alpha_{n, -, i}^{(s)} \bm{k}_{n, i}^{(s)\top} \Big) \Big\vert \\
    &\le \vert \langle \bm{q}_+^{(T_1)}, \bm{q}_-^{(T_1)} \rangle \vert \\
    &+ \sum\limits_{s = T_1}^{t} \vert \alpha_{+, +}^{(s)} \vert \vert \langle \bm{q}_-^{(s)}, \bm{k}_+^{(s)} \rangle \vert + \sum\limits_{n \in S_+} \sum\limits_{i=2}^M \sum\limits_{s = T_1}^{t} \vert \alpha_{n, +, i}^{(s)} \vert \vert \langle \bm{q}_-^{(s)}, \bm{k}_{n, i}^{(s)} \rangle \vert \\
    &+ \sum\limits_{s = T_1}^{t} \vert \alpha_{-, -}^{(s)} \vert \vert \langle \bm{q}_+^{(s)}, \bm{k}_-^{(s)} \rangle \vert + \sum\limits_{n \in S_-} \sum\limits_{i=2}^M \sum\limits_{s = T_1}^{t} \vert \alpha_{n, -, i}^{(s)} \vert \vert \langle \bm{q}_+^{(s)}, \bm{k}_{n, i}^{(s)} \rangle \vert \\
    &+ \{lower \ order \ term\} \\
    &\le \vert \langle \bm{q}_+^{(T_1)}, \bm{q}_-^{(T_1)} \rangle \vert \\
    &+ O \Big( N^{\frac{1}{2}} d_h^{-\frac{1}{4}} \Big) \cdot o(1) + N \cdot M \cdot O \Big( N^{-\frac{1}{2}} d_h^{-\frac{1}{4}} \Big) \cdot \log (d_h^{\frac{1}{2}}) \\
    &= \vert \langle \bm{q}_+^{(T_1)}, \bm{q}_-^{(T_1)} \rangle \vert + O \Big( N^{\frac{1}{2}} d_h^{-\frac{1}{4}} \log (d_h^{\frac{1}{2}}) \Big) \\
    &= o(1),
\end{align*}
where the first inequality is triangle inequality, the second inequality is by \eqref{q_p_q_m_dy}, the last equality is by \(d_h = \widetilde{\Omega} \Big( \max \{\mathrm{SNR}^4, \mathrm{SNR}^{-4}\} N^2 \epsilon^{-2} \Big)\).
\begin{align*}
    &\vert \langle \bm{q}_+^{(t+1)}, \bm{q}_{n, i}^{(t+1)} \rangle \vert \le \vert \langle \bm{q}_+^{(T_1)}, \bm{q}_{n, i}^{(T_1)} \rangle \vert + \sum\limits_{s = T_1}^{t} \Big\vert \langle \bm{q}_+^{(s+1)}, \bm{q}_{n, i}^{(s+1)} \rangle - \langle \bm{q}_+^{(s)}, \bm{q}_{n, i}^{(s)} \rangle \Big\vert \\
    &\le \vert \langle \bm{q}_+^{(T_1)}, \bm{q}_{n, i}^{(T_1)} \rangle \vert \\
    &+ \sum\limits_{s = T_1}^{t} \Big\vert \alpha_{+, +}^{(s)} \langle \bm{q}_{n, i}^{(s)}, \bm{k}_+^{(s)} \rangle + \sum\limits_{n^\prime \in S_+} \sum\limits_{l=2}^M \alpha_{n^\prime, +, l}^{(s)} \langle \bm{q}_{n, i}^{(s)}, \bm{k}_{n^\prime, l}^{(s)} \rangle \\
    &+ \alpha_{n, i, +}^{(s)} \langle \bm{q}_+^{(s)}, \bm{k}_+^{(s)} \rangle + \alpha_{n, i, -}^{(s)} \langle \bm{q}_+^{(s)}, \bm{k}_-^{(s)} \rangle + \sum\limits_{n^\prime = 1}^N \sum\limits_{l=2}^M \alpha_{n, i, n^\prime, l}^{(s)} \langle \bm{q}_+^{(s)}, \bm{k}_{n^\prime, l}^{(s)} \rangle \\
    &+ \Big( \alpha_{+, +}^{(s)} \bm{k}_+^{(s)} + \sum\limits_{n \in S_+} \sum\limits_{i=2}^M \alpha_{n, +, i}^{(s)} \bm{k}_{n, i}^{(s)} \Big) \\
    &\cdot \Big( \alpha_{n, i, +}^{(s)} \bm{k}_+^{(s)\top} + \alpha_{n, i, -}^{(s)} \bm{k}_-^{(s)\top} + \sum\limits_{n^\prime=1}^N \sum\limits_{l=2}^M \alpha_{n, i, n^\prime, l}^{(s)} \bm{k}_{n^\prime, l}^{(s)\top} \Big) \Big\vert \\
    &\le \vert \langle \bm{q}_+^{(T_1)}, \bm{q}_{n, i}^{(T_1)} \rangle \vert \\
    &+ \sum\limits_{s = T_1}^{t} \vert \alpha_{+, +}^{(s)} \vert \vert \langle \bm{q}_{n, i}^{(s)}, \bm{k}_+^{(s)} \rangle \vert + \sum\limits_{l=2}^M \sum\limits_{s = T_1}^{t} \vert \alpha_{n, +, l}^{(s)} \vert \vert \langle \bm{q}_{n, i}^{(s)}, \bm{k}_{n, l}^{(s)} \rangle \vert + \sum\limits_{n^\prime \in S_+ \wedge n^\prime \ne n} \sum\limits_{l=2}^M \sum\limits_{s = T_1}^{t} \vert \alpha_{n^\prime, +, l}^{(s)} \vert \vert \langle \bm{q}_{n, i}^{(s)}, \bm{k}_{n^\prime, l}^{(s)} \rangle \vert \\
    &+ \sum\limits_{s = T_1}^{t} \vert \alpha_{n, i, +}^{(s)} \vert \vert \langle \bm{q}_+^{(s)}, \bm{k}_+^{(s)} \rangle \vert + \sum\limits_{s = T_1}^{t} \vert \alpha_{n, i, -}^{(s)} \vert \vert \langle \bm{q}_+^{(s)}, \bm{k}_-^{(s)} \rangle \vert + \sum\limits_{l=2}^M \sum\limits_{s = T_1}^{t} \vert \alpha_{n, i, n, l}^{(s)} \vert \vert \langle \bm{q}_+^{(s)}, \bm{k}_{n, l}^{(s)} \rangle \vert \\
    &+ \sum\limits_{n^\prime \ne n} \sum\limits_{l=2}^M \sum\limits_{s = T_1}^{t} \vert \alpha_{n, i, n^\prime, l}^{(s)} \vert \vert \langle \bm{q}_+^{(s)}, \bm{k}_{n^\prime, l}^{(s)} \rangle \vert \\
    &+ \{lower \ order \ term\} \\
    &\le \vert \langle \bm{q}_+^{(T_1)}, \bm{q}_{n, i}^{(T_1)} \rangle \vert \\
    &+ O \Big( N^{\frac{1}{2}} d_h^{-\frac{1}{4}} \Big) \cdot \log (d_h^{\frac{1}{2}}) + M \cdot O \Big( N^{-\frac{1}{2}} d_h^{-\frac{1}{4}} \Big) \cdot \log (d_h^{\frac{1}{2}}) \\
    &+ N \cdot M \cdot O \Big( N^{-\frac{1}{2}} d_h^{-\frac{1}{4}} \Big) \cdot o(1) + O \Big( d_h^{-\frac{1}{4}} \Big) \cdot \log (d_h^{\frac{1}{2}}) \\
    &+ M \cdot O \Big( d_h^{-\frac{1}{4}} \Big) \cdot \log (d_h^{\frac{1}{2}}) + N \cdot M \cdot O \Big( d^{-\frac{1}{2}} d_h^{-\frac{1}{4}} \log (6N^2M^2 / \delta) \Big) \cdot \log (d_h^{\frac{1}{2}}) \\
    &= \vert \langle \bm{q}_+^{(T_1)}, \bm{q}_{n, i}^{(T_1)} \rangle \vert + O \Big( N^{\frac{1}{2}} d_h^{-\frac{1}{4}} \Big) + O \Big( N d^{-\frac{1}{2}} d_h^{-\frac{1}{4}} \log (6N^2M^2 / \delta) \log (d_h^{\frac{1}{2}}) \Big) \\
    &= o(1),
\end{align*}
where the first inequality is triangle inequality, the second inequality is by \eqref{q_p_q_i_dy}, the last equality is by \(d_h = \widetilde{\Omega} \Big( \max \{\mathrm{SNR}^4, \mathrm{SNR}^{-4}\} N^2 \epsilon^{-2} \Big)\) and \(d = \widetilde{\Omega} \Big( \epsilon^{-2} N^2 d_h \Big) \). Similarly, we have \(\vert \langle \bm{q}_-^{(t+1)}, \bm{q}_{n, i}^{(t+1)} \rangle \vert = o(1)\).
\begin{align*}
    &\vert \langle \bm{q}_{n, i}^{(t+1)}, \bm{q}_{n, j}^{(t+1)} \rangle \vert \le \vert \langle \bm{q}_{n, i}^{(T_1)}, \bm{q}_{n, j}^{(T_1)} \rangle \vert + \sum\limits_{s = T_1}^{t} \Big\vert \langle \bm{q}_{n, i}^{(s+1)}, \bm{q}_{n, j}^{(s+1)} \rangle - \langle \bm{q}_{n, i}^{(s)}, \bm{q}_{n, j}^{(s)} \rangle \Big\vert \\
    &\le \vert \langle \bm{q}_{n, i}^{(T_1)}, \bm{q}_{n, j}^{(T_1)} \rangle \vert \\
    &+ \sum\limits_{s = T_1}^{t} \Big\vert \alpha_{n, i, +}^{(s)} \langle \bm{q}_{n, j}^{(s)}, \bm{k}_+^{(s)} \rangle + \alpha_{n, i, -}^{(s)} \langle \bm{q}_{n, j}^{(s)}, \bm{k}_-^{(s)} \rangle + \sum\limits_{n^\prime = 1}^N \sum\limits_{l=2}^M \alpha_{n, i, n^\prime, l}^{(s)} \langle \bm{q}_{n, j}^{(s)}, \bm{k}_{n^\prime, l}^{(s)} \rangle \\
    &+ \alpha_{n, j, +}^{(s)} \langle \bm{q}_{n, i}^{(s)}, \bm{k}_+^{(s)} \rangle + \alpha_{n, j, -}^{(s)} \langle \bm{q}_{n, i}^{(s)}, \bm{k}_-^{(s)} \rangle + \sum\limits_{n^\prime = 1}^N \sum\limits_{l=2}^M \alpha_{n, j, n^\prime, l}^{(s)} \langle \bm{q}_{n, i}^{(s)}, \bm{k}_{n^\prime, l}^{(s)} \rangle \\
    &+ \Big( \alpha_{n, i, +}^{(s)} \bm{k}_+^{(s)} + \alpha_{n, i, -}^{(s)} \bm{k}_-^{(s)} + \sum\limits_{n^\prime = 1}^N \sum\limits_{l=2}^M \alpha_{n, i, n^\prime, l}^{(s)} \bm{k}_{n^\prime, l}^{(s)} \Big) \\
    &\cdot \Big( \alpha_{n, j, +}^{(s)} \bm{k}_+^{(s)\top} + \alpha_{n, j, -}^{(s)} \bm{k}_-^{(s)\top} + \sum\limits_{n^\prime = 1}^N \sum\limits_{l=2}^M \alpha_{n, j, n^\prime, l}^{(s)} \bm{k}_{n^\prime, l}^{(s)\top} \Big) \Big\vert \\
    &\le \vert \langle \bm{q}_{n, i}^{(T_1)}, \bm{q}_{n, j}^{(T_1)} \rangle \vert \\
    &+ \sum\limits_{s = T_1}^{t} \vert \alpha_{n, i, +}^{(s)} \vert \vert \langle \bm{q}_{n, j}^{(s)}, \bm{k}_+^{(s)} \rangle \vert + \sum\limits_{s = T_1}^{t} \vert \alpha_{n, i, -}^{(s)} \vert \vert \langle \bm{q}_{n, j}^{(s)}, \bm{k}_-^{(s)} \rangle \vert + \sum\limits_{l=2}^M \sum\limits_{s = T_1}^{t} \vert \alpha_{n, i, n, l}^{(s)} \vert \vert \langle \bm{q}_{n, j}^{(s)}, \bm{k}_{n, l}^{(s)} \rangle \vert \\
    &+ \sum\limits_{n^\prime \ne n} \sum\limits_{l=2}^M \sum\limits_{s = T_1}^{t} \vert \alpha_{n, i, n^\prime, l}^{(s)} \vert \vert \langle \bm{q}_{n, j}^{(s)}, \bm{k}_{n^\prime, l}^{(s)} \rangle \vert + \sum\limits_{s = T_1}^{t} \vert \alpha_{n, j, +}^{(s)} \vert \vert \langle \bm{q}_{n, i}^{(s)}, \bm{k}_+^{(s)} \rangle \vert + \sum\limits_{s = T_1}^{t} \vert \alpha_{n, j, -}^{(s)} \vert \vert \langle \bm{q}_{n, i}^{(s)}, \bm{k}_-^{(s)} \rangle \vert \\
    &+ \sum\limits_{l=2}^M \sum\limits_{s = T_1}^{t} \vert \alpha_{n, j, n, l}^{(s)} \vert \vert \langle \bm{q}_{n, i}^{(s)}, \bm{k}_{n, l}^{(s)} \rangle \vert + \sum\limits_{n^\prime \ne n} \sum\limits_{l=2}^M \sum\limits_{s = T_1}^{t} \vert \alpha_{n, j, n^\prime, l}^{(s)} \vert \vert \langle \bm{q}_{n, i}^{(s)}, \bm{k}_{n^\prime, l}^{(s)} \rangle \vert \\
    &+ \{lower \ order \ term\} \\
    &\le \vert \langle \bm{q}_{n, i}^{(T_1)}, \bm{q}_{n, j}^{(T_1)} \rangle \vert \\
    &+ O \Big( d_h^{-\frac{1}{4}} \Big) \cdot \log (d_h^{\frac{1}{2}}) + M \cdot O \Big( d_h^{-\frac{1}{4}} \Big) \cdot \log (d_h^{\frac{1}{2}}) \\
    &+ N \cdot M \cdot O \Big( d^{-\frac{1}{2}} d_h^{-\frac{1}{4}} \log (6N^2M^2 / \delta) \Big) \cdot o(1) \\
    &= \vert \langle \bm{q}_{n, i}^{(T_1)}, \bm{q}_{n, j}^{(T_1)} \rangle \vert + O \Big( d_h^{-\frac{1}{4}} \log (d_h^{\frac{1}{2}}) \Big) + o \Big( N d^{-\frac{1}{2}} d_h^{-\frac{1}{4}} \log (6N^2M^2 / \delta) \Big) \\
    &= o(1)
\end{align*}
for \(i, j \in [M] \backslash \{1\}, i \ne j, n \in [N]\). The first inequality is triangle inequality, the second inequality is by \eqref{q_i_q_j_dy}, the last equality is by \(d_h = \widetilde{\Omega} \Big( \max \{\mathrm{SNR}^4, \mathrm{SNR}^{-4}\} N^2 \epsilon^{-2} \Big)\) and \(d = \widetilde{\Omega} \Big( \epsilon^{-2} N^2 d_h \Big) \).
\begin{align*}
    &\vert \langle \bm{q}_{n, i}^{(t+1)}, \bm{q}_{\overline{n}, j}^{(t+1)} \rangle \vert \le \vert \langle \bm{q}_{n, i}^{(T_1)}, \bm{q}_{\overline{n}, j}^{(T_1)} \rangle \vert + \sum\limits_{s = T_1}^{t} \Big\vert \langle \bm{q}_{n, i}^{(s+1)}, \bm{q}_{\overline{n}, j}^{(s+1)} \rangle - \langle \bm{q}_{n, i}^{(s)}, \bm{q}_{\overline{n}, j}^{(s)} \rangle \Big\vert \\
    &\le \vert \langle \bm{q}_{n, i}^{(T_1)}, \bm{q}_{\overline{n}, j}^{(T_1)} \rangle \vert \\
    &+ \sum\limits_{s = T_1}^{t} \Big\vert \alpha_{n, i, +}^{(s)} \langle \bm{q}_{\overline{n}, j}^{(s)}, \bm{k}_+^{(s)} \rangle + \alpha_{n, i, -}^{(s)} \langle \bm{q}_{\overline{n}, j}^{(s)}, \bm{k}_-^{(s)} \rangle + \sum\limits_{n^\prime = 1}^N \sum\limits_{l=2}^M \alpha_{n, i, n^\prime, l}^{(s)} \langle \bm{q}_{\overline{n}, j}^{(s)}, \bm{k}_{n^\prime, l}^{(s)} \rangle \\
    &+ \alpha_{\overline{n}, j, +}^{(s)} \langle \bm{q}_{n, i}^{(s)}, \bm{k}_+^{(s)} \rangle + \alpha_{\overline{n}, j, -}^{(s)} \langle \bm{q}_{n, i}^{(s)}, \bm{k}_-^{(s)} \rangle + \sum\limits_{n^\prime = 1}^N \sum\limits_{l=2}^M \alpha_{\overline{n}, j, n^\prime, l}^{(s)} \langle \bm{q}_{n, i}^{(s)}, \bm{k}_{n^\prime, l}^{(s)} \rangle \\
    &+ \Big( \alpha_{n, i, +}^{(s)} \bm{k}_+^{(s)} + \alpha_{n, i, -}^{(s)} \bm{k}_-^{(s)} + \sum\limits_{n^\prime = 1}^N \sum\limits_{l=2}^M \alpha_{n, i, n^\prime, l}^{(s)} \bm{k}_{n^\prime, l}^{(s)} \Big) \\
    &\cdot \Big( \alpha_{\overline{n}, j, +}^{(s)} \bm{k}_+^{(s)\top} + \alpha_{\overline{n}, j, -}^{(s)} \bm{k}_-^{(s)\top} + \sum\limits_{n^\prime = 1}^N \sum\limits_{l=2}^M \alpha_{\overline{n}, j, n^\prime, l}^{(s)} \bm{k}_{n^\prime, l}^{(s)\top} \Big) \Big\vert \\
    &\le \vert \langle \bm{q}_{n, i}^{(T_1)}, \bm{q}_{\overline{n}, j}^{(T_1)} \rangle \vert \\
    &+ \sum\limits_{s = T_1}^{t} \vert \alpha_{n, i, +}^{(s)} \vert \vert \langle \bm{q}_{\overline{n}, j}^{(s)}, \bm{k}_+^{(s)} \rangle \vert + \sum\limits_{s = T_1}^{t} \vert \alpha_{n, i, -}^{(s)} \vert \vert \langle \bm{q}_{\overline{n}, j}^{(s)}, \bm{k}_-^{(s)} \rangle \vert + \sum\limits_{l=2}^M \sum\limits_{s = T_1}^{t} \vert \alpha_{n, i, \overline{n}, l}^{(s)} \vert \vert \langle \bm{q}_{\overline{n}, j}^{(s)}, \bm{k}_{\overline{n}, l}^{(s)} \rangle \vert \\
    &+ \sum\limits_{l=2}^M \sum\limits_{s = T_1}^{t} \vert \alpha_{n, i, n, l}^{(s)} \vert \vert \langle \bm{q}_{\overline{n}, j}^{(s)}, \bm{k}_{n, l}^{(s)} \rangle \vert + \sum\limits_{n^\prime \ne n \wedge n^\prime \ne \overline{n}} \sum\limits_{l=2}^M \sum\limits_{s = T_1}^{t} \vert \alpha_{n, i, n^\prime, l}^{(s)} \vert \vert \langle \bm{q}_{\overline{n}, j}^{(s)}, \bm{k}_{n^\prime, l}^{(s)} \rangle \vert \\
    &+ \sum\limits_{s = T_1}^{t} \vert \alpha_{\overline{n}, j, +}^{(s)} \vert \vert \langle \bm{q}_{n, i}^{(s)}, \bm{k}_+^{(s)} \rangle \vert + \sum\limits_{s = T_1}^{t} \vert \alpha_{\overline{n}, j, -}^{(s)} \vert \vert \langle \bm{q}_{n, i}^{(s)}, \bm{k}_-^{(s)} \rangle \vert + \sum\limits_{l=2}^M \sum\limits_{s = T_1}^{t} \vert \alpha_{\overline{n}, j, n, l}^{(s)} \vert \vert \langle \bm{q}_{n, i}^{(s)}, \bm{k}_{n, l}^{(s)} \rangle \vert \\
    &+ \sum\limits_{l=2}^M \sum\limits_{s = T_1}^{t} \vert \alpha_{\overline{n}, j, \overline{n}, l}^{(s)} \vert \vert \langle \bm{q}_{n, i}^{(s)}, \bm{k}_{\overline{n}, l}^{(s)} \rangle \vert + \sum\limits_{n^\prime \ne n \wedge n^\prime \ne \overline{n}} \sum\limits_{l=2}^M \sum\limits_{s = T_1}^{t} \vert \alpha_{\overline{n}, j, n^\prime, l}^{(s)} \vert \vert \langle \bm{q}_{n, i}^{(s)}, \bm{k}_{n^\prime, l}^{(s)} \rangle \vert \\
    &+ \{lower \ order \ term\} \\
    &= \vert \langle \bm{q}_{n, i}^{(T_1)}, \bm{q}_{\overline{n}, j}^{(T_1)} \rangle \vert \\
    &+ O \Big( d_h^{-\frac{1}{4}} \Big) \cdot \log (d_h^{\frac{1}{2}}) + M \cdot O \Big( d^{-\frac{1}{2}} d_h^{-\frac{1}{4}} \log (6N^2M^2 / \delta) \Big) \cdot \log (d_h^{\frac{1}{2}}) \\
    &+ M \cdot O \Big( d_h^{-\frac{1}{4}} \Big) \cdot o(1) + N \cdot M \cdot O \Big( d^{-\frac{1}{2}} d_h^{-\frac{1}{4}} \log (6N^2M^2 / \delta) \Big) \cdot o(1) \\
    &= \vert \langle \bm{q}_{n, i}^{(T_1)}, \bm{q}_{\overline{n}, j}^{(T_1)} \rangle \vert + O \Big( d_h^{-\frac{1}{4}} \log (d_h^{\frac{1}{2}}) \Big) + o \Big( N d^{-\frac{1}{2}} d_h^{-\frac{1}{4}} \log (6N^2M^2 / \delta) \Big) \\
    &= o(1)
\end{align*}
for \(i, j \in [M] \backslash \{1\}, n, \overline{n} \in [N], n \ne \overline{n}\). The first inequality is triangle inequality, the second inequality is by \eqref{q_i_q_j_n_dy}, the last equality is by \(d_h = \widetilde{\Omega} \Big( \max \{\mathrm{SNR}^4, \mathrm{SNR}^{-4}\} N^2 \epsilon^{-2} \Big)\) and \(d = \widetilde{\Omega} \Big( \epsilon^{-2} N^2 d_h \Big) \).
\begin{align*}
    &\vert \langle \bm{k}_+^{(t+1)}, \bm{k}_-^{(t+1)} \rangle \vert \le \vert \langle \bm{k}_+^{(T_1)}, \bm{k}_-^{(T_1)} \rangle \vert + \sum\limits_{s = T_1}^{t} \Big\vert \langle \bm{k}_+^{(s+1)}, \bm{k}_-^{(s+1)} \rangle - \langle \bm{k}_+^{(s)}, \bm{k}_-^{(s)} \rangle \Big\vert \\
    &\le \vert \langle \bm{k}_+^{(T_1)}, \bm{k}_-^{(T_1)} \rangle \vert \\
    &+ \sum\limits_{s = T_1}^{t} \Big\vert \beta_{+, +}^{(s)} \langle \bm{q}_+^{(s)}, \bm{k}_-^{(s)} \rangle + \sum\limits_{n \in S_+} \sum\limits_{i=2}^M \beta_{n, +, i}^{(s)} \langle \bm{q}_{n, i}^{(s)}, \bm{k}_-^{(s)} \rangle \\
    &+ \beta_{-, -}^{(s)} \langle \bm{q}_-^{(s)}, \bm{k}_+^{(s)} \rangle + \sum\limits_{n \in S_-} \sum\limits_{i=2}^M \beta_{n, -, i}^{(s)} \langle \bm{q}_{n, i}^{(s)}, \bm{k}_+^{(s)} \rangle \\
    &+ \Big( \beta_{+, +}^{(s)} \bm{q}_+^{(s)} + \sum\limits_{n \in S_+} \sum\limits_{i=2}^M \beta_{n, +, i}^{(s)} \bm{q}_{n, i}^{(s)} \Big) \\
    &\cdot \Big( \beta_{-, -}^{(s)} \bm{q}_-^{(s)\top} + \sum\limits_{n \in S_-} \sum\limits_{i=2}^M \beta_{n, -, i}^{(s)} \bm{q}_{n, i}^{(s)\top} \Big) \Big\vert \\
    &\le \vert \langle \bm{k}_+^{(T_1)}, \bm{k}_-^{(T_1)} \rangle \vert \\
    &+ \sum\limits_{s = T_1}^{t} \vert \beta_{+, +}^{(s)} \vert \vert \langle \bm{q}_+^{(s)}, \bm{k}_-^{(s)} \rangle \vert + \sum\limits_{n \in S_+} \sum\limits_{i=2}^M \sum\limits_{s = T_1}^{t} \vert \beta_{n, +, i}^{(s)} \vert \vert \langle \bm{q}_{n, i}^{(s)}, \bm{k}_-^{(s)} \rangle \vert \\
    &+ \sum\limits_{s = T_1}^{t} \vert \beta_{-, -}^{(s)} \vert \vert \langle \bm{q}_-^{(s)}, \bm{k}_+^{(s)} \rangle \vert + \sum\limits_{n \in S_-} \sum\limits_{i=2}^M \sum\limits_{s = T_1}^{t} \vert \beta_{n, -, i}^{(s)} \vert \vert \langle \bm{q}_{n, i}^{(s)}, \bm{k}_+^{(s)} \rangle \vert \\
    &+ \{lower \ order \ term\} \\
    &= \vert \langle \bm{k}_+^{(T_1)}, \bm{k}_-^{(T_1)} \rangle \vert \\
    &+ O \Big( N^{\frac{1}{2}} d_h^{-\frac{1}{4}} \Big) \cdot \log (d_h^{\frac{1}{2}}) + N \cdot M \cdot O \Big(\mathrm{SNR} \cdot N^{-\frac{1}{2}} d_h^{-\frac{1}{4}} \Big) \cdot \log (d_h^{\frac{1}{2}}) \\
    &= \vert \langle \bm{k}_+^{(T_1)}, \bm{k}_-^{(T_1)} \rangle \vert + O \Big(\mathrm{SNR} \cdot N^{\frac{1}{2}} d_h^{-\frac{1}{4}} \log (d_h^{\frac{1}{2}}) \Big) \\
    &= o(1),
\end{align*}
where the first inequality is triangle inequality, the second inequality is by \eqref{k_p_k_m_dy}, the last equality is by \(d_h = \widetilde{\Omega} \Big( \max \{\mathrm{SNR}^4, \mathrm{SNR}^{-4}\} N^2 \epsilon^{-2} \Big)\).
\begin{align*}
    &\vert \langle \bm{k}_+^{(t+1)}, \bm{k}_{n, i}^{(t+1)} \rangle \vert \le \vert \langle \bm{k}_+^{(T_1)}, \bm{k}_{n, i}^{(T_1)} \rangle \vert + \sum\limits_{s = T_1}^{t} \Big\vert \langle \bm{k}_+^{(s+1)}, \bm{k}_{n, i}^{(s+1)} \rangle - \langle \bm{k}_+^{(s)}, \bm{k}_{n, i}^{(s)} \rangle \Big\vert \\
    &\le \vert \langle \bm{k}_+^{(T_1)}, \bm{k}_{n, i}^{(T_1)} \rangle \vert \\
    &+ \sum\limits_{s = T_1}^{t} \Big\vert \beta_{+, +}^{(s)} \langle \bm{q}_+^{(s)}, \bm{k}_{n, i}^{(s)} \rangle + \sum\limits_{n^\prime \in S_+} \sum\limits_{l=2}^M \beta_{n^\prime, +, l}^{(s)} \langle \bm{q}_{n^\prime, l}^{(s)}, \bm{k}_{n, i}^{(s)} \rangle \\
    &+ \beta_{n, i, +}^{(s)} \langle \bm{q}_+^{(s)}, \bm{k}_+^{(s)} \rangle + \beta_{n, i, -}^{(s)} \langle \bm{q}_-^{(s)}, \bm{k}_+^{(s)} \rangle + \sum\limits_{n^\prime = 1}^N \sum\limits_{l=2}^M \beta_{n, i, n^\prime, l}^{(s)} \langle \bm{q}_{n^\prime, l}^{(s)}, \bm{k}_+^{(s)} \rangle \\
    &+ \Big( \beta_{+, +}^{(s)} \bm{q}_+^{(s)} + \sum\limits_{n^\prime \in S_+} \sum\limits_{l=2}^M \beta_{n^\prime, +, l}^{(s)} \bm{q}_{n^\prime, l}^{(s)} \Big) \\
    &\cdot \Big( \beta_{n, i, +}^{(s)} \bm{q}_+^{(s)\top} + \beta_{n, i, -}^{(s)} \bm{q}_-^{(s)\top} + \sum\limits_{n^\prime = 1}^N \sum\limits_{l=2}^M \beta_{n, i, n^\prime, l}^{(s)} \bm{q}_{n^\prime, l}^{(s)\top} \Big) \Big\vert \\
    &\le \vert \langle \bm{k}_+^{(T_1)}, \bm{k}_{n, i}^{(T_1)} \rangle \vert \\
    &+ \sum\limits_{s = T_1}^{t} \vert \beta_{+, +}^{(s)} \vert \vert \langle \bm{q}_+^{(s)}, \bm{k}_{n, i}^{(s)} \rangle \vert + \sum\limits_{l=2}^M \sum\limits_{s = T_1}^{t} \vert \beta_{n, +, l}^{(s)} \vert \vert \langle \bm{q}_{n, l}^{(s)}, \bm{k}_{n, i}^{(s)} \rangle \vert \\
    &+ \sum\limits_{n^\prime \in S_+ \wedge n^\prime \ne n} \sum\limits_{l=2}^M \sum\limits_{s = T_1}^{t} \vert \beta_{n^\prime, +, l}^{(s)} \vert \vert \langle \bm{q}_{n^\prime, l}^{(s)}, \bm{k}_{n, i}^{(s)} \rangle \vert + \sum\limits_{s = T_1}^{t} \vert \beta_{n, i, +}^{(s)} \vert \vert \langle \bm{q}_+^{(s)}, \bm{k}_+^{(s)} \rangle \vert \\
    &+ \sum\limits_{s = T_1}^{t} \vert \beta_{n, i, -}^{(s)} \vert \vert \langle \bm{q}_-^{(s)}, \bm{k}_+^{(s)} \rangle \vert + \sum\limits_{l=2}^M \sum\limits_{s = T_1}^{t} \vert \beta_{n, i, n, l}^{(s)} \vert \vert \langle \bm{q}_{n, l}^{(s)}, \bm{k}_+^{(s)} \rangle \vert \\
    &+ \sum\limits_{n^\prime \ne n}^N \sum\limits_{l=2}^M \sum\limits_{s = T_1}^{t} \vert \beta_{n, i, n^\prime, l}^{(s)} \vert \vert \langle \bm{q}_{n^\prime, l}^{(s)}, \bm{k}_+^{(s)} \rangle \vert \\
    &+ \{lower \ order \ term\} \\
    &= \vert \langle \bm{k}_+^{(T_1)}, \bm{k}_{n, i}^{(T_1)} \rangle \vert \\
    &+ O \Big( N^{\frac{1}{2}} d_h^{-\frac{1}{4}} \Big) \cdot \log (d_h^{\frac{1}{2}}) + M \cdot O \Big(\mathrm{SNR} \cdot N^{-\frac{1}{2}} d_h^{-\frac{1}{4}} \Big) \cdot \log (d_h^{\frac{1}{2}}) \\
    &+ N \cdot M \cdot O \Big(\mathrm{SNR} \cdot N^{-\frac{1}{2}} d_h^{-\frac{1}{4}} \Big) \cdot o(1) + O \Big( N^{\frac{1}{2}} d_h^{-\frac{1}{4}} \Big) \cdot \log (d_h^{\frac{1}{2}}) \\
    &+ M \cdot O \Big( d_h^{-\frac{1}{4}} \Big) \cdot \log (d_h^{\frac{1}{2}}) + N \cdot M \cdot O \Big( d^{-\frac{1}{2}} d_h^{-\frac{1}{4}} \log (6N^2M^2 / \delta) \Big) \cdot \log (d_h^{\frac{1}{2}}) \\
    &= \vert \langle \bm{k}_+^{(T_1)}, \bm{k}_{n, i}^{(T_1)} \rangle \vert + o \Big(\mathrm{SNR} \cdot N^{\frac{1}{2}} d_h^{-\frac{1}{4}} \Big) + O \Big( N^{\frac{1}{2}} d_h^{-\frac{1}{4}} \log (d_h^{\frac{1}{2}}) \Big) \\
    &+ O \Big( N d^{-\frac{1}{2}} d_h^{-\frac{1}{4}} \log (6N^2M^2 / \delta) \cdot \log (d_h^{\frac{1}{2}}) \Big) \\
    &= o(1)
\end{align*}
where the first inequality is triangle inequality, the second inequality is by \eqref{k_p_k_i_dy}, the last equality is by \(d_h = \widetilde{\Omega} \Big( \max \{\mathrm{SNR}^4, \mathrm{SNR}^{-4}\} N^2 \epsilon^{-2} \Big)\) and \(d = \widetilde{\Omega} \Big( \epsilon^{-2} N^2 d_h \Big) \). Similarly, we have \(\vert \langle \bm{k}_-^{(t+1)}, \bm{k}_{n, i}^{(t+1)} \rangle \vert = o(1)\).
\begin{align*}
    &\vert \langle \bm{k}_{n, i}^{(t+1)}, \bm{k}_{n, j}^{(t+1)} \rangle \vert \le \vert \langle \bm{k}_{n, i}^{(T_1)}, \bm{k}_{n, j}^{(T_1)} \rangle \vert + \sum\limits_{s = T_1}^{t} \Big\vert \langle \bm{k}_{n, i}^{(s+1)}, \bm{k}_{n, j}^{(s+1)} \rangle - \langle \bm{k}_{n, i}^{(s)}, \bm{k}_{n, j}^{(s)} \rangle \Big\vert \\
    &\le \vert \langle \bm{k}_{n, i}^{(T_1)}, \bm{k}_{n, j}^{(T_1)} \rangle \vert \\
    &+ \sum\limits_{s = T_1}^{t} \Big\vert \beta_{n, i, +}^{(s)} \langle \bm{q}_+^{(s)}, \bm{k}_{n, j}^{(s)} \rangle + \beta_{n, i, -}^{(s)} \langle \bm{q}_-^{(s)}, \bm{k}_{n, j}^{(s)} \rangle + \sum\limits_{n^\prime = 1}^N \sum\limits_{l=2}^M \beta_{n, i, n^\prime, l}^{(s)} \langle \bm{q}_{n^\prime, l}^{(s)}, \bm{k}_{n, j}^{(s)} \rangle \\
    &+ \beta_{n, j, +}^{(s)} \langle \bm{q}_+^{(s)}, \bm{k}_{n, i}^{(s)} \rangle + \beta_{n, j, -}^{(s)} \langle \bm{q}_-^{(s)}, \bm{k}_{n, i}^{(s)} \rangle + \sum\limits_{n^\prime = 1}^N \sum\limits_{l=2}^M \beta_{n, j, n^\prime, l}^{(s)} \langle \bm{q}_{n^\prime, l}^{(s)}, \bm{k}_{n, i}^{(s)} \rangle \\
    &+ \Big( \beta_{n, i, +}^{(s)} \bm{q}_+^{(s)} + \beta_{n, i, -}^{(s)} \bm{q}_-^{(s)} + \sum\limits_{n^\prime = 1}^N \sum\limits_{l=2}^M \beta_{n, i, n^\prime, l}^{(s)} \bm{q}_{n^\prime, l}^{(s)} \Big) \\
    &\cdot \Big( \beta_{n, j, +}^{(s)} \bm{q}_+^{(s)\top} + \beta_{n, j, -}^{(s)} \bm{q}_-^{(s)\top} + \sum\limits_{n^\prime = 1}^N \sum\limits_{l=2}^M \beta_{n, j, n^\prime, l}^{(s)} \bm{q}_{n^\prime, l}^{(s)\top} \Big) \Big\vert \\
    &\le \vert \langle \bm{k}_{n, i}^{(T_1)}, \bm{k}_{n, j}^{(T_1)} \rangle \vert \\
    &+ \sum\limits_{s = T_1}^{t} \vert \beta_{n, i, +}^{(s)} \vert \vert \langle \bm{q}_+^{(s)}, \bm{k}_{n, j}^{(s)} \rangle \vert + \sum\limits_{s = T_1}^{t} \vert \beta_{n, i, -}^{(s)} \vert \vert \langle \bm{q}_-^{(s)}, \bm{k}_{n, j}^{(s)} \rangle \vert + \sum\limits_{l=2}^M \sum\limits_{s = T_1}^{t} \vert \beta_{n, i, n, l}^{(s)} \vert \vert \langle \bm{q}_{n, l}^{(s)}, \bm{k}_{n, j}^{(s)} \rangle \vert \\
    &+ \sum\limits_{n^\prime \ne n}^N \sum\limits_{l=2}^M \sum\limits_{s = T_1}^{t} \vert \beta_{n, i, n^\prime, l}^{(s)} \vert \vert \langle \bm{q}_{n^\prime, l}^{(s)}, \bm{k}_{n, j}^{(s)} \rangle \vert + \sum\limits_{s = T_1}^{t} \vert \beta_{n, j, +}^{(s)} \vert \vert \langle \bm{q}_+^{(s)}, \bm{k}_{n, i}^{(s)} \rangle \vert + \sum\limits_{s = T_1}^{t} \vert \beta_{n, j, -}^{(s)} \vert \vert \langle \bm{q}_-^{(s)}, \bm{k}_{n, i}^{(s)} \rangle \vert \\
    &+ \sum\limits_{l=2}^M \sum\limits_{s = T_1}^{t} \vert \beta_{n, j, n, l}^{(s)} \vert \vert \langle \bm{q}_{n, l}^{(s)}, \bm{k}_{n, i}^{(s)} \rangle \vert + \sum\limits_{n^\prime \ne n}^N \sum\limits_{l=2}^M \sum\limits_{s = T_1}^{t} \vert \beta_{n, j, n^\prime, l}^{(s)} \vert \vert \langle \bm{q}_{n^\prime, l}^{(s)}, \bm{k}_{n, i}^{(s)} \rangle \vert \\
    &+ \{lower \ order \ term\} \\
    &= \vert \langle \bm{k}_{n, i}^{(T_1)}, \bm{k}_{n, j}^{(T_1)} \rangle \vert \\
    &+ O \Big( N^{\frac{1}{2}} d_h^{-\frac{1}{4}} \Big) \cdot \log (d_h^{\frac{1}{2}}) + M \cdot O \Big( d_h^{-\frac{1}{4}} \Big) \cdot \log (d_h^{\frac{1}{2}}) \\
    &+ N \cdot M \cdot O \Big( d^{-\frac{1}{2}} d_h^{-\frac{1}{4}} \log (6N^2M^2 / \delta) \Big) \cdot o(1)\\
    &= \vert \langle \bm{k}_{n, i}^{(T_1)}, \bm{k}_{n, j}^{(T_1)} \rangle \vert + O \Big( N^{\frac{1}{2}} d_h^{-\frac{1}{4}} \log (d_h^{\frac{1}{2}}) \Big) + o \Big( N d^{-\frac{1}{2}} d_h^{-\frac{1}{4}} \log (6N^2M^2 / \delta) \Big) \\
    &= o(1)
\end{align*}
for \(i, j \in [M] \backslash \{1\}, i \ne j, n \in [N]\). The first inequality is triangle inequality, the second inequality is by \eqref{k_i_k_j_dy}, the last equality is by \(d_h = \widetilde{\Omega} \Big( \max \{\mathrm{SNR}^4, \mathrm{SNR}^{-4}\} N^2 \epsilon^{-2} \Big)\) and \(d = \widetilde{\Omega} \Big( \epsilon^{-2} N^2 d_h \Big) \).
\begin{align*}
    &\vert \langle \bm{k}_{n, i}^{(t+1)}, \bm{k}_{\overline{n}, j}^{(t+1)} \rangle \vert \le \vert \langle \bm{k}_{n, i}^{(T_1)}, \bm{k}_{\overline{n}, j}^{(T_1)} \rangle \vert + \sum\limits_{s = T_1}^{t} \Big\vert \langle \bm{k}_{n, i}^{(s+1)}, \bm{k}_{\overline{n}, j}^{(s+1)} \rangle - \langle \bm{k}_{n, i}^{(s)}, \bm{k}_{\overline{n}, j}^{(s)} \rangle \Big\vert \\
    &\le \vert \langle \bm{k}_{n, i}^{(T_1)}, \bm{k}_{\overline{n}, j}^{(T_1)} \rangle \vert \\
    &+ \sum\limits_{s = T_1}^{t} \Big\vert \beta_{n, i, +}^{(s)} \langle \bm{q}_+^{(s)}, \bm{k}_{\overline{n}, j}^{(s)} \rangle + \beta_{n, i, -}^{(s)} \langle \bm{q}_-^{(s)}, \bm{k}_{\overline{n}, j}^{(s)} \rangle + \sum\limits_{n^\prime = 1}^N \sum\limits_{l=2}^M \beta_{n, i, n^\prime, l}^{(s)} \langle \bm{q}_{n^\prime, l}^{(s)}, \bm{k}_{\overline{n}, j}^{(s)} \rangle \\
    &+ \beta_{\overline{n}, j, +}^{(s)} \langle \bm{q}_+^{(s)}, \bm{k}_{n, i}^{(s)} \rangle + \beta_{\overline{n}, j, -}^{(s)} \langle \bm{q}_-^{(s)}, \bm{k}_{n, i}^{(s)} \rangle + \sum\limits_{n^\prime = 1}^N \sum\limits_{l=2}^M \beta_{\overline{n}, j, n^\prime, l}^{(s)} \langle \bm{q}_{n^\prime, l}^{(s)}, \bm{k}_{n, i}^{(s)} \rangle \\
    &+ \Big( \beta_{n, i, +}^{(s)} \bm{q}_+^{(s)} + \beta_{n, i, -}^{(s)} \bm{q}_-^{(s)} + \sum\limits_{n^\prime = 1}^N \sum\limits_{l=2}^M \beta_{n, i, n^\prime, l}^{(s)} \bm{q}_{n^\prime, l}^{(s)} \Big) \\
    &\cdot \Big( \beta_{\overline{n}, j, +}^{(s)} \bm{q}_+^{(s)\top} + \beta_{\overline{n}, j, -}^{(s)} \bm{q}_-^{(s)\top} + \sum\limits_{n^\prime = 1}^N \sum\limits_{l=2}^M \beta_{\overline{n}, j, n^\prime, l}^{(s)} \bm{q}_{n^\prime, l}^{(s)\top} \Big) \Big\vert \\
    &\le \vert \langle \bm{k}_{n, i}^{(T_1)}, \bm{k}_{\overline{n}, j}^{(T_1)} \rangle \vert \\
    &+ \sum\limits_{s = T_1}^{t} \vert \beta_{n, i, +}^{(s)} \vert \vert \langle \bm{q}_+^{(s)}, \bm{k}_{\overline{n}, j}^{(s)} \rangle \vert + \sum\limits_{s = T_1}^{t} \vert \beta_{n, i, -}^{(s)} \vert \vert \langle \bm{q}_-^{(s)}, \bm{k}_{\overline{n}, j}^{(s)} \rangle \vert + \sum\limits_{l=2}^M \sum\limits_{s = T_1}^{t} \vert \beta_{n, i, \overline{n}, l}^{(s)} \vert \vert \langle \bm{q}_{\overline{n}, l}^{(s)}, \bm{k}_{\overline{n}, j}^{(s)} \rangle \vert \\
    &+ \sum\limits_{l=2}^M \sum\limits_{s = T_1}^{t} \vert \beta_{n, i, n, l}^{(s)} \vert \vert \langle \bm{q}_{n, l}^{(s)}, \bm{k}_{\overline{n}, j}^{(s)} \rangle \vert + \sum\limits_{n^\prime \ne n \wedge n^\prime \overline{n}} \sum\limits_{l=2}^M \sum\limits_{s = T_1}^{t} \vert \beta_{n, i, n^\prime, l}^{(s)} \vert \vert \langle \bm{q}_{n^\prime, l}^{(s)}, \bm{k}_{\overline{n}, j}^{(s)} \rangle \vert \\
    &+ \sum\limits_{s = T_1}^{t} \vert \beta_{\overline{n}, j, +}^{(s)} \vert \vert \langle \bm{q}_+^{(s)}, \bm{k}_{n, i}^{(s)} \rangle \vert + \sum\limits_{s = T_1}^{t} \vert \beta_{\overline{n}, j, -}^{(s)} \vert \vert \langle \bm{q}_-^{(s)}, \bm{k}_{n, i}^{(s)} \rangle \vert + \sum\limits_{l=2}^M \sum\limits_{s = T_1}^{t} \vert \beta_{\overline{n}, j, n, l}^{(s)} \vert \vert \langle \bm{q}_{n, l}^{(s)}, \bm{k}_{n, i}^{(s)} \rangle \vert \\
    &+ \sum\limits_{l=2}^M \sum\limits_{s = T_1}^{t} \vert \beta_{\overline{n}, j, \overline{n}, l}^{(s)} \vert \vert \langle \bm{q}_{\overline{n}, l}^{(s)}, \bm{k}_{n, i}^{(s)} \rangle \vert + \sum\limits_{n^\prime \ne n \wedge n^\prime \overline{n}} \sum\limits_{l=2}^M \sum\limits_{s = T_1}^{t} \vert \beta_{\overline{n}, j, n^\prime, l}^{(s)} \vert \vert \langle \bm{q}_{n^\prime, l}^{(s)}, \bm{k}_{n, i}^{(s)} \rangle \vert \\
    &+ \{lower \ order \ term\} \\
    &= \vert \langle \bm{k}_{n, i}^{(T_1)}, \bm{k}_{\overline{n}, j}^{(T_1)} \rangle \vert \\
    &+ O \Big( N^{\frac{1}{2}} d_h^{-\frac{1}{4}} \Big) \cdot \log (d_h^{\frac{1}{2}}) + M \cdot O \Big( d^{-\frac{1}{2}} d_h^{-\frac{1}{4}} \log (6N^2M^2 / \delta) \Big) \cdot \log (d_h^{\frac{1}{2}}) \\
    &+ M \cdot O \Big( d_h^{-\frac{1}{4}} \Big) \cdot o(1) + N \cdot M \cdot O \Big( d^{-\frac{1}{2}} d_h^{-\frac{1}{4}} \log (6N^2M^2 / \delta) \Big) \cdot o(1) \\
    &= \vert \langle \bm{k}_{n, i}^{(T_1)}, \bm{k}_{\overline{n}, j}^{(T_1)} \rangle \vert \\
    &+ O \Big( N^{\frac{1}{2}} d_h^{-\frac{1}{4}} \log (d_h^{\frac{1}{2}}) \Big) + o \Big( N d^{-\frac{1}{2}} d_h^{-\frac{1}{4}} \log (6N^2M^2 / \delta) \Big) \\
    &= o(1)
\end{align*}
for \(i, j \in [M] \backslash \{1\}, n, \overline{n} \in [N], n \ne \overline{n}\). The first inequality is triangle inequality, the second inequality is by \eqref{k_i_k_j_n_dy}, the last equality is by \(d_h = \widetilde{\Omega} \Big( \max \{\mathrm{SNR}^4, \mathrm{SNR}^{-4}\} N^2 \epsilon^{-2} \Big)\) and \(d = \widetilde{\Omega} \Big( \epsilon^{-2} N^2 d_h \Big) \).

\subsection{Upper Bounds of \(\langle\) q, k \(\rangle\)}
\label{upper_bound_of_qk2}
In order to give the upper bounds for \( \langle \bm{q}, \bm{k} \rangle \) in stage \uppercase\expandafter{\romannumeral3}, we need to give the upper bounds of \(\alpha\) and \(\beta\) based on the equations in \ref{alpha_and_beta}. The main difference between this subsection and \ref{upper_bound_of_qk} is that the bounds of \(\vert V_\pm \vert, \vert V_{n, i} \vert\) is \(\log \big( O(\frac{1}{\epsilon}) \big)\) in this subsection, while the bounds of \(\vert V_\pm \vert, \vert V_{n, i} \vert\) is \(\log \big( O(\frac{1}{\epsilon}) \big)\) in \ref{upper_bound_of_qk}, resulting in different bounds for \(\alpha\) and \(\beta\). Now we take \(\alpha_{+, +}^{(s)}\) as an example
\begin{align*}
    &\alpha_{+, +}^{(s)} = \frac{\eta}{NM} \sum\limits_{n \in S_+} - \ell_n^{\prime(s)} \Vert \bm{\mu} \Vert_2^2 \\
    &\cdot \Big( V_+^{(s)} \big( \frac{ \exp (\langle \bm{q}_+^{(s)}, \bm{k}_+^{(s)} \rangle)}{ \exp  (\langle \bm{q}_+^{(s)}, \bm{k}_+^{(s)} \rangle) + \sum\limits_{j=2}^M  \exp  ( \langle \bm{q}_+^{(s)}, \bm{k}_{n, j}^{(s)} \rangle ) } \\
    &- (\frac{ \exp (\langle \bm{q}_+^{(s)}, \bm{k}_+^{(s)} \rangle)}{ \exp  (\langle \bm{q}_+^{(s)}, \bm{k}_+^{(s)} \rangle) + \sum\limits_{j=2}^M  \exp  ( \langle \bm{q}_+^{(s)}, \bm{k}_{n, j}^{(s)} \rangle ) })^2 \big) \\
    &- \sum\limits_{i=2}^M \big( V_{n, i}^{(s)} \cdot \frac{ \exp (\langle \bm{q}_+^{(s)}, \bm{k}_+^{(s)} \rangle)}{ \exp  (\langle \bm{q}_+^{(s)}, \bm{k}_+^{(s)} \rangle) + \sum\limits_{j=2}^M  \exp  ( \langle \bm{q}_+^{(s)}, \bm{k}_{n, j}^{(s)} \rangle ) } \\
    &\cdot \frac{ \exp (\langle \bm{q}_+^{(s)}, \bm{k}_{n, i}^{(s)} \rangle)}{ \exp  (\langle \bm{q}_+^{(s)}, \bm{k}_+^{(s)} \rangle) + \sum\limits_{j=2}^M  \exp  ( \langle \bm{q}_+^{(s)}, \bm{k}_{n, j}^{(s)} \rangle ) } \big) \Big) \\
    &= \frac{\eta}{NM} \sum\limits_{n \in S_+} - \ell_n^{\prime(s)} \Vert \bm{\mu} \Vert_2^2 \frac{ \exp (\langle \bm{q}_+^{(s)}, \bm{k}_+^{(s)} \rangle)}{ \exp  (\langle \bm{q}_+^{(s)}, \bm{k}_+^{(s)} \rangle) + \sum\limits_{j=2}^M  \exp  ( \langle \bm{q}_+^{(s)}, \bm{k}_{n, j}^{(s)} \rangle ) } \\
    &\cdot \Big( V_+^{(s)} \cdot \frac{ \sum\limits_{j=2}^M  \exp  ( \langle \bm{q}_+^{(s)}, \bm{k}_{n, j}^{(s)} \rangle ) }{ \exp  (\langle \bm{q}_+^{(s)}, \bm{k}_+^{(s)} \rangle) + \sum\limits_{j=2}^M  \exp  ( \langle \bm{q}_+^{(s)}, \bm{k}_{n, j}^{(s)} \rangle ) } \\
    &- \sum\limits_{i=2}^M V_{n, i}^{(s)} \cdot \frac{ \exp (\langle \bm{q}_+^{(s)}, \bm{k}_{n, i}^{(s)} \rangle)}{ \exp  (\langle \bm{q}_+^{(s)}, \bm{k}_+^{(s)} \rangle) + \sum\limits_{j=2}^M  \exp  ( \langle \bm{q}_+^{(s)}, \bm{k}_{n, j}^{(s)} \rangle ) } \Big) \\
    &\le \frac{\eta}{NM} \sum\limits_{n \in S_+} \Vert \bm{\mu} \Vert_2^2 \cdot \Big( V_+^{(s)} \cdot \frac{ \sum\limits_{j=2}^M  \exp  ( \langle \bm{q}_+^{(s)}, \bm{k}_{n, j}^{(s)} \rangle ) }{ \exp  (\langle \bm{q}_+^{(s)}, \bm{k}_+^{(s)} \rangle) + \sum\limits_{j=2}^M  \exp  ( \langle \bm{q}_+^{(s)}, \bm{k}_{n, j}^{(s)} \rangle ) } \\
    &+ \max\limits_{i} \vert V_{n, i}^{(s)} \vert \cdot \frac{ \sum\limits_{j=2}^M  \exp  ( \langle \bm{q}_+^{(s)}, \bm{k}_{n, j}^{(s)} \rangle ) }{ \exp  (\langle \bm{q}_+^{(s)}, \bm{k}_+^{(s)} \rangle) + \sum\limits_{j=2}^M  \exp  ( \langle \bm{q}_+^{(s)}, \bm{k}_{n, j}^{(s)} \rangle ) } \Big) \\
    &\le \frac{\eta}{NM} \cdot \frac{3N}{4} \cdot \Vert \bm{\mu} \Vert_2^2 \cdot \Big( V_+^{(s)} \cdot \frac{ C }{ \exp  (\langle \bm{q}_+^{(s)}, \bm{k}_+^{(s)} \rangle) } + \max\limits_{i} \vert V_{n, i}^{(s)} \vert \cdot \frac{ C }{ \exp  (\langle \bm{q}_+^{(s)}, \bm{k}_+^{(s)} \rangle) } \Big) \\
    &\le \frac{\eta C_9 \Vert \bm{\mu} \Vert_2^2 \log \big( O(\frac{1}{\epsilon}) \big) }{\exp ( \langle \bm{q}_+^{(s)}, \bm{k}_+^{(s)} \rangle)},
\end{align*}
where the first inequality is by \(- \ell_n^{\prime(s)} \le 1\) and \(softmax(\langle \bm{q}_+^{(s)}, \bm{k}_+^{(s)} \rangle) \le 1\). For the second inequality, we first consider \(\frac{ \sum\limits_{j=2}^M  \exp  ( \langle \bm{q}_+^{(s)}, \bm{k}_{n, j}^{(s)} \rangle ) }{ \exp  (\langle \bm{q}_+^{(s)}, \bm{k}_+^{(s)} \rangle) + \sum\limits_{j=2}^M  \exp  ( \langle \bm{q}_+^{(s)}, \bm{k}_{n, j}^{(s)} \rangle ) } \le \frac{ \sum\limits_{j=2}^M  \exp  ( \langle \bm{q}_+^{(s)}, \bm{k}_{n, j}^{(s)} \rangle ) }{ \exp  (\langle \bm{q}_+^{(s)}, \bm{k}_+^{(s)} \rangle) }\), then by the monotonicity of \(\langle \bm{q}_+^{(s)}, \bm{k}_{n, j}^{(s)} \rangle\) and \(\langle \bm{q}_+^{(T_1)}, \bm{k}_{n, j}^{(T_1)} \rangle = o(1)\) we have \(\sum\limits_{j=2}^M  \exp  ( \langle \bm{q}_+^{(s)}, \bm{k}_{n, j}^{(s)} \rangle ) \le C\) for \(t \in [T_1, T_3]\). The last inequality is by \(V_+^{(s)}, \vert V_{n, i}^{(s)} \vert \le 2\log \big( O(\frac{1}{\epsilon}) \big)\) for \(t \in [T_2, T_3]\) and absorbing the constant factors. 
Similar to \ref{upper_bound_of_qk}, we can give the bounds for the other \(\alpha\) and \(\beta\) as follows:
\begin{equation*}
    \alpha_{-, -}^{(s)} \le \frac{\eta C_9 \Vert \bm{\mu} \Vert_2^2 \log \big( O(\frac{1}{\epsilon}) \big) }{\exp ( \langle \bm{q}_-^{(s)}, \bm{k}_-^{(s)} \rangle)},
\end{equation*}
\begin{equation*}
    \beta_{+, +}^{(s)} \le \frac{\eta C_9 \Vert \bm{\mu} \Vert_2^2 \log \big( O(\frac{1}{\epsilon}) \big) }{\exp ( \langle \bm{q}_+^{(s)}, \bm{k}_+^{(s)} \rangle)},
\end{equation*}
\begin{equation*}
    \beta_{-, -}^{(s)} \le \frac{\eta C_9 \Vert \bm{\mu} \Vert_2^2 \log \big( O(\frac{1}{\epsilon}) \big) }{\exp ( \langle \bm{q}_-^{(s)}, \bm{k}_-^{(s)} \rangle)}.
\end{equation*}

\begin{align*}
    \alpha_{n, +, j}^{(s)} \ge - \frac{\eta C_9 \Vert \bm{\mu} \Vert_2^2 \log \big( O(\frac{1}{\epsilon}) \big)}{N} \cdot \exp (\langle \bm{q}_+^{(s)}, \bm{k}_{n, j}^{(s)} \rangle),
\end{align*}

\begin{align*}
    \alpha_{n, -, j}^{(s)} \ge - \frac{\eta C_9 \Vert \bm{\mu} \Vert_2^2 \log \big( O(\frac{1}{\epsilon}) \big)}{N} \cdot \exp (\langle \bm{q}_-^{(s)}, \bm{k}_{n, j}^{(s)} \rangle),
\end{align*}

\begin{align*}
    \beta_{n, +, i}^{(s)} \le \frac{\eta C_9 \Vert \bm{\mu} \Vert_2^2 \log \big( O(\frac{1}{\epsilon}) \big)}{N \exp  (\langle \bm{q}_{n, i}^{(s)}, \bm{k}_+^{(s)} \rangle)},
\end{align*}

\begin{align*}
    \beta_{n, -, i}^{(s)} \le \frac{\eta C_9 \Vert \bm{\mu} \Vert_2^2 \log \big( O(\frac{1}{\epsilon}) \big)}{N \exp  (\langle \bm{q}_{n, i}^{(s)}, \bm{k}_-^{(s)} \rangle)},
\end{align*}

\begin{align*}
    \alpha_{n, i, +}^{(s)} \le \frac{\eta C_9 \sigma_p^2 d \log \big( O(\frac{1}{\epsilon}) \big)}{N \exp  (\langle \bm{q}_{n, i}^{(s)}, \bm{k}_+^{(s)} \rangle)},
\end{align*}

\begin{align*}
    \alpha_{n, i, -}^{(s)} \le \frac{\eta C_9 \sigma_p^2 d \log \big( O(\frac{1}{\epsilon}) \big)}{N \exp  (\langle \bm{q}_{n, i}^{(s)}, \bm{k}_-^{(s)} \rangle)},
\end{align*}

\begin{align*}
    \beta_{n, i, +}^{(s)} \ge - \frac{\eta C_9 \sigma_p^2 d \log \big( O(\frac{1}{\epsilon}) \big)}{N} \exp (\langle \bm{q}_+^{(s)}, \bm{k}_{n, i}^{(s)} \rangle),
\end{align*}

\begin{align*}
    \beta_{n, i, -}^{(s)} \ge - \frac{\eta C_9 \sigma_p^2 d \log \big( O(\frac{1}{\epsilon}) \big)}{N} \exp (\langle \bm{q}_-^{(s)}, \bm{k}_{n, i}^{(s)} \rangle),
\end{align*}

\begin{align*}
    \alpha_{n, i, n, j}^{(s)} \ge - \frac{\eta C_9 \sigma_p^2 d \log \big( O(\frac{1}{\epsilon}) \big)}{N} \exp (\langle \bm{q}_{n, i}^{(s)}, \bm{k}_{n, j}^{(s)} \rangle),
\end{align*}

\begin{align*}
    \beta_{n, j, n, i}^{(s)} \ge - \frac{\eta C_9 \sigma_p^2 d \log \big( O(\frac{1}{\epsilon}) \big)}{N} \exp (\langle \bm{q}_{n, i}^{(s)}, \bm{k}_{n, j}^{(s)} \rangle).
\end{align*}

Similar to \ref{lower_bound_of_qk}, we apply the bounds of \(\alpha\) and \(\beta\) above to give the upper bounds for the dynamics \(\langle \bm{q}, \bm{k} \rangle\).

\begin{equation}
\begin{split}
    &\langle \bm{q}_+^{(s+1)}, \bm{k}_+^{(s+1)} \rangle - \langle \bm{q}_+^{(s)}, \bm{k}_+^{(s)} \rangle \\
    &= \alpha_{+, +}^{(s)} \Vert \bm{k}_+^{(s)} \Vert_2^2 + \sum\limits_{n \in S_+} \sum\limits_{i=2}^M \alpha_{n, +, i}^{(s)} \langle \bm{k}_+^{(s)}, \bm{k}_{n, i}^{(s)} \rangle \\
    &+ \beta_{+, +}^{(s)} \Vert \bm{q}_+^{(s)} \Vert_2^2 + \sum\limits_{n \in S_+} \sum\limits_{i=2}^M \beta_{n, +, i}^{(s)} \langle \bm{q}_+^{(s)}, \bm{q}_{n, i}^{(s)} \rangle \\
    &+ \Big( \alpha_{+, +}^{(s)} \bm{k}_+^{(s)} + \sum\limits_{n \in S_+} \sum\limits_{i=2}^M \alpha_{n, +, i}^{(s)} \bm{k}_{n, i}^{(s)} \Big) \\
    &\cdot \Big( \beta_{+, +}^{(s)} \bm{q}_+^{(s)\top} + \sum\limits_{n \in S_+} \sum\limits_{i=2}^M \beta_{n, +, i}^{(s)} \bm{q}_{n, i}^{(s)\top} \Big) \\
    &= \alpha_{+, +}^{(s)} \Vert \bm{k}_+^{(s)} \Vert_2^2 + \beta_{+, +}^{(s)} \Vert \bm{q}_+^{(s)} \Vert_2^2 + \{ lower \ order \ term \} \\
    &\le \frac{2 \eta C_9 \Vert \bm{\mu} \Vert_2^2 \log \big( O(\frac{1}{\epsilon}) \big) }{\exp ( \langle \bm{q}_+^{(s)}, \bm{k}_+^{(s)} \rangle)} \cdot \Theta (\Vert \bm{\mu} \Vert_2^2 \sigma_h^2 d_h) + \{ lower \ order \ term \} \\
    &\le \frac{\eta C_{10} \Vert \bm{\mu} \Vert_2^4 \sigma_h^2 d_h \log \big( O(\frac{1}{\epsilon}) \big)}{\exp ( \langle \bm{q}_+^{(s)}, \bm{k}_+^{(s)} \rangle)},
\end{split}
\end{equation}
similarly, we have
\begin{equation}
\begin{split}
    &\langle \bm{q}_-^{(s+1)}, \bm{k}_-^{(s+1)} \rangle - \langle \bm{q}_-^{(s)}, \bm{k}_-^{(s)} \rangle \le \frac{\eta C_{10} \Vert \bm{\mu} \Vert_2^4 \sigma_h^2 d_h \log \big( O(\frac{1}{\epsilon}) \big)}{\exp ( \langle \bm{q}_-^{(s)}, \bm{k}_-^{(s)} \rangle)}.
\end{split}
\end{equation}

\begin{equation}
\begin{split}
    &\langle \bm{q}_+^{(s+1)}, \bm{k}_{n, j}^{(s+1)} \rangle - \langle \bm{q}_+^{(s)}, \bm{k}_{n, j}^{(s)} \rangle \\
    &= \alpha_{+, +}^{(s)} \langle \bm{k}_+^{(s)}, \bm{k}_{n, j}^{(s)} \rangle + \sum\limits_{n^\prime \in S_+} \sum\limits_{l=2}^M \alpha_{n^\prime, +, l}^{(s)} \langle \bm{k}_{n, j}^{(s)}, \bm{k}_{n^\prime, l}^{(s)} \rangle \\
    &+ \beta_{n, j, +}^{(s)} \Vert \bm{q}_+^{(s)} \Vert_2^2 + \beta_{n, j, -}^{(s)} \langle \bm{q}_+^{(s)}, \bm{q}_-^{(s)} \rangle + \sum\limits_{n^\prime = 1}^N \sum\limits_{l=2}^M \beta_{n, j, n^\prime, l}^{(s)} \langle \bm{q}_+^{(s)}, \bm{q}_{n^\prime, l}^{(s)} \rangle \\
    &+ \Big( \alpha_{+, +}^{(s)} \bm{k}_+^{(s)} + \sum\limits_{n^\prime \in S_+} \sum\limits_{l=2}^M \alpha_{n^\prime, +, l}^{(s)} \bm{k}_{n^\prime, l}^{(s)} \Big) \\
    &\cdot \Big( \beta_{n, j, +}^{(s)} \bm{q}_+^{(s)\top} + \beta_{n, j, -}^{(s)} \bm{q}_-^{(s)\top} + \sum\limits_{n^\prime = 1}^N \sum\limits_{l=2}^M \beta_{n, j, n^\prime, l}^{(s)} \bm{q}_{n^\prime, l}^{(s)\top} \Big) \\
    &= \alpha_{n, +, j}^{(s)} \Vert \bm{k}_{n, j}^{(s)} \Vert_2^2 + \beta_{n, j, +}^{(s)} \Vert \bm{q}_+^{(s)} \Vert_2^2 + \{ lower \ order \ term \} \\
    &\ge - \frac{\eta C_9 \Vert \bm{\mu} \Vert_2^2 \log \big( O(\frac{1}{\epsilon}) \big)}{N} \cdot \exp (\langle \bm{q}_+^{(s)}, \bm{k}_{n, j}^{(s)} \rangle) \cdot \Theta \Big( \sigma_p^2 \sigma_h^2 d d_h \Big) \\
    &- \frac{\eta C_9 \sigma_p^2 d \log \big( O(\frac{1}{\epsilon}) \big)}{N} \exp (\langle \bm{q}_+^{(s)}, \bm{k}_{n, j}^{(s)} \rangle) \cdot \Theta \Big( \Vert \bm{\mu} \Vert_2^2 \sigma_h^2 d_h \Big) \\
    &+ \{ lower \ order \ term \} \\
    &\ge - \frac{\eta C_{10} \sigma_p^2 d \Vert \bm{\mu} \Vert_2^2 \sigma_h^2 d_h \log \big( O(\frac{1}{\epsilon}) \big)}{N} \cdot \exp (\langle \bm{q}_+^{(s)}, \bm{k}_{n, j}^{(s)} \rangle),
\end{split}
\end{equation}
similarly, we have
\begin{equation}
\begin{split}
    &\langle \bm{q}_-^{(s+1)}, \bm{k}_{n, j}^{(s+1)} \rangle - \langle \bm{q}_-^{(s)}, \bm{k}_{n, j}^{(s)} \rangle \ge - \frac{\eta C_{10} \sigma_p^2 d \Vert \bm{\mu} \Vert_2^2 \sigma_h^2 d_h \log \big( O(\frac{1}{\epsilon}) \big)}{N} \cdot \exp (\langle \bm{q}_-^{(s)}, \bm{k}_{n, j}^{(s)} \rangle).
\end{split}
\end{equation}

\begin{equation}
\begin{split}
    &\langle \bm{q}_{n, i}^{(s+1)}, \bm{k}_+^{(s+1)} \rangle - \langle \bm{q}_{n, i}^{(s)}, \bm{k}_+^{(s)} \rangle \\
    &= \alpha_{n, i, +}^{(s)} \Vert \bm{k}_+^{(s)} \Vert_2^2 + \alpha_{n, i, -}^{(s)} \langle \bm{k}_+^{(s)}, \bm{k}_-^{(s)} \rangle + \sum\limits_{n^\prime =1}^N \sum\limits_{l=2}^M \alpha_{n, i, n^\prime, l}^{(s)} \langle \bm{k}_+^{(s)}, \bm{k}_{n^\prime, l}^{(s)} \rangle \\
    &+ \beta_{+, +}^{(s)} \langle \bm{q}_+^{(s)}, \bm{q}_{n, i}^{(s)} \rangle + \sum\limits_{n^\prime \in S_+} \sum\limits_{l=2}^M \beta_{n^\prime, +, l}^{(s)} \langle \bm{q}_{n, i}^{(s)}, \bm{q}_{n^\prime, l}^{(s)} \rangle \\
    &+ \Big( \alpha_{n, i, +}^{(s)} \bm{k}_+^{(s)} + \alpha_{n, i, -}^{(s)} \bm{k}_-^{(s)} + \sum\limits_{n^\prime =1}^N \sum\limits_{l=2}^M \alpha_{n, i, n^\prime, l}^{(s)} \bm{k}_{n^\prime, l}^{(s)} \Big) \\
    &\cdot \Big( \beta_{+, +}^{(s)} \bm{q}_+^{(s)\top} + \sum\limits_{n^\prime \in S_+} \sum\limits_{l=2}^M \beta_{n^\prime, +, l}^{(s)} \bm{q}_{n^\prime, l}^{(s)\top} \Big) \\
    &= \alpha_{n, i, +}^{(s)} \Vert \bm{k}_+^{(s)} \Vert_2^2 + \beta_{n, +, i}^{(s)} \Vert \bm{q}_{n, i}^{(s)} \Vert_2^2 + \{ lower \ order \ term \} \\
    &\le \frac{\eta C_9 \sigma_p^2 d \log \big( O(\frac{1}{\epsilon}) \big)}{N \exp  (\langle \bm{q}_{n, i}^{(s)}, \bm{k}_+^{(s)} \rangle)} \cdot \Theta \Big( \Vert \bm{\mu} \Vert_2^2 \sigma_h^2 d_h \Big) \\
    &+ \frac{\eta C_9 \Vert \bm{\mu} \Vert_2^2 \log \big( O(\frac{1}{\epsilon}) \big)}{N \exp (\langle \bm{q}_{n, i}^{(s)}, \bm{k}_+^{(s)} \rangle)} \cdot \Theta \Big( \sigma_p^2 \sigma_h^2 d d_h \Big) \\
    &+ \{ lower \ order \ term \} \\
    &\le \frac{\eta C_{10} \sigma_p^2 d \Vert \bm{\mu} \Vert_2^2 \sigma_h^2 d_h \log \big( O(\frac{1}{\epsilon}) \big)}{N \exp (\langle \bm{q}_{n, i}^{(s)}, \bm{k}_+^{(s)} \rangle)},
\end{split}
\end{equation}
similarly, we have
\begin{equation}
\begin{split}
    &\langle \bm{q}_{n, i}^{(s+1)}, \bm{k}_-^{(s+1)} \rangle - \langle \bm{q}_{n, i}^{(s)}, \bm{k}_-^{(s)} \rangle \le \frac{\eta C_{10} \sigma_p^2 d \Vert \bm{\mu} \Vert_2^2 \sigma_h^2 d_h \log \big( O(\frac{1}{\epsilon}) \big)}{N \exp (\langle \bm{q}_{n, i}^{(s)}, \bm{k}_-^{(s)} \rangle)}.
\end{split}
\end{equation}

\begin{equation}
\begin{split}
    &\langle \bm{q}_{n, i}^{(s+1)}, \bm{k}_{n, j}^{(s+1)} \rangle - \langle \bm{q}_{n, i}^{(s)}, \bm{k}_{n, j}^{(s)} \rangle \\
    &= \alpha_{n, i, +}^{(s)} \langle \bm{k}_+^{(s)}, \bm{k}_{n, j}^{(s)} \rangle + \alpha_{n, i, -}^{(s)} \langle \bm{k}_-^{(s)}, \bm{k}_{n, j}^{(s)} \rangle + \sum\limits_{n^\prime =1}^N \sum\limits_{l=2}^M \alpha_{n, i, n^\prime, l}^{(s)} \langle \bm{k}_{n^\prime, l}^{(s)}, \bm{k}_{n, j}^{(s)} \rangle \\
    &+ \beta_{n, j, +}^{(s)} \langle \bm{q}_+^{(s)}, \bm{q}_{n, i}^{(s)} \rangle + \beta_{n, j, -}^{(s)} \langle \bm{q}_-^{(s)}, \bm{q}_{n, i}^{(s)} \rangle + \sum\limits_{n^\prime = 1}^N \sum\limits_{l=2}^M \beta_{n, j, n^\prime, l}^{(s)} \langle \bm{q}_{n^\prime, l}^{(s)}, \bm{q}_{n, i}^{(s)} \rangle \\
    &+ \Big( \alpha_{n, i, +}^{(s)} \bm{k}_+^{(s)} + \alpha_{n, i, -}^{(s)} \bm{k}_-^{(s)} + \sum\limits_{n^\prime =1}^N \sum\limits_{l=2}^M \alpha_{n, i, n^\prime, l}^{(s)} \bm{k}_{n^\prime, l}^{(s)} \Big) \\
    &\cdot \Big( \beta_{n, j, +}^{(s)} \bm{q}_+^{(s)\top} + \beta_{n, j, -}^{(s)} \bm{q}_-^{(s)\top} + \sum\limits_{n^\prime = 1}^N \sum\limits_{l=2}^M \beta_{n, j, n^\prime, l}^{(s)} \bm{q}_{n^\prime, l}^{(s)\top} \Big) \\
    &= \alpha_{n, i, n, j}^{(s)} \Vert \bm{k}_{n, j}^{(s)} \Vert_2^2 + \beta_{n, j, n, i}^{(s)} \Vert \bm{q}_{n, i}^{(s)} \Vert_2^2 \\
    &+ \{ lower \ order \ term \} \\
    &\ge - \frac{2\eta C_9 \sigma_p^2 d \log \big( O(\frac{1}{\epsilon}) \big)}{N} \exp (\langle \bm{q}_{n, i}^{(s)}, \bm{k}_{n, j}^{(s)} \rangle) \cdot \Theta \Big( \sigma_p^2 \sigma_h^2 d d_h \Big) \\
    &+ \{ lower \ order \ term \} \\
    &\ge - \frac{\eta C_{10} \sigma_p^4 d^2 \sigma_h^2 d_h \log \big( O(\frac{1}{\epsilon}) \big)}{N} \cdot \exp (\langle \bm{q}_{n, i}^{(s)}, \bm{k}_{n, j}^{(s)} \rangle).
\end{split}
\end{equation}

\subsection{Bounds for the Sum of \(\alpha\) and \(\beta\)}
\label{sum_alpha_beta2}
Assume that the propositions \(\mathcal{F}(T_2), \dots, \mathcal{F}(s)\), \(\mathcal{H}(T_2), \dots, \mathcal{H}(s - 1)\) hold (\(s \in [T_1, t]\)), we have
\begin{equation}
    \label{V_p_upper_bound}
    \vert V_\pm^{(s)} \vert \le 2 \log \big( O(\frac{1}{\epsilon}) \big),
\end{equation}
\begin{equation}
    \label{V_i_upper_bound}
    \vert V_{n, i}^{(s)} \vert = O(1),
\end{equation}
\begin{equation}
    \label{lambda_pm_nj2}
    \Lambda_{n, \pm, j}^{(s)} \ge \Lambda_{n, \pm, j}^{(T_2)} \ge \log \Big( \exp(\Lambda_{n, \pm, j}^{(T_1)}) + \Theta \Big( \frac{ d_h^{\frac{1}{2}}}{N \big( \log (6N^2M^2 / \delta) \big)^{3} } \Big) \Big),
\end{equation}
\begin{equation}
    \label{lambda_nij_pm2}
    \Lambda_{n, i, \pm, j}^{(s)} \ge \Lambda_{n, i, \pm, j}^{(T_2)} \ge \log \Big( \exp(\Lambda_{n, i, \pm, j}^{(T_1)}) + \Theta \Big( \frac{ \sigma_p^2 d d_h^{\frac{1}{2}} }{N \Vert \bm{\mu} \Vert_2^2 \big( \log (6N^2M^2 / \delta) \big)^{3}} \Big) \Big)
\end{equation}
for \(i, j \in [M] \backslash \{1\}, n \in [N], s \in [T_2, t] \). Similar to \eqref{q_p_k_j_upper_bound_stage3} and \eqref{q_i_k_j_upper_bound_stage3}, we have 
\begin{equation}
\begin{split}
    \label{q_p_k_j_upper_bound_stage3_2}
    \frac{ \exp (\langle \bm{q}_\pm^{(s)}, \bm{k}_{n, j}^{(s)} \rangle)}{ \exp  (\langle \bm{q}_\pm^{(s)}, \bm{k}_\pm^{(s)} \rangle) + \sum\limits_{j^\prime=2}^M  \exp  ( \langle \bm{q}_\pm^{(s)}, \bm{k}_{n, j^\prime}^{(s)} \rangle ) } = O \Big( \frac{N \big( \log (6N^2M^2 / \delta) \big)^{3} }{d_h^{\frac{1}{2}}} \Big)
\end{split}
\end{equation}

\begin{equation}
\begin{split}
    \label{q_i_k_j_upper_bound_stage3_2}
    \frac{ \exp (\langle \bm{q}_{n, i}^{(s)}, \bm{k}_{n, j}^{(s)} \rangle)}{ \exp  (\langle \bm{q}_{n, i}^{(s)}, \bm{k}_+^{(s)} \rangle) + \sum\limits_{j^\prime=2}^M  \exp  ( \langle \bm{q}_{n, i}^{(s)}, \bm{k}_{n, j^\prime}^{(s)} \rangle ) } = O \Big( \frac{N \Vert \bm{\mu} \Vert_2^2 \big( \log (6N^2M^2 / \delta) \big)^{3}}{ \sigma_p^2 d d_h^{\frac{1}{2}} } \Big)
\end{split}
\end{equation}
Plugging \eqref{q_p_k_j_upper_bound_stage3_2},\eqref{q_i_k_j_upper_bound_stage3_2} into the expressions of \(\alpha\), \(\beta\) and letting \(O \Big( \log \big( O(\frac{1}{\epsilon}) \big) \Big)\) be the upper bound for \(\vert V_\pm^{(s)} \vert, \vert V_{n, i}^{(s)} \vert\) we have

\begin{equation}
\begin{split}
    \label{alpha_p_p_upper_bound2}
    &\vert \alpha_{+, +}^{(s)} \vert = \Big\vert \frac{\eta}{NM} \sum\limits_{n \in S_+} - \ell_n^{\prime(s)} \Vert \bm{\mu} \Vert_2^2 \\
    &\cdot \Big( V_+^{(s)} \big( \frac{ \exp (\langle \bm{q}_+^{(s)}, \bm{k}_+^{(s)} \rangle)}{ \exp  (\langle \bm{q}_+^{(s)}, \bm{k}_+^{(s)} \rangle) + \sum\limits_{j=2}^M  \exp  ( \langle \bm{q}_+^{(s)}, \bm{k}_{n, j}^{(s)} \rangle ) } \\
    & - (\frac{ \exp (\langle \bm{q}_+^{(s)}, \bm{k}_+^{(s)} \rangle)}{ \exp  (\langle \bm{q}_+^{(s)}, \bm{k}_+^{(s)} \rangle) + \sum\limits_{j=2}^M  \exp  ( \langle \bm{q}_+^{(s)}, \bm{k}_{n, j}^{(s)} \rangle ) })^2 \big) \\
    & - \sum\limits_{i=2}^M \big( V_{n, i}^{(s)} \cdot \frac{ \exp (\langle \bm{q}_+^{(s)}, \bm{k}_+^{(s)} \rangle)}{ \exp  (\langle \bm{q}_+^{(s)}, \bm{k}_+^{(s)} \rangle) + \sum\limits_{j=2}^M  \exp  ( \langle \bm{q}_+^{(s)}, \bm{k}_{n, j}^{(s)} \rangle ) } \\
    & \cdot \frac{ \exp (\langle \bm{q}_+^{(s)}, \bm{k}_{n, i}^{(s)} \rangle)}{ \exp  (\langle \bm{q}_+^{(s)}, \bm{k}_+^{(s)} \rangle) + \sum\limits_{j=2}^M  \exp  ( \langle \bm{q}_+^{(s)}, \bm{k}_{n, j}^{(s)} \rangle ) } \big) \Big) \Big\vert \\
    &\le \frac{\eta \Vert \bm{\mu} \Vert_2^2}{NM} \cdot \frac{3N}{4} \cdot O \Big( \log \big( O(\frac{1}{\epsilon}) \big) \Big) \cdot O \Big( \frac{N \big( \log (6N^2M^2 / \delta) \big)^{3} }{d_h^{\frac{1}{2}}} \Big) \\
    &= O \Big( \frac{\eta N \Vert \bm{\mu} \Vert_2^2 \big( \log (6N^2M^2 / \delta) \big)^{3} \log \big( O(\frac{1}{\epsilon}) \big) }{d_h^{\frac{1}{2}}} \Big) \\
\end{split}
\end{equation}
Similarly, we have
\begin{equation}
\begin{split}
    \vert \alpha_{-, -}^{(s)} \vert, \vert \beta_{+, +}^{(s)} \vert, \vert \beta_{-, -}^{(s)} \vert = O \Big( \frac{\eta N \Vert \bm{\mu} \Vert_2^2 \big( \log (6N^2M^2 / \delta) \big)^{3} \log \big( O(\frac{1}{\epsilon}) \big) }{d_h^{\frac{1}{2}}} \Big),
\end{split}
\end{equation}
\begin{equation}
\begin{split}
    \vert \alpha_{n, +, i}^{(s)} \vert, \vert \alpha_{n, -, i}^{(s)} \vert = O \Big( \frac{\eta \Vert \bm{\mu} \Vert_2^2 \big( \log (6N^2M^2 / \delta) \big)^{3} \log \big( O(\frac{1}{\epsilon}) \big) }{d_h^{\frac{1}{2}}} \Big),
\end{split}
\end{equation}
\begin{equation}
\begin{split}
    \vert \beta_{n, +, i}^{(s)} \vert, \vert \beta_{n, -, i}^{(s)} \vert &= O \Big( \frac{\eta \Vert \bm{\mu} \Vert_2^4 \big( \log (6N^2M^2 / \delta) \big)^{3} \log \big( O(\frac{1}{\epsilon}) \big) }{ \sigma_p^2 d d_h^{\frac{1}{2}} } \Big) \\
    &= O \Big( \frac{\eta \Vert \bm{\mu} \Vert_2^2 \cdot \mathrm{SNR}^2 \big( \log (6N^2M^2 / \delta) \big)^{3} \log \big( O(\frac{1}{\epsilon}) \big) }{ d_h^{\frac{1}{2}} } \Big),
\end{split}
\end{equation}
for \(i \in [M] \backslash \{1\}, n \in S_\pm\).
\begin{equation}
\begin{split}
    \vert \alpha_{n, i, +}^{(s)} \vert, \vert \alpha_{n, i, -}^{(s)} \vert = O \Big( \frac{\eta \Vert \bm{\mu} \Vert_2^2 \big( \log (6N^2M^2 / \delta) \big)^{3} \log \big( O(\frac{1}{\epsilon}) \big) }{ d_h^{\frac{1}{2}} } \Big),
\end{split}
\end{equation}
for \(i \in [M] \backslash \{1\}, n \in S_\pm\).
\begin{equation}
\begin{split}
    \vert \beta_{n, i, +}^{(s)} \vert, \vert \beta_{n, i, -}^{(s)} \vert &= O \Big( \frac{\eta \sigma_p^2 d \big( \log (6N^2M^2 / \delta) \big)^{3} \log \big( O(\frac{1}{\epsilon}) \big) }{d_h^{\frac{1}{2}}} \Big) \\
    &= O \Big( \frac{\eta N \Vert \bm{\mu} \Vert_2^2 \big( \log (6N^2M^2 / \delta) \big)^{3} \log \big( O(\frac{1}{\epsilon}) \big) }{d_h^{\frac{1}{2}}} \Big),
\end{split}
\end{equation}
for \(i \in [M] \backslash \{1\}, n \in S_\pm\), the last equality is by \(N \cdot \mathrm{SNR}^2 \ge \Omega(1)\).
\begin{equation}
\begin{split}
    \vert \alpha_{n, i, n, j}^{(s)} \vert, \vert \beta_{n, j, n, i}^{(s)} \vert = O \Big( \frac{\eta \Vert \bm{\mu} \Vert_2^2 \big( \log (6N^2M^2 / \delta) \big)^{3} \log \big( O(\frac{1}{\epsilon}) \big) }{ d_h^{\frac{1}{2}} } \Big),
\end{split}
\end{equation}
for \(i, j \in [M] \backslash \{1\}, n \in [N]\).
\begin{equation}
\begin{split}
    \vert \alpha_{n, i, n^\prime, j}^{(s)} \vert, \vert \beta_{n, j, n^\prime, i}^{(s)} \vert = O \Big( \frac{\eta \Vert \bm{\mu} \Vert_2^2 \big( \log (6N^2M^2 / \delta) \big)^{4} \log \big( O(\frac{1}{\epsilon}) \big) }{ d^{\frac{1}{2}} d_h^{\frac{1}{2}} } \Big),
\end{split}
\end{equation}
for \(i, j \in [M] \backslash \{1\}, n, n^\prime \in [N], n \ne n^\prime\). Taking a summation we obtain that

\begin{equation}
\begin{split}
    \sum\limits_{s = T_2}^{t}\vert \alpha_{+, +}^{(s)} \vert &= O \Big( \frac{1}{\eta \epsilon \Vert \bm{\mu} \Vert_2^2 \Vert \bm{w}_O \Vert_2^2 } \Big) \cdot O \Big( \frac{\eta \Vert \bm{\mu} \Vert_2^2 N \big( \log (6N^2M^2 / \delta) \big)^{3} \log \big( O(\frac{1}{\epsilon}) \big) }{d_h^{\frac{1}{2}}} \Big) \\
    &= O \Big( \frac{ N \big( \log (6N^2M^2 / \delta) \big)^{3} \log \big( O(\frac{1}{\epsilon}) \big) }{\epsilon d_h^{\frac{1}{2}}} \Big),
\end{split}
\end{equation}
where the last equality is by \(\Vert \bm{w}_O \Vert = \Theta (1)\). Similarly, we have
\begin{equation}
\begin{split}
    &\sum\limits_{s = T_2}^{t}\vert \beta_{n, +, i}^{(s)} \vert, \sum\limits_{s = T_2}^{t}\vert \beta_{n, -, i}^{(s)} \vert = O \Big( \frac{\mathrm{SNR}^2 \big( \log (6N^2M^2 / \delta) \big)^{3} \log \big( O(\frac{1}{\epsilon}) \big) }{ \epsilon d_h^{\frac{1}{2}} } \Big),
\end{split}
\end{equation}
for \(i \in [M] \backslash \{1\}, n \in S_\pm\).
\begin{equation}
\begin{split}
    &\sum\limits_{s = T_2}^{t}\vert \alpha_{-, -}^{(s)} \vert, \sum\limits_{s = T_2}^{t}\vert \beta_{+, +}^{(s)} \vert, \sum\limits_{s = T_2}^{t}\vert \beta_{-, -}^{(s)} \vert, \sum\limits_{s = T_2}^{t}\vert \beta_{n, +, i}^{(s)} \vert, \sum\limits_{s = T_2}^{t}\vert \beta_{n, -, i}^{(s)} \vert, \sum\limits_{s = T_2}^{t}\vert \beta_{n, i, +}^{(s)} \vert, \sum\limits_{s = T_2}^{t}\vert \beta_{n, i, -}^{(s)} \vert \\
    &= O \Big( \frac{ N \big( \log (6N^2M^2 / \delta) \big)^{3} \log \big( O(\frac{1}{\epsilon}) \big) }{\epsilon d_h^{\frac{1}{2}}} \Big),
\end{split}
\end{equation}
for \(i \in [M] \backslash \{1\}, n \in S_\pm\).
\begin{equation}
\begin{split}
    &\sum\limits_{s = T_2}^{t}\vert \alpha_{n, +, i}^{(s)} \vert, \sum\limits_{s = T_2}^{t}\vert \alpha_{n, -, i}^{(s)} \vert, \sum\limits_{s = T_2}^{t}\vert \alpha_{n, i, +}^{(s)} \vert, \sum\limits_{s = T_2}^{t}\vert \alpha_{n, i, -}^{(s)} \vert, \sum\limits_{s = T_2}^{t}\vert \alpha_{n, i, n, j}^{(s)} \vert, \sum\limits_{s = T_2}^{t}\vert \beta_{n, j, n, i}^{(s)} \vert \\
    &= O \Big( \frac{ \big( \log (6N^2M^2 / \delta) \big)^{3} \log \big( O(\frac{1}{\epsilon}) \big) }{\epsilon d_h^{\frac{1}{2}}} \Big),
\end{split}
\end{equation}
for \(i, j \in [M] \backslash \{1\}, n \in S_\pm\).
\begin{equation}
\begin{split}
    \sum\limits_{s = T_2}^{t}\vert \alpha_{n, i, n^\prime, j}^{(s)} \vert, \sum\limits_{s = T_2}^{t}\vert \beta_{n, j, n^\prime, i}^{(s)} \vert = O \Big( \frac{ \big( \log (6N^2M^2 / \delta) \big)^{4} \log \big( O(\frac{1}{\epsilon}) \big) }{\epsilon d^{\frac{1}{2}} d_h^{\frac{1}{2}}} \Big)
\end{split}
\end{equation}
for \(i, j \in [M] \backslash \{1\}, n, n^\prime \in [N], n \ne n^\prime\).

With these sums of \(\alpha\) and \(\beta\) above, we can easily prove \textbf{Claim} \ref{claim7} and \textbf{Claim} \ref{claim8}.

\subsection{Proof of Claim \ref{claim7}}
\label{proof_claim7}
In this subsection, we assume that \( \mathcal{I}(T_2), \dots, \mathcal{I}(t) \) hold, and then proof that \( \mathcal{G}(t+1) \) is true with the result of \ref{sum_alpha_beta2}.

\begin{align*}
    &\Big\vert \Vert \bm{q}_+^{(t+1)} \Vert_2^2 - \Vert \bm{q}_+^{(t+1)} \Vert_2^2 \Big\vert \le \sum\limits_{s = T_2}^{t} \Big\vert \Vert \bm{q}_+^{(s+1)} \Vert_2^2 - \Vert \bm{q}_+^{(s)} \Vert_2^2 \Big\vert \\
    &\le \sum\limits_{s = T_2}^{t} \Big\vert 2 \alpha_{+, +}^{(s)} \langle \bm{q}_+^{(s)}, \bm{k}_+^{(s)} \rangle + 2 \sum\limits_{n \in S_+} \sum\limits_{i=2}^M \alpha_{n, +, i}^{(s)} \langle \bm{q}_+^{(s)}, \bm{k}_{n, i}^{(s)} \rangle \\
    &+ \Big( \alpha_{+, +}^{(s)} \bm{k}_+^{(s)} + \sum\limits_{n \in S_+} \sum\limits_{i=2}^M \alpha_{n, +, i}^{(s)} \bm{k}_{n, i}^{(s)} \Big) \\
    &\cdot \Big( \alpha_{+, +}^{(s)} \bm{k}_+^{(s)\top} + \sum\limits_{n \in S_+} \sum\limits_{i=2}^M \alpha_{n, +, i}^{(s)} \bm{k}_{n, i}^{(s)\top} \Big) \Big\vert \\
    &\le 2 \sum\limits_{s = T_2}^{t} \vert \alpha_{+, +}^{(s)} \vert \vert \langle \bm{q}_+^{(s)}, \bm{k}_+^{(s)} \rangle \vert + 2 \sum\limits_{n \in S_+} \sum\limits_{i=2}^M \sum\limits_{s = T_2}^{t} \vert \alpha_{n, +, i}^{(s)} \vert \vert \langle \bm{q}_+^{(s)}, \bm{k}_{n, i}^{(s)} \rangle \vert \\
    &+ \{lower \ order \ term\} \\
    &= O \Big( \frac{ N \big( \log (6N^2M^2 / \delta) \big)^{3} \log \big( O(\frac{1}{\epsilon}) \big) }{\epsilon d_h^{\frac{1}{2}}} \Big) \cdot \log ( \epsilon^{-1} d_h^{\frac{1}{2}} ) \\
    &+ N \cdot M \cdot O \Big( \frac{ \big( \log (6N^2M^2 / \delta) \big)^{3} \log \big( O(\frac{1}{\epsilon}) \big) }{\epsilon d_h^{\frac{1}{2}}} \Big) \cdot \log ( \epsilon^{-1} d_h^{\frac{1}{2}} ) \\
    &= O \Big( \frac{ N \big( \log (6N^2M^2 / \delta) \big)^{3} \log \big( O(\frac{1}{\epsilon}) \big) \log ( \epsilon^{-1} d_h^{\frac{1}{2}} ) }{\epsilon d_h^{\frac{1}{2}}} \Big) \\
\end{align*}
where the first inequality is by triangle inequality, the second inequality is by \ref{q_p_sq_dy}, the third inequality is by \(t \le T_3\).
Since \(\sigma_h^2 \ge \big( \max\{\sigma_p^2 d, \Vert \mu \Vert_2^2\} \big)^{-1} \cdot d_h^{-\frac{1}{2}} (\log (6N^2M^2 / \delta))^{-2}\) and \(d_h = \widetilde{\Omega} \Big( \max \{\mathrm{SNR}^4, \mathrm{SNR}^{-4}\} N^2 \epsilon^{-2} \Big)\), we have \(\frac{ N \big( \log (6N^2M^2 / \delta) \big)^{3} \log \big( O(\frac{1}{\epsilon}) \big) \log ( \epsilon^{-1} d_h^{\frac{1}{2}} ) }{\epsilon d_h^{\frac{1}{2}}} = o (\Vert \bm{\mu} \Vert_2^2 \sigma_h^2 d_h)\), so \(\Vert \bm{q}_+^{(t+1)} \Vert_2^2 = \Vert \bm{q}_+^{(T_2)} \Vert_2^2 + o (\Vert \bm{\mu} \Vert_2^2 \sigma_h^2 d_h) = \Theta (\Vert \bm{\mu} \Vert_2^2 \sigma_h^2 d_h)\).
Similarly, we have
\[
    \Big\vert \Vert \bm{q}_-^{(t+1)} \Vert_2^2 - \Vert \bm{q}_-^{(t+1)} \Vert_2^2 \Big\vert = O \Big( \frac{ N \big( \log (6N^2M^2 / \delta) \big)^{3} \log \big( O(\frac{1}{\epsilon}) \big) \log ( \epsilon^{-1} d_h^{\frac{1}{2}} ) }{\epsilon d_h^{\frac{1}{2}}} \Big) = o (\Vert \bm{\mu} \Vert_2^2 \sigma_h^2 d_h),
\]
\[
    \Big\vert \Vert \bm{k}_\pm^{(t+1)} \Vert_2^2 - \Vert \bm{k}_\pm^{(t+1)} \Vert_2^2 \Big\vert = O \Big( \frac{ \big( 1 + SNR^2 \big) N \big( \log (6N^2M^2 / \delta) \big)^{3} \log \big( O(\frac{1}{\epsilon}) \big) \log ( \epsilon^{-1} d_h^{\frac{1}{2}} ) }{\epsilon d_h^{\frac{1}{2}}} \Big) = o (\Vert \bm{\mu} \Vert_2^2 \sigma_h^2 d_h),
\]
\[
    \Big\vert \Vert \bm{q}_{n, i}^{(t+1)} \Vert_2^2 - \Vert \bm{q}_{n, i}^{(t+1)} \Vert_2^2 \Big\vert = O \Big( \frac{ \big( \log (6N^2M^2 / \delta) \big)^{3} \log \big( O(\frac{1}{\epsilon}) \big) \log ( \epsilon^{-1} d_h^{\frac{1}{2}} ) }{\epsilon d_h^{\frac{1}{2}}} \Big) = o (\sigma_p^2 \sigma_h^2 d d_h),
\]
\[
    \Big\vert \Vert \bm{k}_{n, i}^{(t+1)} \Vert_2^2 - \Vert \bm{k}_{n, i}^{(t+1)} \Vert_2^2 \Big\vert = O \Big( \frac{ N \big( \log (6N^2M^2 / \delta) \big)^{3} \log \big( O(\frac{1}{\epsilon}) \big) \log ( \epsilon^{-1} d_h^{\frac{1}{2}} ) }{\epsilon d_h^{\frac{1}{2}}} \Big) = o (\sigma_p^2 \sigma_h^2 d d_h),
\]
so we have
\[
    \Vert \bm{q}_\pm^{(t+1)} \Vert_2^2, \Vert \bm{k}_\pm^{(t+1)} \Vert_2^2 = \Theta (\Vert \bm{\mu} \Vert_2^2 \sigma_h^2 d_h),
\]
\[
    \Vert \bm{q}_{n, i}^{(t+1)} \Vert_2^2, \Vert \bm{k}_{n, i}^{(t+1)} \Vert_2^2 = \Theta ( \sigma_p^2 \sigma_h^2 d d_h )
\]
for \(i \in [M] \backslash \{1\}, n \in [N]\).

\begin{align*}
    &\vert \langle \bm{q}_+^{(t+1)}, \bm{q}_-^{(t+1)} \rangle \vert \le \vert \langle \bm{q}_+^{(T_2)}, \bm{q}_-^{(T_2)} \rangle \vert + \sum\limits_{s = T_2}^{t} \Big\vert \langle \bm{q}_+^{(s+1)}, \bm{q}_-^{(s+1)} \rangle - \langle \bm{q}_+^{(s)}, \bm{q}_-^{(s)} \rangle \Big\vert \\
    &\le \vert \langle \bm{q}_+^{(T_2)}, \bm{q}_-^{(T_2)} \rangle \vert \\
    &+ \sum\limits_{s = T_2}^{t} \Big\vert \alpha_{+, +}^{(s)} \langle \bm{q}_-^{(s)}, \bm{k}_+^{(s)} \rangle + \sum\limits_{n \in S_+} \sum\limits_{i=2}^M \alpha_{n, +, i}^{(s)} \langle \bm{q}_-^{(s)}, \bm{k}_{n, i}^{(s)} \rangle \\
    &+ \alpha_{-, -}^{(s)} \langle \bm{q}_+^{(s)}, \bm{k}_-^{(s)} \rangle + \sum\limits_{n \in S_-} \sum\limits_{i=2}^M \alpha_{n, -, i}^{(s)} \langle \bm{q}_+^{(s)}, \bm{k}_{n, i}^{(s)} \rangle \\
    &+ \Big( \alpha_{+, +}^{(s)} \bm{k}_+^{(s)} + \sum\limits_{n \in S_+} \sum\limits_{i=2}^M \alpha_{n, +, i}^{(s)} \bm{k}_{n, i}^{(s)} \Big) \\
    &\cdot \Big( \alpha_{-, -}^{(s)} \bm{k}_-^{(s)\top} + \sum\limits_{n \in S_-} \sum\limits_{i=2}^M \alpha_{n, -, i}^{(s)} \bm{k}_{n, i}^{(s)\top} \Big) \Big\vert \\
    &\le \vert \langle \bm{q}_+^{(T_2)}, \bm{q}_-^{(T_2)} \rangle \vert \\
    &+ \sum\limits_{s = T_2}^{t} \vert \alpha_{+, +}^{(s)} \vert \vert \langle \bm{q}_-^{(s)}, \bm{k}_+^{(s)} \rangle \vert + \sum\limits_{n \in S_+} \sum\limits_{i=2}^M \sum\limits_{s = T_2}^{t} \vert \alpha_{n, +, i}^{(s)} \vert \vert \langle \bm{q}_-^{(s)}, \bm{k}_{n, i}^{(s)} \rangle \vert \\
    &+ \sum\limits_{s = T_2}^{t} \vert \alpha_{-, -}^{(s)} \vert \vert \langle \bm{q}_+^{(s)}, \bm{k}_-^{(s)} \rangle \vert + \sum\limits_{n \in S_-} \sum\limits_{i=2}^M \sum\limits_{s = T_2}^{t} \vert \alpha_{n, -, i}^{(s)} \vert \vert \langle \bm{q}_+^{(s)}, \bm{k}_{n, i}^{(s)} \rangle \vert \\
    &+ \{lower \ order \ term\} \\
    &\le \vert \langle \bm{q}_+^{(T_2)}, \bm{q}_-^{(T_2)} \rangle \vert \\
    &+ O \Big( \frac{ N \big( \log (6N^2M^2 / \delta) \big)^{3} \log \big( O(\frac{1}{\epsilon}) \big) }{\epsilon d_h^{\frac{1}{2}}} \Big) \cdot o(1) \\
    &+ N \cdot M \cdot O \Big( \frac{ \big( \log (6N^2M^2 / \delta) \big)^{3} \log \big( O(\frac{1}{\epsilon}) \big) }{\epsilon d_h^{\frac{1}{2}}} \Big) \cdot \log ( \epsilon^{-1} d_h^{\frac{1}{2}} ) \\
    &= \vert \langle \bm{q}_+^{(T_2)}, \bm{q}_-^{(T_2)} \rangle \vert + O \Big( \frac{ N \big( \log (6N^2M^2 / \delta) \big)^{3} \log \big( O(\frac{1}{\epsilon}) \big) \log ( \epsilon^{-1} d_h^{\frac{1}{2}} ) }{\epsilon d_h^{\frac{1}{2}}} \Big) \\
    &= o(1),
\end{align*}
where the first inequality is triangle inequality, the second inequality is by \eqref{q_p_q_m_dy}, the last equality is by \(d_h = \widetilde{\Omega} \Big( \max \{\mathrm{SNR}^4, \mathrm{SNR}^{-4}\} N^2 \epsilon^{-2} \Big)\).
\begin{align*}
    &\vert \langle \bm{q}_+^{(t+1)}, \bm{q}_{n, i}^{(t+1)} \rangle \vert \le \vert \langle \bm{q}_+^{(T_2)}, \bm{q}_{n, i}^{(T_2)} \rangle \vert + \sum\limits_{s = T_2}^{t} \Big\vert \langle \bm{q}_+^{(s+1)}, \bm{q}_{n, i}^{(s+1)} \rangle - \langle \bm{q}_+^{(s)}, \bm{q}_{n, i}^{(s)} \rangle \Big\vert \\
    &\le \vert \langle \bm{q}_+^{(T_2)}, \bm{q}_{n, i}^{(T_2)} \rangle \vert \\
    &+ \sum\limits_{s = T_2}^{t} \Big\vert \alpha_{+, +}^{(s)} \langle \bm{q}_{n, i}^{(s)}, \bm{k}_+^{(s)} \rangle + \sum\limits_{n^\prime \in S_+} \sum\limits_{l=2}^M \alpha_{n^\prime, +, l}^{(s)} \langle \bm{q}_{n, i}^{(s)}, \bm{k}_{n^\prime, l}^{(s)} \rangle \\
    &+ \alpha_{n, i, +}^{(s)} \langle \bm{q}_+^{(s)}, \bm{k}_+^{(s)} \rangle + \alpha_{n, i, -}^{(s)} \langle \bm{q}_+^{(s)}, \bm{k}_-^{(s)} \rangle + \sum\limits_{n^\prime = 1}^N \sum\limits_{l=2}^M \alpha_{n, i, n^\prime, l}^{(s)} \langle \bm{q}_+^{(s)}, \bm{k}_{n^\prime, l}^{(s)} \rangle \\
    &+ \Big( \alpha_{+, +}^{(s)} \bm{k}_+^{(s)} + \sum\limits_{n \in S_+} \sum\limits_{i=2}^M \alpha_{n, +, i}^{(s)} \bm{k}_{n, i}^{(s)} \Big) \\
    &\cdot \Big( \alpha_{n, i, +}^{(s)} \bm{k}_+^{(s)\top} + \alpha_{n, i, -}^{(s)} \bm{k}_-^{(s)\top} + \sum\limits_{n^\prime=1}^N \sum\limits_{l=2}^M \alpha_{n, i, n^\prime, l}^{(s)} \bm{k}_{n^\prime, l}^{(s)\top} \Big) \Big\vert \\
    &\le \vert \langle \bm{q}_+^{(T_2)}, \bm{q}_{n, i}^{(T_2)} \rangle \vert \\
    &+ \sum\limits_{s = T_2}^{t} \vert \alpha_{+, +}^{(s)} \vert \vert \langle \bm{q}_{n, i}^{(s)}, \bm{k}_+^{(s)} \rangle \vert + \sum\limits_{l=2}^M \sum\limits_{s = T_2}^{t} \vert \alpha_{n, +, l}^{(s)} \vert \vert \langle \bm{q}_{n, i}^{(s)}, \bm{k}_{n, l}^{(s)} \rangle \vert + \sum\limits_{n^\prime \in S_+ \wedge n^\prime \ne n} \sum\limits_{l=2}^M \sum\limits_{s = T_2}^{t} \vert \alpha_{n^\prime, +, l}^{(s)} \vert \vert \langle \bm{q}_{n, i}^{(s)}, \bm{k}_{n^\prime, l}^{(s)} \rangle \vert \\
    &+ \sum\limits_{s = T_2}^{t} \vert \alpha_{n, i, +}^{(s)} \vert \vert \langle \bm{q}_+^{(s)}, \bm{k}_+^{(s)} \rangle \vert + \sum\limits_{s = T_2}^{t} \vert \alpha_{n, i, -}^{(s)} \vert \vert \langle \bm{q}_+^{(s)}, \bm{k}_-^{(s)} \rangle \vert + \sum\limits_{l=2}^M \sum\limits_{s = T_2}^{t} \vert \alpha_{n, i, n, l}^{(s)} \vert \vert \langle \bm{q}_+^{(s)}, \bm{k}_{n, l}^{(s)} \rangle \vert \\
    &+ \sum\limits_{n^\prime \ne n} \sum\limits_{l=2}^M \sum\limits_{s = T_2}^{t} \vert \alpha_{n, i, n^\prime, l}^{(s)} \vert \vert \langle \bm{q}_+^{(s)}, \bm{k}_{n^\prime, l}^{(s)} \rangle \vert \\
    &+ \{lower \ order \ term\} \\
    &\le \vert \langle \bm{q}_+^{(T_2)}, \bm{q}_{n, i}^{(T_2)} \rangle \vert \\
    &+ O \Big( \frac{ N \big( \log (6N^2M^2 / \delta) \big)^{3} \log \big( O(\frac{1}{\epsilon}) \big) }{\epsilon d_h^{\frac{1}{2}}} \Big) \cdot \log ( \epsilon^{-1} d_h^{\frac{1}{2}} ) + M \cdot O \Big( \frac{ \big( \log (6N^2M^2 / \delta) \big)^{3} \log \big( O(\frac{1}{\epsilon}) \big) }{\epsilon d_h^{\frac{1}{2}}} \Big) \cdot \log ( \epsilon^{-1} d_h^{\frac{1}{2}} ) \\
    &+ N \cdot M \cdot O \Big( \frac{ \big( \log (6N^2M^2 / \delta) \big)^{3} \log \big( O(\frac{1}{\epsilon}) \big) }{\epsilon d_h^{\frac{1}{2}}} \Big) \cdot o(1) \\
    &+ M \cdot O \Big( \frac{ \big( \log (6N^2M^2 / \delta) \big)^{3} \log \big( O(\frac{1}{\epsilon}) \big) }{\epsilon d_h^{\frac{1}{2}}} \Big) \cdot \log ( \epsilon^{-1} d_h^{\frac{1}{2}} ) \\
    &+ N \cdot M \cdot O \Big( \frac{ \big( \log (6N^2M^2 / \delta) \big)^{4} \log \big( O(\frac{1}{\epsilon}) \big) }{\epsilon d^{\frac{1}{2}} d_h^{\frac{1}{2}}} \Big) \cdot \log ( \epsilon^{-1} d_h^{\frac{1}{2}} ) \\
    &= \vert \langle \bm{q}_+^{(T_2)}, \bm{q}_{n, i}^{(T_2)} \rangle \vert + O \Big( \frac{ N \big( \log (6N^2M^2 / \delta) \big)^{3} \log \big( O(\frac{1}{\epsilon}) \big) \log ( \epsilon^{-1} d_h^{\frac{1}{2}} ) }{\epsilon d_h^{\frac{1}{2}}} \Big) \\
    &+ O \Big( \frac{ N \big( \log (6N^2M^2 / \delta) \big)^{4} \log \big( O(\frac{1}{\epsilon}) \big) \log ( \epsilon^{-1} d_h^{\frac{1}{2}} ) }{\epsilon d^{\frac{1}{2}} d_h^{\frac{1}{2}}} \Big) \\
    &= o(1),
\end{align*}

where the first inequality is triangle inequality, the second inequality is by \eqref{q_p_q_i_dy}, the last equality is by \(d_h = \widetilde{\Omega} \Big( \max \{\mathrm{SNR}^4, \mathrm{SNR}^{-4}\} N^2 \epsilon^{-2} \Big)\) and \(d = \widetilde{\Omega} \Big( \epsilon^{-2} N^2 d_h \Big) \). Similarly, we have \(\vert \langle \bm{q}_-^{(t+1)}, \bm{q}_{n, i}^{(t+1)} \rangle \vert = o(1)\).
\begin{align*}
    &\vert \langle \bm{q}_{n, i}^{(t+1)}, \bm{q}_{n, j}^{(t+1)} \rangle \vert \le \vert \langle \bm{q}_{n, i}^{(T_2)}, \bm{q}_{n, j}^{(T_2)} \rangle \vert + \sum\limits_{s = T_2}^{t} \Big\vert \langle \bm{q}_{n, i}^{(s+1)}, \bm{q}_{n, j}^{(s+1)} \rangle - \langle \bm{q}_{n, i}^{(s)}, \bm{q}_{n, j}^{(s)} \rangle \Big\vert \\
    &\le \vert \langle \bm{q}_{n, i}^{(T_2)}, \bm{q}_{n, j}^{(T_2)} \rangle \vert \\
    &+ \sum\limits_{s = T_2}^{t} \Big\vert \alpha_{n, i, +}^{(s)} \langle \bm{q}_{n, j}^{(s)}, \bm{k}_+^{(s)} \rangle + \alpha_{n, i, -}^{(s)} \langle \bm{q}_{n, j}^{(s)}, \bm{k}_-^{(s)} \rangle + \sum\limits_{n^\prime = 1}^N \sum\limits_{l=2}^M \alpha_{n, i, n^\prime, l}^{(s)} \langle \bm{q}_{n, j}^{(s)}, \bm{k}_{n^\prime, l}^{(s)} \rangle \\
    &+ \alpha_{n, j, +}^{(s)} \langle \bm{q}_{n, i}^{(s)}, \bm{k}_+^{(s)} \rangle + \alpha_{n, j, -}^{(s)} \langle \bm{q}_{n, i}^{(s)}, \bm{k}_-^{(s)} \rangle + \sum\limits_{n^\prime = 1}^N \sum\limits_{l=2}^M \alpha_{n, j, n^\prime, l}^{(s)} \langle \bm{q}_{n, i}^{(s)}, \bm{k}_{n^\prime, l}^{(s)} \rangle \\
    &+ \Big( \alpha_{n, i, +}^{(s)} \bm{k}_+^{(s)} + \alpha_{n, i, -}^{(s)} \bm{k}_-^{(s)} + \sum\limits_{n^\prime = 1}^N \sum\limits_{l=2}^M \alpha_{n, i, n^\prime, l}^{(s)} \bm{k}_{n^\prime, l}^{(s)} \Big) \\
    &\cdot \Big( \alpha_{n, j, +}^{(s)} \bm{k}_+^{(s)\top} + \alpha_{n, j, -}^{(s)} \bm{k}_-^{(s)\top} + \sum\limits_{n^\prime = 1}^N \sum\limits_{l=2}^M \alpha_{n, j, n^\prime, l}^{(s)} \bm{k}_{n^\prime, l}^{(s)\top} \Big) \Big\vert \\
    &\le \vert \langle \bm{q}_{n, i}^{(T_2)}, \bm{q}_{n, j}^{(T_2)} \rangle \vert \\
    &+ \sum\limits_{s = T_2}^{t} \vert \alpha_{n, i, +}^{(s)} \vert \vert \langle \bm{q}_{n, j}^{(s)}, \bm{k}_+^{(s)} \rangle \vert + \sum\limits_{s = T_2}^{t} \vert \alpha_{n, i, -}^{(s)} \vert \vert \langle \bm{q}_{n, j}^{(s)}, \bm{k}_-^{(s)} \rangle \vert + \sum\limits_{l=2}^M \sum\limits_{s = T_2}^{t} \vert \alpha_{n, i, n, l}^{(s)} \vert \vert \langle \bm{q}_{n, j}^{(s)}, \bm{k}_{n, l}^{(s)} \rangle \vert \\
    &+ \sum\limits_{n^\prime \ne n} \sum\limits_{l=2}^M \sum\limits_{s = T_2}^{t} \vert \alpha_{n, i, n^\prime, l}^{(s)} \vert \vert \langle \bm{q}_{n, j}^{(s)}, \bm{k}_{n^\prime, l}^{(s)} \rangle \vert + \sum\limits_{s = T_2}^{t} \vert \alpha_{n, j, +}^{(s)} \vert \vert \langle \bm{q}_{n, i}^{(s)}, \bm{k}_+^{(s)} \rangle \vert + \sum\limits_{s = T_2}^{t} \vert \alpha_{n, j, -}^{(s)} \vert \vert \langle \bm{q}_{n, i}^{(s)}, \bm{k}_-^{(s)} \rangle \vert \\
    &+ \sum\limits_{l=2}^M \sum\limits_{s = T_2}^{t} \vert \alpha_{n, j, n, l}^{(s)} \vert \vert \langle \bm{q}_{n, i}^{(s)}, \bm{k}_{n, l}^{(s)} \rangle \vert + \sum\limits_{n^\prime \ne n} \sum\limits_{l=2}^M \sum\limits_{s = T_2}^{t} \vert \alpha_{n, j, n^\prime, l}^{(s)} \vert \vert \langle \bm{q}_{n, i}^{(s)}, \bm{k}_{n^\prime, l}^{(s)} \rangle \vert \\
    &+ \{lower \ order \ term\} \\
    &\le \vert \langle \bm{q}_{n, i}^{(T_2)}, \bm{q}_{n, j}^{(T_2)} \rangle \vert \\
    &+ O \Big( \frac{ \big( \log (6N^2M^2 / \delta) \big)^{3} \log \big( O(\frac{1}{\epsilon}) \big) }{\epsilon d_h^{\frac{1}{2}}} \Big) \cdot \log ( \epsilon^{-1} d_h^{\frac{1}{2}} ) \\
    &+ M \cdot O \Big( \frac{ \big( \log (6N^2M^2 / \delta) \big)^{3} \log \big( O(\frac{1}{\epsilon}) \big) }{\epsilon d_h^{\frac{1}{2}}} \Big) \cdot \log ( \epsilon^{-1} d_h^{\frac{1}{2}} ) \\
    &+ N \cdot M \cdot O \Big( \frac{ \big( \log (6N^2M^2 / \delta) \big)^{4} \log \big( O(\frac{1}{\epsilon}) \big) }{\epsilon d^{\frac{1}{2}} d_h^{\frac{1}{2}}} \Big) \cdot o(1) \\
    &= \vert \langle \bm{q}_{n, i}^{(T_2)}, \bm{q}_{n, j}^{(T_2)} \rangle \vert + O \Big( \frac{ \big( \log (6N^2M^2 / \delta) \big)^{3} \log \big( O(\frac{1}{\epsilon}) \big) \log ( \epsilon^{-1} d_h^{\frac{1}{2}} ) }{\epsilon d_h^{\frac{1}{2}}} \Big) \\
    &+ o \Big( \frac{ N \big( \log (6N^2M^2 / \delta) \big)^{4} \log \big( O(\frac{1}{\epsilon}) \big) }{\epsilon d^{\frac{1}{2}} d_h^{\frac{1}{2}}} \Big) \\
    &= o(1)
\end{align*}
for \(i, j \in [M] \backslash \{1\}, i \ne j, n \in [N]\). The first inequality is triangle inequality, the second inequality is by \eqref{q_i_q_j_dy}, the last equality is by \(d_h = \widetilde{\Omega} \Big( \max \{\mathrm{SNR}^4, \mathrm{SNR}^{-4}\} N^2 \epsilon^{-2} \Big)\) and \(d = \widetilde{\Omega} \Big( \epsilon^{-2} N^2 d_h \Big) \).
\begin{align*}
    &\vert \langle \bm{q}_{n, i}^{(t+1)}, \bm{q}_{\overline{n}, j}^{(t+1)} \rangle \vert \le \vert \langle \bm{q}_{n, i}^{(T_2)}, \bm{q}_{\overline{n}, j}^{(T_2)} \rangle \vert + \sum\limits_{s = T_2}^{t} \Big\vert \langle \bm{q}_{n, i}^{(s+1)}, \bm{q}_{\overline{n}, j}^{(s+1)} \rangle - \langle \bm{q}_{n, i}^{(s)}, \bm{q}_{\overline{n}, j}^{(s)} \rangle \Big\vert \\
    &\le \vert \langle \bm{q}_{n, i}^{(T_2)}, \bm{q}_{\overline{n}, j}^{(T_2)} \rangle \vert \\
    &+ \sum\limits_{s = T_2}^{t} \Big\vert \alpha_{n, i, +}^{(s)} \langle \bm{q}_{\overline{n}, j}^{(s)}, \bm{k}_+^{(s)} \rangle + \alpha_{n, i, -}^{(s)} \langle \bm{q}_{\overline{n}, j}^{(s)}, \bm{k}_-^{(s)} \rangle + \sum\limits_{n^\prime = 1}^N \sum\limits_{l=2}^M \alpha_{n, i, n^\prime, l}^{(s)} \langle \bm{q}_{\overline{n}, j}^{(s)}, \bm{k}_{n^\prime, l}^{(s)} \rangle \\
    &+ \alpha_{\overline{n}, j, +}^{(s)} \langle \bm{q}_{n, i}^{(s)}, \bm{k}_+^{(s)} \rangle + \alpha_{\overline{n}, j, -}^{(s)} \langle \bm{q}_{n, i}^{(s)}, \bm{k}_-^{(s)} \rangle + \sum\limits_{n^\prime = 1}^N \sum\limits_{l=2}^M \alpha_{\overline{n}, j, n^\prime, l}^{(s)} \langle \bm{q}_{n, i}^{(s)}, \bm{k}_{n^\prime, l}^{(s)} \rangle \\
    &+ \Big( \alpha_{n, i, +}^{(s)} \bm{k}_+^{(s)} + \alpha_{n, i, -}^{(s)} \bm{k}_-^{(s)} + \sum\limits_{n^\prime = 1}^N \sum\limits_{l=2}^M \alpha_{n, i, n^\prime, l}^{(s)} \bm{k}_{n^\prime, l}^{(s)} \Big) \\
    &\cdot \Big( \alpha_{\overline{n}, j, +}^{(s)} \bm{k}_+^{(s)\top} + \alpha_{\overline{n}, j, -}^{(s)} \bm{k}_-^{(s)\top} + \sum\limits_{n^\prime = 1}^N \sum\limits_{l=2}^M \alpha_{\overline{n}, j, n^\prime, l}^{(s)} \bm{k}_{n^\prime, l}^{(s)\top} \Big) \Big\vert \\
    &\le \vert \langle \bm{q}_{n, i}^{(T_2)}, \bm{q}_{\overline{n}, j}^{(T_2)} \rangle \vert \\
    &+ \sum\limits_{s = T_2}^{t} \vert \alpha_{n, i, +}^{(s)} \vert \vert \langle \bm{q}_{\overline{n}, j}^{(s)}, \bm{k}_+^{(s)} \rangle \vert + \sum\limits_{s = T_2}^{t} \vert \alpha_{n, i, -}^{(s)} \vert \vert \langle \bm{q}_{\overline{n}, j}^{(s)}, \bm{k}_-^{(s)} \rangle \vert + \sum\limits_{l=2}^M \sum\limits_{s = T_2}^{t} \vert \alpha_{n, i, \overline{n}, l}^{(s)} \vert \vert \langle \bm{q}_{\overline{n}, j}^{(s)}, \bm{k}_{\overline{n}, l}^{(s)} \rangle \vert \\
    &+ \sum\limits_{l=2}^M \sum\limits_{s = T_2}^{t} \vert \alpha_{n, i, n, l}^{(s)} \vert \vert \langle \bm{q}_{\overline{n}, j}^{(s)}, \bm{k}_{n, l}^{(s)} \rangle \vert + \sum\limits_{n^\prime \ne n \wedge n^\prime \ne \overline{n}} \sum\limits_{l=2}^M \sum\limits_{s = T_2}^{t} \vert \alpha_{n, i, n^\prime, l}^{(s)} \vert \vert \langle \bm{q}_{\overline{n}, j}^{(s)}, \bm{k}_{n^\prime, l}^{(s)} \rangle \vert \\
    &+ \sum\limits_{s = T_2}^{t} \vert \alpha_{\overline{n}, j, +}^{(s)} \vert \vert \langle \bm{q}_{n, i}^{(s)}, \bm{k}_+^{(s)} \rangle \vert + \sum\limits_{s = T_2}^{t} \vert \alpha_{\overline{n}, j, -}^{(s)} \vert \vert \langle \bm{q}_{n, i}^{(s)}, \bm{k}_-^{(s)} \rangle \vert + \sum\limits_{l=2}^M \sum\limits_{s = T_2}^{t} \vert \alpha_{\overline{n}, j, n, l}^{(s)} \vert \vert \langle \bm{q}_{n, i}^{(s)}, \bm{k}_{n, l}^{(s)} \rangle \vert \\
    &+ \sum\limits_{l=2}^M \sum\limits_{s = T_2}^{t} \vert \alpha_{\overline{n}, j, \overline{n}, l}^{(s)} \vert \vert \langle \bm{q}_{n, i}^{(s)}, \bm{k}_{\overline{n}, l}^{(s)} \rangle \vert + \sum\limits_{n^\prime \ne n \wedge n^\prime \ne \overline{n}} \sum\limits_{l=2}^M \sum\limits_{s = T_2}^{t} \vert \alpha_{\overline{n}, j, n^\prime, l}^{(s)} \vert \vert \langle \bm{q}_{n, i}^{(s)}, \bm{k}_{n^\prime, l}^{(s)} \rangle \vert \\
    &+ \{lower \ order \ term\} \\
    &= \vert \langle \bm{q}_{n, i}^{(T_2)}, \bm{q}_{\overline{n}, j}^{(T_2)} \rangle \vert \\
    &+ O \Big( \frac{ \big( \log (6N^2M^2 / \delta) \big)^{3} \log \big( O(\frac{1}{\epsilon}) \big) }{\epsilon d_h^{\frac{1}{2}}} \Big) \cdot \log ( \epsilon^{-1} d_h^{\frac{1}{2}} ) \\
    &+ M \cdot O \Big( \frac{ \big( \log (6N^2M^2 / \delta) \big)^{4} \log \big( O(\frac{1}{\epsilon}) \big) }{\epsilon d^{\frac{1}{2}} d_h^{\frac{1}{2}}} \Big) \cdot \log ( \epsilon^{-1} d_h^{\frac{1}{2}} ) \\
    &+ M \cdot O \Big( \frac{ \big( \log (6N^2M^2 / \delta) \big)^{3} \log \big( O(\frac{1}{\epsilon}) \big) }{\epsilon d_h^{\frac{1}{2}}} \Big) \cdot o(1) + N \cdot M \cdot O \Big( \frac{ \big( \log (6N^2M^2 / \delta) \big)^{4} \log \big( O(\frac{1}{\epsilon}) \big) }{\epsilon d^{\frac{1}{2}} d_h^{\frac{1}{2}}} \Big) \cdot o(1) \\
    &= \vert \langle \bm{q}_{n, i}^{(T_2)}, \bm{q}_{\overline{n}, j}^{(T_2)} \rangle \vert + O \Big( \frac{ \big( \log (6N^2M^2 / \delta) \big)^{3} \log \big( O(\frac{1}{\epsilon}) \big) \log ( \epsilon^{-1} d_h^{\frac{1}{2}} ) }{\epsilon d_h^{\frac{1}{2}}} \Big) \\
    &+ o \Big( \frac{ N \big( \log (6N^2M^2 / \delta) \big)^{4} \log \big( O(\frac{1}{\epsilon}) \big) }{\epsilon d^{\frac{1}{2}} d_h^{\frac{1}{2}}} \Big) \\
    &= o(1)
\end{align*}
for \(i, j \in [M] \backslash \{1\}, n, \overline{n} \in [N], n \ne \overline{n}\). The first inequality is triangle inequality, the second inequality is by \eqref{q_i_q_j_n_dy}, the last equality is by \(d_h = \widetilde{\Omega} \Big( \max \{\mathrm{SNR}^4, \mathrm{SNR}^{-4}\} N^2 \epsilon^{-2} \Big)\) and \(d = \widetilde{\Omega} \Big( \epsilon^{-2} N^2 d_h \Big) \).
\begin{align*}
    &\vert \langle \bm{k}_+^{(t+1)}, \bm{k}_-^{(t+1)} \rangle \vert \le \vert \langle \bm{k}_+^{(T_2)}, \bm{k}_-^{(T_2)} \rangle \vert + \sum\limits_{s = T_2}^{t} \Big\vert \langle \bm{k}_+^{(s+1)}, \bm{k}_-^{(s+1)} \rangle - \langle \bm{k}_+^{(s)}, \bm{k}_-^{(s)} \rangle \Big\vert \\
    &\le \vert \langle \bm{k}_+^{(T_2)}, \bm{k}_-^{(T_2)} \rangle \vert \\
    &+ \sum\limits_{s = T_2}^{t} \Big\vert \beta_{+, +}^{(s)} \langle \bm{q}_+^{(s)}, \bm{k}_-^{(s)} \rangle + \sum\limits_{n \in S_+} \sum\limits_{i=2}^M \beta_{n, +, i}^{(s)} \langle \bm{q}_{n, i}^{(s)}, \bm{k}_-^{(s)} \rangle \\
    &+ \beta_{-, -}^{(s)} \langle \bm{q}_-^{(s)}, \bm{k}_+^{(s)} \rangle + \sum\limits_{n \in S_-} \sum\limits_{i=2}^M \beta_{n, -, i}^{(s)} \langle \bm{q}_{n, i}^{(s)}, \bm{k}_+^{(s)} \rangle \\
    &+ \Big( \beta_{+, +}^{(s)} \bm{q}_+^{(s)} + \sum\limits_{n \in S_+} \sum\limits_{i=2}^M \beta_{n, +, i}^{(s)} \bm{q}_{n, i}^{(s)} \Big) \\
    &\cdot \Big( \beta_{-, -}^{(s)} \bm{q}_-^{(s)\top} + \sum\limits_{n \in S_-} \sum\limits_{i=2}^M \beta_{n, -, i}^{(s)} \bm{q}_{n, i}^{(s)\top} \Big) \Big\vert \\
    &\le \vert \langle \bm{k}_+^{(T_2)}, \bm{k}_-^{(T_2)} \rangle \vert \\
    &+ \sum\limits_{s = T_2}^{t} \vert \beta_{+, +}^{(s)} \vert \vert \langle \bm{q}_+^{(s)}, \bm{k}_-^{(s)} \rangle \vert + \sum\limits_{n \in S_+} \sum\limits_{i=2}^M \sum\limits_{s = T_2}^{t} \vert \beta_{n, +, i}^{(s)} \vert \vert \langle \bm{q}_{n, i}^{(s)}, \bm{k}_-^{(s)} \rangle \vert \\
    &+ \sum\limits_{s = T_2}^{t} \vert \beta_{-, -}^{(s)} \vert \vert \langle \bm{q}_-^{(s)}, \bm{k}_+^{(s)} \rangle \vert + \sum\limits_{n \in S_-} \sum\limits_{i=2}^M \sum\limits_{s = T_2}^{t} \vert \beta_{n, -, i}^{(s)} \vert \vert \langle \bm{q}_{n, i}^{(s)}, \bm{k}_+^{(s)} \rangle \vert \\
    &+ \{lower \ order \ term\} \\
    &= \vert \langle \bm{k}_+^{(T_2)}, \bm{k}_-^{(T_2)} \rangle \vert + O \Big( \frac{ N \big( \log (6N^2M^2 / \delta) \big)^{3} \log \big( O(\frac{1}{\epsilon}) \big) }{\epsilon d_h^{\frac{1}{2}}} \Big) \cdot \log ( \epsilon^{-1} d_h^{\frac{1}{2}} ) \\
    &+ N \cdot M \cdot O \Big( \frac{\mathrm{SNR}^2 \big( \log (6N^2M^2 / \delta) \big)^{3} \log \big( O(\frac{1}{\epsilon}) \big) }{ \epsilon d_h^{\frac{1}{2}} } \Big) \cdot \log ( \epsilon^{-1} d_h^{\frac{1}{2}} ) \\
    &= \vert \langle \bm{k}_+^{(T_2)}, \bm{k}_-^{(T_2)} \rangle \vert + O \Big( \frac{ N \big( \log (6N^2M^2 / \delta) \big)^{3} \log \big( O(\frac{1}{\epsilon}) \big) \log ( \epsilon^{-1} d_h^{\frac{1}{2}} ) }{\epsilon d_h^{\frac{1}{2}}} \Big) \\
    &+ O \Big( \frac{N \cdot \mathrm{SNR}^2 \big( \log (6N^2M^2 / \delta) \big)^{3} \log \big( O(\frac{1}{\epsilon}) \big) \log ( \epsilon^{-1} d_h^{\frac{1}{2}} ) }{ \epsilon d_h^{\frac{1}{2}} } \Big) \\
    &= o(1),
\end{align*}
where the first inequality is triangle inequality, the second inequality is by \eqref{k_p_k_m_dy}, the last equality is by \(d_h = \widetilde{\Omega} \Big( \max \{\mathrm{SNR}^4, \mathrm{SNR}^{-4}\} N^2 \epsilon^{-2} \Big)\).
\begin{align*}
    &\vert \langle \bm{k}_+^{(t+1)}, \bm{k}_{n, i}^{(t+1)} \rangle \vert \le \vert \langle \bm{k}_+^{(T_2)}, \bm{k}_{n, i}^{(T_2)} \rangle \vert + \sum\limits_{s = T_2}^{t} \Big\vert \langle \bm{k}_+^{(s+1)}, \bm{k}_{n, i}^{(s+1)} \rangle \Big\vert \\
    &\le \vert \langle \bm{k}_+^{(T_2)}, \bm{k}_{n, i}^{(T_2)} \rangle \vert \\
    &+ \sum\limits_{s = T_2}^{t} \Big\vert \beta_{+, +}^{(s)} \langle \bm{q}_+^{(s)}, \bm{k}_{n, i}^{(s)} \rangle + \sum\limits_{n^\prime \in S_+} \sum\limits_{l=2}^M \beta_{n^\prime, +, l}^{(s)} \langle \bm{q}_{n^\prime, l}^{(s)}, \bm{k}_{n, i}^{(s)} \rangle \\
    &+ \beta_{n, i, +}^{(s)} \langle \bm{q}_+^{(s)}, \bm{k}_+^{(s)} \rangle + \beta_{n, i, -}^{(s)} \langle \bm{q}_-^{(s)}, \bm{k}_+^{(s)} \rangle + \sum\limits_{n^\prime = 1}^N \sum\limits_{l=2}^M \beta_{n, i, n^\prime, l}^{(s)} \langle \bm{q}_{n^\prime, l}^{(s)}, \bm{k}_+^{(s)} \rangle \\
    &+ \Big( \beta_{+, +}^{(s)} \bm{q}_+^{(s)} + \sum\limits_{n^\prime \in S_+} \sum\limits_{l=2}^M \beta_{n^\prime, +, l}^{(s)} \bm{q}_{n^\prime, l}^{(s)} \Big) \\
    &\cdot \Big( \beta_{n, i, +}^{(s)} \bm{q}_+^{(s)\top} + \beta_{n, i, -}^{(s)} \bm{q}_-^{(s)\top} + \sum\limits_{n^\prime = 1}^N \sum\limits_{l=2}^M \beta_{n, i, n^\prime, l}^{(s)} \bm{q}_{n^\prime, l}^{(s)\top} \Big) \Big\vert \\
    &\le \vert \langle \bm{k}_+^{(T_2)}, \bm{k}_{n, i}^{(T_2)} \rangle \vert \\
    &+ \sum\limits_{s = T_2}^{t} \vert \beta_{+, +}^{(s)} \vert \vert \langle \bm{q}_+^{(s)}, \bm{k}_{n, i}^{(s)} \rangle \vert + \sum\limits_{l=2}^M \sum\limits_{s = T_2}^{t} \vert \beta_{n, +, l}^{(s)} \vert \vert \langle \bm{q}_{n, l}^{(s)}, \bm{k}_{n, i}^{(s)} \rangle \vert \\
    &+ \sum\limits_{n^\prime \in S_+ \wedge n^\prime \ne n} \sum\limits_{l=2}^M \sum\limits_{s = T_2}^{t} \vert \beta_{n^\prime, +, l}^{(s)} \vert \vert \langle \bm{q}_{n^\prime, l}^{(s)}, \bm{k}_{n, i}^{(s)} \rangle \vert + \sum\limits_{s = T_2}^{t} \vert \beta_{n, i, +}^{(s)} \vert \vert \langle \bm{q}_+^{(s)}, \bm{k}_+^{(s)} \rangle \vert \\
    &+ \sum\limits_{s = T_2}^{t} \vert \beta_{n, i, -}^{(s)} \vert \vert \langle \bm{q}_-^{(s)}, \bm{k}_+^{(s)} \rangle \vert + \sum\limits_{l=2}^M \sum\limits_{s = T_2}^{t} \vert \beta_{n, i, n, l}^{(s)} \vert \vert \langle \bm{q}_{n, l}^{(s)}, \bm{k}_+^{(s)} \rangle \vert \\
    &+ \sum\limits_{n^\prime \ne n}^N \sum\limits_{l=2}^M \sum\limits_{s = T_2}^{t} \vert \beta_{n, i, n^\prime, l}^{(s)} \vert \vert \langle \bm{q}_{n^\prime, l}^{(s)}, \bm{k}_+^{(s)} \rangle \vert \\
    &+ \{lower \ order \ term\} \\
    &= \vert \langle \bm{k}_+^{(T_2)}, \bm{k}_{n, i}^{(T_2)} \rangle \vert \\
    &+ O \Big( \frac{ N \big( \log (6N^2M^2 / \delta) \big)^{3} \log \big( O(\frac{1}{\epsilon}) \big) }{\epsilon d_h^{\frac{1}{2}}} \Big) \cdot \log ( \epsilon^{-1} d_h^{\frac{1}{2}} ) \\
    &+ M \cdot O \Big( \frac{\mathrm{SNR}^2 \big( \log (6N^2M^2 / \delta) \big)^{3} \log \big( O(\frac{1}{\epsilon}) \big) }{ \epsilon d_h^{\frac{1}{2}} } \Big) \cdot \log ( \epsilon^{-1} d_h^{\frac{1}{2}} ) \\
    &+ N \cdot M \cdot O \Big( \frac{\mathrm{SNR}^2 \big( \log (6N^2M^2 / \delta) \big)^{3} \log \big( O(\frac{1}{\epsilon}) \big) }{ \epsilon d_h^{\frac{1}{2}} } \Big) \cdot o(1) \\
    &+ O \Big( \frac{ N \big( \log (6N^2M^2 / \delta) \big)^{3} \log \big( O(\frac{1}{\epsilon}) \big) }{\epsilon d_h^{\frac{1}{2}}} \Big) \cdot \log ( \epsilon^{-1} d_h^{\frac{1}{2}} ) \\
    &+ M \cdot O \Big( \frac{ \big( \log (6N^2M^2 / \delta) \big)^{3} \log \big( O(\frac{1}{\epsilon}) \big) }{\epsilon d_h^{\frac{1}{2}}} \Big) \cdot \log ( \epsilon^{-1} d_h^{\frac{1}{2}} ) \\
    &+ N \cdot M \cdot O \Big( \frac{ \big( \log (6N^2M^2 / \delta) \big)^{4} \log \big( O(\frac{1}{\epsilon}) \big) }{\epsilon d^{\frac{1}{2}} d_h^{\frac{1}{2}}} \Big) \cdot \log ( \epsilon^{-1} d_h^{\frac{1}{2}} ) \\
    &= \vert \langle \bm{k}_+^{(T_2)}, \bm{k}_{n, i}^{(T_2)} \rangle \vert + O \Big( \frac{ N \big( \log (6N^2M^2 / \delta) \big)^{3} \log \big( O(\frac{1}{\epsilon}) \big) \log ( \epsilon^{-1} d_h^{\frac{1}{2}} ) }{\epsilon d_h^{\frac{1}{2}}} \Big) \\
    &+ o \Big( \frac{N \cdot \mathrm{SNR}^2 \big( \log (6N^2M^2 / \delta) \big)^{3} \log \big( O(\frac{1}{\epsilon}) \big) }{ \epsilon d_h^{\frac{1}{2}} } \Big) \\
    &= o(1)
\end{align*}
where the first inequality is triangle inequality, the second inequality is by \eqref{k_p_k_i_dy}, the last equality is by \(d_h = \widetilde{\Omega} \Big( \max \{\mathrm{SNR}^4, \mathrm{SNR}^{-4}\} N^2 \epsilon^{-2} \Big)\). Similarly, we have \(\vert \langle \bm{k}_-^{(t+1)}, \bm{k}_{n, i}^{(t+1)} \rangle \vert = o(1)\).
\begin{align*}
    &\vert \langle \bm{k}_{n, i}^{(t+1)}, \bm{k}_{n, j}^{(t+1)} \rangle \vert \le \vert \langle \bm{k}_{n, i}^{(T_2)}, \bm{k}_{n, j}^{(T_2)} \rangle \vert + \sum\limits_{s = T_2}^{t} \Big\vert \langle \bm{k}_{n, i}^{(s+1)}, \bm{k}_{n, j}^{(s+1)} \rangle - \langle \bm{k}_{n, i}^{(s)}, \bm{k}_{n, j}^{(s)} \rangle \Big\vert \\
    &\le \vert \langle \bm{k}_{n, i}^{(T_2)}, \bm{k}_{n, j}^{(T_2)} \rangle \vert \\
    &+ \sum\limits_{s = T_2}^{t} \Big\vert \beta_{n, i, +}^{(s)} \langle \bm{q}_+^{(s)}, \bm{k}_{n, j}^{(s)} \rangle + \beta_{n, i, -}^{(s)} \langle \bm{q}_-^{(s)}, \bm{k}_{n, j}^{(s)} \rangle + \sum\limits_{n^\prime = 1}^N \sum\limits_{l=2}^M \beta_{n, i, n^\prime, l}^{(s)} \langle \bm{q}_{n^\prime, l}^{(s)}, \bm{k}_{n, j}^{(s)} \rangle \\
    &+ \beta_{n, j, +}^{(s)} \langle \bm{q}_+^{(s)}, \bm{k}_{n, i}^{(s)} \rangle + \beta_{n, j, -}^{(s)} \langle \bm{q}_-^{(s)}, \bm{k}_{n, i}^{(s)} \rangle + \sum\limits_{n^\prime = 1}^N \sum\limits_{l=2}^M \beta_{n, j, n^\prime, l}^{(s)} \langle \bm{q}_{n^\prime, l}^{(s)}, \bm{k}_{n, i}^{(s)} \rangle \\
    &+ \Big( \beta_{n, i, +}^{(s)} \bm{q}_+^{(s)} + \beta_{n, i, -}^{(s)} \bm{q}_-^{(s)} + \sum\limits_{n^\prime = 1}^N \sum\limits_{l=2}^M \beta_{n, i, n^\prime, l}^{(s)} \bm{q}_{n^\prime, l}^{(s)} \Big) \\
    &\cdot \Big( \beta_{n, j, +}^{(s)} \bm{q}_+^{(s)\top} + \beta_{n, j, -}^{(s)} \bm{q}_-^{(s)\top} + \sum\limits_{n^\prime = 1}^N \sum\limits_{l=2}^M \beta_{n, j, n^\prime, l}^{(s)} \bm{q}_{n^\prime, l}^{(s)\top} \Big) \Big\vert \\
    &\le \vert \langle \bm{k}_{n, i}^{(T_2)}, \bm{k}_{n, j}^{(T_2)} \rangle \vert \\
    &+ \sum\limits_{s = T_2}^{t} \vert \beta_{n, i, +}^{(s)} \vert \vert \langle \bm{q}_+^{(s)}, \bm{k}_{n, j}^{(s)} \rangle \vert + \sum\limits_{s = T_2}^{t} \vert \beta_{n, i, -}^{(s)} \vert \vert \langle \bm{q}_-^{(s)}, \bm{k}_{n, j}^{(s)} \rangle \vert + \sum\limits_{l=2}^M \sum\limits_{s = T_2}^{t} \vert \beta_{n, i, n, l}^{(s)} \vert \vert \langle \bm{q}_{n, l}^{(s)}, \bm{k}_{n, j}^{(s)} \rangle \vert \\
    &+ \sum\limits_{n^\prime \ne n}^N \sum\limits_{l=2}^M \sum\limits_{s = T_2}^{t} \vert \beta_{n, i, n^\prime, l}^{(s)} \vert \vert \langle \bm{q}_{n^\prime, l}^{(s)}, \bm{k}_{n, j}^{(s)} \rangle \vert + \sum\limits_{s = T_2}^{t} \vert \beta_{n, j, +}^{(s)} \vert \vert \langle \bm{q}_+^{(s)}, \bm{k}_{n, i}^{(s)} \rangle \vert + \sum\limits_{s = T_2}^{t} \vert \beta_{n, j, -}^{(s)} \vert \vert \langle \bm{q}_-^{(s)}, \bm{k}_{n, i}^{(s)} \rangle \vert \\
    &+ \sum\limits_{l=2}^M \sum\limits_{s = T_2}^{t} \vert \beta_{n, j, n, l}^{(s)} \vert \vert \langle \bm{q}_{n, l}^{(s)}, \bm{k}_{n, i}^{(s)} \rangle \vert + \sum\limits_{n^\prime \ne n}^N \sum\limits_{l=2}^M \sum\limits_{s = T_2}^{t} \vert \beta_{n, j, n^\prime, l}^{(s)} \vert \vert \langle \bm{q}_{n^\prime, l}^{(s)}, \bm{k}_{n, i}^{(s)} \rangle \vert \\
    &+ \{lower \ order \ term\} \\
    &= \vert \langle \bm{k}_{n, i}^{(T_2)}, \bm{k}_{n, j}^{(T_2)} \rangle \vert \\
    &+ O \Big( \frac{ N \big( \log (6N^2M^2 / \delta) \big)^{3} \log \big( O(\frac{1}{\epsilon}) \big) }{\epsilon d_h^{\frac{1}{2}}} \Big) \cdot \log ( \epsilon^{-1} d_h^{\frac{1}{2}} ) \\
    &+ M \cdot O \Big( \frac{ \big( \log (6N^2M^2 / \delta) \big)^{3} \log \big( O(\frac{1}{\epsilon}) \big) }{\epsilon d_h^{\frac{1}{2}}} \Big) \cdot \log ( \epsilon^{-1} d_h^{\frac{1}{2}} ) \\
    &+ N \cdot M \cdot O \Big( \frac{ \big( \log (6N^2M^2 / \delta) \big)^{4} \log \big( O(\frac{1}{\epsilon}) \big) }{\epsilon d^{\frac{1}{2}} d_h^{\frac{1}{2}}} \Big) \cdot o(1)\\
    &= \vert \langle \bm{k}_{n, i}^{(T_2)}, \bm{k}_{n, j}^{(T_2)} \rangle \vert + O \Big( \frac{ N \big( \log (6N^2M^2 / \delta) \big)^{3} \log \big( O(\frac{1}{\epsilon}) \big) \log ( \epsilon^{-1} d_h^{\frac{1}{2}} ) }{\epsilon d_h^{\frac{1}{2}}} \Big) \\
    &+ o \Big( \frac{ N \big( \log (6N^2M^2 / \delta) \big)^{4} \log \big( O(\frac{1}{\epsilon}) \big) }{\epsilon d^{\frac{1}{2}} d_h^{\frac{1}{2}}} \Big) \\
    &= o(1)
\end{align*}
for \(i, j \in [M] \backslash \{1\}, i \ne j, n \in [N]\). The first inequality is triangle inequality, the second inequality is by \eqref{k_i_k_j_dy}, the last equality is by \(d_h = \widetilde{\Omega} \Big( \max \{\mathrm{SNR}^4, \mathrm{SNR}^{-4}\} N^2 \epsilon^{-2} \Big)\) and \(d = \widetilde{\Omega} \Big( \epsilon^{-2} N^2 d_h \Big) \).
\begin{align*}
    &\vert \langle \bm{k}_{n, i}^{(t+1)}, \bm{k}_{\overline{n}, j}^{(t+1)} \rangle \vert \le \vert \langle \bm{k}_{n, i}^{(T_2)}, \bm{k}_{\overline{n}, j}^{(T_2)} \rangle \vert + \sum\limits_{s = T_2}^{t} \Big\vert \langle \bm{k}_{n, i}^{(s+1)}, \bm{k}_{\overline{n}, j}^{(s+1)} \rangle - \langle \bm{k}_{n, i}^{(s)}, \bm{k}_{\overline{n}, j}^{(s)} \rangle \Big\vert \\
    &\le \vert \langle \bm{k}_{n, i}^{(T_2)}, \bm{k}_{\overline{n}, j}^{(T_2)} \rangle \vert \\
    &+ \sum\limits_{s = T_2}^{t} \Big\vert \beta_{n, i, +}^{(s)} \langle \bm{q}_+^{(s)}, \bm{k}_{\overline{n}, j}^{(s)} \rangle + \beta_{n, i, -}^{(s)} \langle \bm{q}_-^{(s)}, \bm{k}_{\overline{n}, j}^{(s)} \rangle + \sum\limits_{n^\prime = 1}^N \sum\limits_{l=2}^M \beta_{n, i, n^\prime, l}^{(s)} \langle \bm{q}_{n^\prime, l}^{(s)}, \bm{k}_{\overline{n}, j}^{(s)} \rangle \\
    &+ \beta_{\overline{n}, j, +}^{(s)} \langle \bm{q}_+^{(s)}, \bm{k}_{n, i}^{(s)} \rangle + \beta_{\overline{n}, j, -}^{(s)} \langle \bm{q}_-^{(s)}, \bm{k}_{n, i}^{(s)} \rangle + \sum\limits_{n^\prime = 1}^N \sum\limits_{l=2}^M \beta_{\overline{n}, j, n^\prime, l}^{(s)} \langle \bm{q}_{n^\prime, l}^{(s)}, \bm{k}_{n, i}^{(s)} \rangle \\
    &+ \Big( \beta_{n, i, +}^{(s)} \bm{q}_+^{(s)} + \beta_{n, i, -}^{(s)} \bm{q}_-^{(s)} + \sum\limits_{n^\prime = 1}^N \sum\limits_{l=2}^M \beta_{n, i, n^\prime, l}^{(s)} \bm{q}_{n^\prime, l}^{(s)} \Big) \\
    &\cdot \Big( \beta_{\overline{n}, j, +}^{(s)} \bm{q}_+^{(s)\top} + \beta_{\overline{n}, j, -}^{(s)} \bm{q}_-^{(s)\top} + \sum\limits_{n^\prime = 1}^N \sum\limits_{l=2}^M \beta_{\overline{n}, j, n^\prime, l}^{(s)} \bm{q}_{n^\prime, l}^{(s)\top} \Big) \Big\vert \\
    &\le \vert \langle \bm{k}_{n, i}^{(T_2)}, \bm{k}_{\overline{n}, j}^{(T_2)} \rangle \vert \\
    &+ \sum\limits_{s = T_2}^{t} \vert \beta_{n, i, +}^{(s)} \vert \vert \langle \bm{q}_+^{(s)}, \bm{k}_{\overline{n}, j}^{(s)} \rangle \vert + \sum\limits_{s = T_2}^{t} \vert \beta_{n, i, -}^{(s)} \vert \vert \langle \bm{q}_-^{(s)}, \bm{k}_{\overline{n}, j}^{(s)} \rangle \vert + \sum\limits_{l=2}^M \sum\limits_{s = T_2}^{t} \vert \beta_{n, i, \overline{n}, l}^{(s)} \vert \vert \langle \bm{q}_{\overline{n}, l}^{(s)}, \bm{k}_{\overline{n}, j}^{(s)} \rangle \vert \\
    &+ \sum\limits_{l=2}^M \sum\limits_{s = T_2}^{t} \vert \beta_{n, i, n, l}^{(s)} \vert \vert \langle \bm{q}_{n, l}^{(s)}, \bm{k}_{\overline{n}, j}^{(s)} \rangle \vert + \sum\limits_{n^\prime \ne n \wedge n^\prime \overline{n}} \sum\limits_{l=2}^M \sum\limits_{s = T_2}^{t} \vert \beta_{n, i, n^\prime, l}^{(s)} \vert \vert \langle \bm{q}_{n^\prime, l}^{(s)}, \bm{k}_{\overline{n}, j}^{(s)} \rangle \vert \\
    &+ \sum\limits_{s = T_2}^{t} \vert \beta_{\overline{n}, j, +}^{(s)} \vert \vert \langle \bm{q}_+^{(s)}, \bm{k}_{n, i}^{(s)} \rangle \vert + \sum\limits_{s = T_2}^{t} \vert \beta_{\overline{n}, j, -}^{(s)} \vert \vert \langle \bm{q}_-^{(s)}, \bm{k}_{n, i}^{(s)} \rangle \vert + \sum\limits_{l=2}^M \sum\limits_{s = T_2}^{t} \vert \beta_{\overline{n}, j, n, l}^{(s)} \vert \vert \langle \bm{q}_{n, l}^{(s)}, \bm{k}_{n, i}^{(s)} \rangle \vert \\
    &+ \sum\limits_{l=2}^M \sum\limits_{s = T_2}^{t} \vert \beta_{\overline{n}, j, \overline{n}, l}^{(s)} \vert \vert \langle \bm{q}_{\overline{n}, l}^{(s)}, \bm{k}_{n, i}^{(s)} \rangle \vert + \sum\limits_{n^\prime \ne n \wedge n^\prime \overline{n}} \sum\limits_{l=2}^M \sum\limits_{s = T_2}^{t} \vert \beta_{\overline{n}, j, n^\prime, l}^{(s)} \vert \vert \langle \bm{q}_{n^\prime, l}^{(s)}, \bm{k}_{n, i}^{(s)} \rangle \vert \\
    &+ \{lower \ order \ term\} \\
    &= \vert \langle \bm{k}_{n, i}^{(T_2)}, \bm{k}_{\overline{n}, j}^{(T_2)} \rangle \vert \\
    &+ O \Big( \frac{ N \big( \log (6N^2M^2 / \delta) \big)^{3} \log \big( O(\frac{1}{\epsilon}) \big) }{\epsilon d_h^{\frac{1}{2}}} \Big) \cdot \log ( \epsilon^{-1} d_h^{\frac{1}{2}} ) + M \cdot O \Big( \frac{ \big( \log (6N^2M^2 / \delta) \big)^{4} \log \big( O(\frac{1}{\epsilon}) \big) }{\epsilon d^{\frac{1}{2}} d_h^{\frac{1}{2}}} \Big) \cdot \log ( \epsilon^{-1} d_h^{\frac{1}{2}} ) \\
    &+ M \cdot O \Big( \frac{ \big( \log (6N^2M^2 / \delta) \big)^{3} \log \big( O(\frac{1}{\epsilon}) \big) }{\epsilon d_h^{\frac{1}{2}}} \Big) \cdot o(1) + N \cdot M \cdot O \Big( \frac{ \big( \log (6N^2M^2 / \delta) \big)^{4} \log \big( O(\frac{1}{\epsilon}) \big) }{\epsilon d^{\frac{1}{2}} d_h^{\frac{1}{2}}} \Big) \cdot o(1) \\
    &= \vert \langle \bm{k}_{n, i}^{(T_2)}, \bm{k}_{\overline{n}, j}^{(T_2)} \rangle \vert \\
    &+ O \Big( \frac{ N \big( \log (6N^2M^2 / \delta) \big)^{3} \log \big( O(\frac{1}{\epsilon}) \big) \log ( \epsilon^{-1} d_h^{\frac{1}{2}} ) }{\epsilon d_h^{\frac{1}{2}}} \Big) + o \Big( \frac{ N \big( \log (6N^2M^2 / \delta) \big)^{4} \log \big( O(\frac{1}{\epsilon}) \big) }{\epsilon d^{\frac{1}{2}} d_h^{\frac{1}{2}}} \Big) \\
    &= o(1)
\end{align*}
for \(i, j \in [M] \backslash \{1\}, n, \overline{n} \in [N], n \ne \overline{n}\). The first inequality is triangle inequality, the second inequality is by \eqref{k_i_k_j_n_dy}, the last equality is by \(d_h = \widetilde{\Omega} \Big( \max \{\mathrm{SNR}^4, \mathrm{SNR}^{-4}\} N^2 \epsilon^{-2} \Big)\) and \(d = \widetilde{\Omega} \Big( \epsilon^{-2} N^2 d_h \Big) \).

\subsection{Explanations of Lower Order Terms}
In this section, we provide some explanations of lower order terms to demonstrate the rigor of our proof.

To bound the so-call \(\{lower \ order \ term\}\), we condition that dimensions \( d \), \(d_h\) are sufficiently large and learning rate \(\eta\) is sufficiently small. Next, we show how we utilize these three parameters.

\paragraph{Sufficiently mall learning rate \(\eta\) :} Recall the dynamics of QK 
\[
    \langle \bm{q}^{(t+1)}, \bm{k}^{(t+1)} \rangle - \langle \bm{q}^{(t)}, \bm{k}^{(t)} \rangle = \langle \Delta \bm{q}^{(t)}, \bm{k}^{(t)} \rangle + \langle \bm{q}^{(t)}, \Delta \bm{k}^{(t)} \rangle + \langle \Delta \bm{q}^{(t)}, \Delta \bm{k}^{(t)} \rangle
\]
Note that terms \(\langle \Delta \bm{q}^{(t)}, \bm{k}^{(t)} \rangle\), \(\langle \bm{q}^{(t)}, \Delta \bm{k}^{(t)} \rangle\) contain factor \(\eta\), and term \(\langle \Delta \bm{q}^{(t)}, \Delta \bm{k}^{(t)} \rangle\) contains factor \(\eta^2\). 
Therefore, as long as \(\eta\) is sufficiently small, \(\langle \Delta \bm{q}^{(t)}, \Delta \bm{k}^{(t)} \rangle\) is sufficiently small than \(\langle \Delta \bm{q}^{(t)}, \bm{k}^{(t)} \rangle\) and \(\langle \bm{q}^{(t)}, \Delta \bm{k}^{(t)} \rangle\). 
Now we take the dynamic of \(\langle \bm{q}_+^{(t)}, \bm{k}_+^{(t)} \rangle\) as an example.

\begin{equation*}
\begin{split}
    &\langle \bm{q}_+^{(t+1)}, \bm{k}_+^{(t+1)} \rangle - \langle \bm{q}_+^{(t)}, \bm{k}_+^{(t)} \rangle \\
    &= \alpha_{+, +}^{(t)} \Vert \bm{k}_+^{(t)} \Vert_2^2 + \sum\limits_{n \in S_+} \sum\limits_{i=2}^M \alpha_{n, +, i}^{(t)} \langle \bm{k}_+^{(t)}, \bm{k}_{n, i}^{(t)} \rangle \\
    &+ \beta_{+, +}^{(t)} \Vert \bm{q}_+^{(t)} \Vert_2^2 + \sum\limits_{n \in S_+} \sum\limits_{i=2}^M \beta_{n, +, i}^{(t)} \langle \bm{q}_+^{(t)}, \bm{q}_{n, i}^{(t)} \rangle \\
    &+ \Big( \alpha_{+, +}^{(t)} \bm{k}_+^{(t)} + \sum\limits_{n \in S_+} \sum\limits_{i=2}^M \alpha_{n, +, i}^{(t)} \bm{k}_{n, i}^{(t)} \Big) \\
    &\cdot \Big( \beta_{+, +}^{(t)} \bm{q}_+^{(t)\top} + \sum\limits_{n \in S_+} \sum\limits_{i=2}^M \beta_{n, +, i}^{(t)} \bm{q}_{n, i}^{(t)\top} \Big),
\end{split}
\end{equation*}
Under benign overfitting regime, we have \(\Vert \bm{k}_+^{(t)} \Vert_2^2 = \Theta (\Vert \bm{\mu} \Vert_2^2 \sigma_h^2 d_h)\), 
\(\langle \bm{q}_+^{(t)}, \bm{k}_+^{(t)} \rangle \le \log ( \epsilon^{-1} d_h^{\frac{1}{2}} ) \) 
and \(\beta_{+, +}^{(t)} = O(\eta \Vert \bm{\mu} \Vert_2^2)\).
Therefore, as long as \(\eta = o( \sigma_h^2 d_h (\log ( \epsilon^{-1} d_h^{\frac{1}{2}} ))^{-1} )\), we have 
\[
    \big( \alpha_{+, +}^{(t)} \bm{k}_+^{(t)} \big) \big( \beta_{+, +}^{(t)} \bm{q}_+^{(t)\top} \big) = o( \alpha_{+, +}^{(t)} \Vert \bm{k}_+^{(t)} \Vert_2^2 ).
\]
Similar method can be used for other items in \( \Big( \alpha_{+, +}^{(t)} \bm{k}_+^{(t)} + \sum\limits_{n \in S_+} \sum\limits_{i=2}^M \alpha_{n, +, i}^{(t)} \bm{k}_{n, i}^{(t)} \Big) \cdot \Big( \beta_{+, +}^{(t)} \bm{q}_+^{(t)\top} + \sum\limits_{n \in S_+} \sum\limits_{i=2}^M \beta_{n, +, i}^{(t)} \bm{q}_{n, i}^{(t)\top} \Big) \).
At last, we have
\begin{equation}
\begin{split}
    \label{example_dh}
    &\langle \bm{q}_+^{(t+1)}, \bm{k}_+^{(t+1)} \rangle - \langle \bm{q}_+^{(t)}, \bm{k}_+^{(t)} \rangle \\
    &= \alpha_{+, +}^{(t)} \Vert \bm{k}_+^{(t)} \Vert_2^2 + \sum\limits_{n \in S_+} \sum\limits_{i=2}^M \alpha_{n, +, i}^{(t)} \langle \bm{k}_+^{(t)}, \bm{k}_{n, i}^{(t)} \rangle \\
    &+ \beta_{+, +}^{(t)} \Vert \bm{q}_+^{(t)} \Vert_2^2 + \sum\limits_{n \in S_+} \sum\limits_{i=2}^M \beta_{n, +, i}^{(t)} \langle \bm{q}_+^{(t)}, \bm{q}_{n, i}^{(t)} \rangle \\
    &+ \{lower \ order \ term\}.
\end{split}
\end{equation}

\paragraph{Sufficiently large dimension \(d_h\) :} Take \eqref{example_dh} as an example, Noting that \(\Vert \bm{k}_+^{(t)} \Vert_2^2 = O(\eta \Vert \bm{\mu} \Vert_2^2)\) and \(\langle \bm{k}_+^{(t)}, \bm{k}_{n, i}^{(t)} \rangle = o(1)\).
Therefore, as long as \(d_h\) is sufficiently large, \(\Vert \bm{k}_+^{(t)} \Vert_2^2\) is much larger than \(\langle \bm{k}_+^{(t)}, \bm{k}_{n, i}^{(t)} \rangle\).
Besides, by the property that the sum of each row and column of matrix \((diag(\bm{\varphi}_{n, 1}^{(t)}) - \bm{\varphi}_{n, 1}^{(t)\top} \bm{\varphi}_{n, 1}^{(t)})\) is 0,
we have \(\alpha_{+, +}^{(t)} + \sum\limits_{n \in S_+} \sum\limits_{i=2}^M \alpha_{n, +, i}^{(t)} = 0\).
We also prove \(\alpha_{+, +}^{(t)} \ge 0\) and \(\alpha_{n, +, i}^{(t)} \le 0\) under benign overfitting regime, thus the magnitude of \(\alpha_{n, +, i}^{(t)}\) is smaller than \(\alpha_{+, +}^{(t)}\).
All in all, it can be proved that \(\sum\limits_{n \in S_+} \sum\limits_{i=2}^M \alpha_{n, +, i}^{(t)} \langle \bm{k}_+^{(t)}, \bm{k}_{n, i}^{(t)} \rangle = o( \alpha_{+, +}^{(t)} \Vert \bm{k}_+^{(t)} \Vert_2^2 )\).
This method can be further applied to \(\beta_{+, +}^{(t)} \Vert \bm{q}_+^{(t)} \Vert_2^2 + \sum\limits_{n \in S_+} \sum\limits_{i=2}^M \beta_{n, +, i}^{(t)} \langle \bm{q}_+^{(t)}, \bm{q}_{n, i}^{(t)} \rangle\).
At last, we can simplify \eqref{example_dh} as follows:
\begin{equation}
\begin{split}
    &\langle \bm{q}_+^{(t+1)}, \bm{k}_+^{(t+1)} \rangle - \langle \bm{q}_+^{(t)}, \bm{k}_+^{(t)} \rangle \\
    &= \alpha_{+, +}^{(t)} \Vert \bm{k}_+^{(t)} \Vert_2^2 + \beta_{+, +}^{(t)} \Vert \bm{q}_+^{(t)} \Vert_2^2 + \{lower \ order \ term\}.
\end{split}
\end{equation}

\paragraph{Sufficiently large dimension \(d\) :} Take \(\alpha_{n^\prime, i^\prime, +}^{(t)}\) as an example:
\begin{equation}
\begin{split}
    \label{example_d}
    &\alpha_{n^\prime, i^\prime, +}^{(t)} = \frac{\eta}{NM} \sum\limits_{n \in S_+} - \ell_n^{\prime(t)} \sum\limits_{i=2}^M \langle \bm{\xi}_{n^\prime, i^\prime}, \bm{\xi}_{n, i} \rangle \\
    &\cdot \Big( V_+^{(t)} \big( \frac{ \exp (\langle \bm{q}_{n, i}^{(t)}, \bm{k}_+^{(t)} \rangle)}{ \exp  (\langle \bm{q}_{n, i}^{(t)}, \bm{k}_+^{(t)} \rangle) + \sum\limits_{j=2}^M  \exp  ( \langle \bm{q}_{n, i}^{(t)}, \bm{k}_{n, j}^{(t)} \rangle ) } \\
    & - (\frac{ \exp (\langle \bm{q}_{n, i}^{(t)}, \bm{k}_+^{(t)} \rangle)}{ \exp  (\langle \bm{q}_{n, i}^{(t)}, \bm{k}_+^{(t)} \rangle) + \sum\limits_{j=2}^M  \exp  ( \langle \bm{q}_{n, i}^{(t)}, \bm{k}_{n, j}^{(t)} \rangle ) })^2 \big) \\
    & -\sum\limits_{k=2}^M \big( V_{n, i}^{(t)} \cdot \frac{ \exp (\langle \bm{q}_{n, i}^{(t)}, \bm{k}_+^{(t)} \rangle)}{ \exp  (\langle \bm{q}_{n, i}^{(t)}, \bm{k}_+^{(t)} \rangle) + \sum\limits_{j=2}^M  \exp  ( \langle \bm{q}_{n, i}^{(t)}, \bm{k}_{n, j}^{(t)} \rangle ) } \\
    & \cdot \frac{ \exp (\langle \bm{q}_{n, i}^{(t)}, \bm{k}_{n, k}^{(t)} \rangle)}{ \exp  (\langle \bm{q}_{n, i}^{(t)}, \bm{k}_+^{(t)} \rangle) + \sum\limits_{j=2}^M  \exp  ( \langle \bm{q}_{n, i}^{(t)}, \bm{k}_{n, j}^{(t)} \rangle ) } \big) \Big),
\end{split}
\end{equation}
Note that \(\langle \bm{\xi}_{n^\prime, i^\prime}, \bm{\xi}_{n, i} \rangle\) can be divided into two types: \(\Vert \bm{\xi}_{n^\prime, i^\prime} \Vert_2^2\) and \(\langle \bm{\xi}_{n^\prime, i^\prime}, \bm{\xi}_{n, i} \rangle\) for \(i \ne i^\prime \ or \ n \ne n^\prime\).

By Lemma \ref{caoyuan}, we have
\[
    \tilde{\sigma}_p^2d / 2 \le \Vert \bm{\xi}_{n, 2} \Vert_2^2 \le 3\tilde{\sigma}_p^2d / 2,
\]
\[
    \sigma_p^2d / 2 \le \Vert \bm{\xi}_{n, i} \Vert_2^2 \le 3\sigma_p^2d / 2,
\]
\[
    \vert \langle \bm{\xi}_{n, i} , \bm{\xi}_{n^\prime, i^\prime} \rangle \vert \le 2 \tilde{\sigma}_p^2 \cdot \sqrt{d \log (4N^2M^2/\delta)}
\]
for \(i, i^\prime \in [M] \backslash \{1\}, n, n^\prime \in [N], i \ne i^\prime \ or \ n \ne n^\prime\).

As long as \(d\) is sufficiently large, \(\Vert \bm{\xi}_{n^\prime, i^\prime} \Vert_2^2\) is much larger than \(\langle \bm{\xi}_{n^\prime, i^\prime}, \bm{\xi}_{n, i} \rangle\) for \(i \ne i^\prime \ or \ n \ne n^\prime\).
Therefore, \eqref{example_d} can be further simplified as follows:
\begin{equation}
\begin{split}
    &\alpha_{n^\prime, i^\prime, +}^{(t)} = - \frac{\eta}{NM} \ell_{n^\prime}^{\prime(t)} \Vert \bm{\xi}_{n^\prime, i^\prime} \Vert_2^2 \\
    &\cdot \Big( V_+^{(t)} \big( \frac{ \exp (\langle \bm{q}_{n^\prime, i^\prime}^{(t)}, \bm{k}_+^{(t)} \rangle)}{ \exp  (\langle \bm{q}_{n^\prime, i^\prime}^{(t)}, \bm{k}_+^{(t)} \rangle) + \sum\limits_{j=2}^M  \exp  ( \langle \bm{q}_{n^\prime, i^\prime}^{(t)}, \bm{k}_{n^\prime, j}^{(t)} \rangle ) } \\
    & - (\frac{ \exp (\langle \bm{q}_{n^\prime, i^\prime}^{(t)}, \bm{k}_+^{(t)} \rangle)}{ \exp  (\langle \bm{q}_{n^\prime, i^\prime}^{(t)}, \bm{k}_+^{(t)} \rangle) + \sum\limits_{j=2}^M  \exp  ( \langle \bm{q}_{n^\prime, i^\prime}^{(t)}, \bm{k}_{n^\prime, j}^{(t)} \rangle ) })^2 \big) \\
    & -\sum\limits_{k=2}^M \big( V_{n^\prime, i^\prime}^{(t)} \cdot \frac{ \exp (\langle \bm{q}_{n^\prime, i^\prime}^{(t)}, \bm{k}_+^{(t)} \rangle)}{ \exp  (\langle \bm{q}_{n^\prime, i^\prime}^{(t)}, \bm{k}_+^{(t)} \rangle) + \sum\limits_{j=2}^M  \exp  ( \langle \bm{q}_{n^\prime, i^\prime}^{(t)}, \bm{k}_{n^\prime, j}^{(t)} \rangle ) } \\
    & \cdot \frac{ \exp (\langle \bm{q}_{n^\prime, i^\prime}^{(t)}, \bm{k}_{n^\prime, k}^{(t)} \rangle)}{ \exp  (\langle \bm{q}_{n^\prime, i^\prime}^{(t)}, \bm{k}_+^{(t)} \rangle) + \sum\limits_{j=2}^M  \exp  ( \langle \bm{q}_{n^\prime, i^\prime}^{(t)}, \bm{k}_{n^\prime, j}^{(t)} \rangle ) } \big) \Big) \\
    &+ \{lower \ order \ term\}.
\end{split}
\end{equation}

\section{Takeaways for Practitioners}
Our theoretical results mainly focuse on the impact of different $N$ and SNR on the generalization performance. So we can provide guidance from the perspective of increasing $N$, SNR and $N \cdot \mathrm{SNR}^2$. The following are some practical scenarios

\textbf{Data Augmentation:} Researchers sometimes employ the technique of data augmentation by introducing controlled noise into their datasets. From the perspective of our paper's results, this method reduce SNR but improve N because we generate "new" data point by adding noises. As reducing SNR may be harmful to generalization performance, we must make sure that we use enough data points to train the model( enough sample size $N$ ).

\textbf{Semi-Supervised Learning:} Semi-supervised learning is useful when you have a small amount of labeled data and a large amount of unlabeled data. Labeled data can be seen as data with high SNR, while unlabeled data with low SNR because for some unlabeled samples, we may mistake their labels, making them equivalent to noises. In this scenario, we need to ensure that we have sufficient unlabeled data (enough sample size $N$) and make full use of labeled data (high SNR data points).

Overall, we need to consider both the sample size $N$ and the signal-to-noise ratio SNR to train the model.

\section{Broader Impacts}
\label{broader_impacts}
This work focus on theoretically studying the training dynamics and generalization of Transformer in Vision. 
The techniques used in this paper may be generalized to study other abilities of Transformer or other network models. 
Besides, the theoretical results in this paper may inspire more attempts at training large foundational models with high-quality data.
We do not foresee any form of negative social impact induced by our work.

\end{document}